\documentclass[10pt,twocolumn,letterpaper]{article}

\usepackage[pagenumbers]{cvpr}
\makeatletter
\@namedef{ver@everyshi.sty}{}
\makeatother

\usepackage{amsmath,amsfonts,bm}

\def\eqref#1{equation~\ref{#1}}

\def\1{\bm{1}}

\DeclareMathAlphabet{\mathsfit}{\encodingdefault}{\sfdefault}{m}{sl}
\SetMathAlphabet{\mathsfit}{bold}{\encodingdefault}{\sfdefault}{bx}{n}

\usepackage{url}
\usepackage{todonotes}
\usepackage{amssymb}
\usepackage{booktabs}
\usepackage{multirow}
\usepackage{minitoc}

\usepackage[pagebackref,breaklinks,colorlinks]{hyperref}

\renewcommand{\paragraph}[1]{\vspace{-0.1em} \smallskip\noindent {\bfseries\boldmath\ignorespaces #1}\hskip 0.9em plus 0.3em minus 0.3em}

\title{Bias in Pruned Vision Models: In-Depth Analysis and Countermeasures} 

\author{
Eugenia Iofinova  \\{\small IST Austria} \\ 
\and Alexandra Peste \\ {\small IST Austria} \\ 
\and Dan Alistarh\\
{\small IST Austria \&  Neural Magic} \\
}

\begin{document}
\doparttoc %
\faketableofcontents %

\maketitle

\begin{abstract}

Pruning---that is, setting a significant subset of the parameters of a neural network to zero---is one of the most popular methods of model compression. 
Yet, several recent works have raised the issue that pruning may induce or exacerbate \emph{bias} in the output of the compressed model. 
 Despite existing evidence for this phenomenon, the relationship between neural network pruning and induced bias is not well-understood. 
 In this work, we systematically investigate and characterize this phenomenon in Convolutional Neural Networks for computer vision.
First, we show that it is in fact possible to obtain highly-sparse models, e.g. with less than $10
\%$ remaining weights, which do not decrease in accuracy nor substantially increase in bias when compared to dense models. 
 At the same time, we also find that, at higher sparsities, pruned models exhibit higher uncertainty in their outputs, as well as increased correlations, which we directly link to increased bias. We propose easy-to-use criteria which, based only on the \emph{uncompressed model},  establish whether bias  will increase with pruning, and identify the samples most susceptible to biased predictions post-compression.

\end{abstract}

\vspace{-1em}
\section{Introduction}

The concept of ``bias'' in machine learning models spans a range of considerations in terms of statistical, performance, and social metrics. 
Different definitions can lead to different relationships between bias and accuracy. 
For instance, if bias is defined in terms of accuracy disparity between identity groups, then accuracy in the ``stronger'' group may have to be reduced in order to reduce model bias. 
Several sources of bias have been identified in this context. For example, bias in datasets commonly used to train machine learning models~\cite{yang_towards_2020, birhane_large_2021, bender_dangers_2021} can severely impact outputs, and may be  difficult or even impossible to correct during training. The choice of model architecture, training methods, evaluation, and deployment can create or exacerbate bias~\cite{mehrabi_exacerbating_2021, barocas-hardt-narayanan, MehrabiBiasSurvey}.

One potential source of bias which is relatively less investigated is the fact that machine learning models, and in particular deep neural networks, are often \emph{compressed} for efficiency before being deployed. 
Seminal work by Hooker et al.~\cite{hooker_characterising_2020} and its follow-ups, e.g.~\cite{liebenwein_lost_2021, hooker2019compressed}  provided examples where model compression, and in particular pruning, can exacerbate bias by leading models to perform poorly on ``unusual'' data, which can frequently coincide with marginalized groups. 
Given the recent popularity of compression methods in deployment settings~\cite{gale2019state, elsen2020fast, hoefler2021sparsity, gholami2021survey} and the fact that, for massive models, compression is often necessary to enable model deployment, these findings raise the question of whether the bias due to compression can be exactly characterized, and in particular whether bias is an inherent side-effect of the model compression process. 

In this paper, we perform an in-depth analysis of bias in compressed vision models, providing new insights on this phenomenon, as well as a set of practical, effective criteria for identifying samples susceptible to biased predictions, which can be used to significantly attenuate bias. 

Our work starts from a common setting to study bias and bias mitigation~\cite{hooker2019compressed, hooker_characterising_2020, Wang2020TowardsFI, Lin2022FairGRAPE}: we study properties of sparse residual convolutional neural networks~\cite{he2016deep}, in particular ResNet18, applied for classification on the CelebA dataset\cite{Liu2015DeepLF}. 
Then, we validate our findings across other CNN architectures and other datasets. 
To study the impact of sparsity, we train highly accurate models with sparsity ranging from 80\% to 99.5\%, using the standard gradual magnitude pruning (GMP) approach~\cite{hagiwara1994, han2015learning, zhu2017prune, gale2019state}.  
We consider bias in dense and sparse models from two perspectives: 
\emph{systematic bias}, which refers to consistent errors in the model output, 
and \emph{category bias}, which refers to violations of fairness metrics associated with protected groups.%

On the positive side, our analysis shows that the GMP approach can produce models that are highly sparse, i.e. 90-95\% of pruned weights, without significant increase in any bias-related metrics. Yet, this requires care: we show that \emph{shared, jointly-trained} representations are significantly less susceptible to bias, and so careful choices of training procedure are needed for good results. 
On the other hand, at very high sparsities (95\%-99.5\%) we do observe non-trivial increase in category bias for the sparse models, for specific protected attributes. 
We perform an in-depth study of this phenomenon, correlating increase in bias with increased uncertainty in the model outputs, induced by sparsity. 
Leveraging insights from our analysis, we provide a simple set of criteria and techniques based on threshold calibration and overriding decisions for sensitive samples, which we show to have a significant effect on bias reduction. The latter only use information found in the original dense model.%

\section{Methodology}

\subsection{Notions of Bias}
\label{sec:bias-describe}

We now define the notions of bias we will use in the rest of the paper. 
We emphasize these categories should not be seen as exclusive: instead, they allow us to study different aspects of the given phenomena. 

\paragraph{Systematic Bias.} A standard, broad meaning of bias is \emph{systematic error}~\cite{dietterich1995machine}: for example, we can measure whether models are biased toward overconfidence in their predictions, or if they tend to generalize poorly to data from a shifted distribution. We call this \emph{Systematic Bias}; a full list of the metrics we use is given in section~\ref{sec:systematic_metrics}.

\paragraph{Category Bias.} 
A complementary approach to defining bias centers around the notion of \emph{subgroup/category} of samples in the dataset. 
Here, bias refers to violations of group fairness metrics with respect to given categories~\cite{barocas-hardt-narayanan} for instance by measuring differences in false positive, false negative, or error rates across subgroups. 
Other related metrics are worst subgroup performance \cite{Sagawa2020Distributionally}, or the standard deviation of accuracy across identity categories \cite{Lin2022FairGRAPE}.  

Inherent to these definitions is that the choice of attributes that define the subgroups must be \emph{meaningful} in a sociological context and \emph{relevant} to the model's application. For example, it is appropriate to measure the accuracy difference with respect to race and gender in facial identification software, since even a moderate difference in accuracy can lead to discrimination in real-world settings. 
Models that are highly-accurate on standard metrics, e.g. top-1 accuracy, may still be considered biased, for instance with respect to demographic parity. 
In order to distinguish the concept of \emph{bias} from that of \emph{fairness}, here we focus on \emph{algorithmic bias}, which we define as cases in which a model amplifies bias found in the training data. 
A classic example is when a model tends to have worse accuracy on samples from poorly-represented subgroups of the dataset. 
We call this type of bias \emph{Category Bias}.

These  notions are complementary: category biases are likely associated with systematic biases, and therefore, studying systematic bias can help us understand cases where models show socially-relevant category bias. 
This is a common assumption that is frequently used to study bias, for instance in the work on compression-identified exemplars of \cite{hooker2019compressed, hooker_characterising_2020}, which first identifies a consistent set of examples on which compressed models frequently struggle, and then demonstrates that these are enriched for certain identity groups. 
Generally, we are also interested in understanding the relationship between statistical notions of bias, examined via specific metrics, and potential systematic bias across protected categories.

\subsection{Category Bias Metric: Bias Amplification}

Following prior work~\cite{zhao-2017-men-shopping, hooker_characterising_2020}, we consider datasets where samples are classified according to binary attributes, and use a subset of these as ``identity'' attributes. 
For this, we introduce as our main metric a variant of \emph{Bias Amplification (BA)}~\cite{zhao-2017-men-shopping}. 
Intuitively, bias amplification will measure the extent to which correlations between identity categories and predicted attributes in the training data are exaggerated by the model. 
While positive correlation between an identity category and a predicted attribute can be reasonable (a model can predict that women wear earrings more frequently than men), models that \emph{amplify} such input relationships in their output may be stereotyping, by relying on identity markers as a proxy for other attributes. 
   
   To encode this formally, we compute bias amplification. We define the function $N(\cdot)$ to provide the \emph{count} of the number of samples with a specific binary attribute value, e.g. $\text{Young} = 1$, over a given sample set. We then define the bias $b$ of a binary attribute $A \in \{0,1\}$ with respect to a binary identity category $I \in \{0,1\}$ as 
   \vspace{-0.6em}
   $$b = \frac{N(A=1,I=1)}{N(A=1)},$$ 
   if the attribute and identity category are positively correlated in the training data, and
   \vspace{-0.6em}$$b = \frac{N(A=1,I=0)}{N(A=1)}, \textnormal{ otherwise.}$$ 
   
   The bias \emph{amplification} is then the difference between the bias computed on the predicted attribute $\Tilde{A}$ and the true value of the attribute $A$, computed on the test set: 
   \vspace{-0.8em}
   \begin{equation*} 
   BA = \frac{N(\tilde{A}=1,I=1)}{N(\tilde{A}=1)} - \frac{N(A=1,I=1)}{N(A=1)},
   \end{equation*} if the predicted attribute is positively correlated with the identity category, and 
   \vspace{-0.8em}
   \begin{equation*} 
    BA = \frac{N(\tilde{A}=1,I=0)}{N(\tilde{A}=1)} - \frac{N(A=1,I=0)}{N(A=1)}, 
   \end{equation*} if the predicted attribute is negatively correlated with the identity category.  
   We do not compute the Bias Amplification on any attribute that is not significantly biased toward either value of the identity category, 
   or if some combination of the predicted and protected attribute is very infrequent (e.g., occurring less than 10 times in the test data). 
   
   \paragraph{Discussion.} This metric has several advantages. Firstly, it is clear that high BA values signal stereotyping by the model. Unlike the original BA metric of \cite{zhao-2017-men-shopping}, our definition uses the label distribution in the \emph{test data} as the true baseline for the predicted label distribution of the model, allowing us to separate the effect of the model itself from the effect of the underlying data, and also allowing us to test the model for bias in settings where the test distribution does not closely resemble the training distribution. 
   
   Additionally, BA is not directly affected by other possible biases in the model, such as a tendency to underpredict rare attributes. 
   Moreover, unlike direct false-positive/negative analysis, BA directly takes into account predictions over both values of the protected attribute, and can be meaningfully aggregated across attributes.

\subsection{Systematic Bias Metrics}
\label{sec:systematic_metrics}
We use several other fine-grained metrics to measure the systematic bias of dense and sparse models.

\paragraph{Threshold Calibration Bias (TCB).} 
On many datasets, the majority of attributes are not evenly split across samples: e.g., for CelebA, the average imbalance is  80\%/20\%. 
We measure the change (typically, decline) of the proportion of predictions into the less common value of the attribute using the default threshold. Note that values near 1 show minimal TCB, while values away from 1 in either direction show higher TCB. 
\vspace{-1em}
\begin{equation*}
    TCB = 
    \begin{cases}
        \frac{N(\tilde{A}=1)}{N(A=1)},& \text{if} \text{Mean}(A) < 0.5\\
        \frac{N(\tilde{A}=0)}{N(A=0)}, & \text{otherwise.}
    \end{cases}
\end{equation*}

\paragraph{Uncertainty and Calibration.} Attribute predictions after applying the sigmoid function range between 0 and 1. For a converged model, they tend to cluster around the extremes, with some smaller number of predictions falling nearer the center of the interval. 
We consider prediction values between 0.1 and 0.9 to be \emph{uncertain}. These uncertainty metrics simply compute the proportion of predictions that fall into the uncertain interval.
We then check if the uncertainty correctly estimates the proportion correct by bucketing~\cite{chen2022wineverythinglottery, Naeini2015ECEObtainingWC}. The prediction range is split into ten equal-width buckets, and average per-bucket difference of the confidence and the proportion correct. These are then weighted by the bucket size and aggregated. The weighted average difference of the accuracy and confidence of the buckets is presented as the Expected Calibration Error (ECE). %
\vspace{-1em}
\begin{equation*}
    ECE=\sum_{m=1}^{10} \frac{|B_m|}{\sum_{n=1}^{10}|B_n|} \left| \text{acc}(B_m) - \text{conf}(B_m)\right|.
\end{equation*}

\paragraph{Label Interrelation.} Finally, we look at the strength of relationship between predicted labels on the various attribute. Specifically, for each attribute $A$, we train a linear regression using all other attributes as the features and $A$ as the variable to be predicted; the coefficient of determination ($R^2$) of this model tells us the extent to which the model output for A can be predicted from the model outputs of the other attributes in a co-trained model. Note that this does not imply a causal relationship - we cannot say that the model is using some of the attributes to predict others. Rather, a high interrelation suggests that the hidden feature layer is less expressive, forcing a closer relationship between linear classifiers using it as the features. 

\subsection{Evaluation Setup}
\label{sec:methods}

\paragraph{CelebA Setup.} In our primary study, we focus on ResNet18~\cite{he2016deep} models that predict human-annotated binary attributes from cropped-and-centered photos of celebrities in the CelebA dataset \cite{Liu2015DeepLF}. %

CelebA attribute prediction is frequently used for bias measurement  \cite{hooker2019compressed, hooker_characterising_2020, Wang2020TowardsFI, Lin2022FairGRAPE}. 
This is in part due to its size and widespread availability. 
Yet, CelebA is an imperfect proxy for real-world human photographs, as it skews substantially in both age and skin color, as well as make-up, hairstyles, and overall presentation of the human subjects.
As previous works have looked at both models that jointly co-train all or most CelebA attributes \cite{Wang2020TowardsFI, Lin2022FairGRAPE} and models that train only a single attribute \cite{hooker_characterising_2020}, we conduct both types of experiments. For the all-in-one/joint training, we train a ResNet18 model with 40 logistic classifiers after the fully-connected layer. Additionally, we train models with a single head for 7 CelebA attributes: Blond, Smiling, Oval Face, Big Nose, Mustache, Receding Hairline, Bags Under Eyes.

We validate our results by repeating our experiments on the ResNet50 and MobileNetV1\cite{howard2017mobilenets} architectures, as well as on structured sparsity (2:4, 1:4 and 1:8) sparsity patterns, which are better supported by current NVIDIA hardware \cite{mishra2021accelerating}. We also validate some of our findings on the uncropped CelebA dataset, as well as on the iWildcam \cite{beery2020iwildcam} and Animals with Attributes2~\cite{Xian2019AwAZeroShotLC} datasets.

 For CelebA, we use four attributes for computing Category Bias: ``Male'', ``Young'', ``Chubby'', and ``Pale Skin''\footnote{The choices to present gender as a binary attribute, and the specific words to describe the attributes were chosen by the creators of the CelebA dataset. We continue their use here to avoid confusion and enable comparisons with other works.}. These attributes were chosen because they loosely correspond to categories traditionally used to measure bias and discrimination. Examples of these categories can be found in Appendix~\ref{appendix:ui_tool}. \textbf{In the rest of the paper, we use ``categories'' to refer to these four attributes when they are used as the group identifier to compute BA, and ``attribute'' to refer to any CelebA attribute that is used as a prediction target.}

\paragraph{Model Architectures.} For both ResNet and MobileNet models, we use the standard model architecture, with only one fully-connected layer and a logit transformation following the convolutional blocks, and Binary Cross-Entropy loss. Unlike other studies using CelebA~\cite{Wang2020TowardsFI}, we found that including an additional fully-connected layer did not improve accuracy. Nor did it increase accuracy to initialize with ImageNet weights as in \cite{Wang2020TowardsFI, Lin2022FairGRAPE}, and therefore all models were randomly initialized following~\cite{he2015delving}. Consistent with other work, we use the cropped-and-centered version of the dataset described in \cite{Liu2015DeepLF}, and perform training data augmentations consistent with \cite{Wang2020TowardsFI}. We also validate on the uncropped version. We report results after running each experiment from 5 random seeds. %

\paragraph{Model Compression.}  We perform unstructured pruning, by gradually removing the lowest magnitude weights during training, known as Global Magnitude Pruning (GMP) \cite{hagiwara1994, zhu2017prune, han2015learning, gale2019state}. GMP is a standard baseline, which, despite its simplicity, is competitive with more complex approaches \cite{gale2019state, singh2020woodfisher, frantar2021efficient, kurtic2022optimal, kurtic2022gmp}. We prune all ResNet18 models to 80\%, 90\%, 95\%, 98\%, 99\%, and 99.5\% sparsity. Following earlier work~\cite{Iofinova2021HowWD}, we considered two variants of GMP. The main variant starts from a random initialization (RI), and gradually removes parameters after the tenth training epoch, while simultaneously training the model~\cite{zhu2017prune}; we refer to this setup as GMP-RI. The second variant starts from a pre-trained dense model, then gradually removes parameters with the lowest global magnitude while continuing to finetune the model at a lower learning rate; this second variant will be referred to as GMP-PT. We train models using SGD with momentum, with the exception of pre-trained (PT) pruning, for which we found Adam~\cite{kingma2014adam} to yield better results. We use the model state at the end of the epoch which reached highest performance on a held-out validation dataset. All the experiments presented are performed for ResNet18 models under the GMP-RI setup; we provide additional validation for GMP-PT in Appendix~\ref{appendix:post-training}, which supports our conclusions. 

Our setup makes some complementary choices relative to prior work~\cite{hooker2019compressed, hooker_characterising_2020}. 
Specifically, we prune weights by magnitude \emph{globally} as opposed to \emph{per-layer}. 
This will allow us to reach much higher sparsity levels relative to~\cite{hooker2019compressed, hooker_characterising_2020} before model breakdown. 
Further, we chose relatively long model training times (100 epochs for 40-attribute dense and GMP-RI models, 80 epochs for GMP-PT models, and 20 epochs for all single-attribute models), as this leads to both higher accuracy and lower bias metrics.

\paragraph{Accuracy Results.} Using GMP and an extended training schedule, we are able to obtain sparse models that match or outperform the dense baseline, both in terms of accuracy and ROC-AUC values, even at high ($\geq 99\%$) sparsities, while providing substantial improvements in theoretical FLOPs (computed as in~\cite{evci2020rigging}), and practical inference speed on CPU when using the DeepSparse inference engine\cite{NM}.  We present our results for dense and sparse (GMP-RI) models trained to predict all 40 attributes in Table~\ref{tab:celeba-acc}, which show that sparse models can outperform the dense one, even at high sparsities. 
This is also confirmed by the more robust AUC metric, which is agnostic to the prediction threshold; at all sparsity levels, except for 99.5\%, we can observe a slight improvement in AUC scores over the dense models. We observe a similar trend regarding the quality of sparse models over the dense baseline with single-attribute training. This is in contrast to previous work~\cite{hooker_characterising_2020}, which observes a degradation of sparse models over dense even at 90\% sparsity. We believe our improved results are due to the use of a better pruner (global over uniform layer-wise magnitude pruning), and improved training schedule. Nevertheless, they further motivate our study of properties of sparse models, beyond accuracy. 

Additionally, we examined randomly-selected images in each category manually, to validate the quality of the human ratings and the images presented to the automated classifier (see Appendix~\ref{appendix:ui_tool} for screenshots).

\begin{table}[t]
    \centering
    \scalebox{0.51}{
    \begin{tabular}{cccccccc}
        \toprule
         \multirow{2}{*}{Metric} &  \multirow{2}{*}{Dense} & \multicolumn{6}{c}{Sparsity (\%)}\\
         & & 80 & 90 & 95 & 98 & 99 & 99.5 \\
         \midrule 
         Accuracy (\%) & 90.4  & 90.8 & 91.0  & 91.3 & 91.5 & 91.5 & 91.1 \\
        AUC (\%) & 80.5$\pm$0.2 & 81.0 $\pm$ 0.2 & 81.3 $\pm$ 0.3 & 81.5 $\pm$ 0.2 & 81.5 $\pm$ 0.2 & 81.0 $\pm$ 0.1 & 79.7 $\pm$ 0.1 \\
        \midrule
        Inf. FLOPs (B) & 3.64 & 1.40 & 0.998 & 0.683 & 0.386 & 0.241 & 0.145 \\
        Inf. items/sec & 130 & 138 & 181 & 234 & 318 & 373 & 403\\
        \bottomrule 
    \end{tabular}
    }
    \caption{Average Accuracy AUC, estimated inference FLOPs, and inference times on CPU (using the DeepSparse Engine~\cite{pmlr-v119-kurtz20a}) for ResNet18 models jointly trained on all 40 binary attributes. We report results after running each experiment from 5 random seeds. For better readability, we present AUC scores as percentages. We omit variances for the accuracies, as they are all $\leq 0.1$.}
    \label{tab:celeba-acc}
\end{table}

\section{The Effects of Sparsity on Bias}
\label{sec:sparse-bias}

\subsection{Baseline: Analysis of Dense Models}
\label{sec:dense-bias}

\begin{figure}[ht]
\centering
\begin{tabular}{cc}
\includegraphics[width=0.22\textwidth]{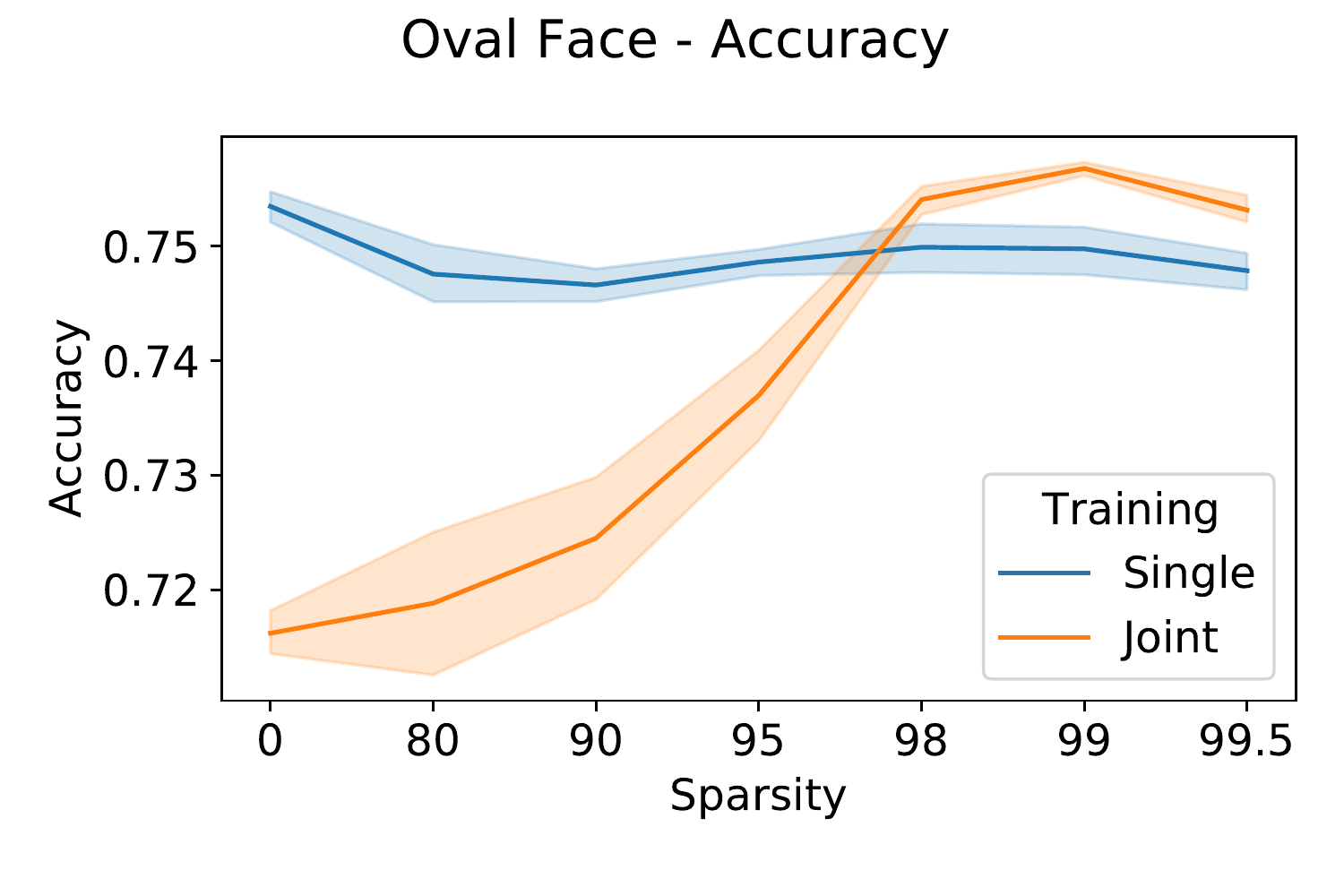} &
\includegraphics[width=0.22\textwidth]{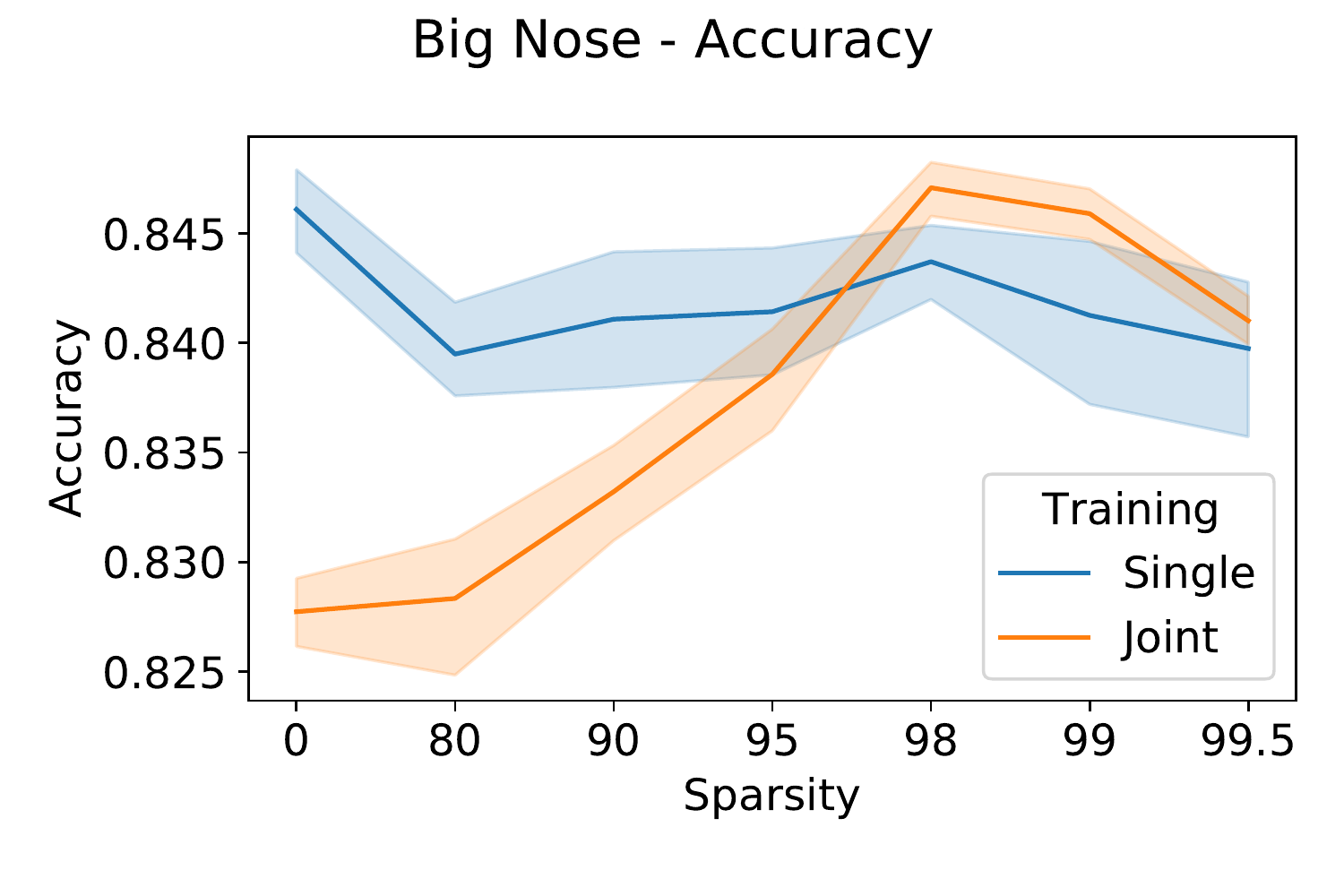} \\
  \includegraphics[width=0.22\textwidth]{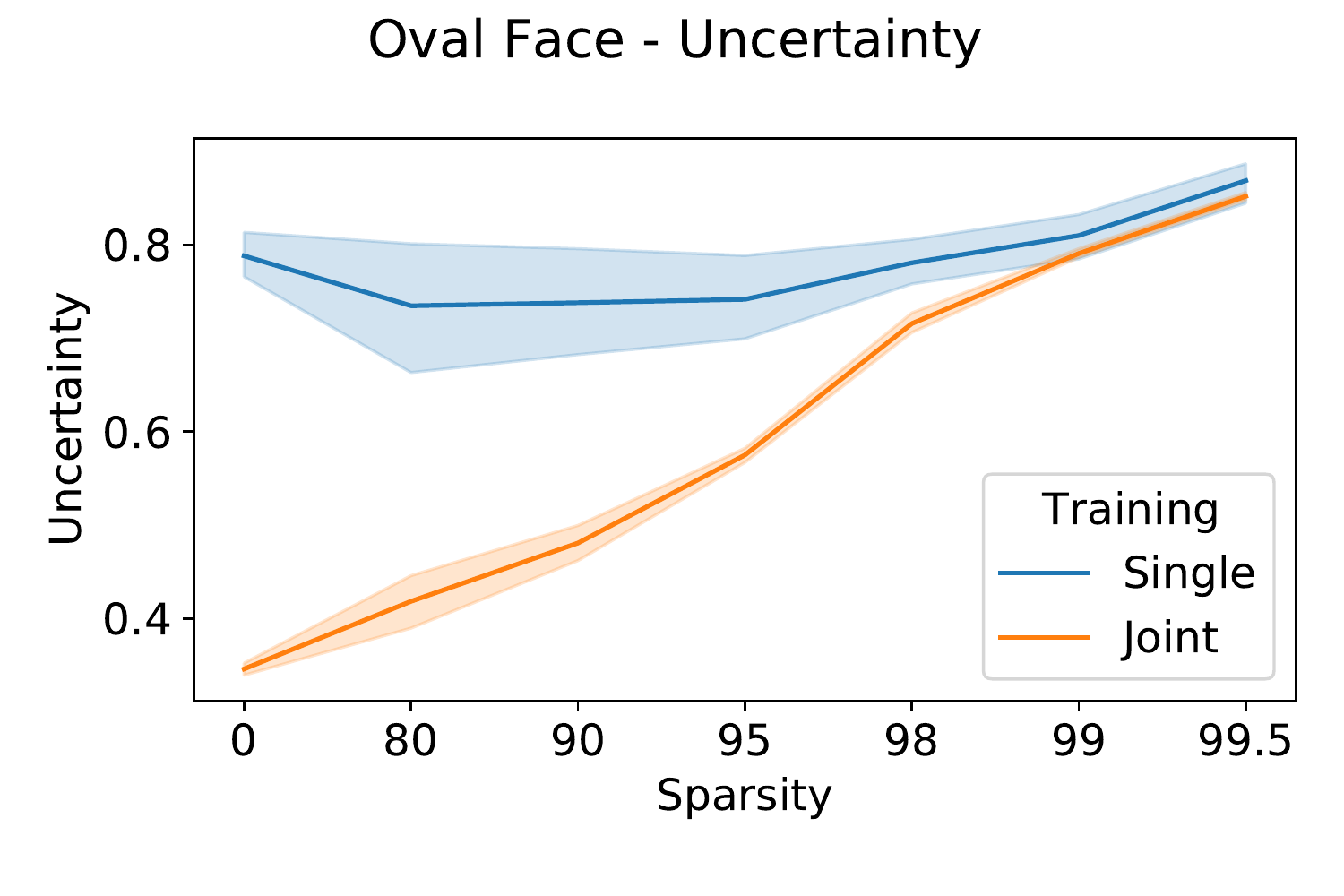} &
      \includegraphics[width=0.22\textwidth]{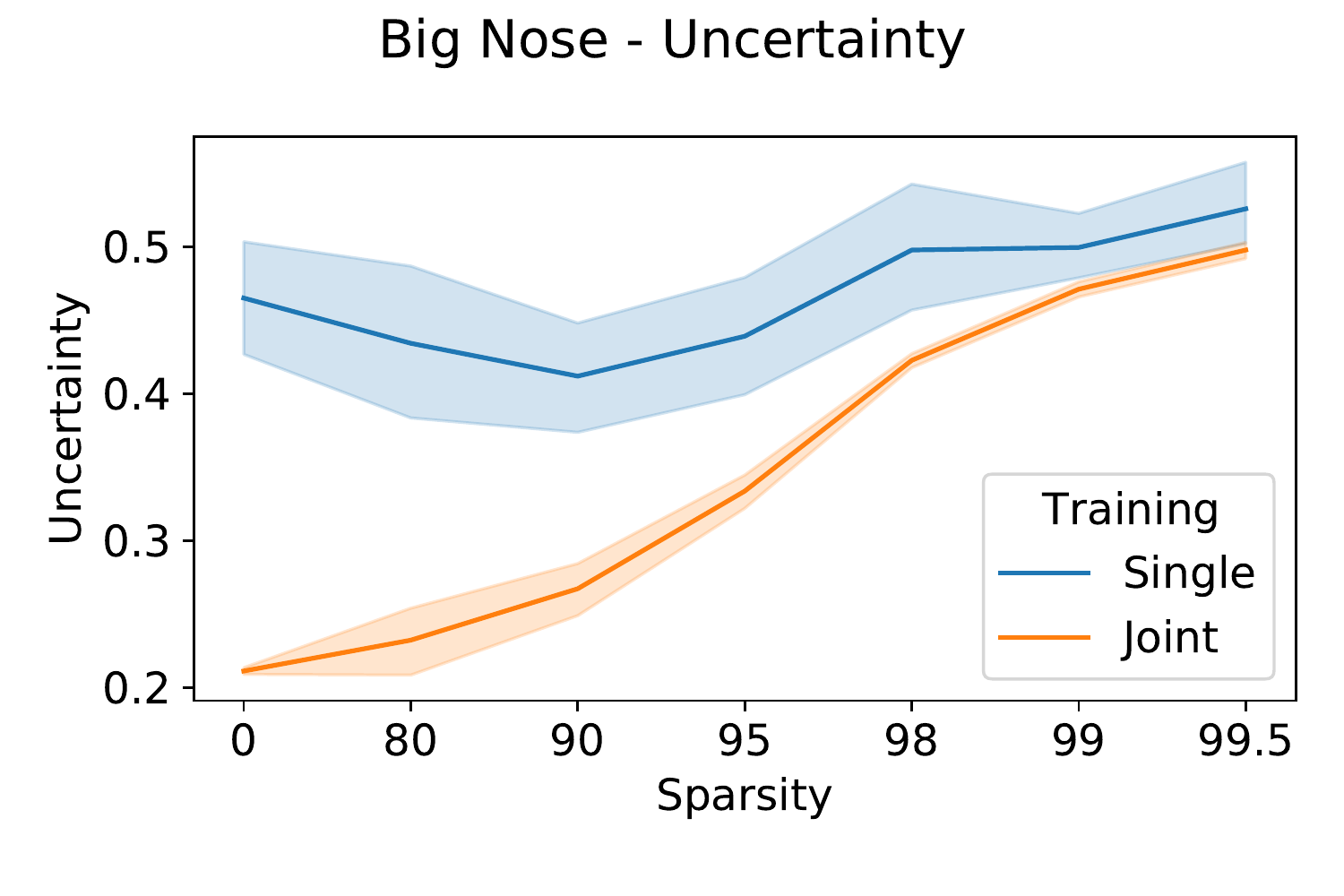} \\
    \includegraphics[width=0.22\textwidth]
  {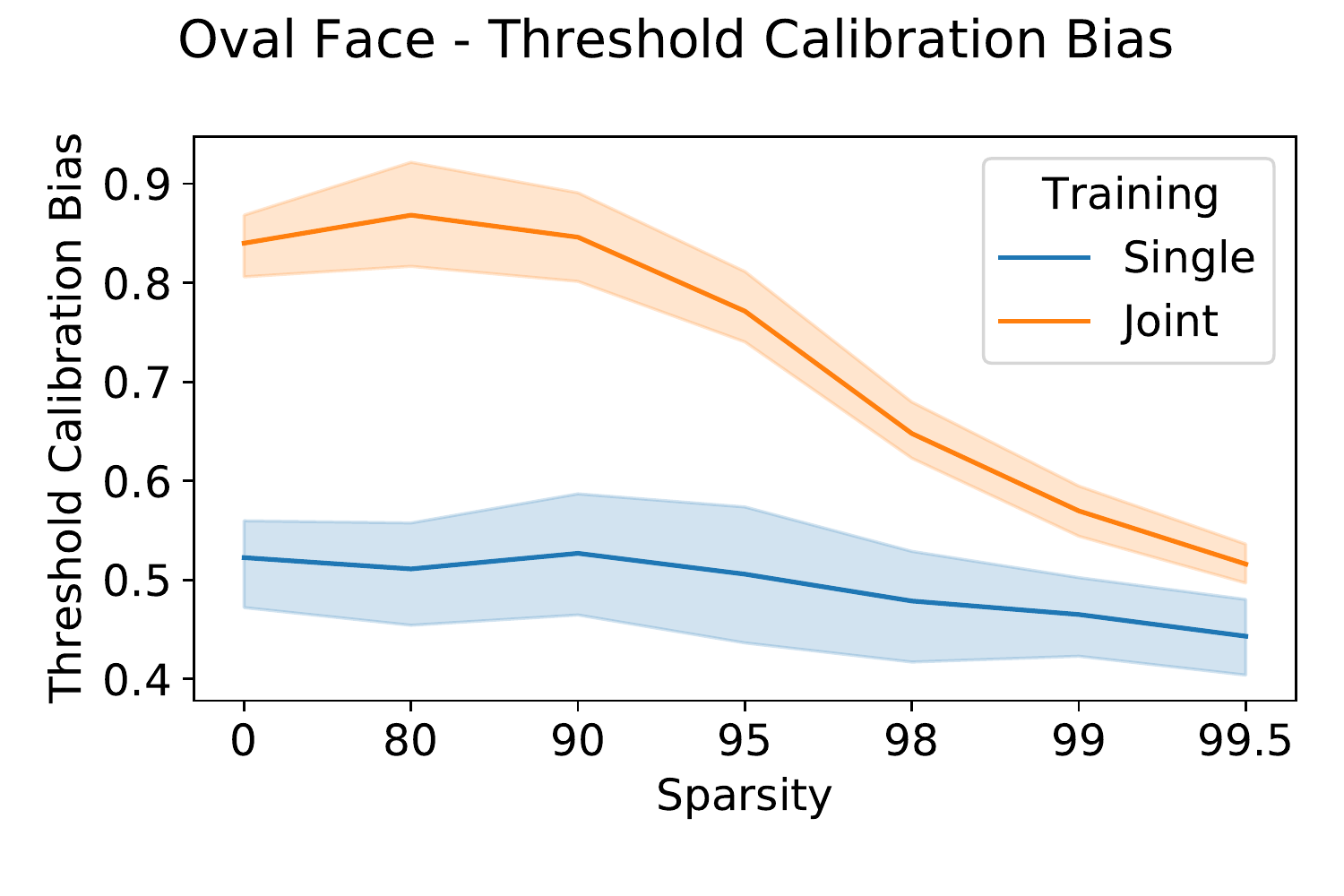} &
  \includegraphics[width=0.22\textwidth]
  {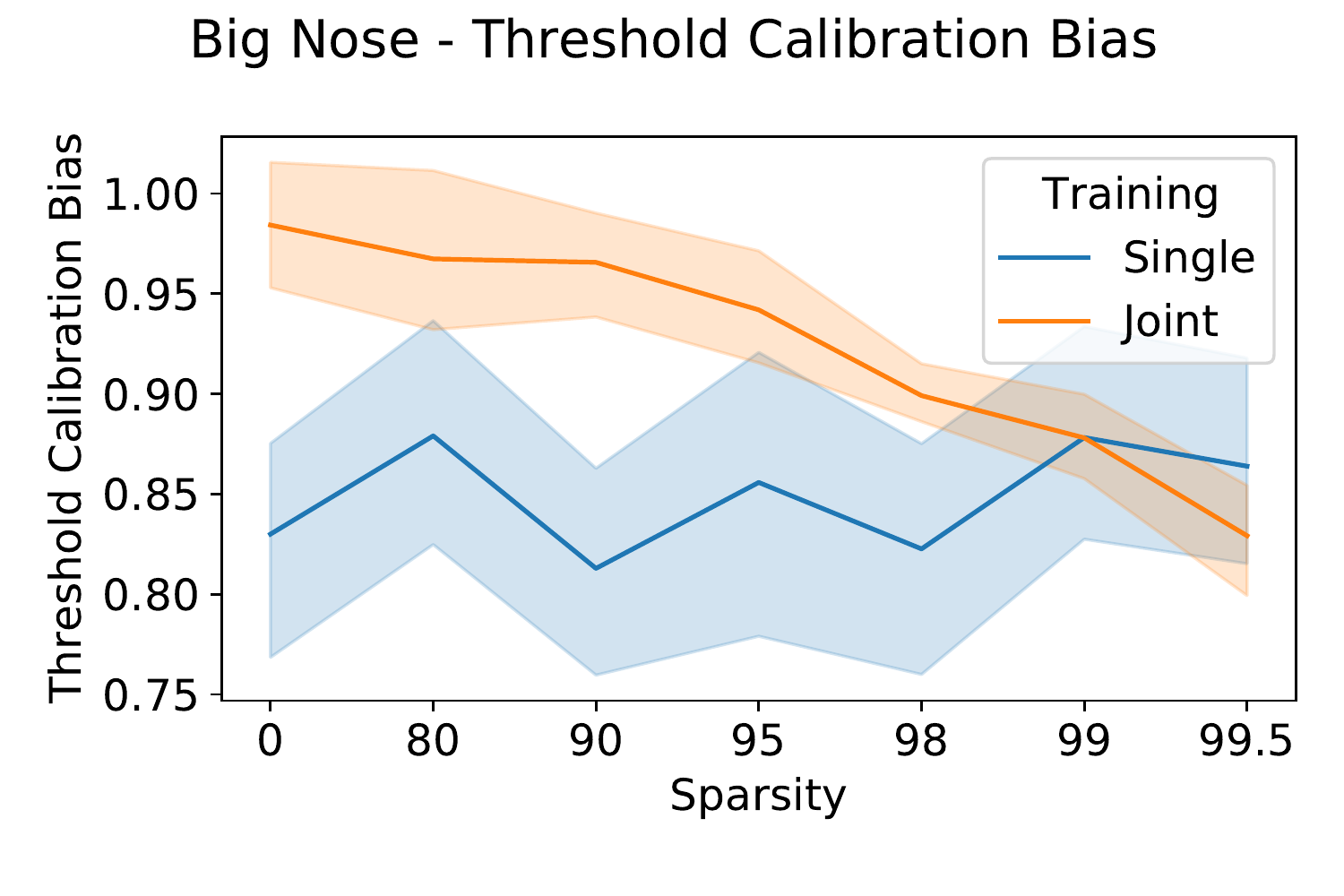} \\
  \includegraphics[width=0.22\textwidth]
  {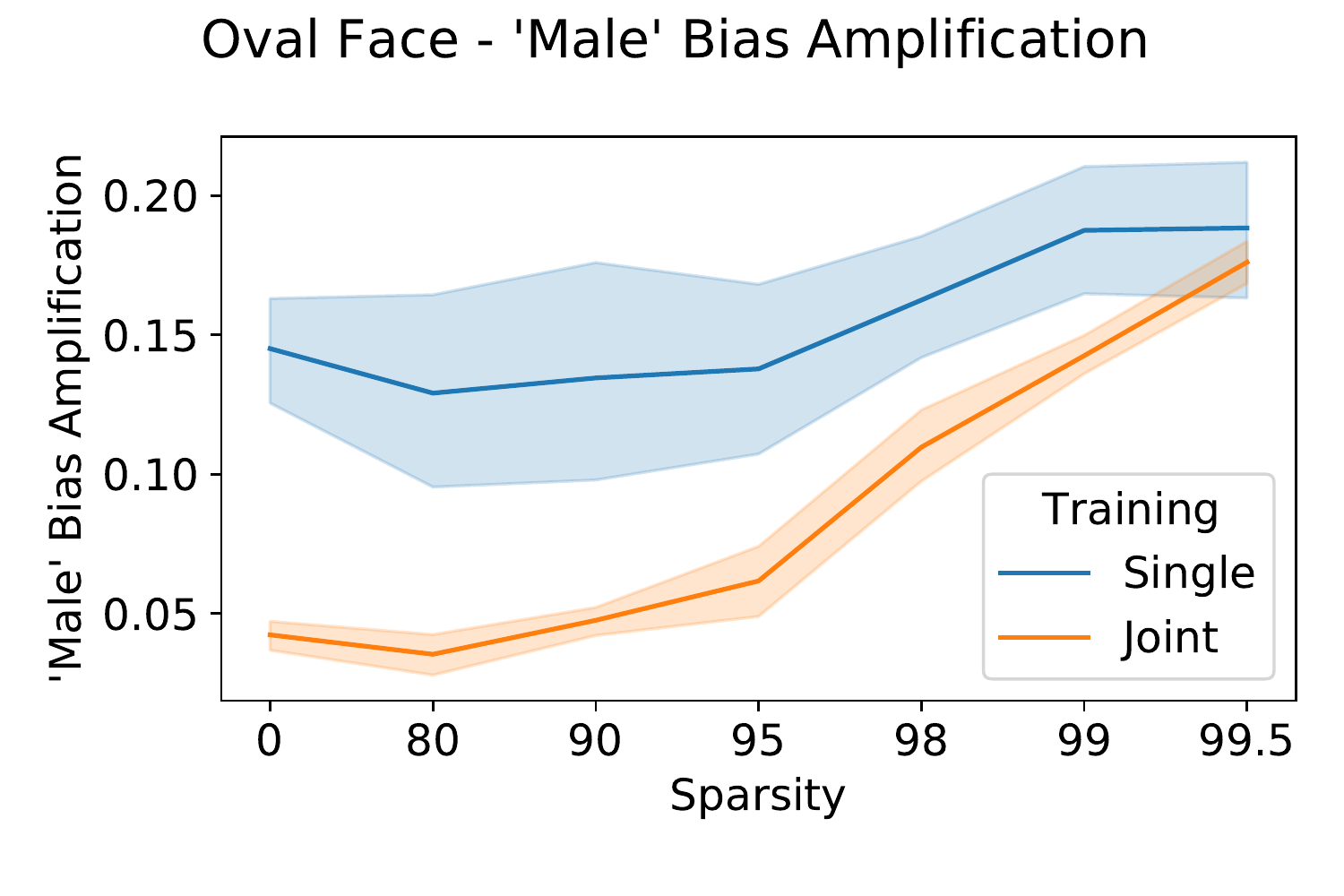} &
    \includegraphics[width=0.22\textwidth]{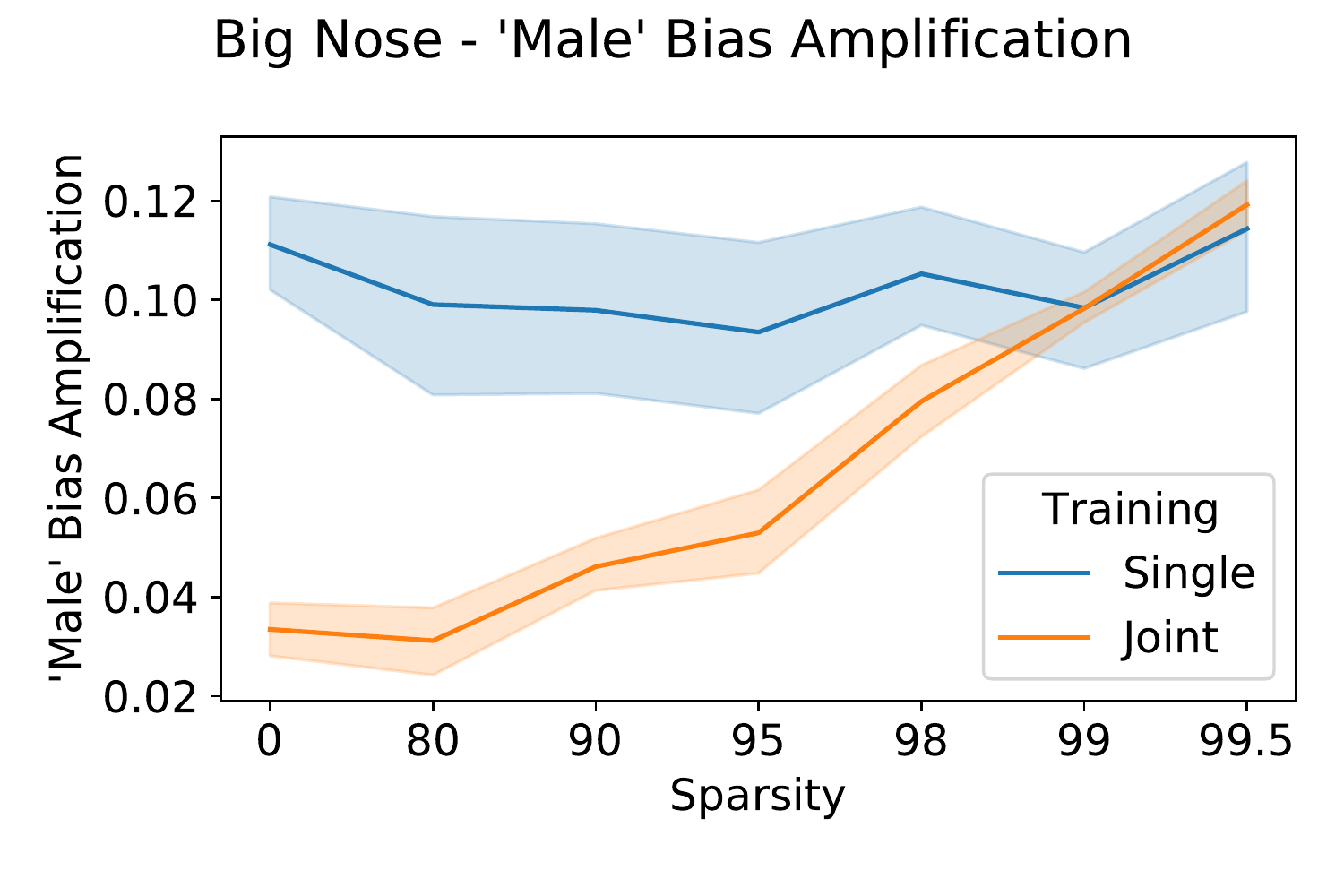}\\
\end{tabular}
    \caption{Effect of single versus joint training of attributes on accuracy (first row), uncertainty (second row), Threshold Calibration Bias (third row), 
    and Bias Amplification for the `Male' attribute (fourth row), on the ResNet18 CelebA model, predicting Oval Face (left) and Big Nose (right).}
    \label{fig:celeba_rn18_single}
\end{figure}

\paragraph{Systematic Bias in Uncompressed Models.}
Examining bias in dense models, we find that, when jointly-trained across all attributes, they tend to under-predict the less prevalent output value for each attribute, with an average TCB of 0.9. Models trained on a single attribute have a worse under-prediction error than jointly-trained models at lower sparsities; for instance, predictions for Oval Face had a TCB of 0.84 when trained jointly with all other attributes, but 0.52 when trained singly. Additionally, dense models were overconfident with respect to the prediction probability, with an average ECE of 0.054 for jointly-trained models. Single-attribute dense models showed higher uncertainty (Figure \ref{fig:celeba_rn18_single} and Appendix \ref{appendix:missing_data}), despite having higher accuracy than jointly-trained models. 

\begin{figure}[ht]
\centering
\begin{tabular}{cc}
  \includegraphics[width=0.22\textwidth]{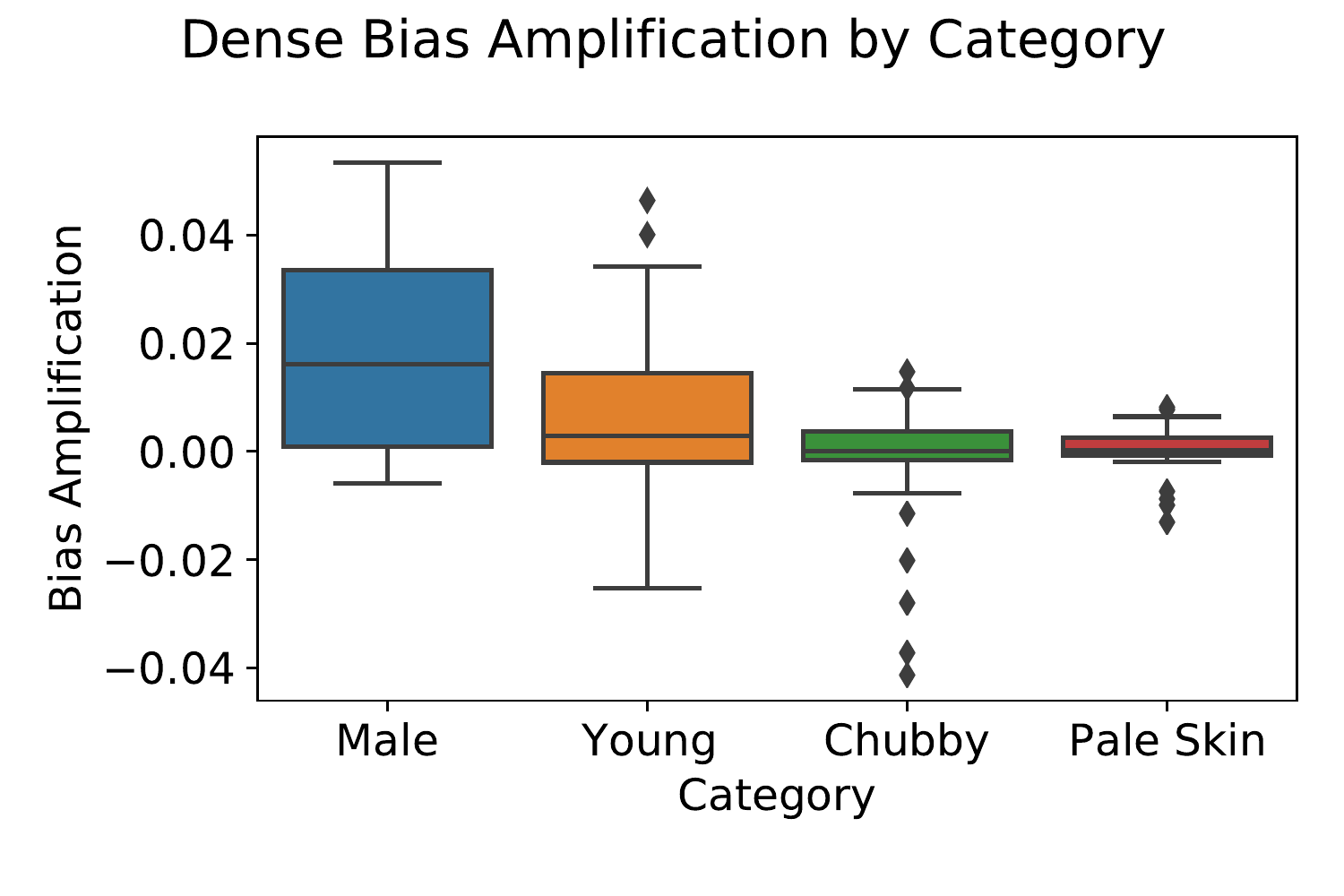}
  &
\includegraphics[width=0.22\textwidth]{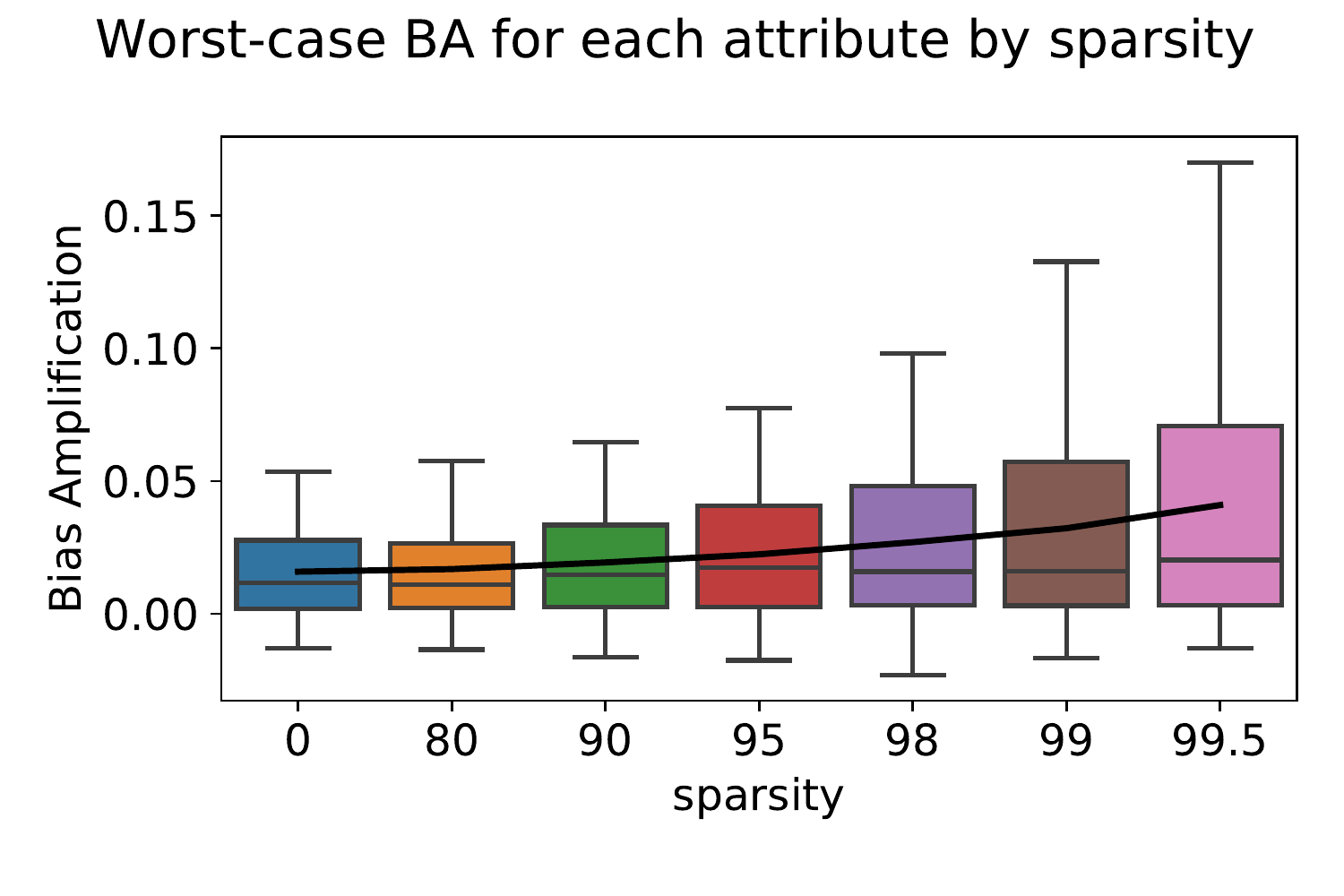}
    \end{tabular}
    \caption{(Left) Bias Amplification by category for dense ResNet18 CelebA models. (Right) Distribution of Worst-Case Bias Amplification across identity categories, for all attributes and sparsities, CelebA on ResNet18. }
    \label{fig:celeba_dense_ba_by_cat_sparse_dense_reg}

\end{figure}

\paragraph{Category Bias in Uncompressed Models.}
Dense models exhibit non-trivial bias amplification (BA), for both singly and jointly-trained attributes. The results show two trends. 
The first, shown in Figure~\ref{fig:celeba_dense_ba_by_cat_sparse_dense_reg} (left), is that BA is substantially higher with respect to specific categories: for instance, with respect to Male and Young, relative to Chubby and Pale Skin. 
The attributes with highest BA value for dense joint training are Double Chin (Male, 0.053), Wavy Hair (Male, 0.047), Wearing Necktie (Young, 0.046), Pointy Nose (Male, 0.045), Chubby (Male, 0.043), and Oval Face (Male, 0.042). (See Appendix~\ref{appendix:tabular} for a full table.) These attributes rank in the top five for several identity categories, suggesting that they are prone to correlations.

The second trend is that single-attribute training shows a much higher BA than joint training. (See the bottom row of Figure~\ref{fig:celeba_rn18_single}, 0\% sparsity.) 
For instance, BA with respect to `Male' is about three times higher when training singly rather than jointly in the case of Oval Face and Big Nose (0.15 vs 0.04 and 0.11 vs 0.03). 

\paragraph{Discussion.} 
It appears that both compressed and uncompressed models are still prone to bias amplification. 
From the point of view of our analysis, the presence of bias in the dense model allows us to compare against sparse models.

\paragraph{Manual Review of Celeb-A Samples.} 
It is tempting to ascribe intuitive explanations to the above correlations. 
However, examining the above attributes more closely, we observe that they have low accuracy and high uncertainty values. Inspecting randomly chosen images, we noticed that attributes such as Pointy Nose often appear difficult to classify, even for human raters. Others, such as Wearing Necktie, are often \emph{impossible to observe directly} on the \emph{cropped version} of the image typically used for this task\footnote{Human raters were asked to assign labels using the uncropped version of the image.}. Finally, an inspection of images shows that Wearing Lipstick appears difficult to judge from the appearance of the mouth, without relying on indirect information, such as the person's gender, or presence of other makeup. Thus, even though we do not detect large bias amplification for this attribute, we consider this measurement unreliable. See Appendix~\ref{appendix:ui_tool} for examples from these categories.

\begin{figure}[ht]
\centering
\begin{tabular}{cc}
    \includegraphics[width=0.22\textwidth]{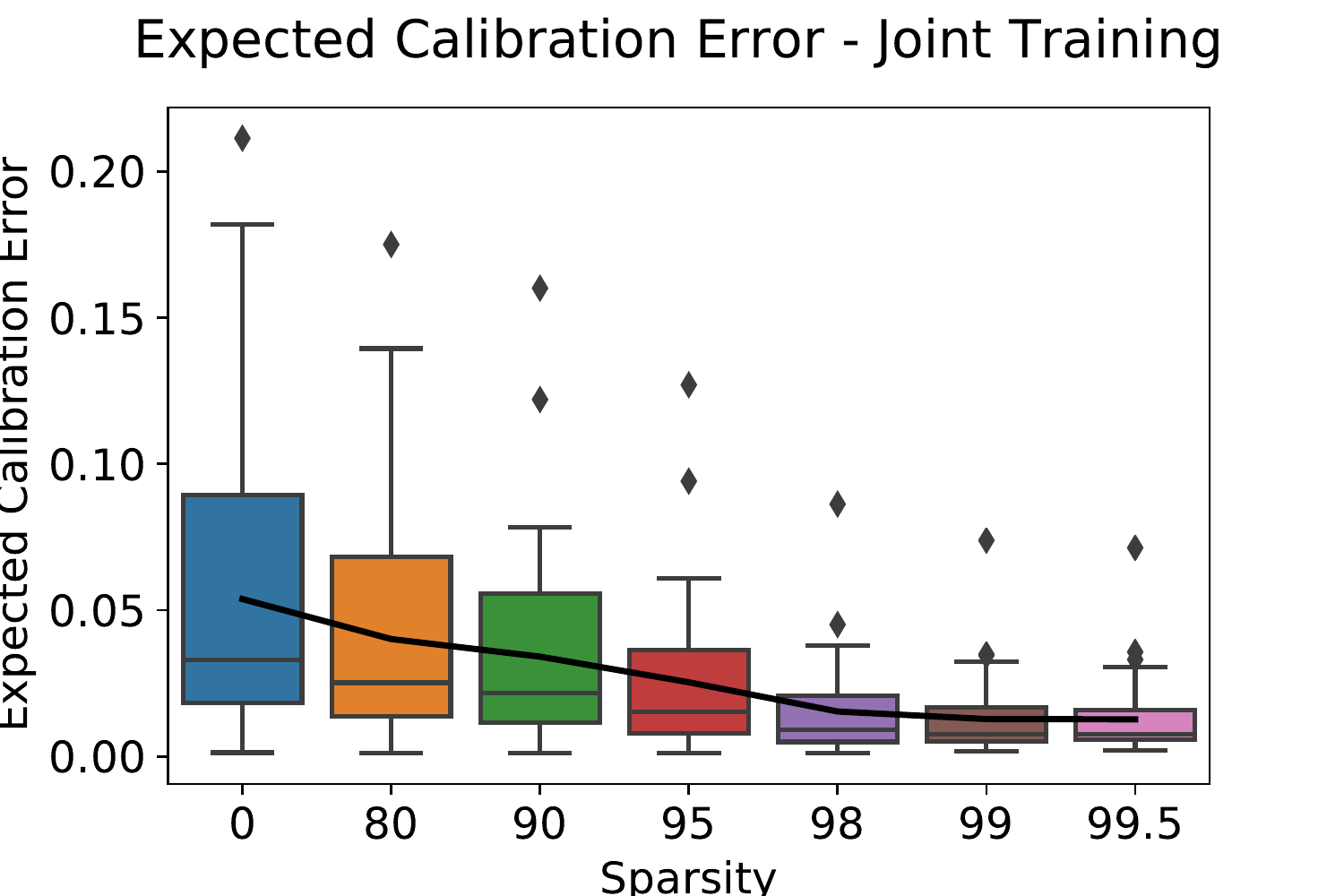} &
    \includegraphics[width=0.22\textwidth]{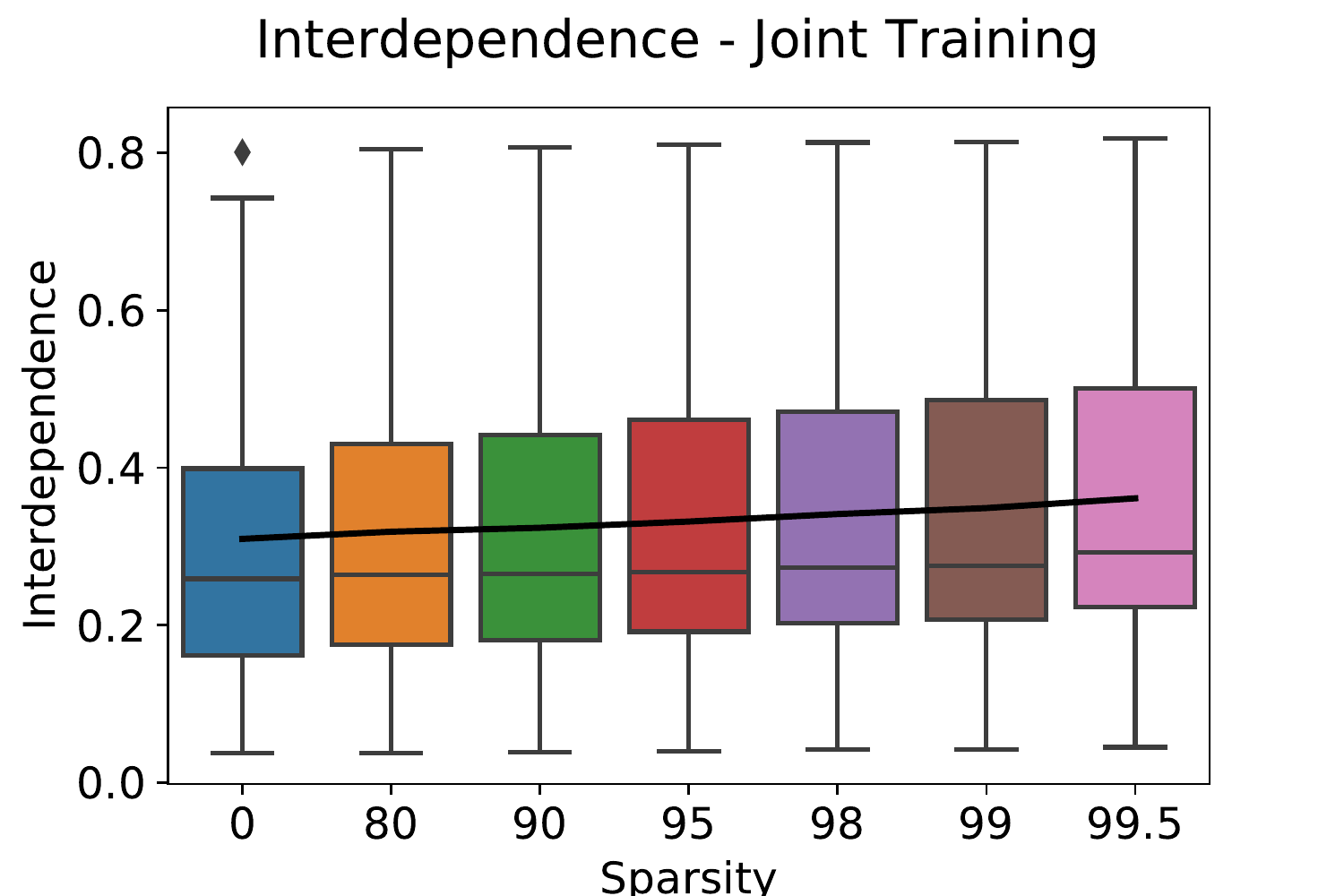}\\
        \includegraphics[width=0.22\textwidth]{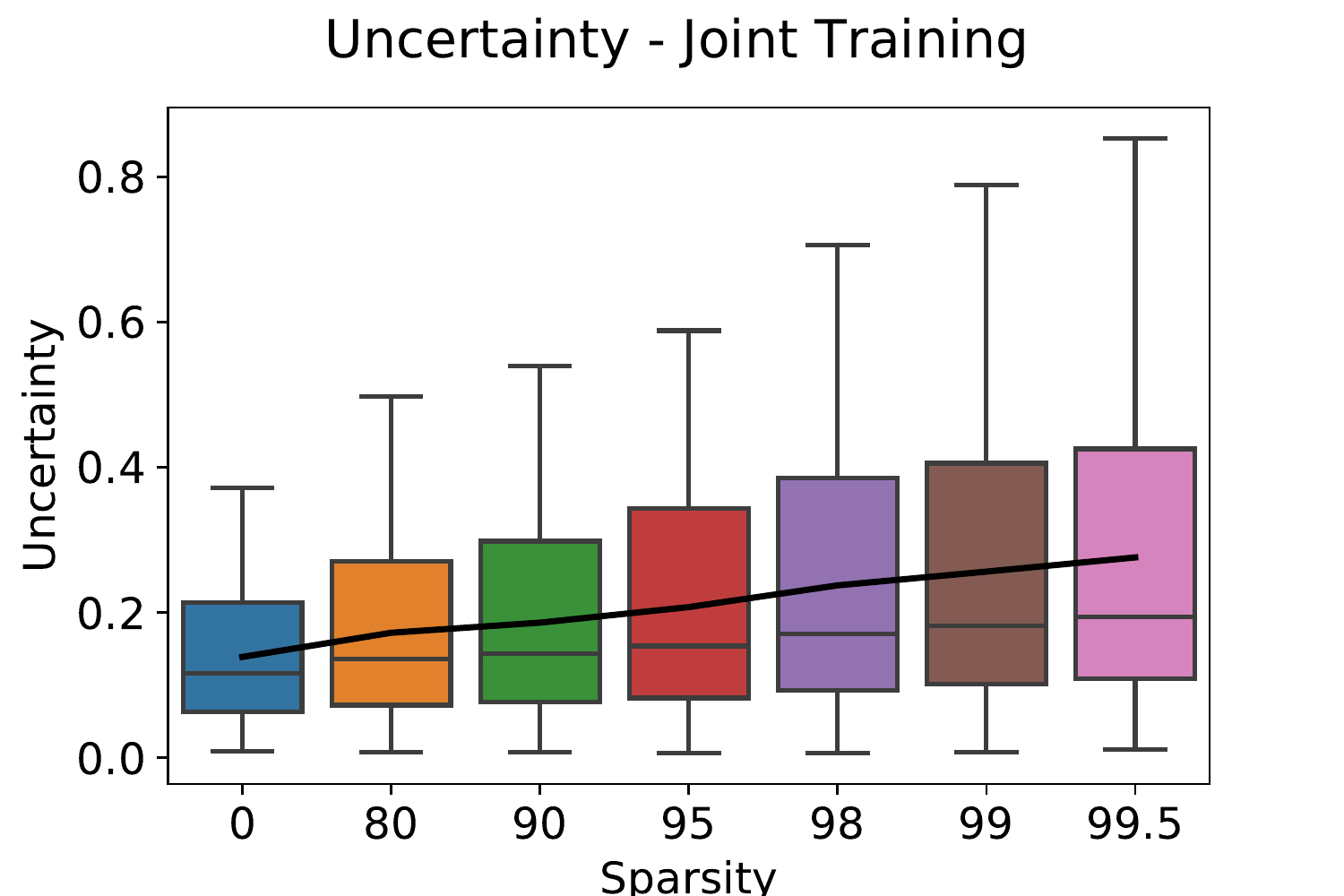} &
    \includegraphics[width=0.22\textwidth]{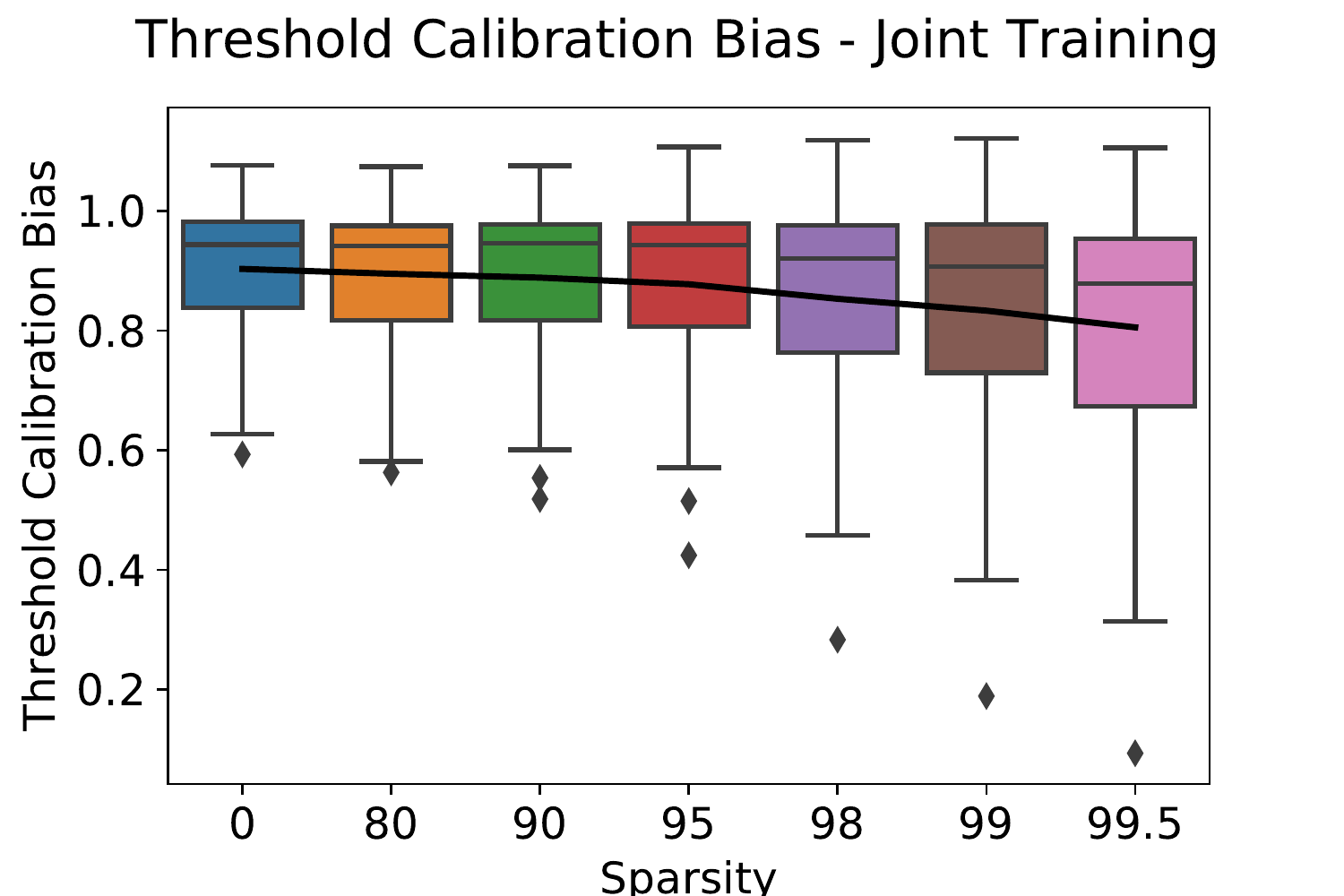}\\
\end{tabular}
    \caption{Systematic Bias metrics (TCB, ECE, Interdependence) of ResNet18 models jointly trained on all CelebA attributes. The thick black line denotes the mean value at each sparsity level.
    In this and all boxplots, the horizontal line represents the median across all CelebA attributes, the edges of the box denote the $25^{th}$ and $75^{th}$ quartiles, and dots indicate all points more than 2.5 times the distance from the mean to the respective quartile.
    }
    \label{fig:celeba_rn18_joint_systematic}
\end{figure}

\subsection{The Effect of Sparsity on Systematic Bias}

Figure~\ref{fig:celeba_rn18_joint_systematic} shows the effect of pruning CelebA models jointly-trained on all attributes on systematic bias, in the random initialization (RI) setup.  
First, notice that, as we increase model sparsity, accuracy stays largely unchanged. 
Yet, other characteristics of the model change considerably. 
Threshold Calibration Bias (TCB) worsens with sparsity for jointly trained models, with an ever-lower proportion of predictions of the less popular value of each attribute. (Consider that the average TCB for dense models is 0.90, while for 99.5\%-sparse models it is 0.81.) %
Uncertainty goes up considerably for almost every attribute, roughly doubling from dense to 99.5\%-sparse models. 

Combining these two observations, we note that in our experiments, jointly-trained sparse models are \emph{better calibrated} than dense with an average ECE of 0.013 for jointly-trained 99.5\% sparse models versus 0.054 for dense models. 
(Note that~\cite{chen2022wineverythinglottery} observe similar behavior of ECE for Lottery Tickets~\cite{frankle2018lottery}, at lower sparsity, and on different datasets.) 
Finally, label interdependence increases with sparsity, from an average $R^2$ of 0.31 to 0.36, suggesting that the more compact feature representation in sparse models results in greater entanglement between the features for every attribute.

For singly-trained models, uncertainty is largely unchanged as sparsity increases, perhaps due to already having high values in the dense model, relative to the jointly-trained model. 
In effect, jointly-trained models have lower uncertainty than singly-trained ones at lower sparsities, but roughly equal uncertainty at higher sparsities. (See Figure~\ref{fig:celeba_rn18_single} and Appendix~\ref{appendix:missing_data} for full data.) 
Threshold Calibration Bias confirms this trend: TCB is roughly constant with sparsity for singly-trained models, but gets worse (decreases) for jointly-trained models. Thus, jointly-trained models are less miscalibrated at lower sparsities relative to singly-trained ones, but similarly miscalibrated at higher sparsities.

\subsection{The Effect of Sparsity on Category Bias}

Next, we focus on the effect of sparsity on bias amplification. 
Here, the expectation is that, if sparse models exhibit more bias, for instance by picking up on spurious correlations, bias amplification should increase. 
We first examine this trend in Figure~\ref{fig:celeba_dense_ba_by_cat_sparse_dense_reg} (right), for jointly-trained models. 
We observe that BA presents a slight increase w.r.t. sparsity between 90 and 95\%, after which the increase is more pronounced. 
The values for BA at the highest sparsity levels are largely determined by the BA values of dense models, with a coefficient of determination $R^2=73.2$.

In contrast, when we examine runs with \emph{single-attribute training} (bottom row of Figure~\ref{fig:celeba_rn18_single} and Appendix~\ref{appendix:missing_data}), 
we observe that, in this case, sparsity has very little effect on bias amplification for the hidden `Male' category, which stays roughly constant, within noise bounds.  
However, recall from our previous discussion that the baseline (dense) bias amplification is significantly higher for single-attribute training relative to jointly-trained attributes. 
Specifically, BA for \emph{dense singly-trained models} is roughly as high as for \emph{99.5\%-sparse jointly-trained} models. 
One interpretation is that the additional prediction heads of the jointly-trained models encourage a more robust feature representation which \emph{discourages bias at low sparsity}; at high sparsity, however, the compactness of representation induces more bias. 
Thus, switching to singly-trained attributes may be a good strategy at high sparsity levels. 

Another observation is the high correlation between the evolution of \emph{uncertainty} (second row in Figure~\ref{fig:celeba_rn18_single}), TCB (third row), and that of bias amplification (fourth row), relative to the sparsity increase. 
Specifically, the increase in output uncertainty is linked to stronger bias amplification. 

\textcolor{black}{We further investigated whether co-training the identity category with the attribute of interest encourages more diversity in the representation. 
In this case, we observed a very similar trend regarding BA as for singly-trained attributes, which indicates that the source of bias goes beyond the relationship between the two attributes.} These results are shown in Appendix~\ref{appendix:ba_combined}.

\subsection{Injecting Backdoor Features in Sparse Models}

\begin{figure}[ht]
\centering
\begin{tabular}{cc}
\includegraphics[width=0.22\textwidth]{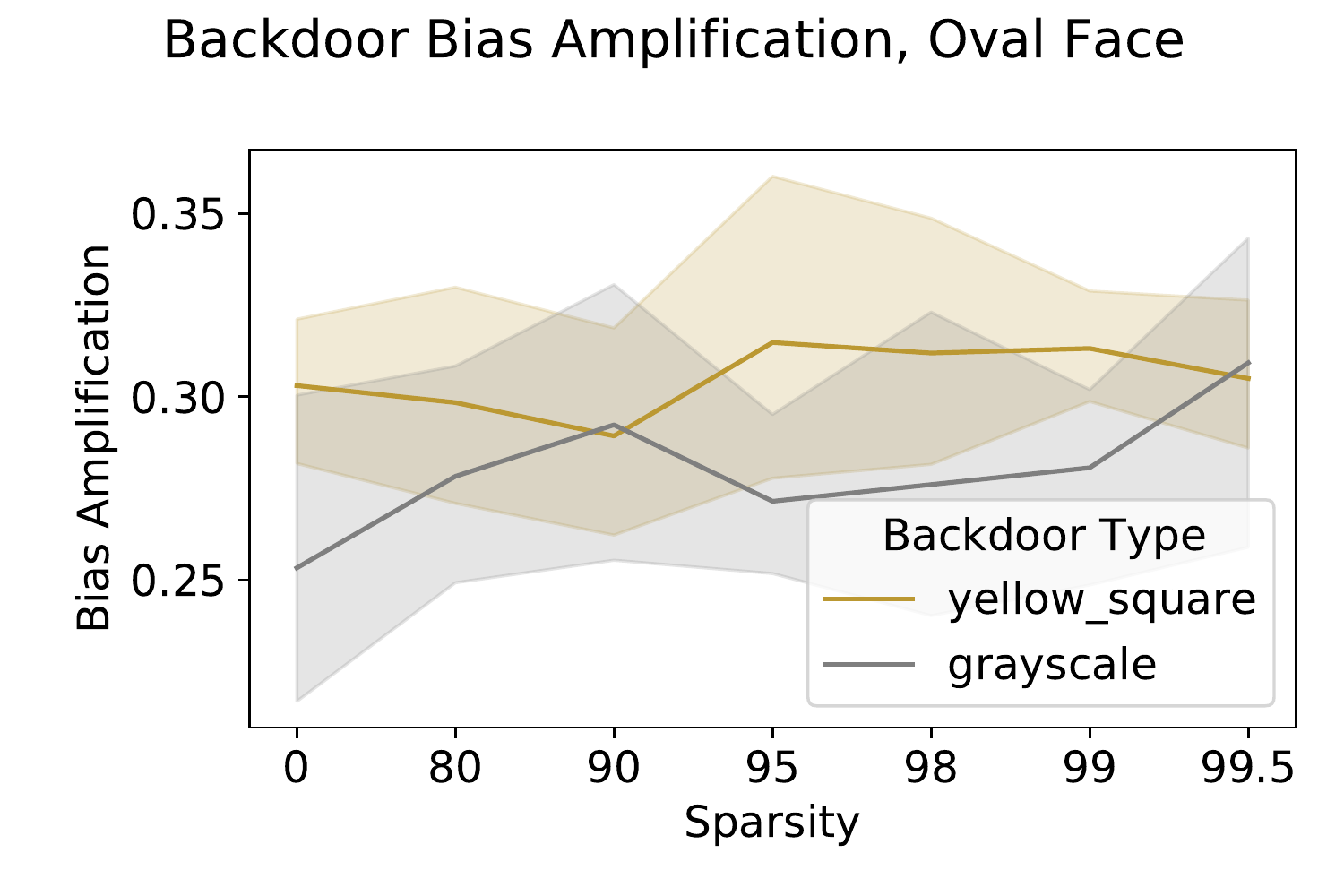} &
\includegraphics[width=0.22\textwidth]{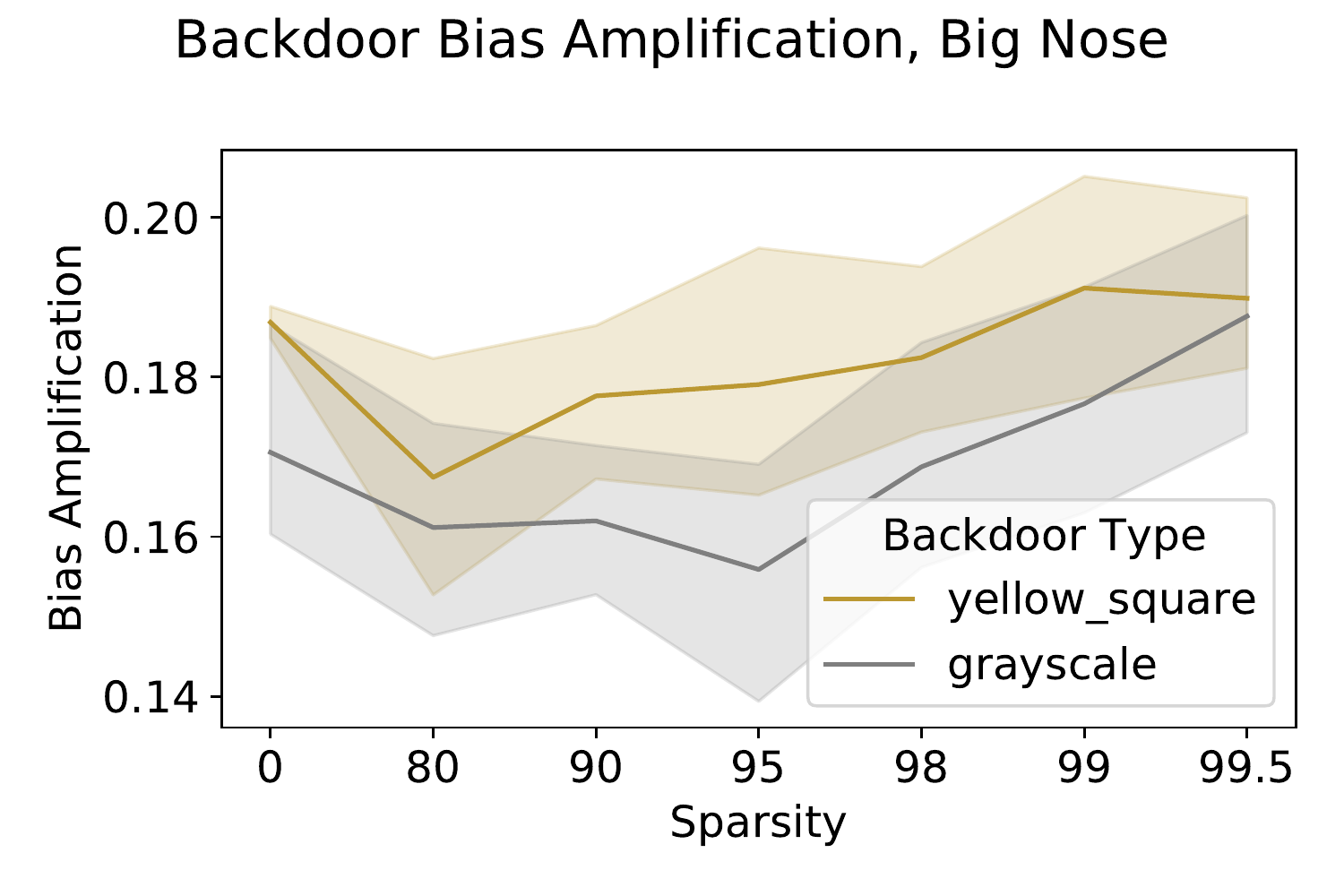} \\
\includegraphics[width=0.22\textwidth]{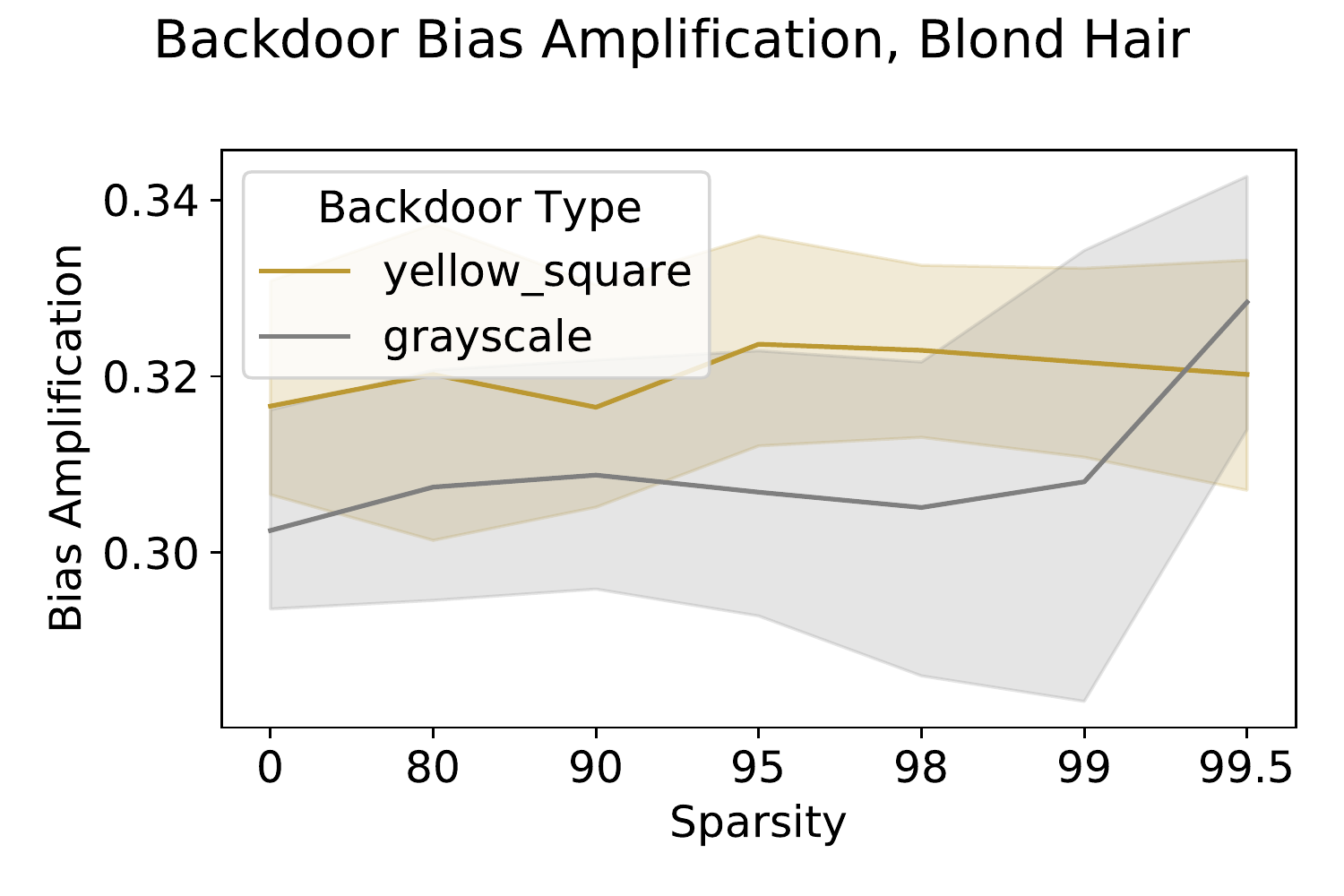} &
\includegraphics[width=0.22\textwidth]{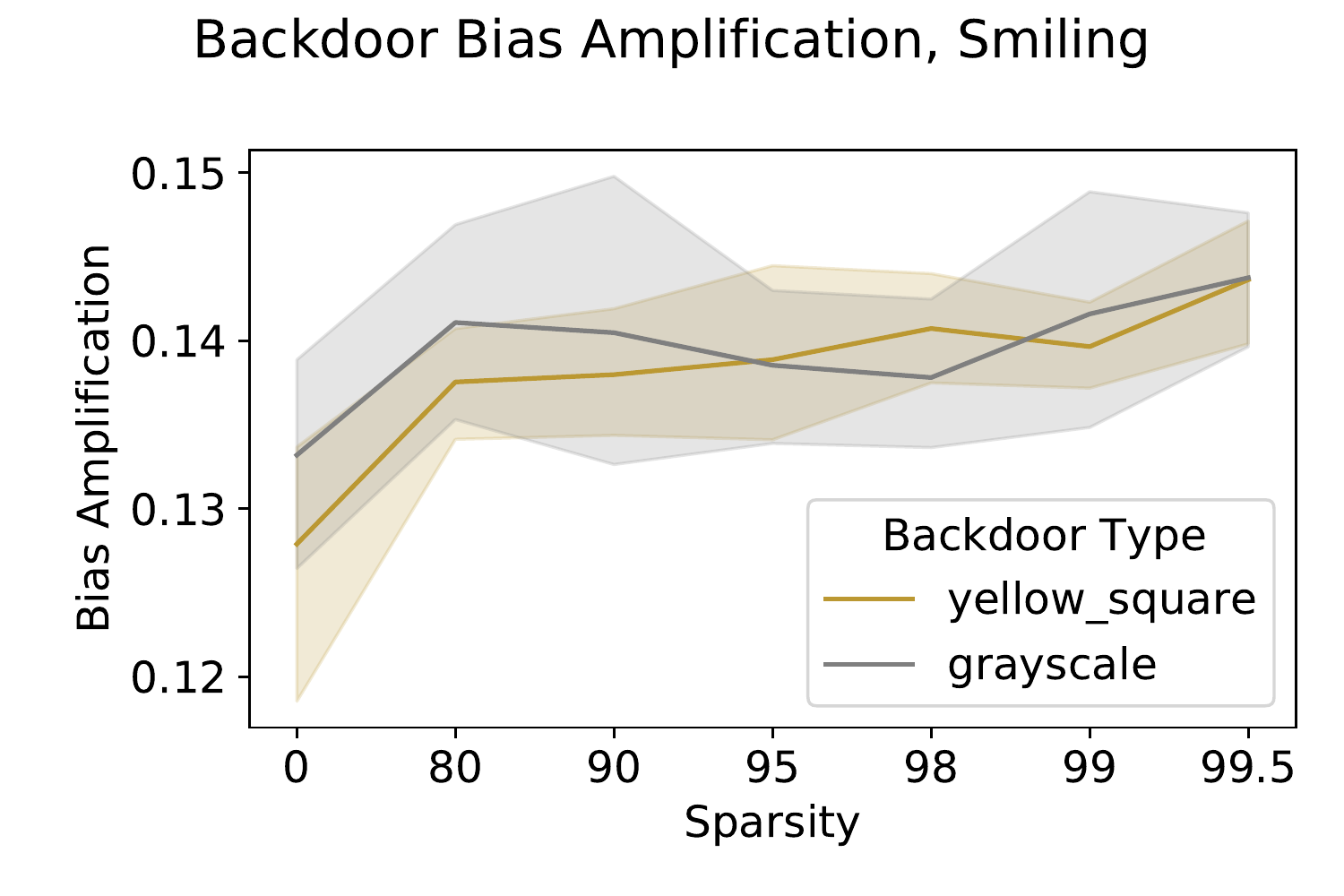} \\
\end{tabular}
    \caption{Effect on BA of adding a backdoor feature when performing single-attribute training for four attributes.}
    \label{fig:celeba_rn18_backdoor}
\end{figure}

To study the amplification of bias by sparse models, we artificially introduce bias in the data through specific modifications to the samples, via ``backdoor attacks''. 
We then measure the effects on a similarly ``backdoored'' test set, for dense and sparse models for single-attribute prediction.  We follow a similar approach to \cite{salman2022does, Wang2020TowardsFI} for backdooring: we apply a fixed transformation---grayscaling of the entire image~\cite{Wang2020TowardsFI}, or inserting a small yellow square~\cite{salman2022does} --- to the majority of training samples with a positive label, and to a smaller subset of those having the negative label. On the test set, we keep an even ratio of backdoored samples. We perform both the grayscale and yellow square backdoor attacks when training with four separate attributes: Blond, Smiling, Oval Face and Big Nose. 
We use a backdooring split of 95\% positive /5\% negative for Blond and Smiling, and 65\% positive /35\% negative for Oval Face and Big Nose. The smaller split prevents the model from simply memorizing the backdoor on harder tasks.

Targeted backdoors enable us to better control and isolate the source of bias introduced in the models.
We consider category bias, and focus on bias amplification (BA) as our main metric. Specifically, in the definition of BA described in Section~\ref{sec:bias-describe} we consider backdooring as our identity category, \emph{i.e.} if a sampled is backdoored, then it has identity category 1, and 0 otherwise. 

Our results in Figure~\ref{fig:celeba_rn18_backdoor} show that, as expected, BA increases substantially for all models considered. Moreover, we observe that bias is slightly amplified with sparsity, for example on the Big Nose or Smiling attributes. Overall, our study on bias for backdoored models results in similar conclusions to the ``clean'' single label experiments. For example, when examining the BA scores for single label training in Figure~\ref{fig:celeba_rn18_single}, we notice that the values have only a slight increase with sparsity. This suggests that bias is more likely to follow from less diverse feature representations, whereas here the  relationship with sparsity is weaker.

\section{Mitigating Sparsity-Induced Bias}

\subsection{Threshold Calibration}

Inspired by our earlier observation that sparser models tend to show worse threshold calibration bias, we consider what happens when we adjust the thresholds to better fit the true distribution of each attribute. We note that the decision to adjust the threshold is not clear-cut; the logistic loss encourages the correct prediction, rather than the correct \emph{ranking} for each attribute. Further, the threshold adjustment does not take the identity feature into account, and should not be confused with fairness-aware threshold adjustments~\cite{Hardt2016EqualityOO}. Instead, we set a single threshold for each attribute so that the predictions are correctly calibrated on the original CelebA validation set. 

\begin{figure}[ht]
\centering
\begin{tabular}{cc}
  \includegraphics[width=0.21\textwidth]{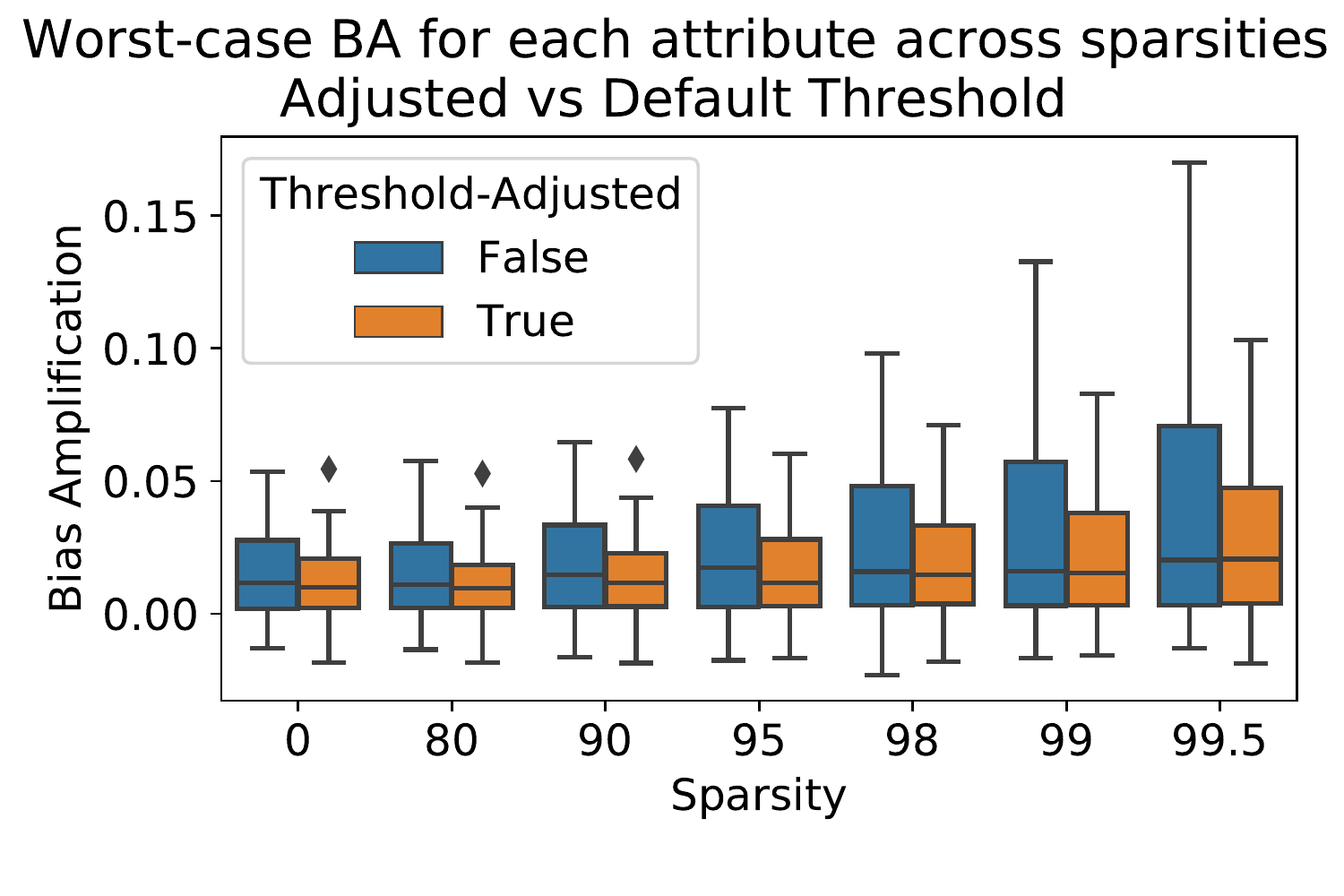} &
  \includegraphics[width=0.21\textwidth]{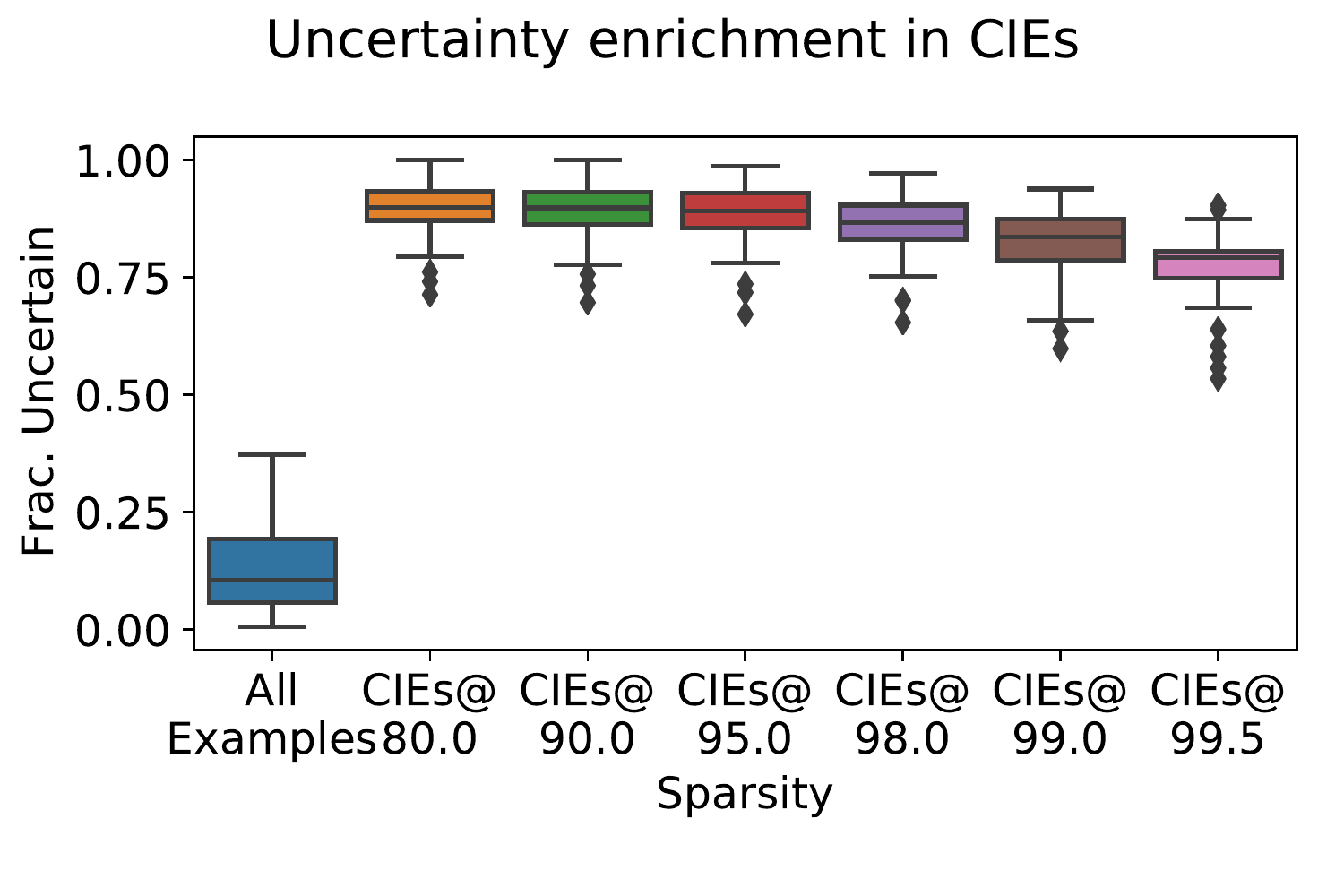}
  \end{tabular}
    \caption{(Left) Effect of threshold calibration on ResNet18 models jointly trained on all attributes. %
    (Right) Proportion of uncertain predictions for \emph{dense} models across all attributes for all elements in the CelebA test set, and for Compression-Identified Exemplars at different sparsities.}
    \label{fig:celeba_rn18_threshold_adj}
\end{figure}

The results of threshold calibration are shown in Figure~\ref{fig:celeba_rn18_threshold_adj} (Left). Despite the fact that the threshold adjustment process is agnostic to identity categories, this simple correction reduces the bias amplification across all sparsities, almost eliminating bias effects at up to 90\% sparsity.%

\subsection{Overriding Sensitive Samples}

\begin{figure}[ht]
\centering

\includegraphics[width=0.47\textwidth]{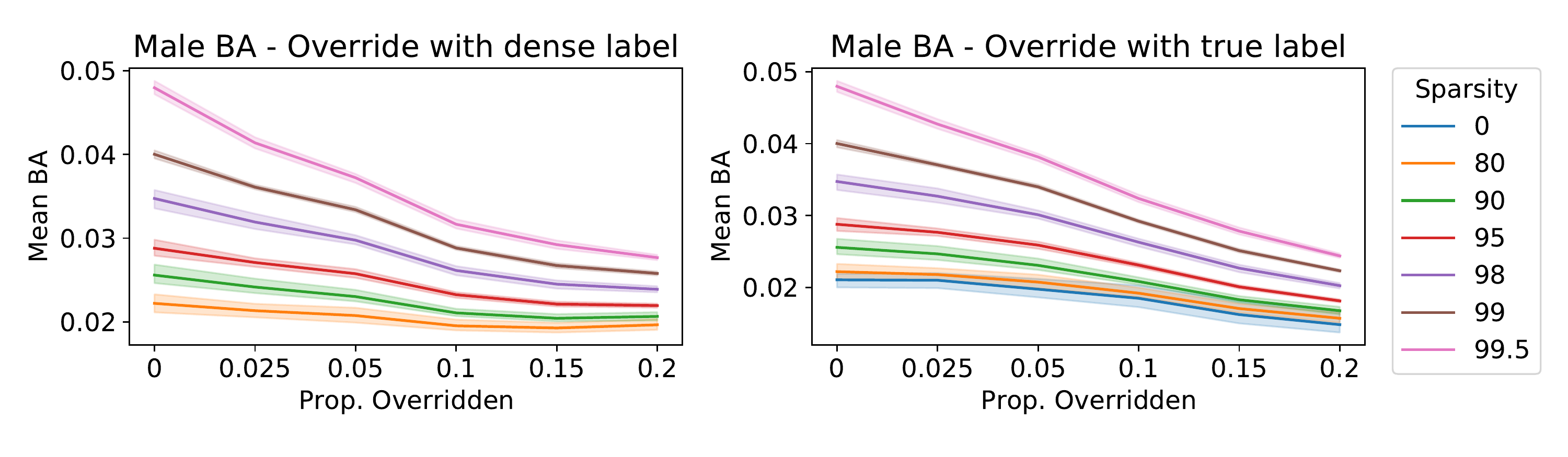}

    \caption{Effect of label overrides on Male Bias Amplification.}
    \label{fig:overrides}
\end{figure}

Since the additional bias amplification in sparse models must be due to test samples whose classification has changed between dense and sparse models, we examine these examples more closely. We focus on Compression-Identified Exemplars (CIEs)\cite{hooker2019compressed, hooker_characterising_2020}, which are the test examples on which the modal dense label across multiple training runs disagrees with the modal sparse label, regardless of which one is correct. For each sparsity, we compute the CIEs across five runs each of the dense and sparse models.
Our results in Figure~\ref{fig:celeba_rn18_threshold_adj} (Right) show that CIEs are greatly enriched for prediction uncertainty, suggesting that improving the predictions of these examples may assist in reducing BA, especially in the sparse models. However, CIEs are expensive to compute due to requiring multiple models for consensus, and are specific to the sparsity level.

Prediction overrides, where a fixed label for a small subset of data is distributed along with the model, and selected over the model prediction at inference time, are common in model deployment. Inspired by our observation that CIEs are highly enriched for uncertain examples, we propose to prioritize the highest-uncertainty data as classified by a dense model, in cases where the dense model already shows positive BA. We replicate this setting on the test dataset. %
This is  consistent with standard practice for override prioritization to improve accuracy, since the most uncertain examples are presumed to have the highest chance of having the wrong label. 

We consider two possible override labels: the correct label, which simulates human overrides, and the dense label, which simulates the best possible label if human labeling is impractical. We apply these overrides to all sparse labels and measure the bias amplification.
Our results (Figure~\ref{fig:overrides} and Figure~\ref{fig:overrides_rn18_RI}) show that overrides with both human and dense labels substantially decrease the bias amplification of models of all sparsities. For instance, using manual overrides for the most uncertain 5\% of examples lowers the mean BA of the 99.5\% sparse model by 23\%, and replacing the top 10\% lowers the mean BA by 35\%. This suggests that the use of uncertainty-based override pipelines is an effective tool for reducing bias amplification on sparse models, even when only the dense model is used to set prioritization.

\section{Additional Validation}

We emphasize the fact that the above observations have been validated on additional datasets and models, so our findings hold generally. 
We discuss these experiments briefly below, and present them in full in the Appendix. 

\paragraph{Additional Validation on CelebA.}%
We experiment with the setup where pruning starts from a pretrained model, for which we include the results in
Appendix~\ref{appendix:post-training}, showing similar results. We additionally prune to N:M (2:4, 1:4 and 1:8) sparsity patterns~\cite{mishra2021accelerating} in Appendix~\ref{appendix:nm_sparsity}, with similar results to lower-sparsity models pruned without this restriction.
Experiments validating our results for singly- and jointly-trained attributes on the MobileNetV1 architecture~\cite{howard2017mobilenets} can be found in Appendix~\ref{appendix:mobilenet}, showing the same trends, but at slightly lower sparsities. We additionally validate the joint training results on the ResNet50 architecture in Appendix~\ref{appendix:resnet50}, with very similar results to ResNet18.
Finally, we repeat the ResNet18 joint training experiments using the \emph{uncropped} CelebA dataset, which ensures that features such as the presence of neckwear are available to the model (as they were to the human labellers). 
We discuss these results in Appendix~\ref{appendix:uncropped_celeba}.

\paragraph{Additional Datasets.} We further validated our findings on two additional datasets. 
The Animals with Attributes (AwA) dataset~\cite{Xian2019AwAZeroShotLC} serves as a useful validation for our observations regarding the effect of sparsity on bias in binary prediction (Appendix~\ref{appendix:awa}). The challenging iWildcam dataset \cite{beery2020iwildcam} validates our observations regarding increased uncertainty relative to sparsity in the context of multiclass classification (Appendix~\ref{appendix:iwildcam}).

\section{Related Work}

\paragraph{Fairness, Bias, and Bias Mitigation.} A number of fairness metrics have been proposed, including individual fairness, which requires that individuals with similar characteristics receive similar outcomes, and group fairness, which requires parity along some metric between individuals in commonly-identified groups~\cite{barocas-hardt-narayanan}. 
Many works propose techniques to remove or mitigate bias in general~\cite{Sagawa2020Distributionally, Wang2020TowardsFI}, while~\cite{Lin2022FairGRAPE} mitigates accuracy bias on compressed models. Notably, \cite{Wang2020TowardsFI}  proposes the use of synthetic benchmarks such as backdooring images. Backdooring is also used by \cite{salman2022does} for evaluating bias in transfer learning. 

\paragraph{Bias Due to Compression.} Seminal work by Hooker et al.\cite{hooker2019compressed, hooker_characterising_2020} initiated the study of compression-induced bias, showing that bias can be amplified by model pruning, and isolate the influence of Compression Identified Exemplars (CIEs) as rare examples in the training data. 
Our work significantly extends this research, by examining compression effects via Bias Amplification, and showing that highly-sparse models may in fact be bias-free for moderate $\leq 90\%$ sparsities, using joint training, global pruning, and additional finetuning. 
In addition, we provide strategies for bias mitigation that do not require knowledge of identity categories, nor any information about compressed models. 

Recent work by Chen et al.~\cite{chen2022wineverythinglottery} studies pruning effects from four aspects: generalization/robustness to distribution shifts, prediction uncertainty, interpretability, and loss landscape, for pruned models obtained via variants of the Lottery Ticket Hypothesis (LTH) approach~\cite{frankle2018lottery, chen2020lottery, chen2021vision}.  
They show that LTH-pruned models match (or slightly outperform) dense models across all these categories. 
Our work is related in that they also study prediction uncertainty for models, noticing that sparse LTH models can be competitive with dense ones in terms of uncertainty, measured as ECE. 
Yet, the focus of our work is different: we perform an in-depth comparison of bias effects, specifically focusing on the high-sparsity range, where we exhibit and carefully analyze the emergence of bias. In addition, we provide a set of techniques for characterizing and mitigating bias in pruned models, which is beyond the scope of~\cite{chen2022wineverythinglottery}. 

Good et al.~\cite{Good2022RecallDI} studies the relative distortions in the recall of a model in relationship with sparsity, and proposed a gradient-based pruning method to decrease the negative effect of sparsity on this metric. Other works analyze the variance in classification error among classes as a proxy for bias in sparse models~\cite{blakeney2021simon}, while others~\cite{joseph2020going, xu2021beyond} use knowledge distillation~\cite{hinton2015distilling} to decrease the misalignment between sparse and dense models. By comparison, our study focuses on characterizing and mitigating bias given a fixed compression scheme, for which we propose different metrics, as well as detection criteria and countermeasures. 

\paragraph{Systematic Bias.} Finally, systematic bias is an important avenue of research that compliments our work by using more sophisticated techniques to identify and categorize hard-to-learn examples~\cite{DOMINO, spotlight, rajani2022seal, Baldock2021DeepLT}. However, these works use finer-grained definitions of systematic bias, and do not consider model compression.

\section{Conclusion}

We performed an in-depth study of bias in sparse models, and showed that it is possible to obtain highly-sparse models without loss in accuracy or AUC. However, these models have higher uncertainty compared to dense ones, and the predicted labels are more interdependent. Bias amplification is often substantially exacerbated at high sparsities ($\geq95\%$) and the bias of individual attributes in sparse models correlates well with their bias in the dense baseline. However, the effect we observe on both systematic and category bias is influenced by the training setting, i.e. joint or individual attribute training.
In future work, we plan to examine the impact of different compression approaches (pruning and quantization techniques) on our bias metrics, more complex countermeasures for mitigating the bias we have shown to arise in highly-compressed models, and further applications, such as language modelling. %

\section*{Acknowledgments}

The authors would like to sincerely thank Sara Hooker for her feedback during the development of this work.
EI was supported in part by the FWF DK VGSCO, grant agreement number W1260-N35. AP and DA acknowledge generous ERC support, via Starting Grant 805223 ScaleML.

\clearpage
{\small
\bibliographystyle{ieee_fullname}
\bibliography{references.bib}
}
\newpage
\appendix
\onecolumn

\addcontentsline{toc}{section}{Appendix} %
\part{Appendix} %
\parttoc %

\setcounter{table}{0}
\renewcommand{\thetable}{\Alph{section}.\arabic{table}}
\setcounter{figure}{0}
\renewcommand{\thefigure}{\Alph{section}.\arabic{figure}}

\section{Full Training Settings}
\label{appendix:training_settings}
In this section we provide the complete details regarding the training setting for our dense and sparse models on CelebA. For all our experiments we used standard random augmentations for CelebA used in \cite{Wang2020TowardsFI}, and we normalized the samples using mean and standard deviation each of 0.5 per channel. Furthermore, we replicated all experiments from five different seeds. We adapted the public implementation for model pruning: \url{https://github.com/IST-DASLab/ACDC} to train with Binary Logistic Loss.

\paragraph{Joint training.} We train the dense model for 100 epochs, using SGD with momentum, with the same hyperparameters (learning rate scheduler, momentum, weight decay, batch size) as the ones used for training ImgageNet in~\cite{kusupati2020soft}, but without label smoothing. Generally, we have noticed that on the held-out CelebA validation set, the dense model tends to overfit after around 40 epochs; therefore, we consider the model with the best validation during training and we use it for our final results on the test set. Likewise, we use the same training hyperparameters for GMP-RI; furthermore, we start pruning from the 10th epoch, using global magnitude pruning on all layers, and increase the sparsity level every 10 epochs, using a standard polynomial schedule~\cite{zhu2017prune}. We finetune the sparse models for the last 20 epochs of training and consider the models with the best validation between epochs 80-100. In the case of GMP-PT models, we use 80 epochs for training, and we increase the sparsity level every 4th epoch, while the final 20 epochs are reserved for finetuning at maximum sparsity. For GMP-PT we use the Adam optimizer, with a fixed learning rate of 0.0001, similar to~\cite{fairgrape}.    

\paragraph{Single label training.} In addition to the joint attribute training, we also train a subset of labels individually. The labels we consider are the following: Bags Under Eyes, Blond, Big nose, Mustache, Oval Face, Receding Hairline, and Smiling. All single label experiments are trained for 20 epochs to avoid overfitting. The dense models were trained using SGD with momentum, with initial learning rate 0.1, batch size 256, momentum value 0.9 and weight decay 0.0001; additionally, we used a cosine annleaning learning rate scheduler. The GMP-RI models were trained using SGD with momentum value 0.9, weight decay 0.0001 and fixed learning rate of 0.1; models were pruned starting from the third epoch, with a gradual increase in sparsity every epoch following a polynomial schedule~\cite{zhu2017prune}, while the final 4 epochs were reserved for finetuning. %

\clearpage
\section{Full Override Results for Jointly-trained ResNet18 Models}
\label{appendix:celeba_rn18_overrides}

\begin{figure}[h]
\centering
\includegraphics[width=0.8\textwidth]{figures/Male_override_improvements.pdf}
\includegraphics[width=0.8\textwidth]{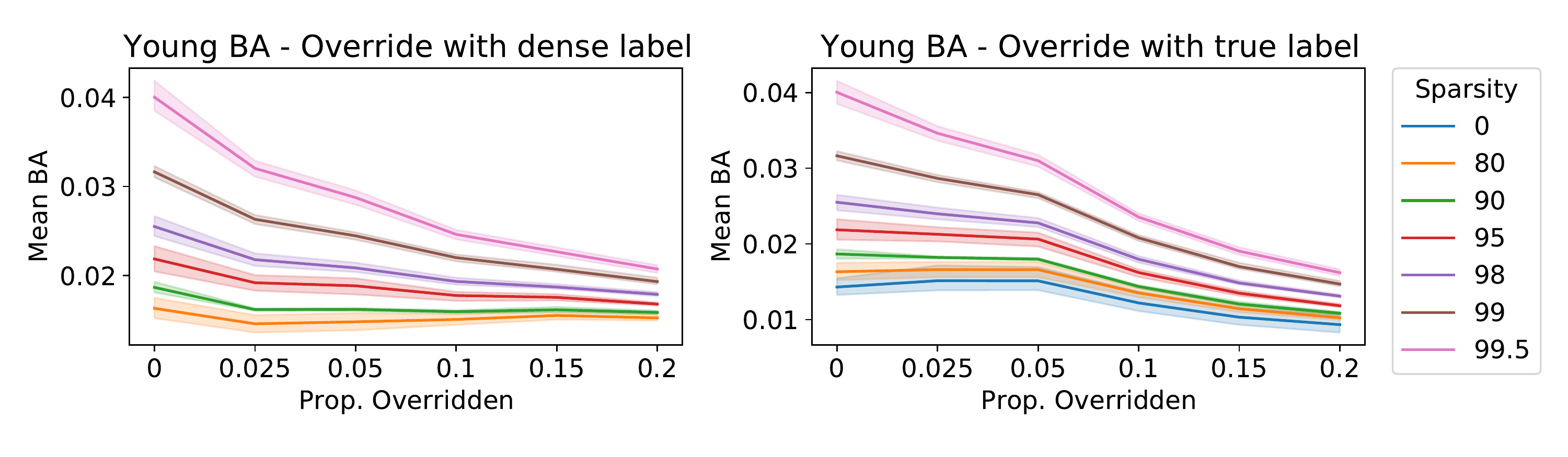}
\includegraphics[width=0.8\textwidth]{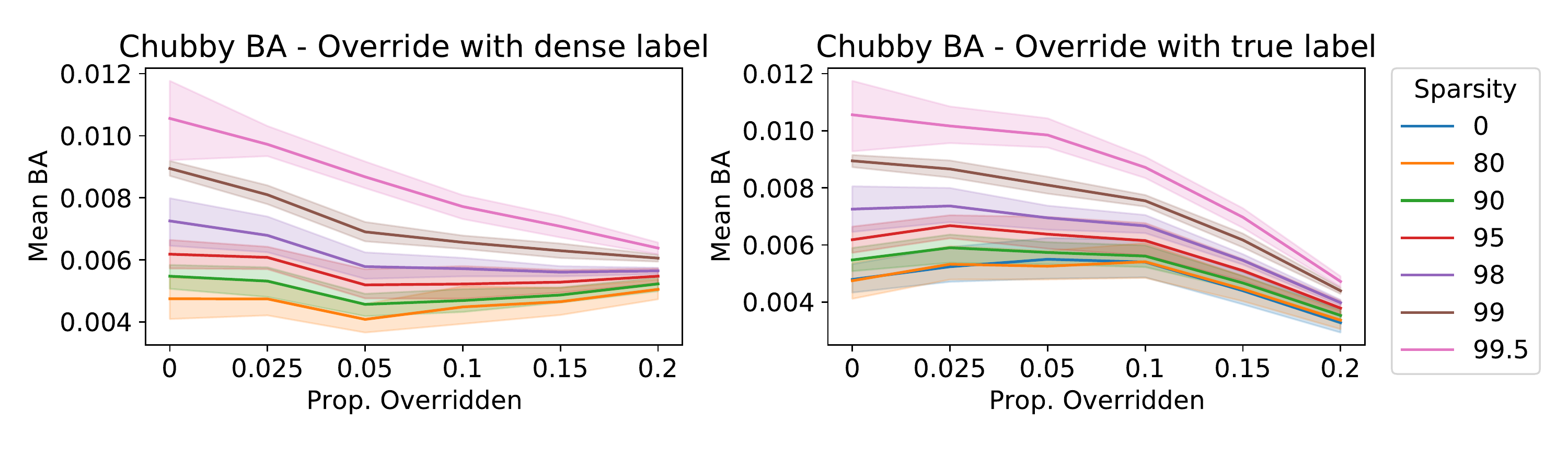}
\includegraphics[width=0.8\textwidth]{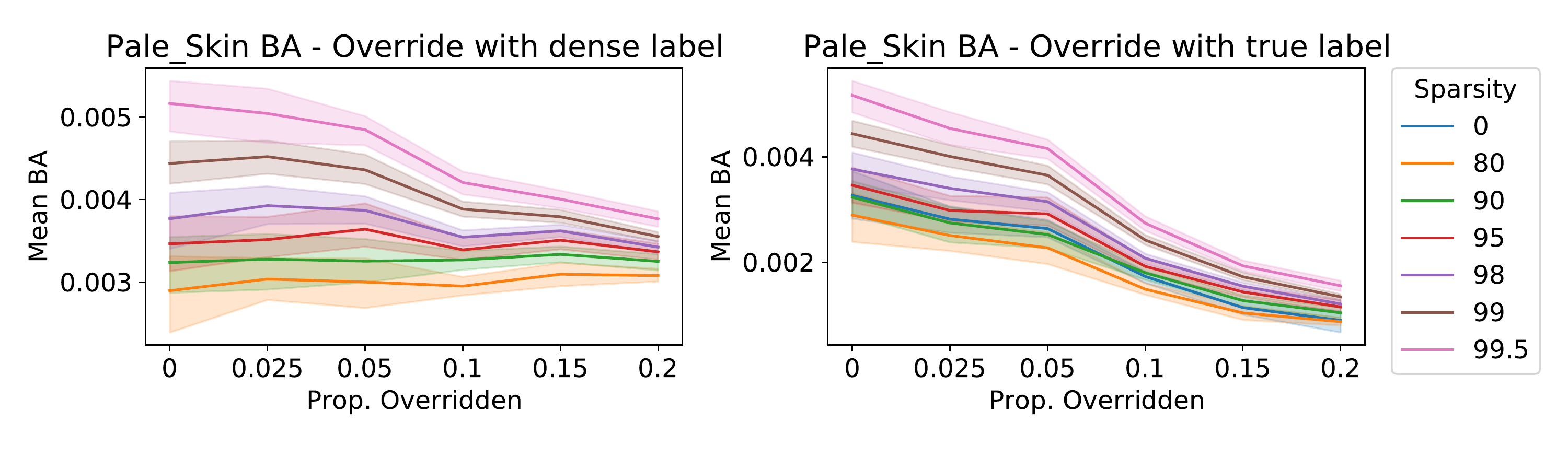}
    \caption{[CelebA / ResNet18 / GMP-RI] Effect of label overrides on Bias Amplification. In all cases, overrides are prioritized by dense model uncertainty.}
    \label{fig:overrides_rn18_RI}
\end{figure}

In this section, we present the full data for the impact on Bias Amplification of selectively overriding model predictions with dense predictions (in the case of sparse models) or correct labels. In all cases, the overridden samples are prioritized by the uncertainty of the \emph{dense} model on that attribute. Further, only predictions for attributes that show positive bias amplification in the dense case are overridden. The results are shown in Figure~\ref{fig:overrides_rn18_RI}. We observe that in general, overrides using dense model predictions are effective in the case of very sparse (99\%-99.5\% sparse) models, but their effectiveness decreases for less sparse models. This is consistent with our observation that less sparse models show less bias amplification relative to dense even without any interventions. Further, we observe that even for categories where the BA is relatively low (Chubby and Pale Skin), overrides are still effective at further reducing relative bias amplification at high sparsity. Overriding with the true label reduces bias amplification throughout.

\clearpage
\section{Full results for Singly-trained CelebA models on ResNet18}
\label{appendix:missing_data}

\begin{figure}[ht]
\centering
\begin{tabular}{ccccccc}
\includegraphics[width=0.12\textwidth]{figures/celeba_rn18_oval-face_acc_single.pdf} &
\includegraphics[width=0.12\textwidth]{figures/celeba_rn18_big-nose_acc_single.pdf} &
\includegraphics[width=0.12\textwidth]{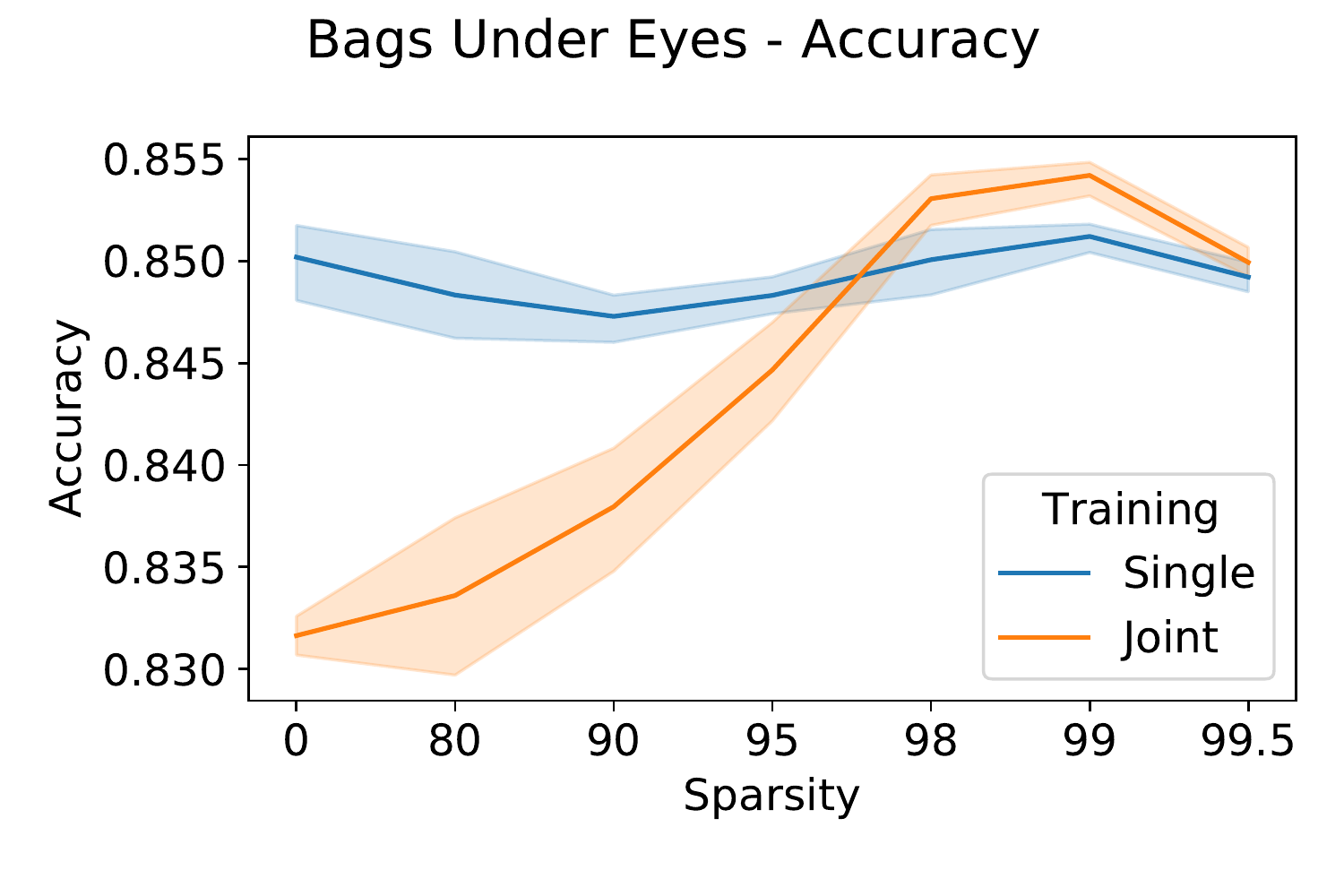} &
\includegraphics[width=0.12\textwidth]{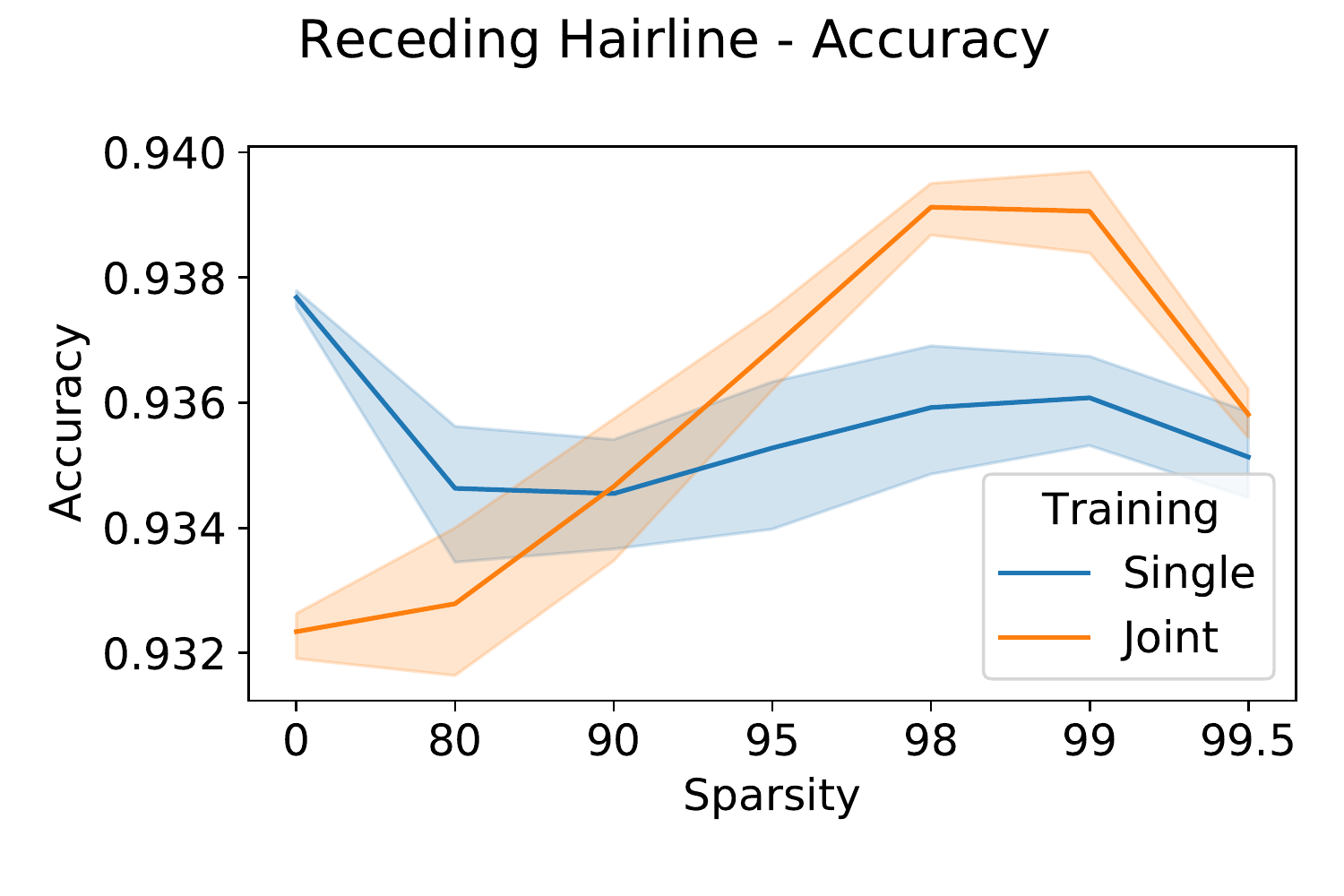} &
\includegraphics[width=0.12\textwidth]{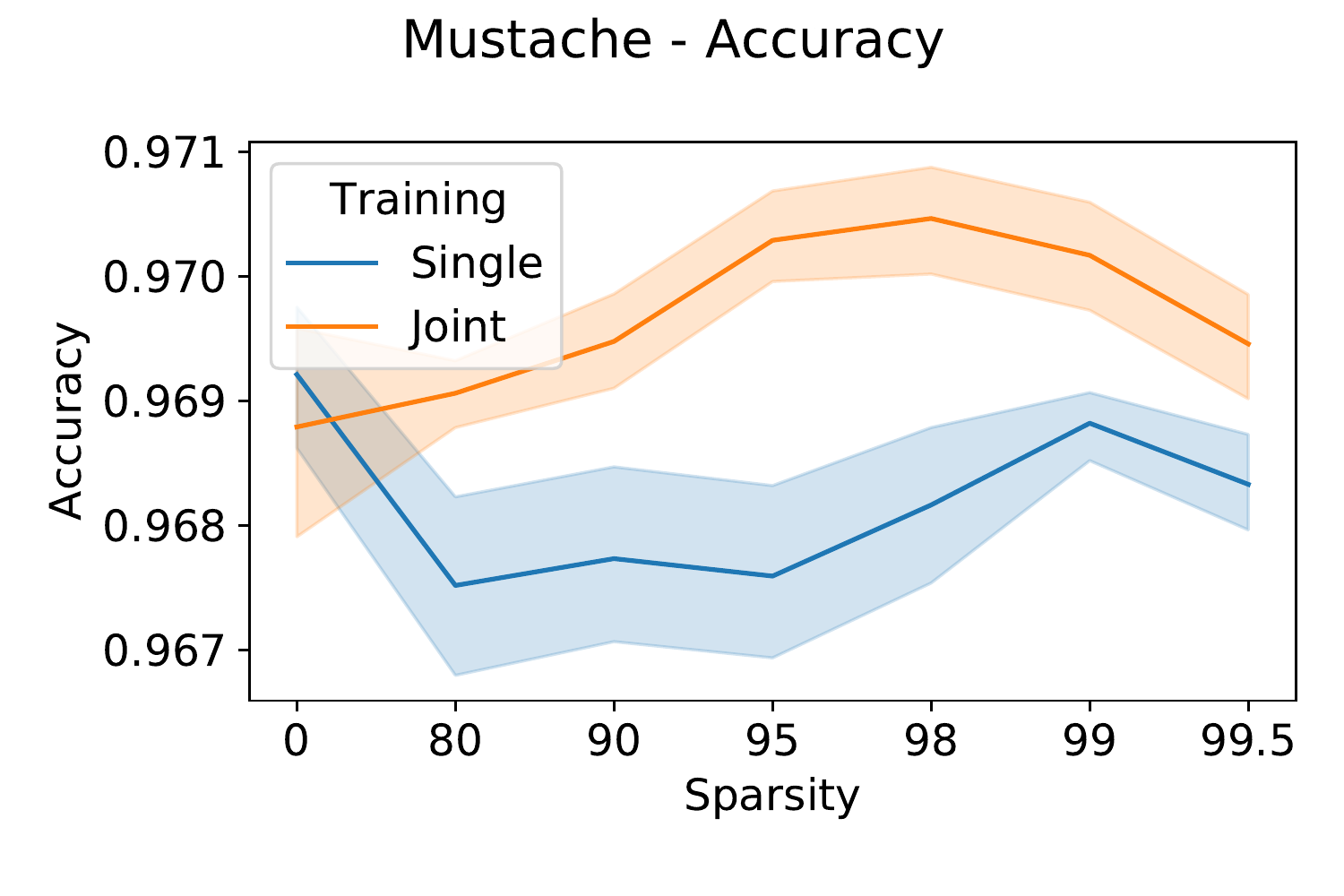} &
\includegraphics[width=0.12\textwidth]{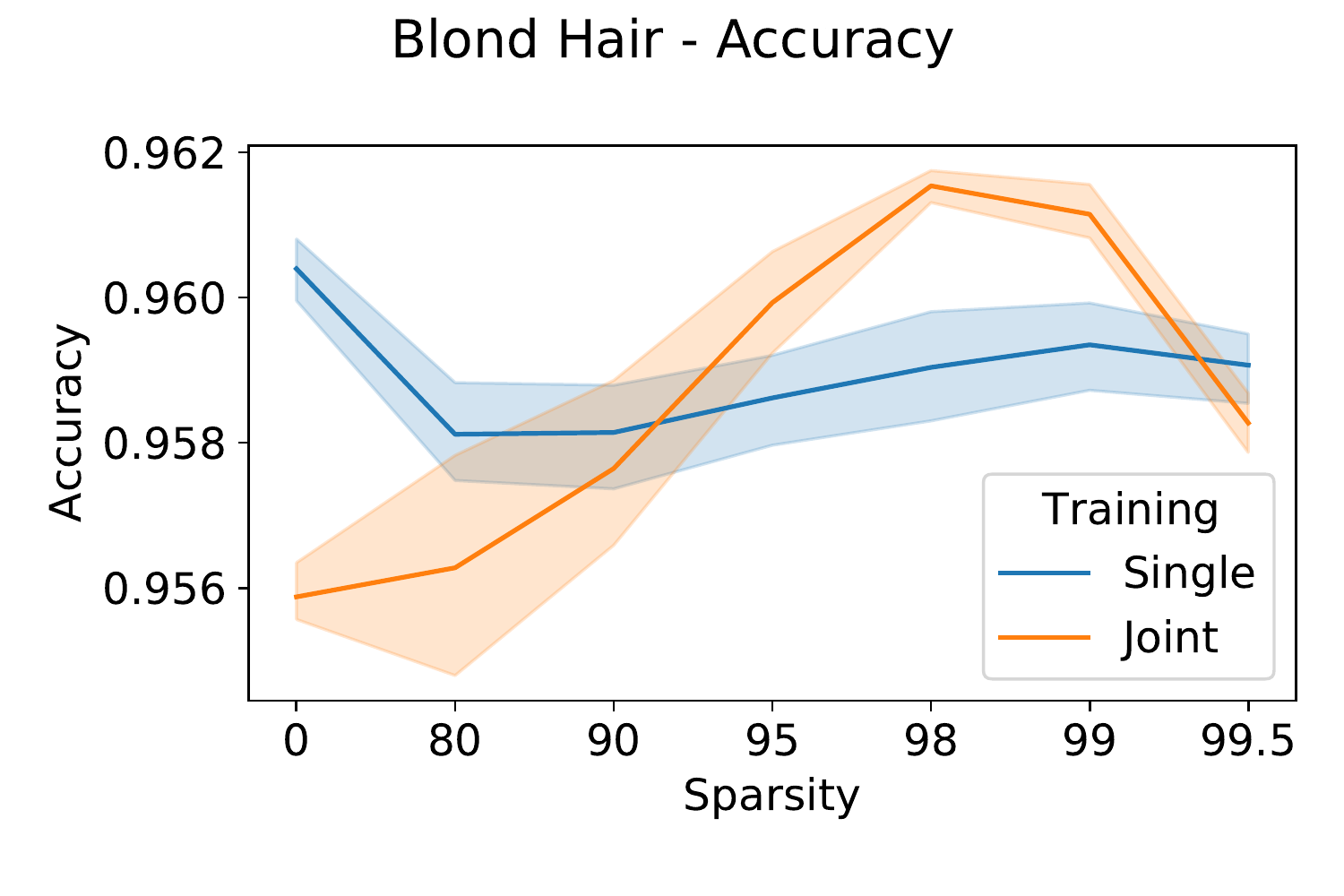} &
\includegraphics[width=0.12\textwidth]{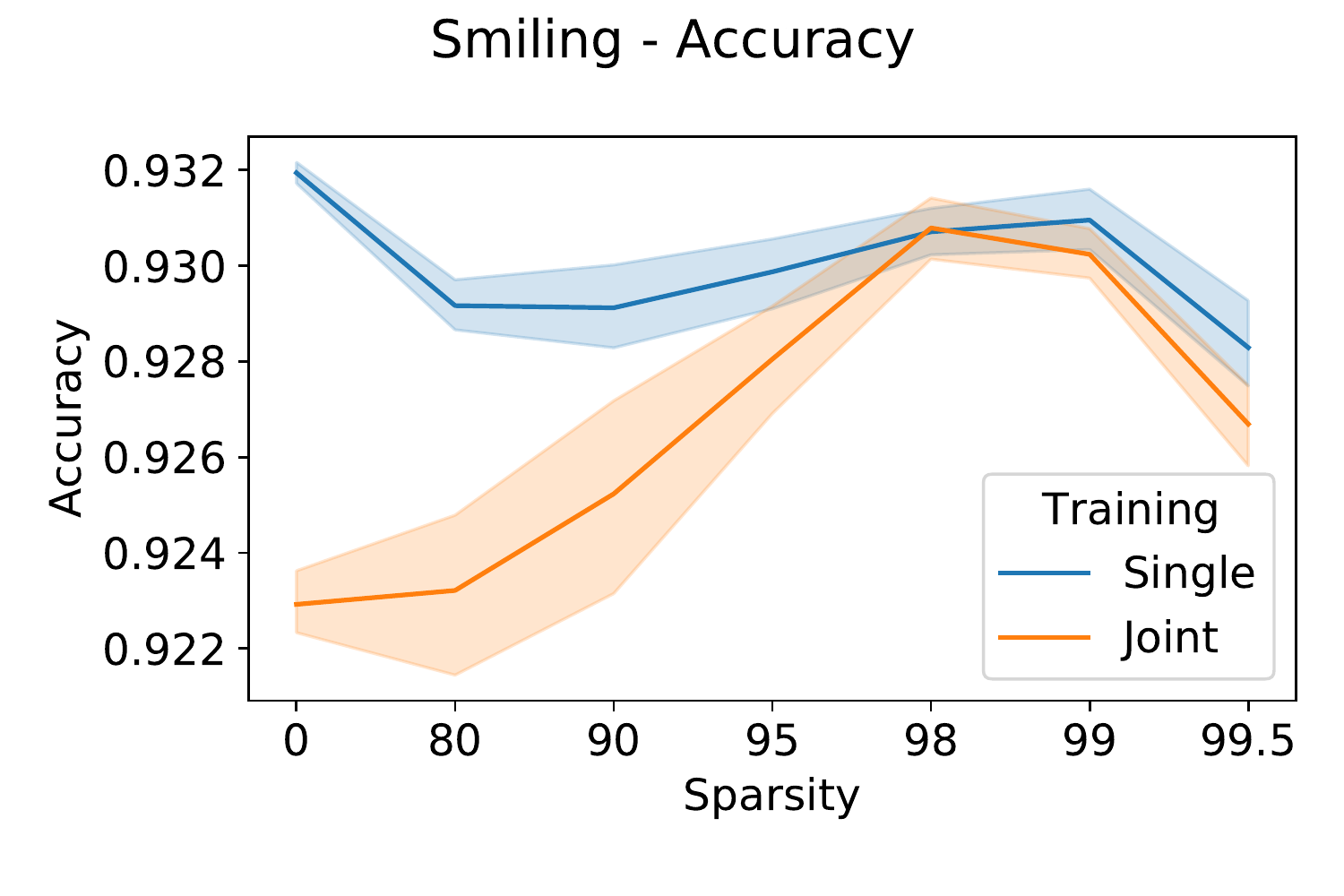}
\\
  \includegraphics[width=0.12\textwidth]{figures/celeba_rn18_oval-face_uncertainty_single.pdf} &
      \includegraphics[width=0.12\textwidth]{figures/celeba_rn18_big-nose_uncertainty_single.pdf} &
    \includegraphics[width=0.12\textwidth]{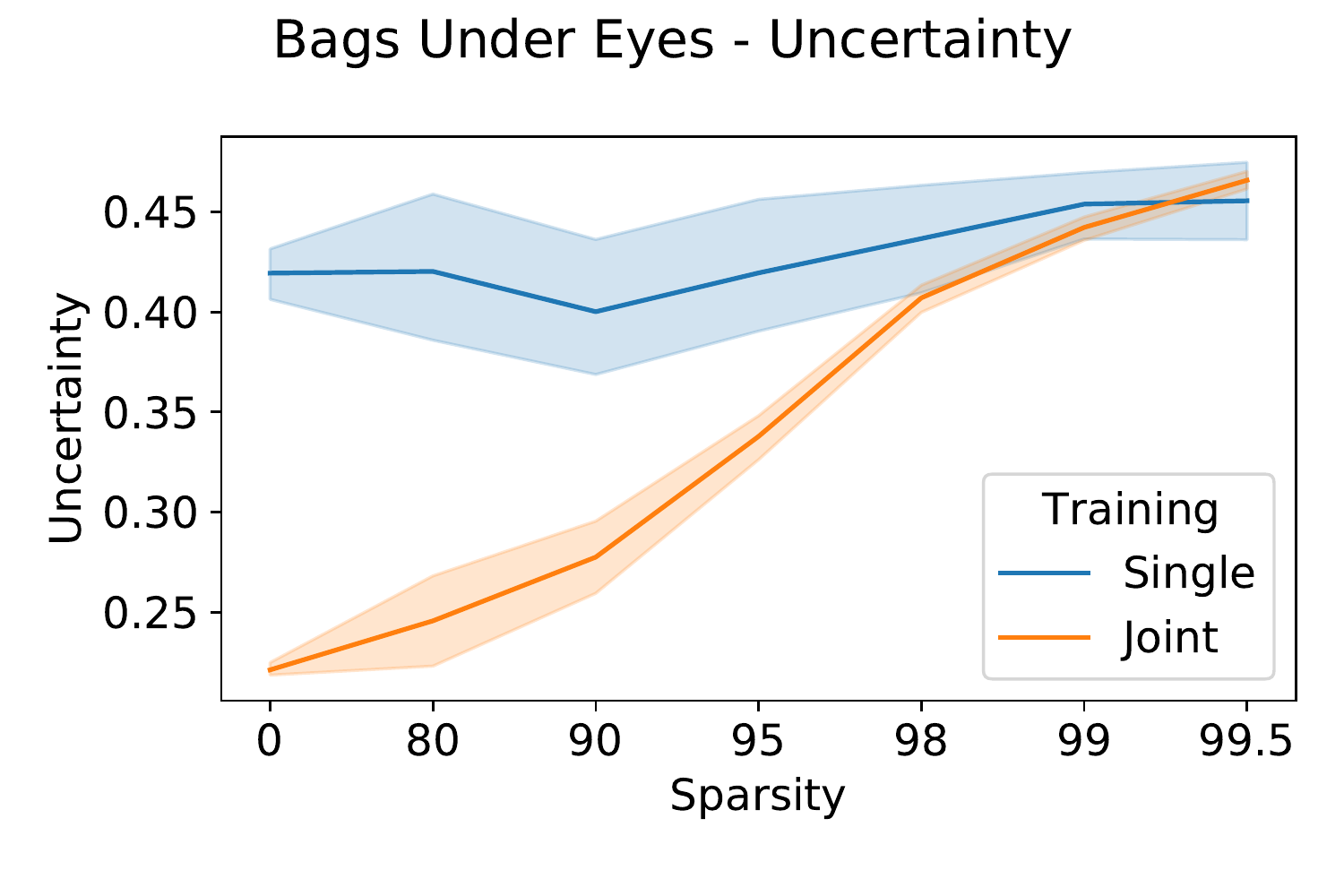} &
      \includegraphics[width=0.12\textwidth]{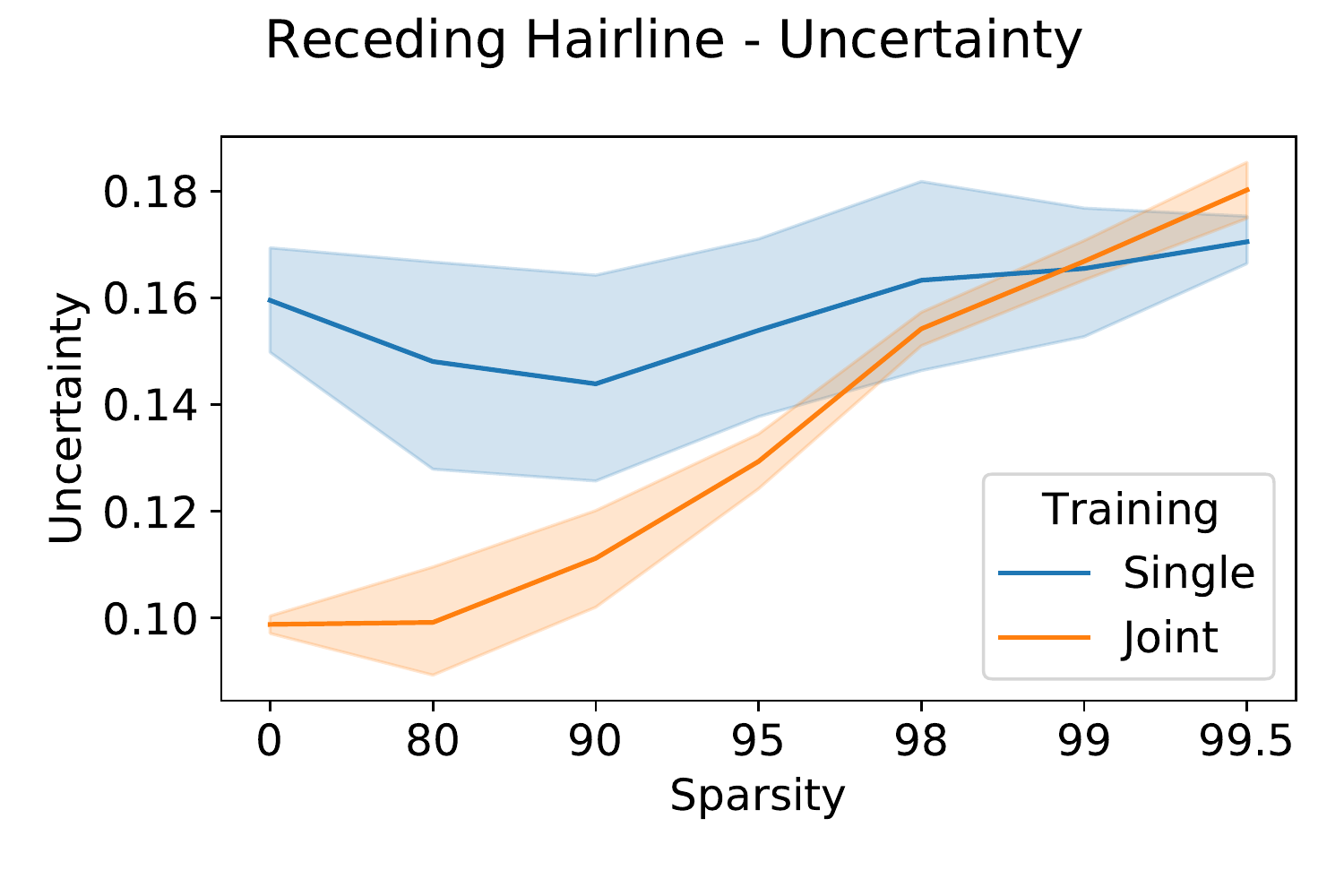} &
      \includegraphics[width=0.12\textwidth]{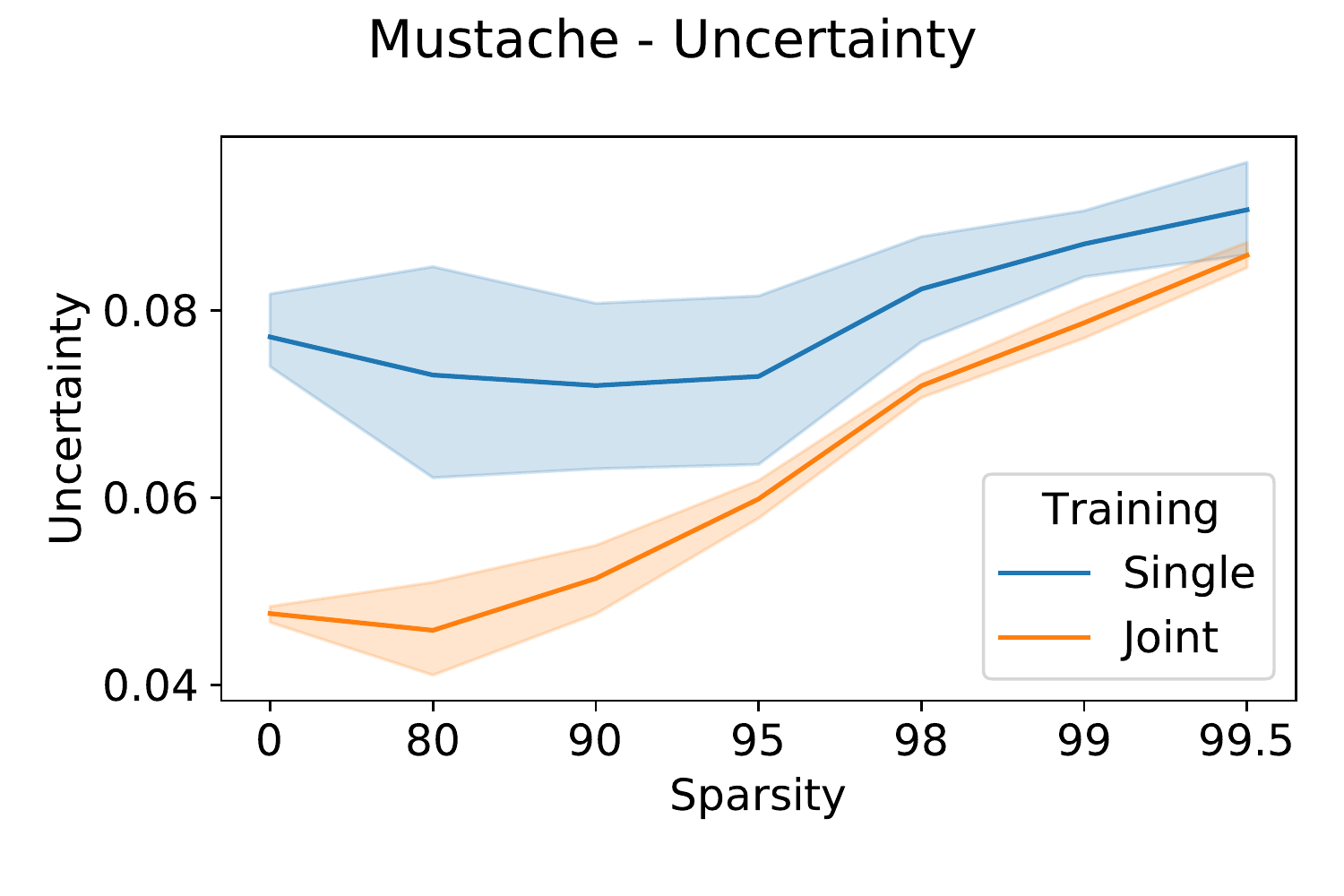} &
      \includegraphics[width=0.12\textwidth]{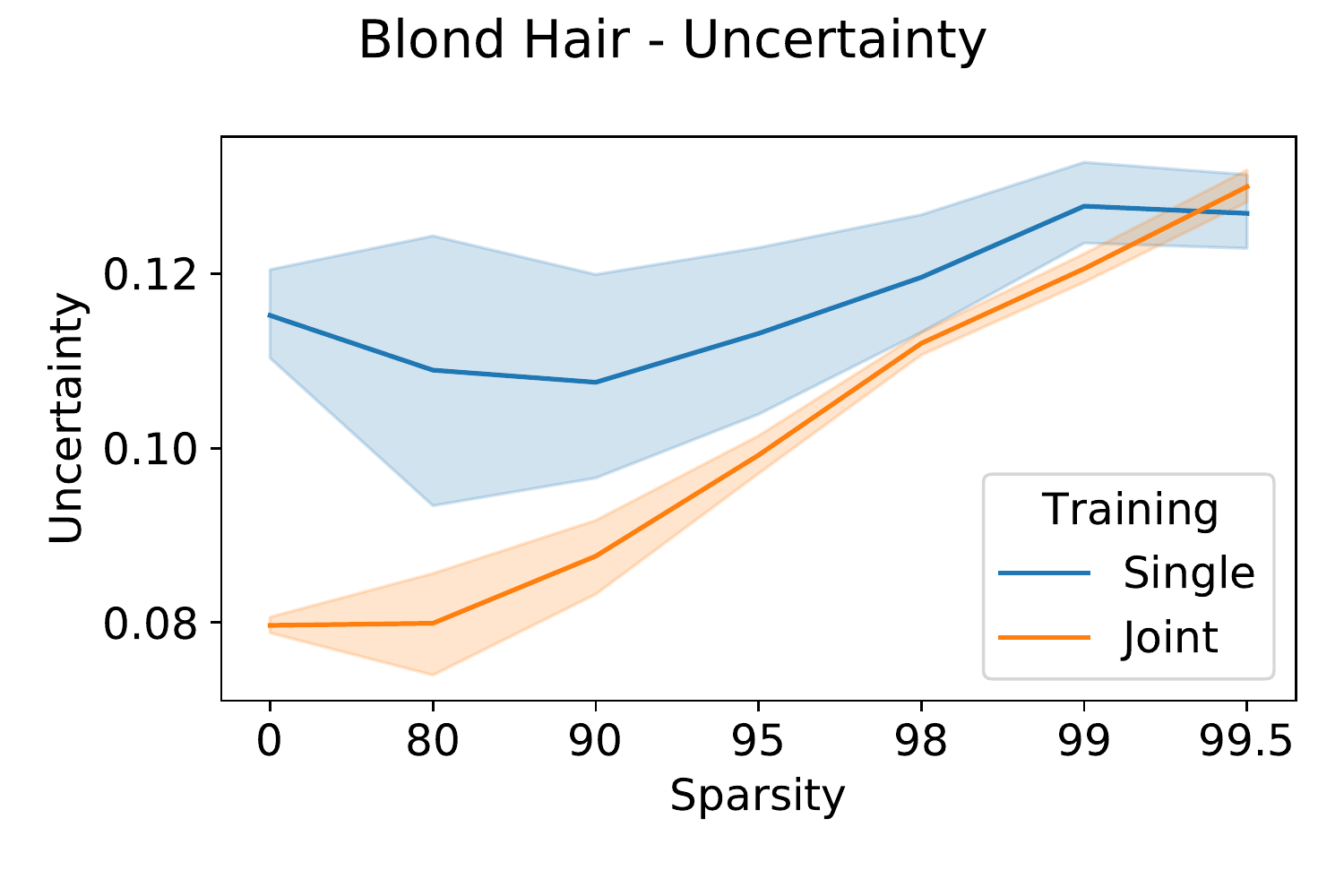} &
      \includegraphics[width=0.12\textwidth]{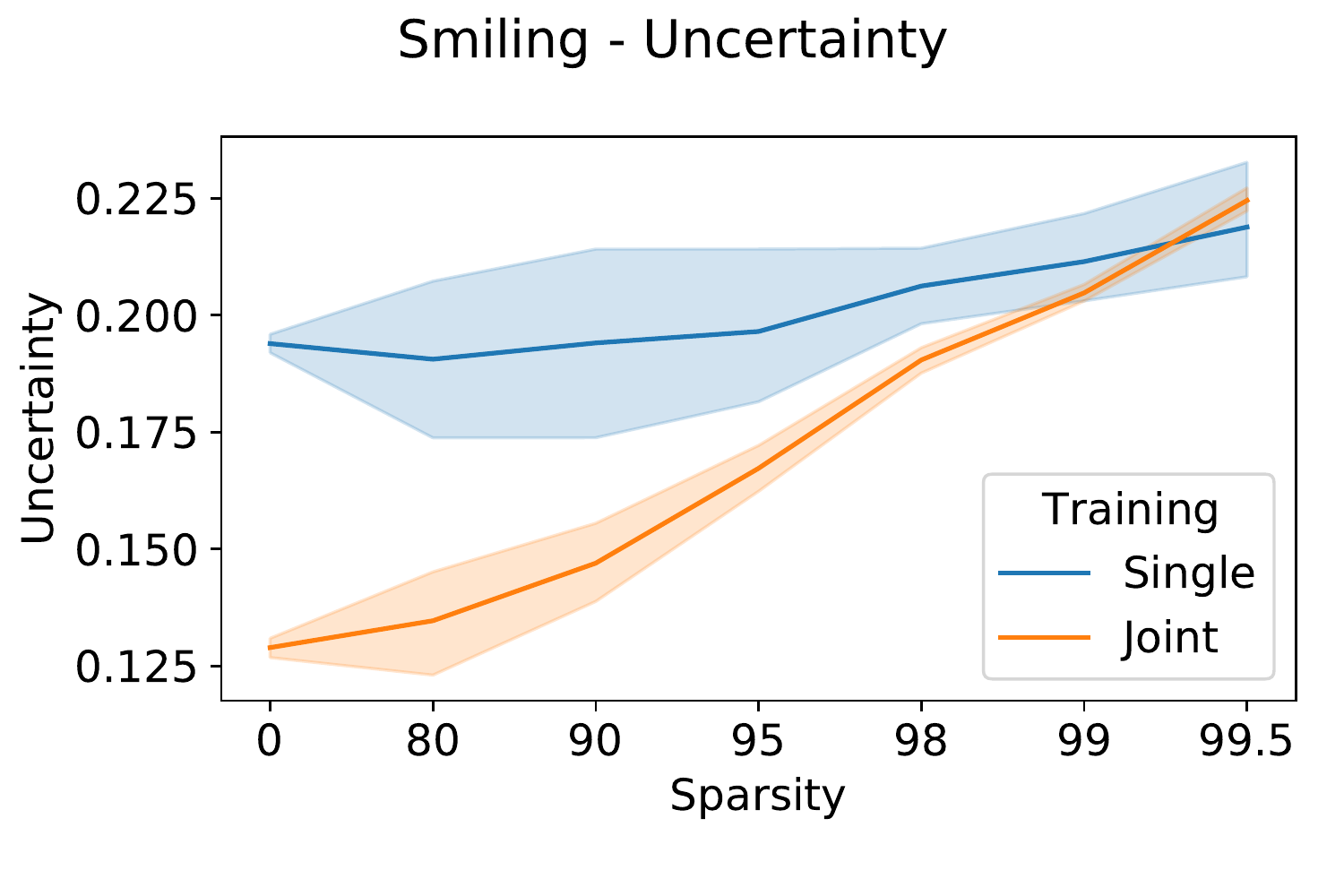}
      \\
  \includegraphics[width=0.12\textwidth]{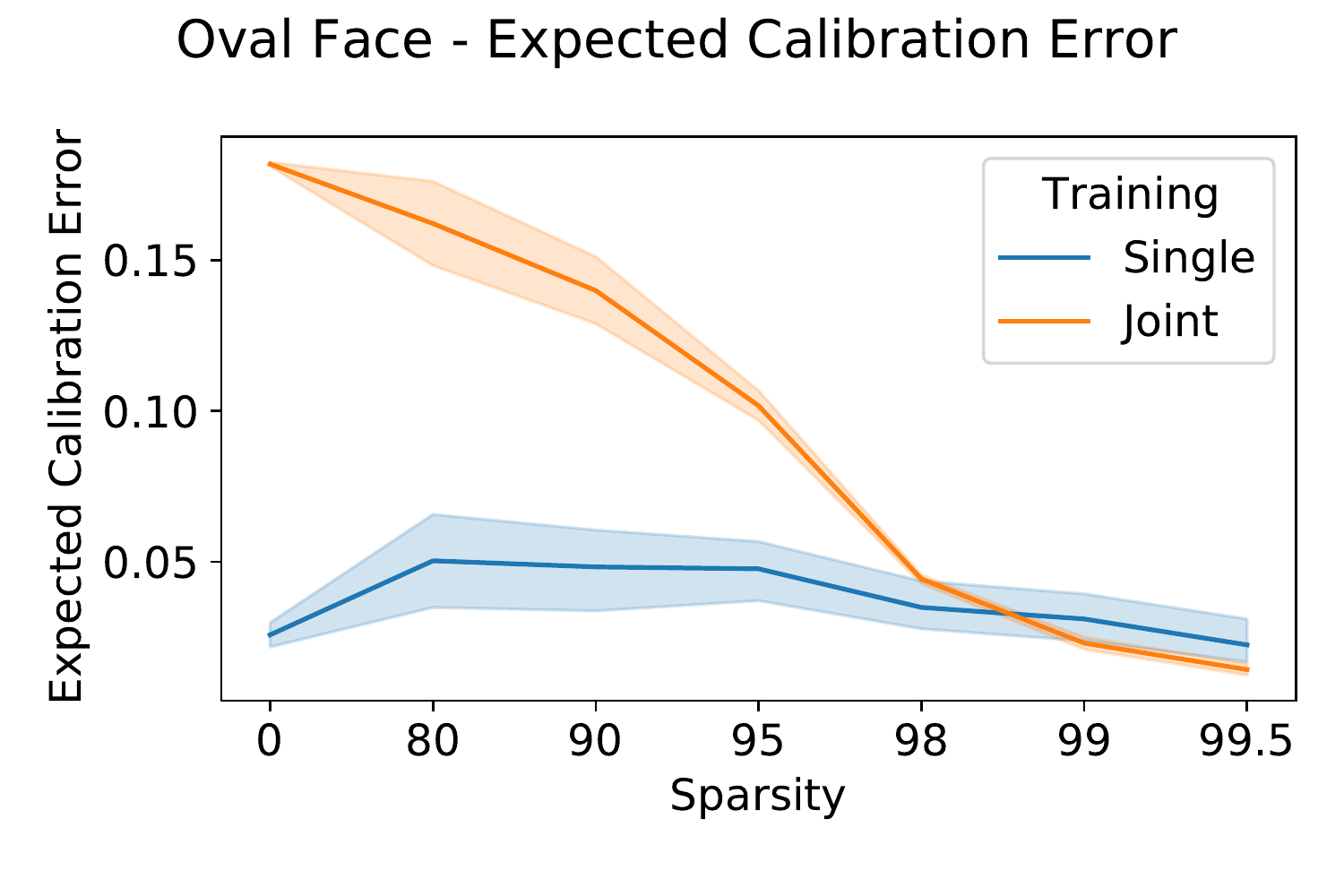} &
      \includegraphics[width=0.12\textwidth]{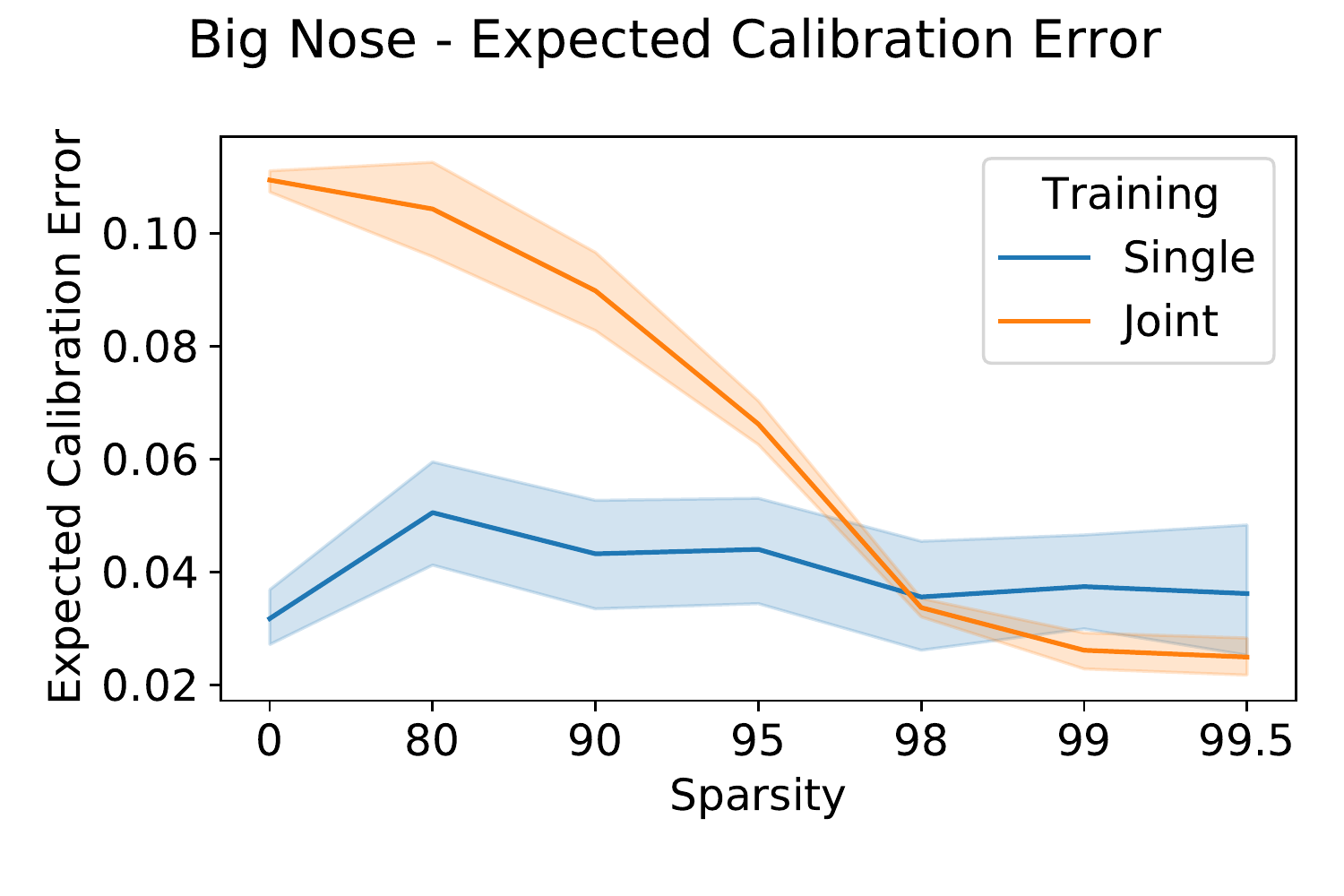} &
    \includegraphics[width=0.12\textwidth]{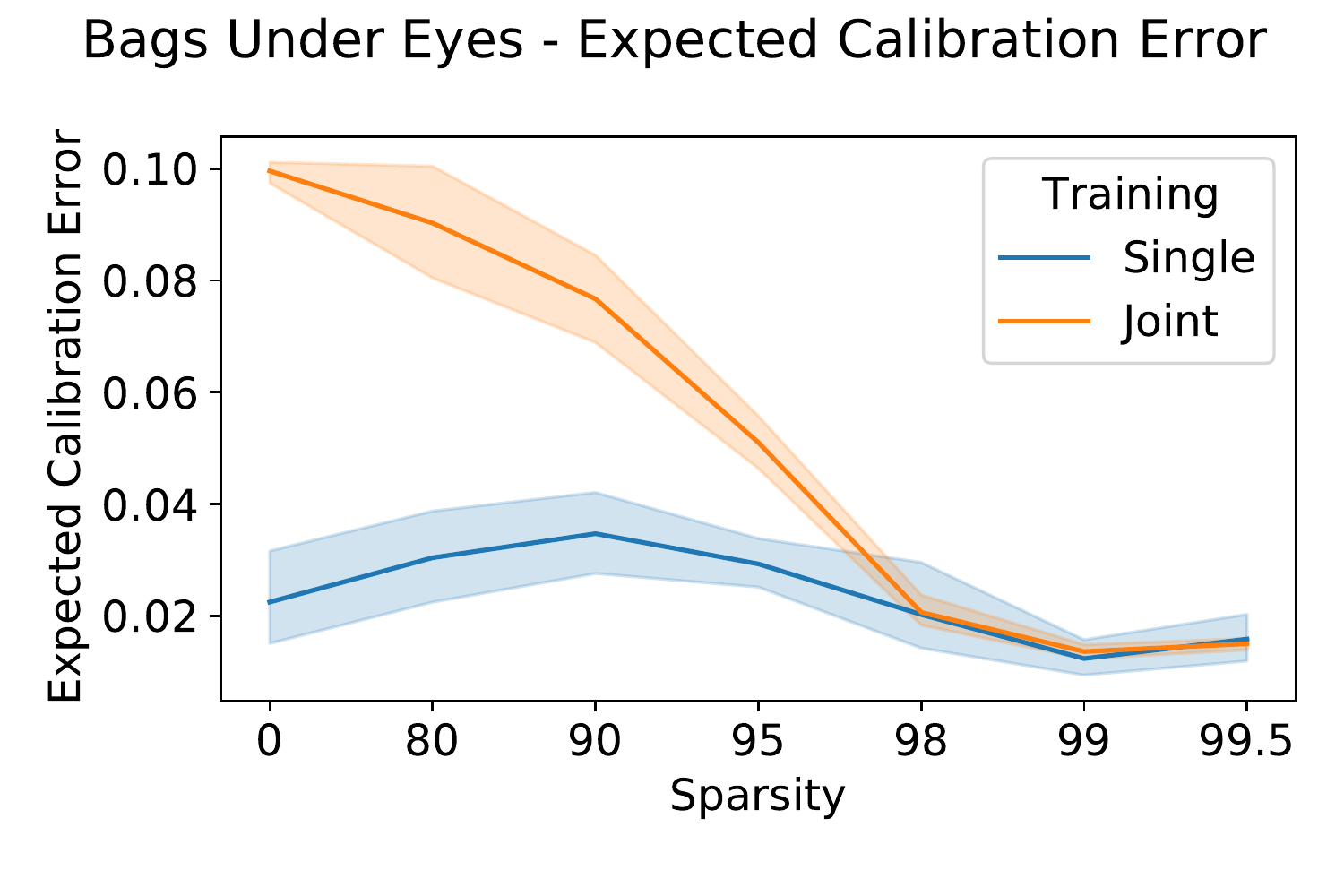} &
      \includegraphics[width=0.12\textwidth]{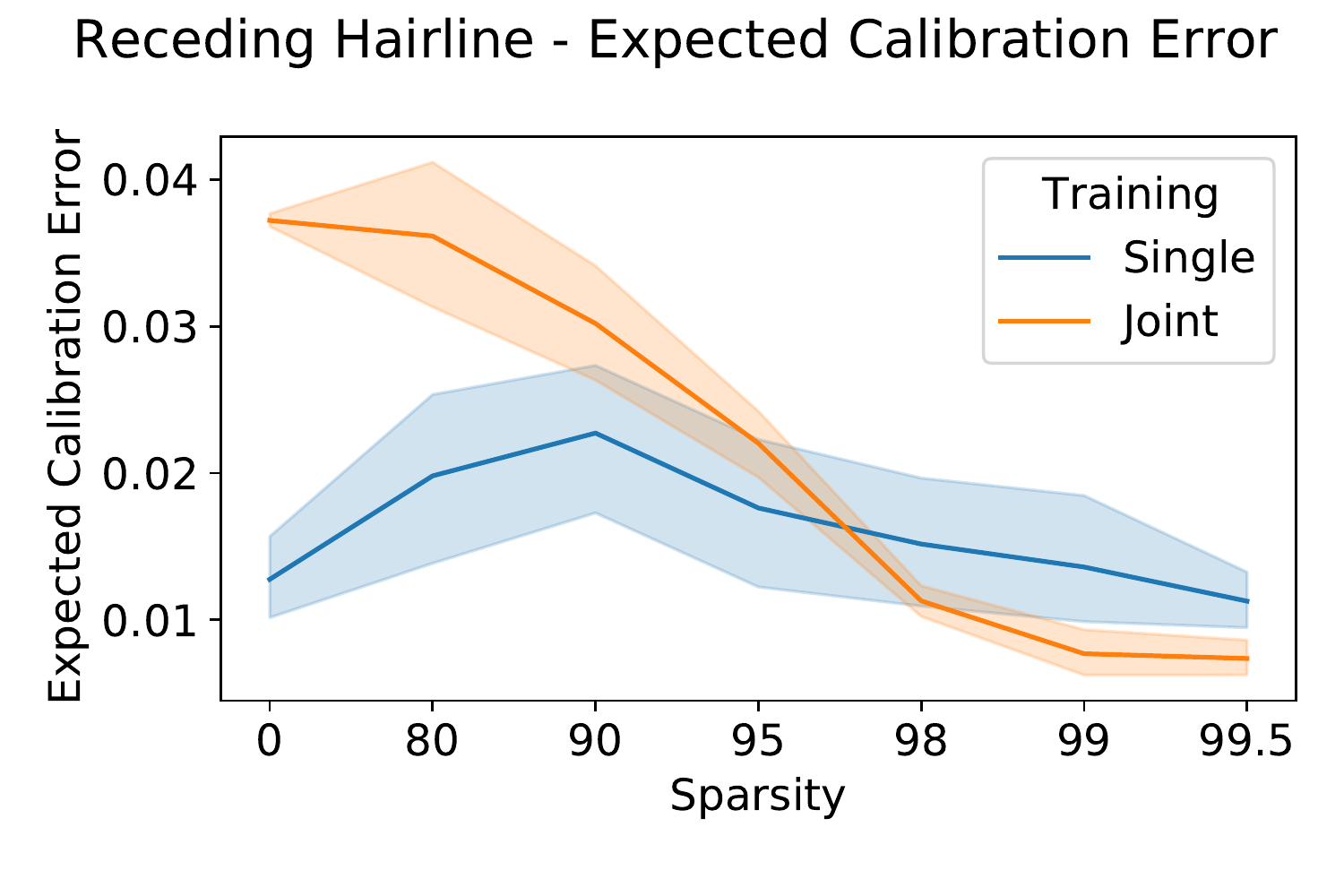} &
      \includegraphics[width=0.12\textwidth]{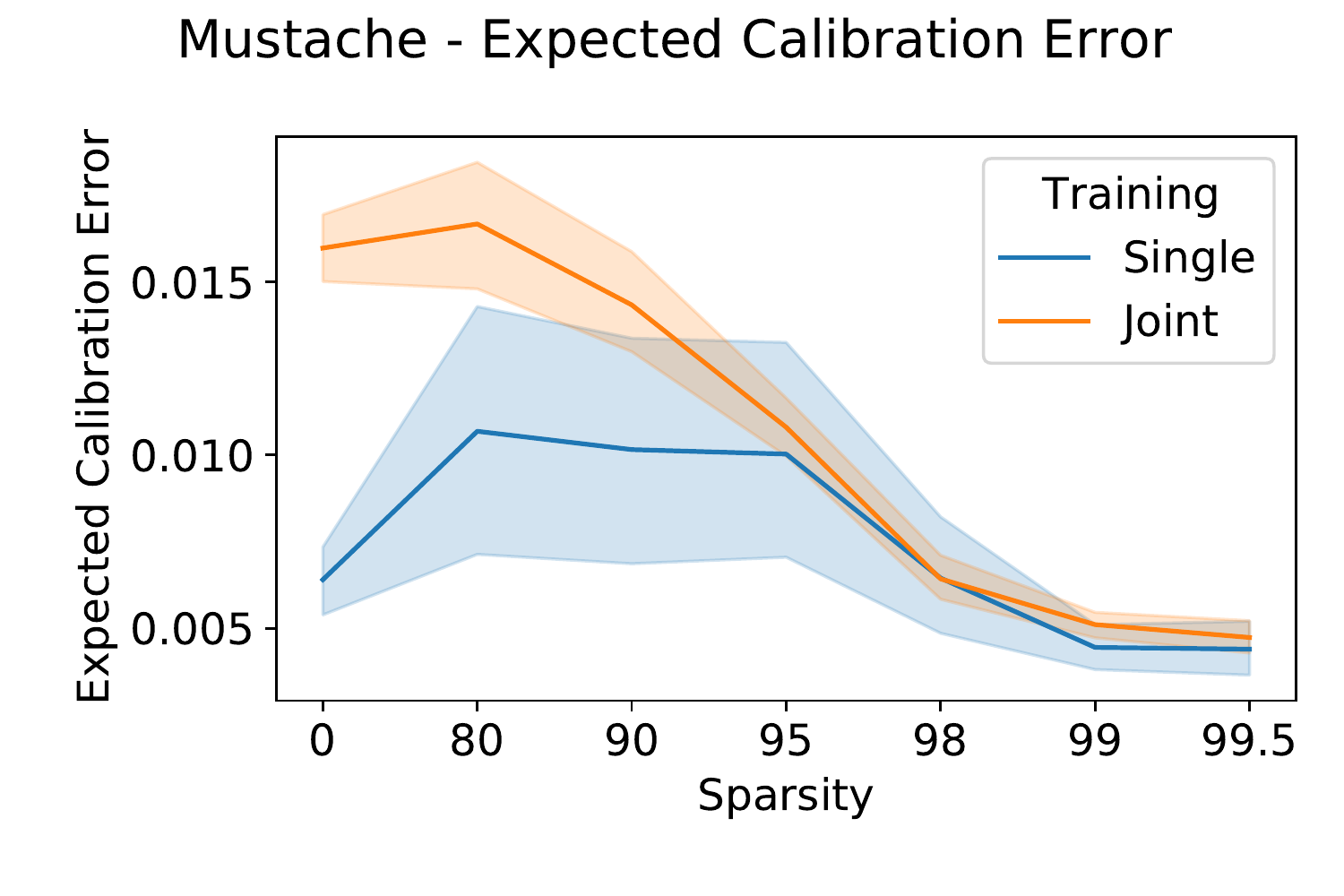} &
      \includegraphics[width=0.12\textwidth]{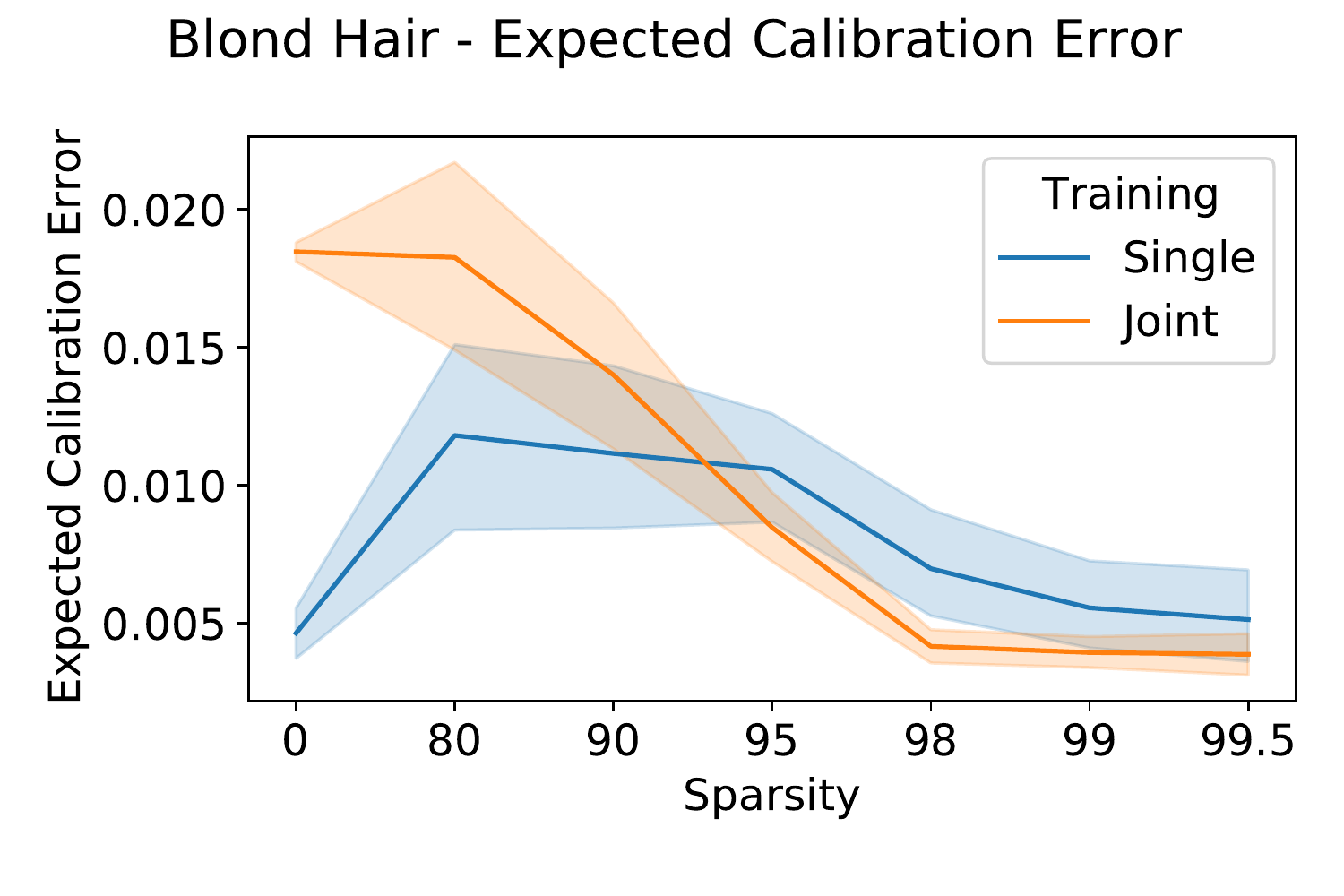} &
      \includegraphics[width=0.12\textwidth]{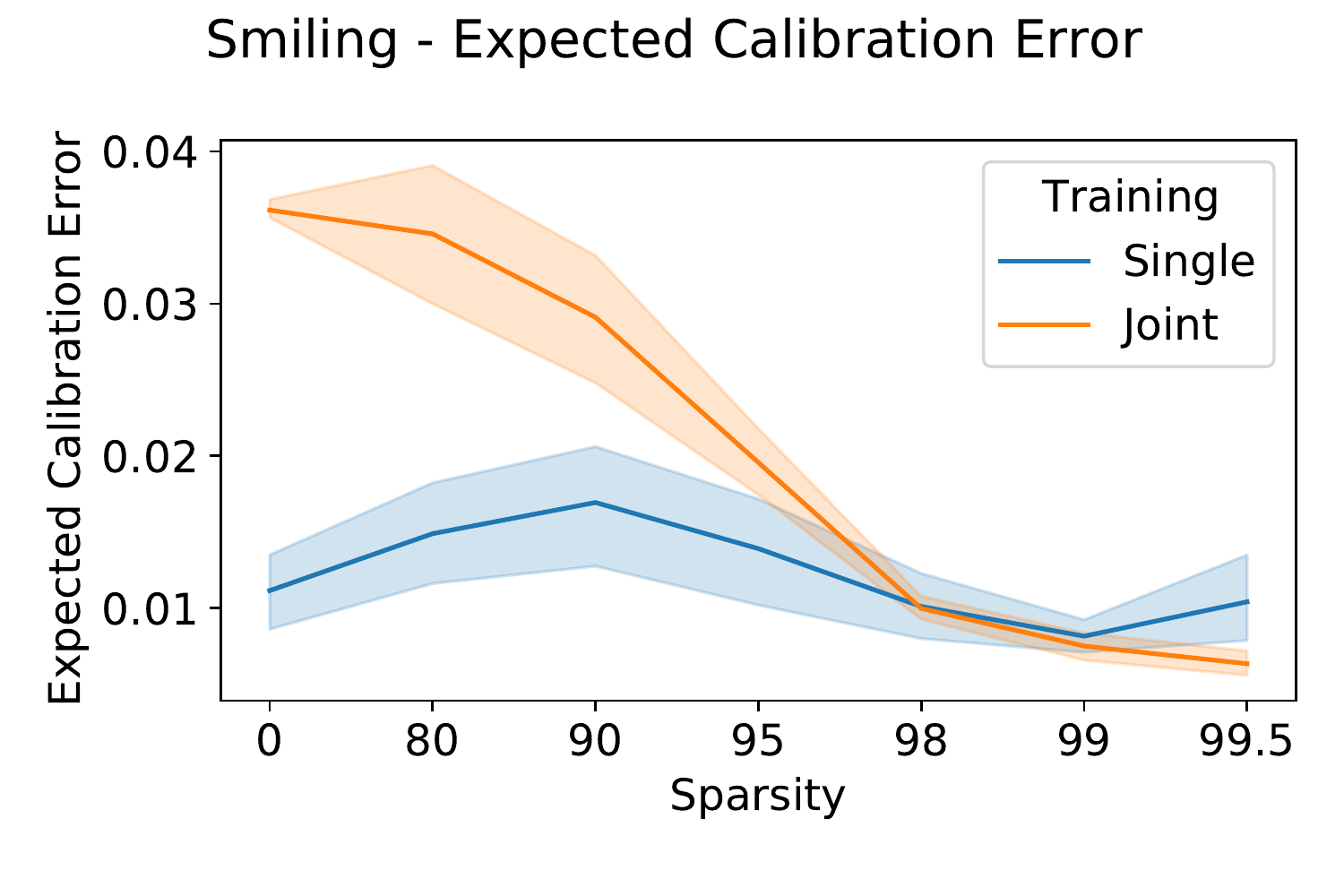}
      \\
    \includegraphics[width=0.12\textwidth]
  {figures/celeba_rn18_oval-face_rare-val-underpredict_single.pdf} &
  \includegraphics[width=0.12\textwidth]
  {figures/celeba_rn18_big-nose_rare-val-underpredict_single.pdf} &
  \includegraphics[width=0.12\textwidth]
  {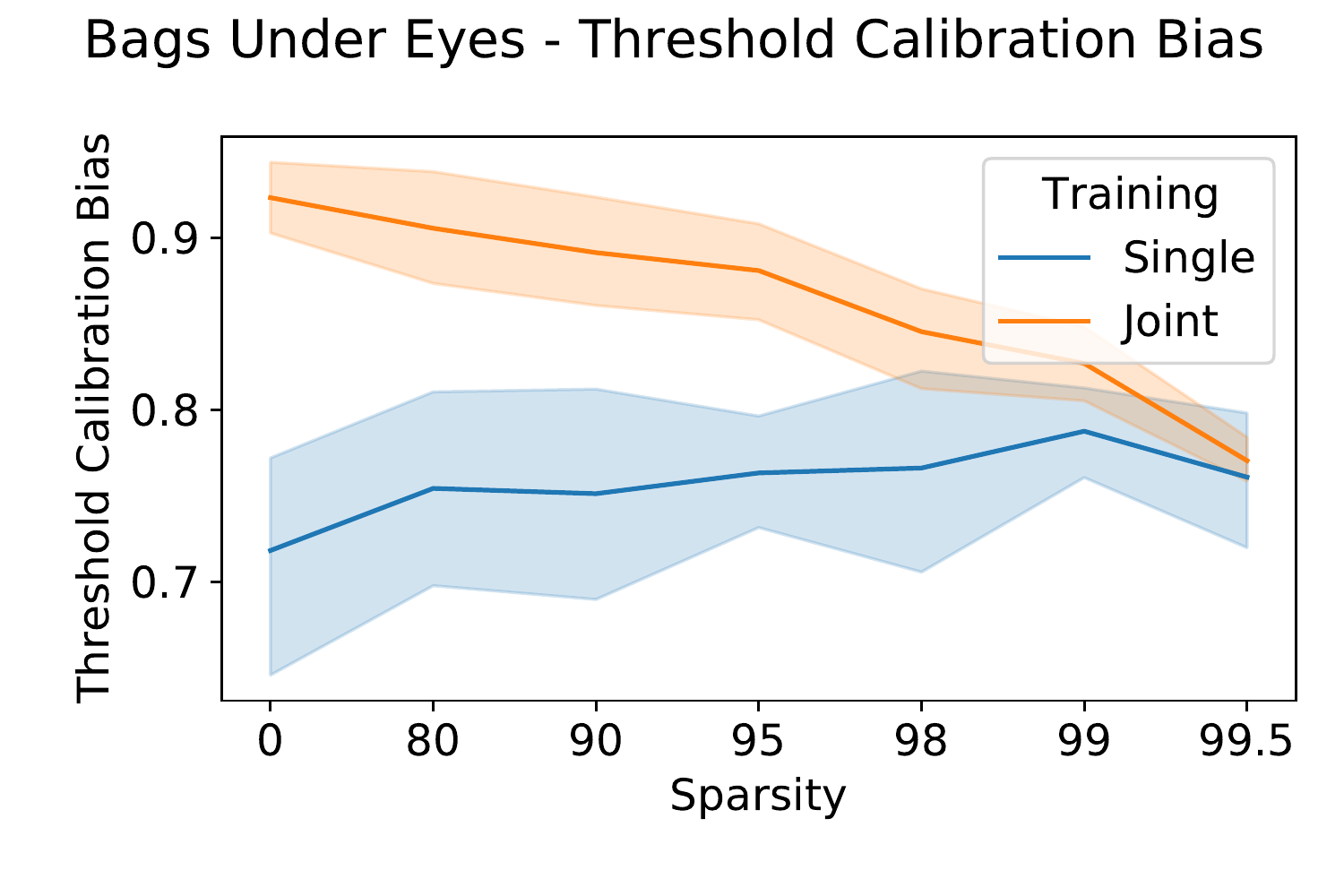} &
  \includegraphics[width=0.12\textwidth]
  {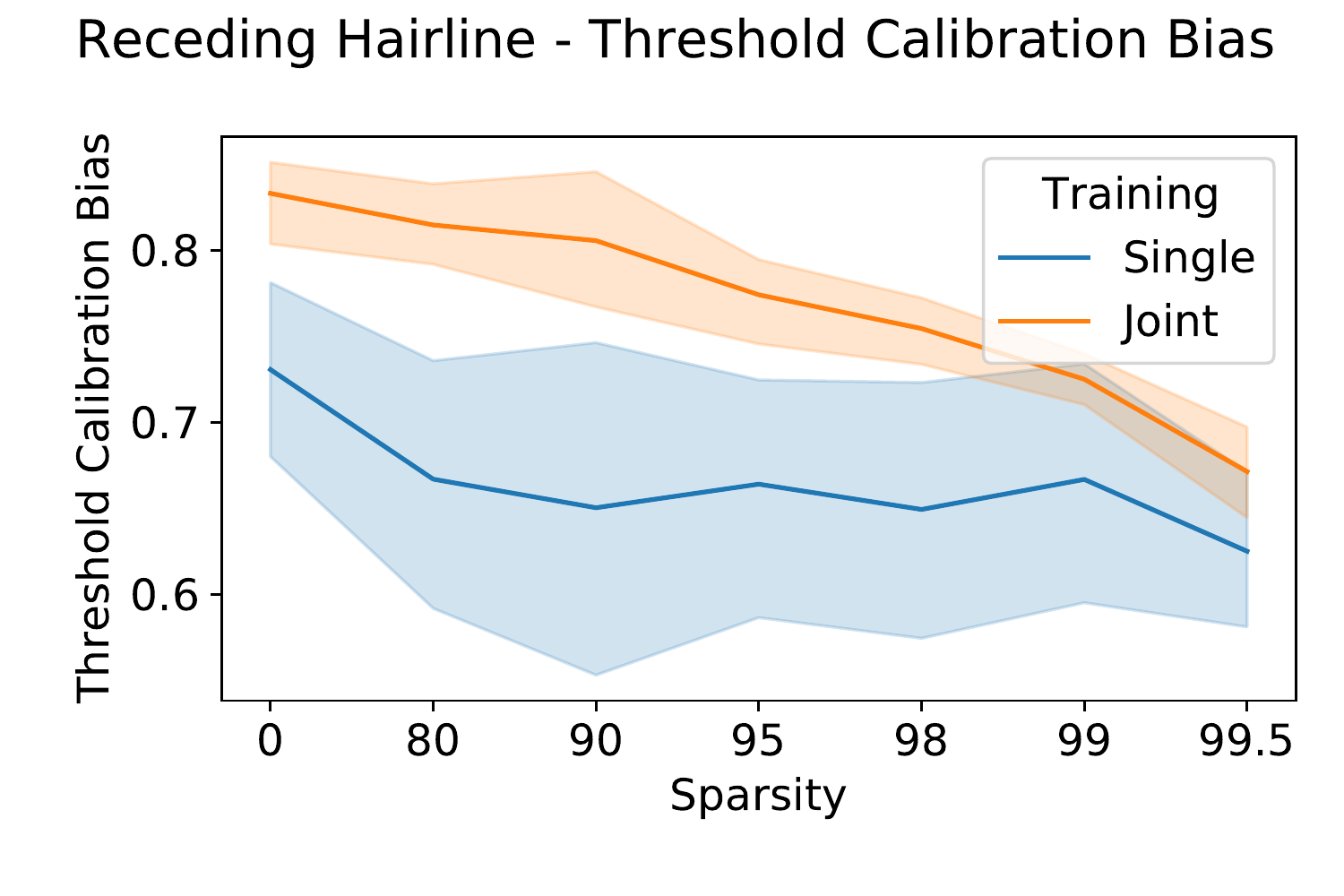} &
  \includegraphics[width=0.12\textwidth]
  {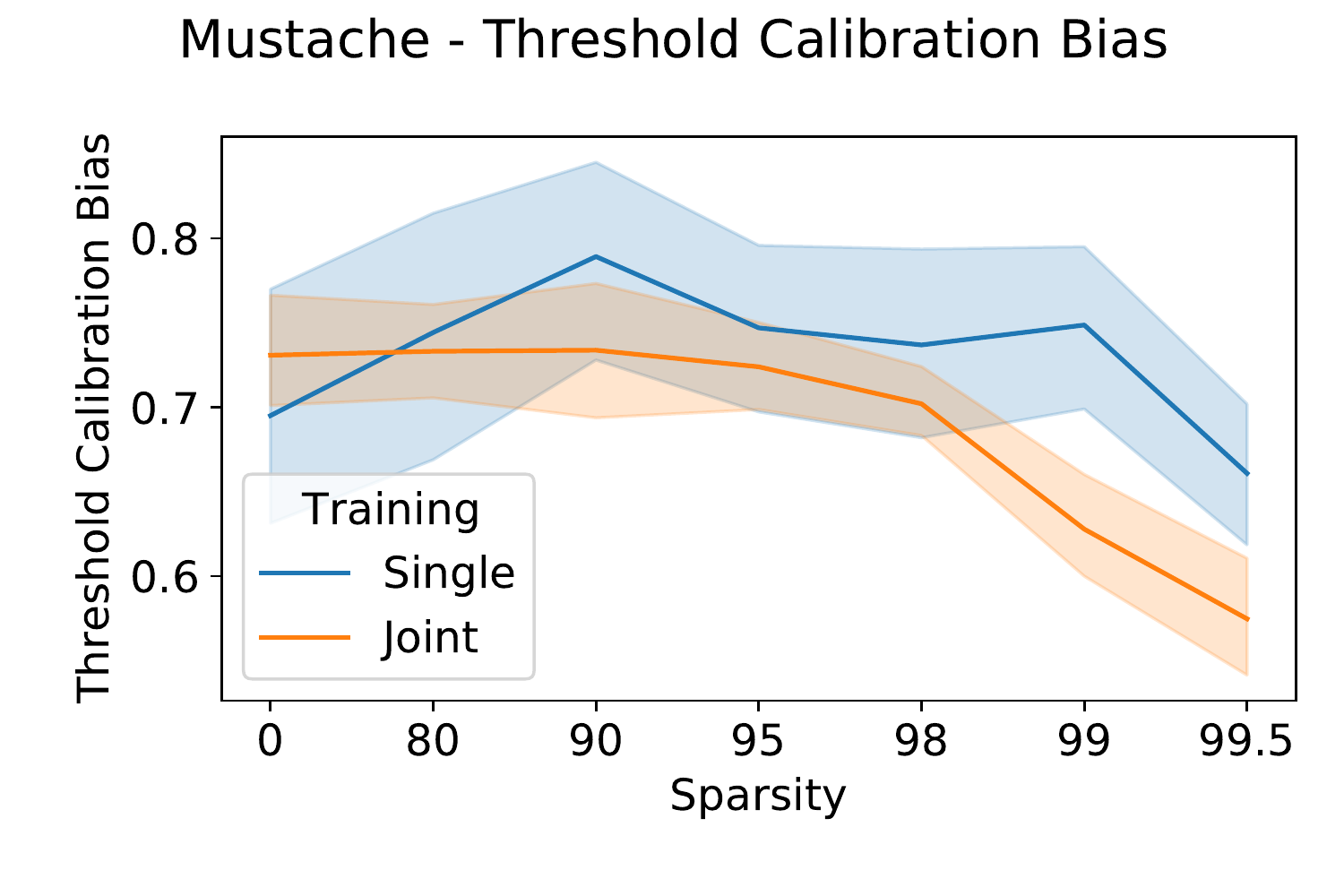} &
  \includegraphics[width=0.12\textwidth]
  {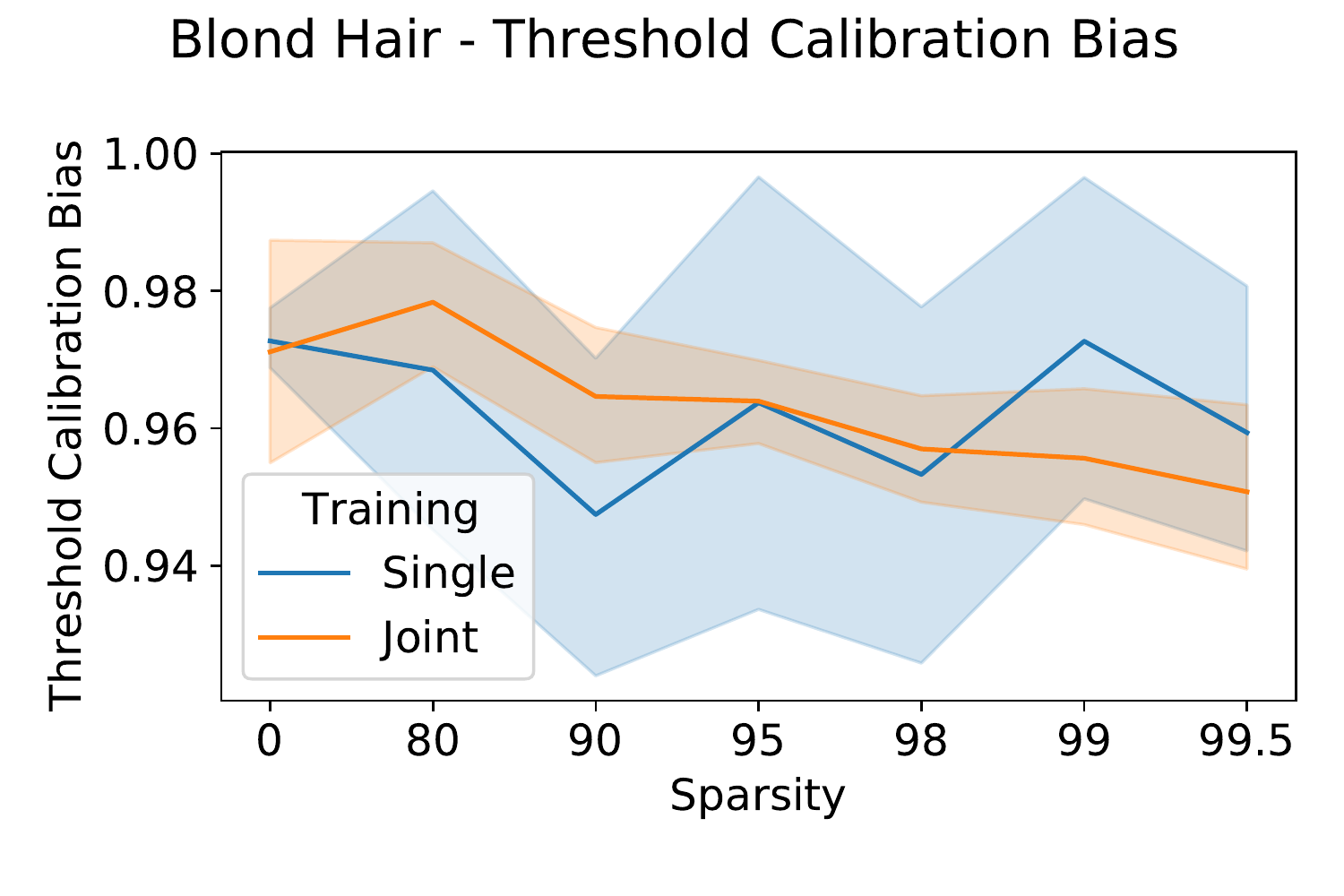} &
  \includegraphics[width=0.12\textwidth]
  {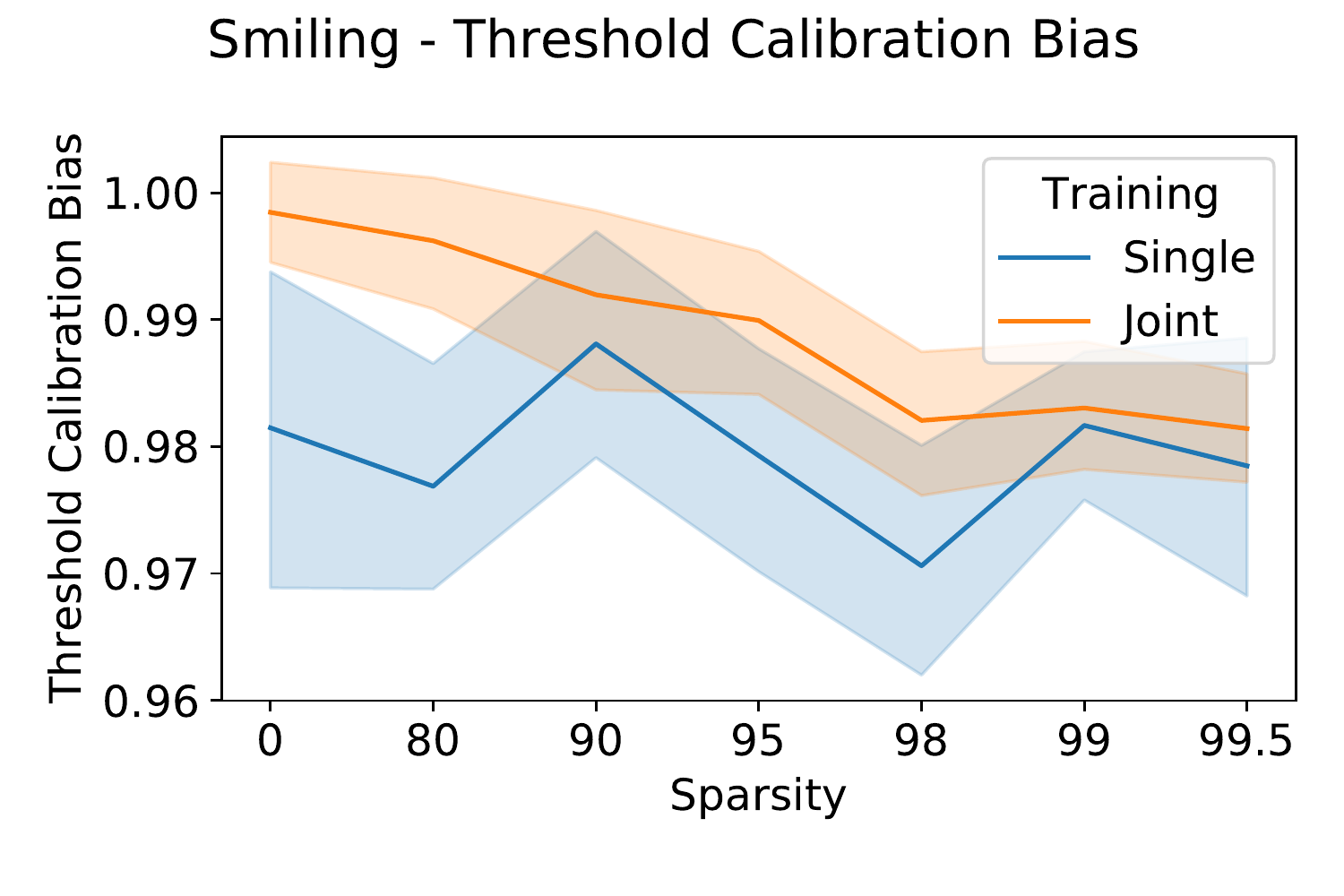}
  \\
  \includegraphics[width=0.12\textwidth]
  {figures/celeba_rn18_oval-face_Male-bas_single.pdf} &
\includegraphics[width=0.12\textwidth]{figures/celeba_rn18_big-nose_Male-bas_single.pdf} &
\includegraphics[width=0.12\textwidth]{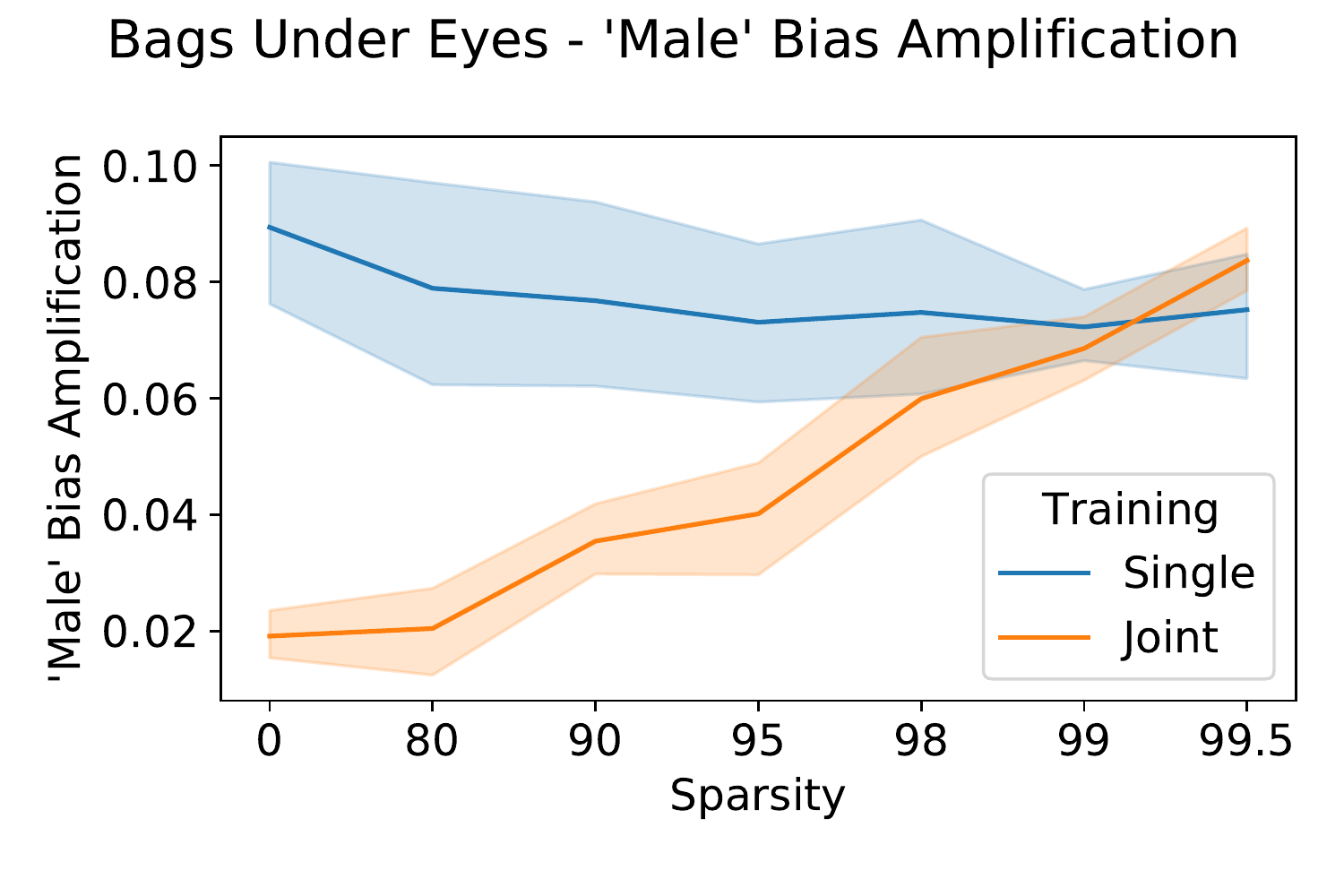} &
\includegraphics[width=0.12\textwidth]{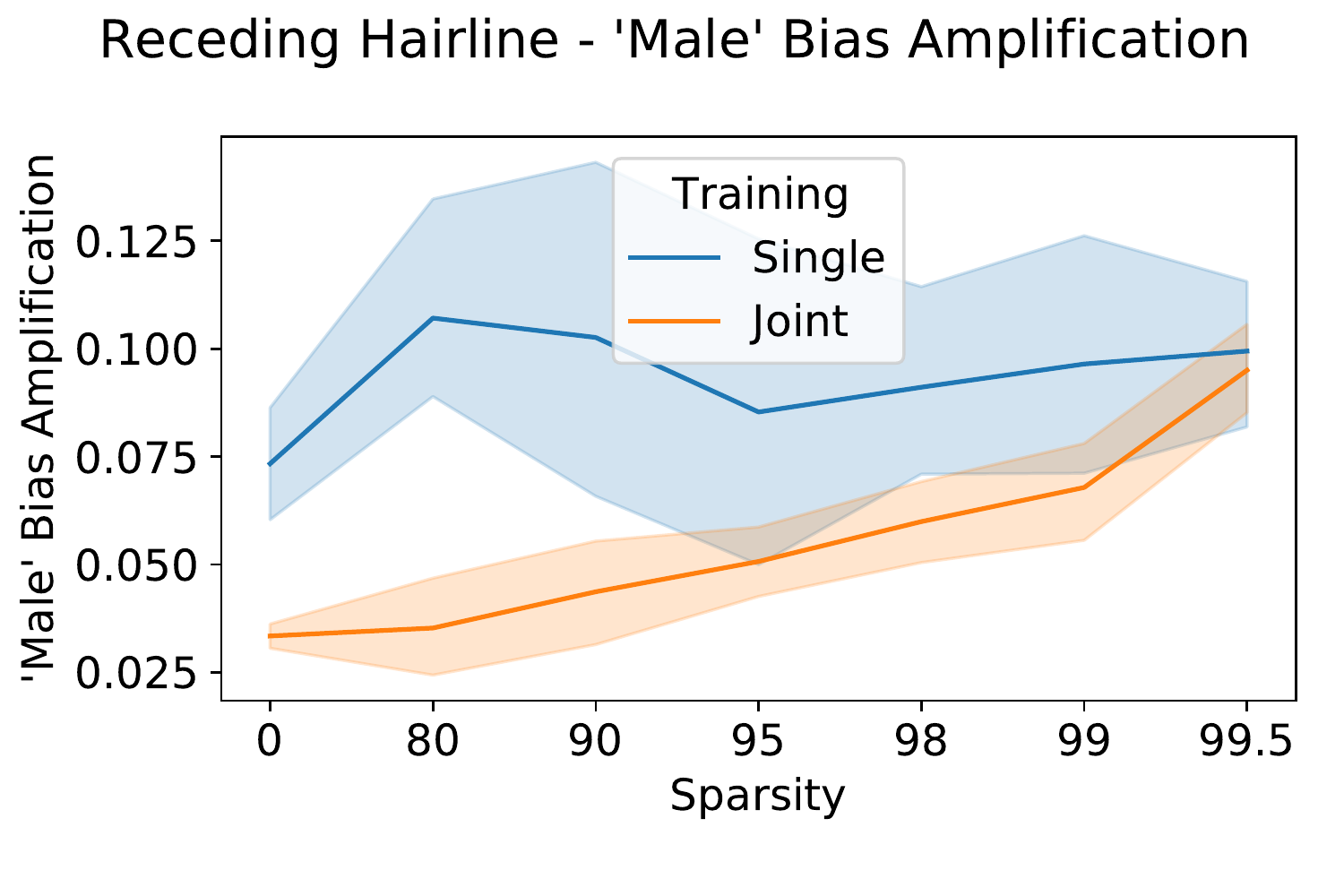} &
&%
\includegraphics[width=0.12\textwidth]{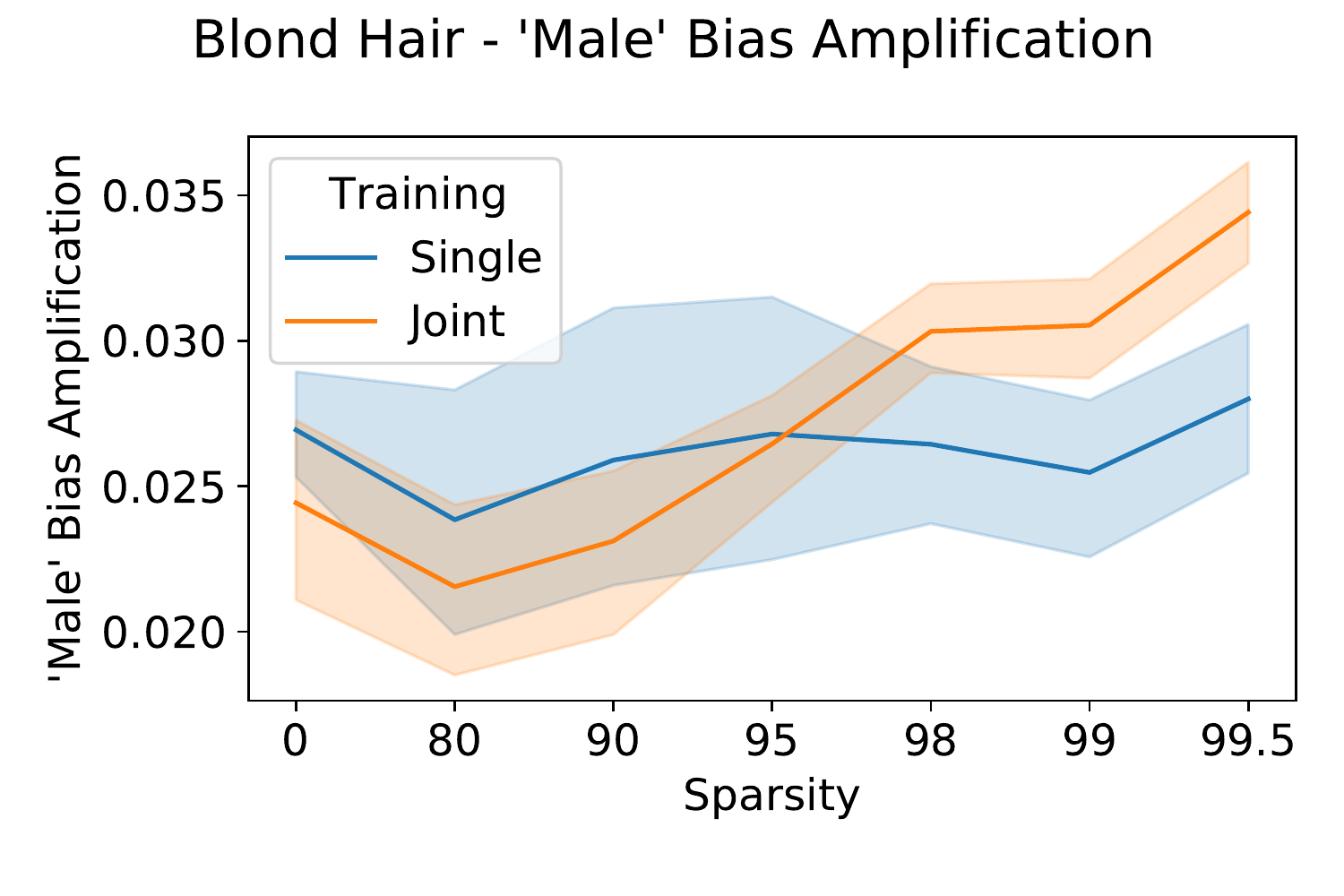} &
\includegraphics[width=0.12\textwidth]{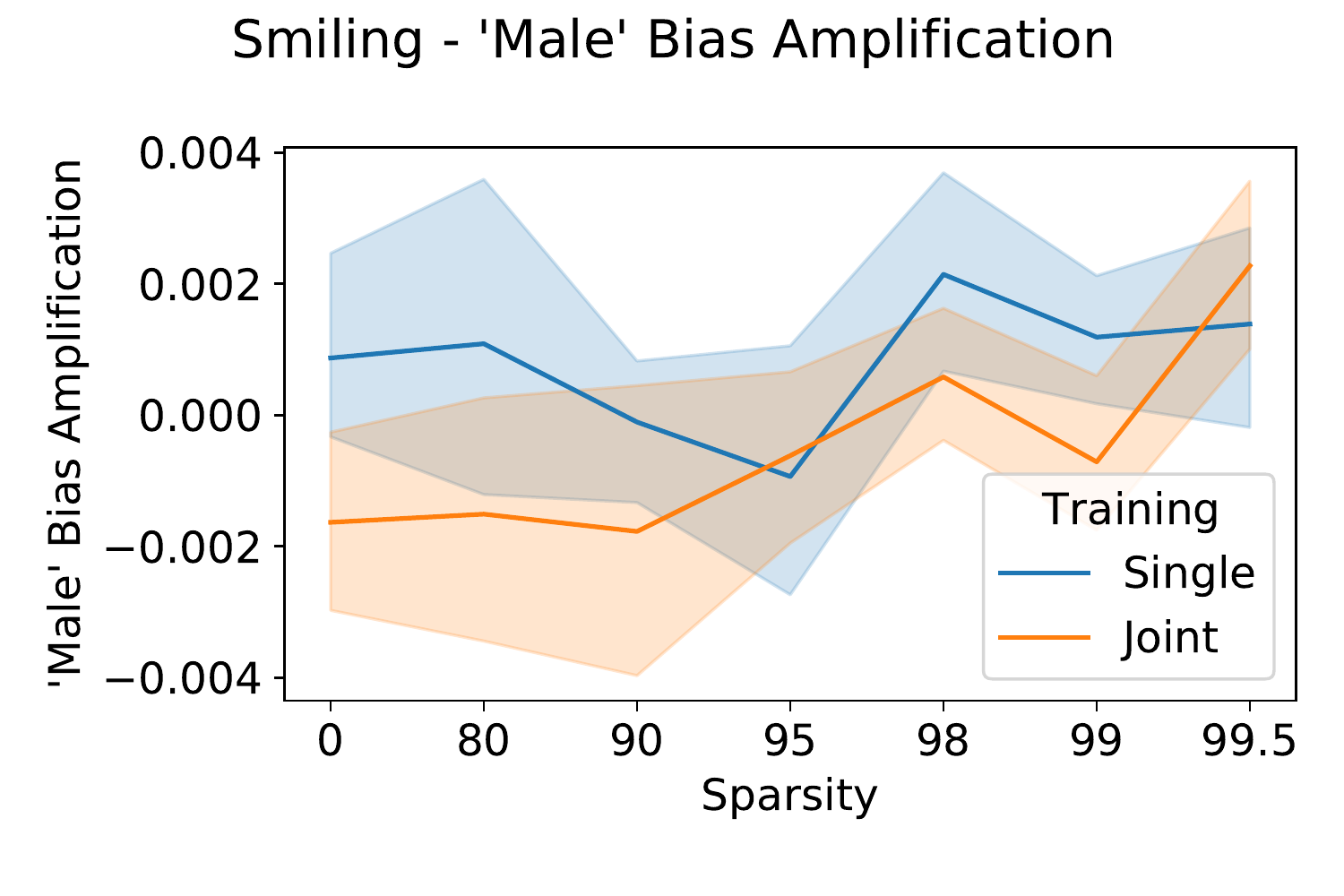}
    \\
  \includegraphics[width=0.12\textwidth]
  {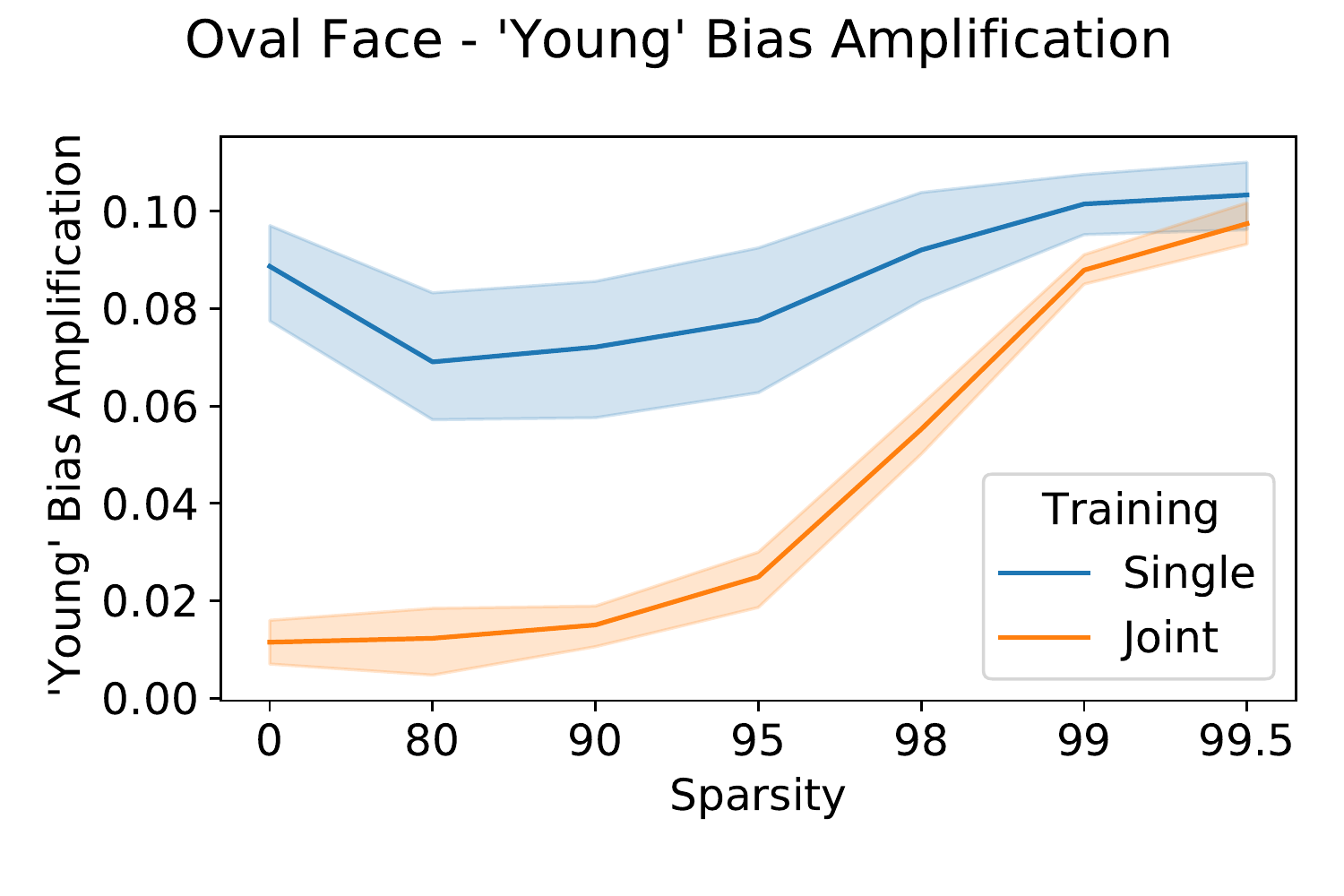} &
\includegraphics[width=0.12\textwidth]{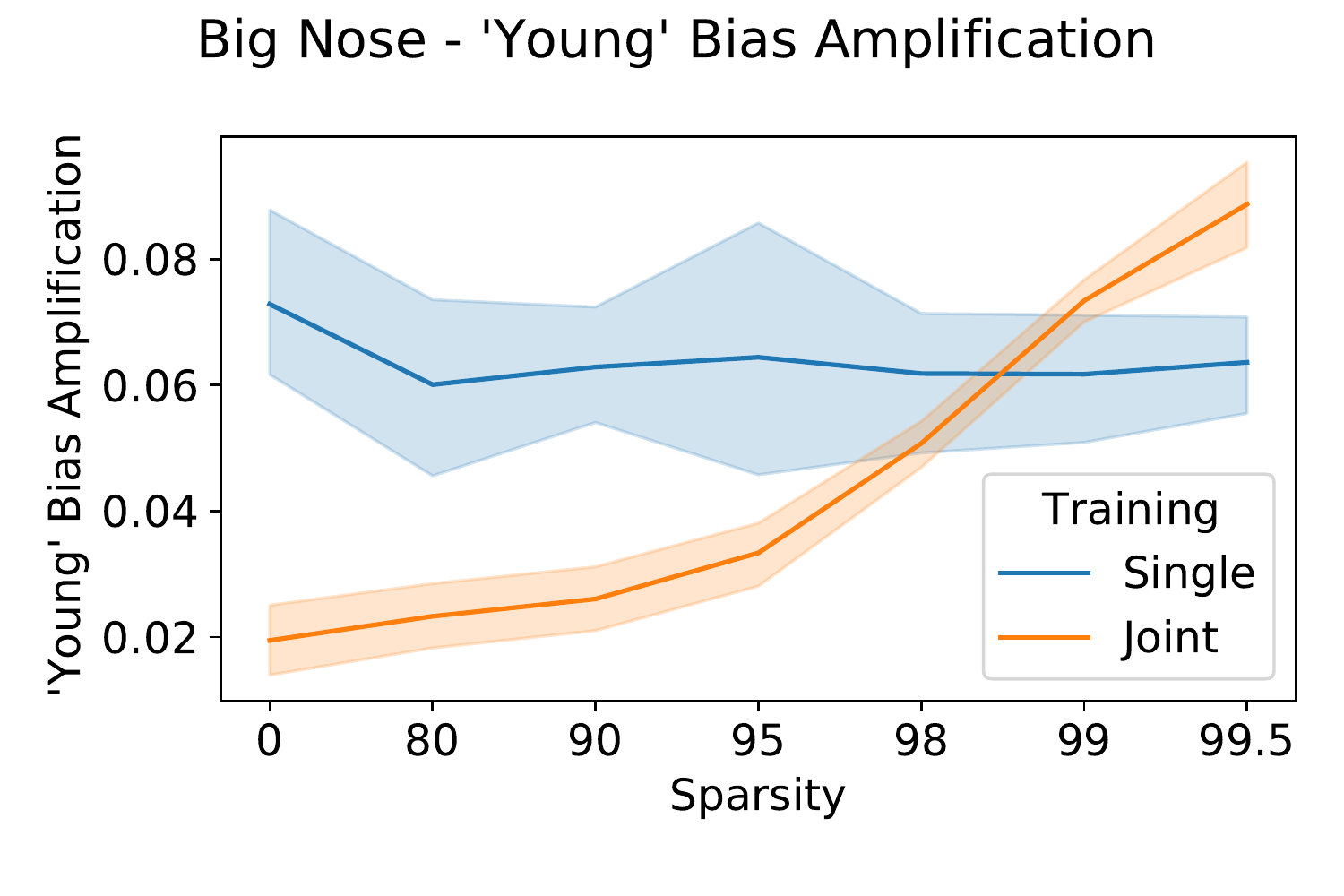} &
\includegraphics[width=0.12\textwidth]{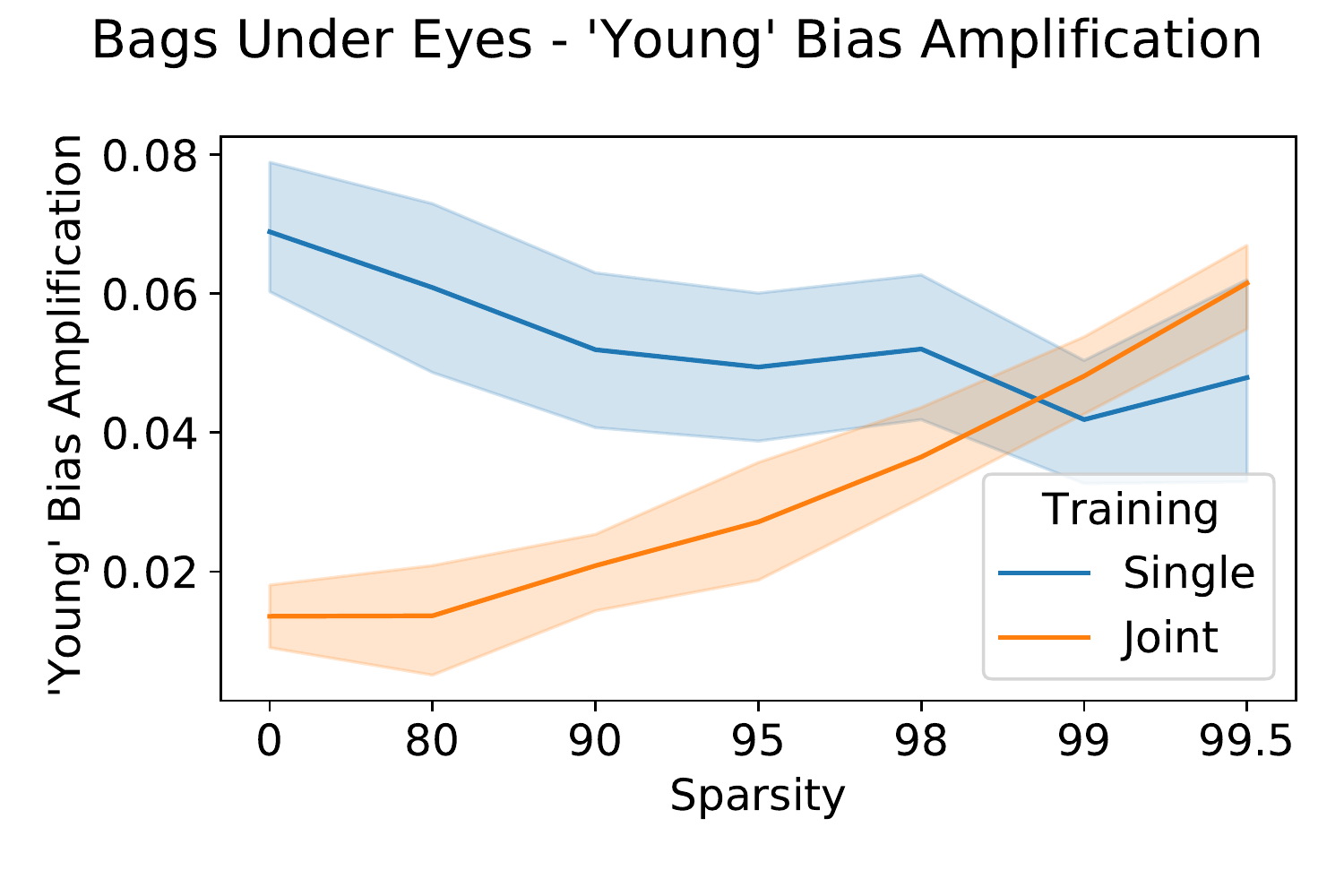} &
\includegraphics[width=0.12\textwidth]{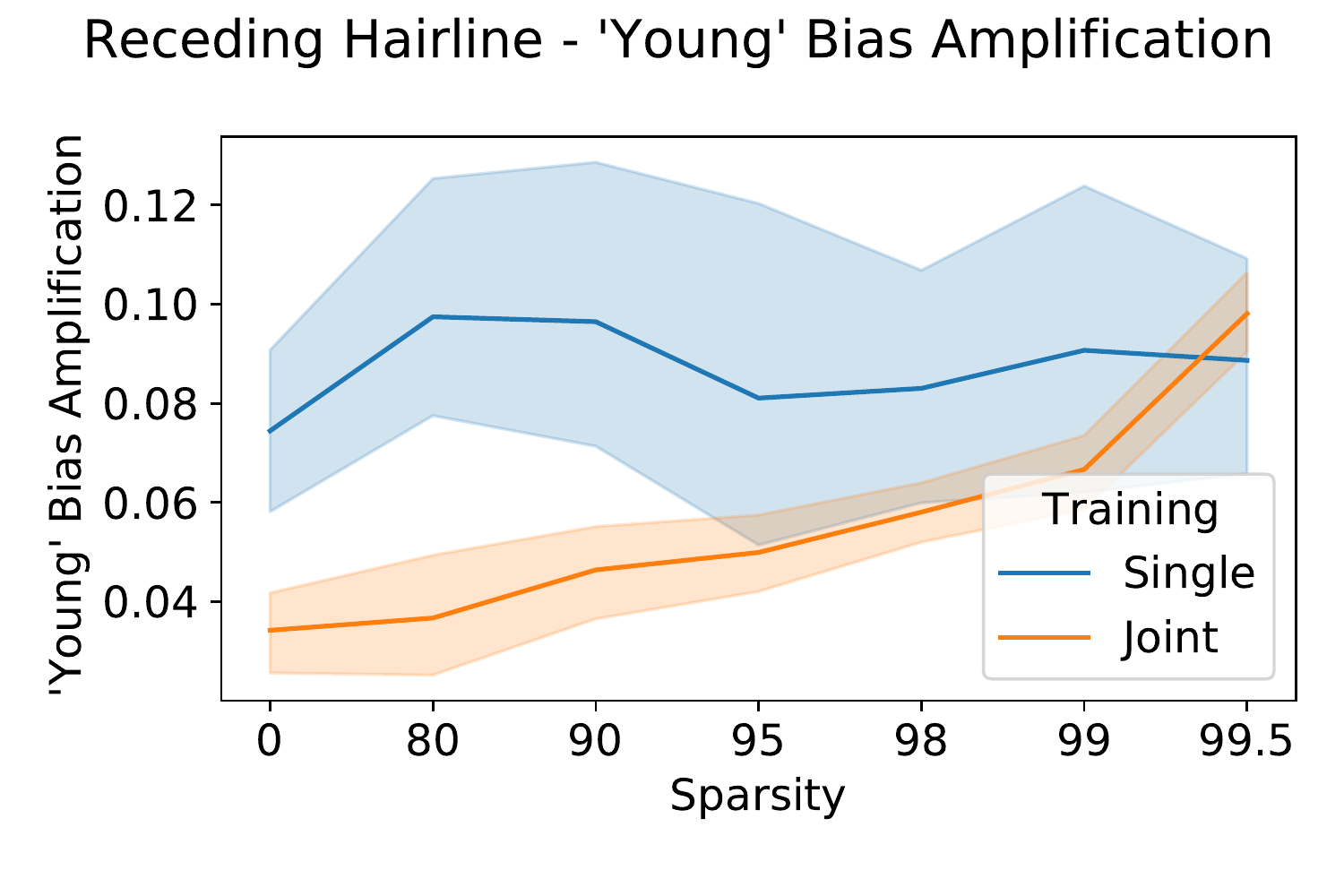} &
\includegraphics[width=0.12\textwidth]{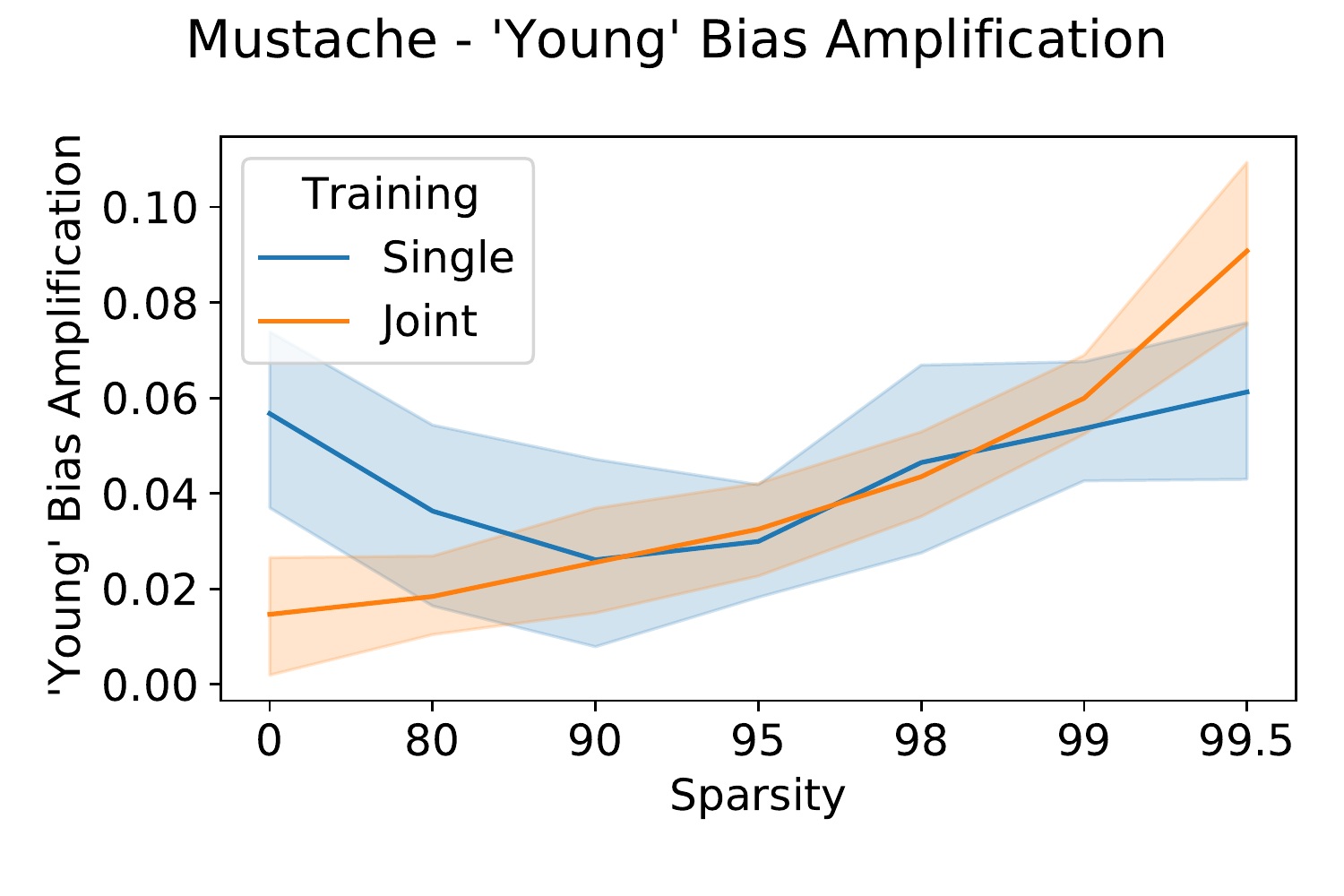} &
\includegraphics[width=0.12\textwidth]{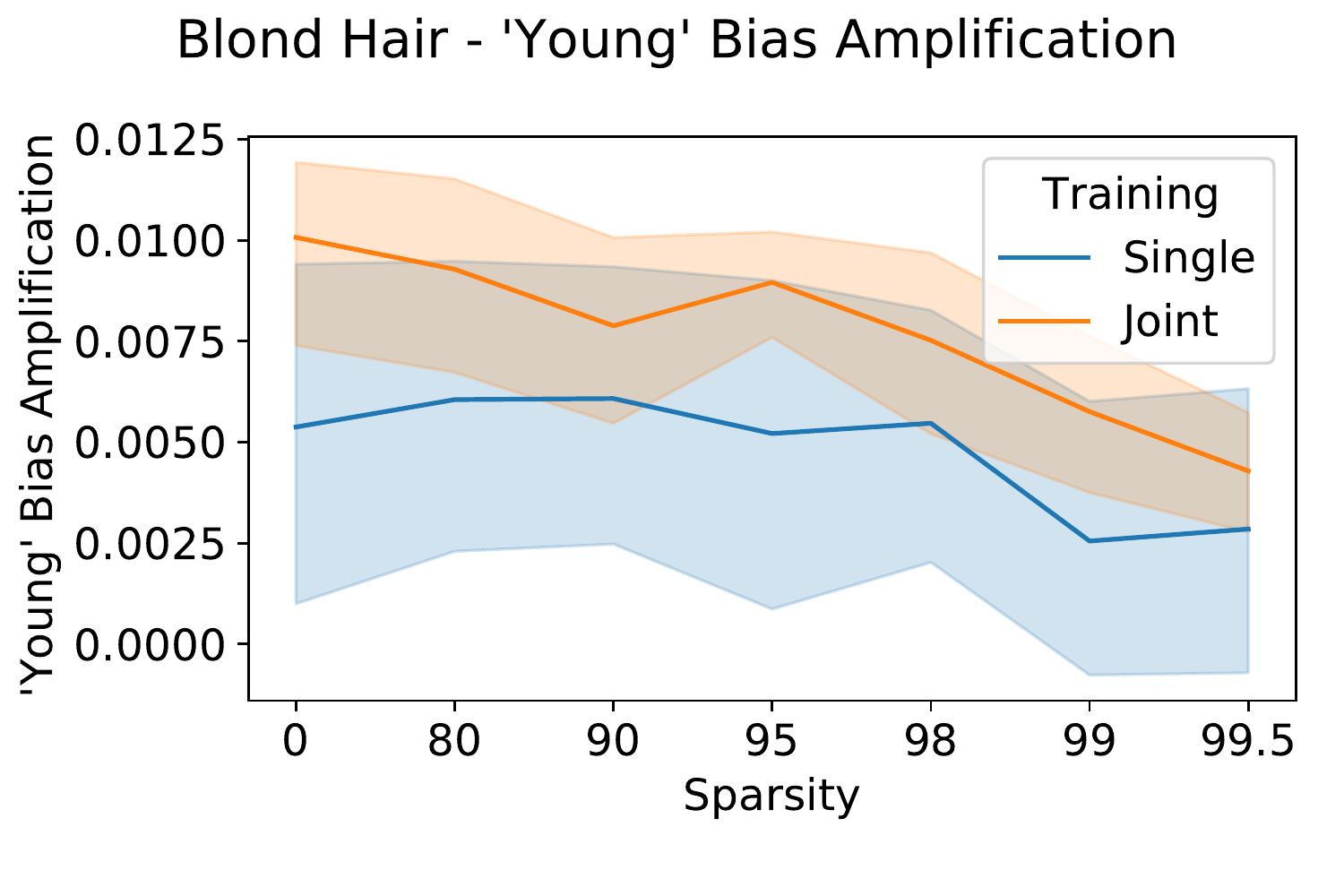} &
    \\
  \includegraphics[width=0.12\textwidth]
  {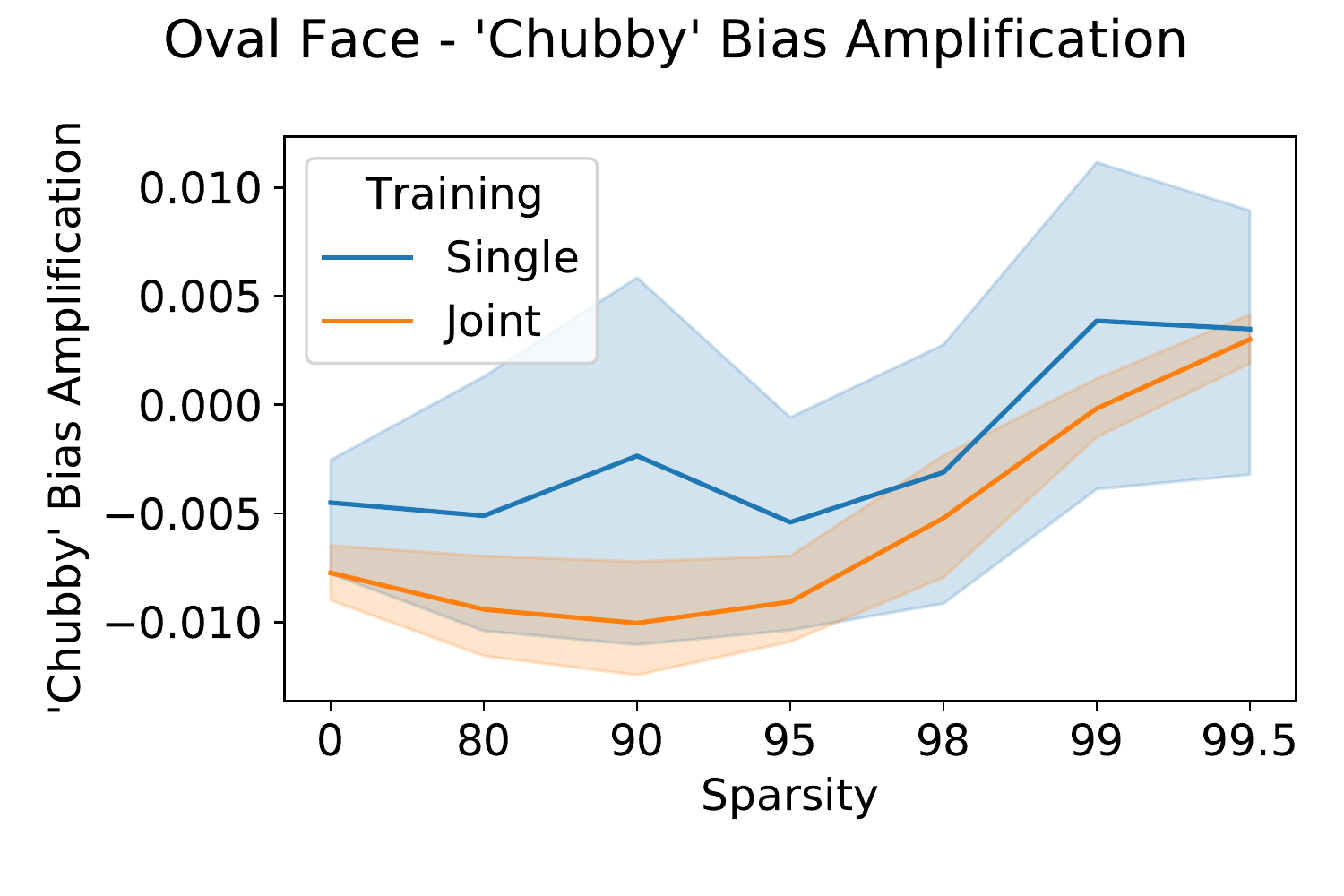} &
\includegraphics[width=0.12\textwidth]{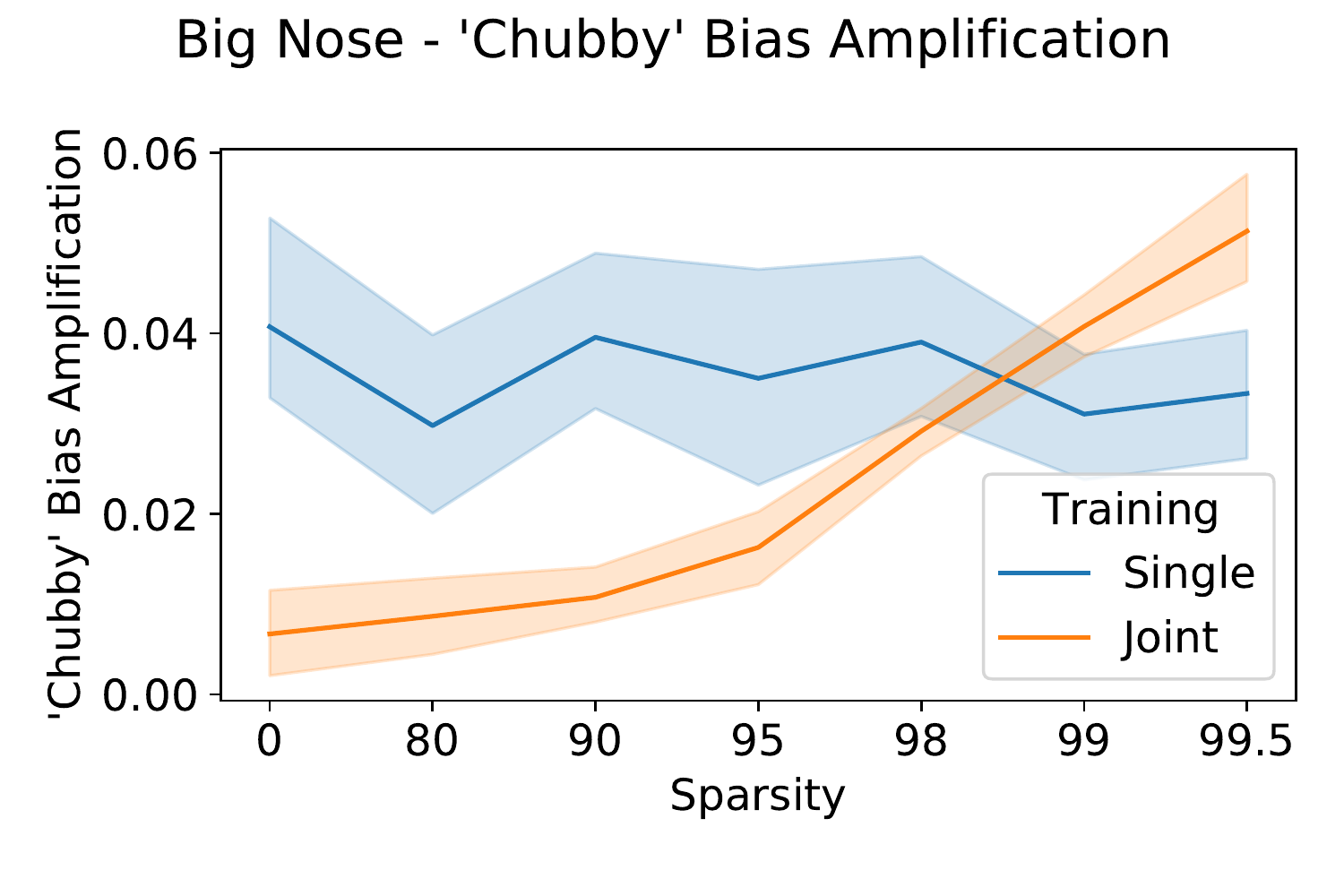} &
\includegraphics[width=0.12\textwidth]{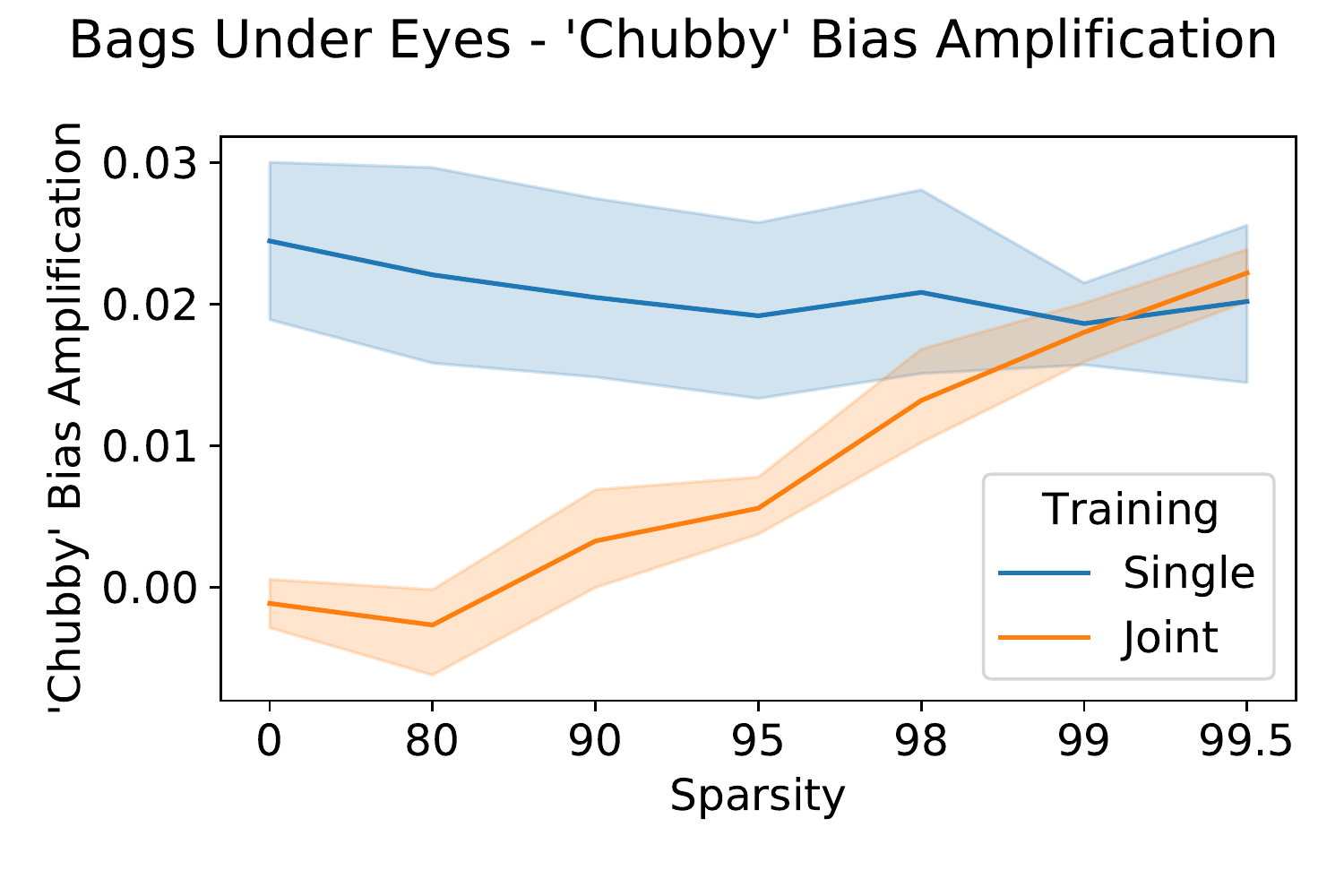} &
\includegraphics[width=0.12\textwidth]{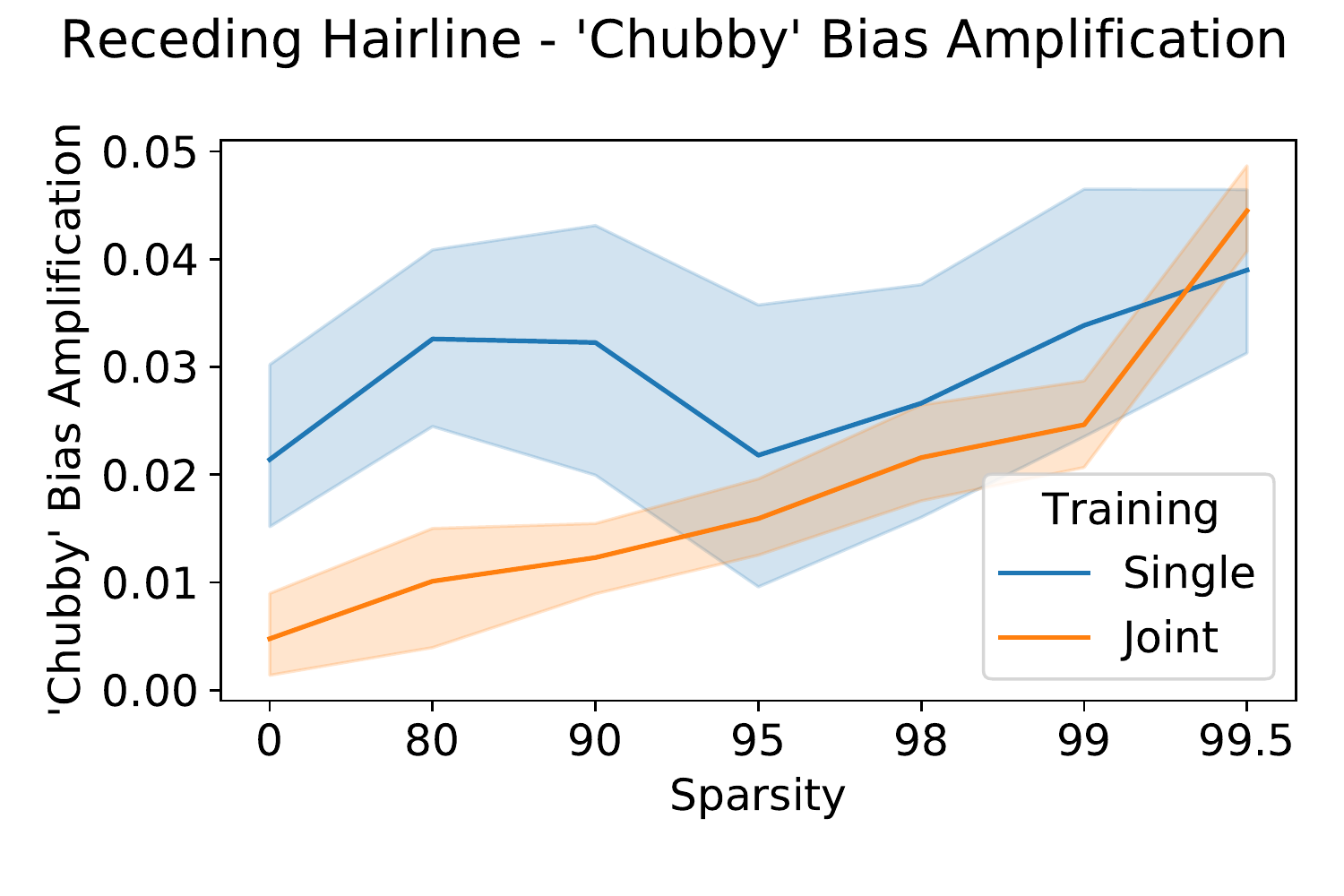} &
\includegraphics[width=0.12\textwidth]{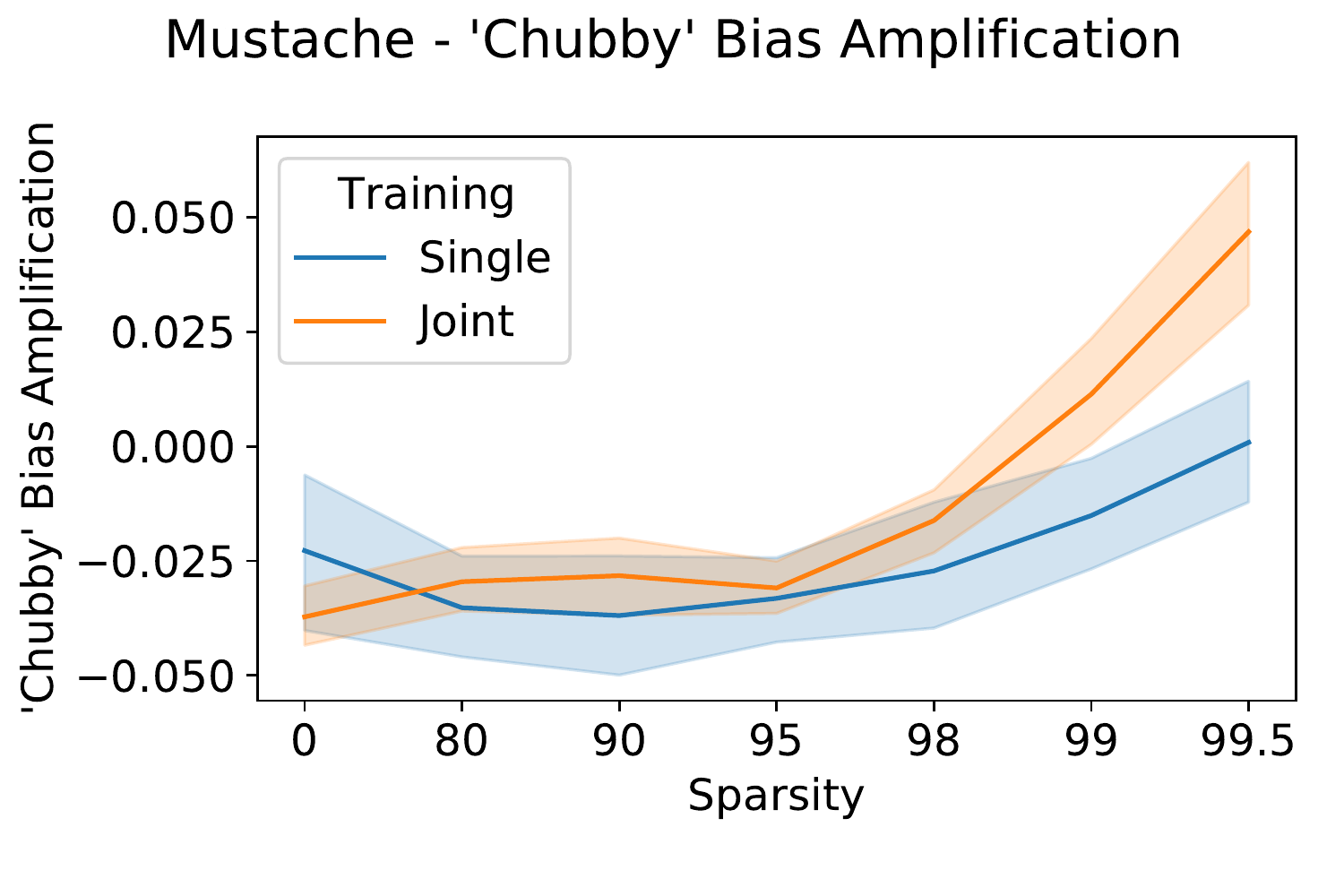} &
\includegraphics[width=0.12\textwidth]{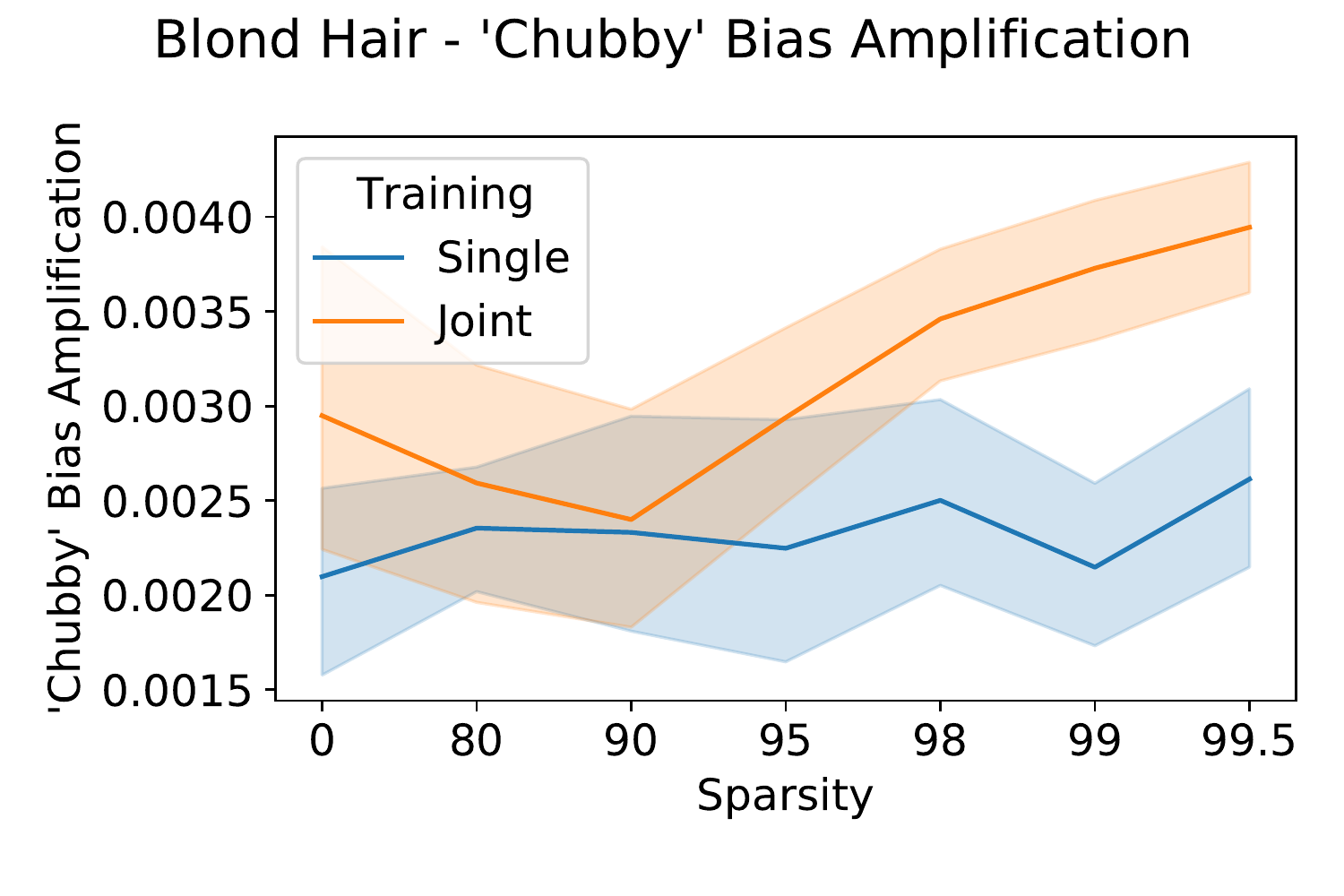} &
\includegraphics[width=0.12\textwidth]{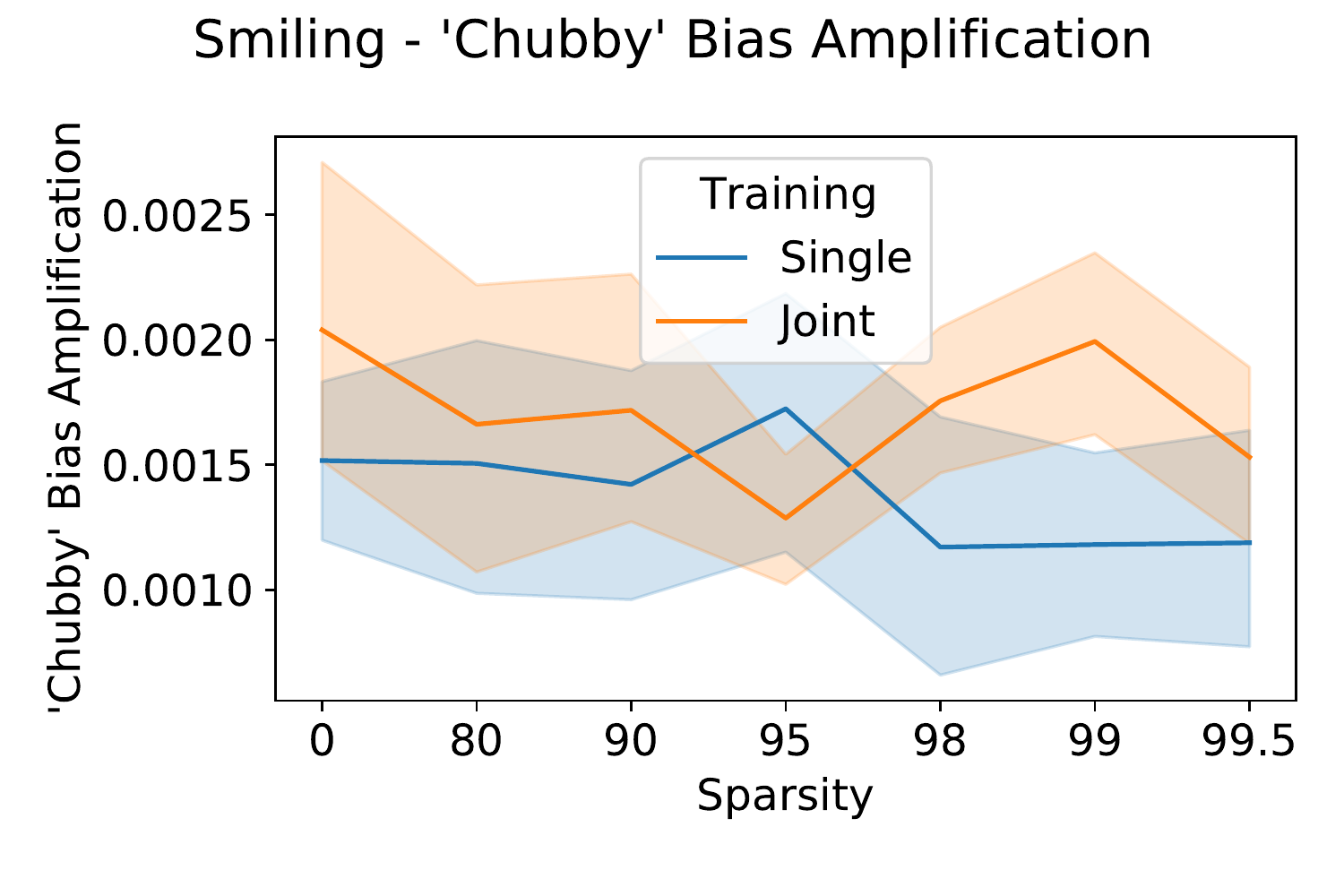}
    \\
      \includegraphics[width=0.12\textwidth]
  {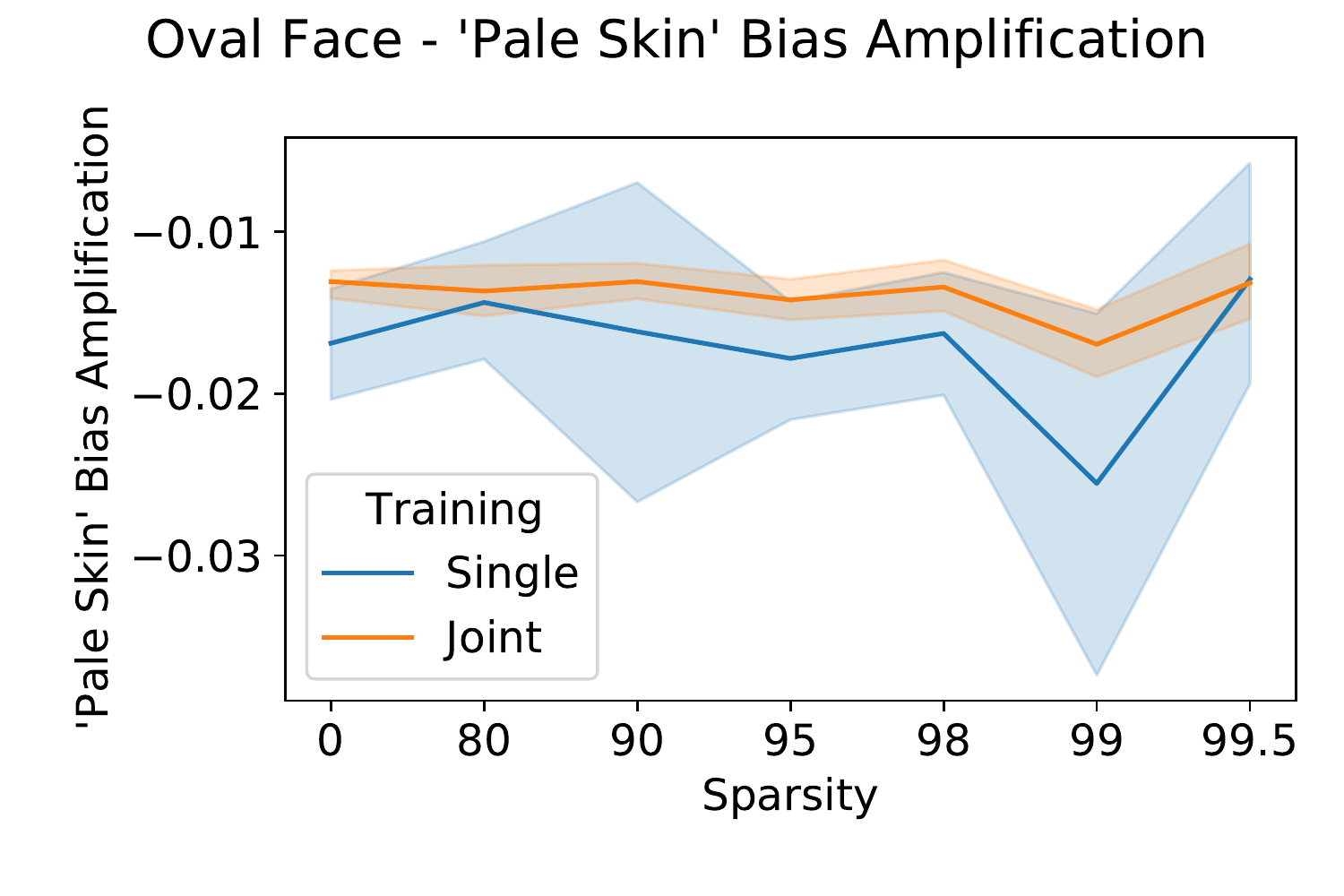} &
\includegraphics[width=0.12\textwidth]{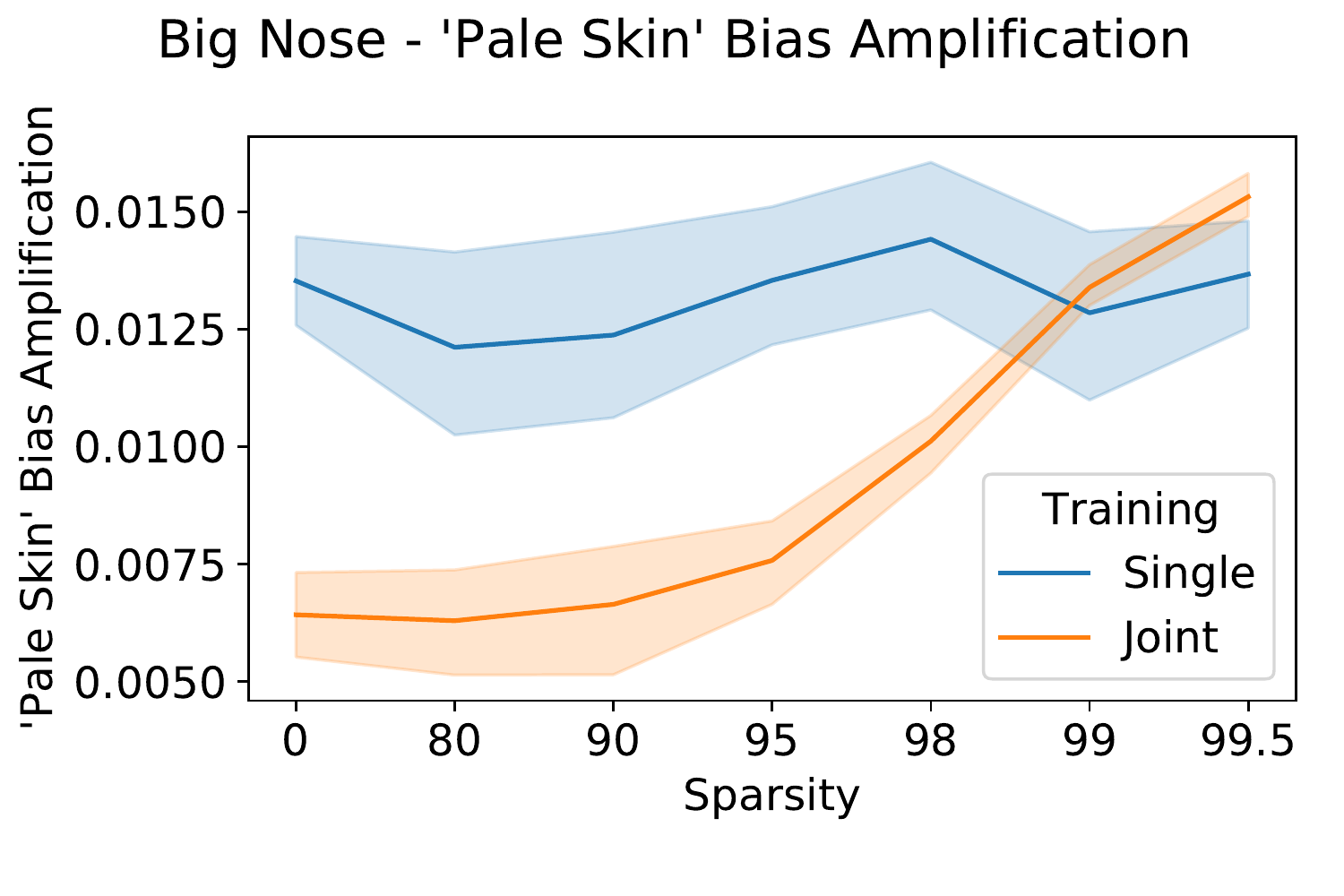} &
\includegraphics[width=0.12\textwidth]{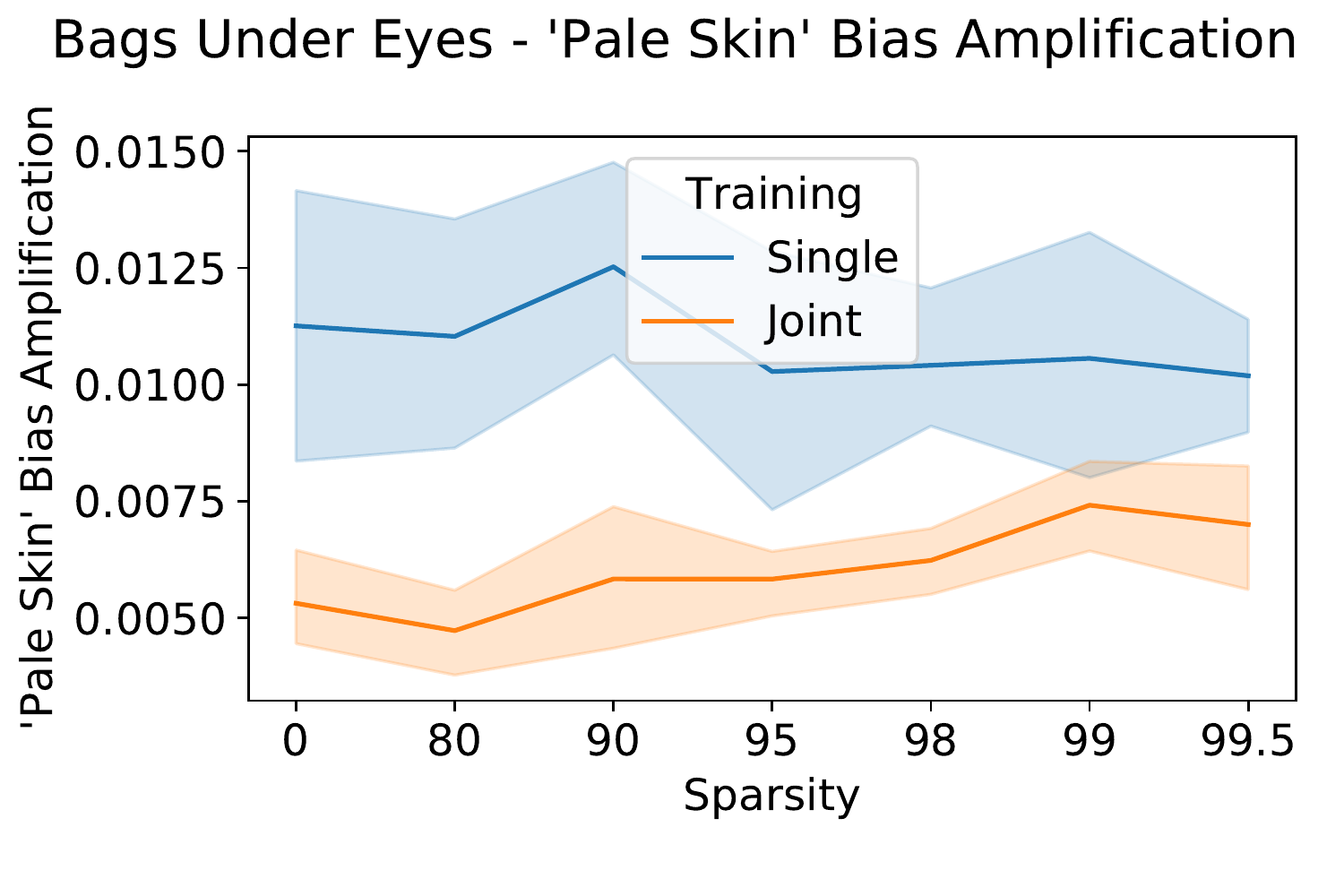} &
\includegraphics[width=0.12\textwidth]{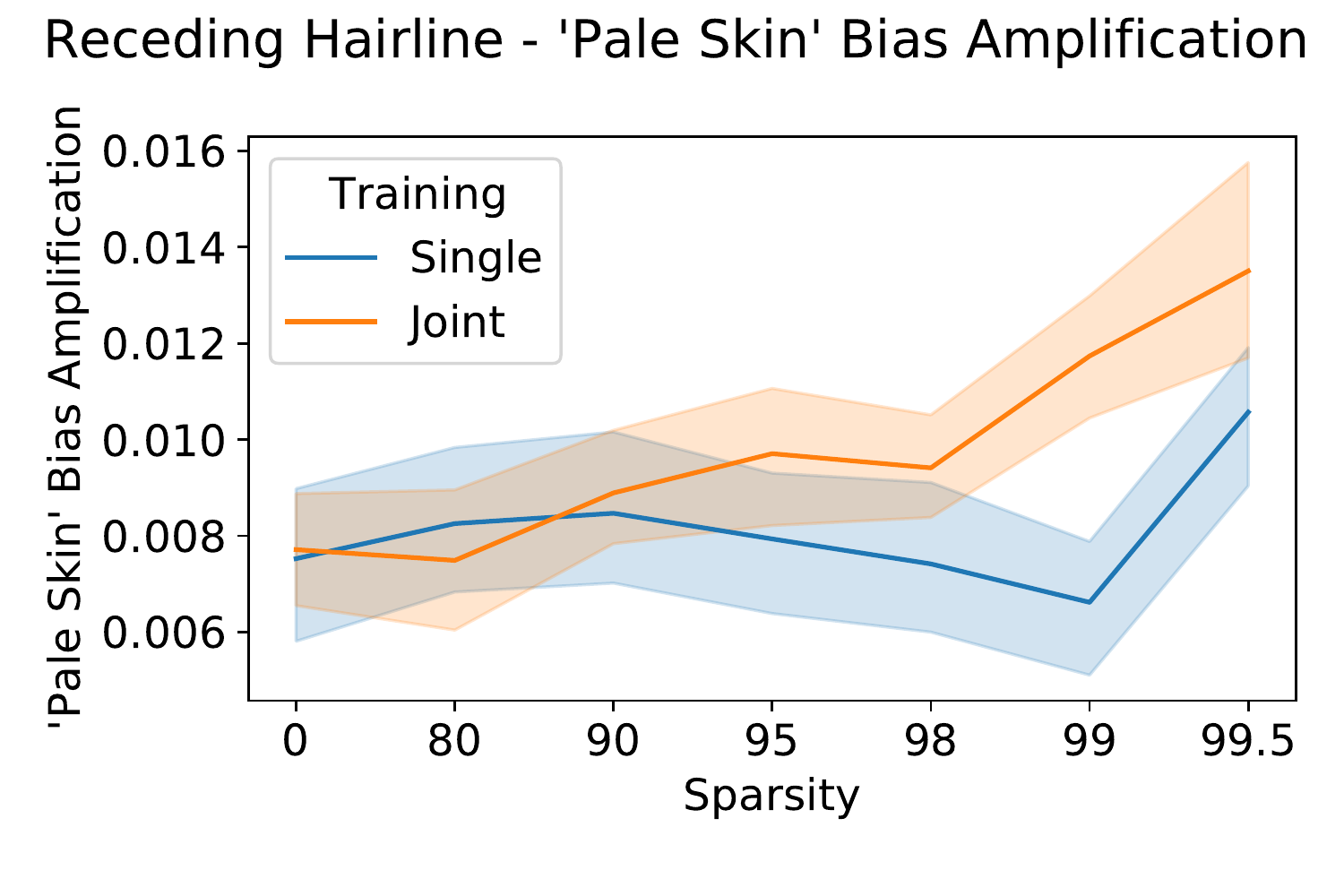} &
&%
\includegraphics[width=0.12\textwidth]{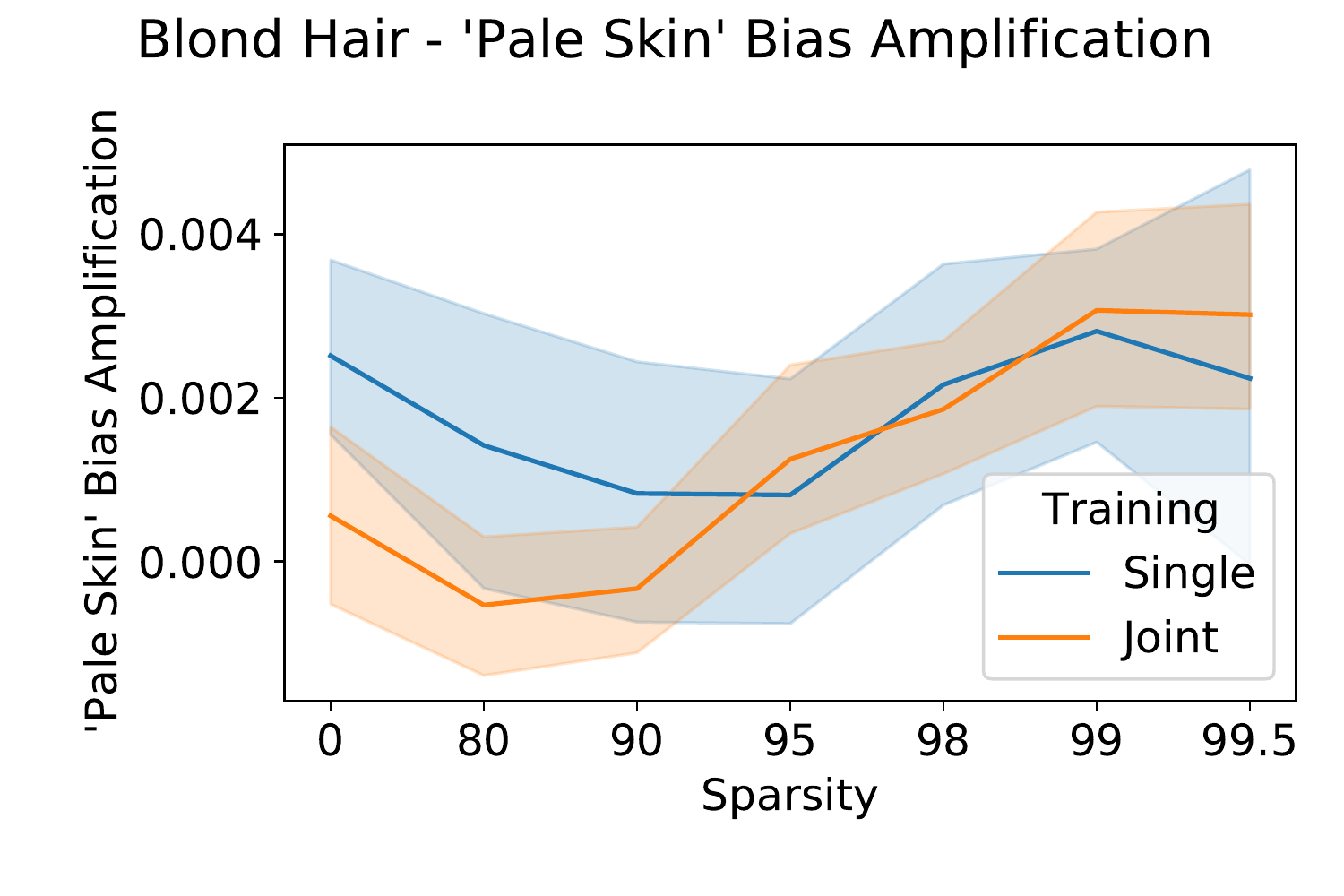} &
\includegraphics[width=0.12\textwidth]{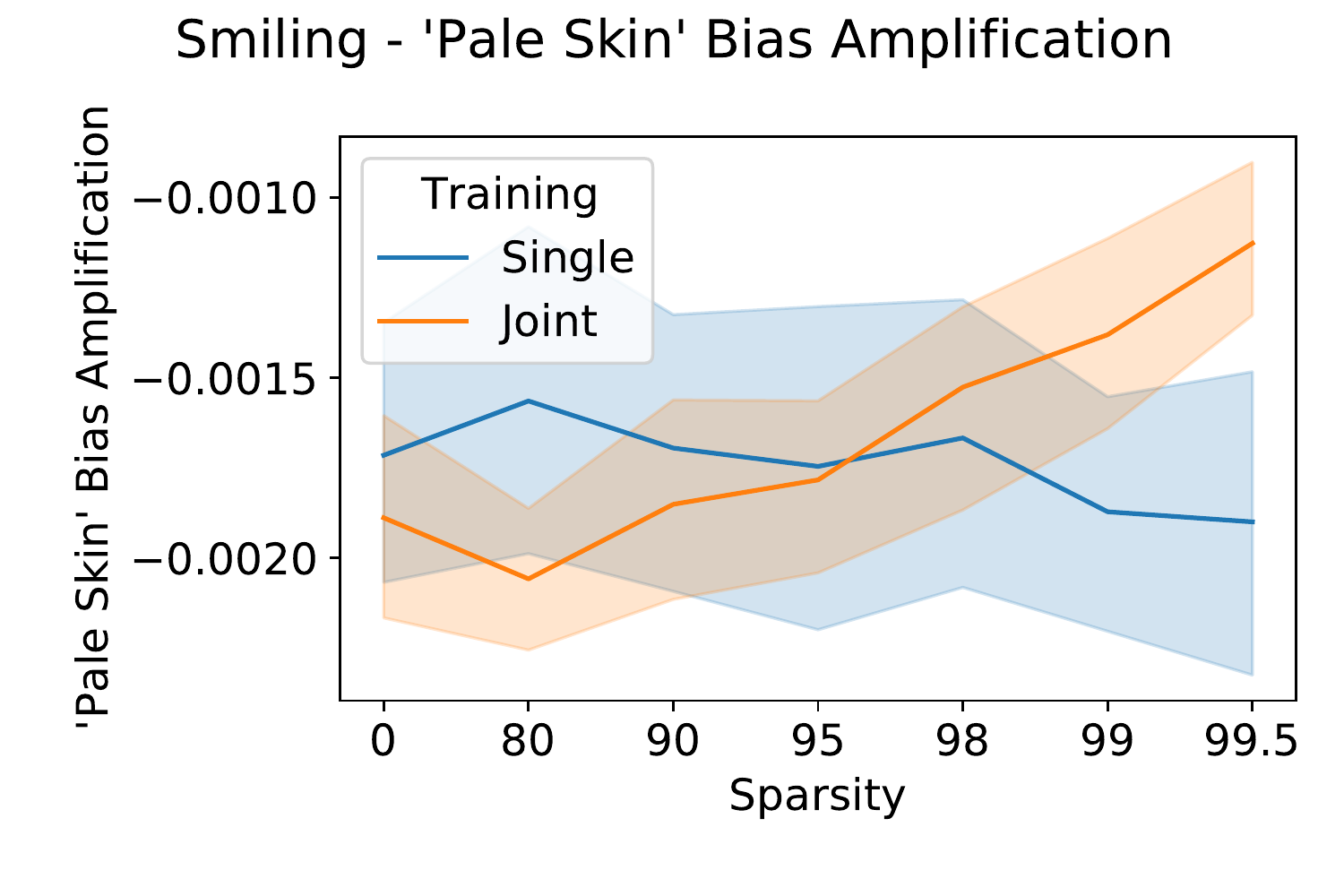}
    \\
\end{tabular}
    \caption{[CelebA / ResNet18 / Single Attribute / GMP-RI] Effect of single versus joint training of attributes on Accuracy (first row), Uncertainty (second row), ECE (third row), Threshold Calibration Bias (fourth row), 
    and Bias Amplification for the `Male', `Young', `Chubby', and `Pale Skin' attributes (fifth-eighth rows), on the ResNet18 CelebA model, predicting, from left to right, Oval Face, Big Nose, Bags Under Eyes, Receding Hairline, Mustache, Blond Hair, and Smiling). Orange denotes results from joint runs and blue denotes results from single runs. Omitted panels are cases where BA cannot be computed, either because there is no relationship between the predicted attribute and the category, or because the attribute is not present for one of the values of the category.}
    \label{fig:celeba_rn18_single_full}
\end{figure}

In this section we provide and discuss Figure \ref{fig:celeba_rn18_single_full}, which is a more complete version of Figure~\ref{fig:celeba_rn18_single} (Accuracy and Bias on singly-trained models); this version includes all seven binary attributes for which we ran the experiment, and all metrics. We observe that the conclusions which we described in Sections~\ref{sec:sparse-bias} for the Oval Face and Big Nose attributes generally hold true for the additional five attributes (Bgs Under Eyes, Receding Hairline, Mustache, Blond Hair, and Smiling) as well. We observe that model accuracy and AUC is generally higher for single-attribute models than joint models, at no or low sparsities, but roughly equal for high sparsities. Further, singly-trained models are much less impacted by sparsity than jointly-trained models when it comes to both Systematic and Categorical bias. However, this manifests as \emph{less} bias in jointly-trained models at low sparsity, and roughly equal bias at high sparsities ($\geq 95\%$).

\begin{figure}[ht]
\centering
\begin{tabular}{ccccccc}
\includegraphics[width=0.12\textwidth]{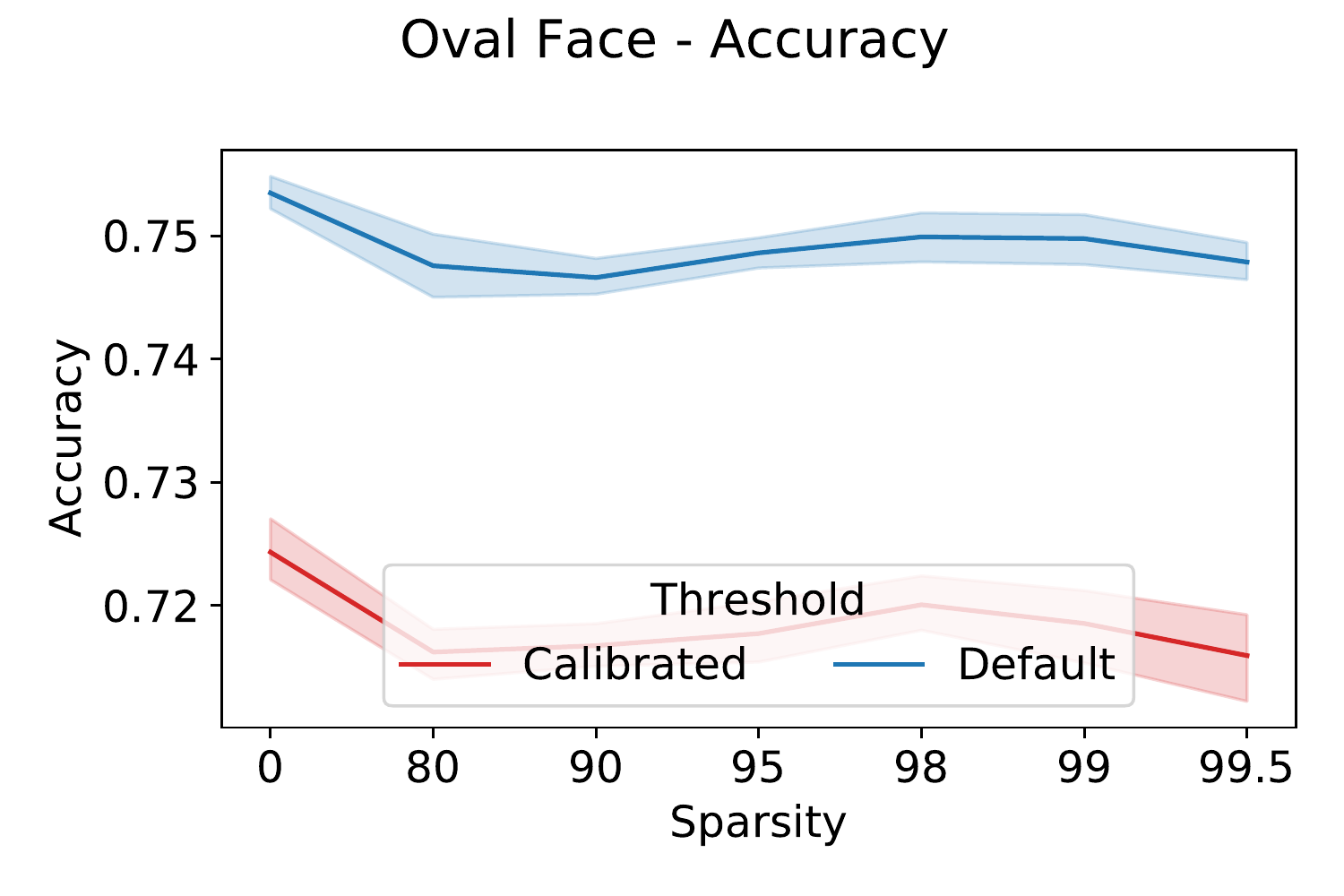} &
\includegraphics[width=0.12\textwidth]{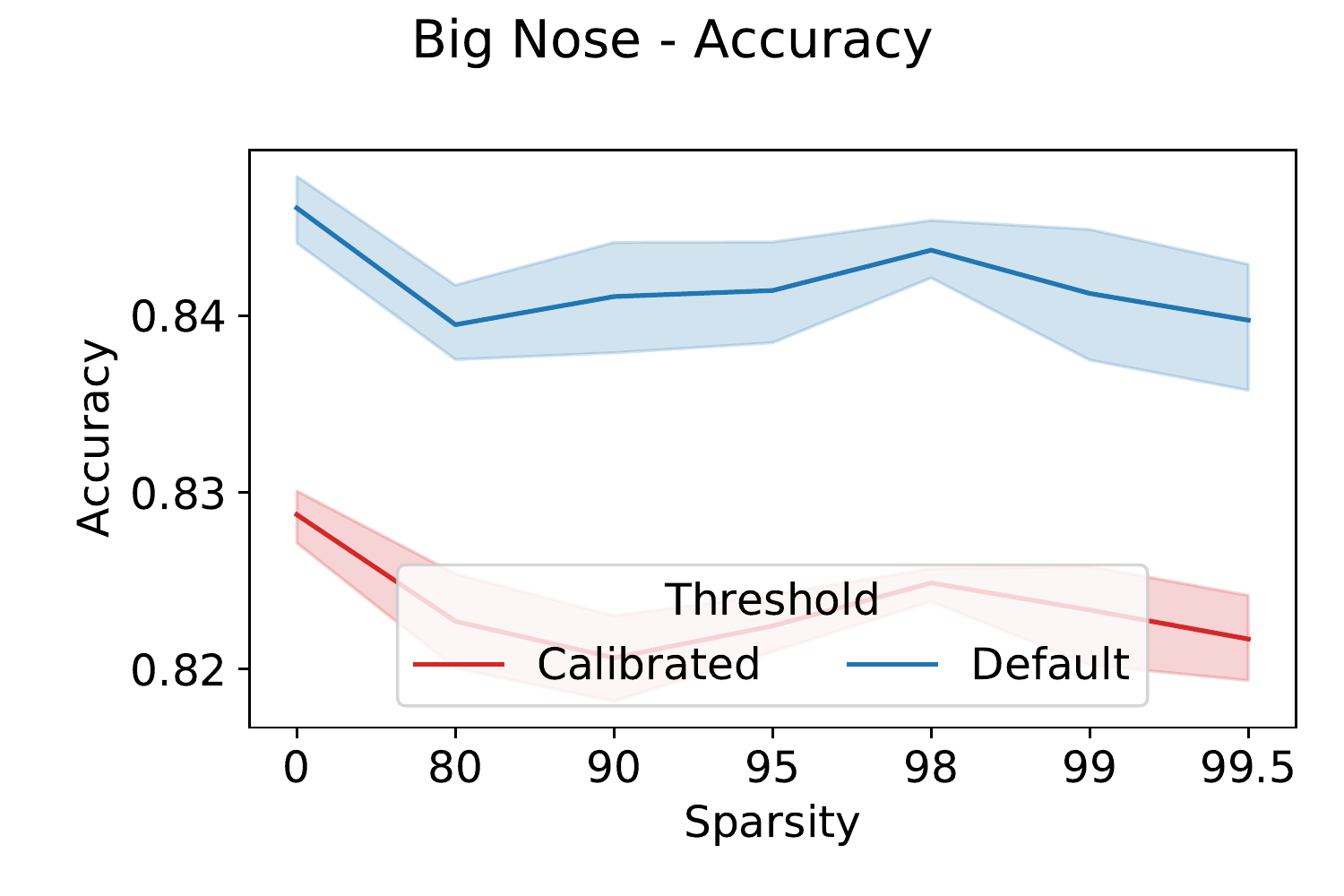} &
\includegraphics[width=0.12\textwidth]{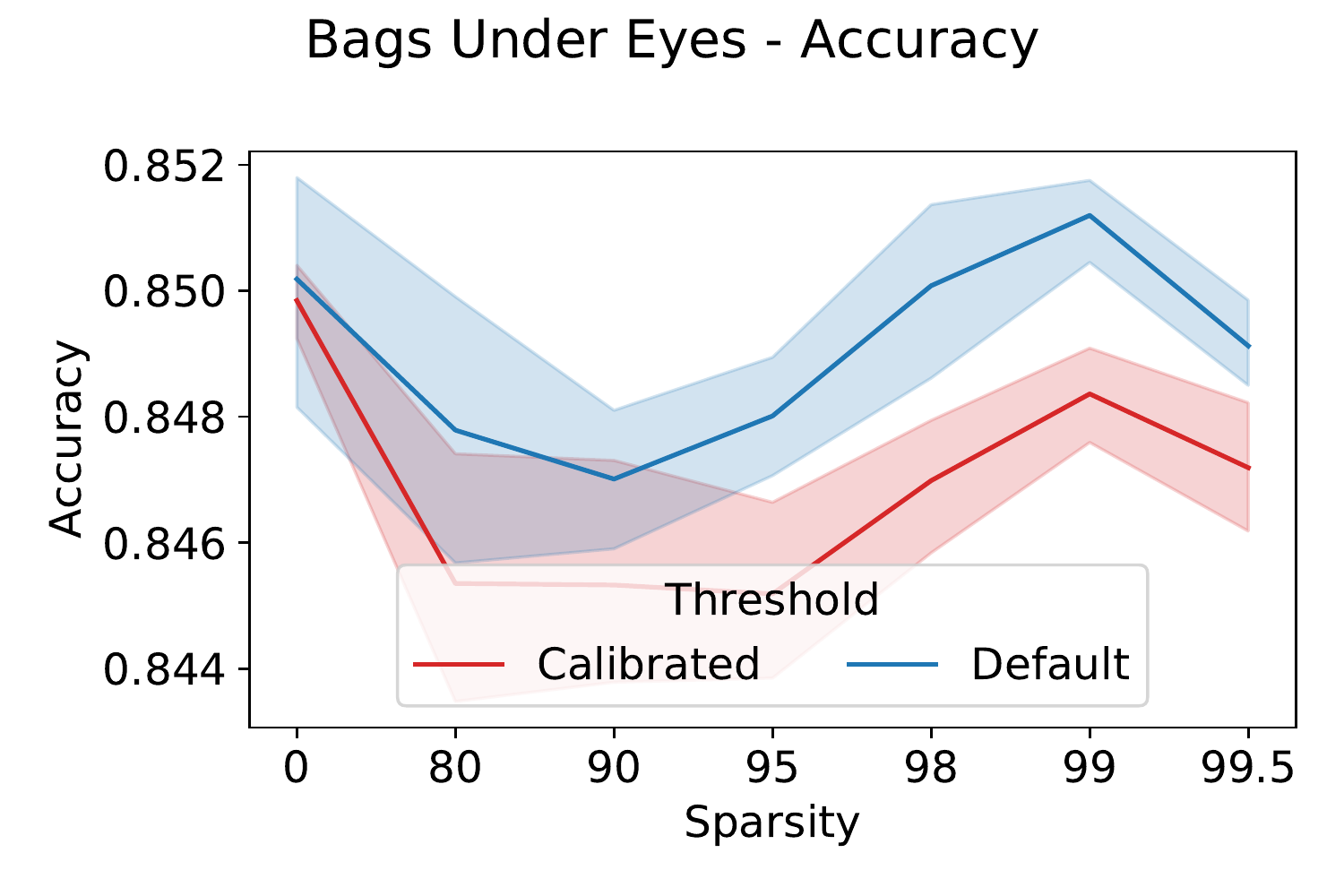} &
\includegraphics[width=0.12\textwidth]{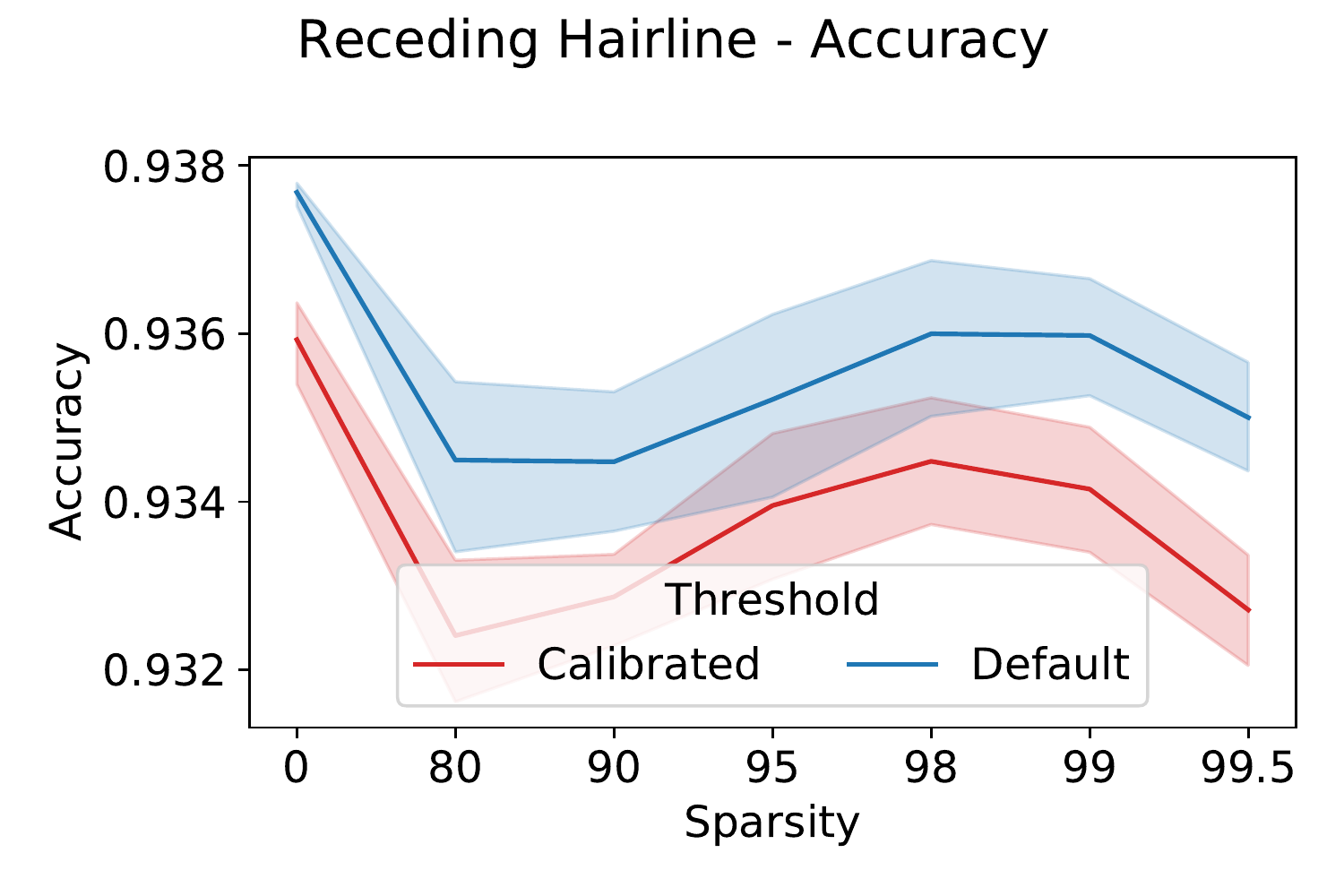} &
\includegraphics[width=0.12\textwidth]{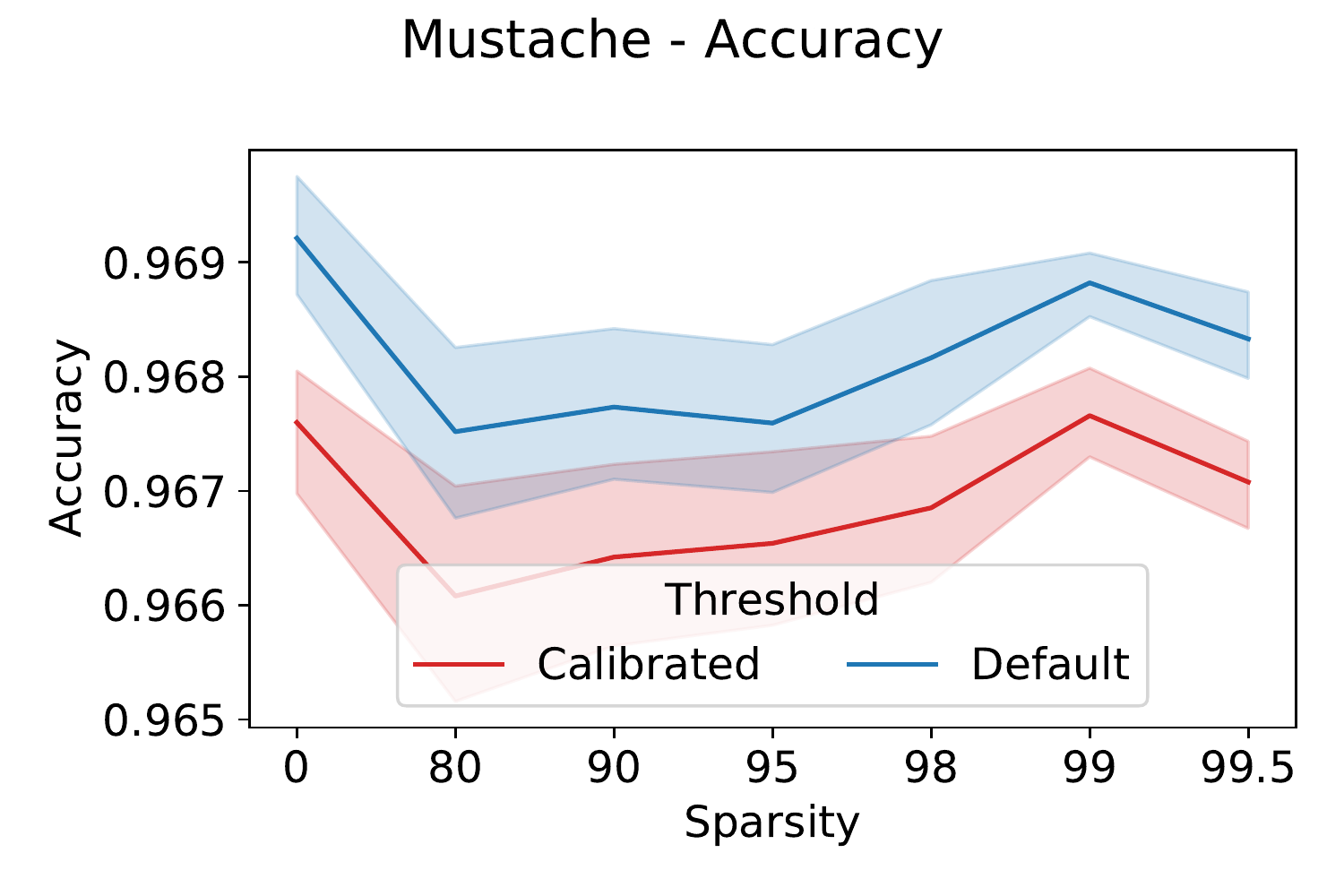} &
\includegraphics[width=0.12\textwidth]{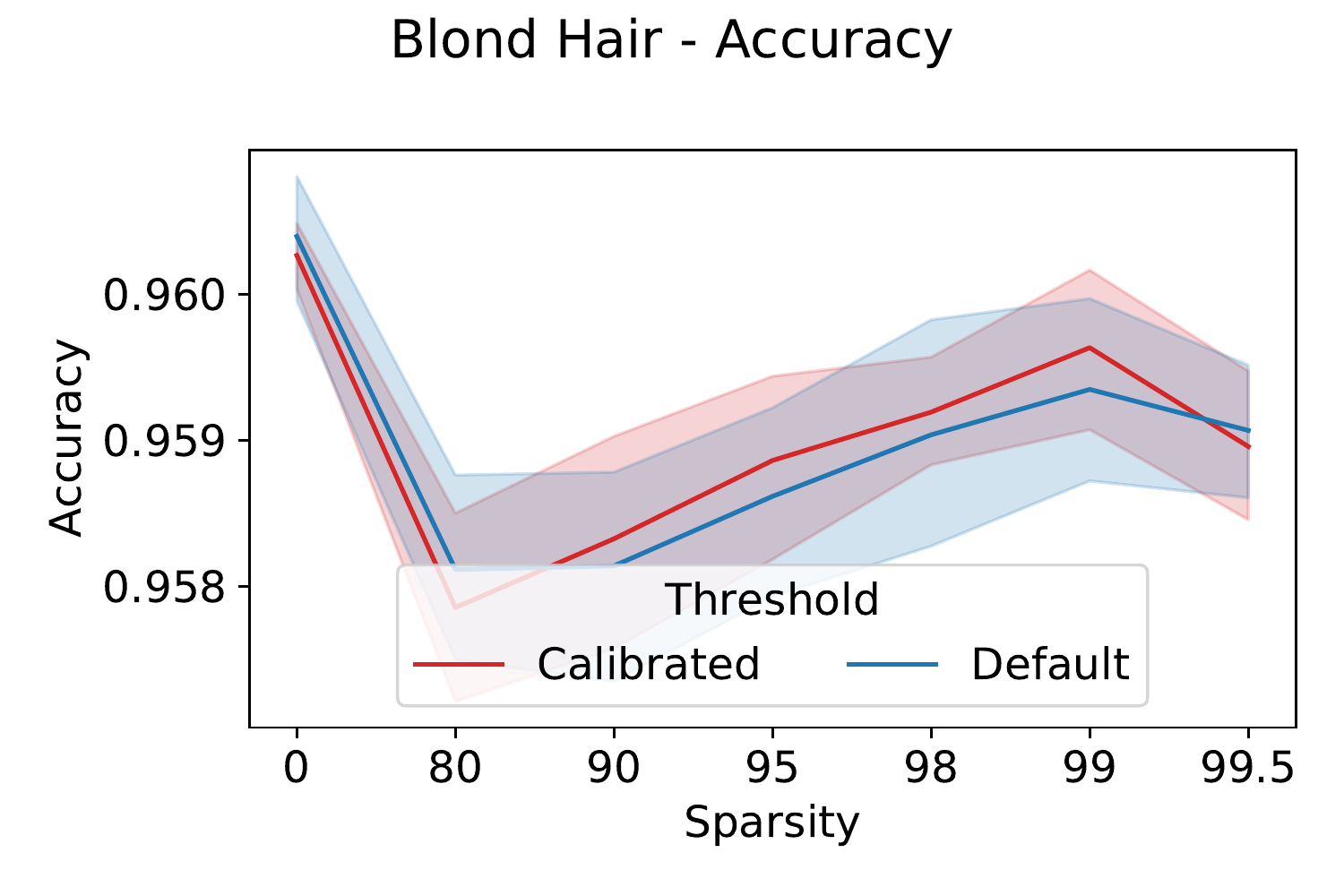} &
\includegraphics[width=0.12\textwidth]{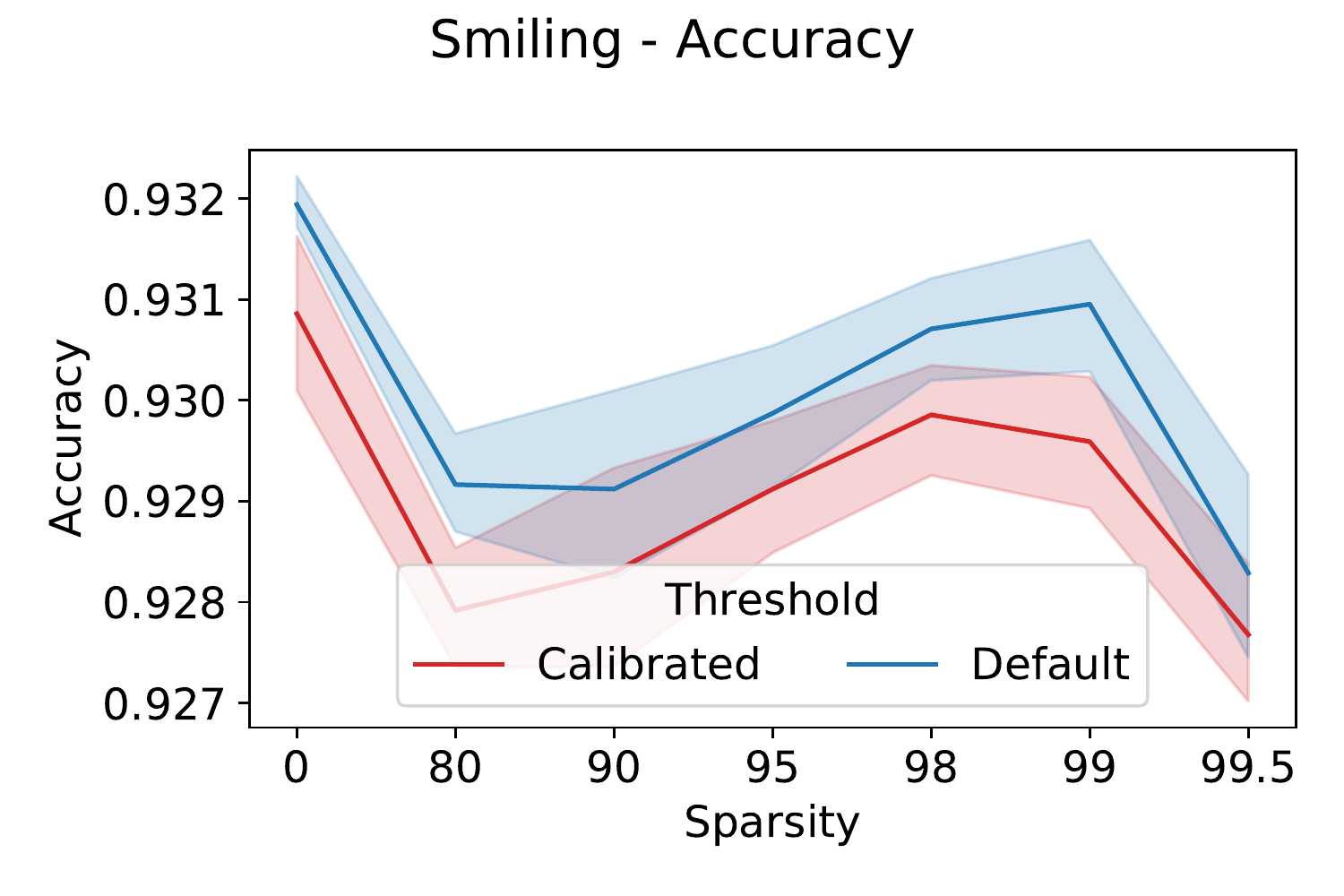}

\\
    \includegraphics[width=0.12\textwidth]
  {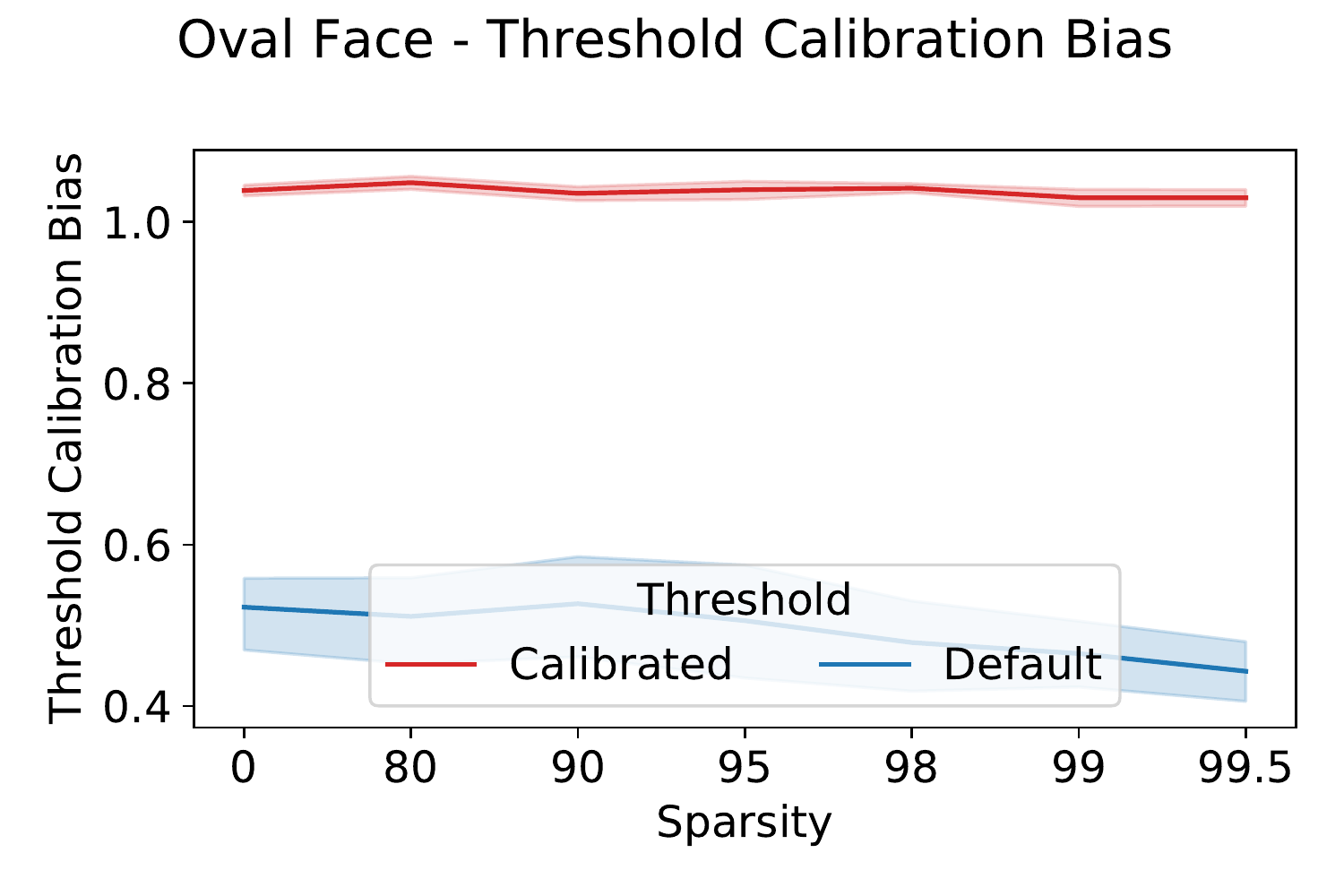} &
  \includegraphics[width=0.12\textwidth]
  {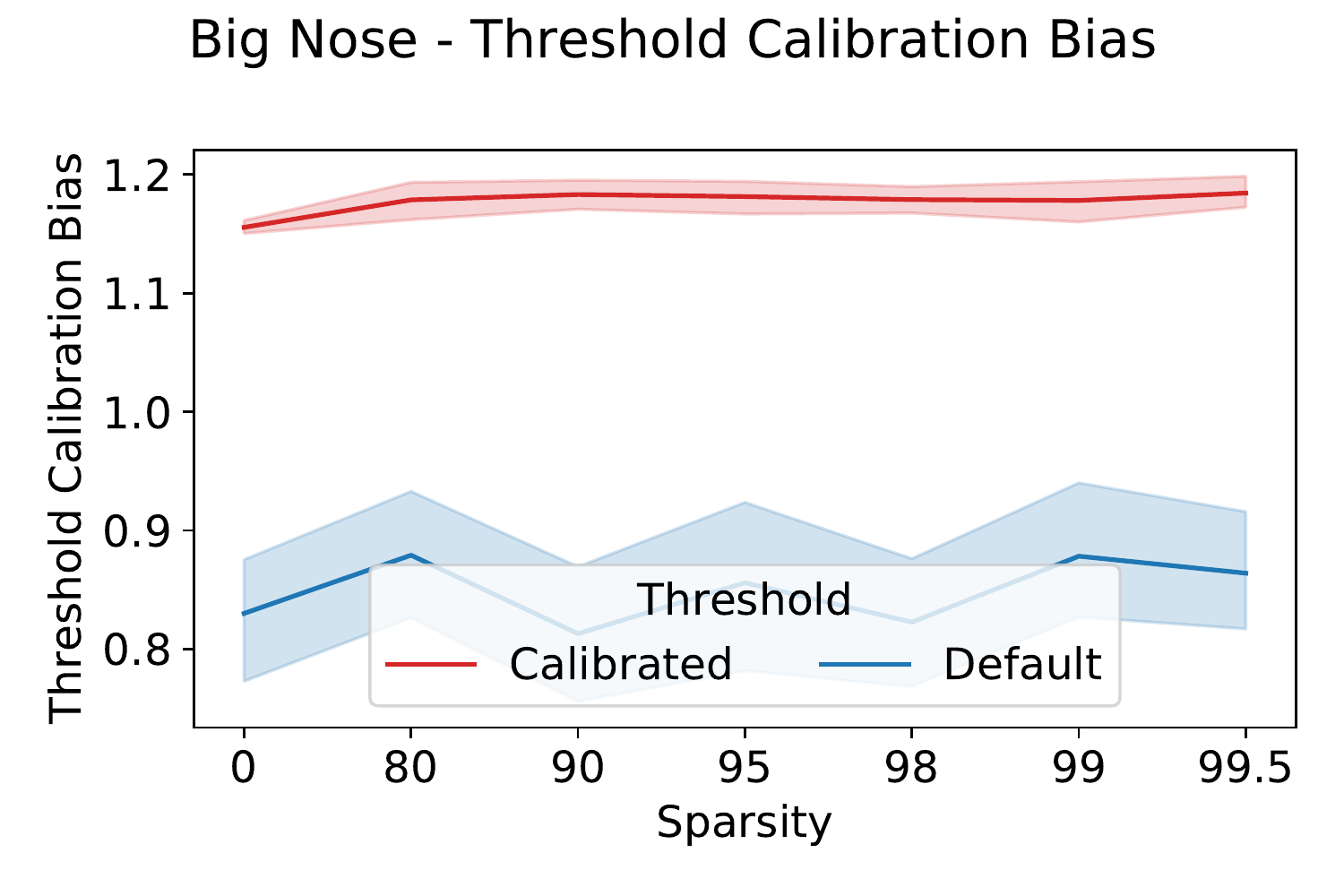} &
  \includegraphics[width=0.12\textwidth]
  {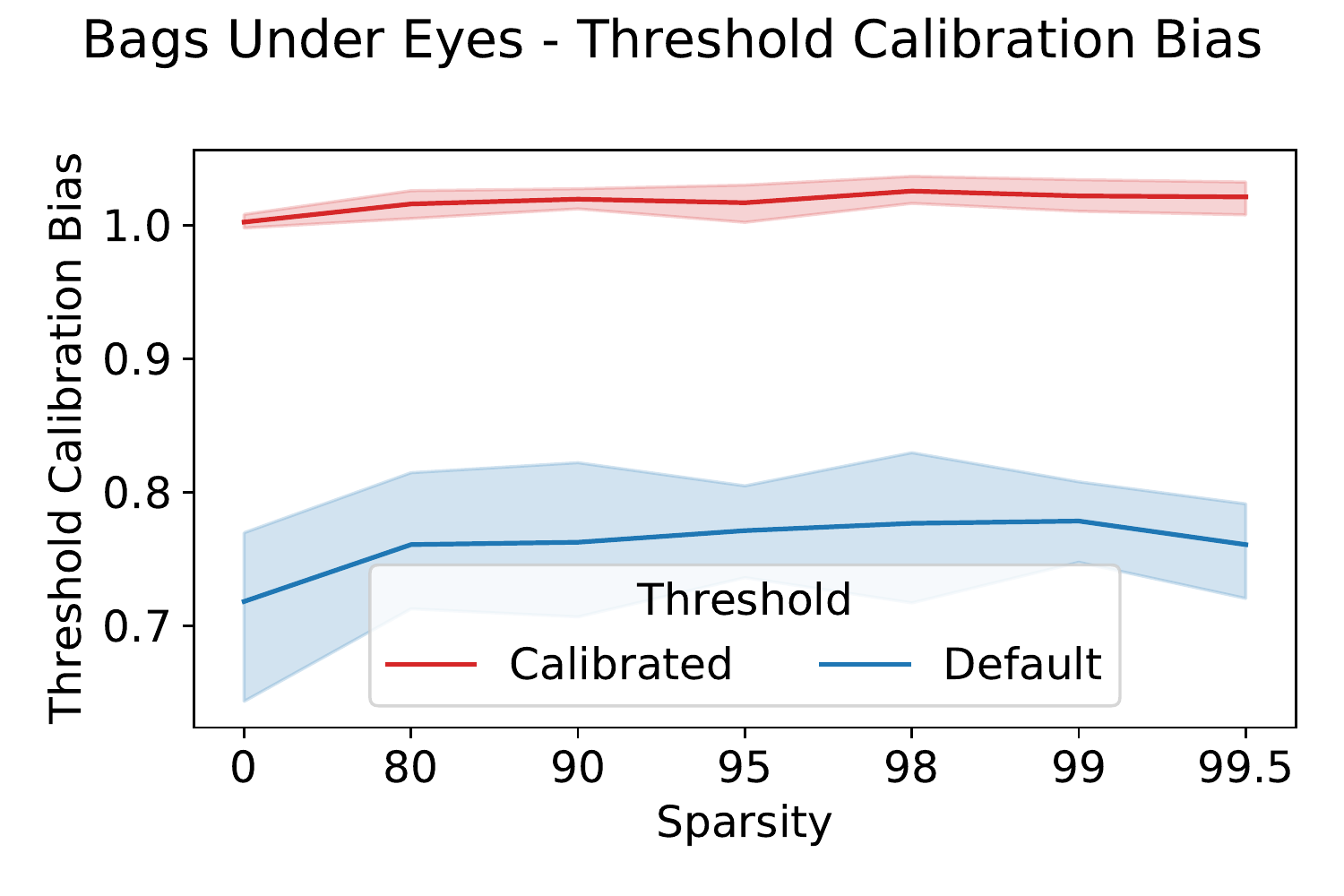} &
  \includegraphics[width=0.12\textwidth]
  {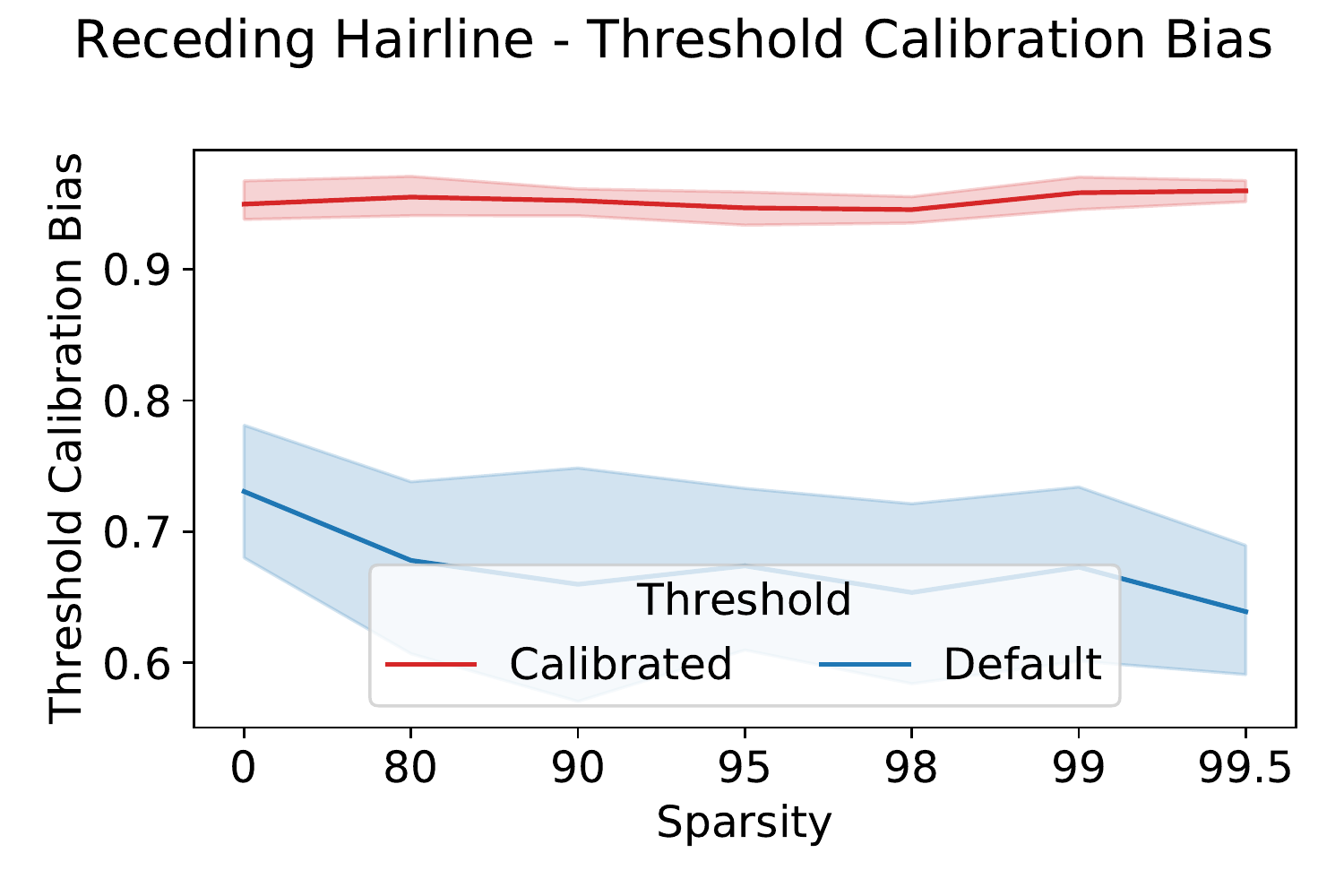} &
  \includegraphics[width=0.12\textwidth]
  {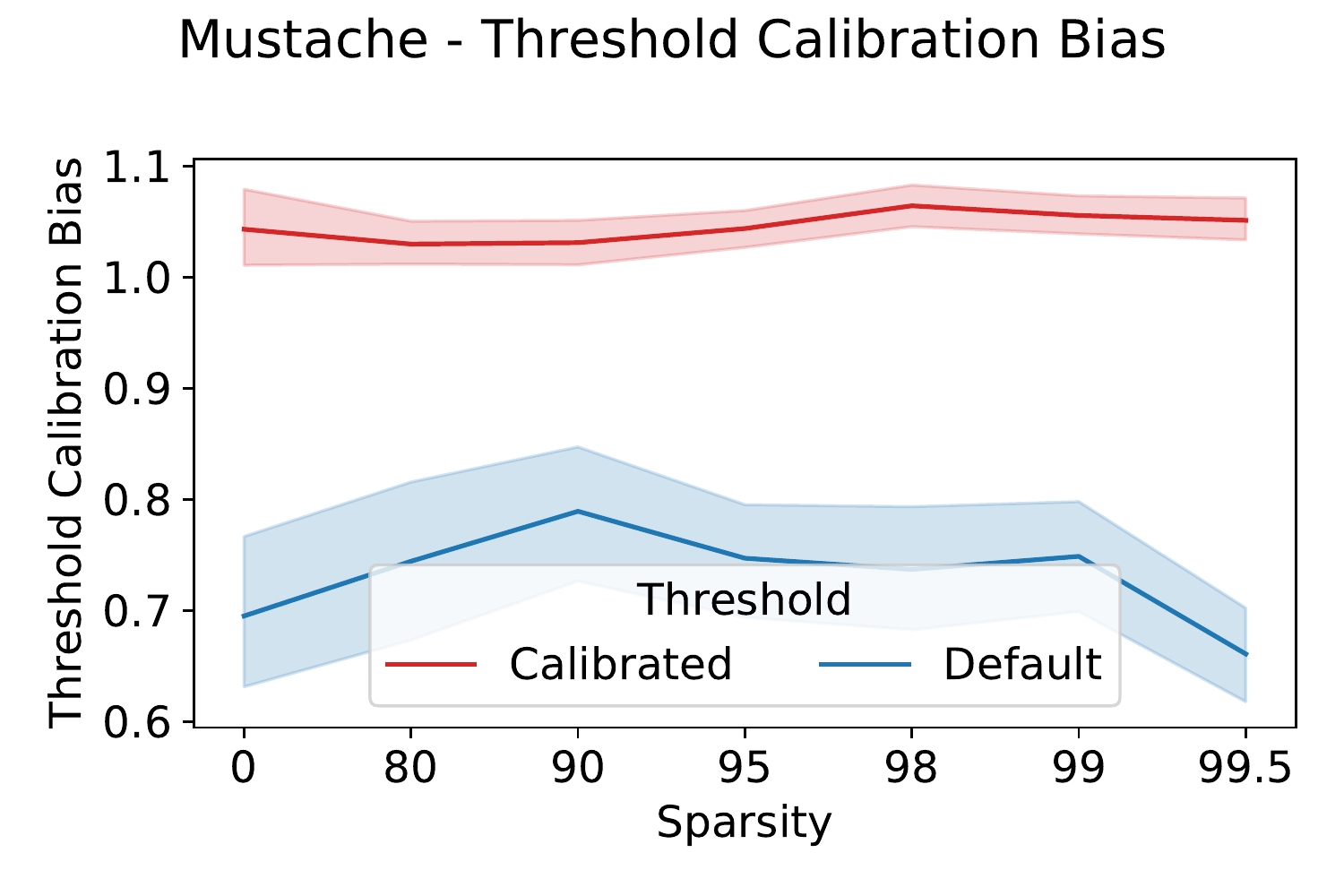} &
  \includegraphics[width=0.12\textwidth]
  {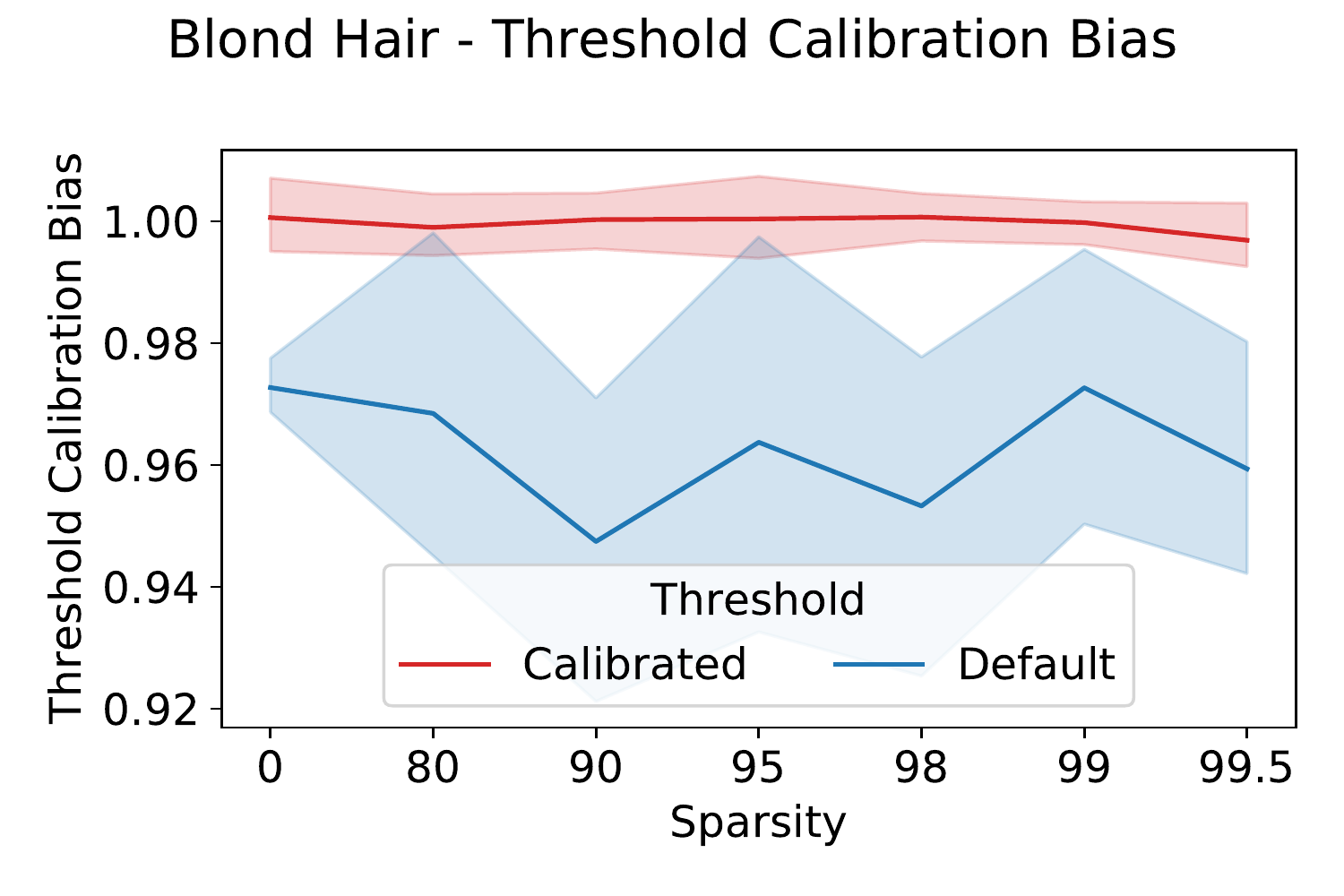} &
  \includegraphics[width=0.12\textwidth]
  {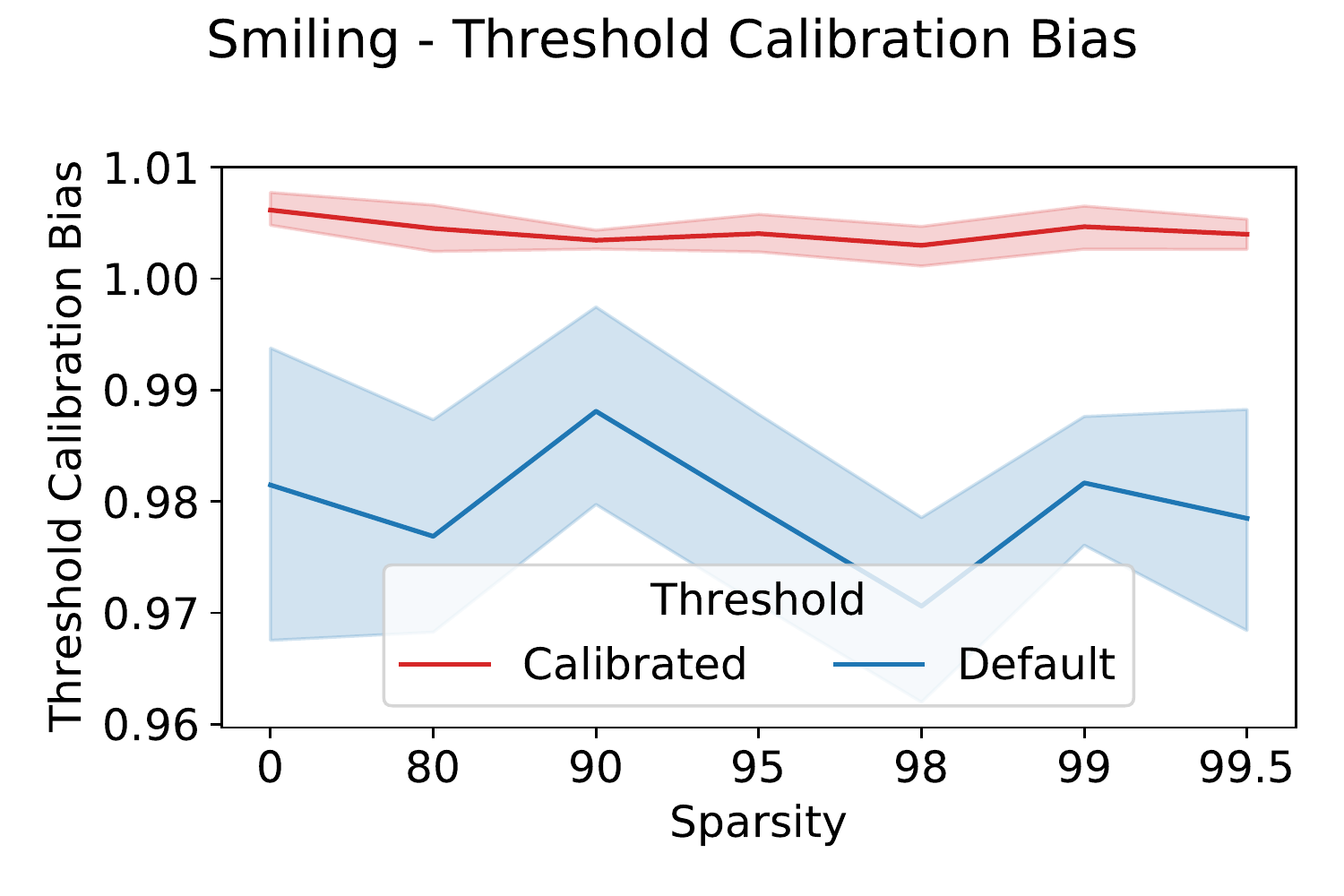}
  \\
  \includegraphics[width=0.12\textwidth]
  {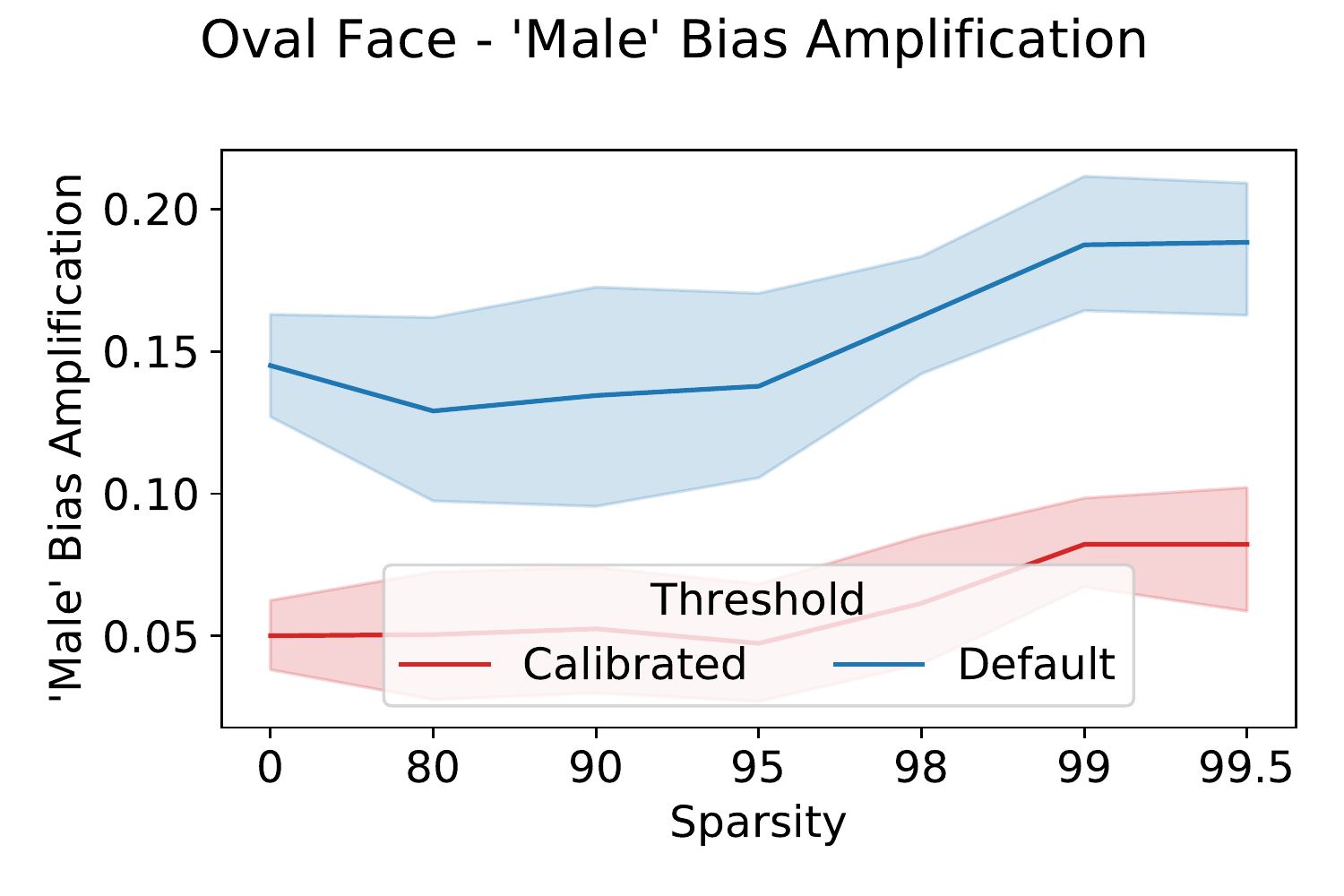} &
\includegraphics[width=0.12\textwidth]{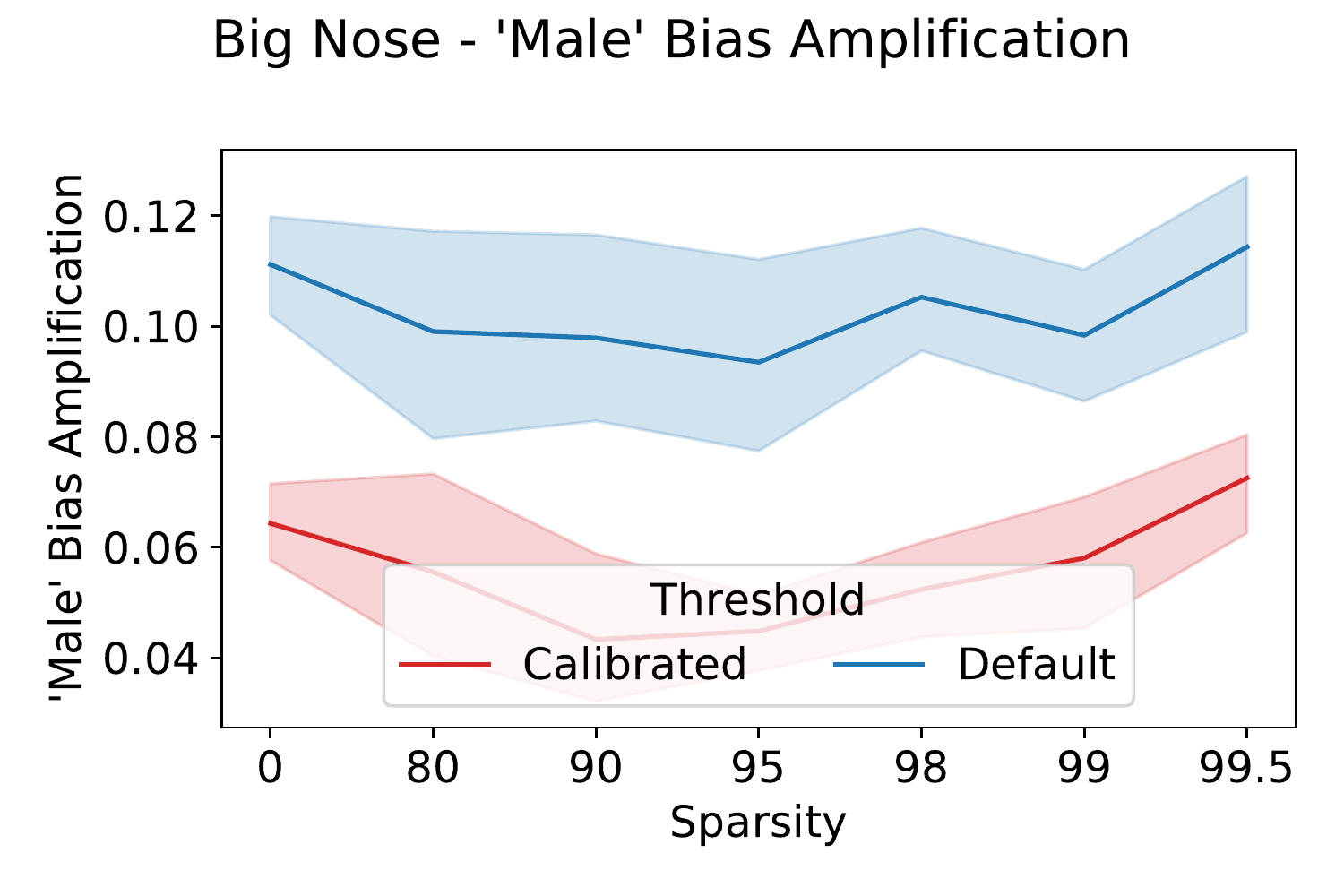} &
\includegraphics[width=0.12\textwidth]{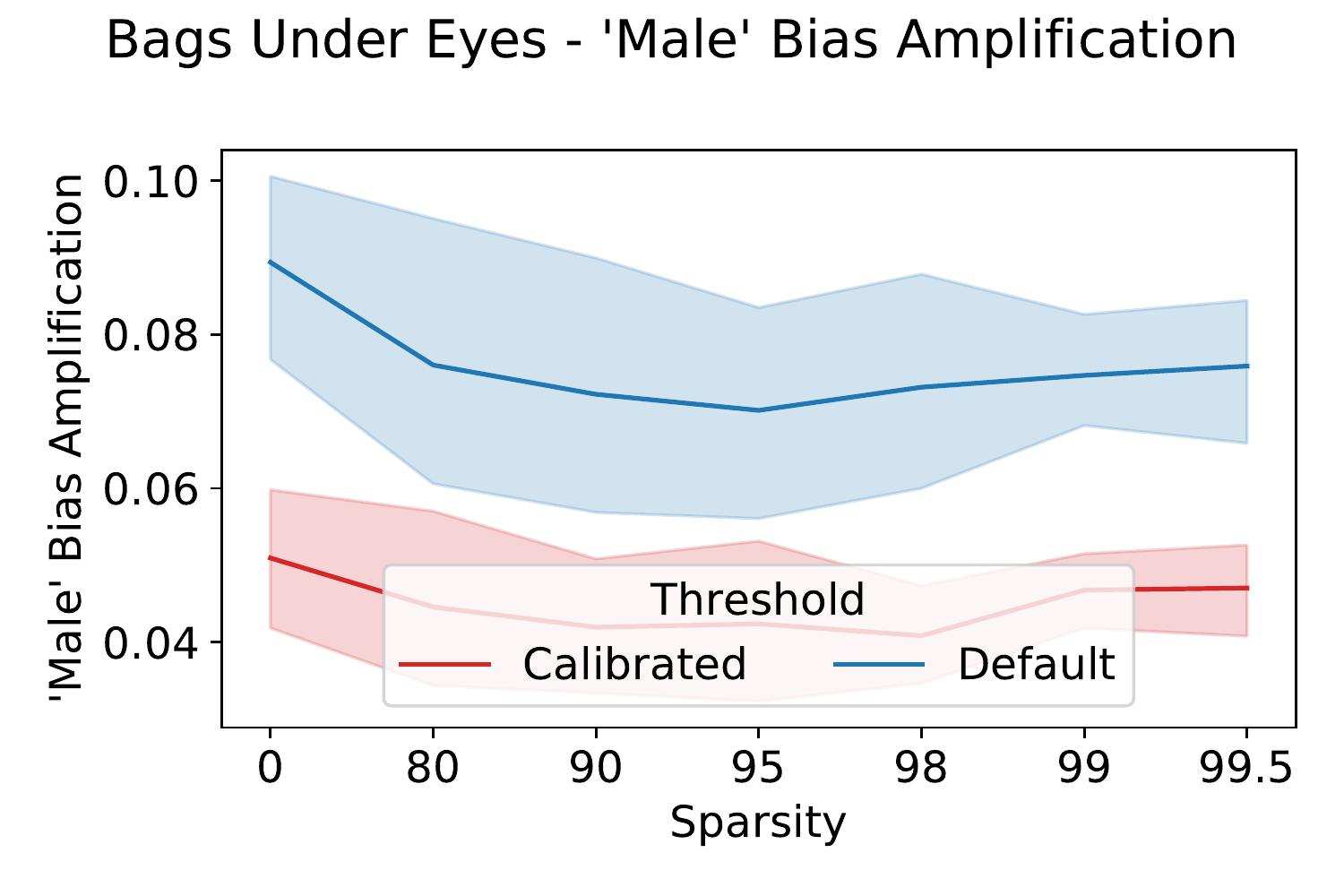} &
\includegraphics[width=0.12\textwidth]{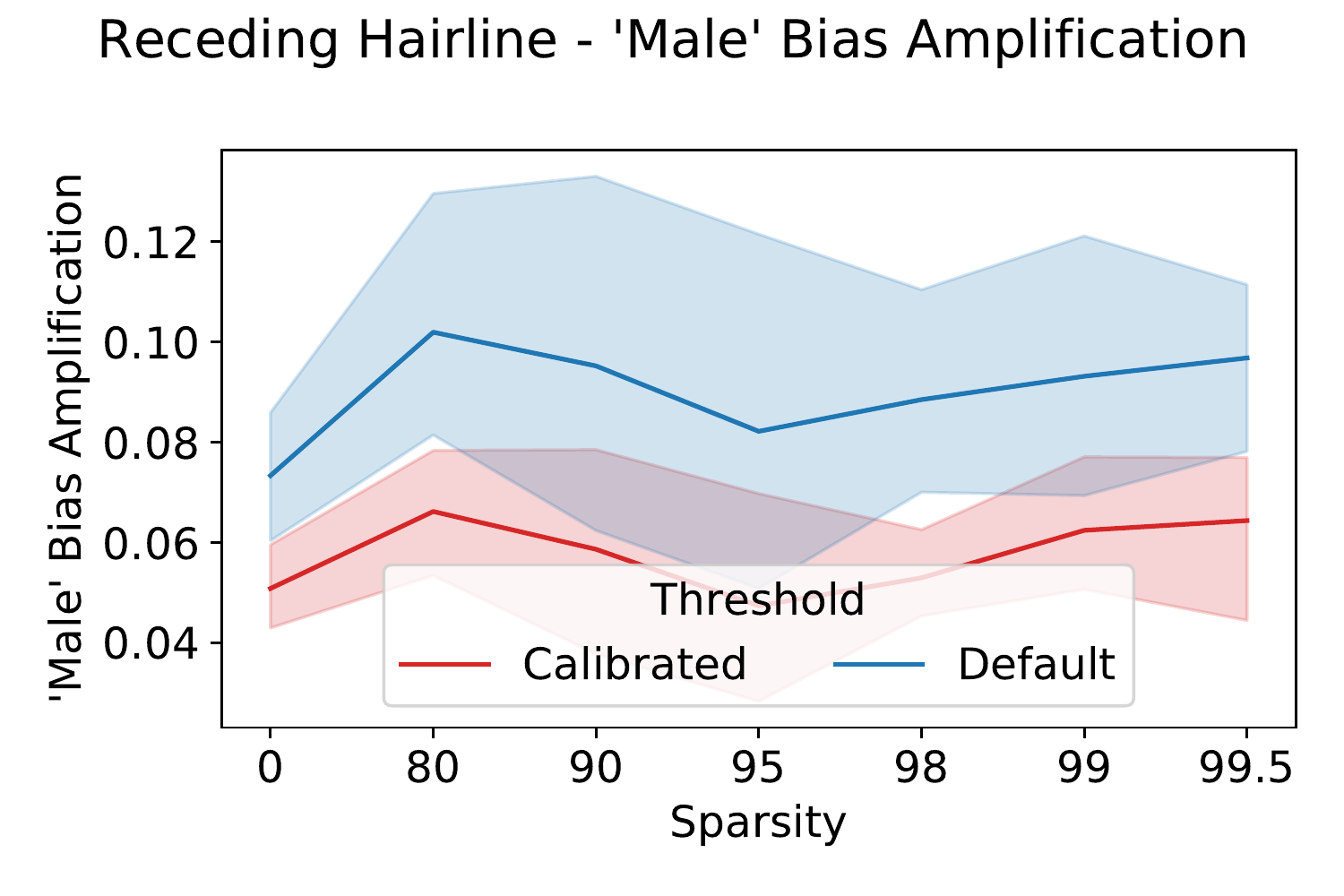} &
&%
\includegraphics[width=0.12\textwidth]{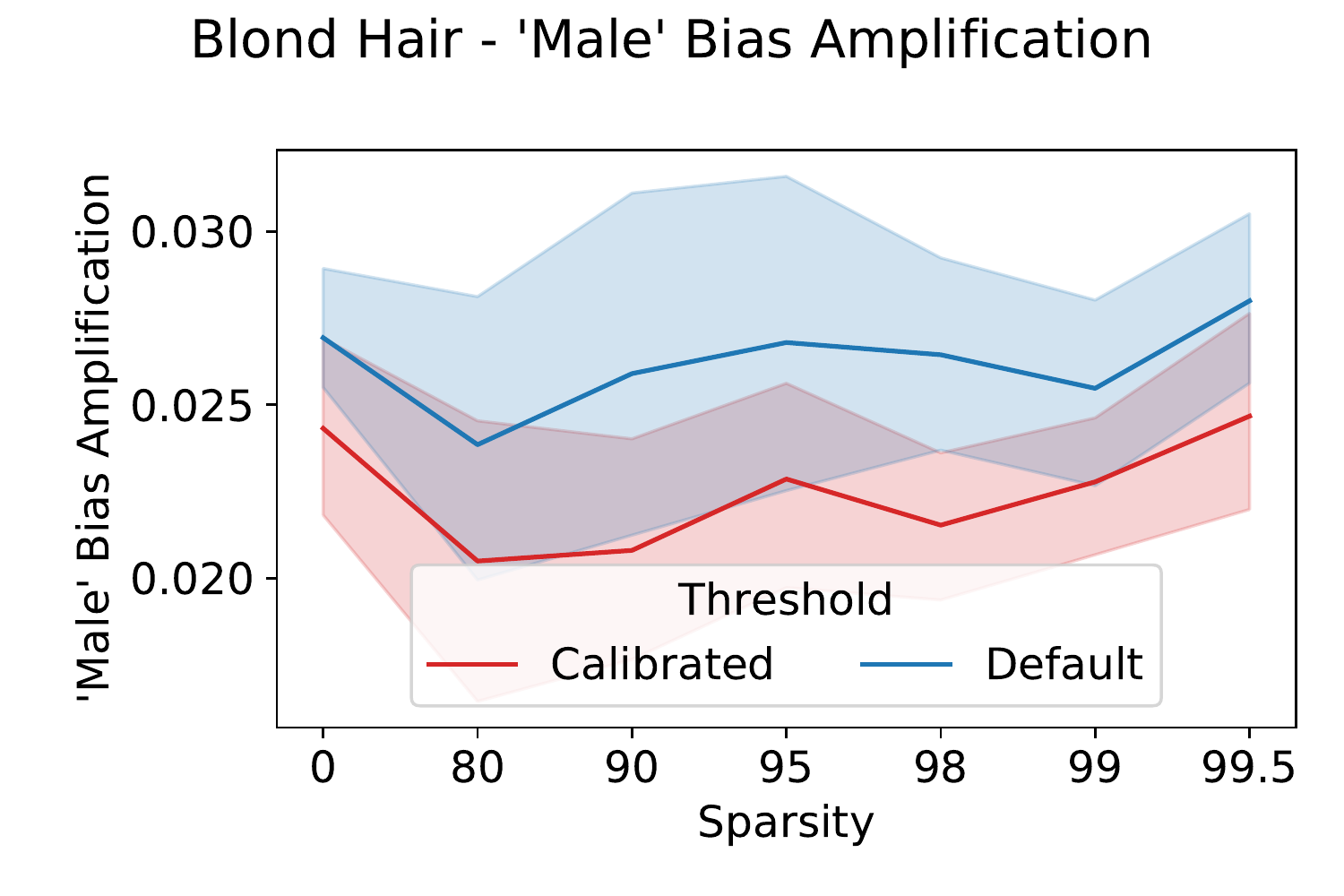} &
\includegraphics[width=0.12\textwidth]{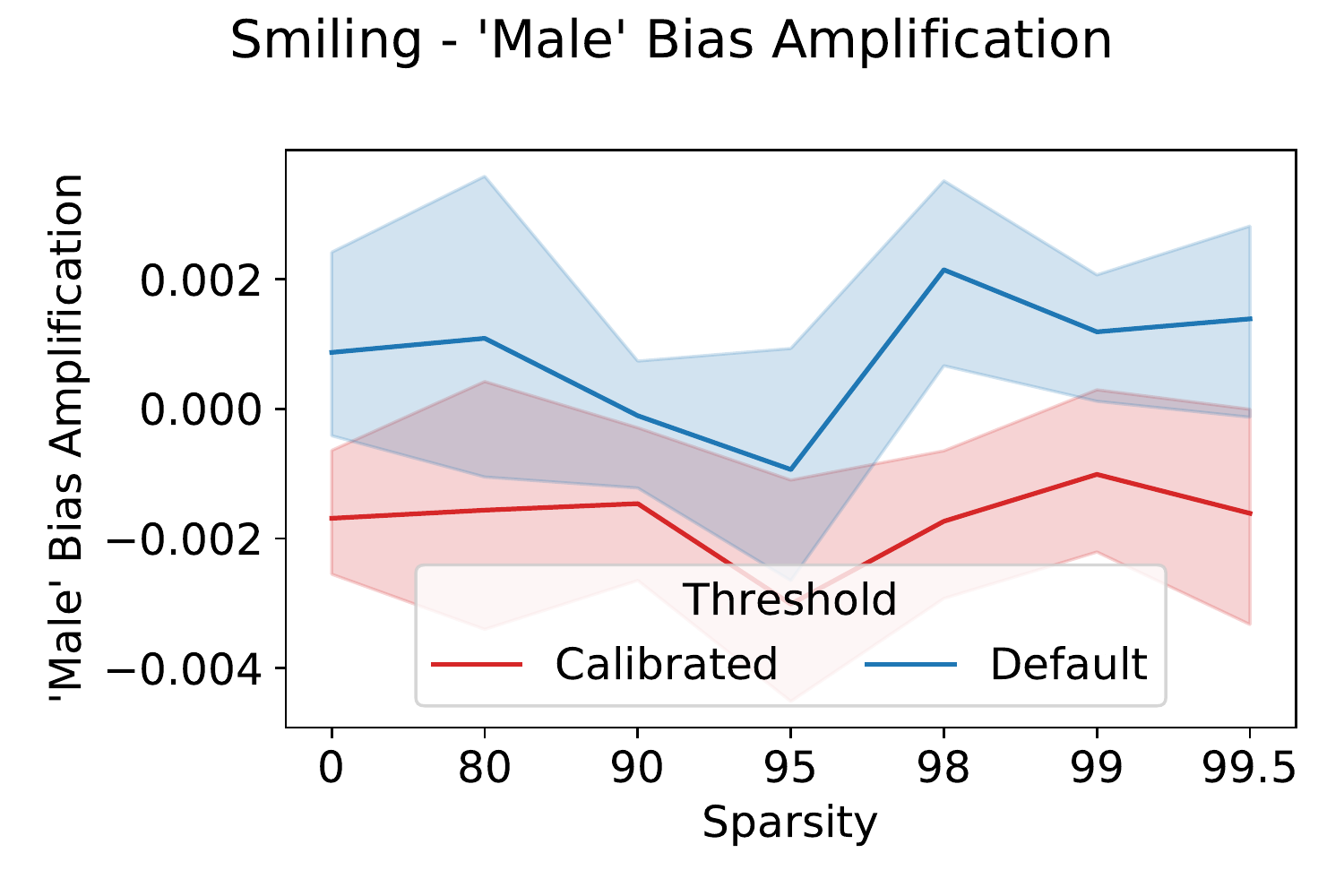}
    \\
  \includegraphics[width=0.12\textwidth]
  {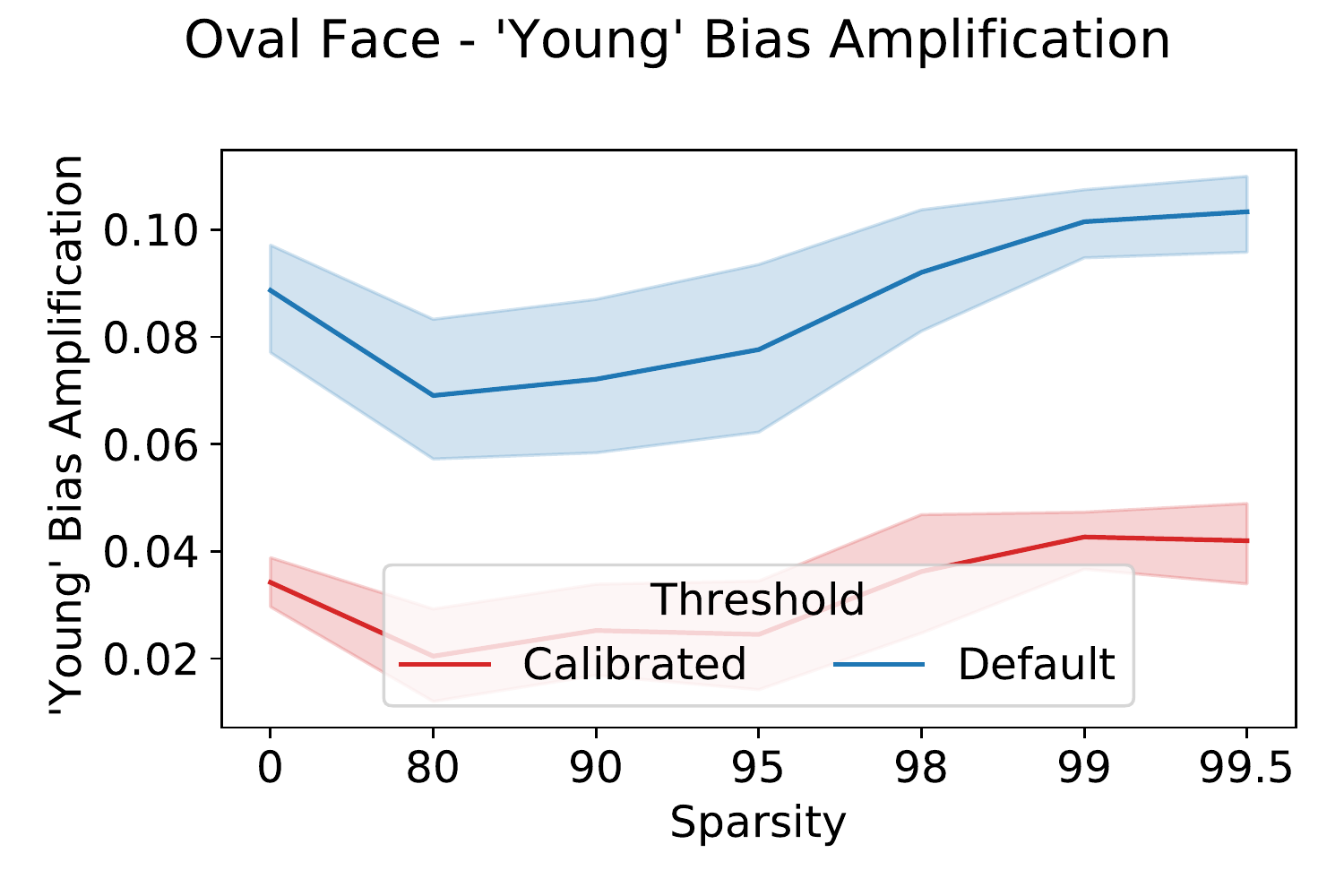} &
\includegraphics[width=0.12\textwidth]{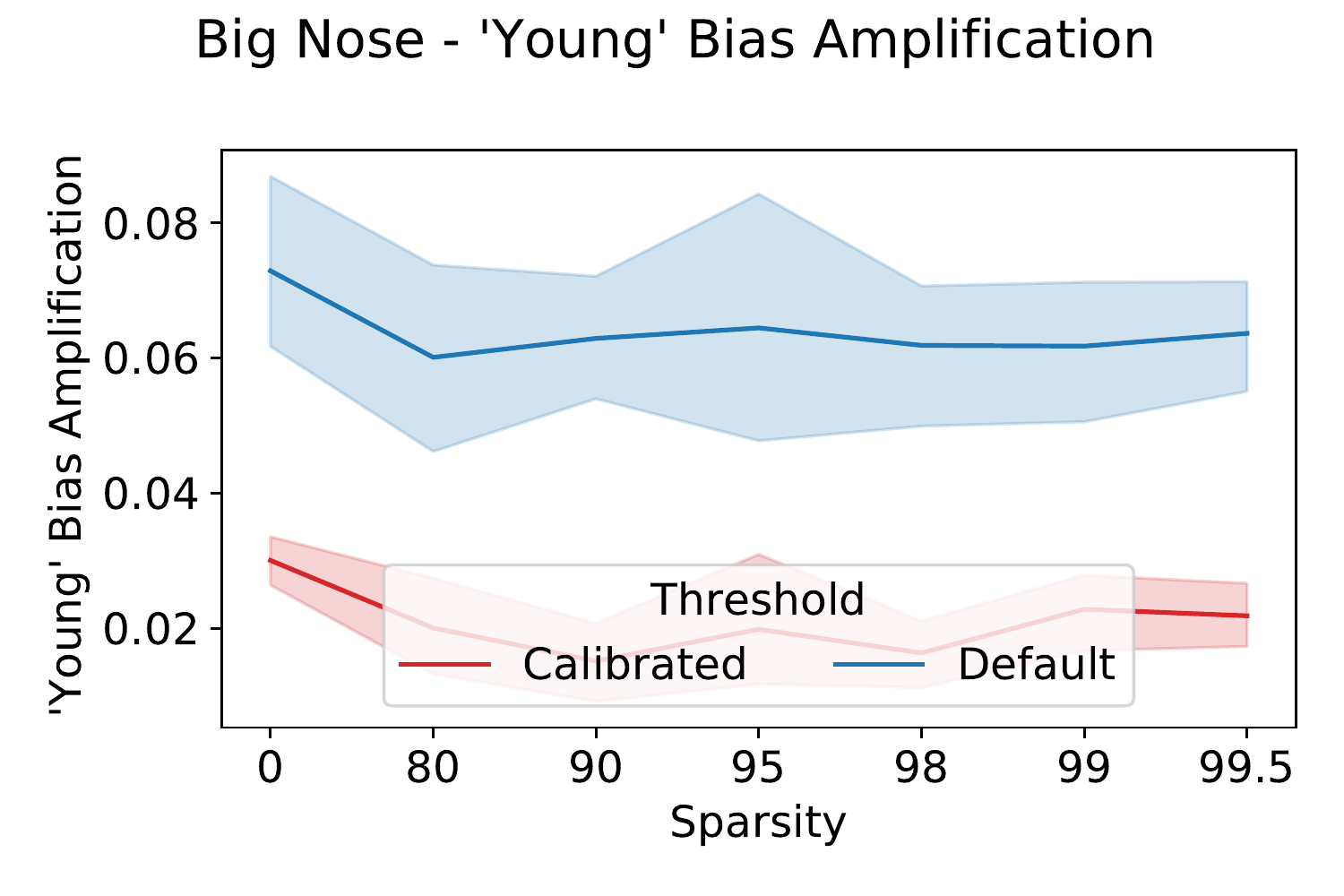} &
\includegraphics[width=0.12\textwidth]{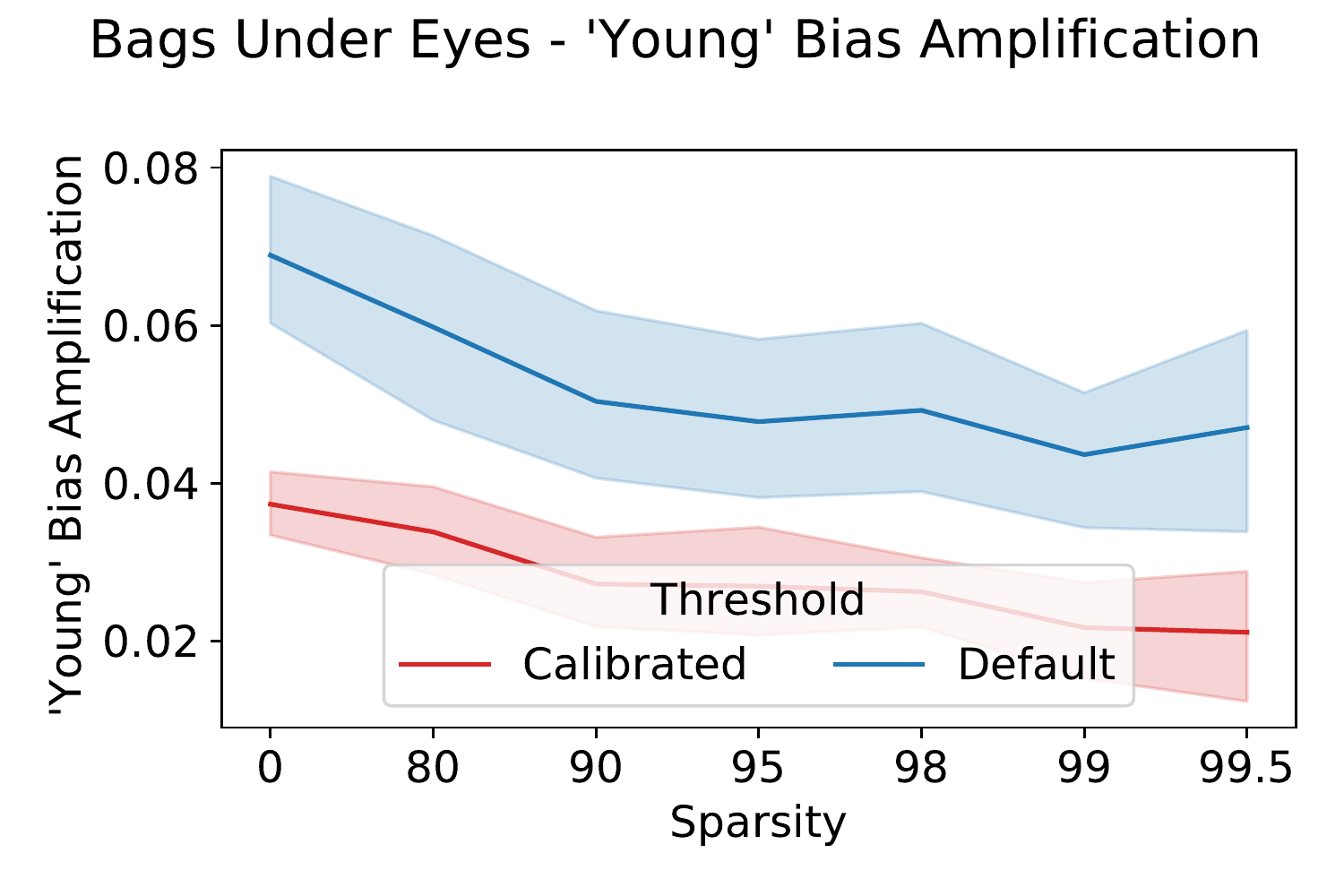} &
\includegraphics[width=0.12\textwidth]{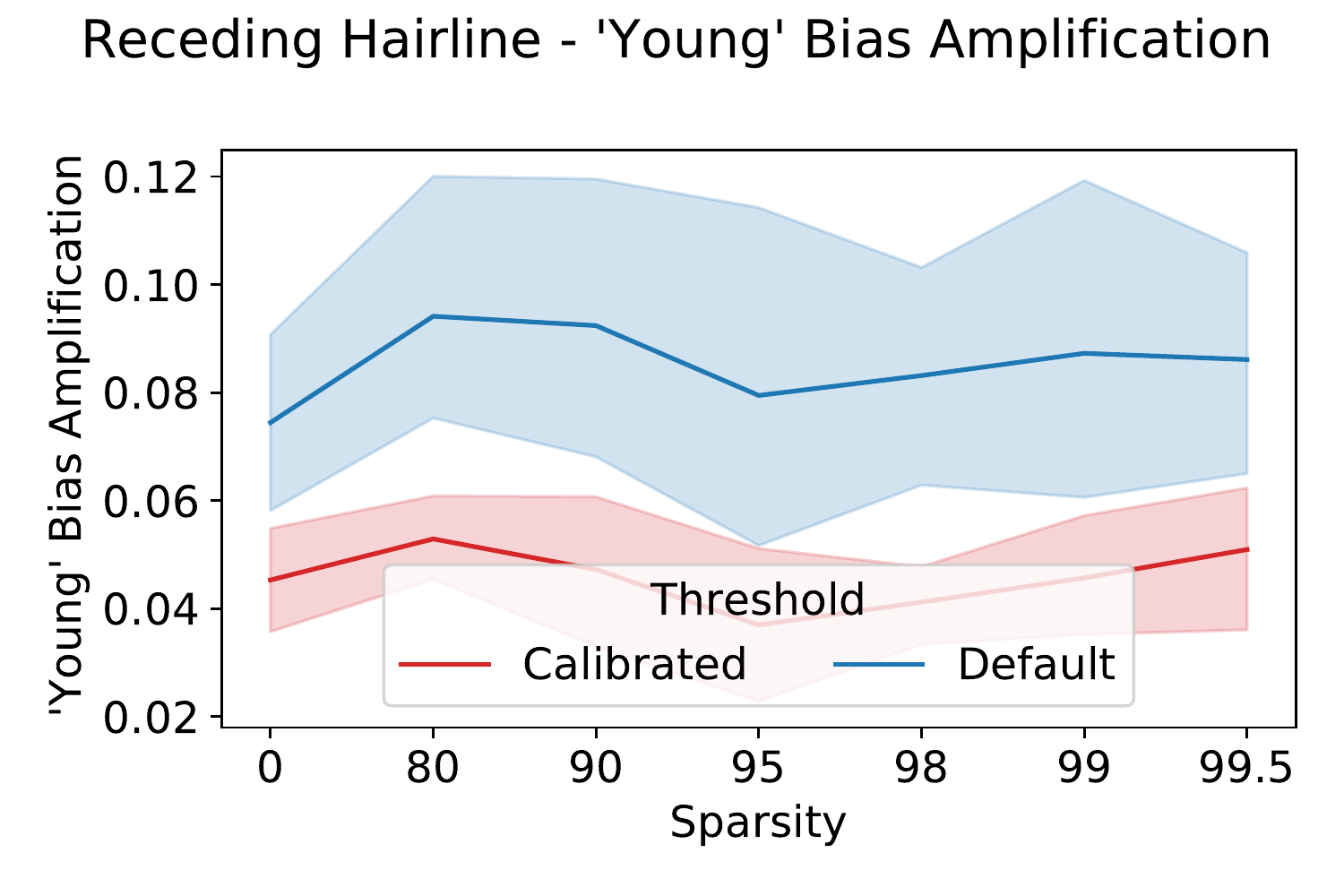} &
\includegraphics[width=0.12\textwidth]{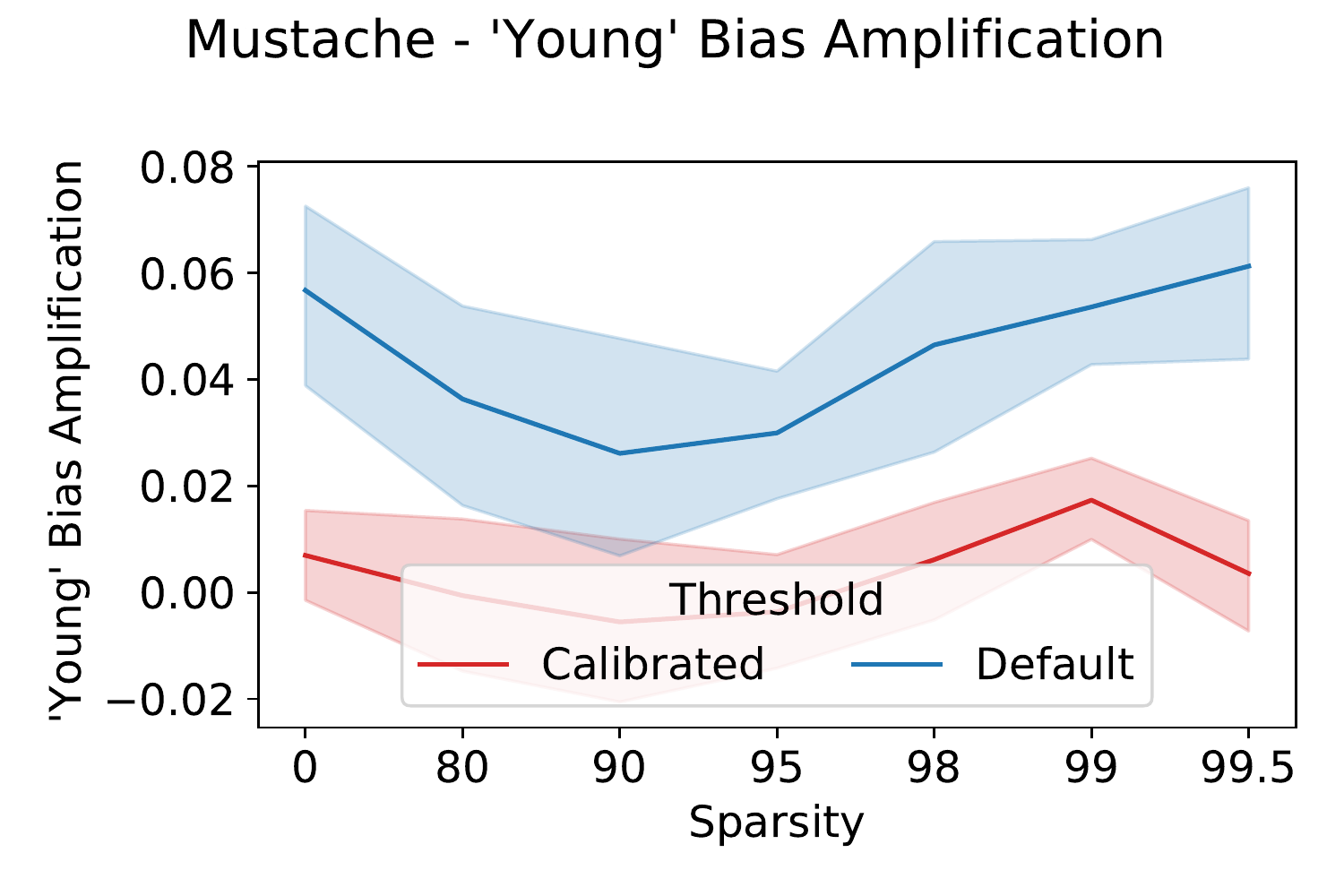} &
\includegraphics[width=0.12\textwidth]{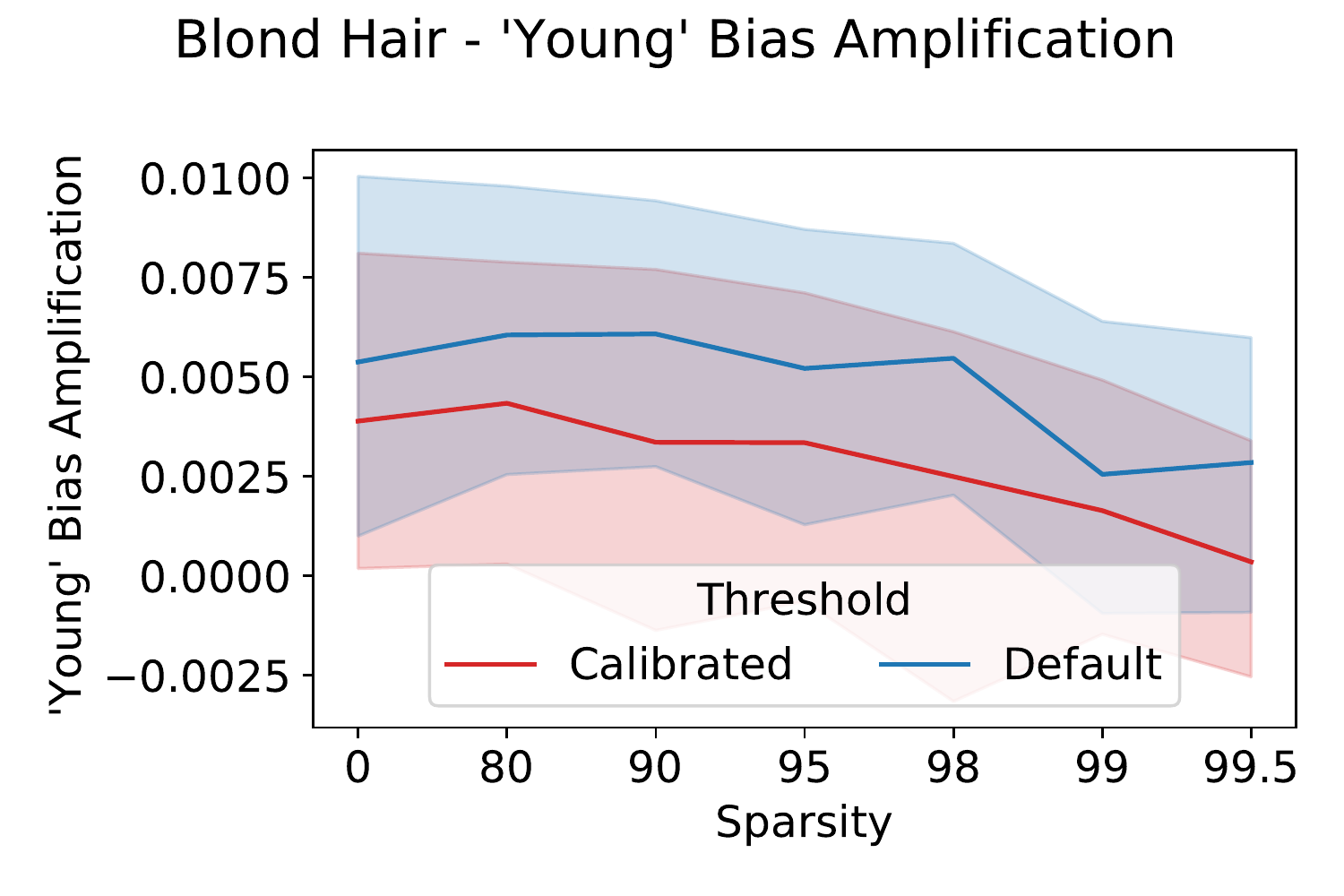} &
    \\
  \includegraphics[width=0.12\textwidth]
  {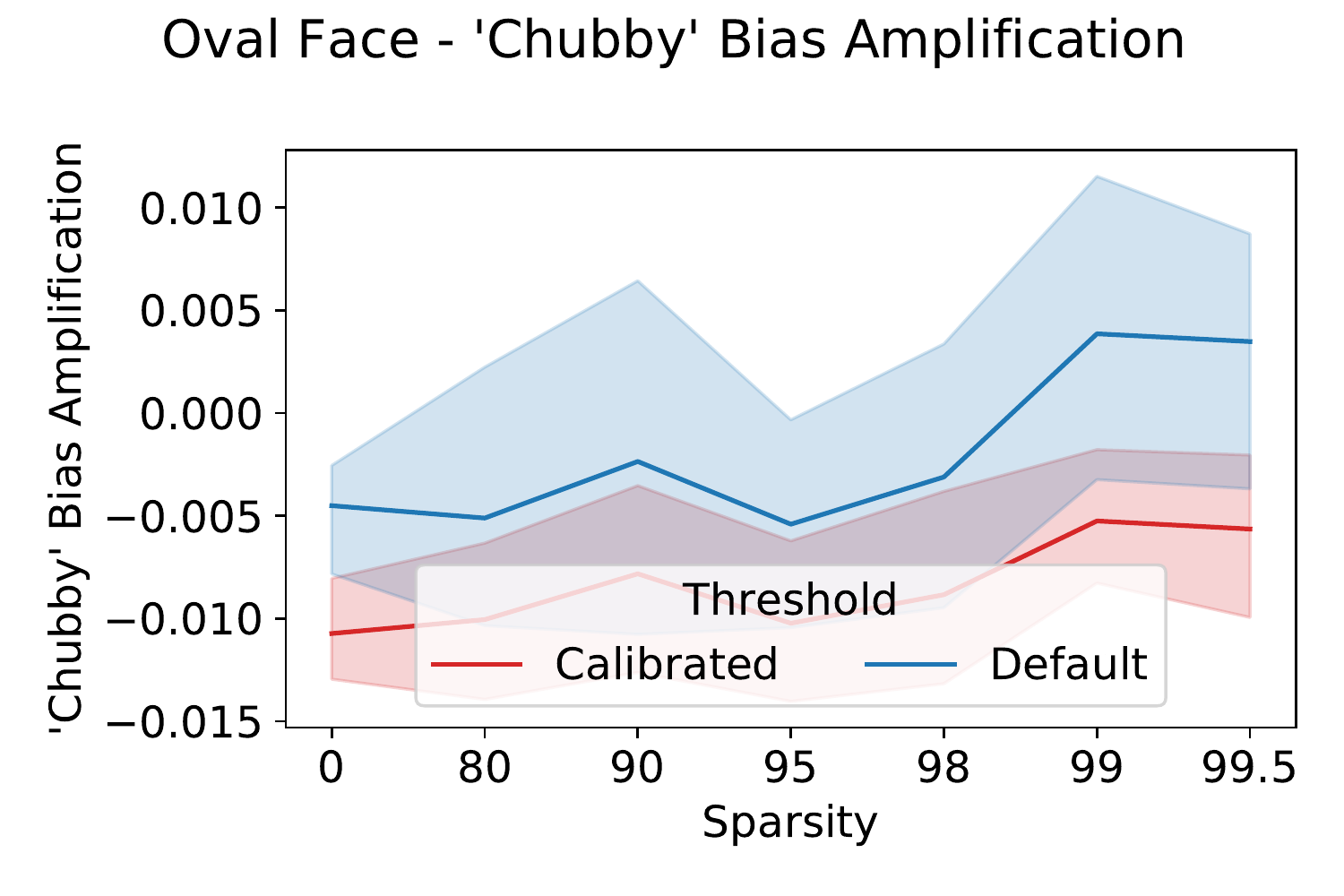} &
\includegraphics[width=0.12\textwidth]{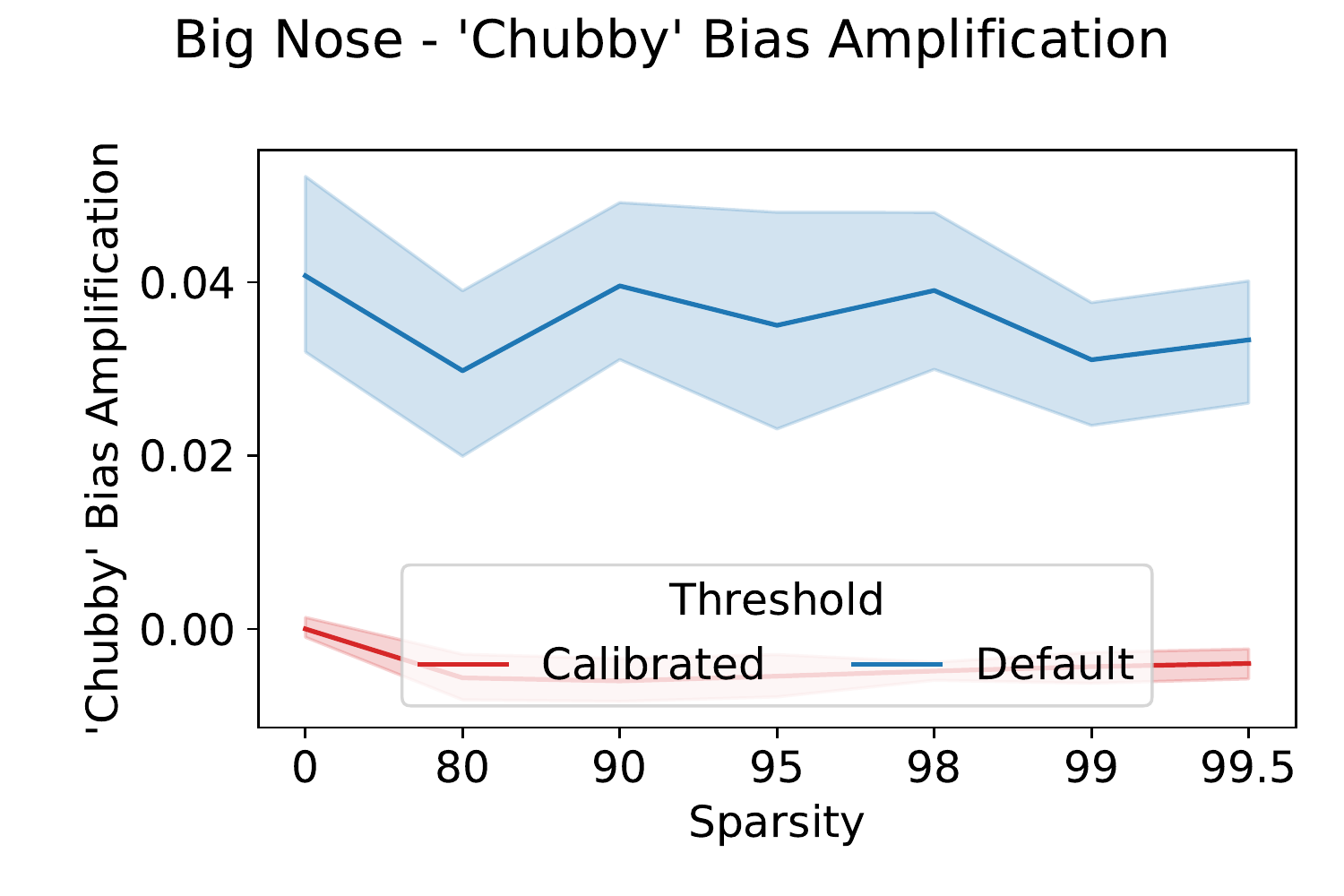} &
\includegraphics[width=0.12\textwidth]{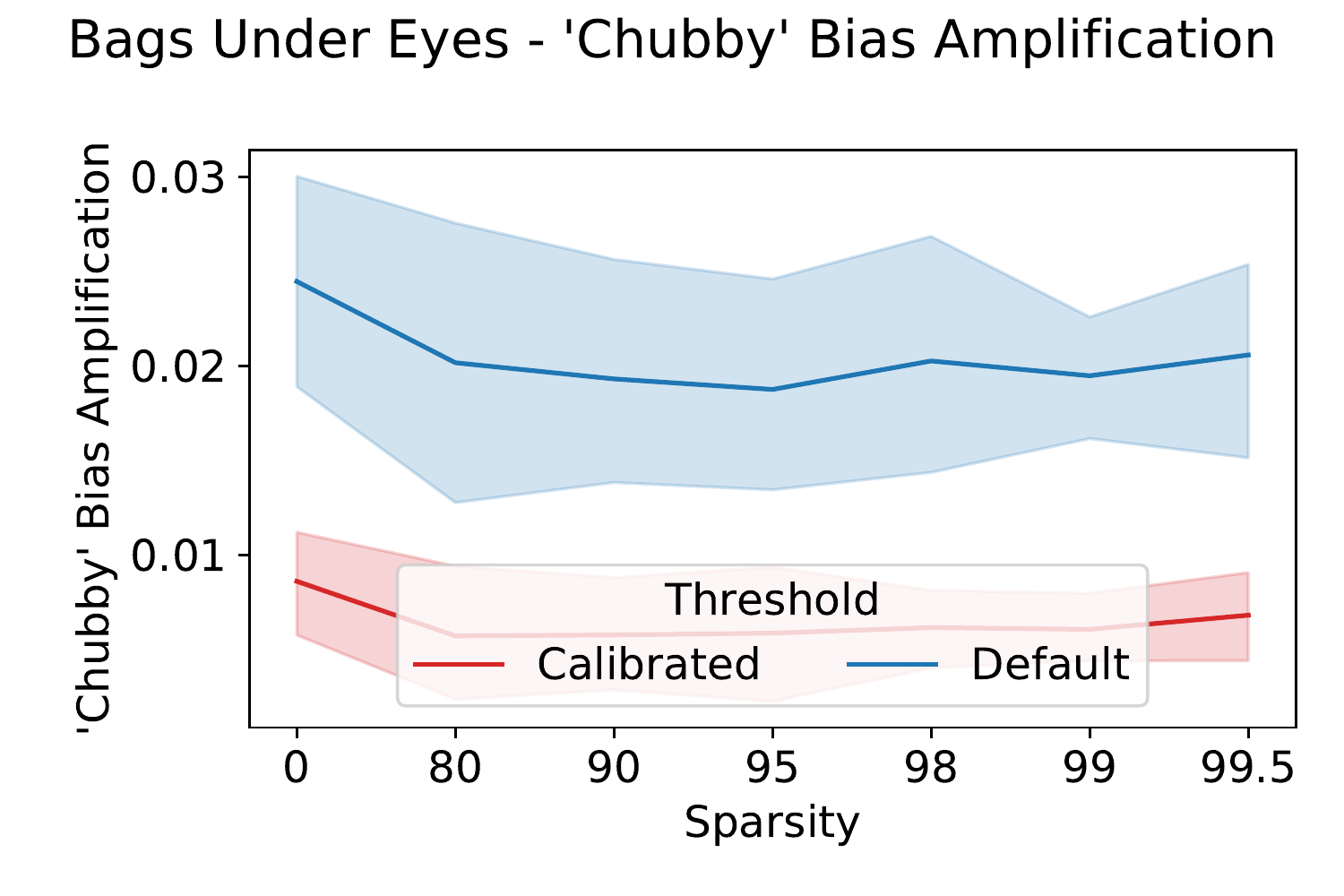} &
\includegraphics[width=0.12\textwidth]{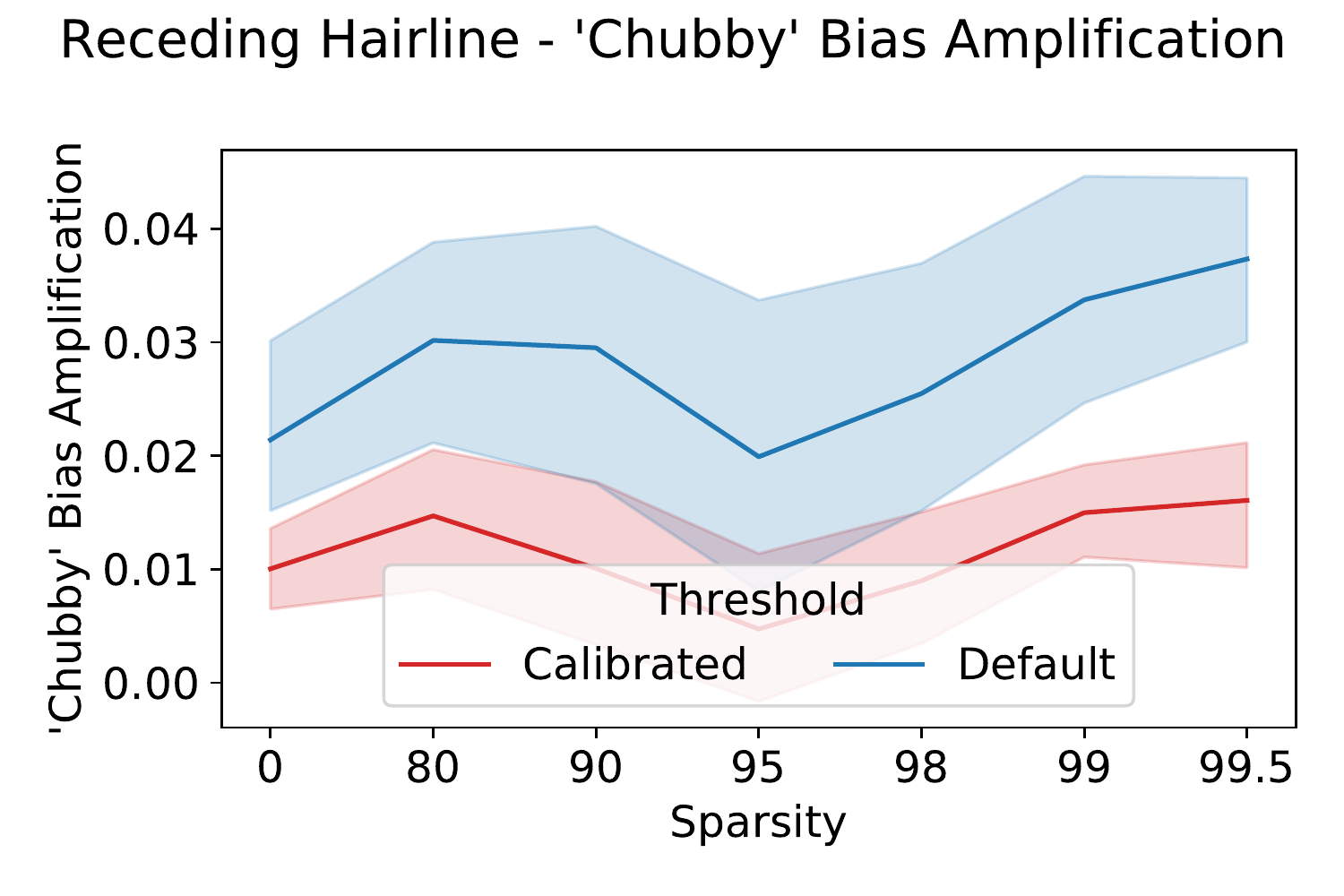} &
\includegraphics[width=0.12\textwidth]{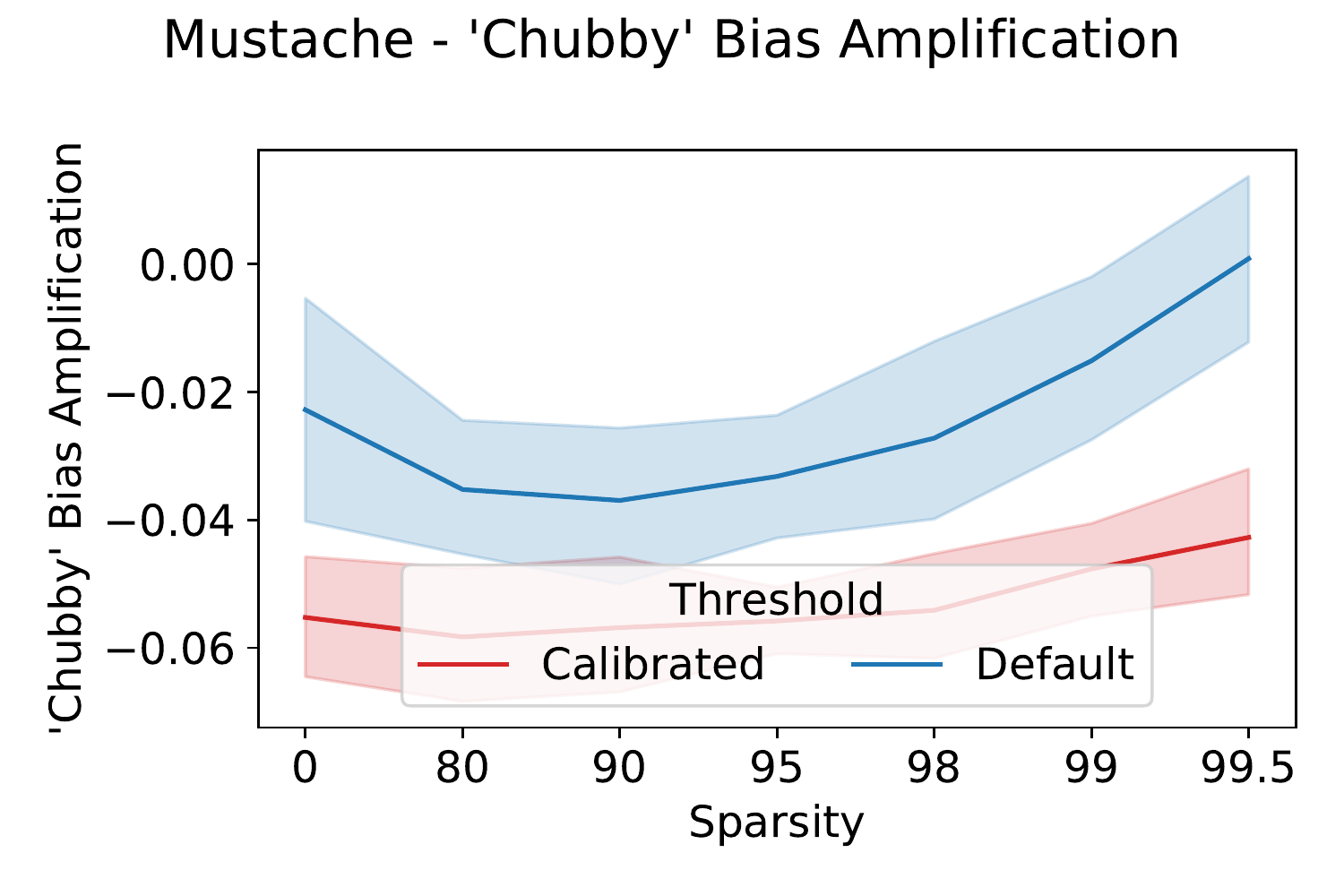} &
\includegraphics[width=0.12\textwidth]{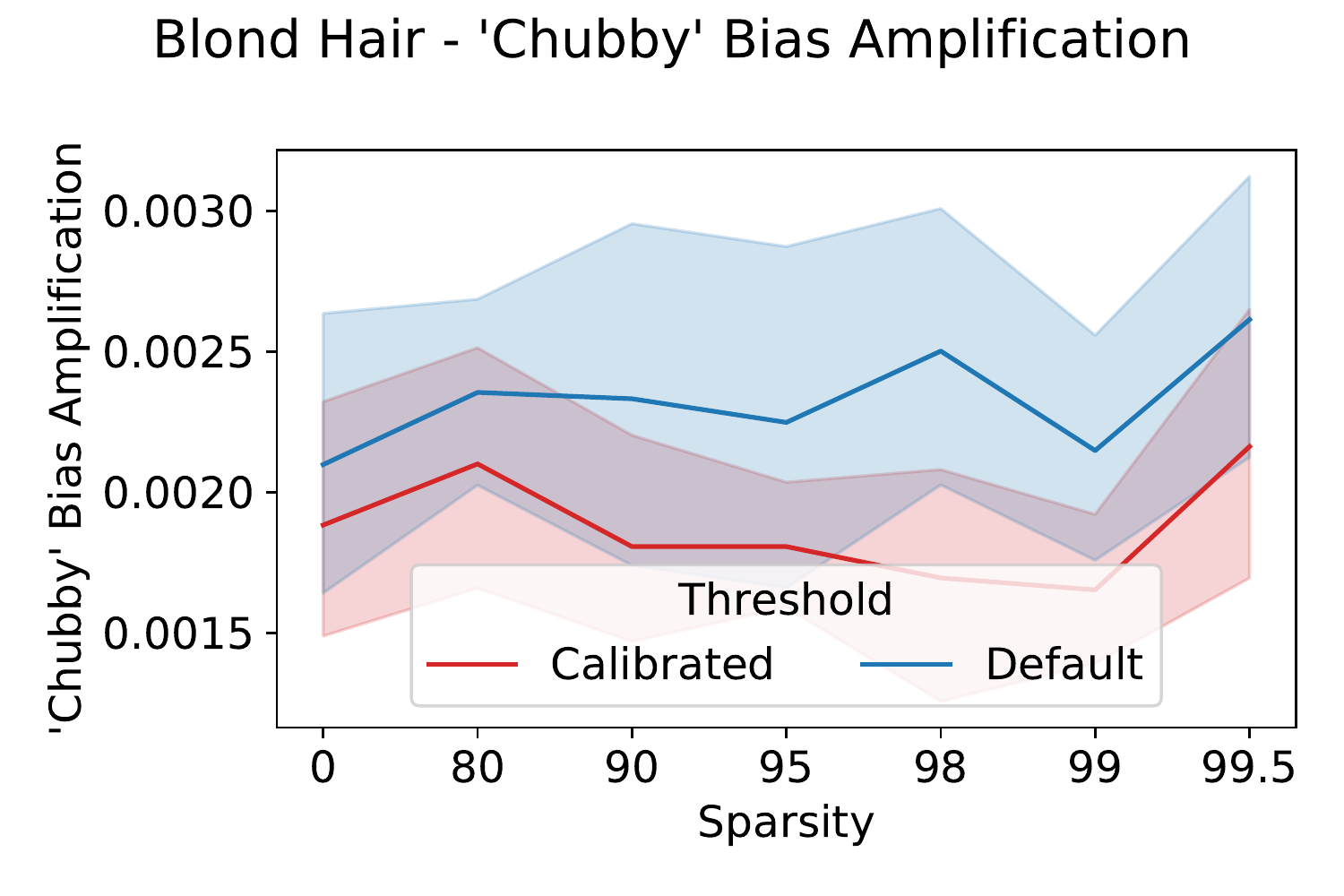} &
\includegraphics[width=0.12\textwidth]{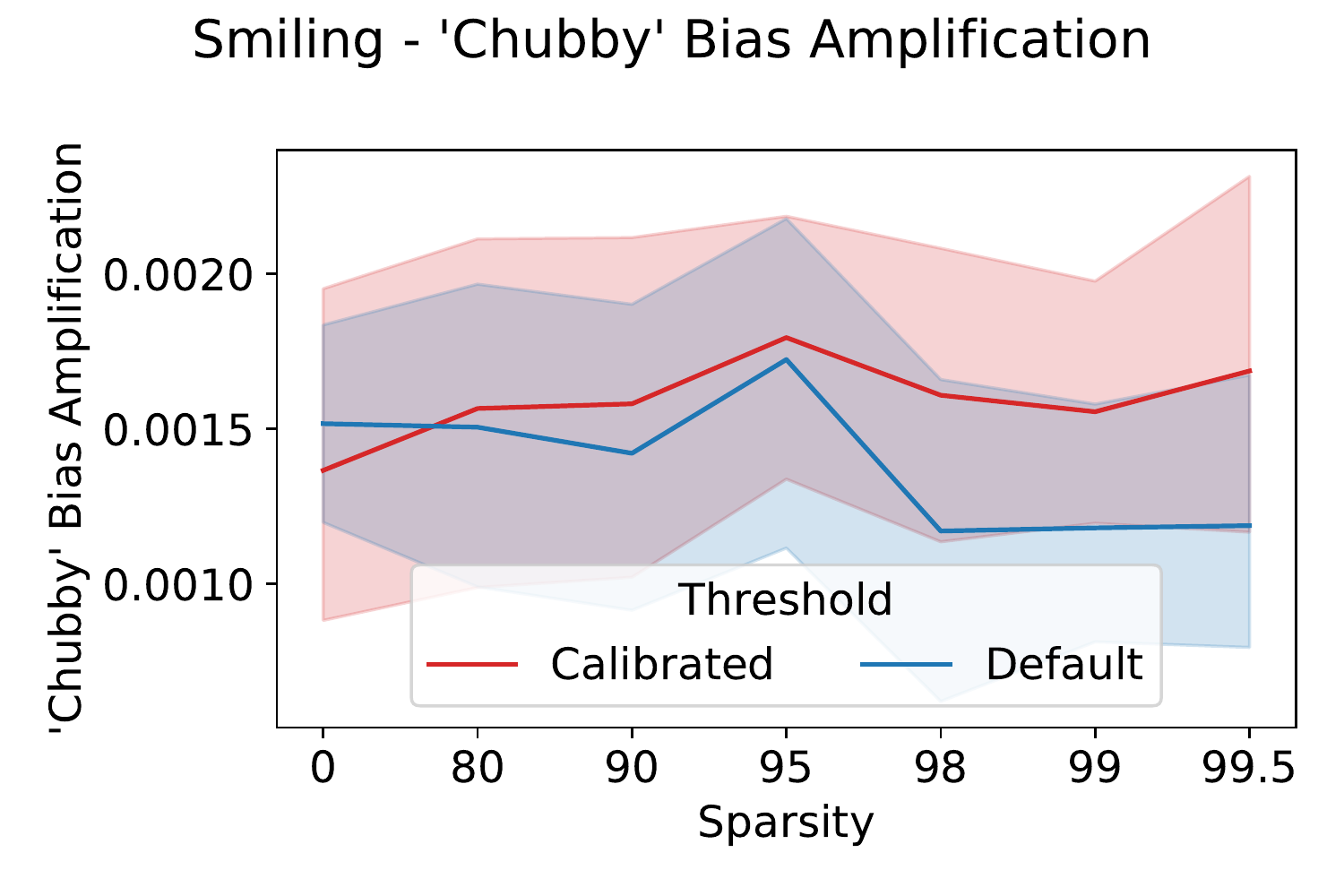}
    \\
      \includegraphics[width=0.12\textwidth]
  {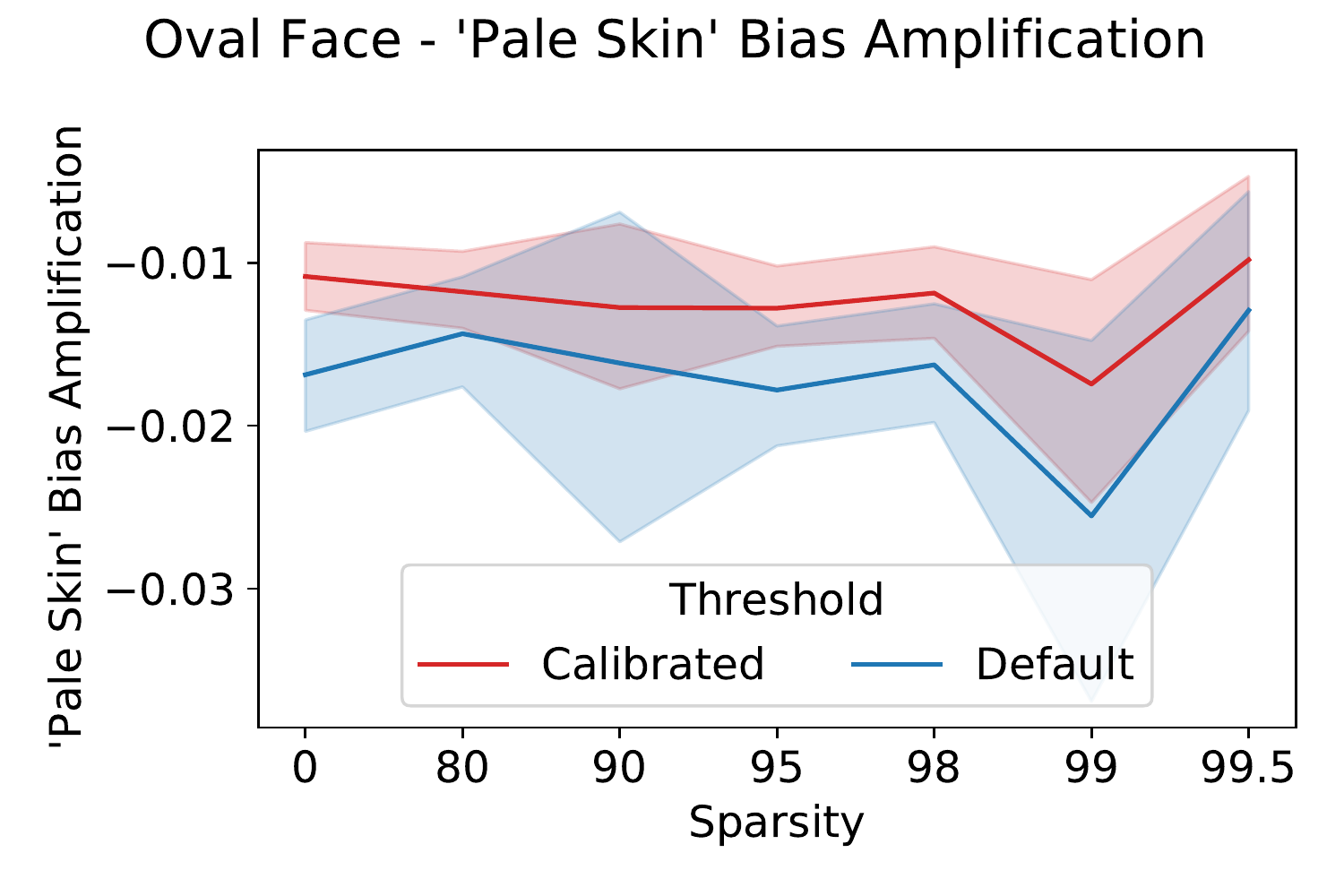} &
\includegraphics[width=0.12\textwidth]{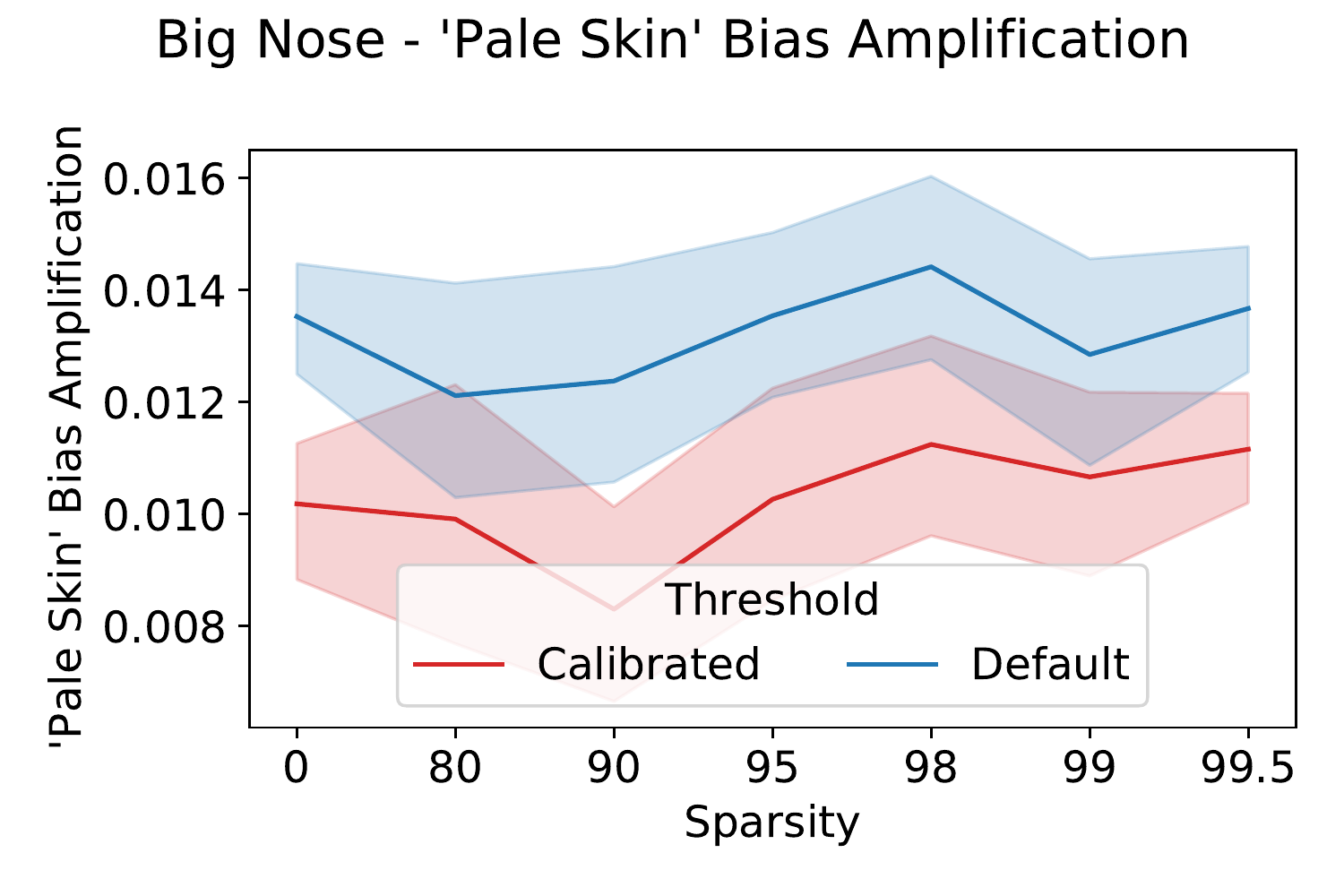} &
\includegraphics[width=0.12\textwidth]{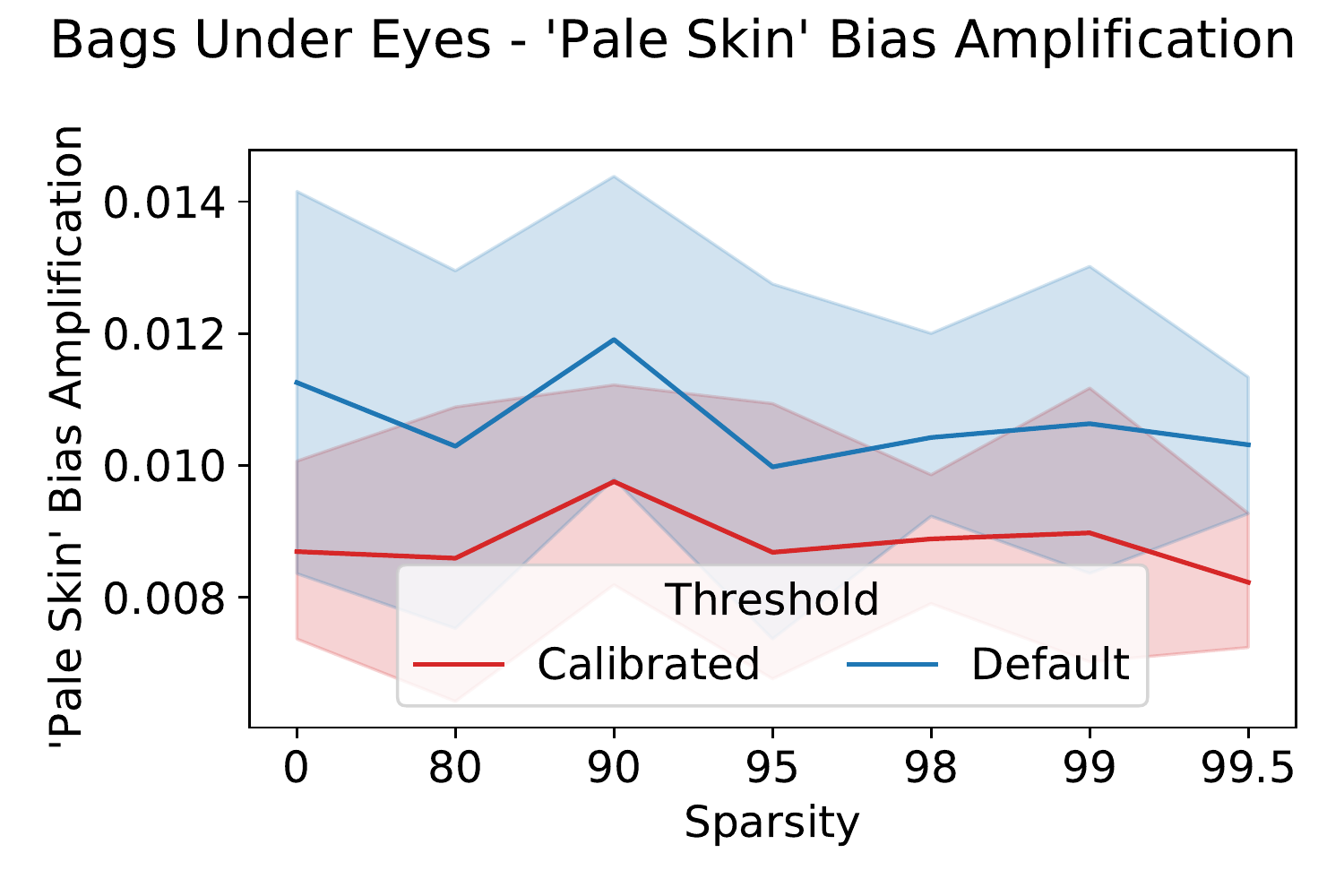} &
\includegraphics[width=0.12\textwidth]{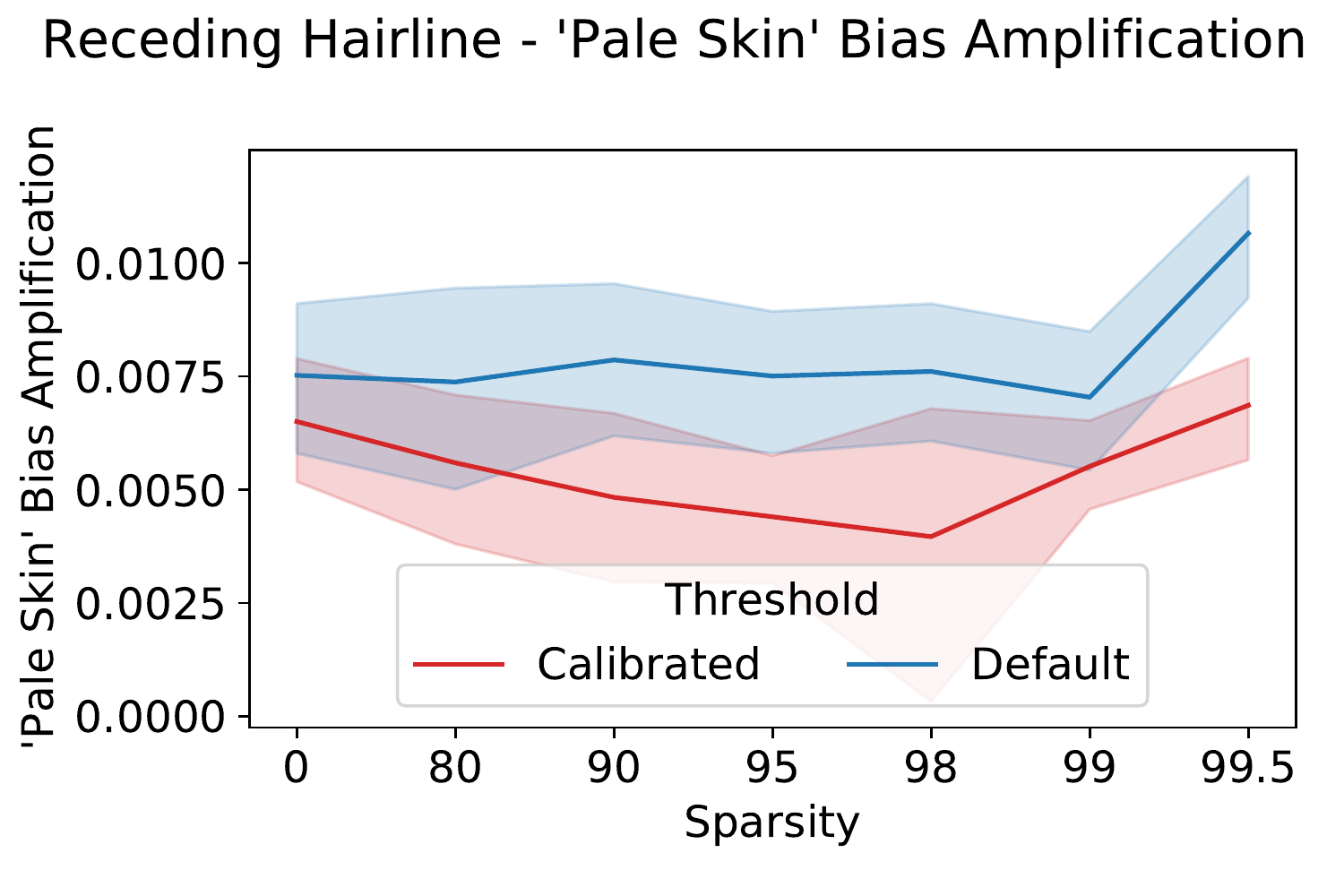} &
&%
\includegraphics[width=0.12\textwidth]{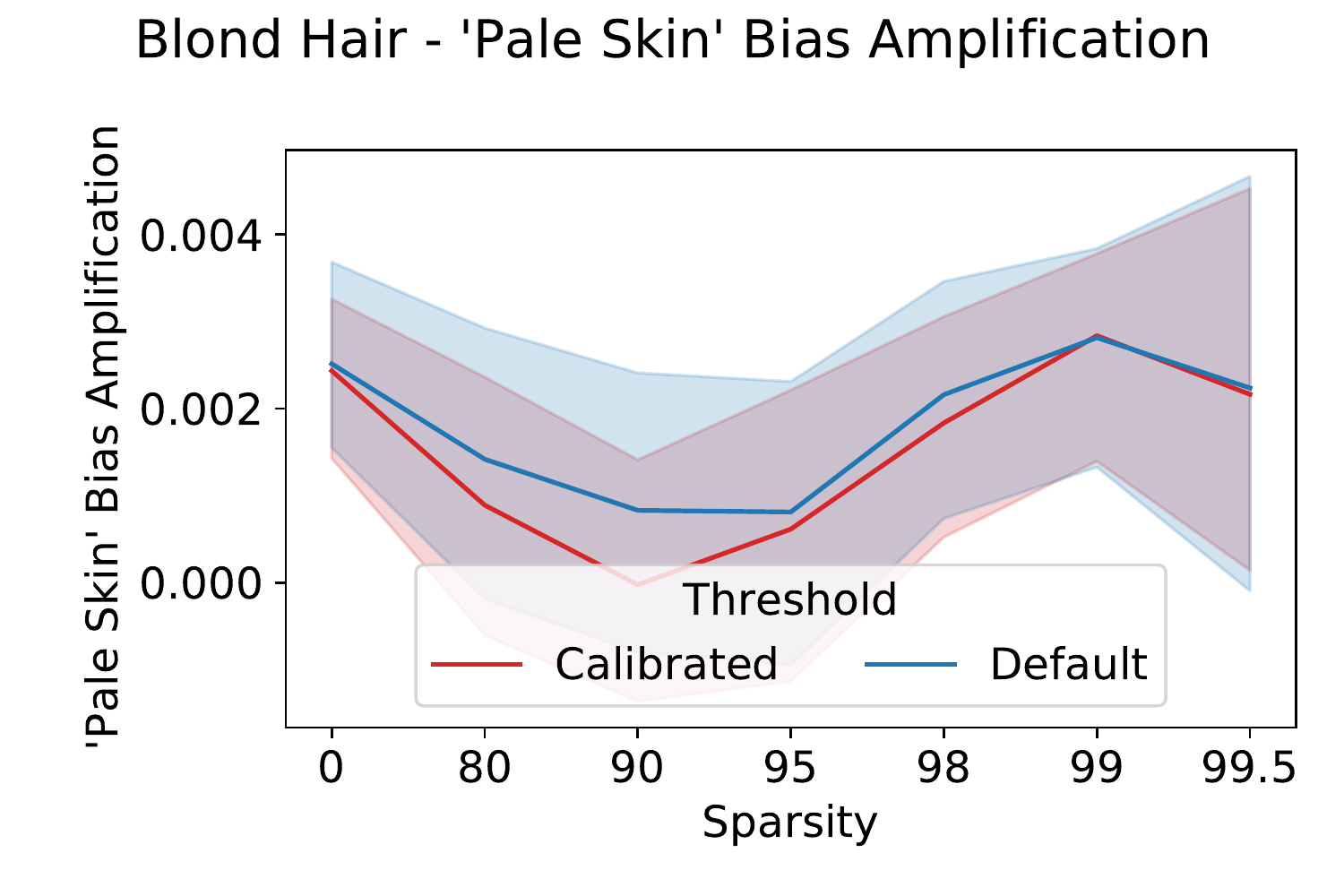} &
\includegraphics[width=0.12\textwidth]{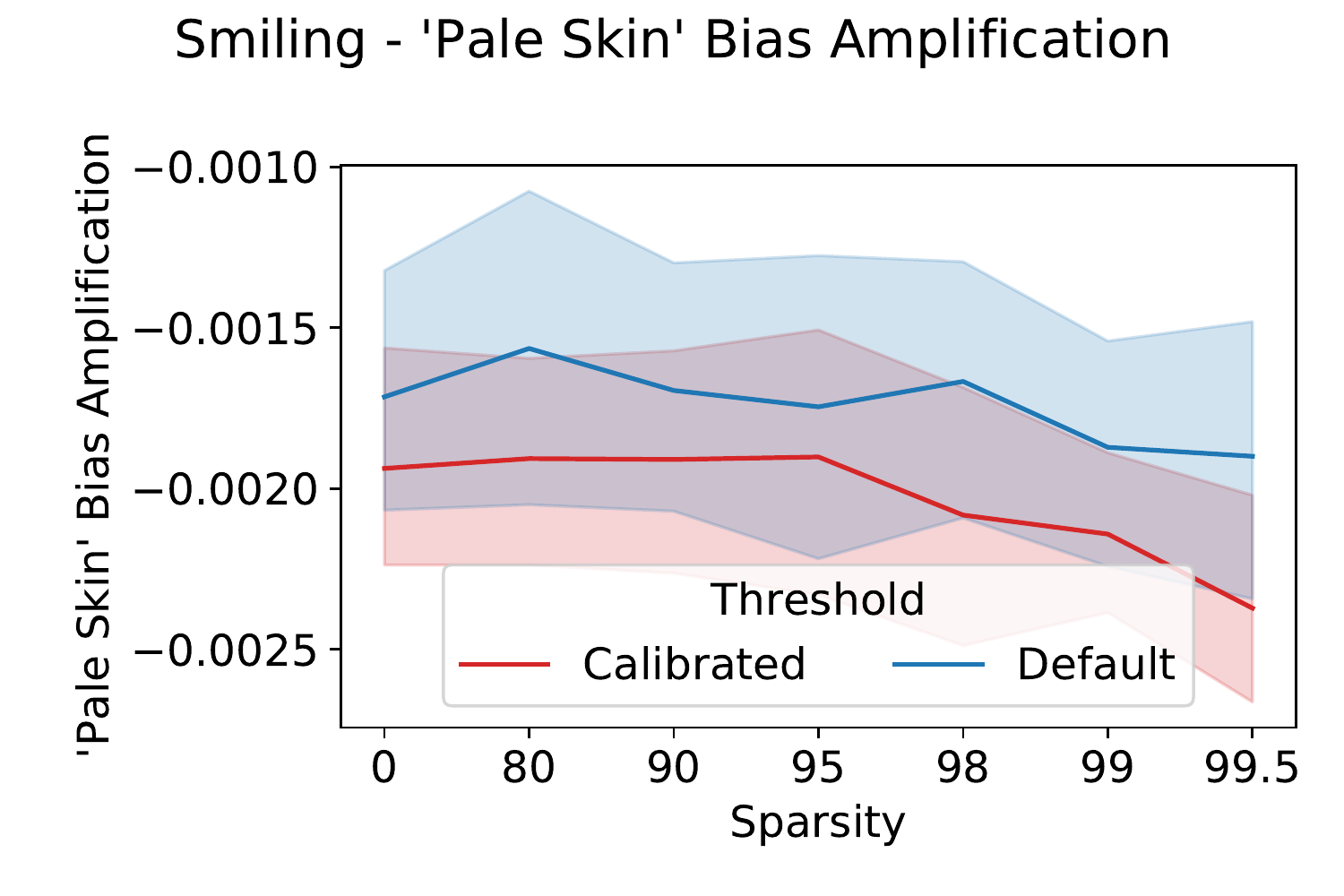}
    \\
\end{tabular}
    \caption{[CelebA / ResNet18 / Single Attribute / GMP-RI] Effect of threshold adjustment on Accuracy (first row), Threshold Calibration Bias (second row), 
    and Bias Amplification for the `Male', `Young', `Chubby', and `Pale Skin' attributes (third-sixth rows), on the ResNet18 CelebA model, predicting, from left to right, Oval Face, Big Nose, Bags Under Eyes, Receding Hairline, Mustache, Blond Hair, and Smiling). Red denotes results where the threshold is calibrated on the validation set, and blue denotes results from runs where the default threshold of 0.5 was used.  Omitted panels are cases where BA cannot be computed, either because there is no relationship between the predicted attribute and the category, or because the attribute is not present for one of the values of the category.}
    \label{fig:ta_celeba_rn18_single_full}
\end{figure}

\clearpage
\section{Bias Amplification Results from Training the Predicted and Category Attribute Together}
\label{appendix:ba_combined}

Inspired by our observation that, at low sparsities, joint training of all 40 attributes results in substantially lower bias amplification, we tested the impact of jointly training two attributes - a predicted attribute that shows high bias amplification in other training scenarios, and the identity category with regard to which high BA was observed. In all, we jointly co-trained five such pairs: Big Nose + Male, Oval Face + Male, Big Nose + Young, Mustache + Young, and Receding Hairline + Young. Except for using two logistic heads in the prediction layer, the training setting matches exactly our training settings for singly-trained models.

\begin{figure}[h]
\centering\begin{tabular}{cc}
\includegraphics[width=0.4\textwidth]{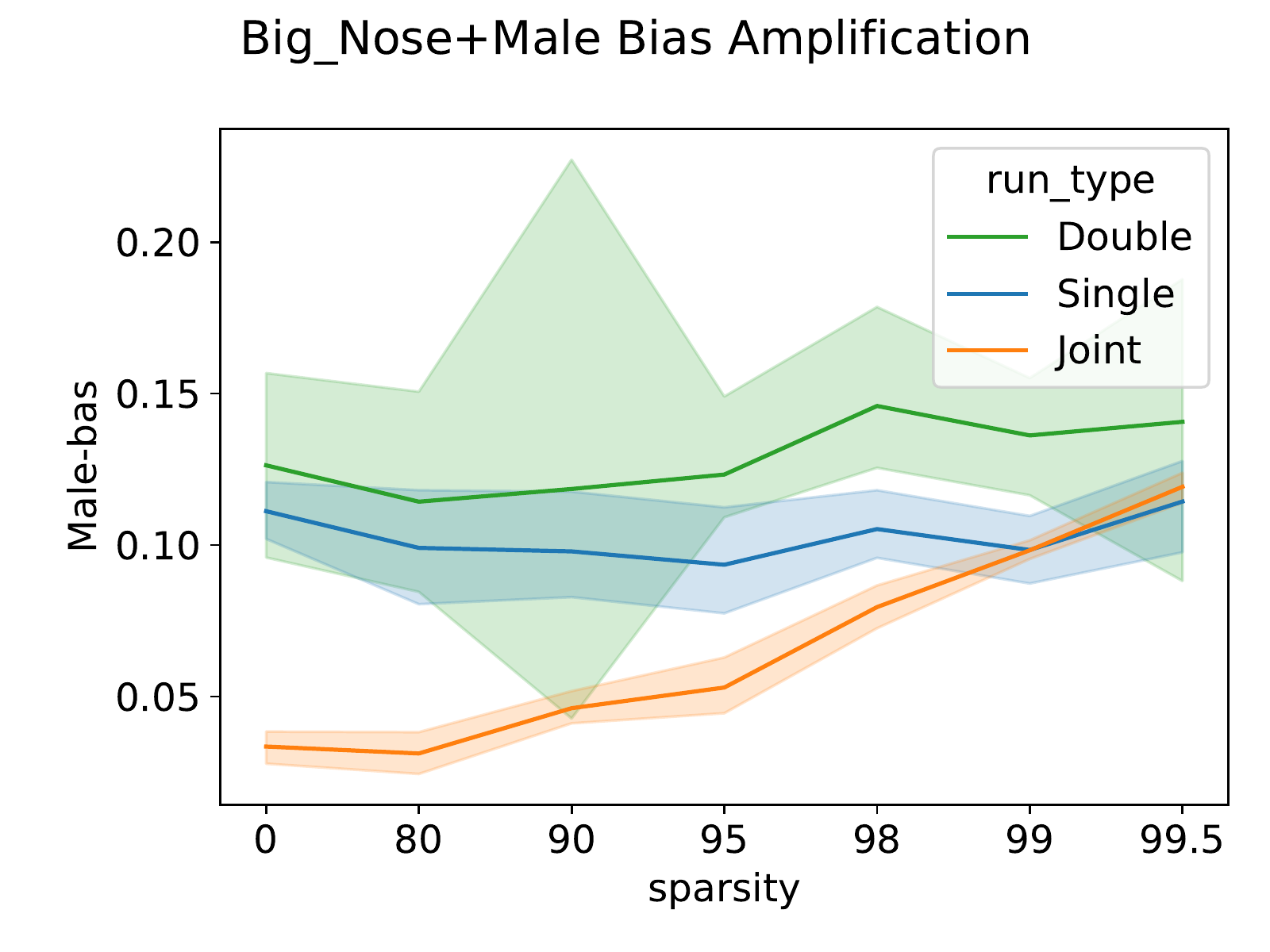} &
\includegraphics[width=0.4\textwidth]{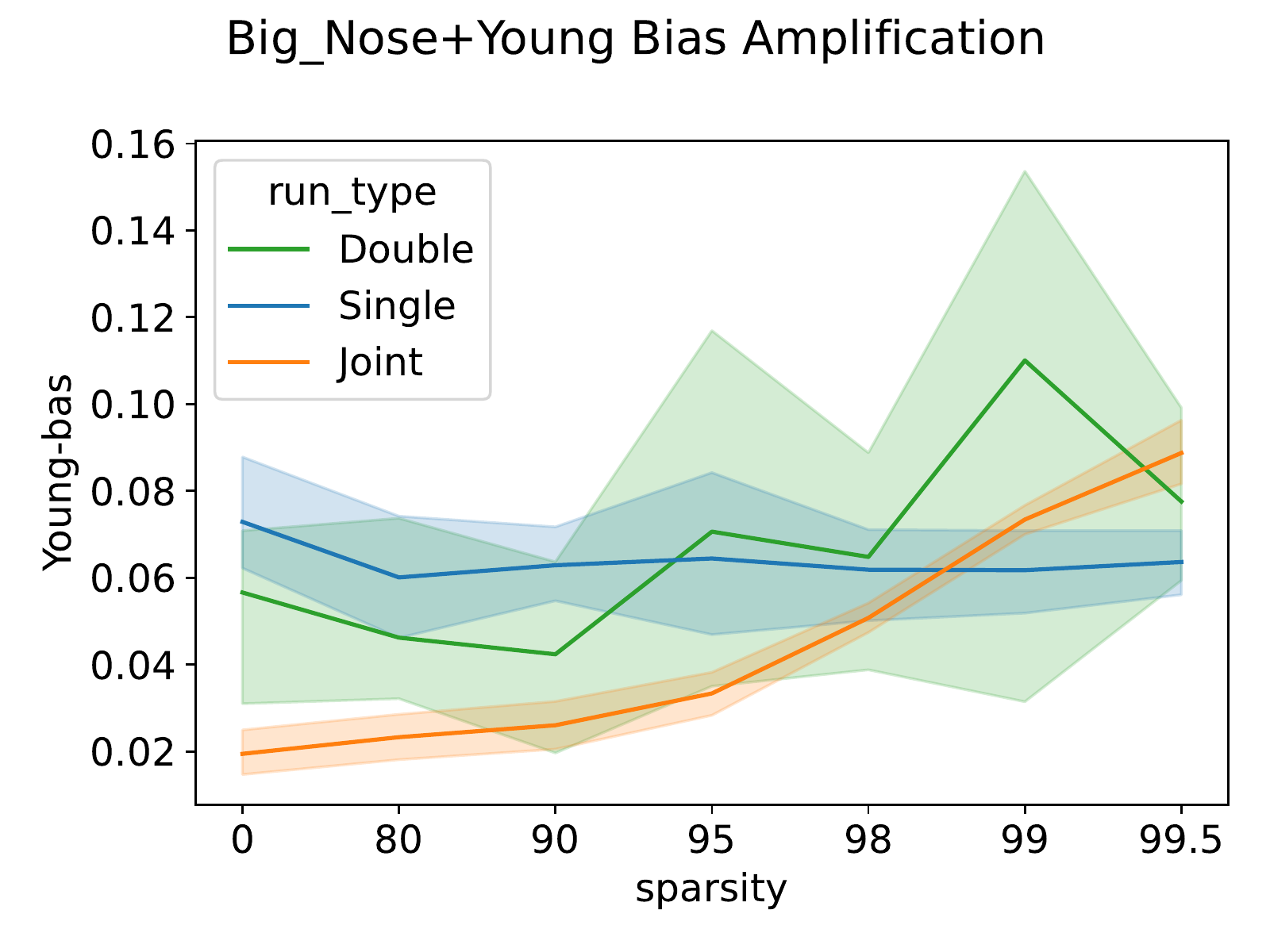} \\
\includegraphics[width=0.4\textwidth]{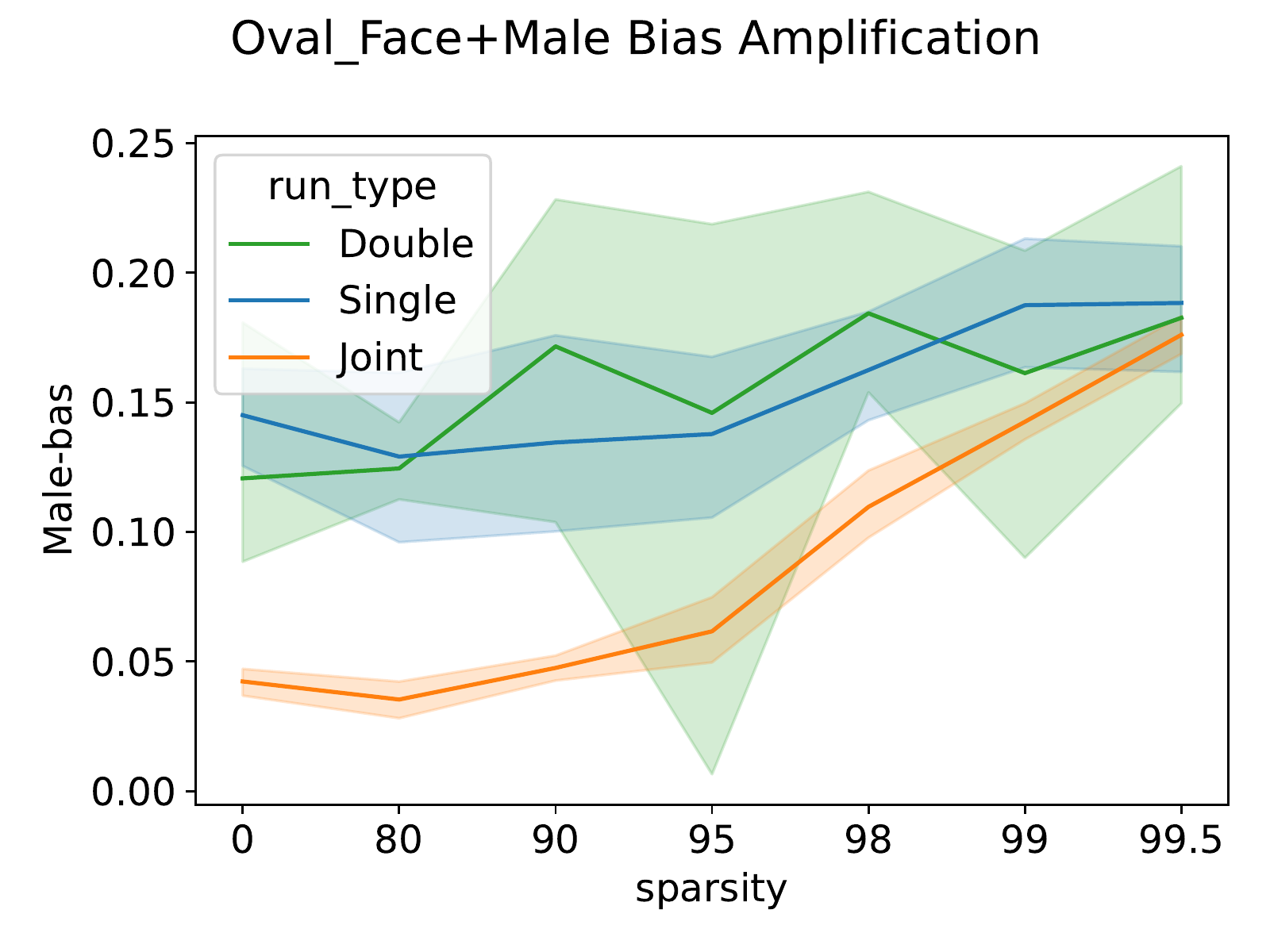} &
\includegraphics[width=0.4\textwidth]{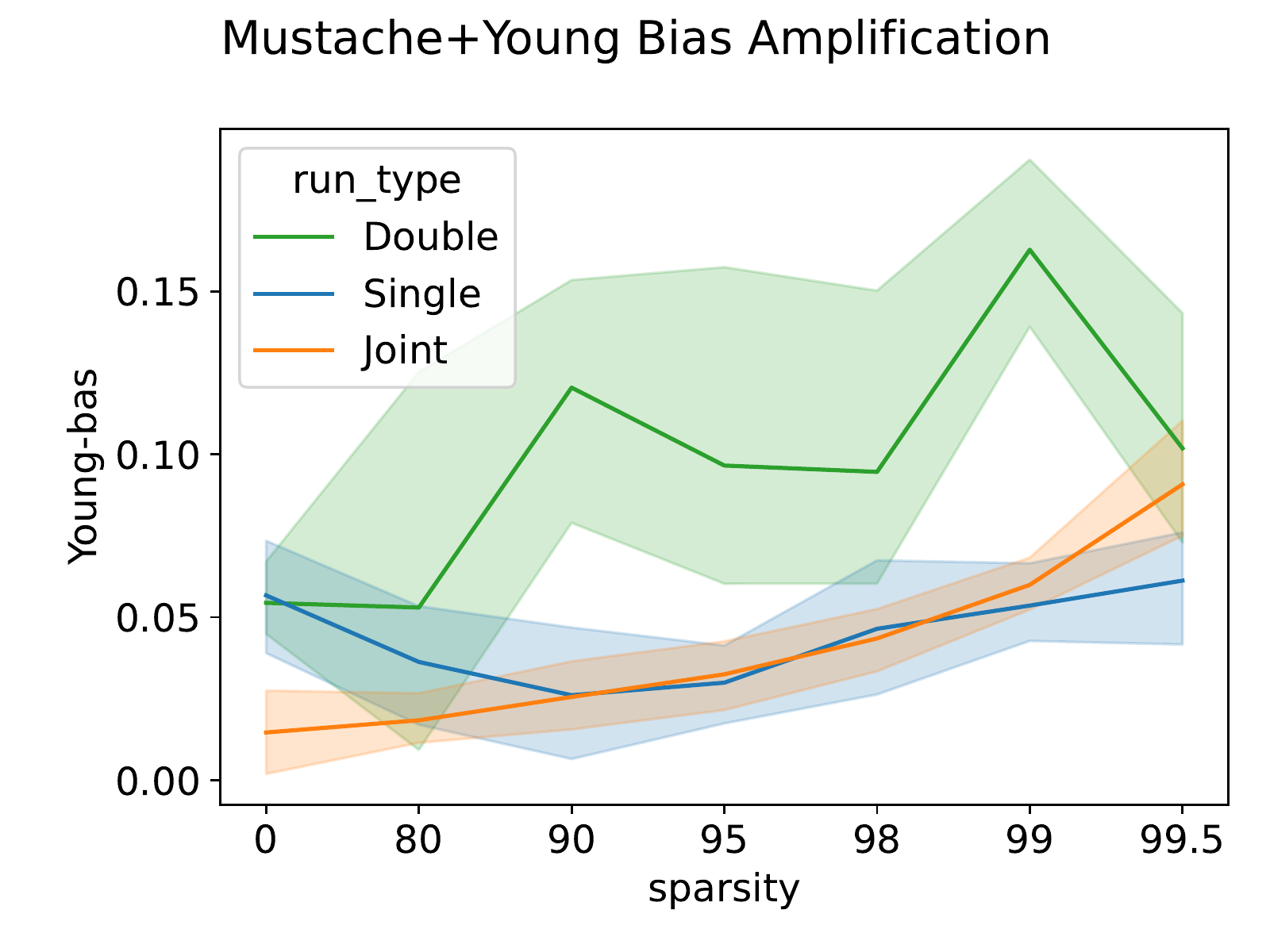} \\
&
\includegraphics[width=0.4\textwidth]{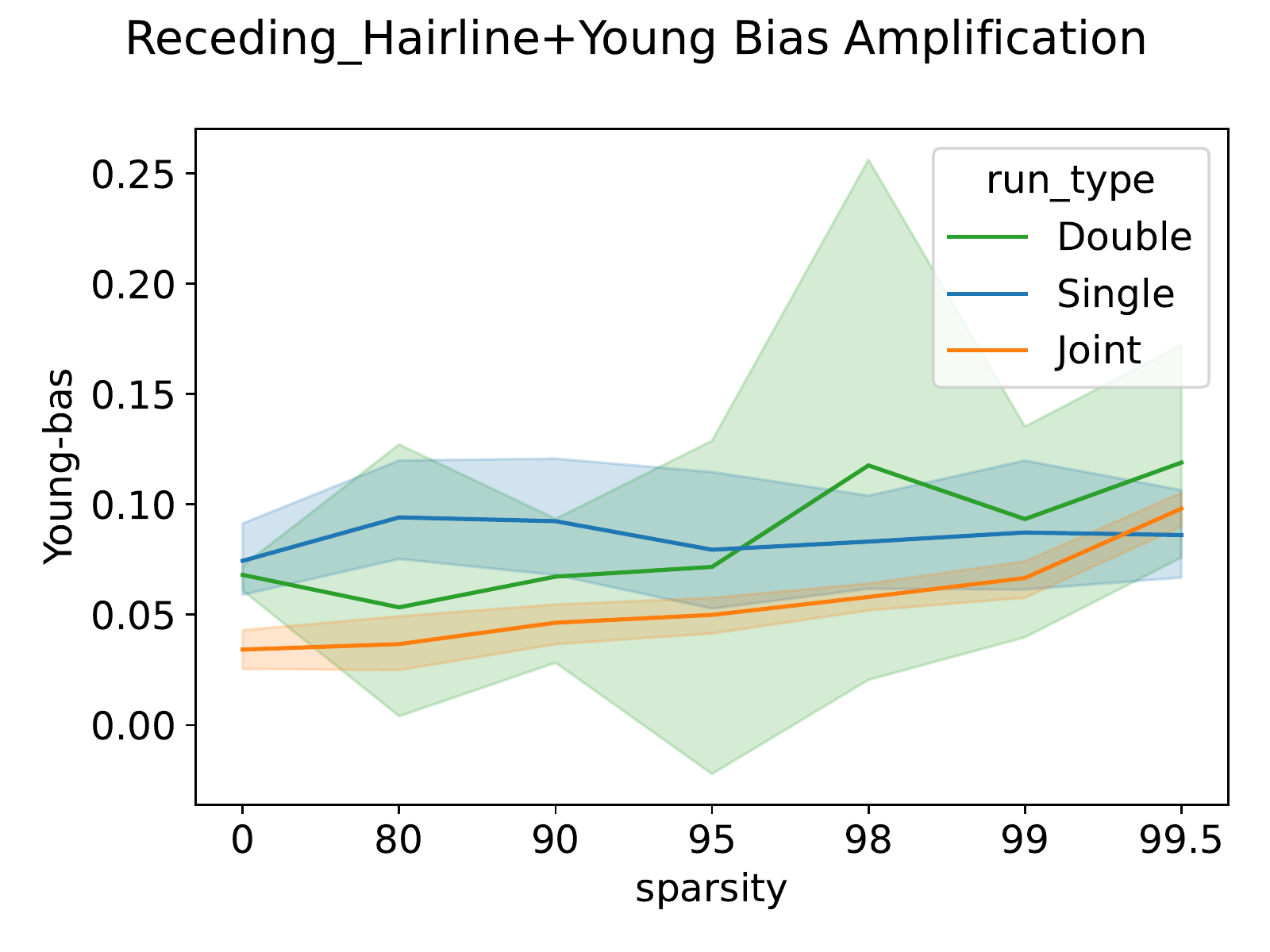} \\

\end{tabular}
    \caption{[CelebA / ResNet18 / Two-Attribute / GMP-RI] Comparison of bias amplification between models that are singly-trained, jointly-trained for all forty attributes, and models that are trained to predict only one attribute + the protected category.}
    \label{fig:celeba_rn18_ba_combined}
\end{figure}

The results of the experiment are shown in Figure~\ref{fig:celeba_rn18_ba_combined}. We observe that in all five cases, the BA of the "double" model, which co-trains the protected and predicted attribute, matches the BA of the single model fairly closely. This result suggests that more attributes looking at various facial features would need to be jointly trained in order to decrease BA at lower sparsities.

\clearpage
\section{Post-training pruning results}
\label{appendix:post-training}

\begin{figure}[h]
    \centering
\begin{tabular}{cccc}
   \includegraphics[width=0.22\textwidth]{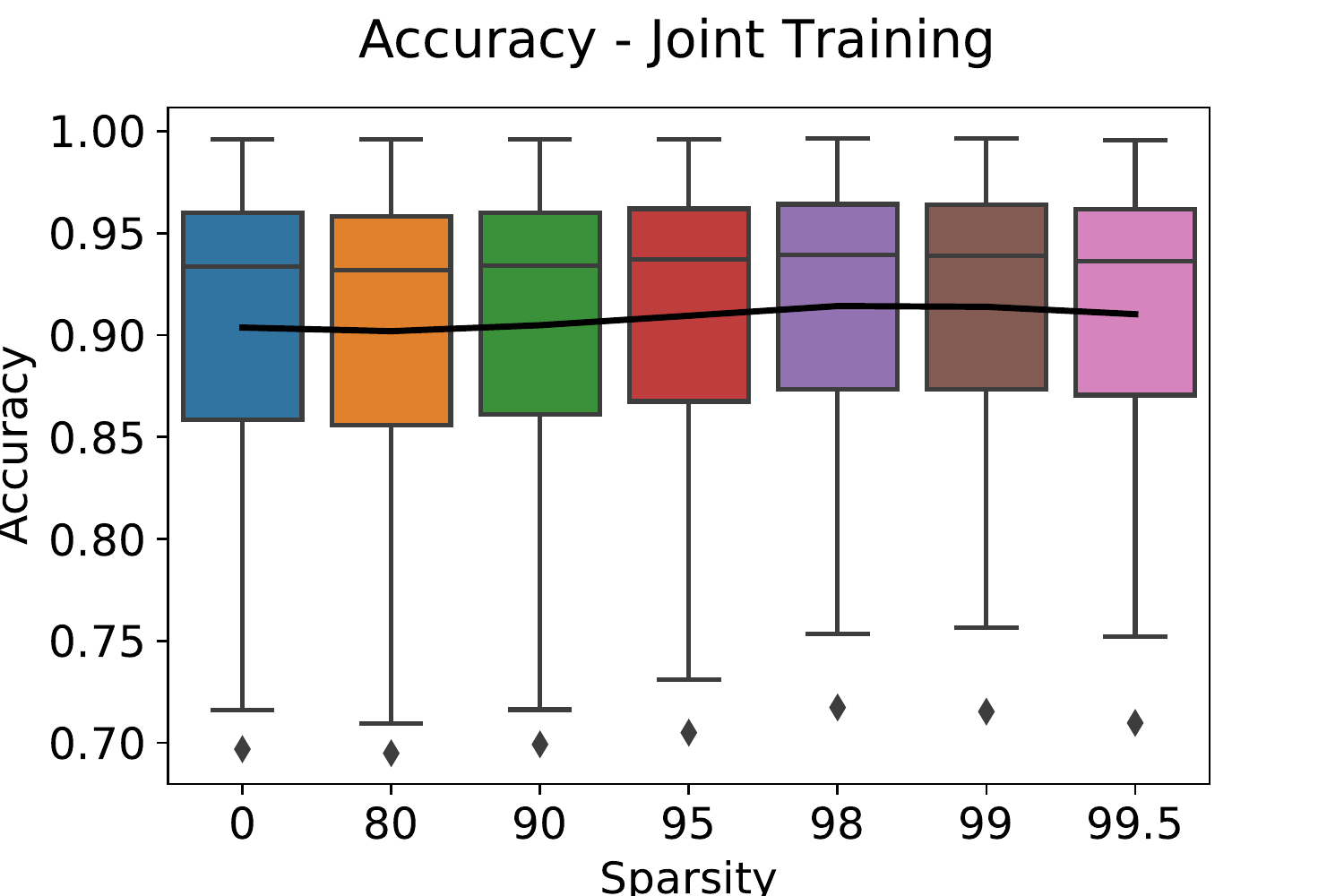} &
   \includegraphics[width=0.22\textwidth]{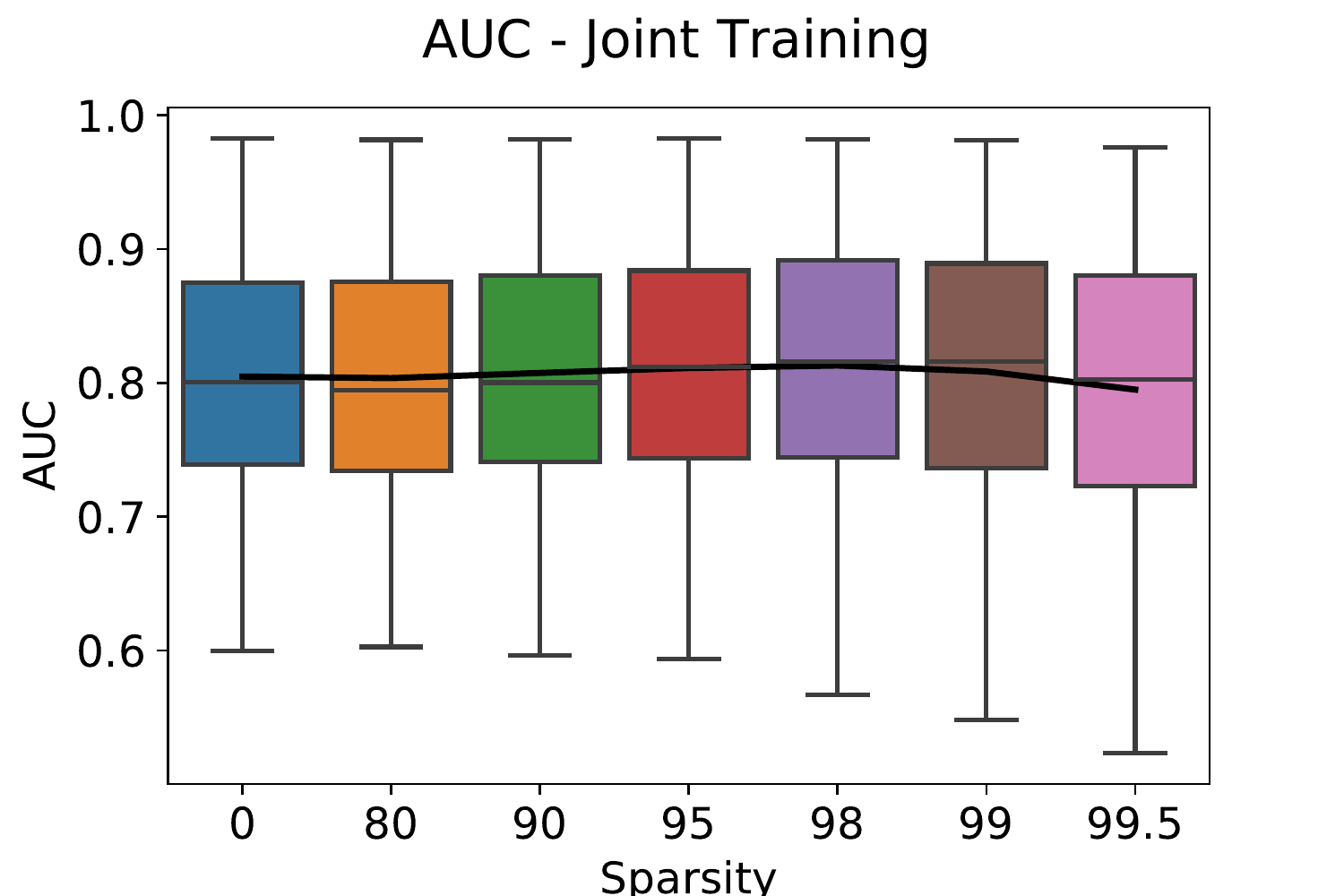} \\
    \includegraphics[width=0.22\textwidth]{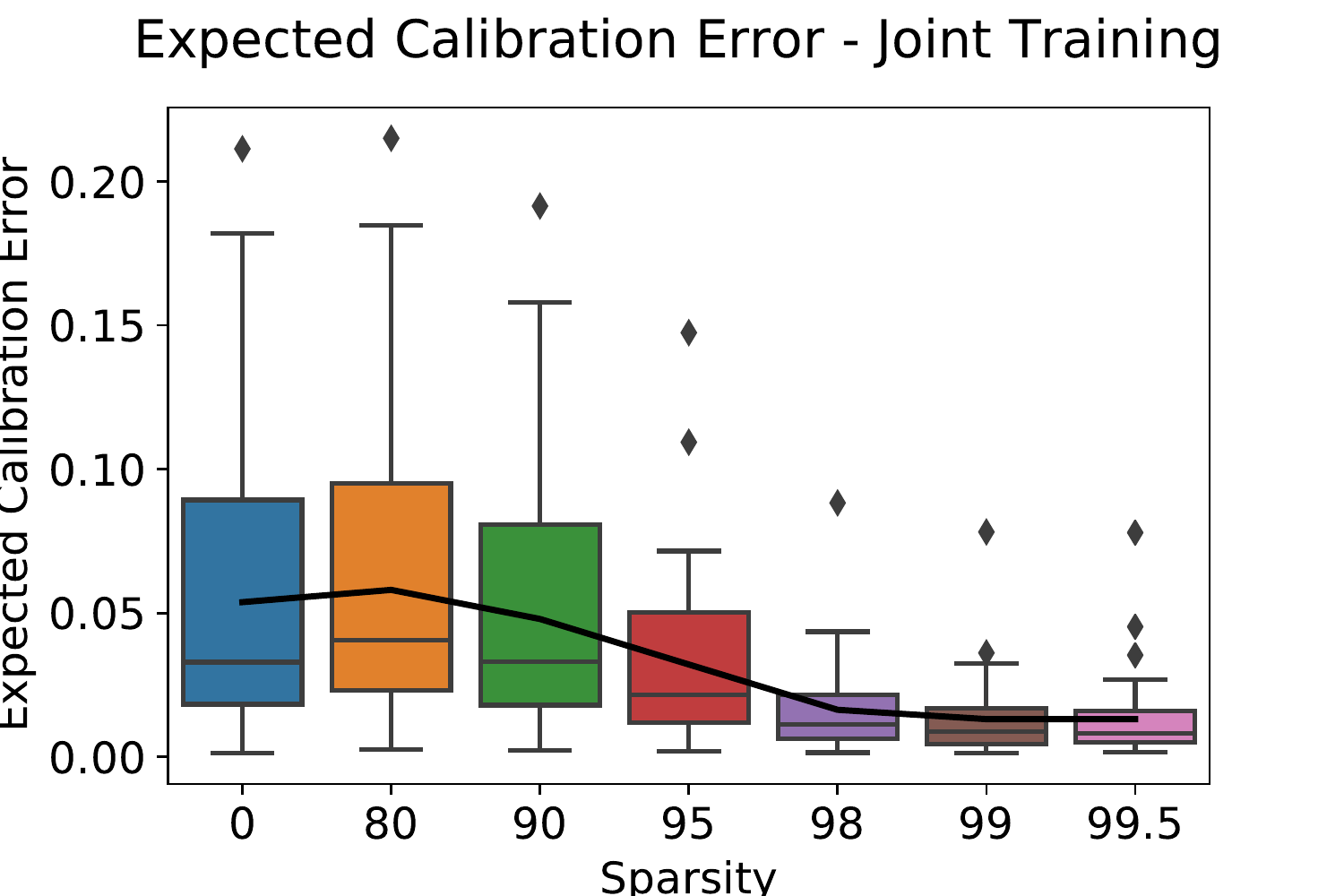} &
    \includegraphics[width=0.22\textwidth]{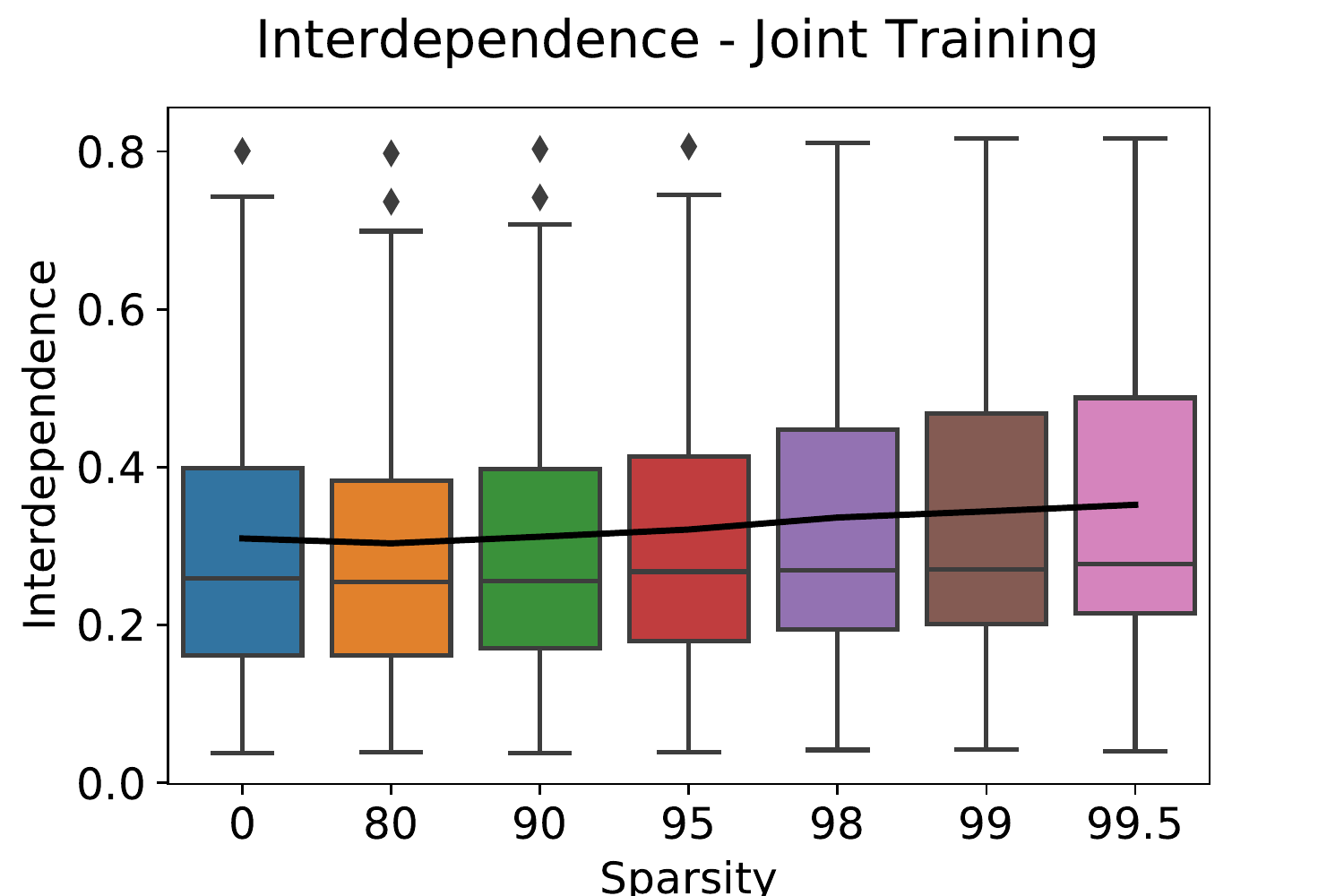} &
        \includegraphics[width=0.22\textwidth]{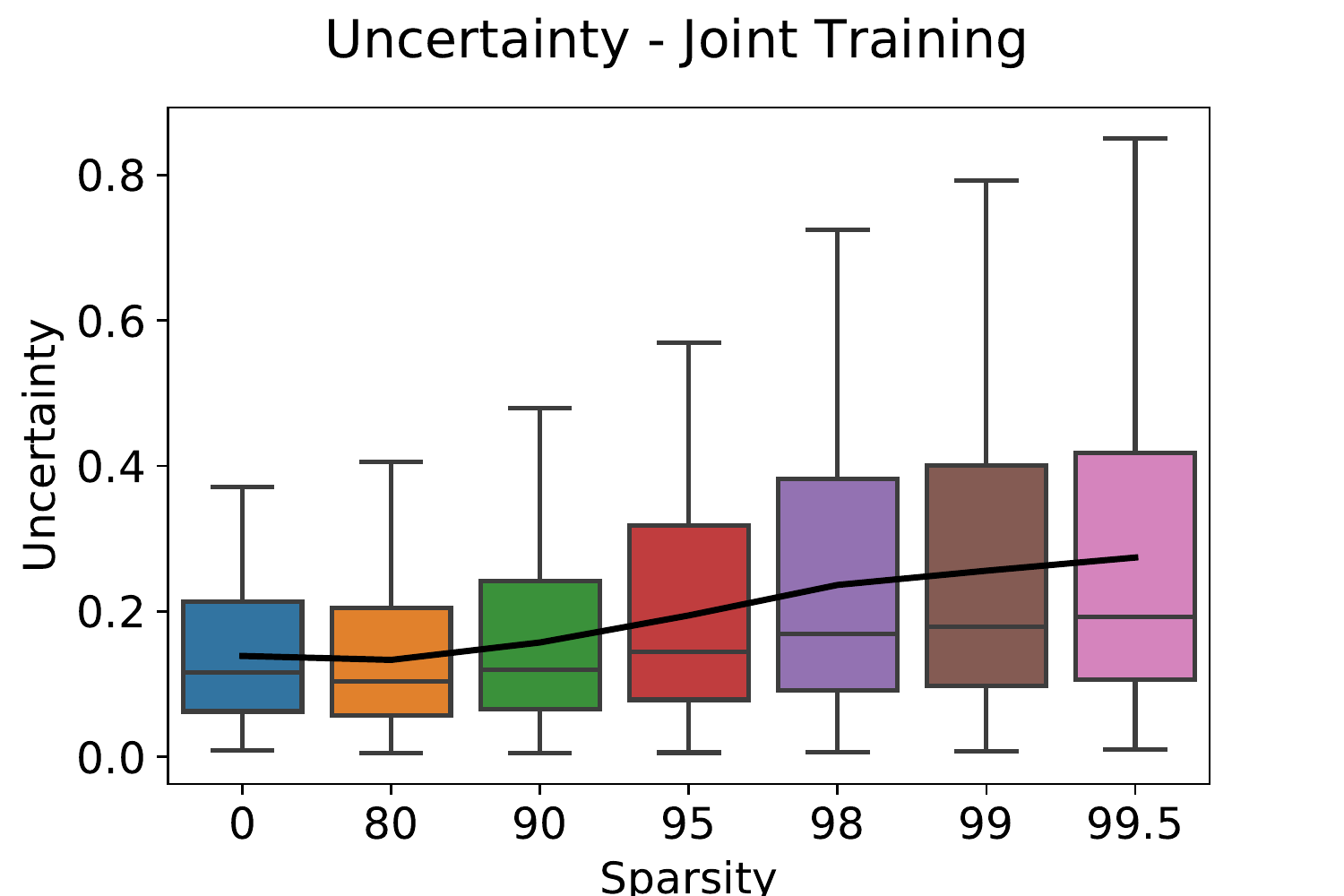} &
    \includegraphics[width=0.22\textwidth]{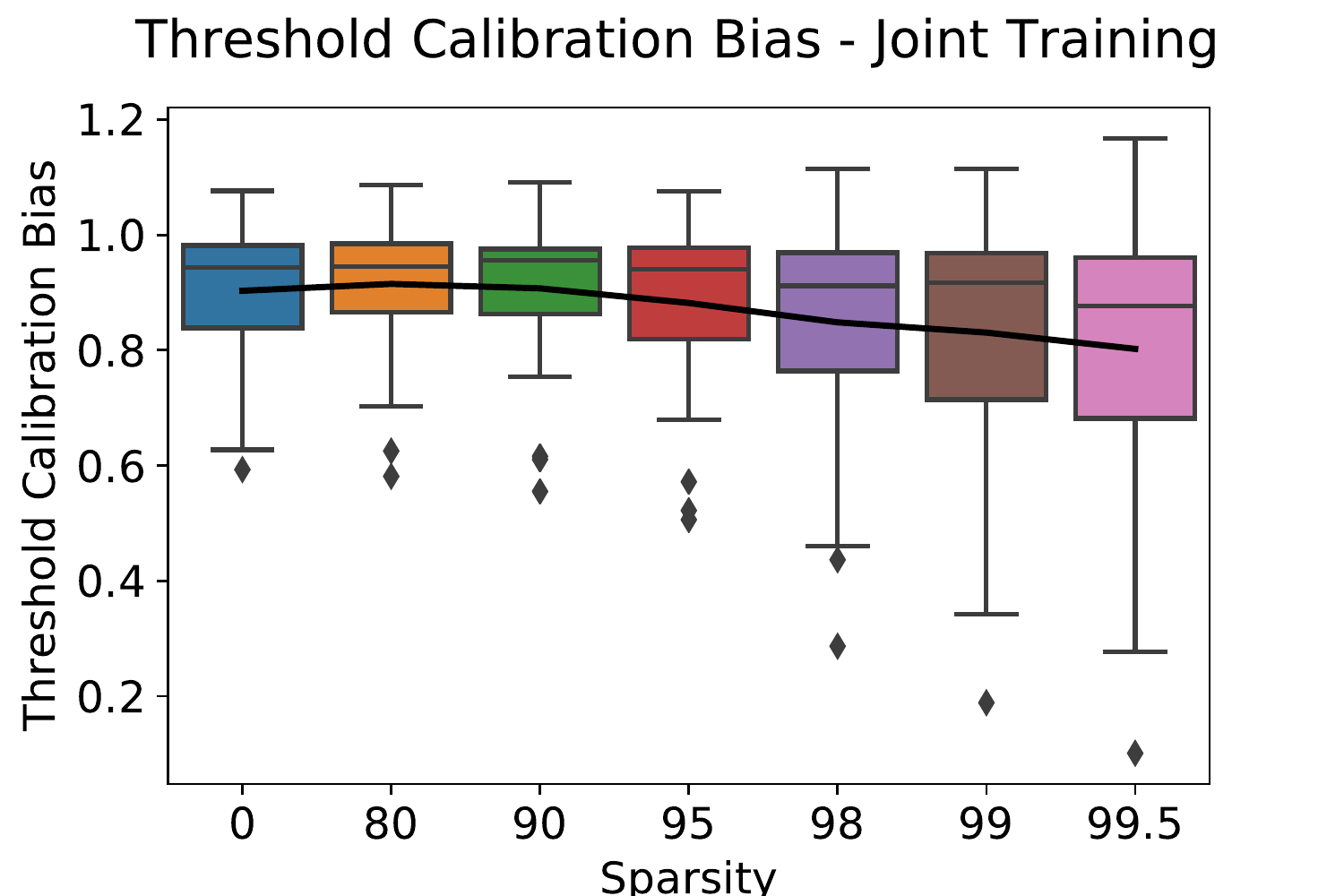}\\
    \includegraphics[width=0.22\textwidth]{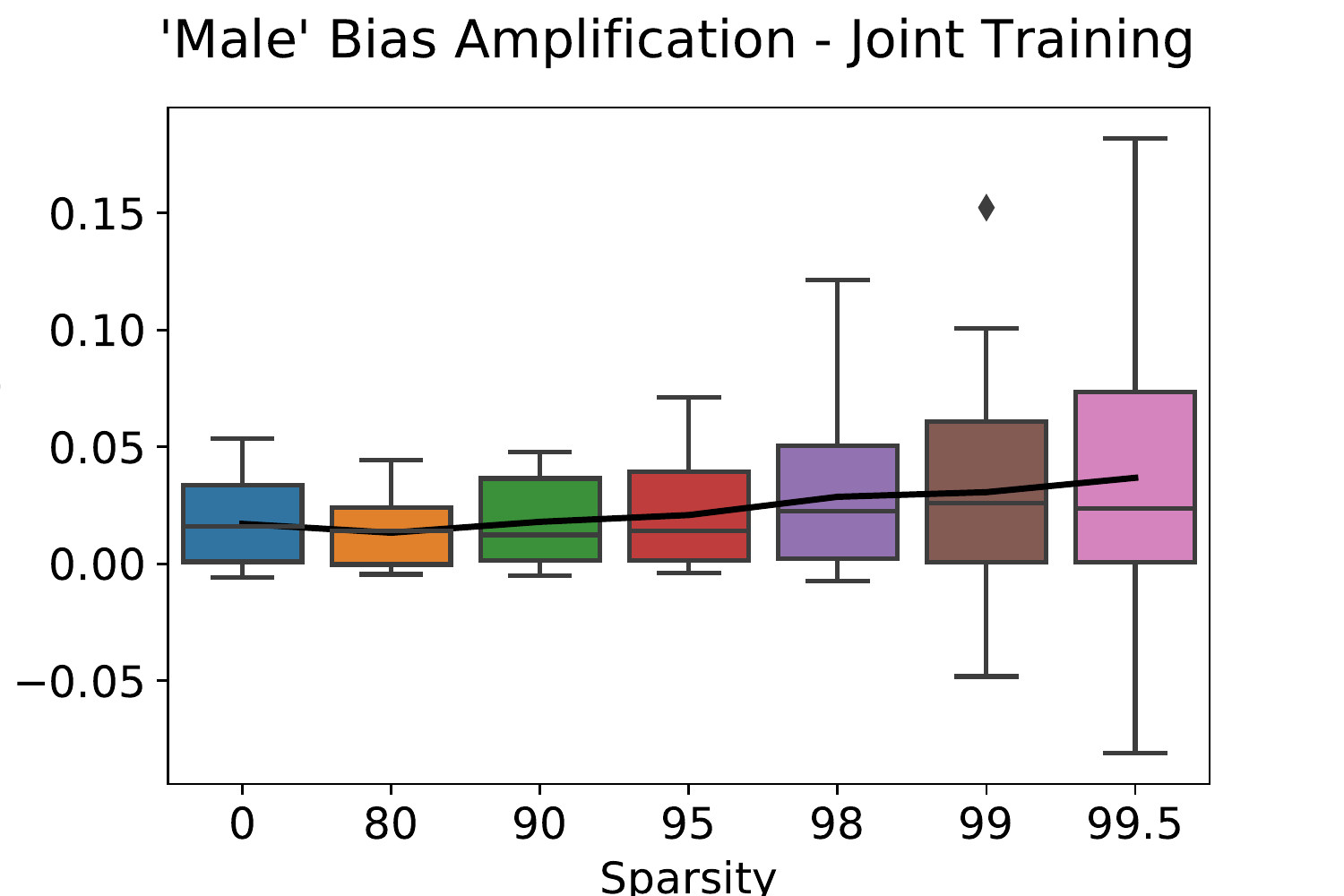} &
    \includegraphics[width=0.22\textwidth]{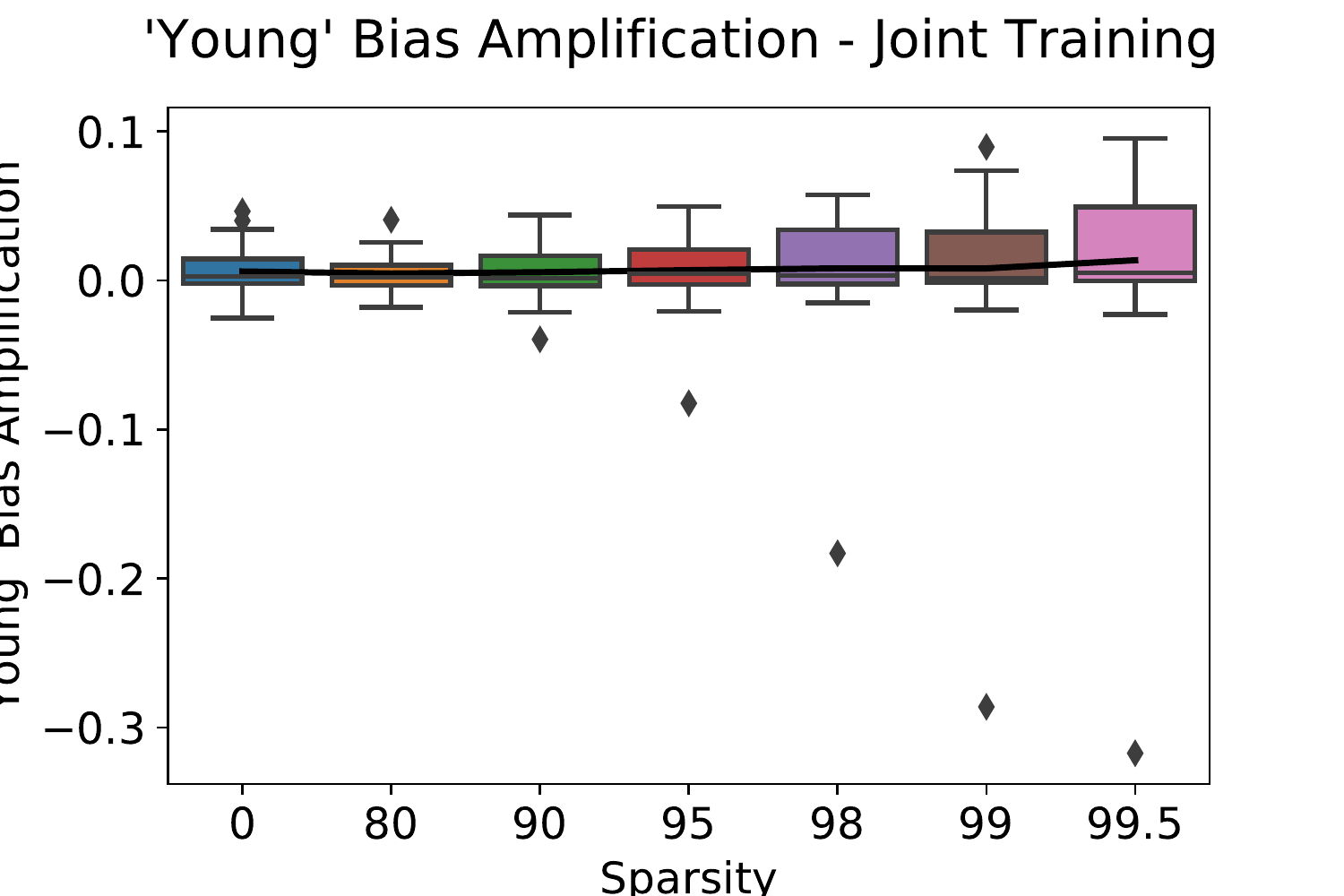} &
        \includegraphics[width=0.22\textwidth]{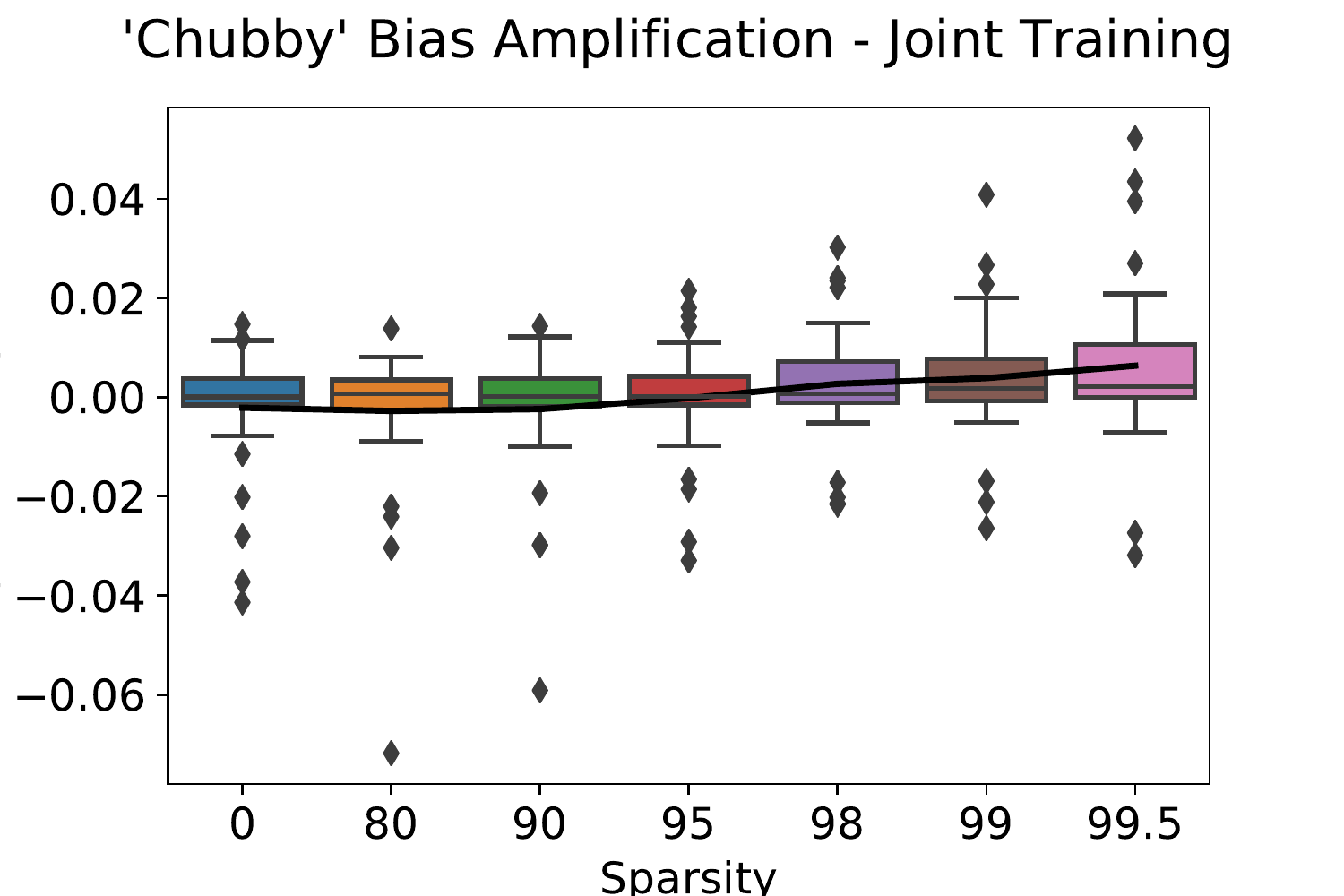} &
    \includegraphics[width=0.22\textwidth]{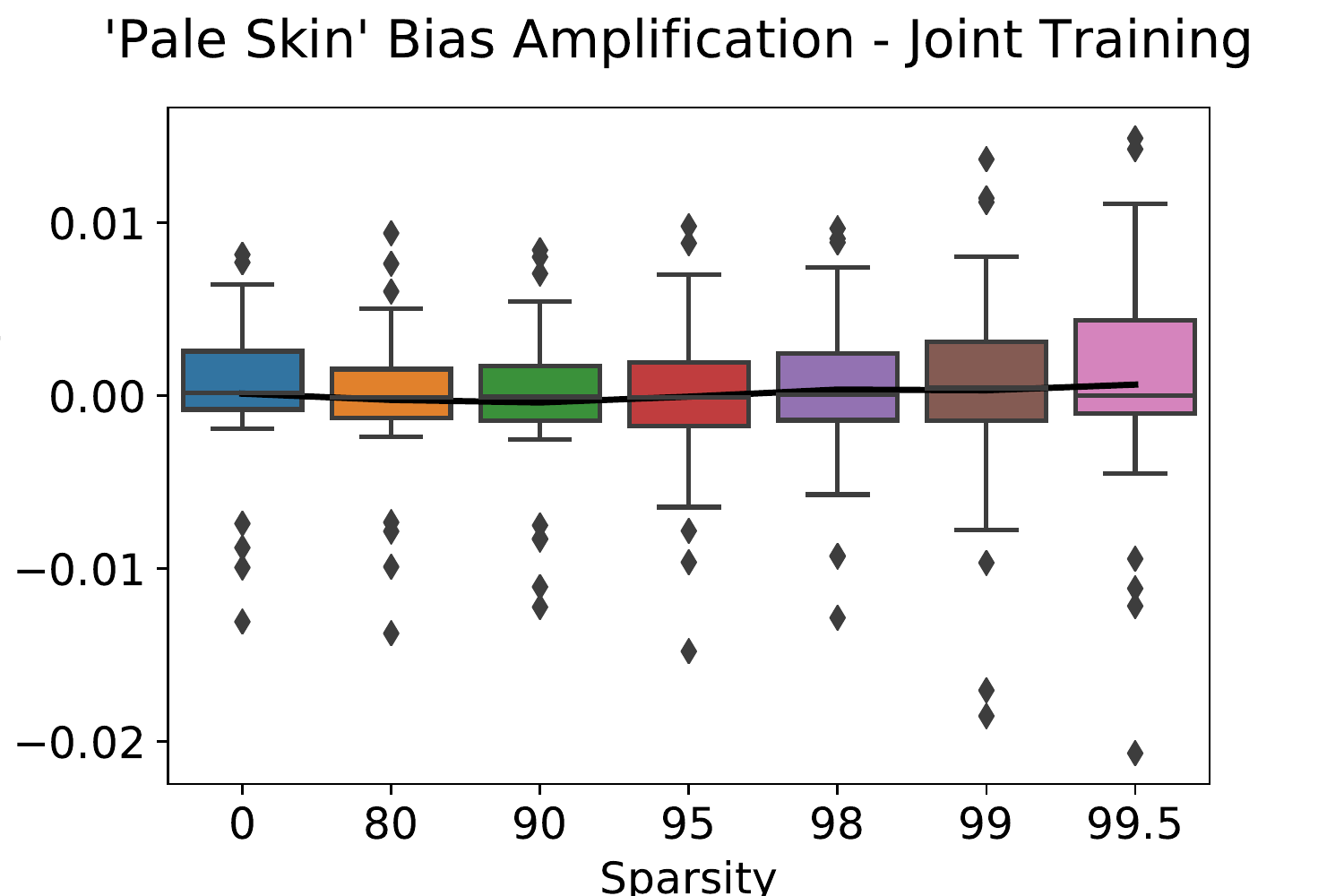}\\
\end{tabular}
    \caption{[CelebA / ResNet18 / GMP-PT] Accuracy and Systematic Bias metrics (TCB, ECE, Interdependence) of ResNet18 models jointly trained on all CelebA attributes, and pruned Post-Training (GMP-PT). The thick black line denotes the mean value at each sparsity level.
    }
    \label{fig:celeba_rn18_joint_PT_systematic}
\end{figure}

\begin{figure}[ht]
\centering
\begin{tabular}{cc}
  \includegraphics[width=0.35\textwidth]{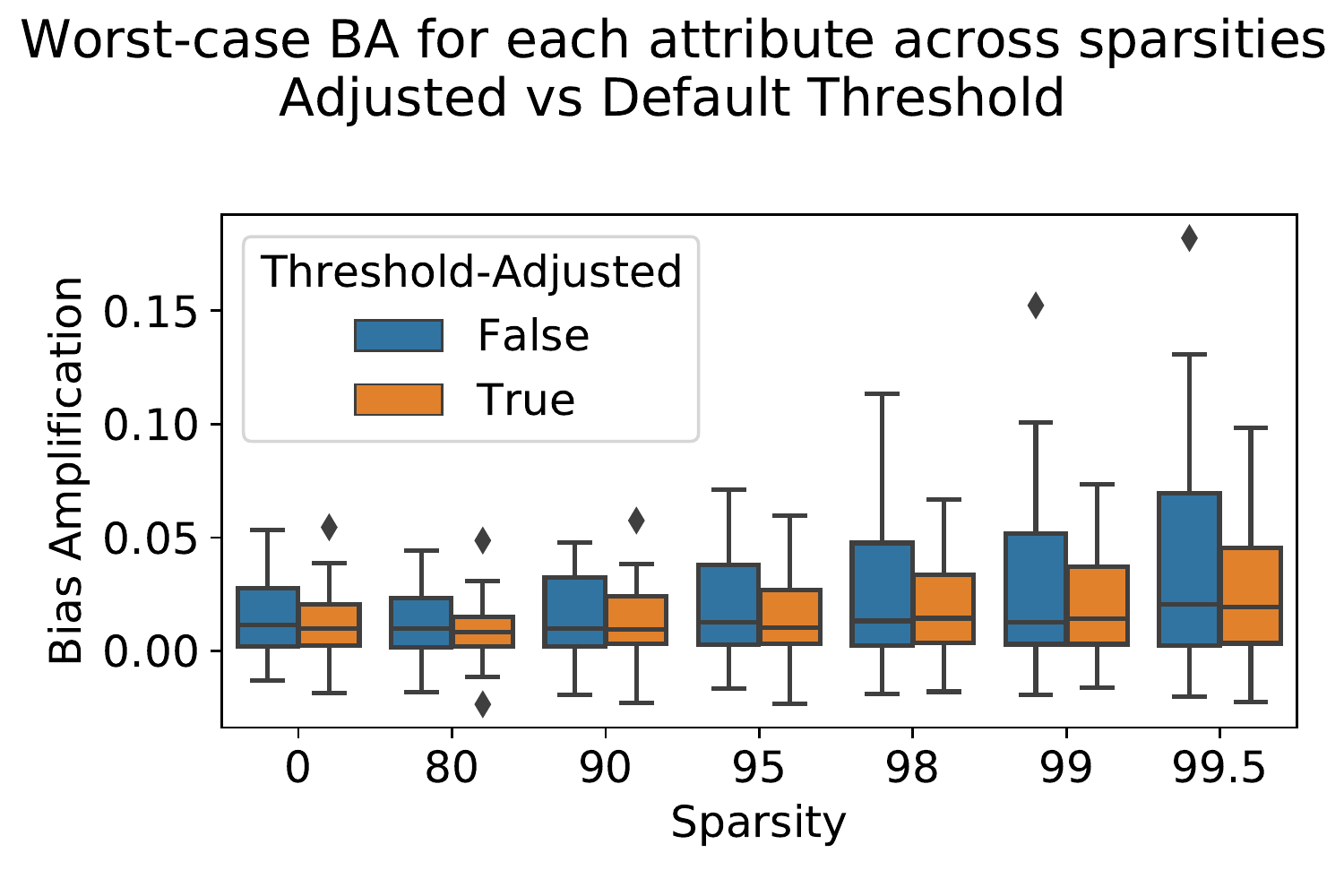} &
  \includegraphics[width=0.35\textwidth]{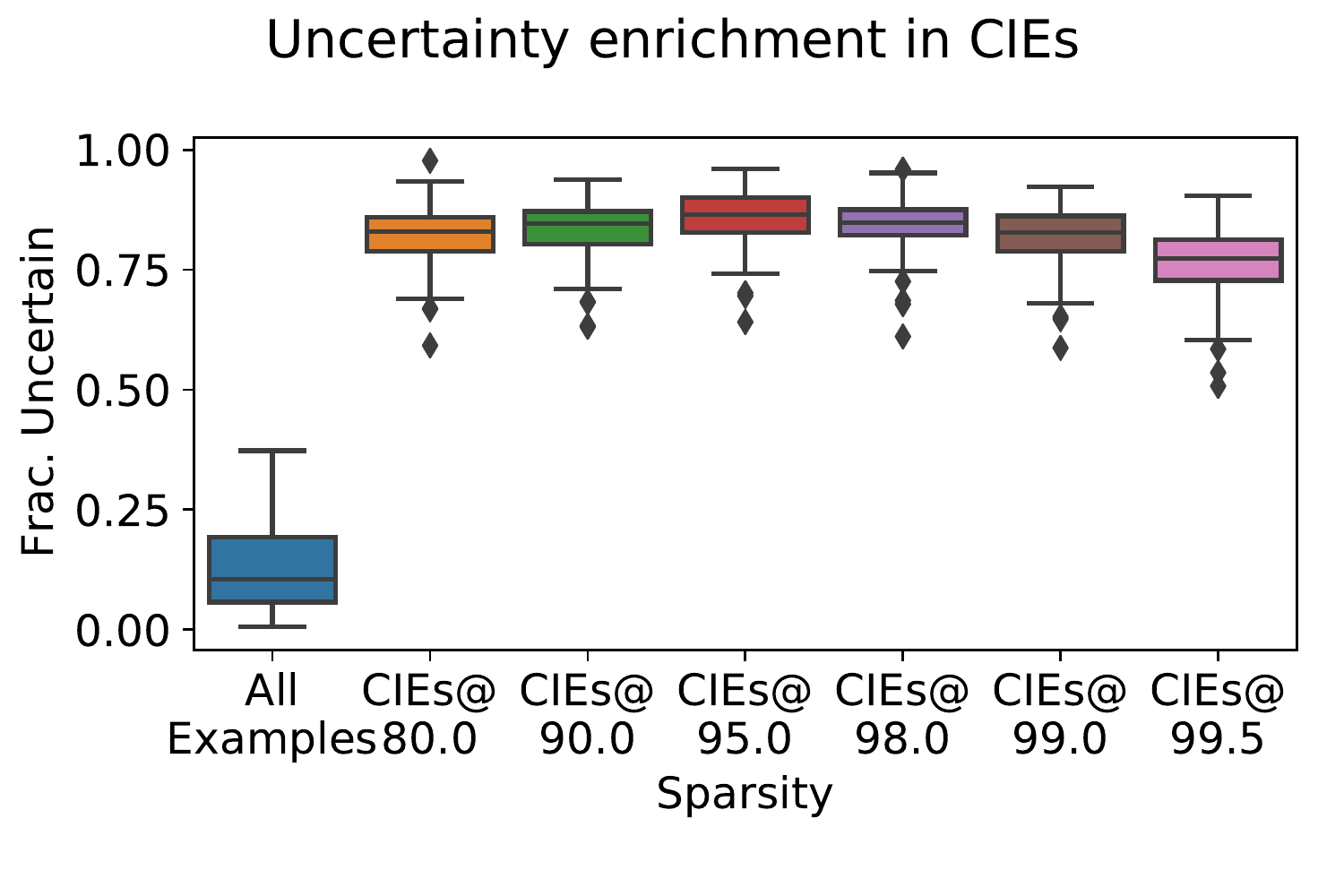}
  \end{tabular}
    \caption{[CelebA / ResNet18 / GMP-PT] (Left) Effect of threshold calibration on models jointly trained on all attributes. %
    (Right) Proportion of uncertain predictions for \emph{dense} models across all attributes for all elements in the CelebA test set, and for Compression-Identified Exemplars at different sparsities.}
    \label{fig:celeba_rn18_gmp_PT_threshold_adj}
\end{figure}

\begin{figure}[h]
\centering
\includegraphics[width=0.8\textwidth]{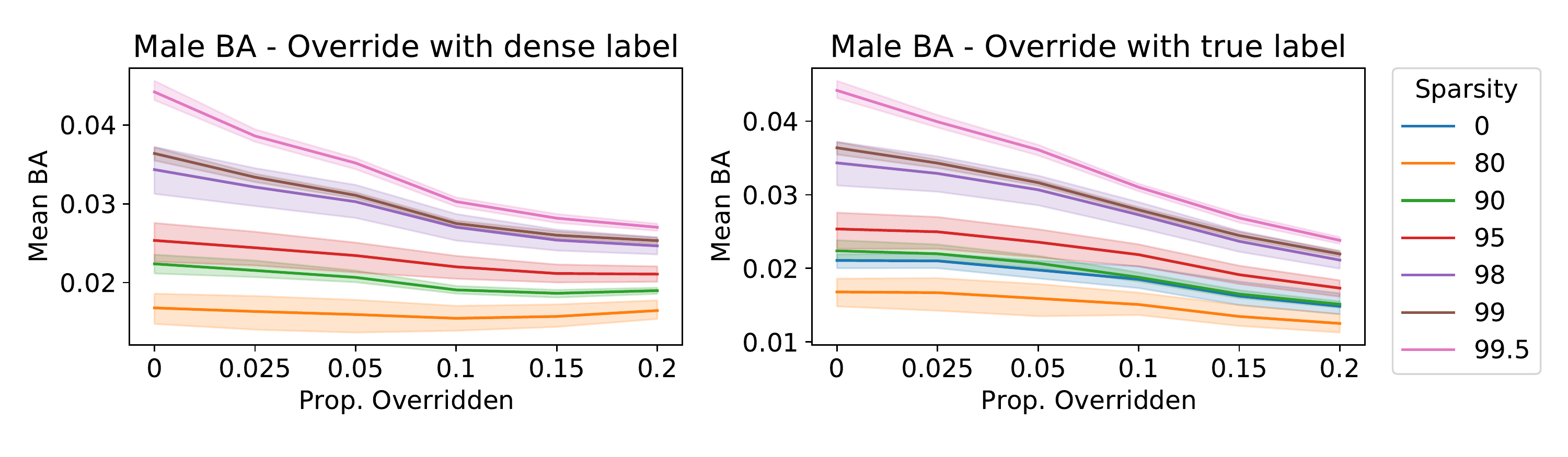}
\includegraphics[width=0.8\textwidth]{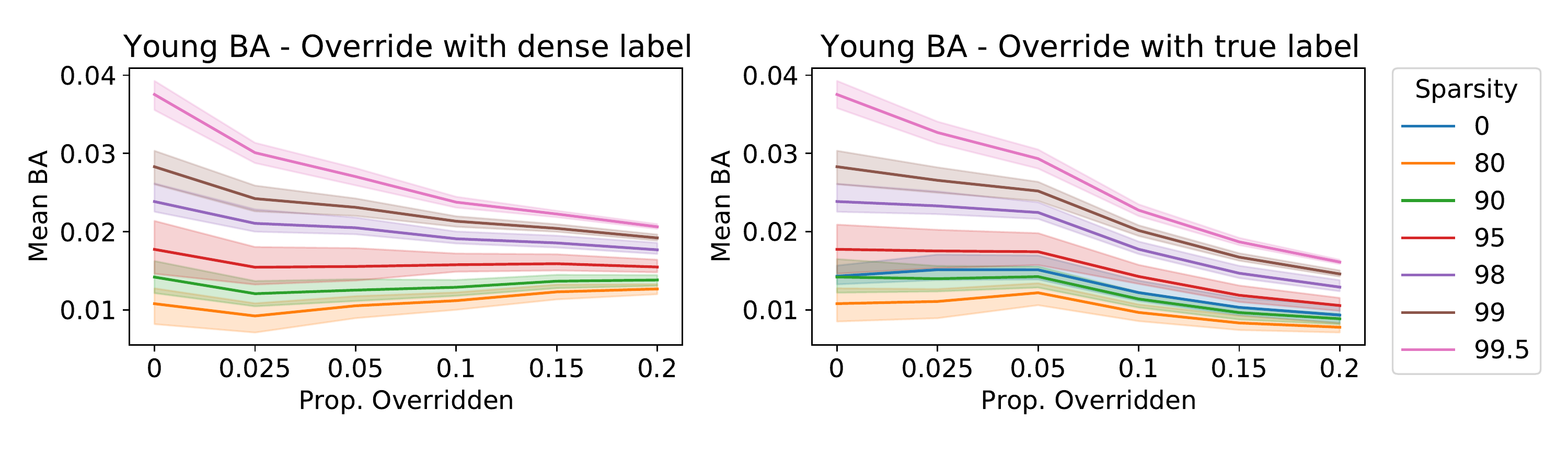}
\includegraphics[width=0.8\textwidth]{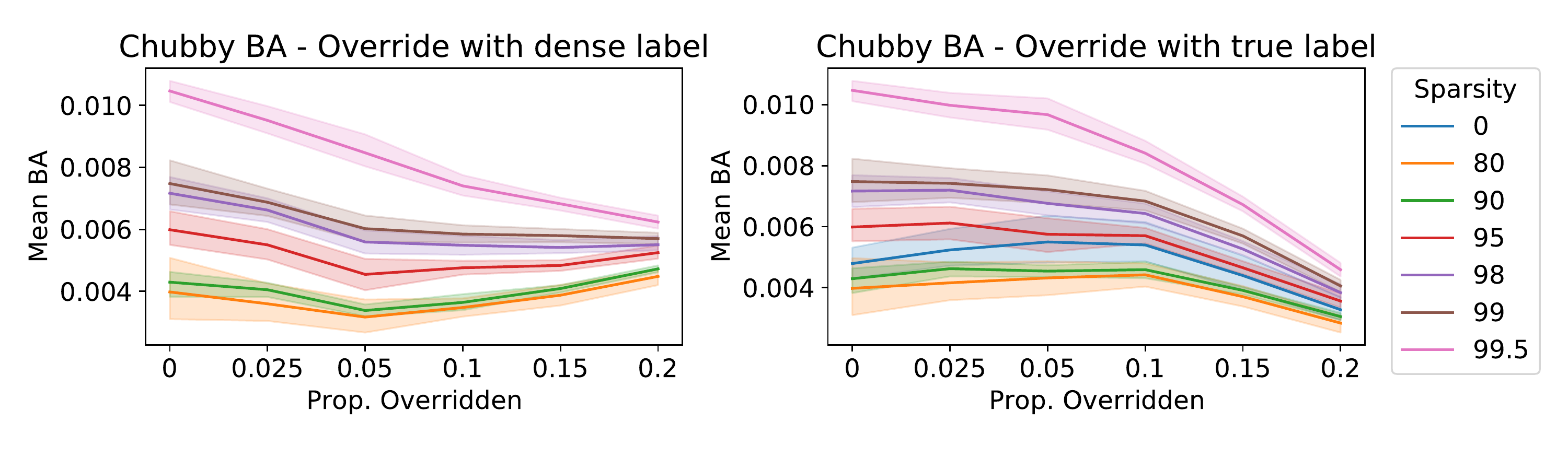}
\includegraphics[width=0.8\textwidth]{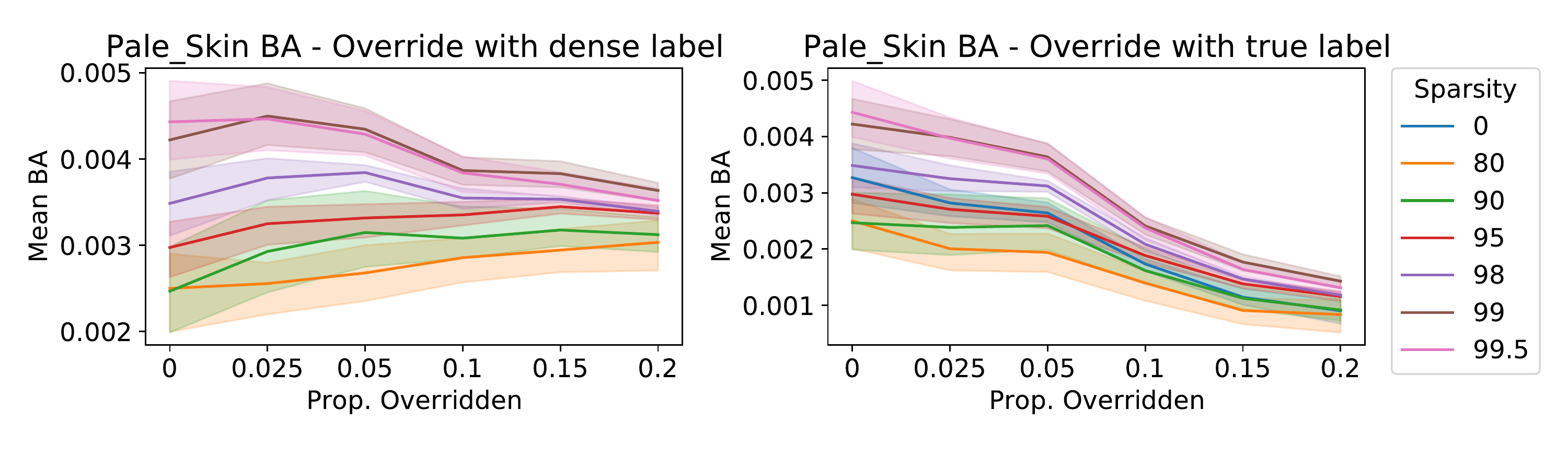}
    \caption{[CelebA / ResNet18 / GMP-PT] Effect of label overrides on Bias Amplification. In all cases, overrides are prioritized by dense model uncertainty.}
    \label{fig:overrides_PT}
\end{figure}

We further extend our analysis of bias in sparse CelebA/ResNet18 models, by using a different pruning procedure. Specifically, we perform gradual magnitude pruning starting from pre-trained dense models (GMP-PT); the full training hyperparameters are explained in Appendix Section~\ref{appendix:training_settings}. Our results for GMP-PT are presented in Figure~\ref{fig:celeba_rn18_joint_PT_systematic}. In terms of accuracy or AUC performance, we obtain good quality models even at high sparsity ($>99\%$), which is in line with our observations for the GMP-RI setting. Similarly, our conclusions hold for Systematic and Category bias. Namely, the ECE and TCB go down with sparsity, while the interdependence slightly increases and the fraction of uncertain samples increases substantially with model sparsity. The Category bias (BA) also increases with sparsity; this can be seen better on the Male attribute. Notably, compared to GMP-RI, the BA values are slightly lower for less sparse models (\emph{e.g} 80\% and 90\% sparse). We further test methods for bias mitigation on the GMP-RT and notice similar effects to the GMP-RI setting; namely, when overriding low confidence samples in the sparse models with either the true or dense label, we observe a substantial decrease in Category bias, as measured by BA, particularly at high sparsity (please see Figure~\ref{fig:overrides_PT}). Lastly, we study the relationship between uncertain samples and compression identified exemplars (CIEs)~\cite{hooker2019compressed, hooker_characterising_2020} in Figure~\ref{fig:celeba_rn18_gmp_PT_threshold_adj} and observe that most of the CIEs are uncertain samples.

\clearpage
\section{N:M Sparsity Results}
\label{appendix:nm_sparsity}

While modern GPU hardware cannot take full advantage of unstructured sparsity, introducing additional constraints can lead to effective speedups. In particular, N:M sparsity patterns, in which N out of every contiguous M values are removed, can be successfully accelerated \cite{mishra2021accelerating}. We validate our findings by evaluating systematic and categorical bias in the N:M sparsity setting. The sparsification algorithm is a variant of the Random-Initialization Global Magnitude Pruning algorithm used in the main body of the paper. Each experiment was repeated from three different random initializations. 

\begin{figure}[h]
    \centering
\begin{tabular}{cccc}
   \includegraphics[width=0.22\textwidth]{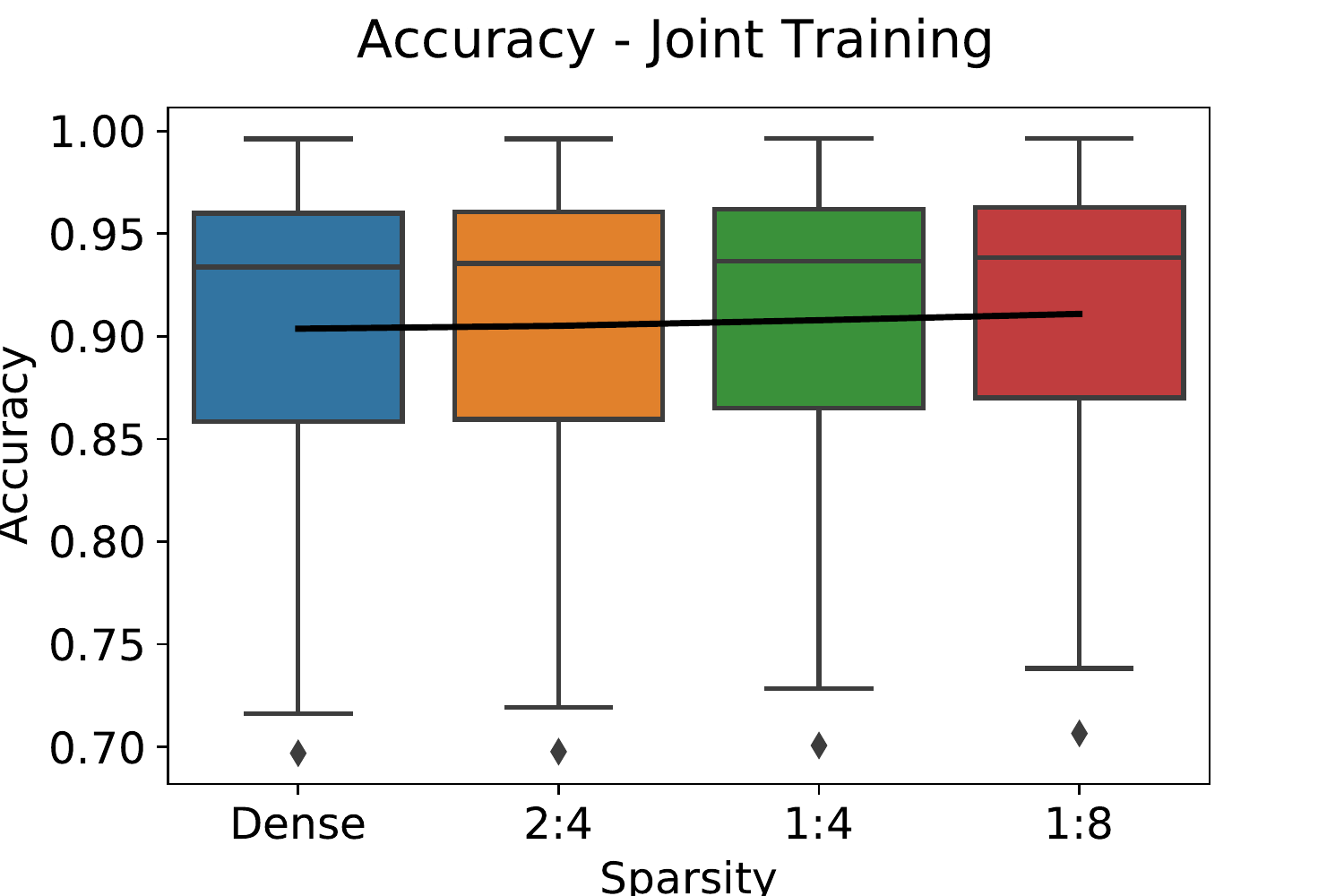} &
   \includegraphics[width=0.22\textwidth]{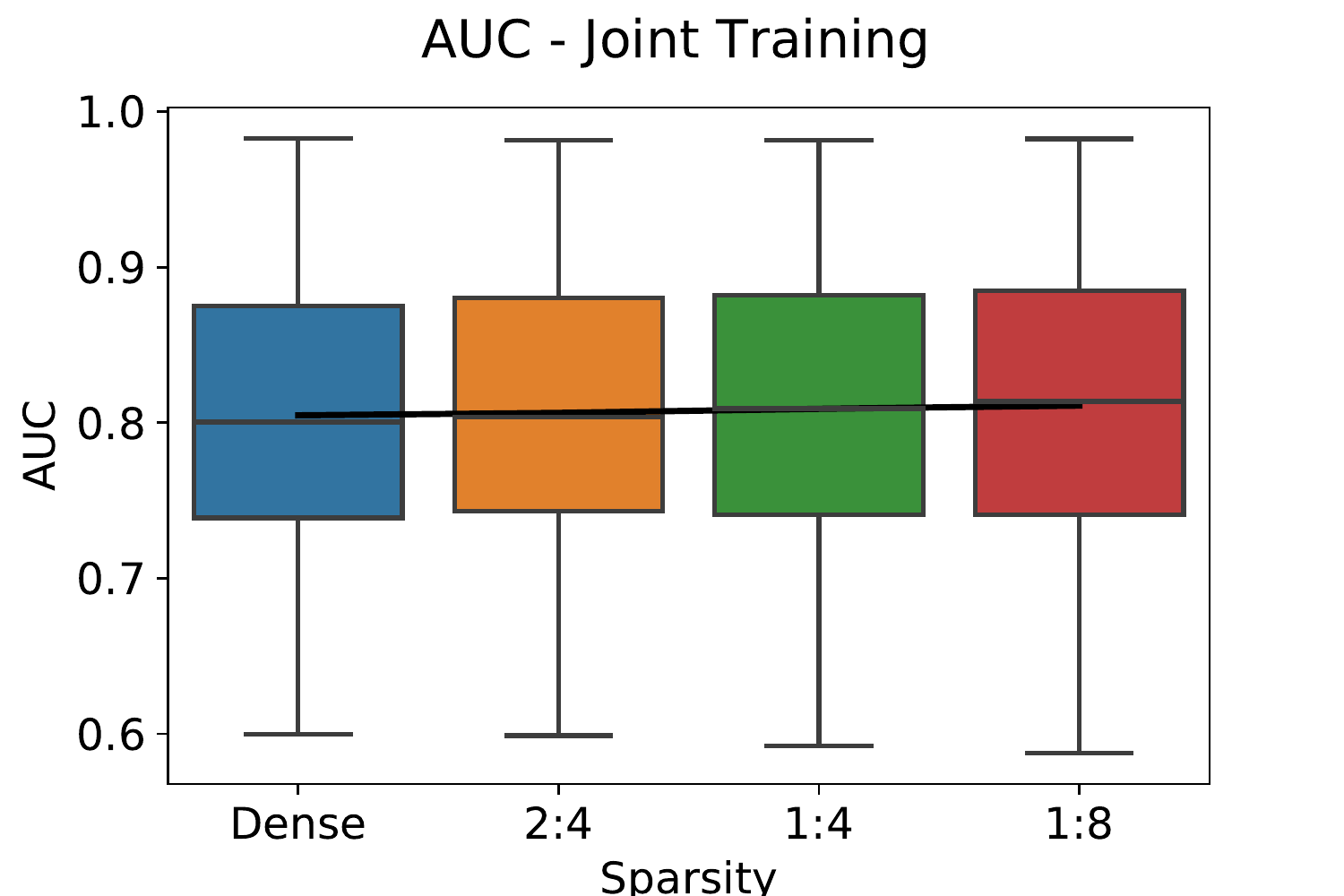} & & \\
    \includegraphics[width=0.22\textwidth]{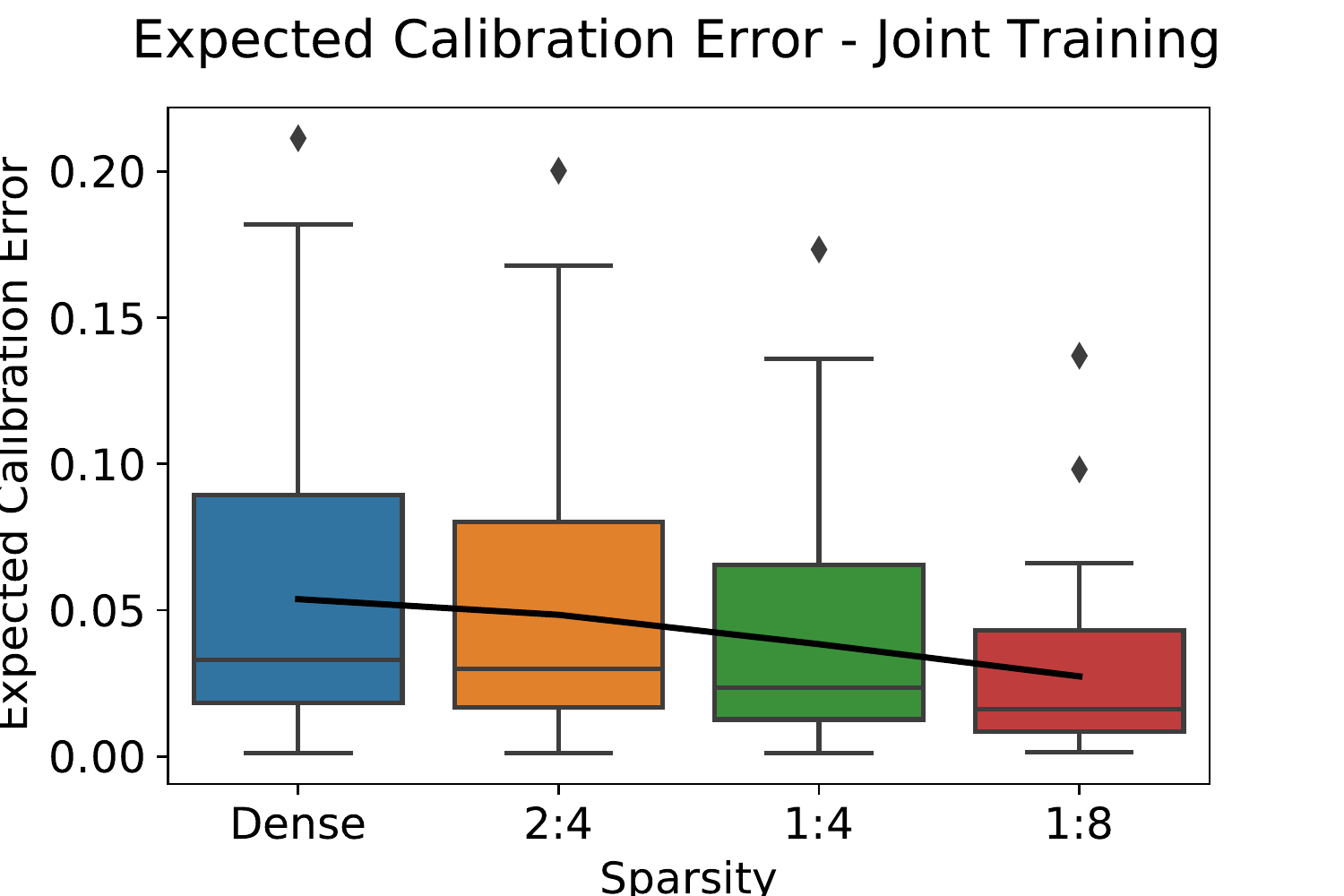} &
    \includegraphics[width=0.22\textwidth]{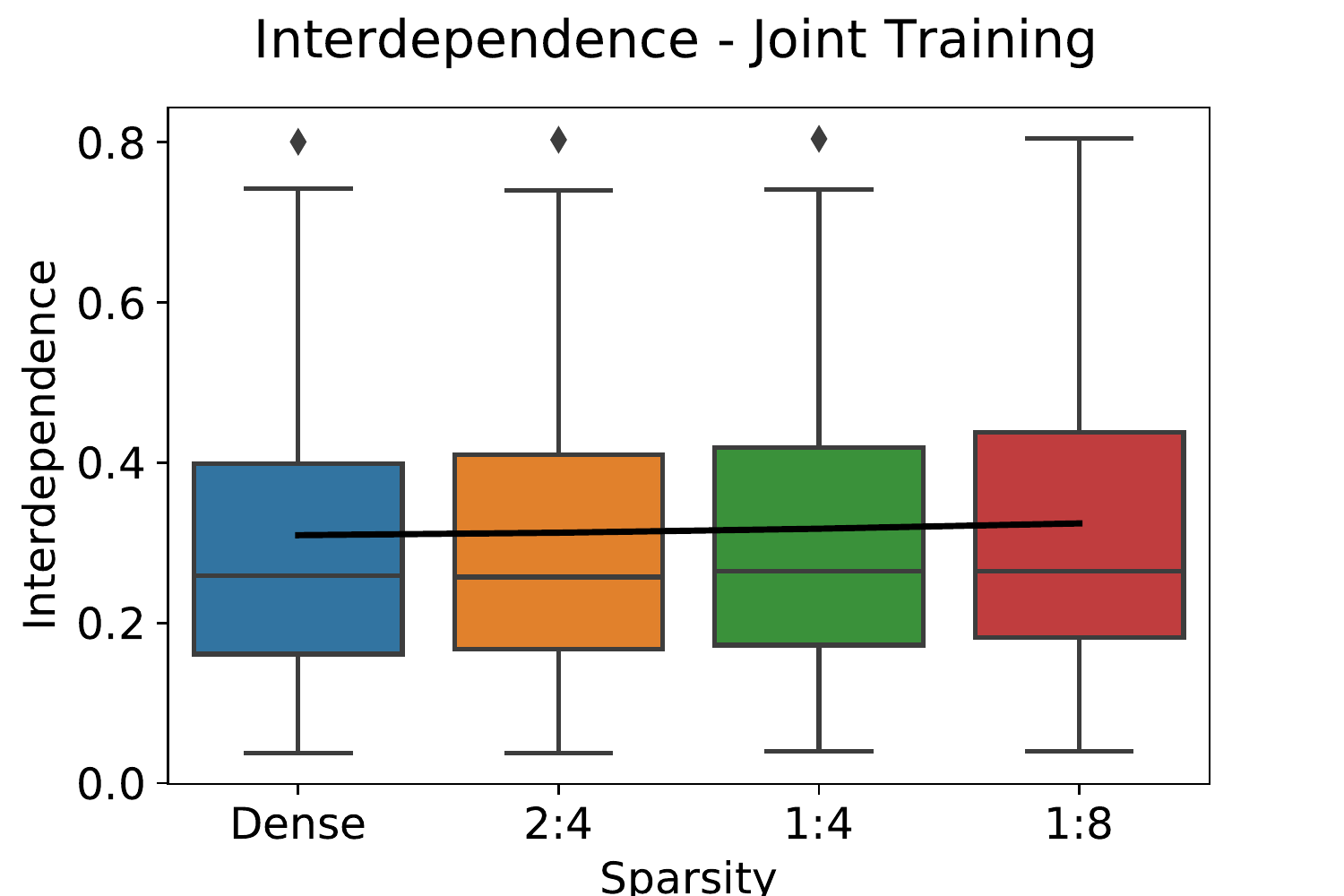} & 
    \includegraphics[width=0.22\textwidth]{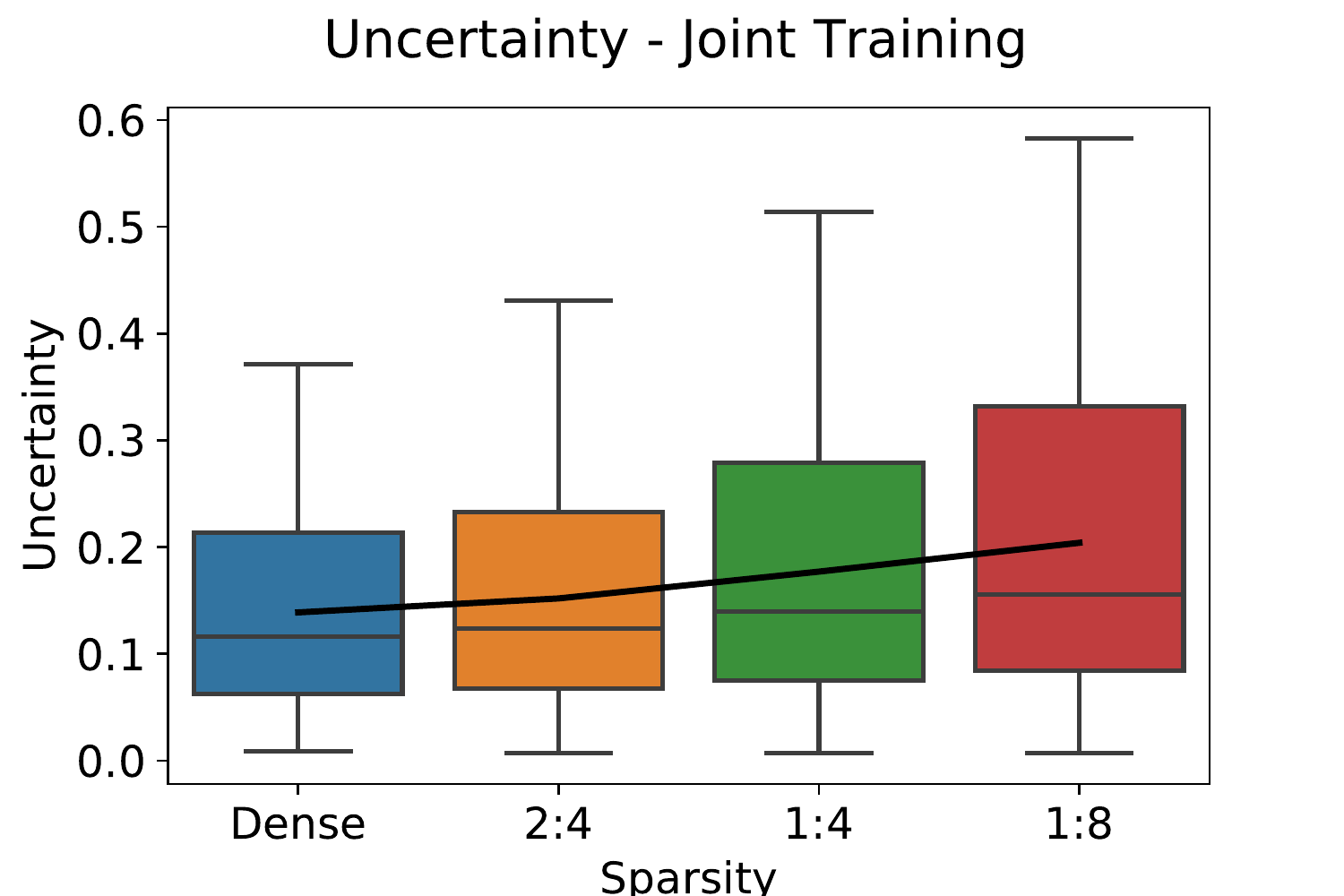} &
    \includegraphics[width=0.22\textwidth]{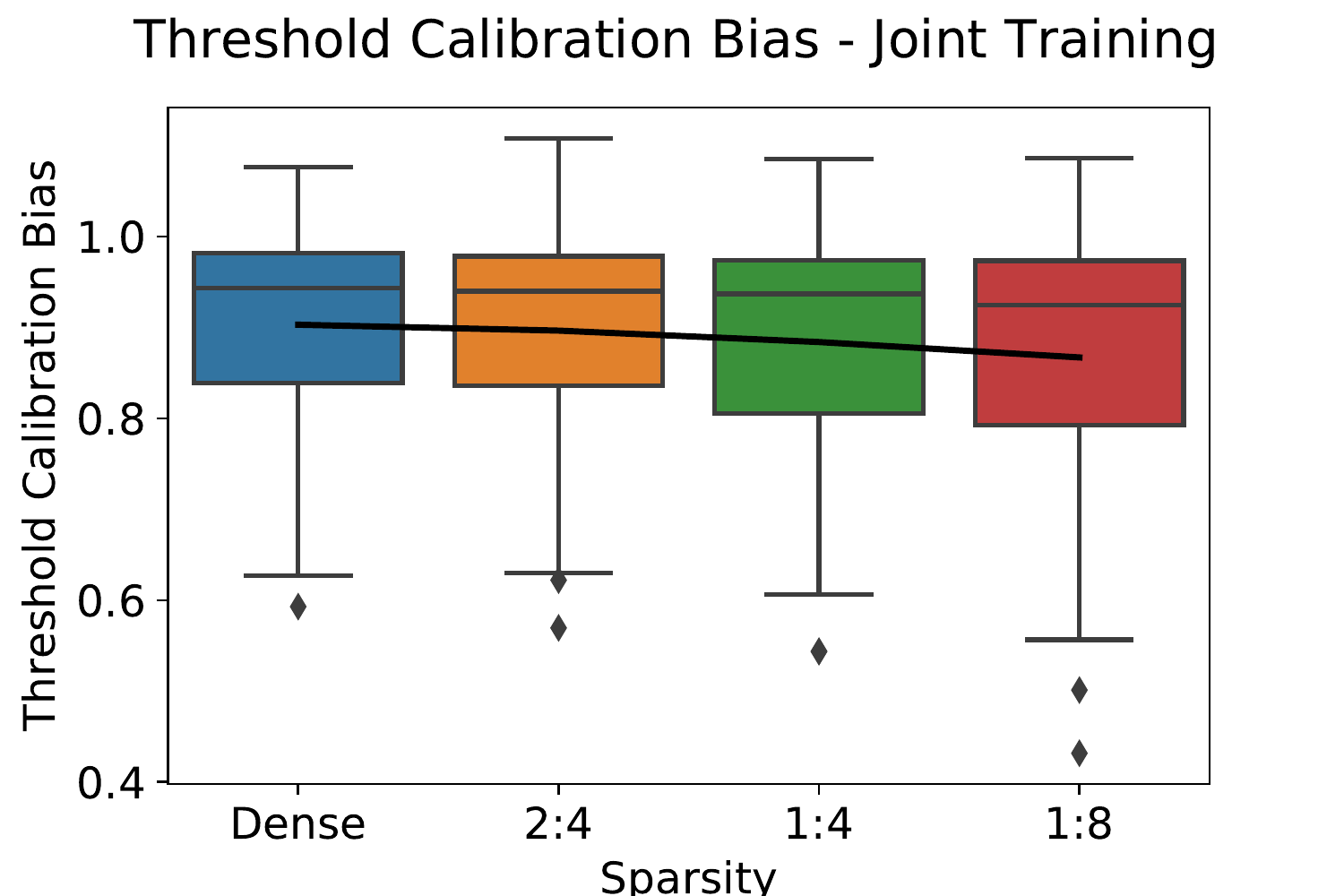}\\
    \includegraphics[width=0.22\textwidth]{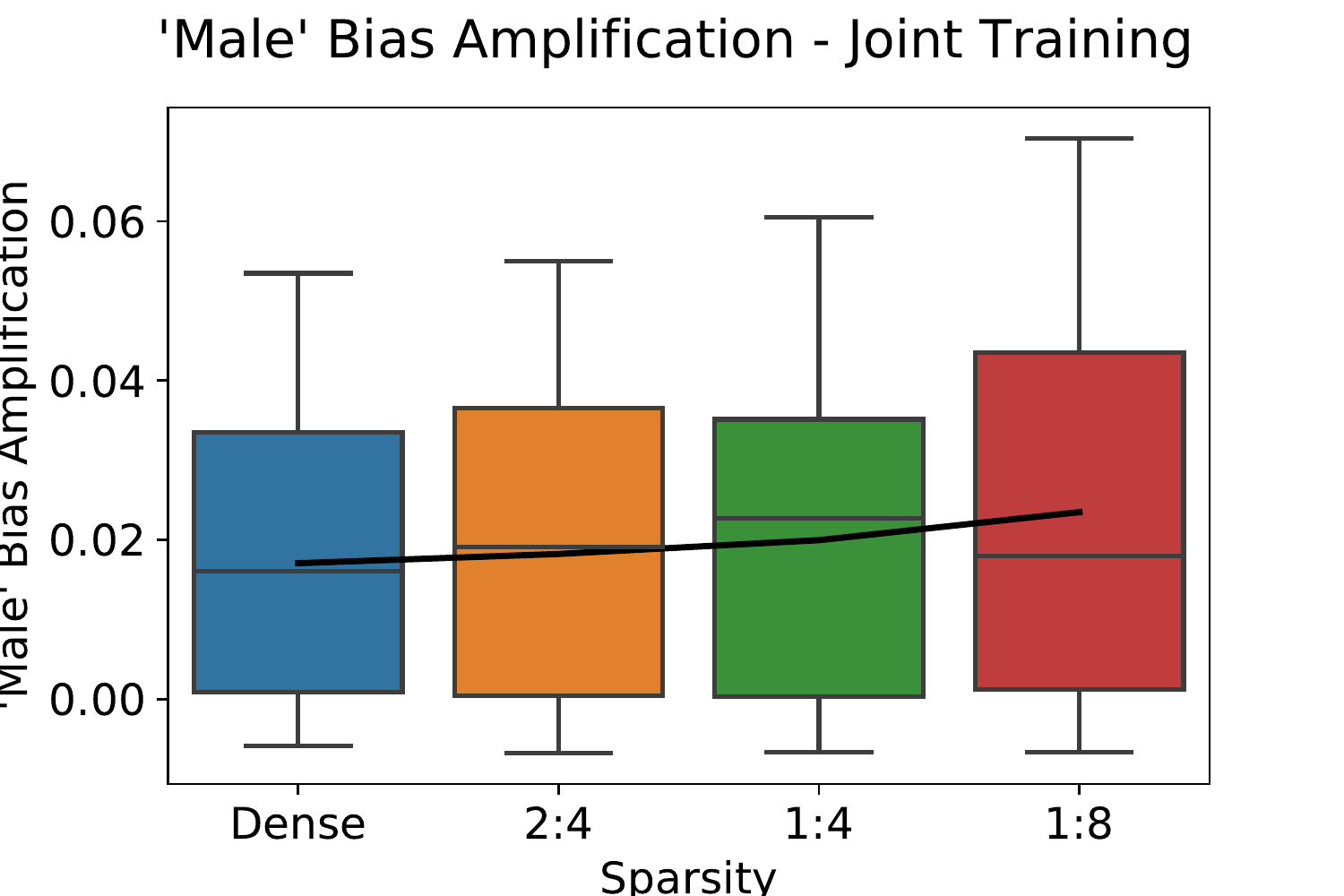} &
    \includegraphics[width=0.22\textwidth]{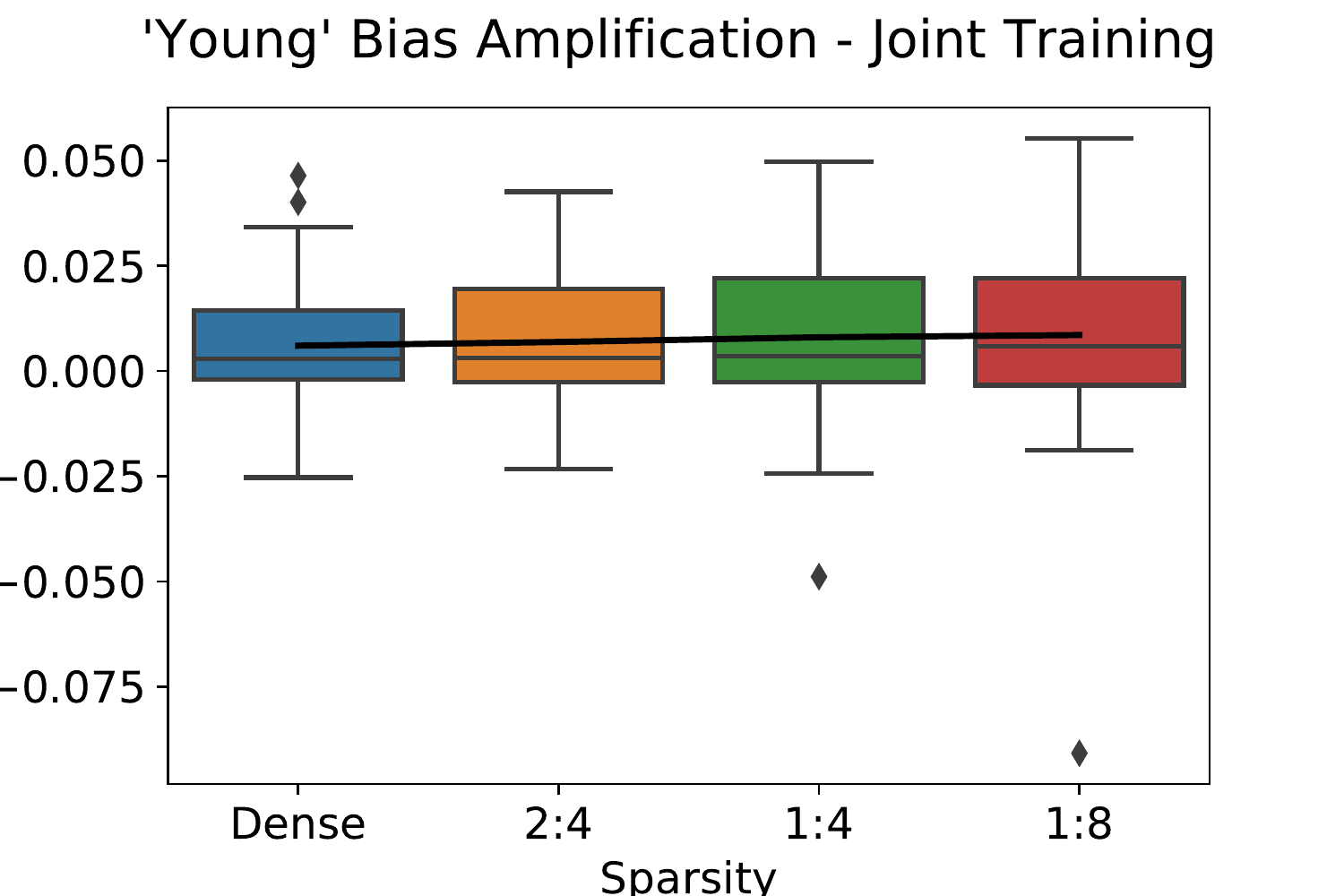} &
    \includegraphics[width=0.22\textwidth]{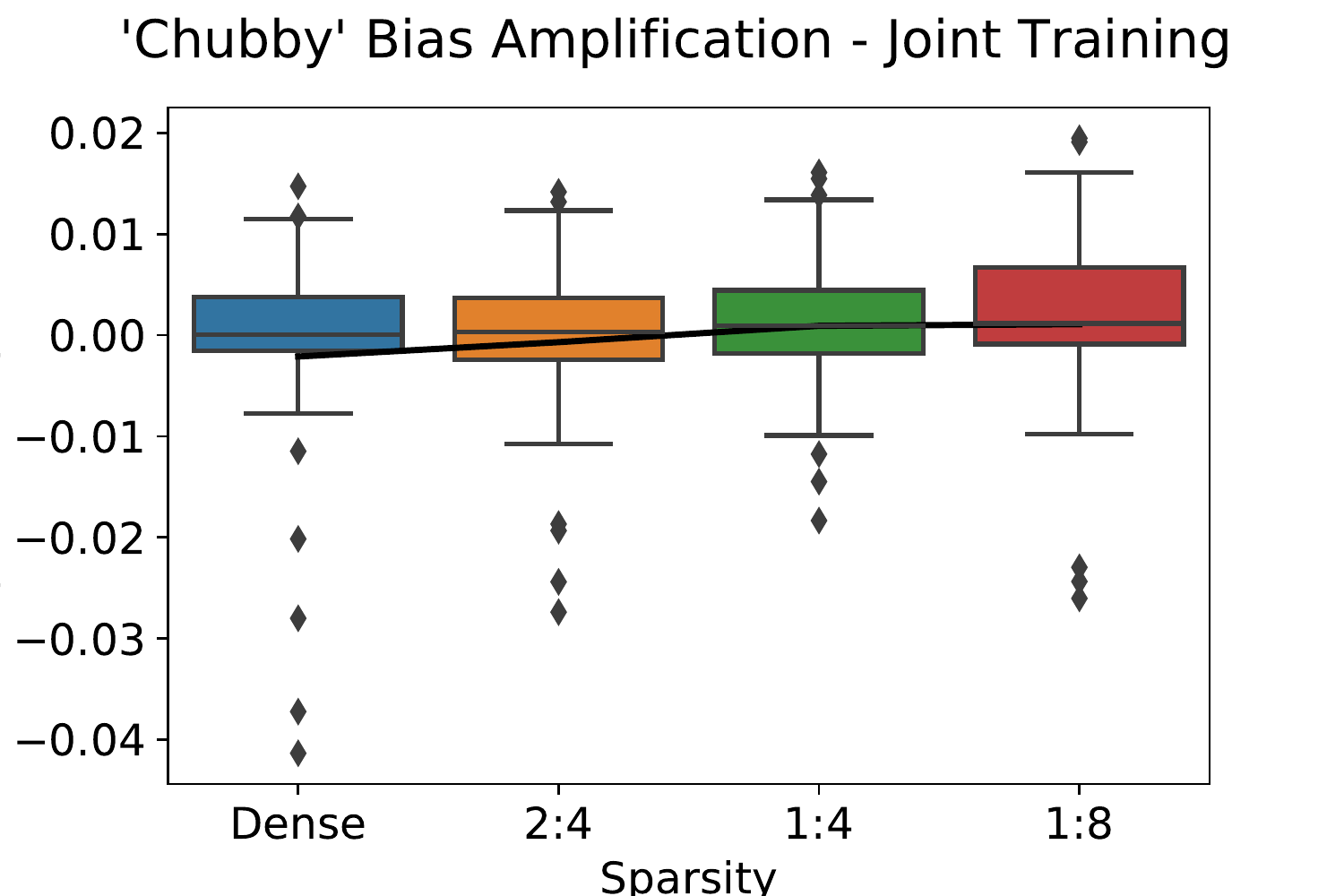} &
    \includegraphics[width=0.22\textwidth]{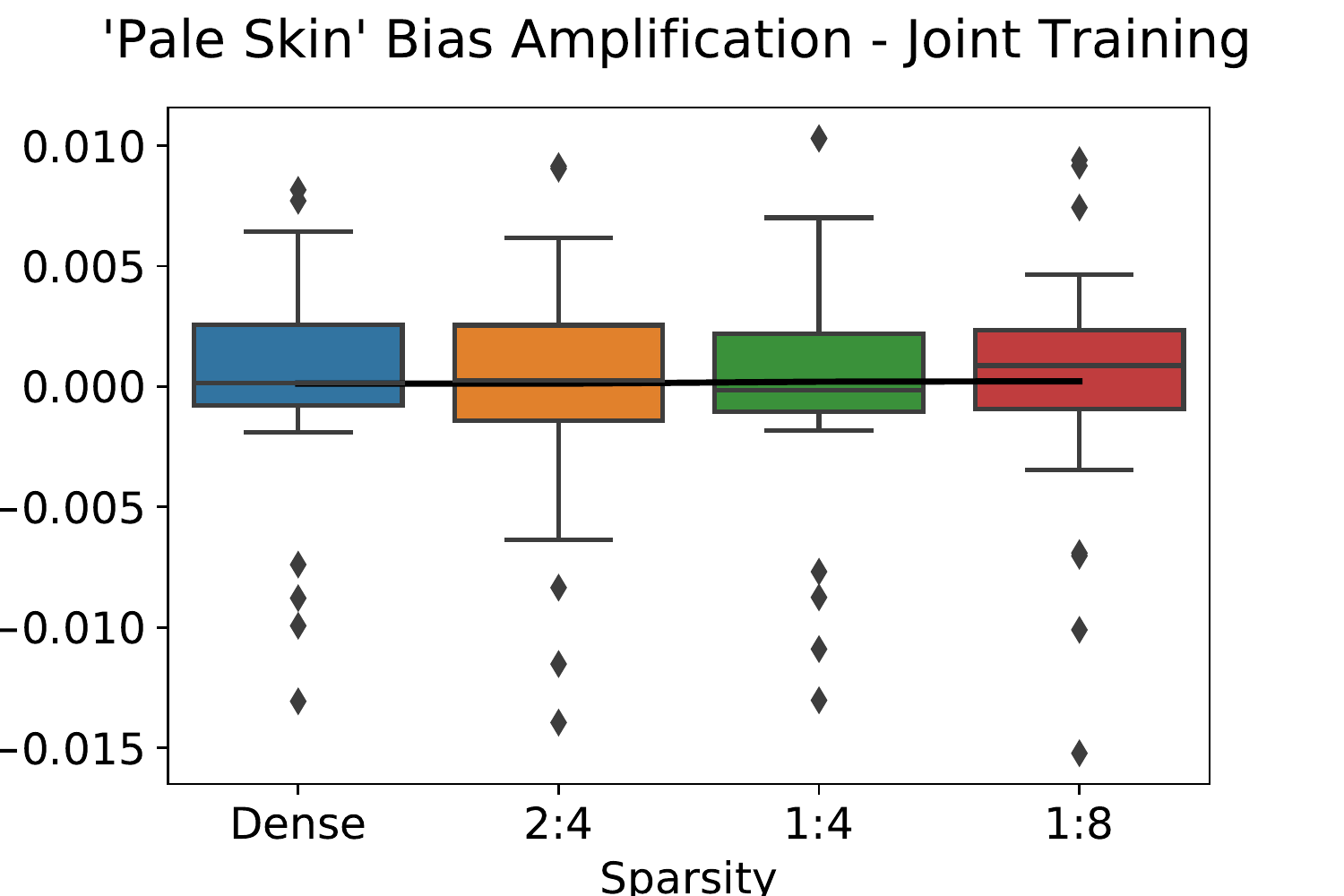}\\
\end{tabular}
    \caption{[CelebA / ResNet18/ N:M/ GMP-RI] Accuracy and Systematic Bias metrics (TCB, ECE, Interdependence) of MobileNetV1 models jointly trained on all CelebA attributes. The thick black line denotes the mean value at each sparsity level.
    }
    \label{fig:nm_celeba_rn18_joint_systematic}
\end{figure}

We present our results in Figure~\ref{fig:nm_celeba_rn18_joint_systematic}. As in our other experiments, we observe little effect on accuracy and AUC even at the highest 1:8 sparsity level; further, we observe that, as with unstructured sparsity, Expected Calibration Error decreases slightly with sparsity, while Uncertainty increases and Threshold Calibration Bias gets slightly worse. As far as Bias Amplification, we observe a slight increase when splitting the data by the Male category, for the 1:4 and 1:8 sparsity pattern. Splitting by the other three categories (Young, Chubby, and Pale Skin) shows minimal, if any, increased BA, likely because even at the highest 1:8 sparsity level, the model is less than 90\% sparse, as compared with up to 99.5\% sparsity for unstructured pruning. We note that this further validates our finding that ResNet18 models predicting CelebA attributes can be pruned to fairly high sparsity without significant effect on BA.

\begin{figure}[ht]
\centering
\begin{tabular}{cc}
  \includegraphics[width=0.35\textwidth]{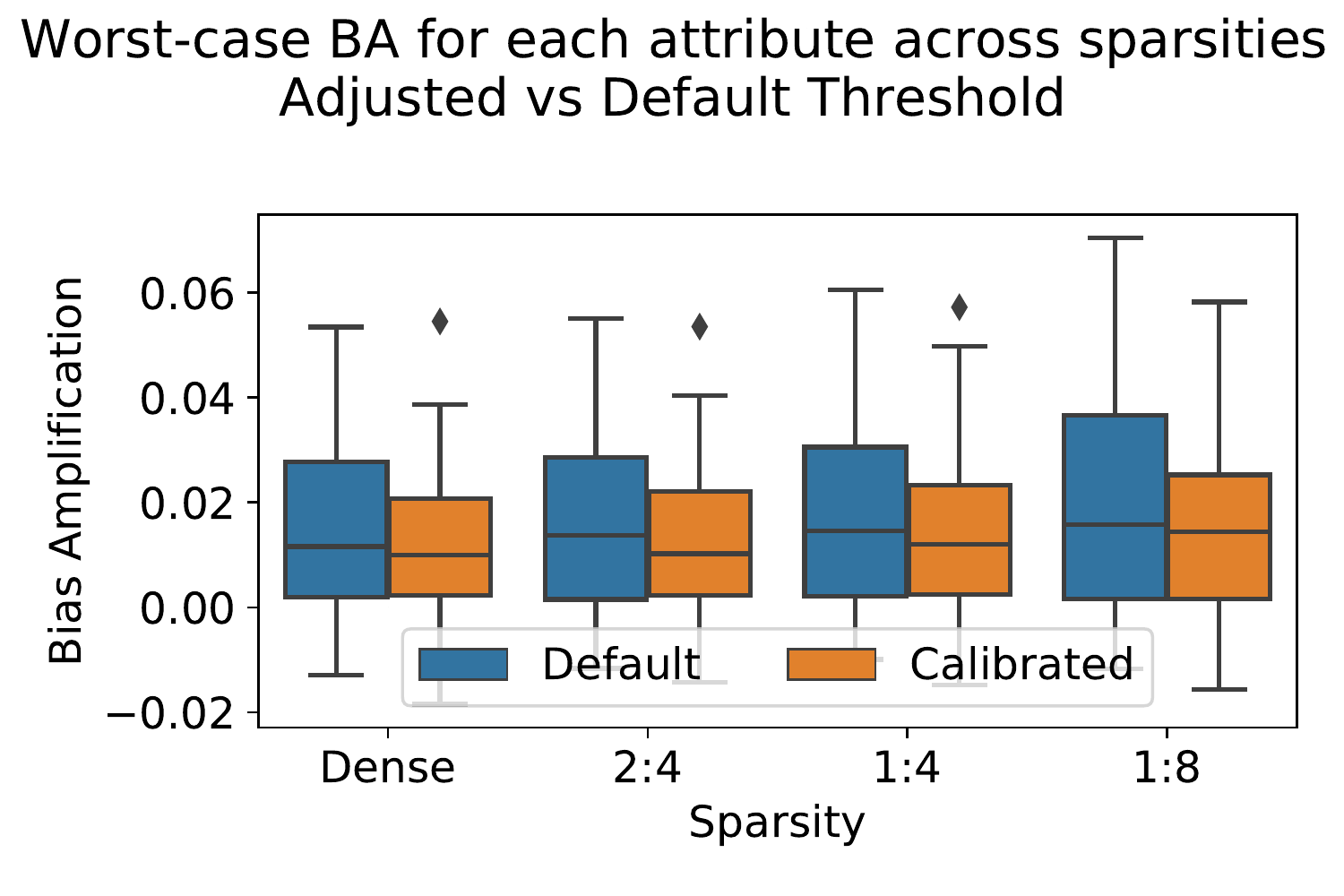} &
  \includegraphics[width=0.35\textwidth]{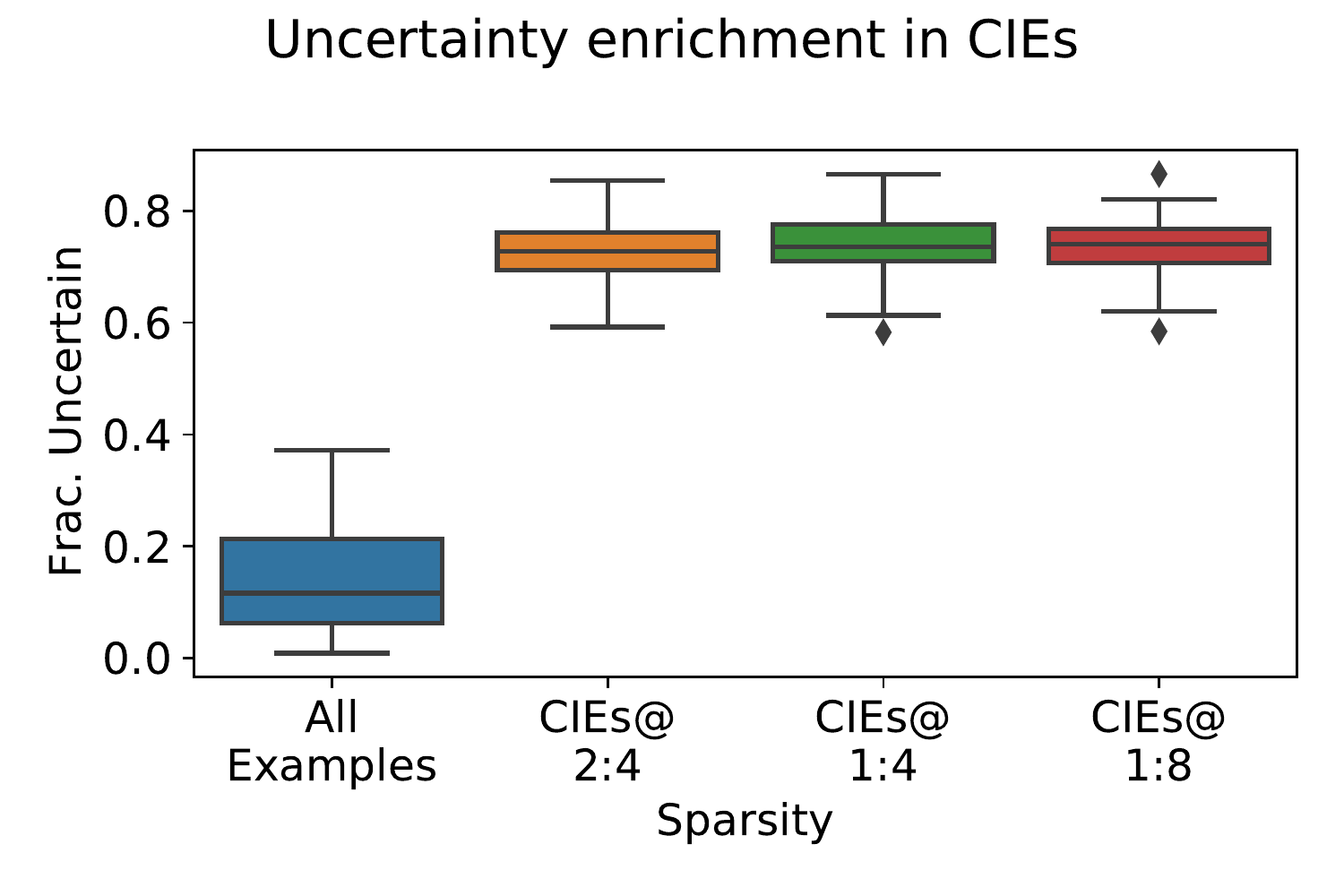}
  \end{tabular}
    \caption{[CelebA / ResNet18 / N:M Sparsity / GMP-RI] (Left) Effect of threshold calibration on ResNet18 N:M sparsity models jointly trained on all attributes. %
    (Right) Proportion of uncertain predictions for \emph{dense} models across all attributes for all elements in the CelebA test set, and for Compression-Identified Exemplars at different sparsities.}
    \label{fig:celeba_rn18_nm_threshold_adj}
\end{figure}

\begin{figure}[h]
\centering
\includegraphics[width=0.8\textwidth]{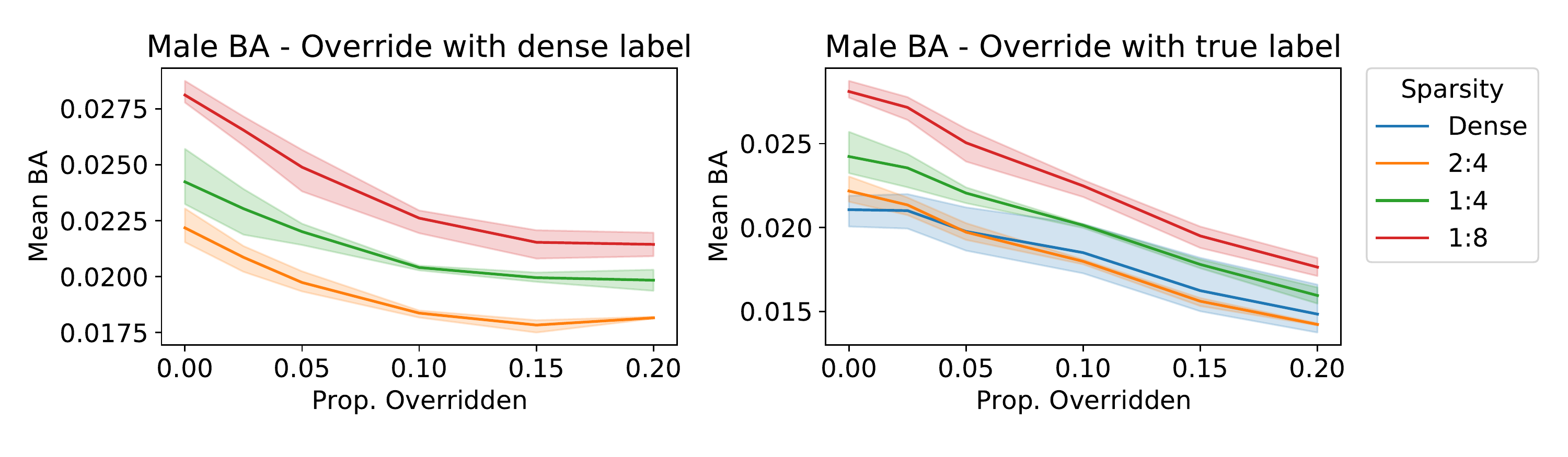}
\includegraphics[width=0.8\textwidth]{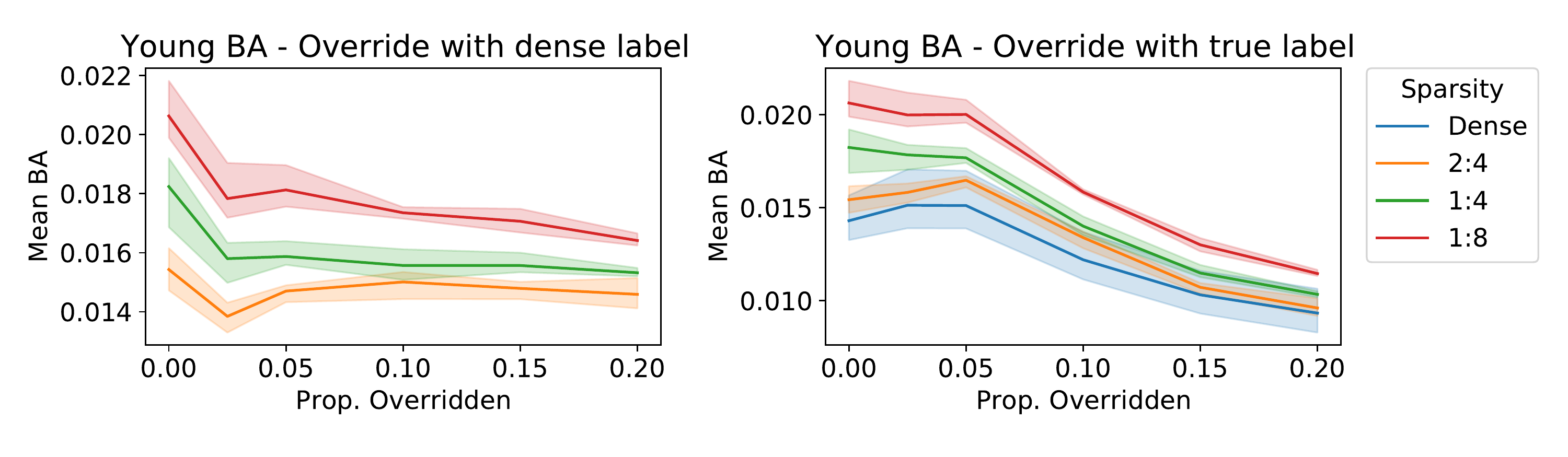}
\includegraphics[width=0.8\textwidth]{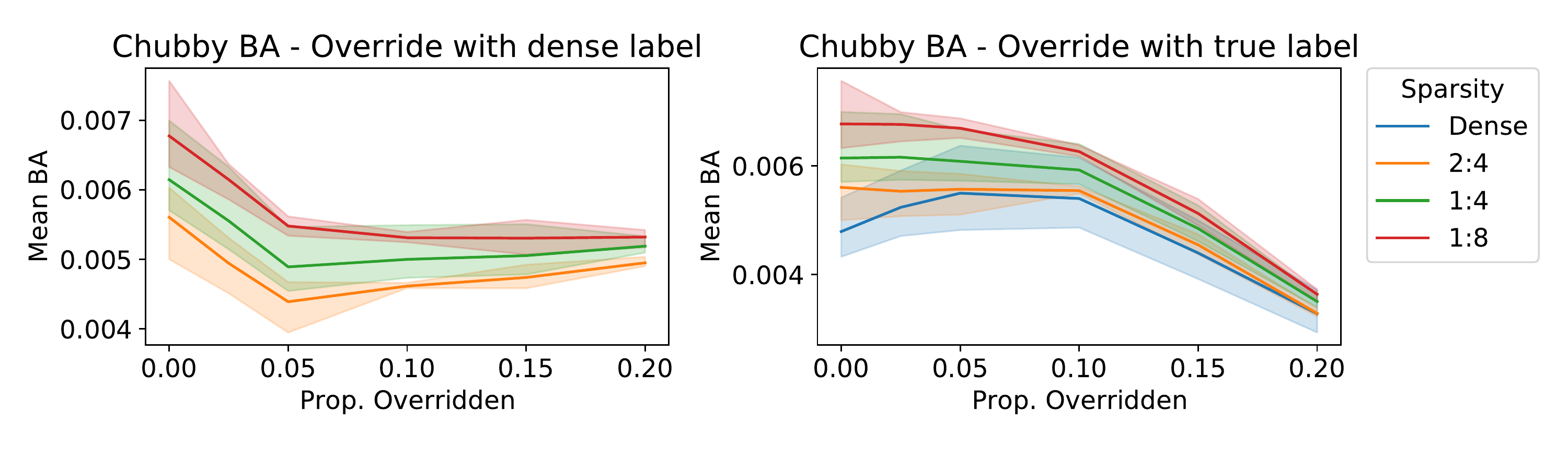}
\includegraphics[width=0.8\textwidth]{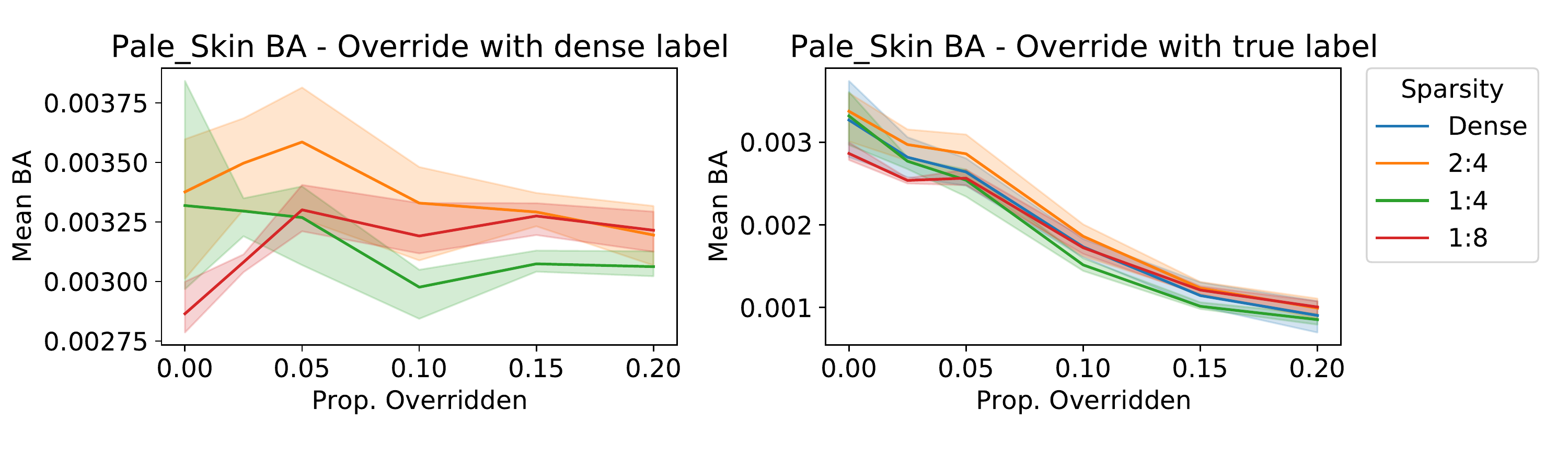}
    \caption{[CelebA / ResNet18 / N:M Sparsity / GMP-RI] Effect of label overrides on Bias Amplification. In all cases, overrides are prioritized by dense model uncertainty.}
    \label{fig:overrides_nm}
\end{figure}

\clearpage
\section{MobileNetV1 results}
\label{appendix:mobilenet}

We additionally validate our results on different architectures, for the joint label training setting. Namely, we choose MobileNet~\cite{howard2017mobilenets}, as it is a smaller model, and known to be more difficult to prune. We train the dense and sparse models using the same hyperparameters described in Appendix Section~\ref{appendix:training_settings}. We show results under the GMP-RI setting.

\begin{figure}[h]
    \centering
\begin{tabular}{cccc}
   \includegraphics[width=0.22\textwidth]{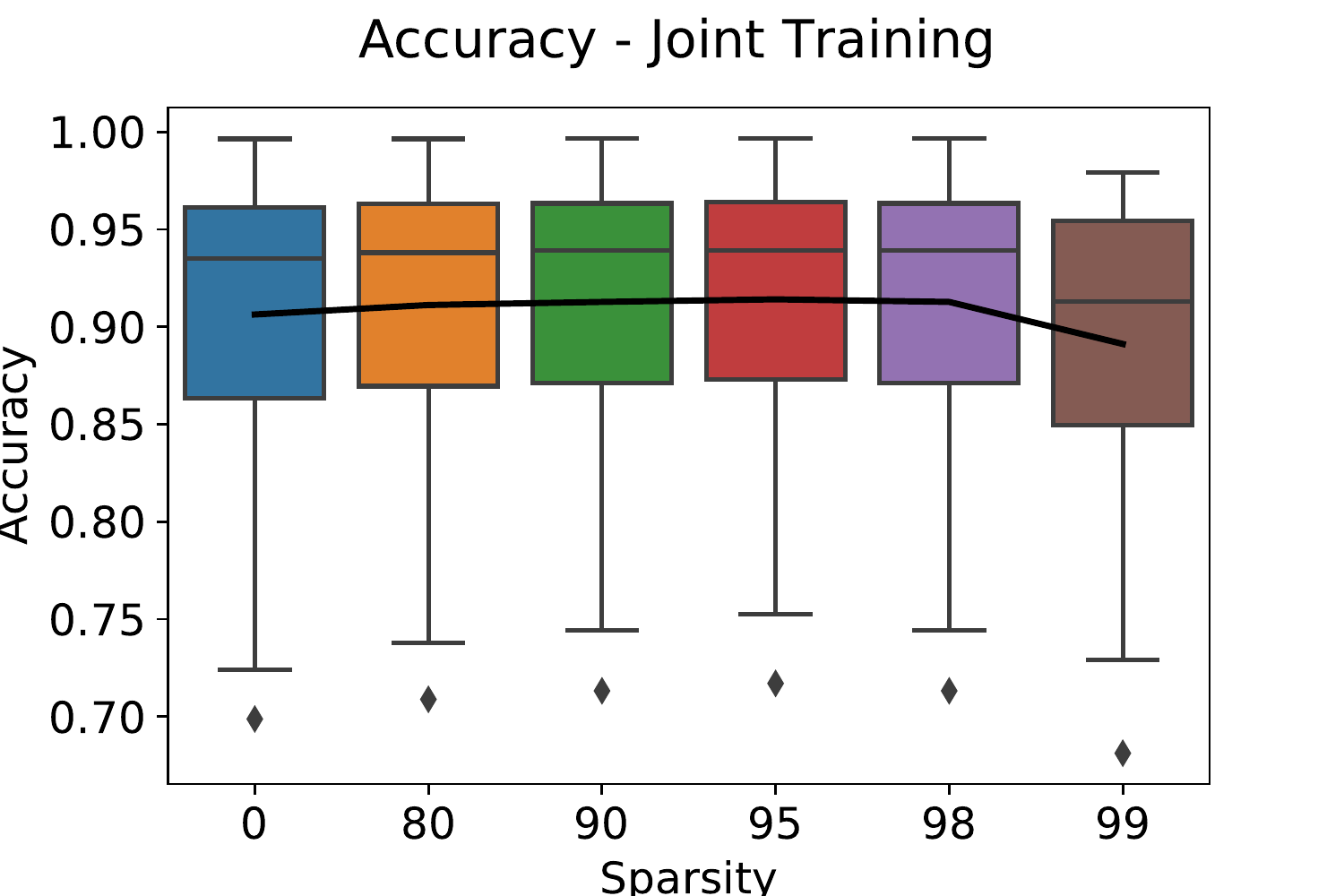} &
   \includegraphics[width=0.22\textwidth]{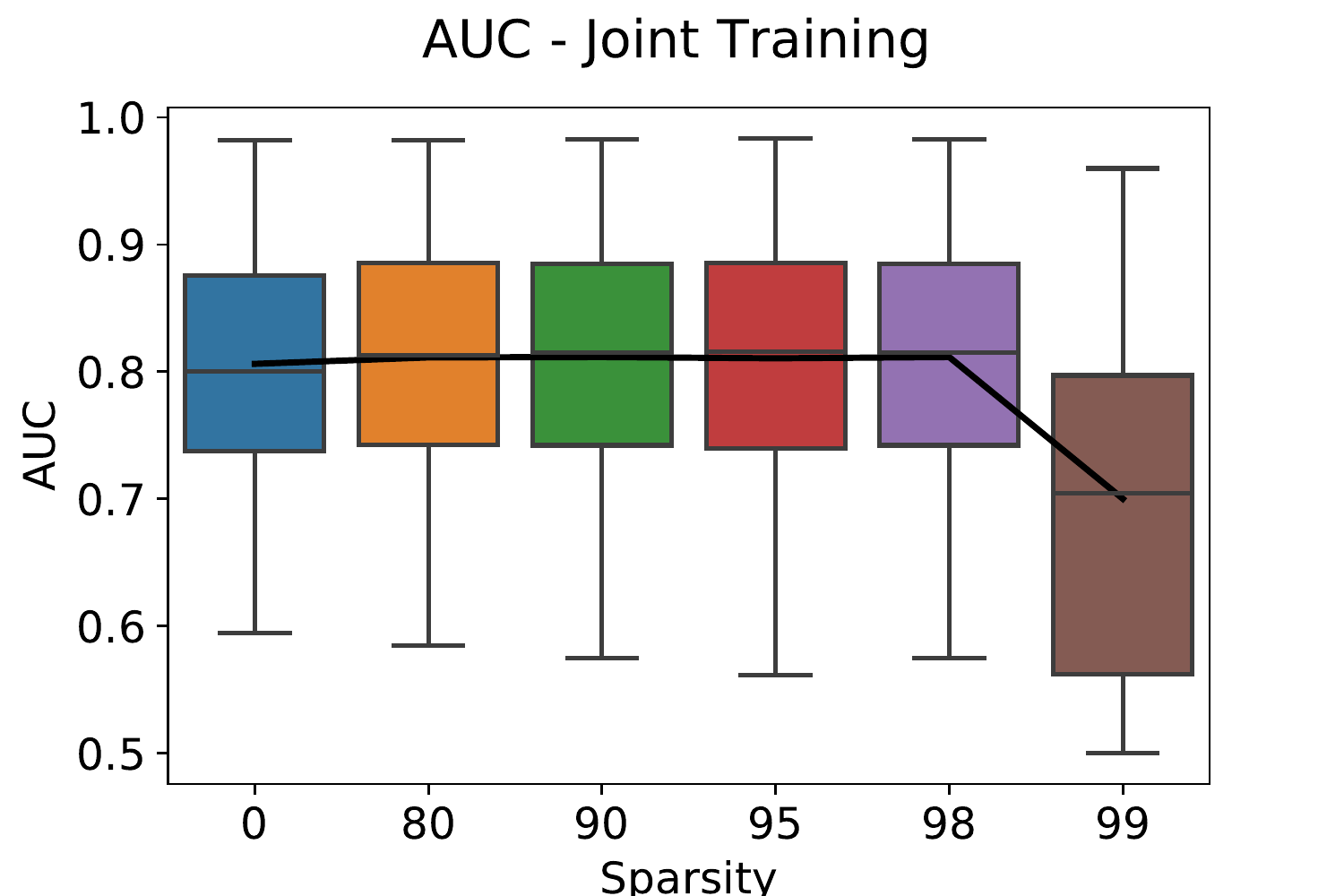} & & \\
    \includegraphics[width=0.22\textwidth]{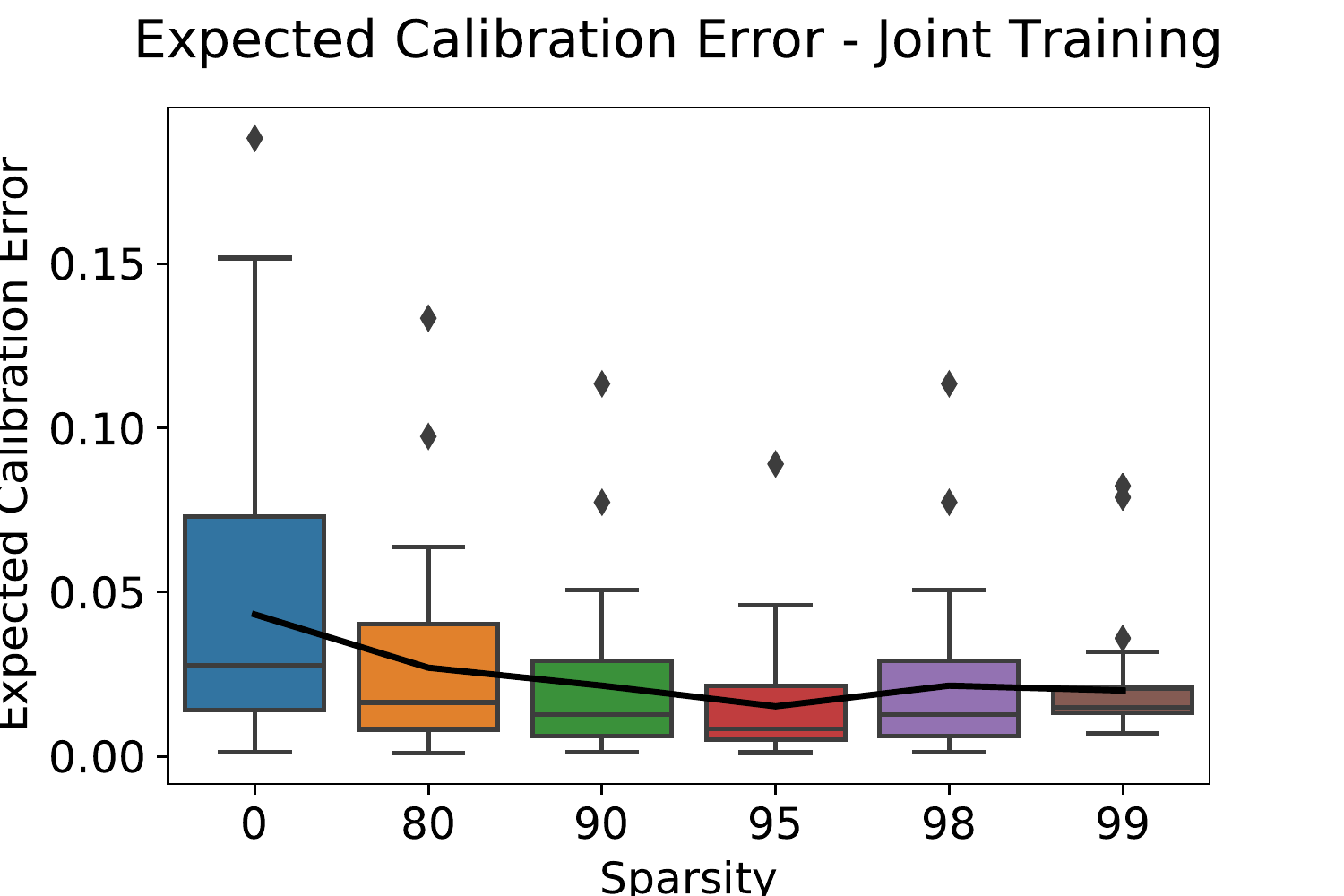} &
    \includegraphics[width=0.22\textwidth]{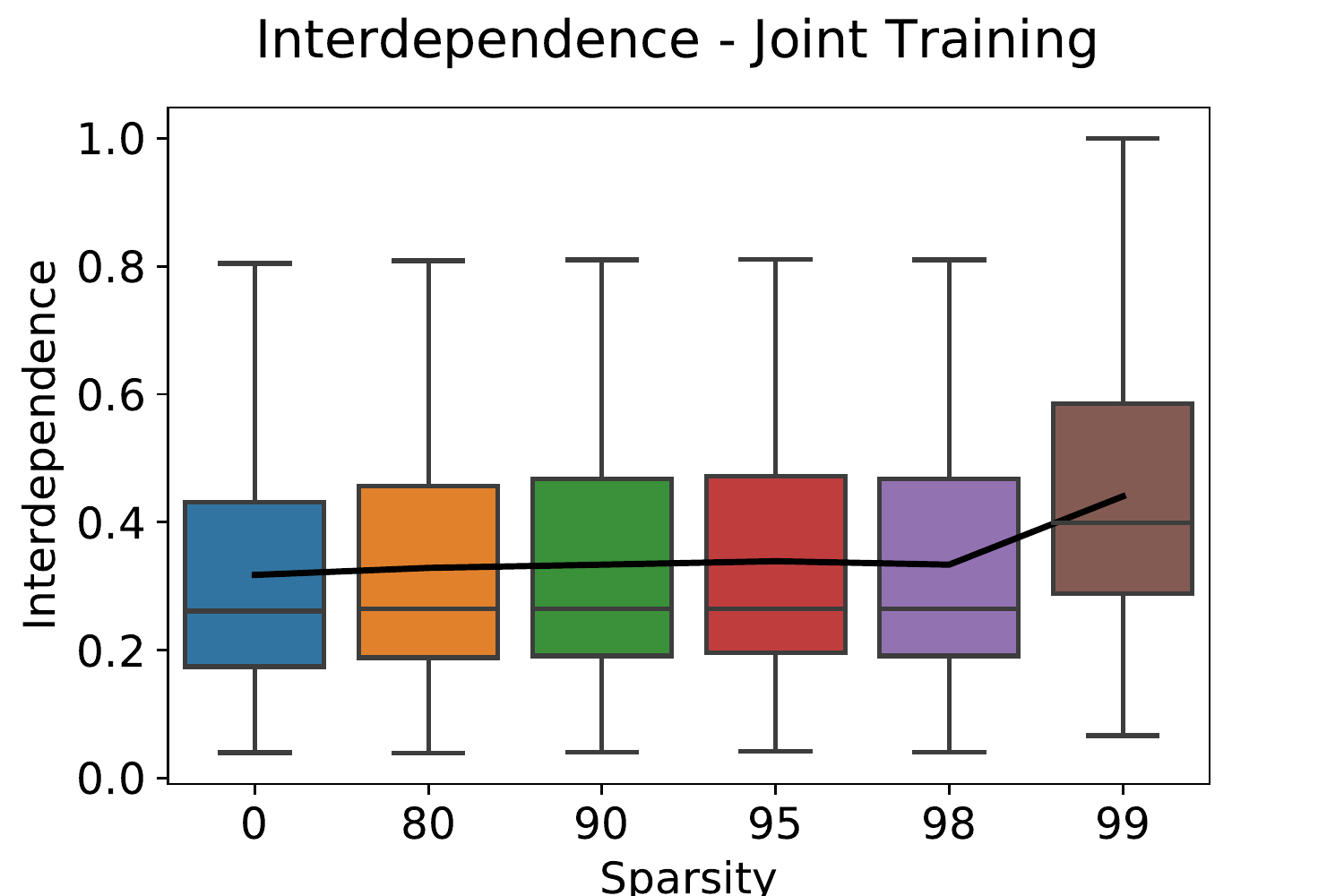} & 
    \includegraphics[width=0.22\textwidth]{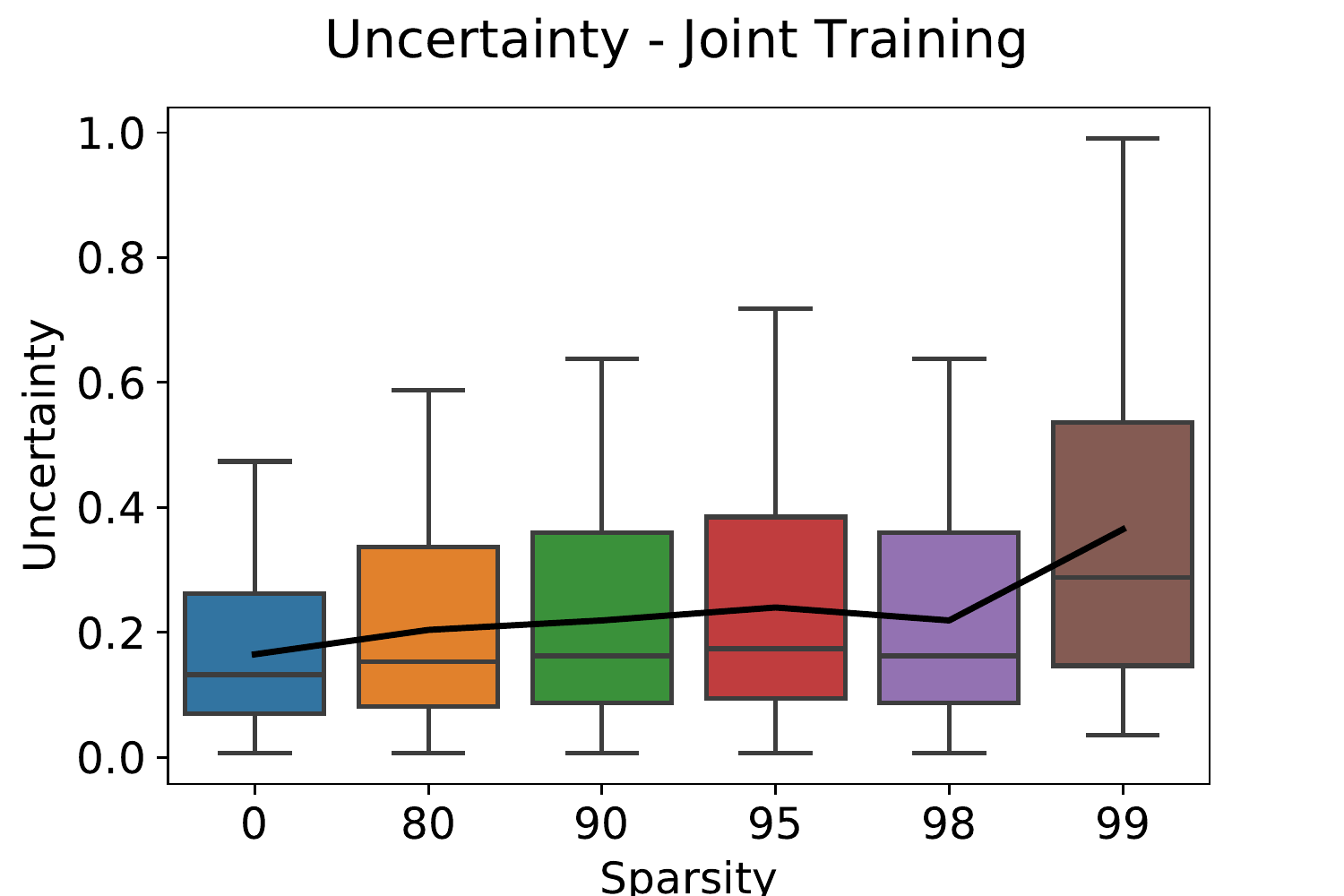} &
    \includegraphics[width=0.22\textwidth]{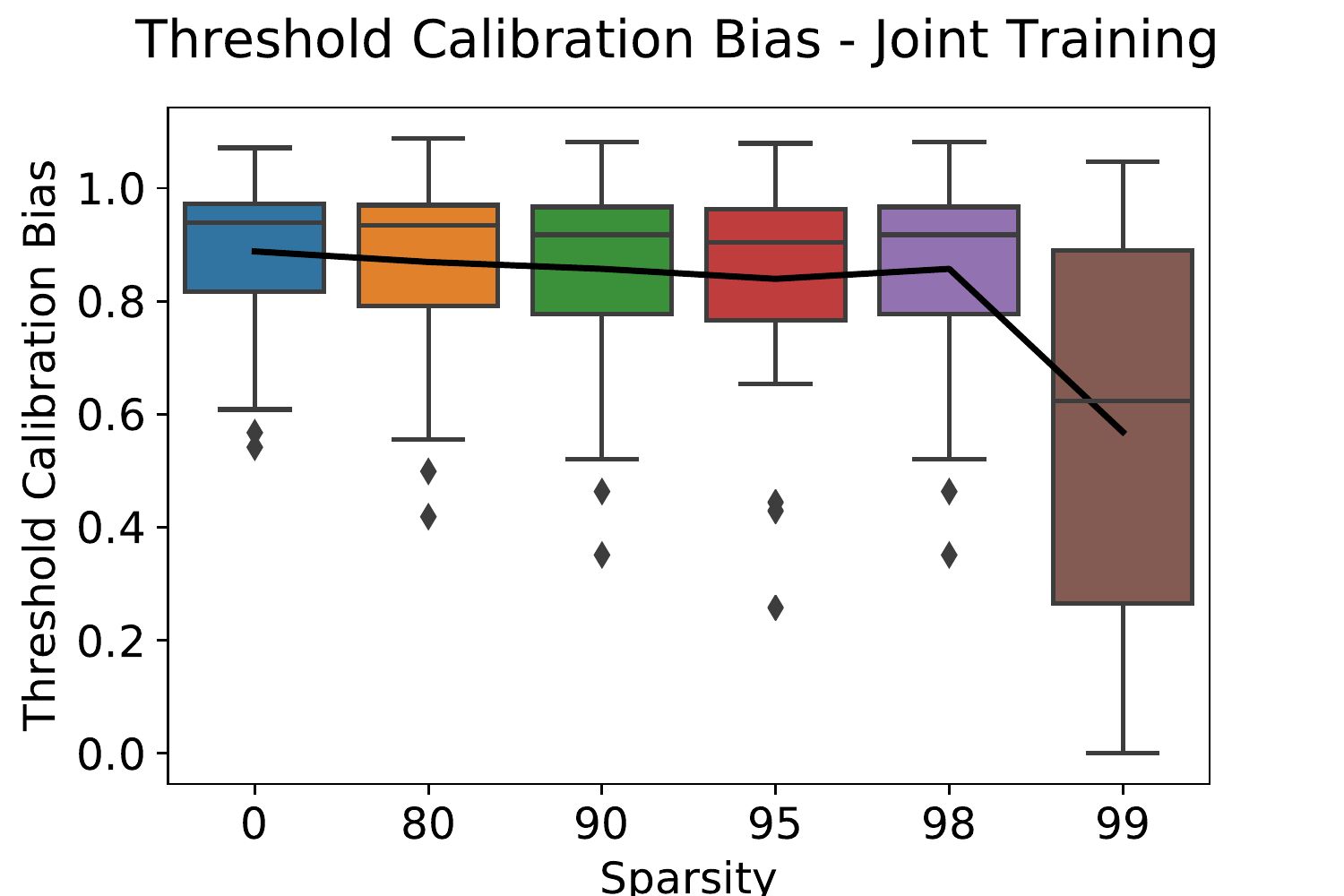}\\
    \includegraphics[width=0.22\textwidth]{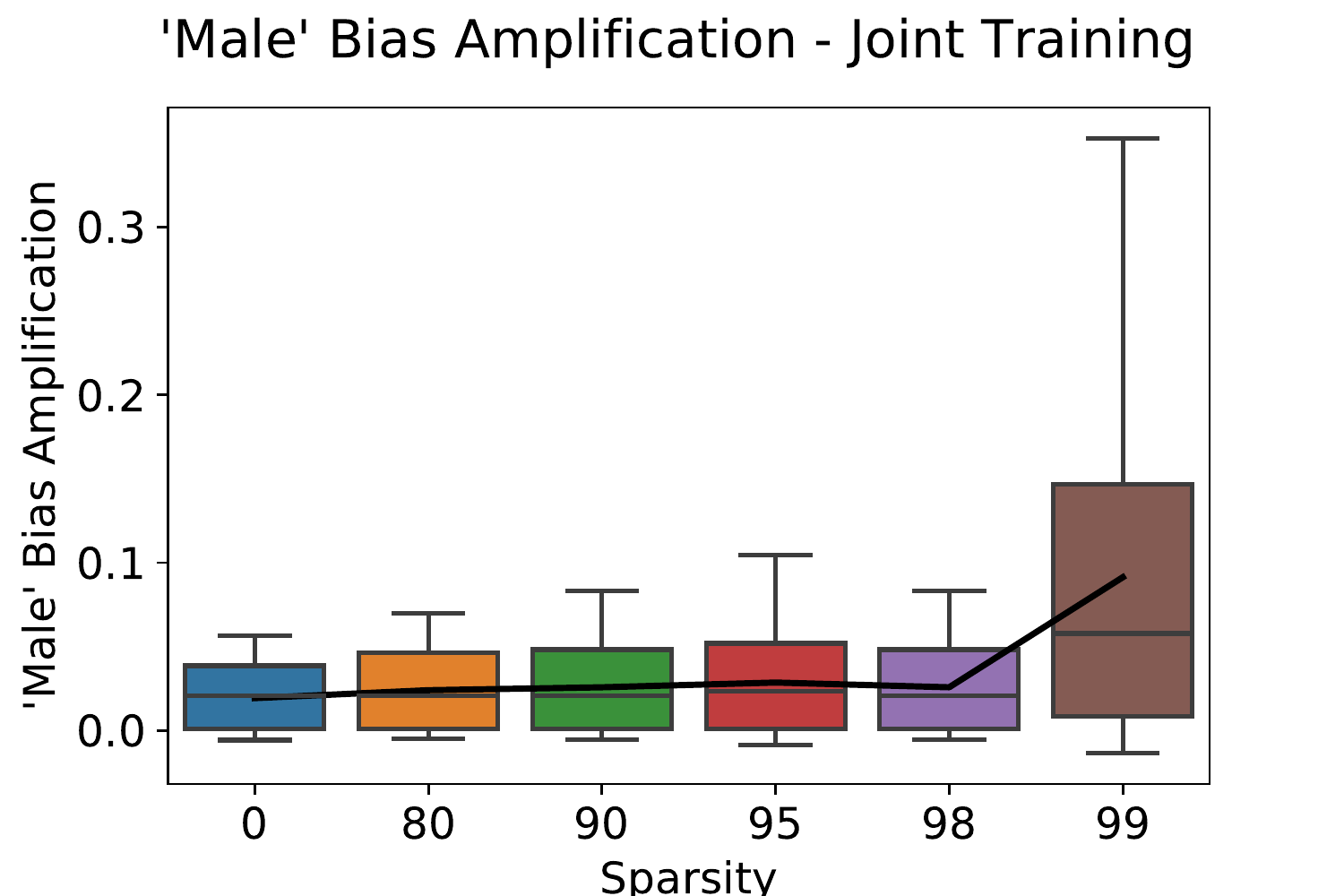} &
    \includegraphics[width=0.22\textwidth]{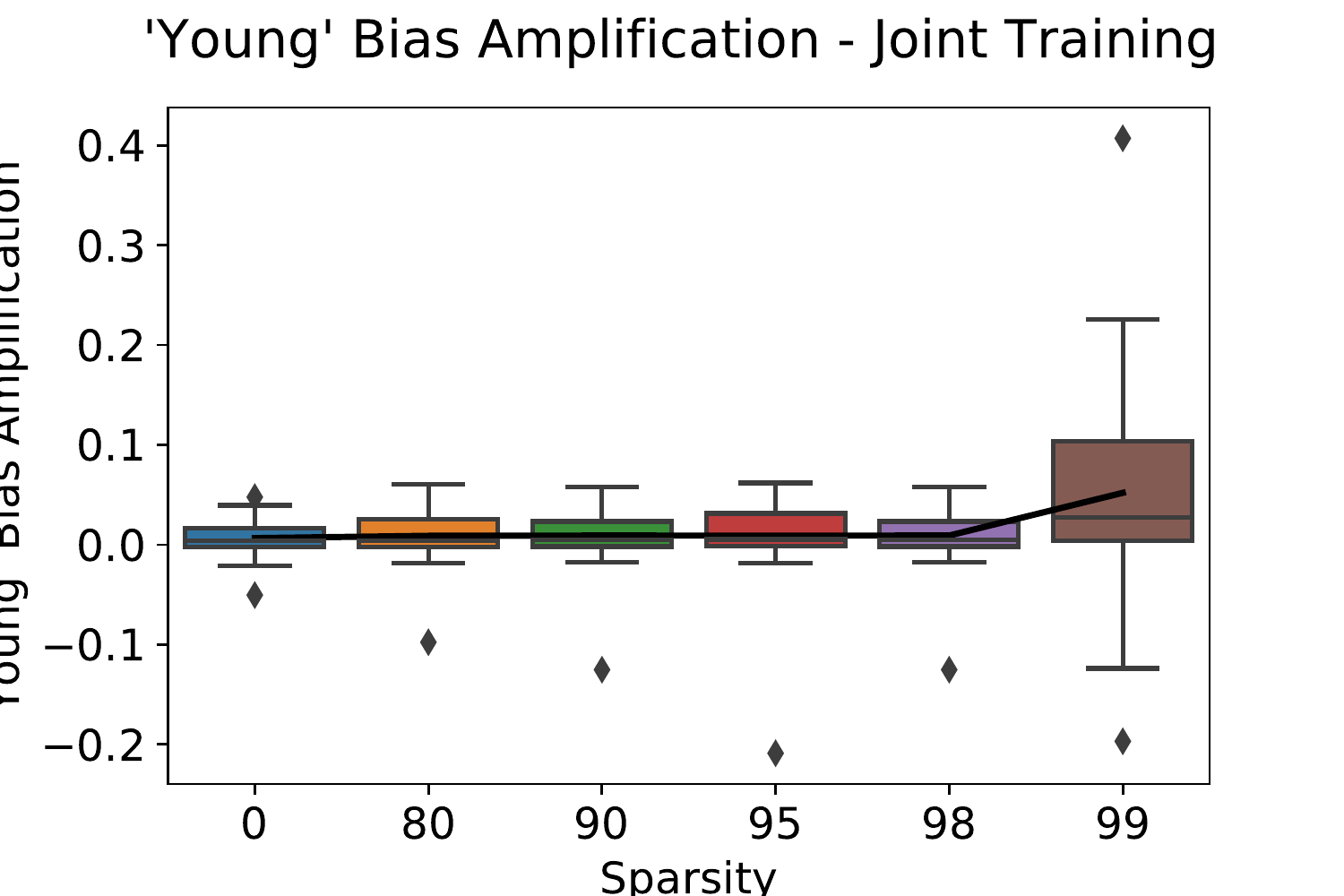} &
        \includegraphics[width=0.22\textwidth]{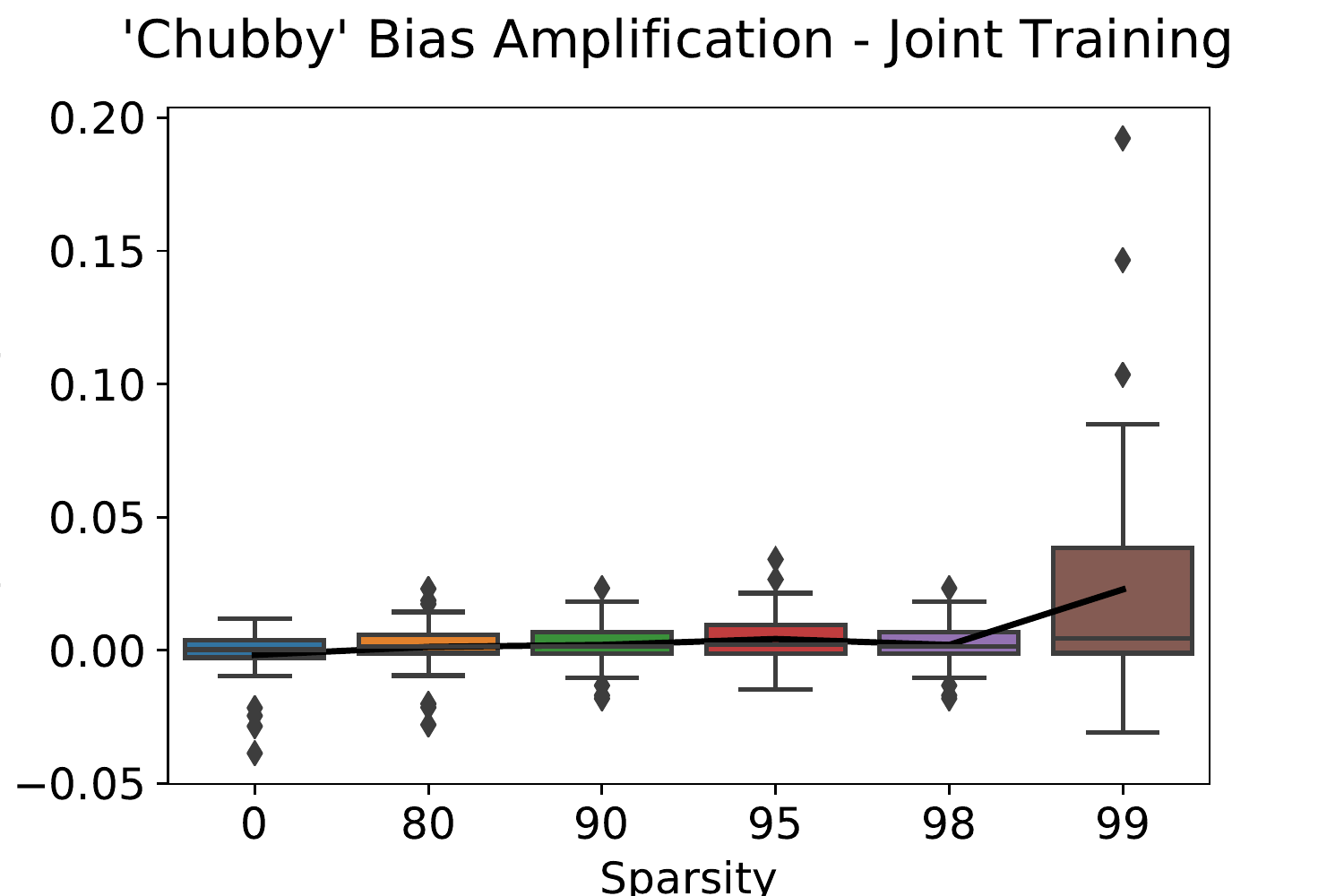} &
    \includegraphics[width=0.22\textwidth]{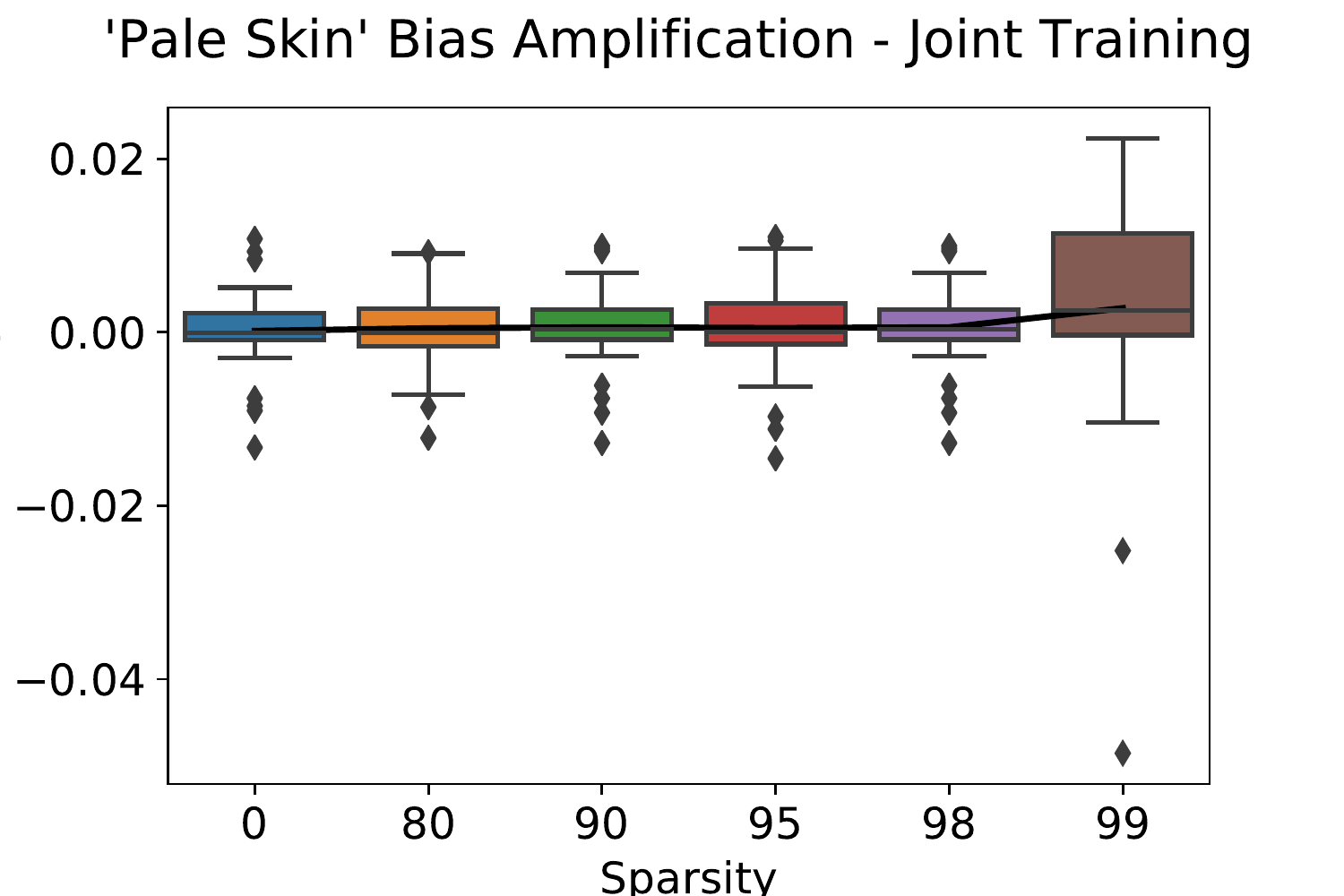}\\
\end{tabular}
    \caption{[CelebA / MobileNetV1 / GMP-RI] Accuracy and Systematic Bias metrics (TCB, ECE, Interdependence) of MobileNetV1 models jointly trained on all CelebA attributes. The thick black line denotes the mean value at each sparsity level.
    }
    \label{fig:celeba_mobilenet_joint_systematic}
\end{figure}

\begin{figure}[ht]
\centering
\begin{tabular}{cc}
  \includegraphics[width=0.35\textwidth]{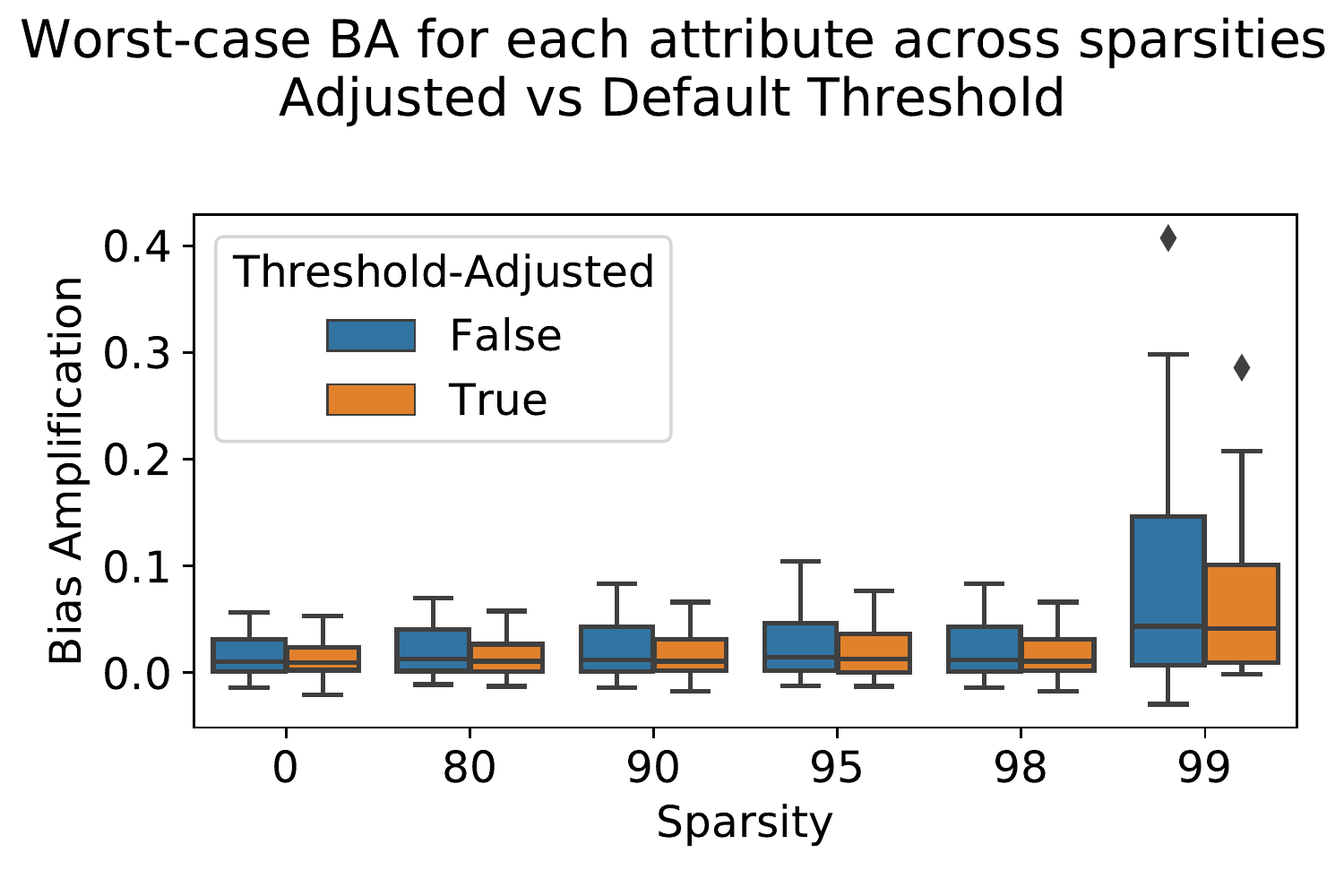} &
  \includegraphics[width=0.35\textwidth]{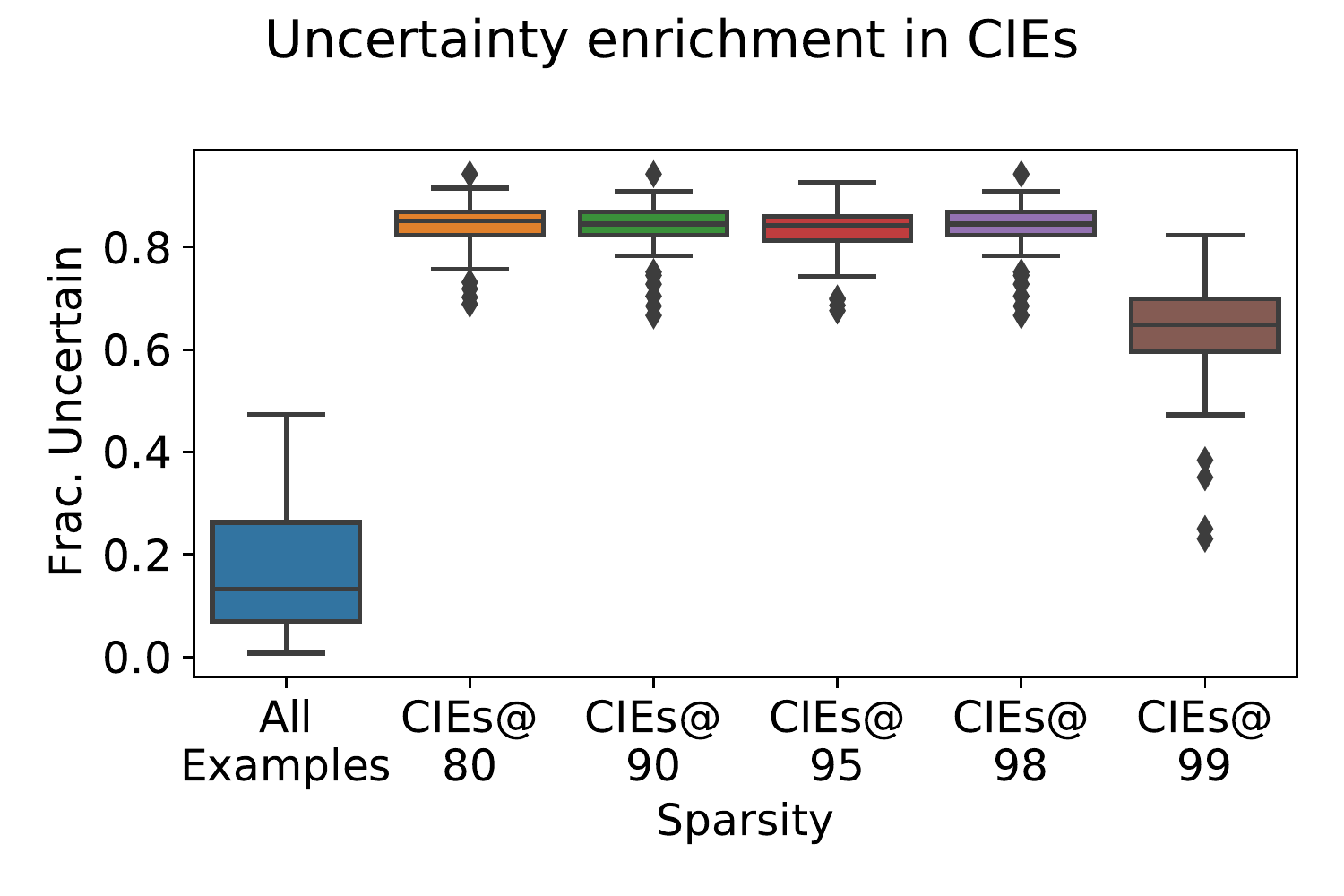}
  \end{tabular}
    \caption{[CelebA / MobileNetV1 / GMP-RI] (Left) Effect of threshold calibration on models jointly trained on all attributes. %
    (Right) Proportion of uncertain predictions for \emph{dense} models across all attributes for all elements in the CelebA test set, and for Compression-Identified Exemplars at different sparsities.}
    \label{fig:celeba_mobilenet_threshold_adj}
\end{figure}

\begin{figure}[h]
\centering
\includegraphics[width=0.8\textwidth]{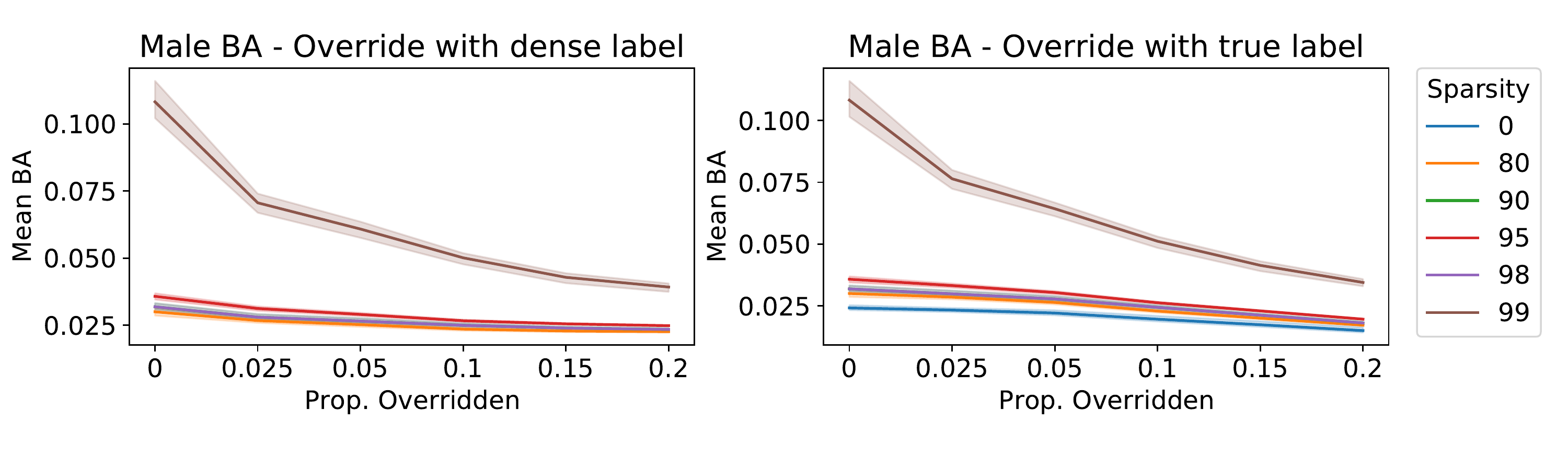}
\includegraphics[width=0.8\textwidth]{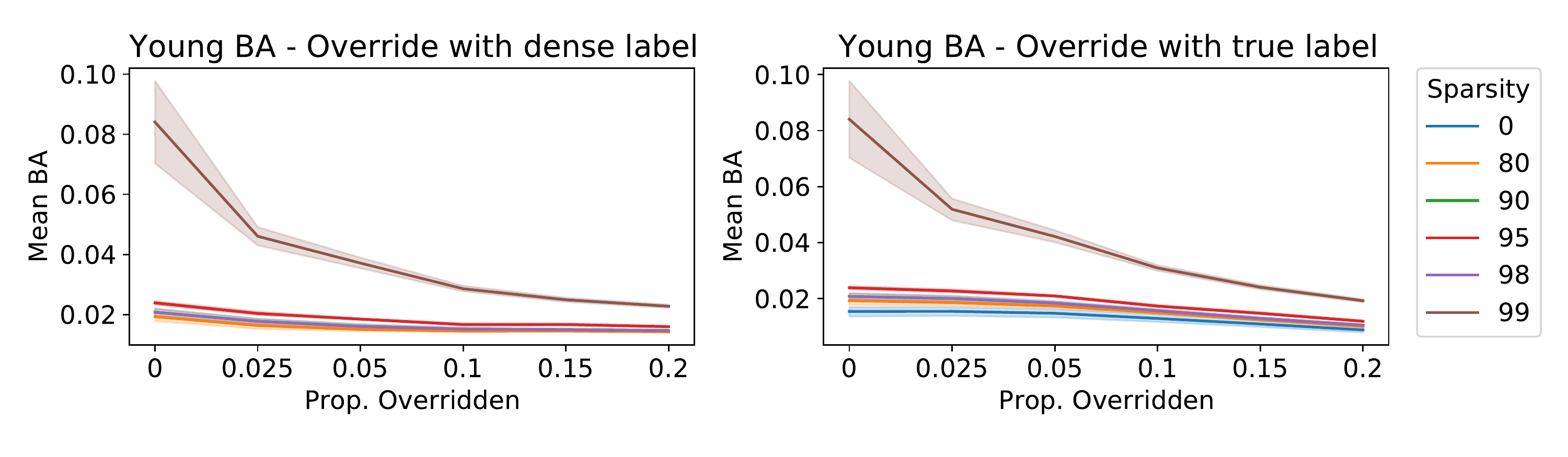}
\includegraphics[width=0.8\textwidth]{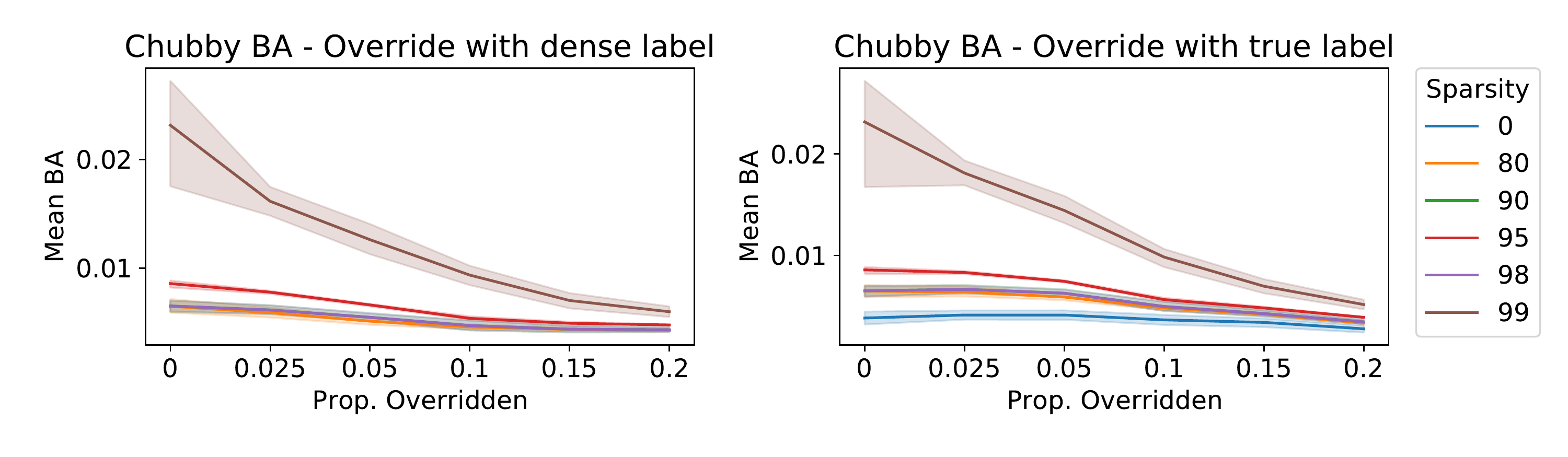}
\includegraphics[width=0.8\textwidth]{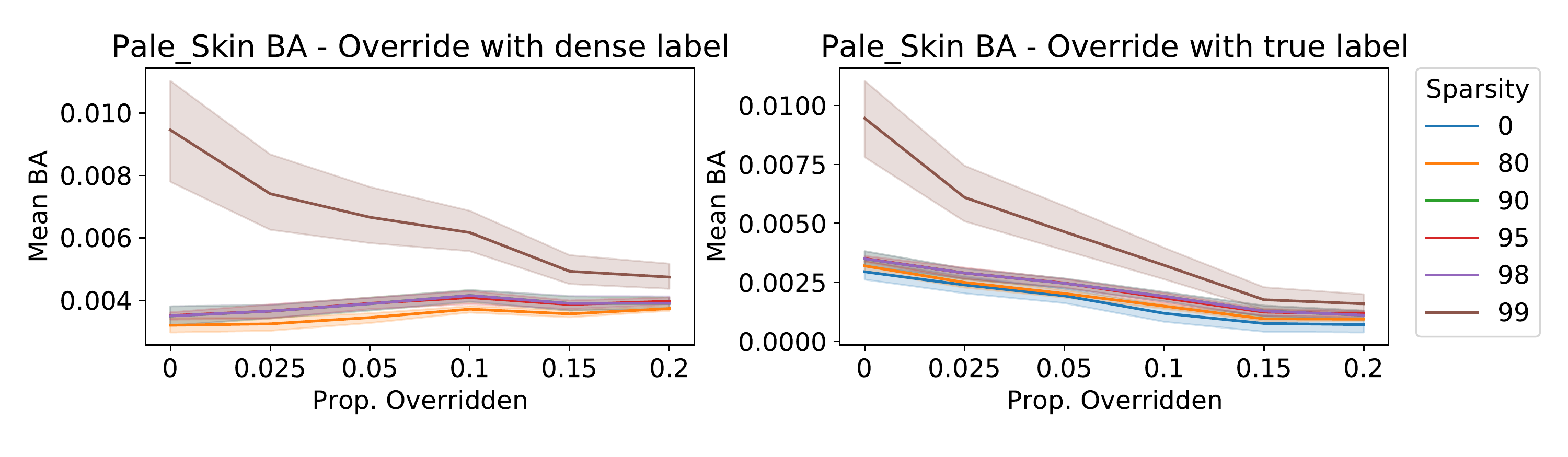}
    \caption{[CelebA / MobileNetV1 / GMP-RI] Effect of label overrides on Bias Amplification. In all cases, overrides are prioritized by dense model uncertainty.}
    \label{fig:overrides_mobilenet}
\end{figure}

For the MobileNet architecture, we note that sparse models maintain a good performance relative to dense, except for 99\% and 99.5\% sparsity, where we observe a decrease in performance, both in terms of accuracy and AUC scores (the 99.5\% models in particular are very poor and are omitted from analysis). The results for systematic and context bias in Figure~\ref{fig:celeba_mobilenet_joint_systematic} show similar trends to those observed for ResNet18; we note that all our bias metrics, including uncertainty, are substantially amplified at 99\% sparsity, which is not surprising given the lower performance of the model. Moreover, we show in Figure~\ref{fig:overrides_mobilenet} that it is possible to decrease the bias in 99\% sparse models by over-ridding the labels of the low confidence samples with their true or dense labels, and we also show that most of CIEs are uncertain samples in Figure~\ref{fig:celeba_mobilenet_threshold_adj}.

\begin{figure}[ht]
\centering
\begin{tabular}{cccccccc}
\includegraphics[width=0.12\textwidth]{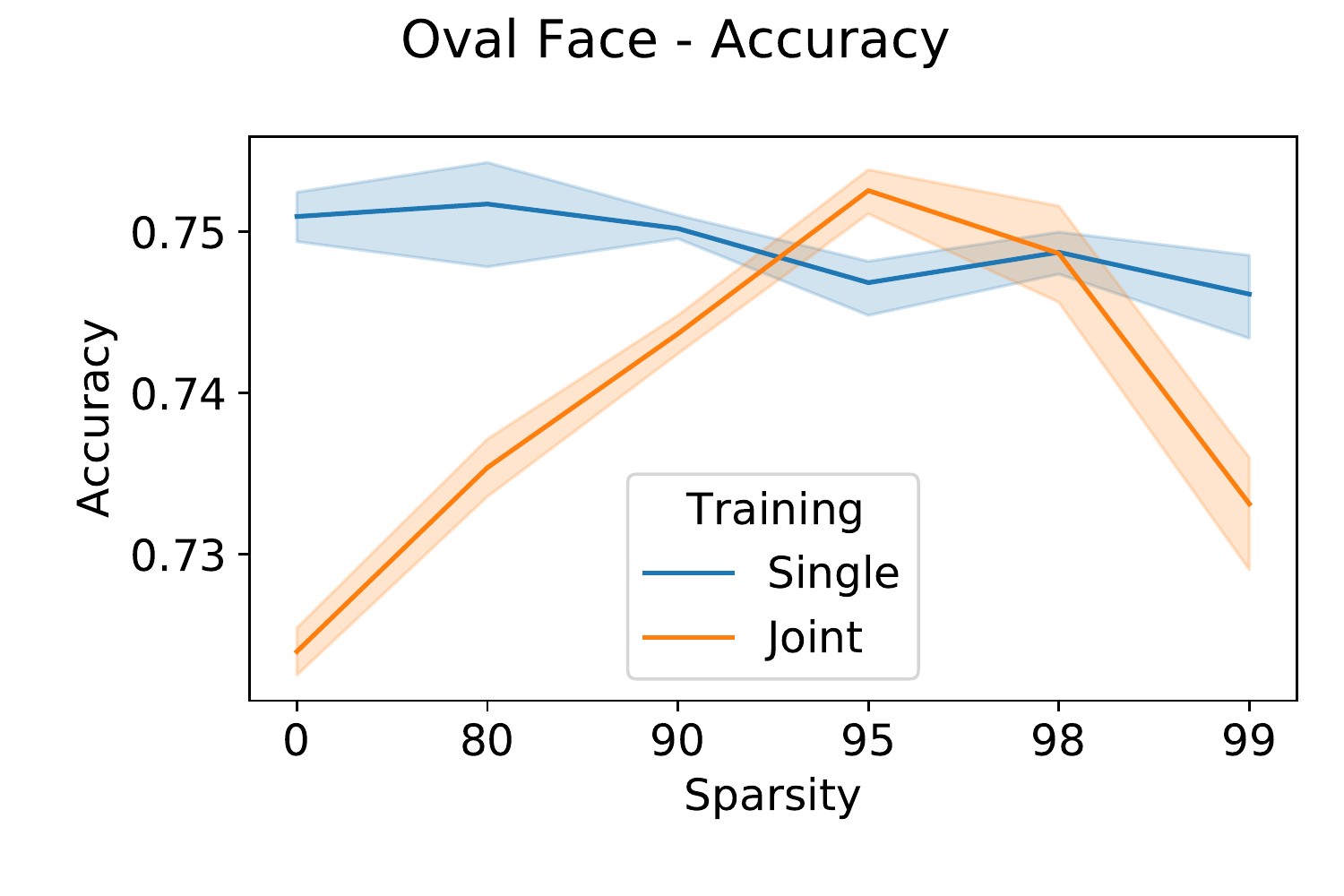} &
\includegraphics[width=0.12\textwidth]{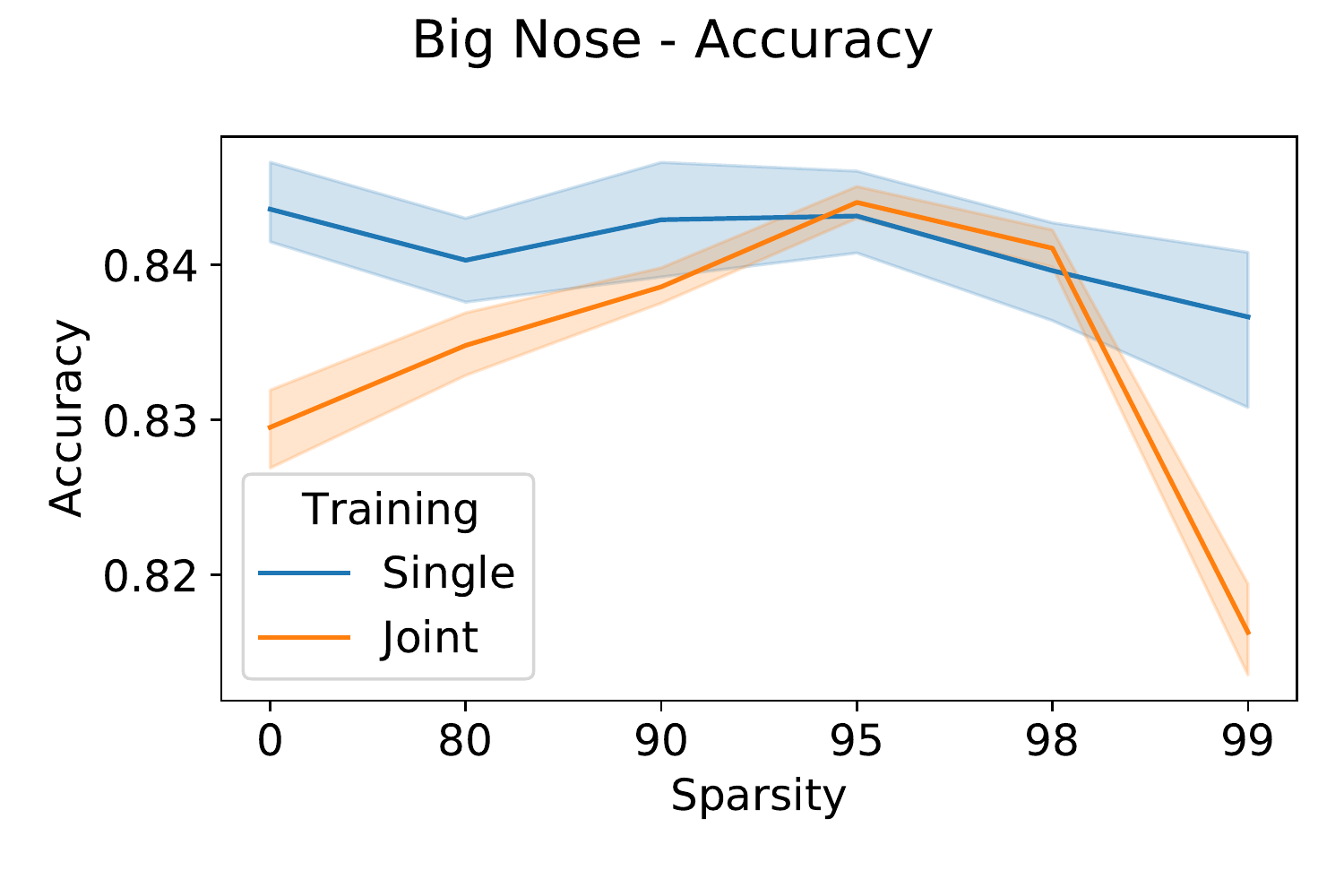} &
\includegraphics[width=0.12\textwidth]{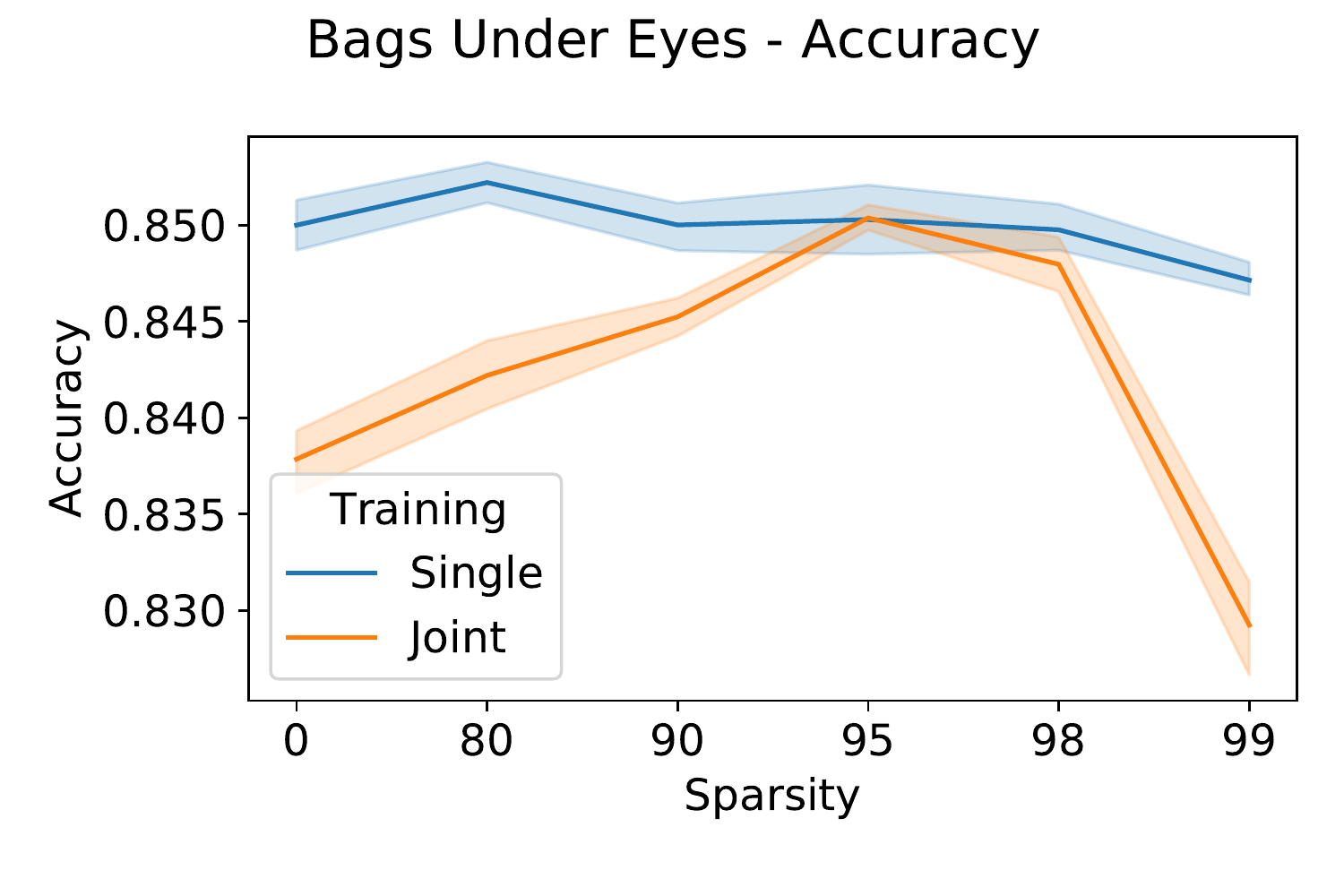} &
\includegraphics[width=0.12\textwidth]{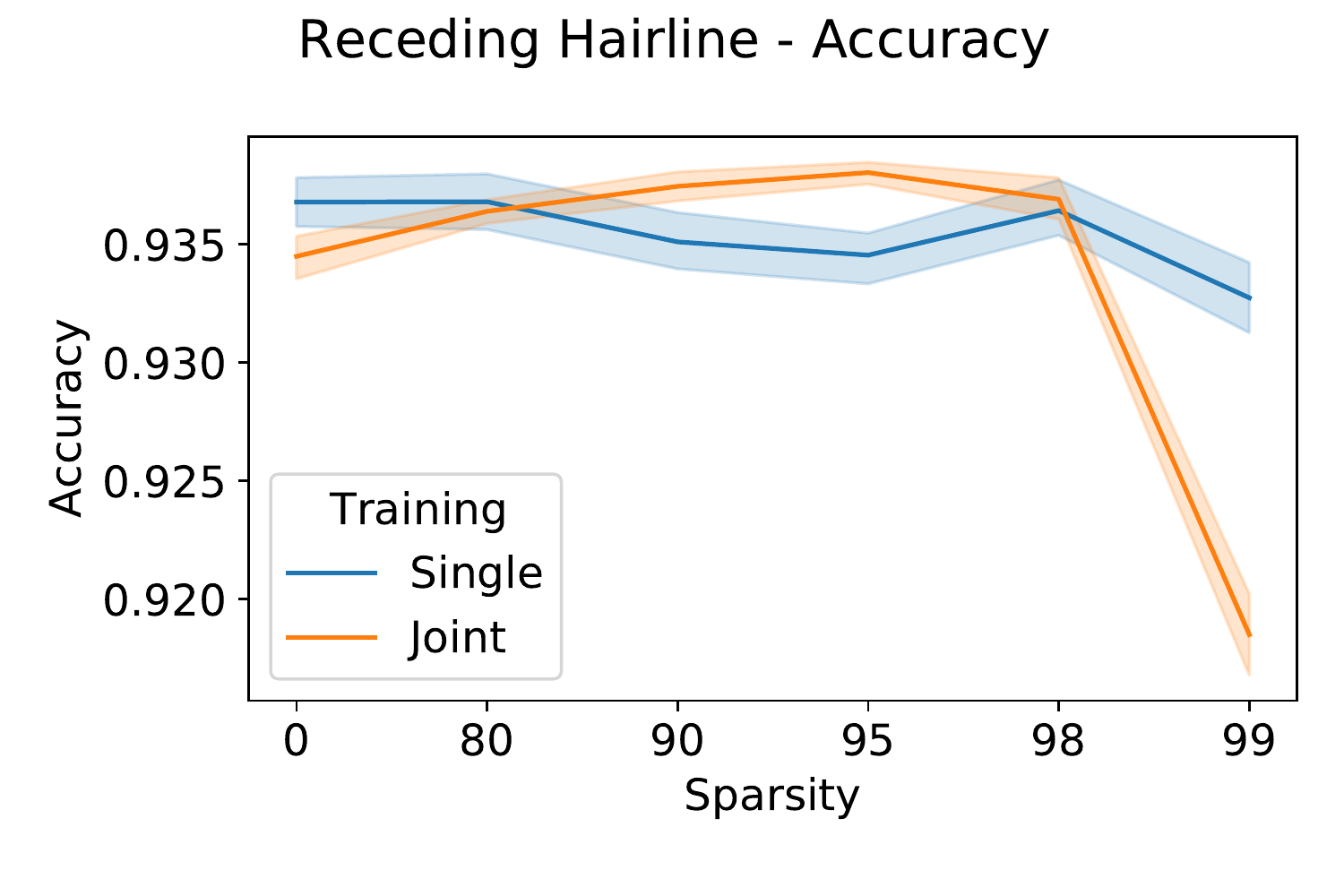} &
\includegraphics[width=0.12\textwidth]{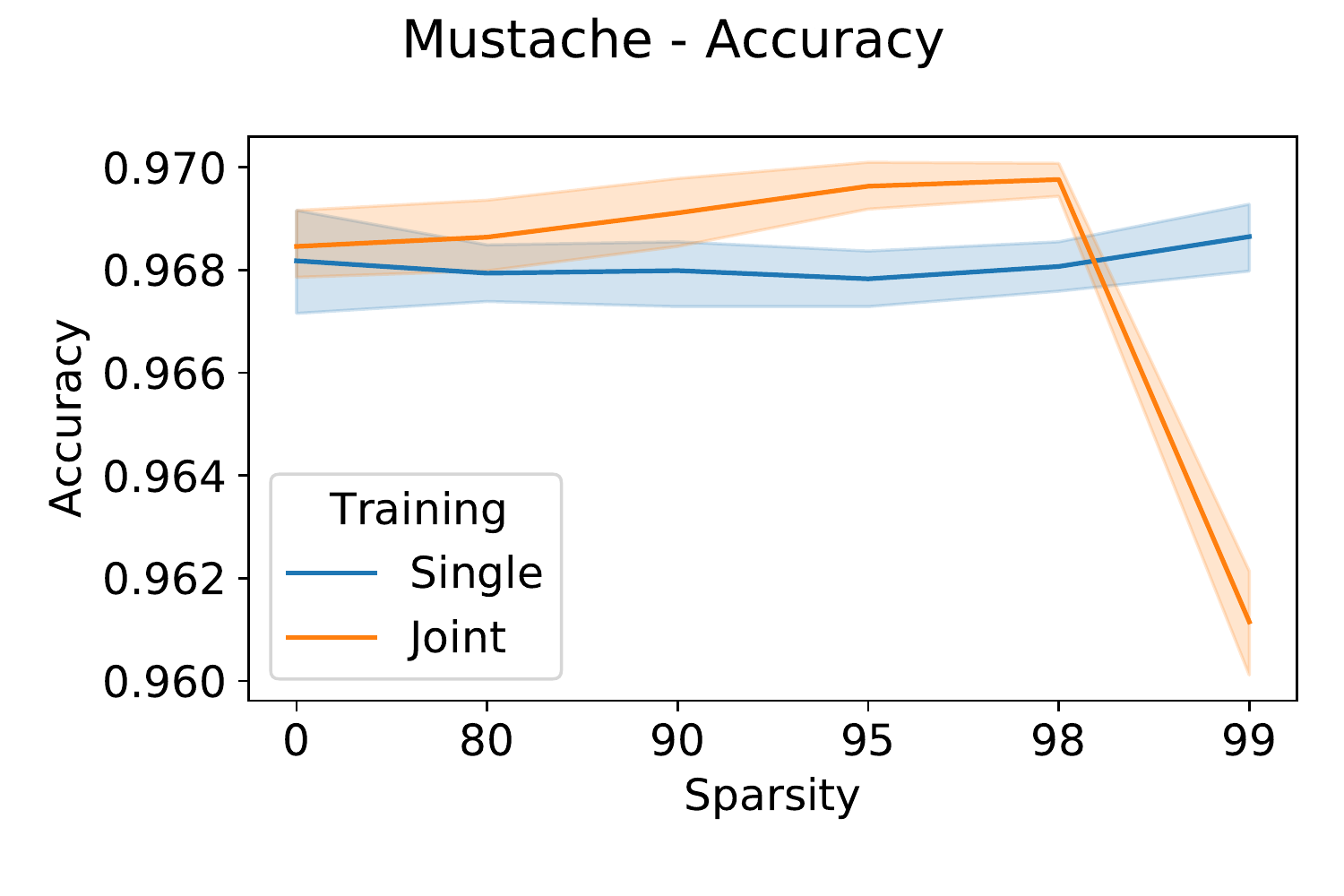} &
\includegraphics[width=0.12\textwidth]{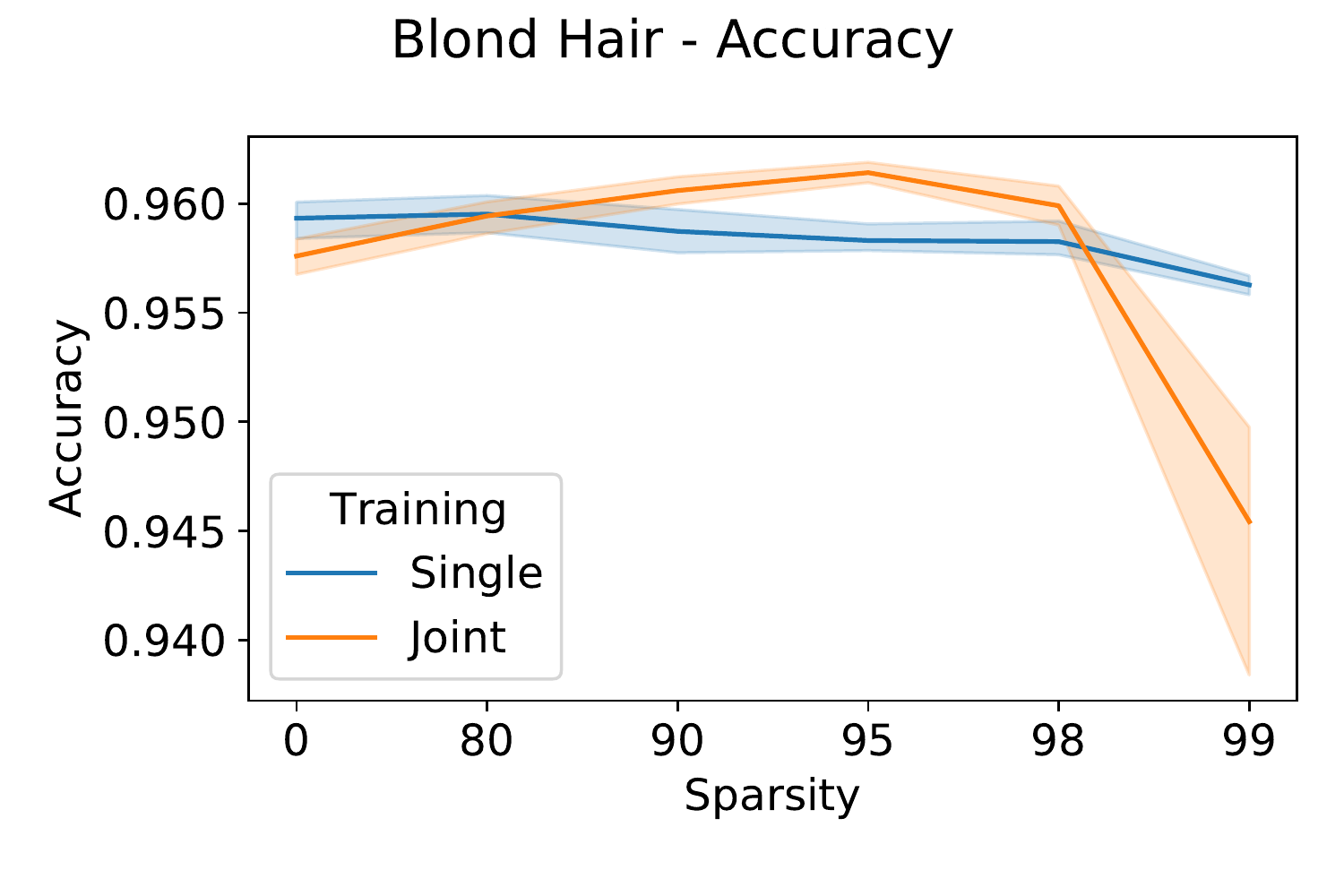} &
\includegraphics[width=0.12\textwidth]{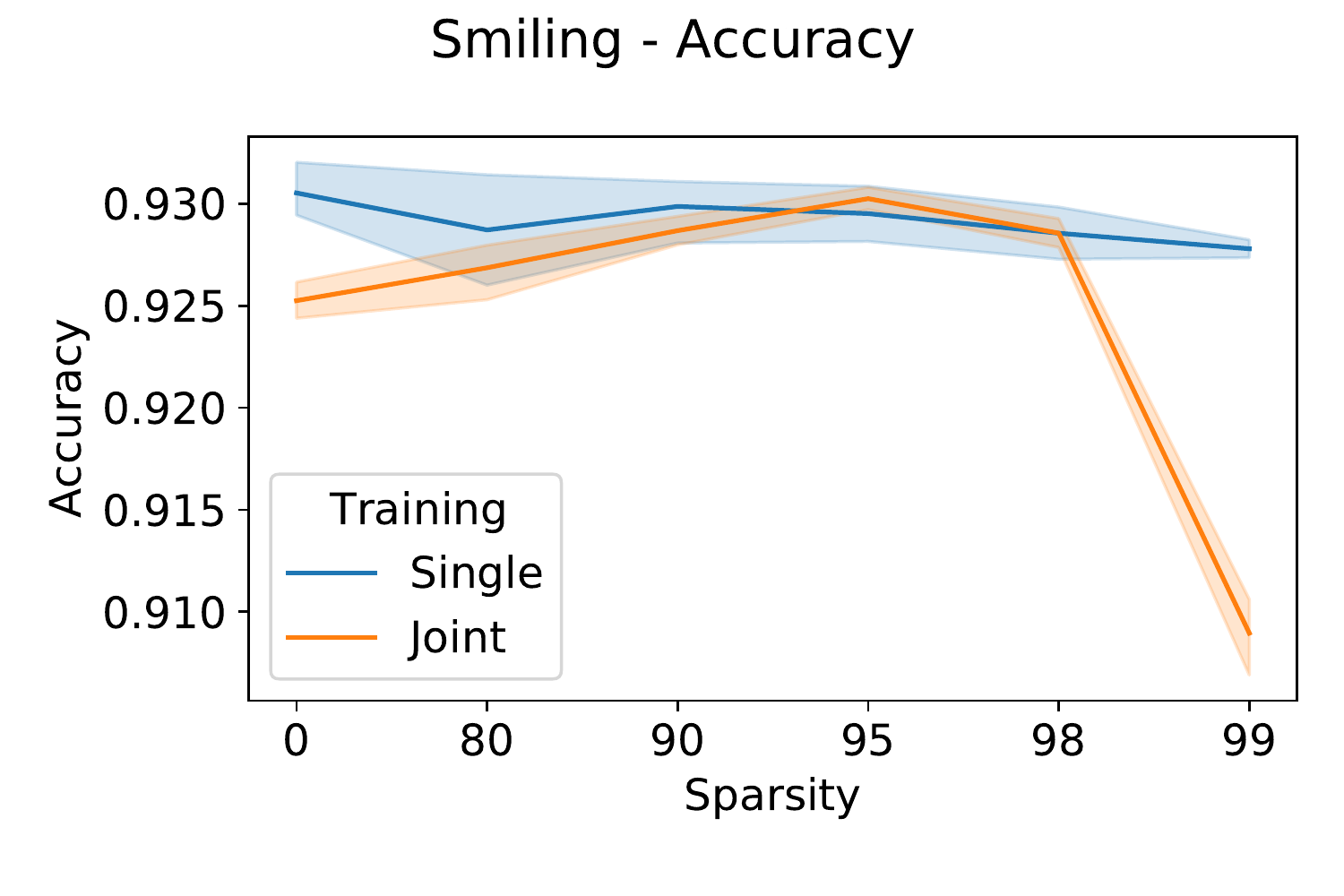}
\\
  \includegraphics[width=0.12\textwidth]{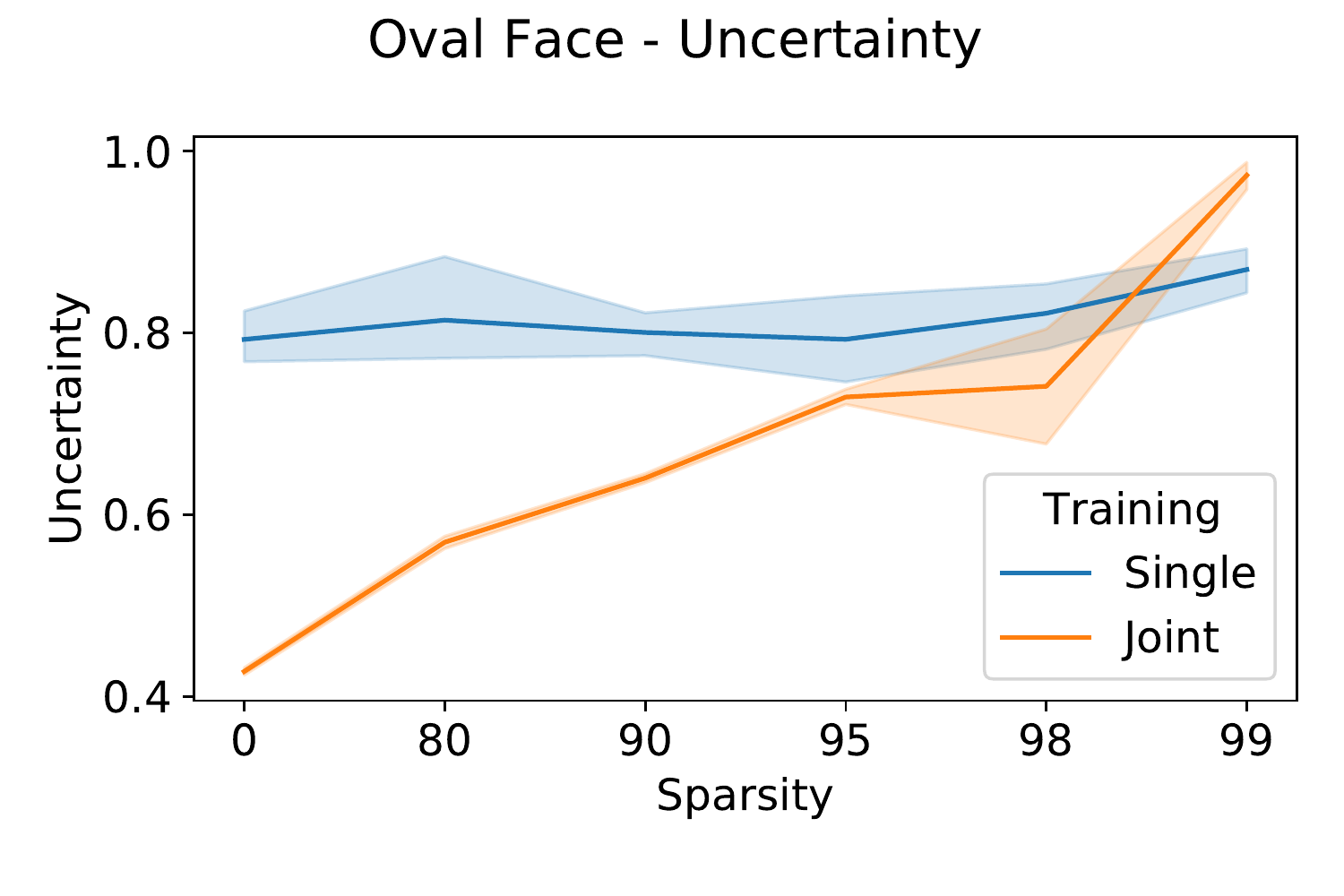} &
      \includegraphics[width=0.12\textwidth]{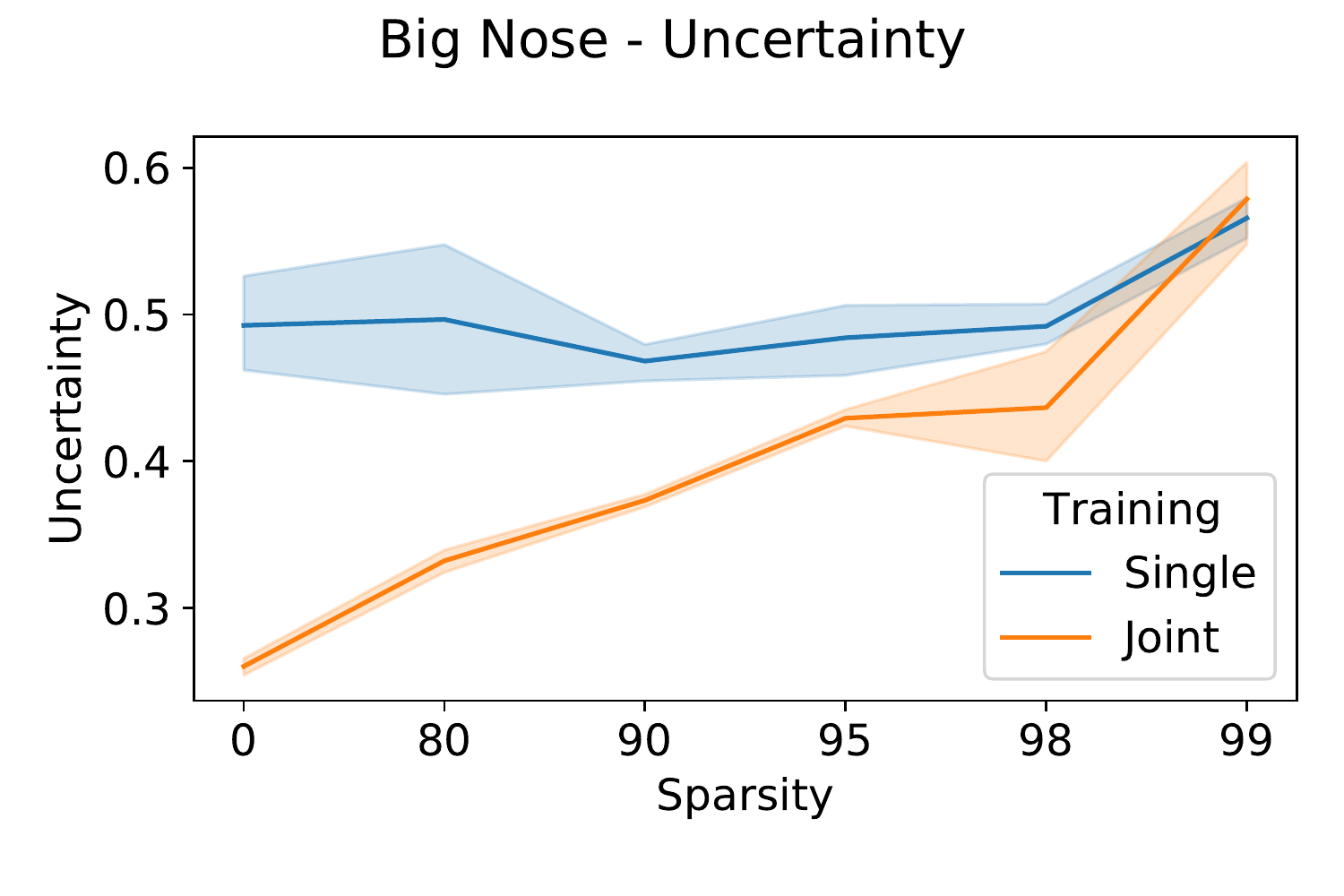} &
    \includegraphics[width=0.12\textwidth]{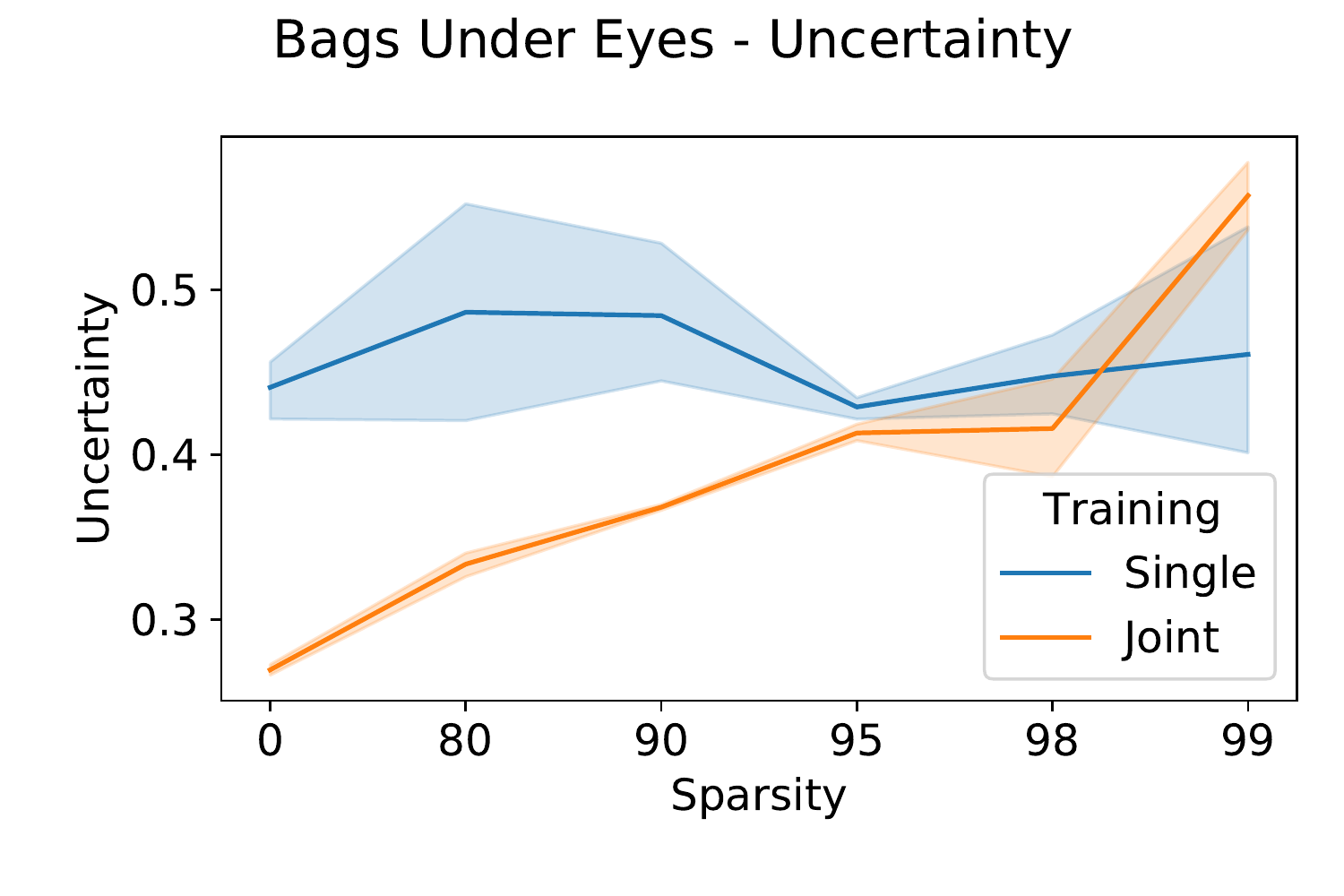} &
      \includegraphics[width=0.12\textwidth]{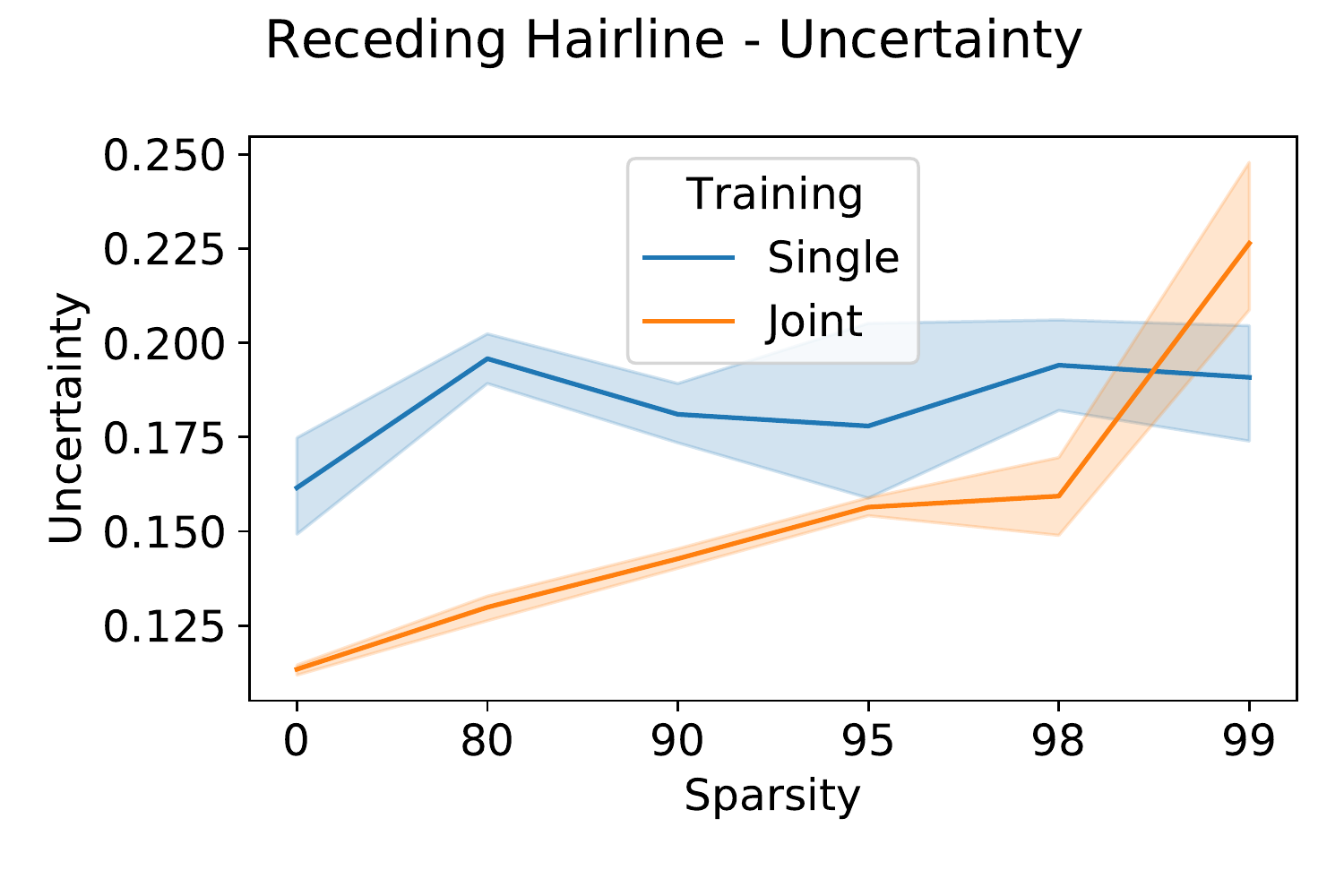} &
      \includegraphics[width=0.12\textwidth]{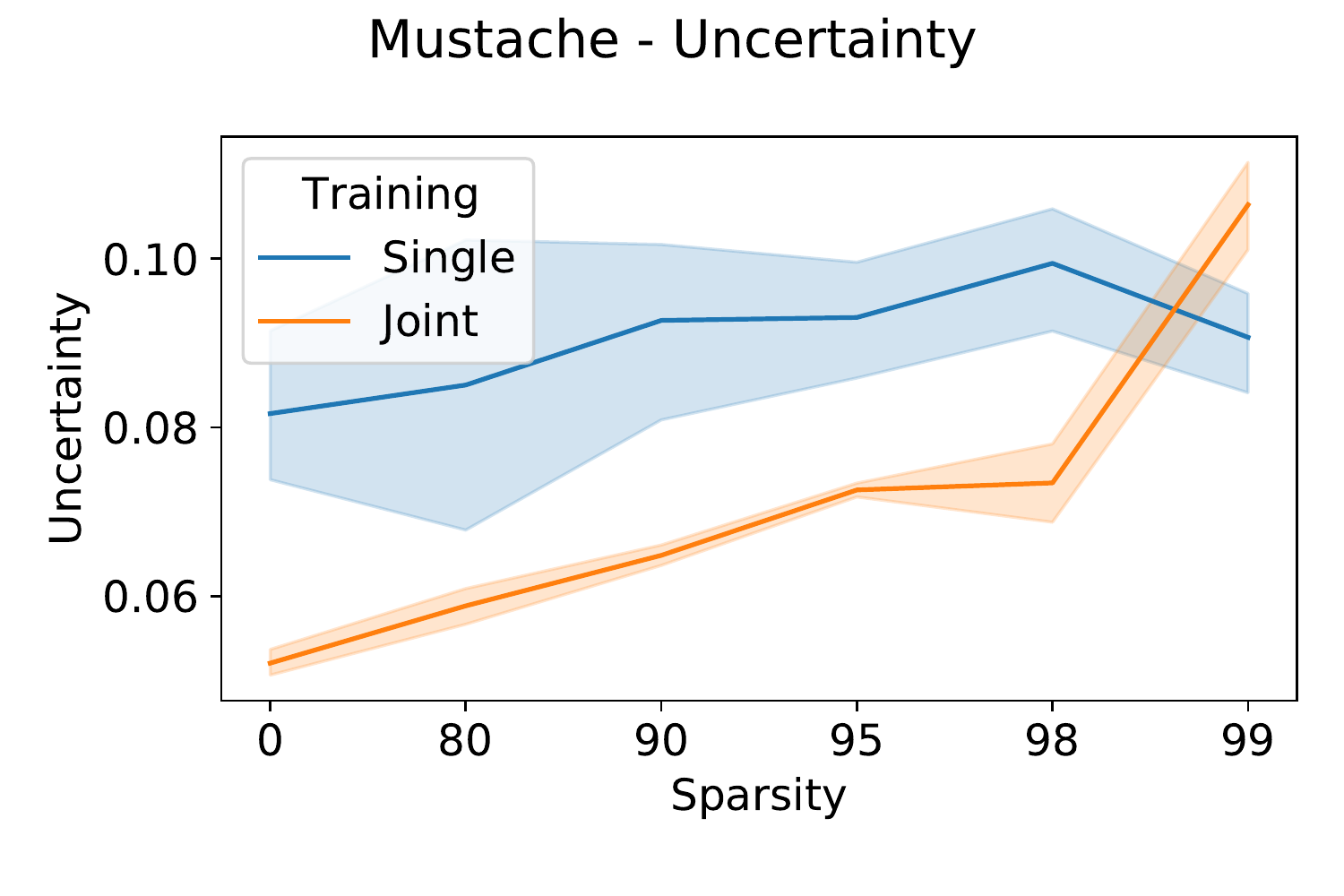} &
      \includegraphics[width=0.12\textwidth]{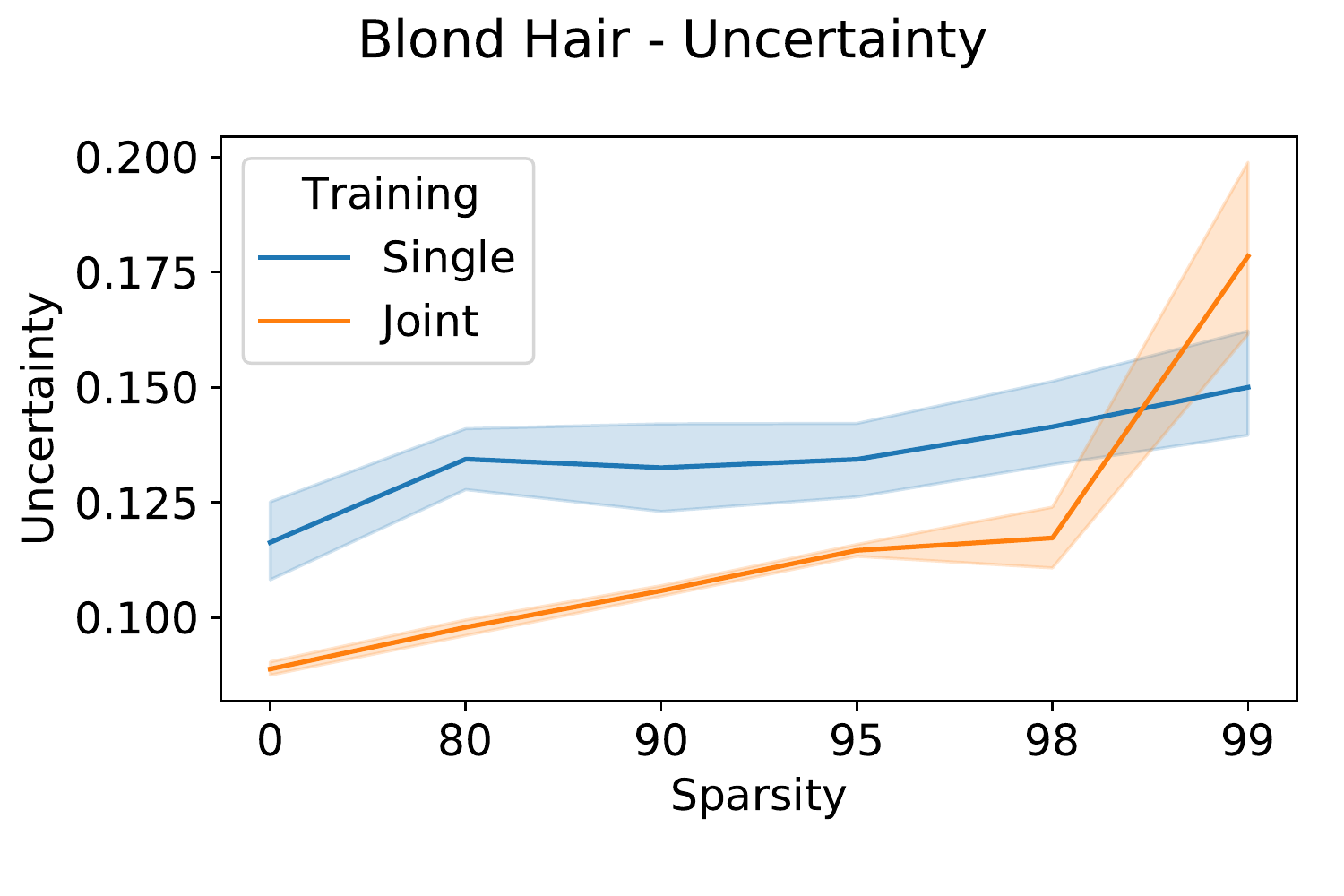} &
      \includegraphics[width=0.12\textwidth]{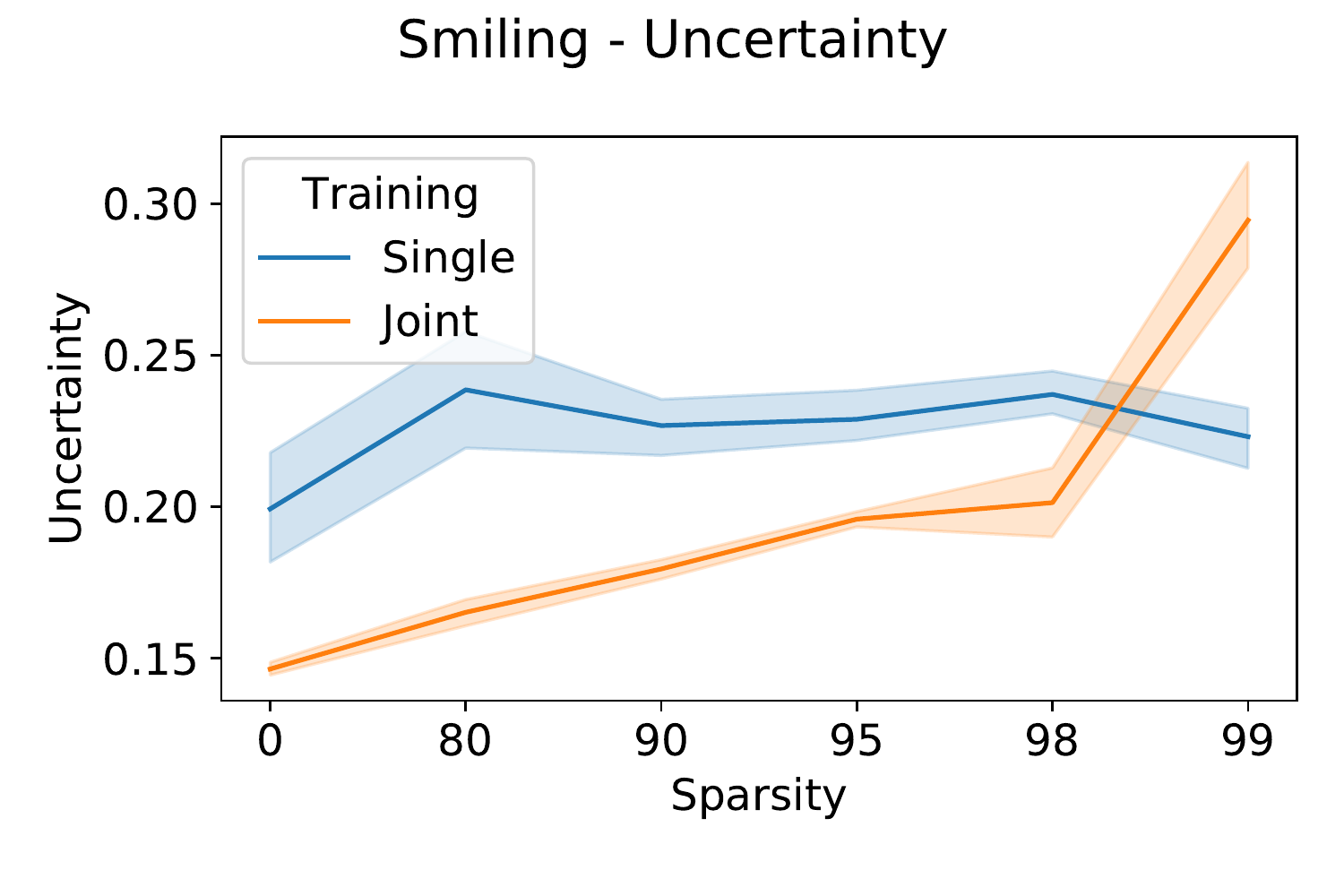}
      \\
  \includegraphics[width=0.12\textwidth]{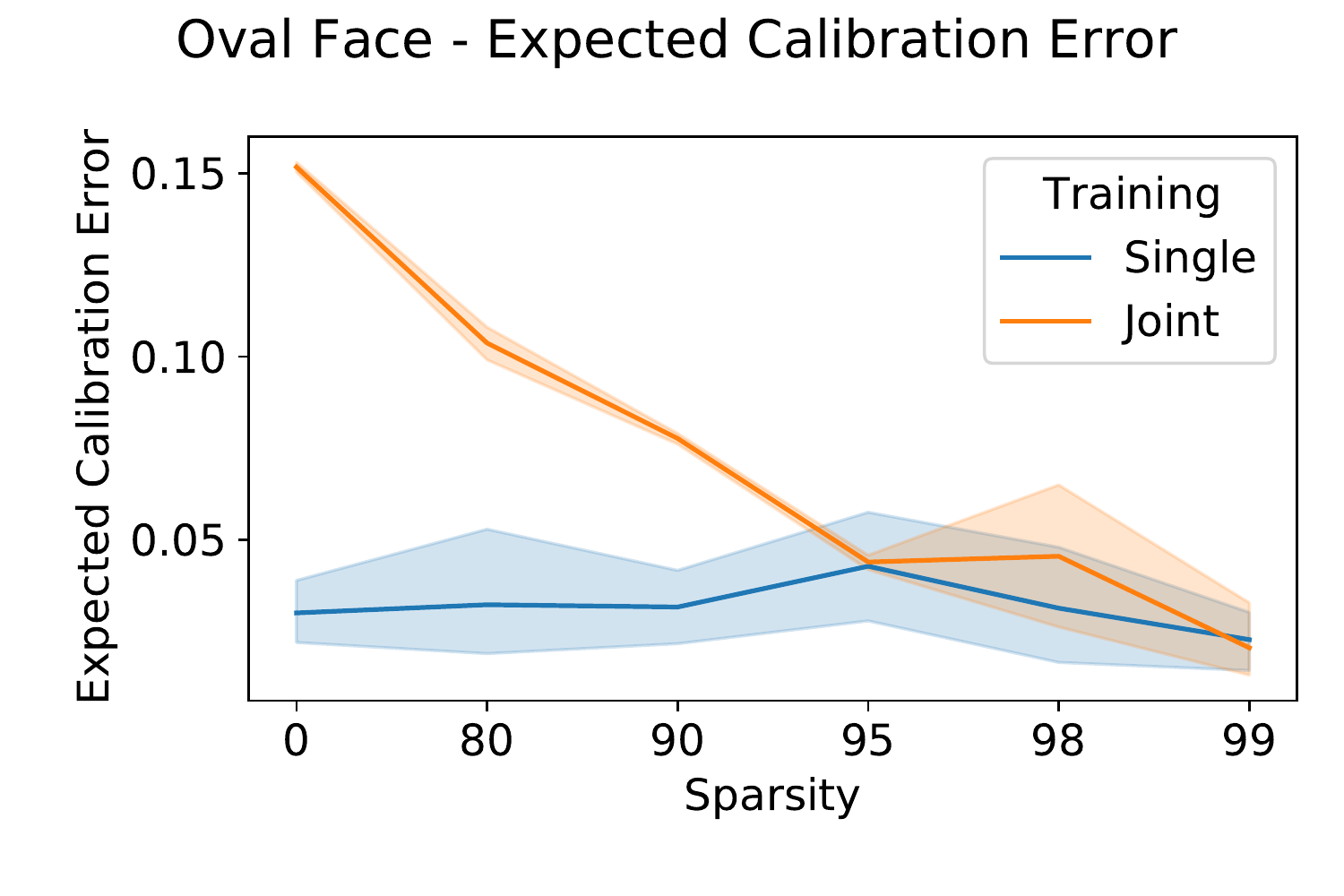} &
      \includegraphics[width=0.12\textwidth]{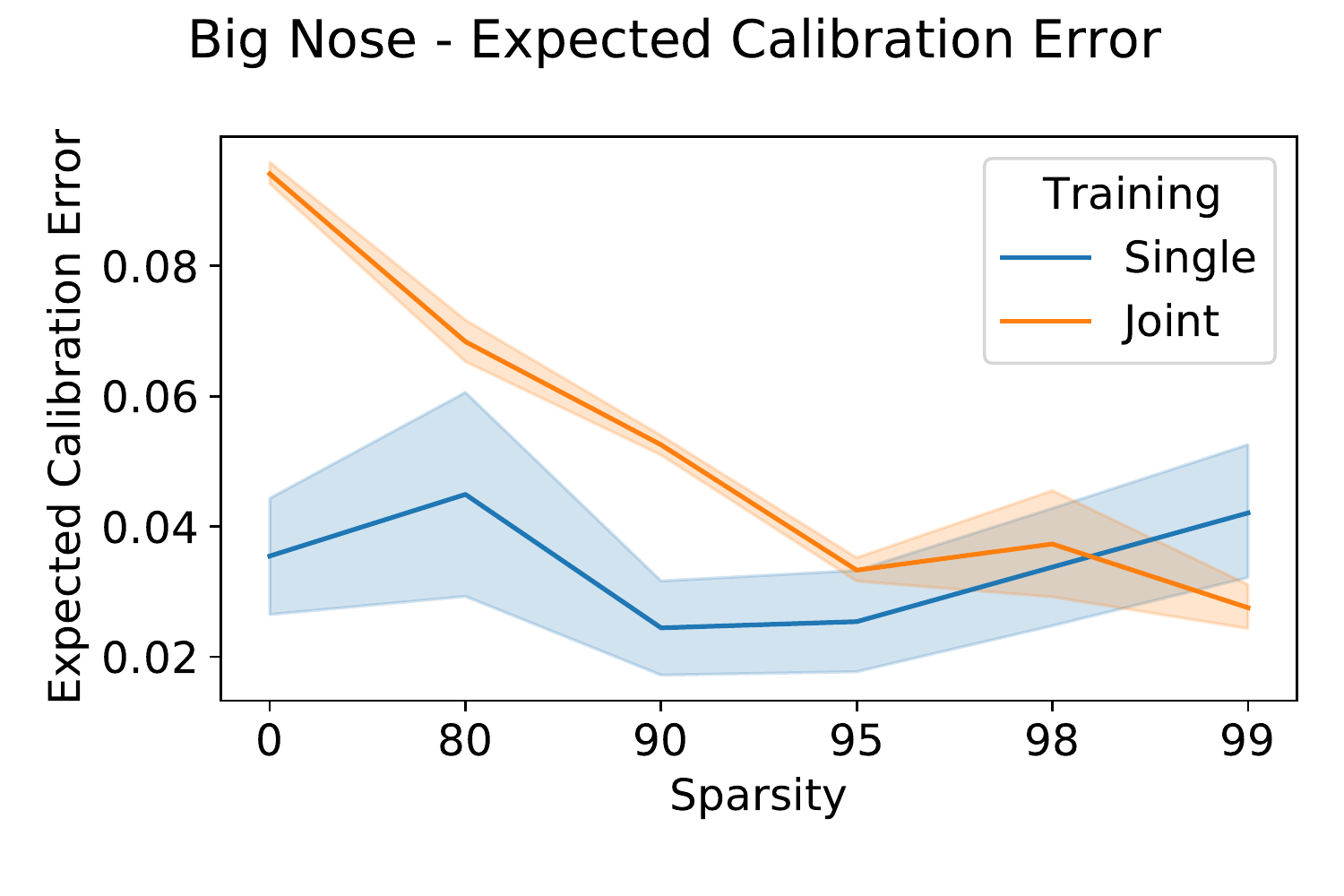} &
    \includegraphics[width=0.12\textwidth]{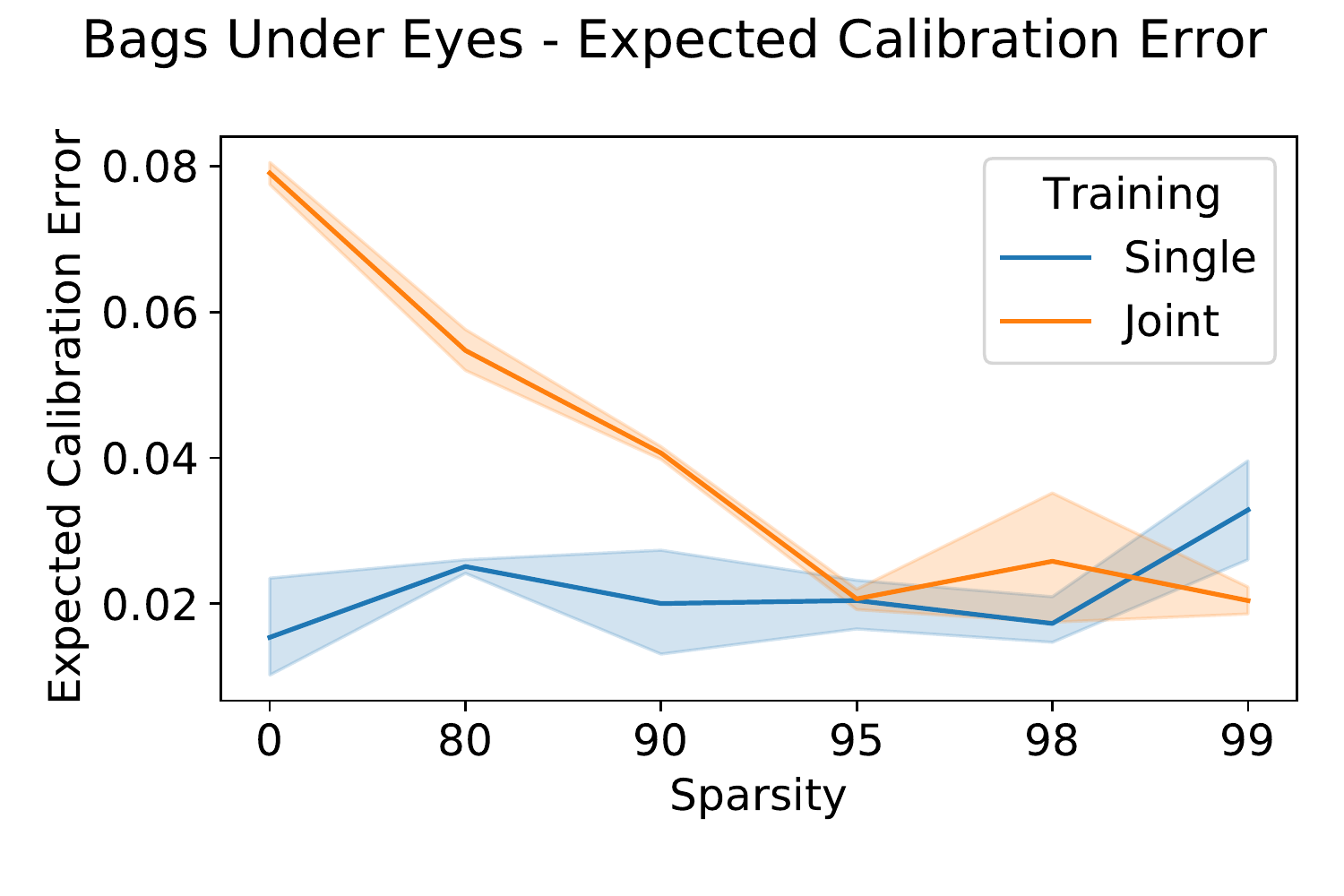} &
      \includegraphics[width=0.12\textwidth]{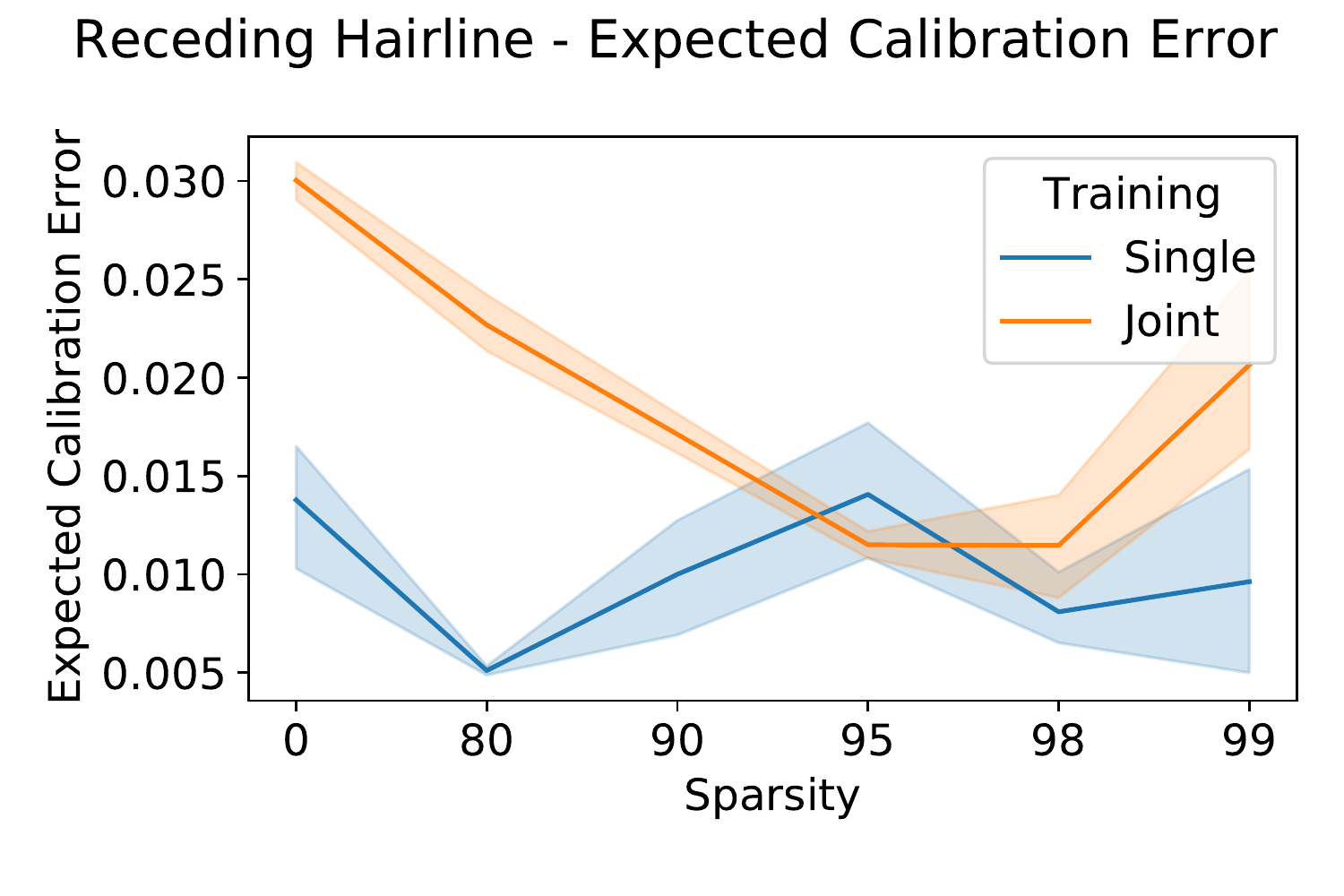} &
      \includegraphics[width=0.12\textwidth]{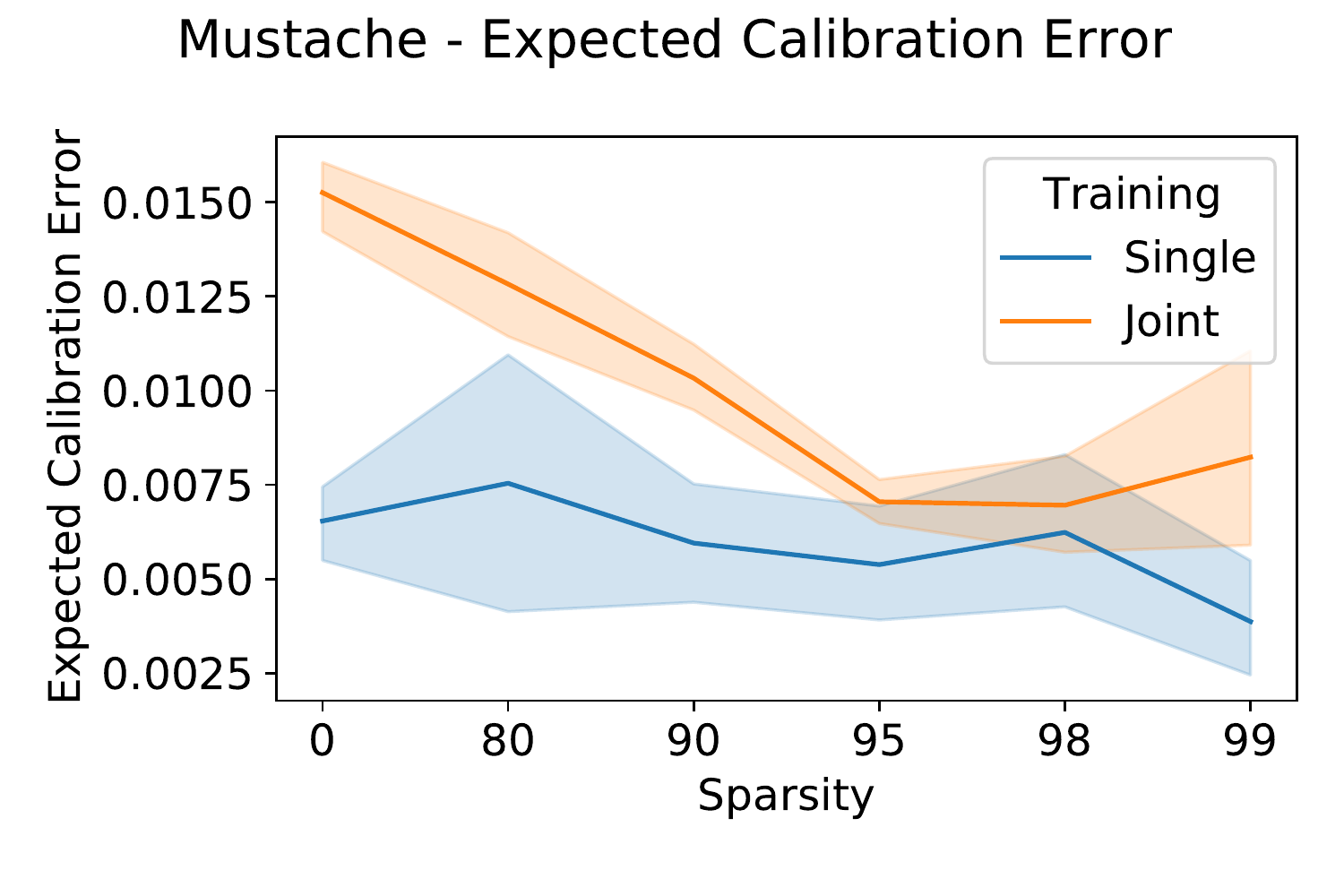} &
      \includegraphics[width=0.12\textwidth]{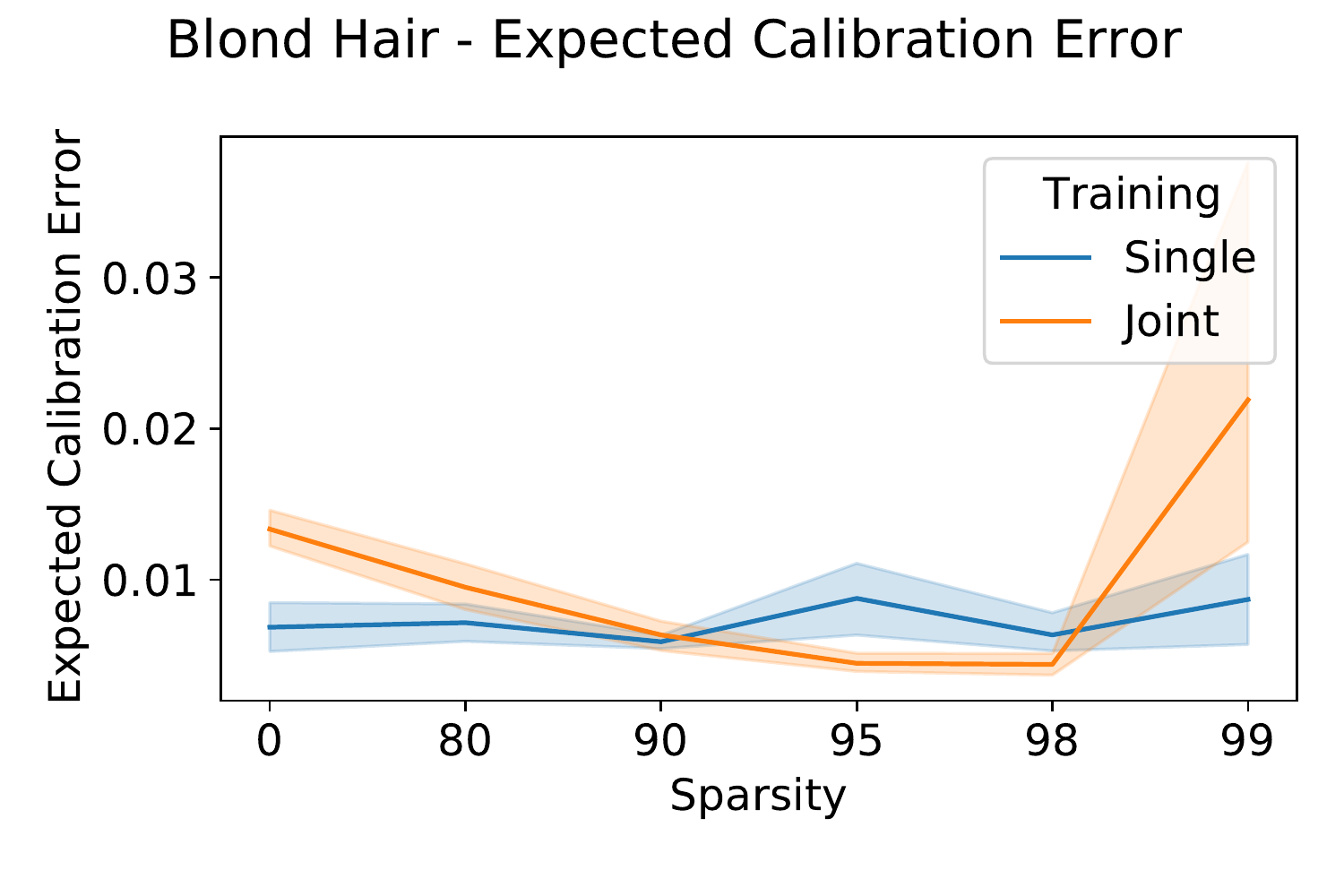} &
      \includegraphics[width=0.12\textwidth]{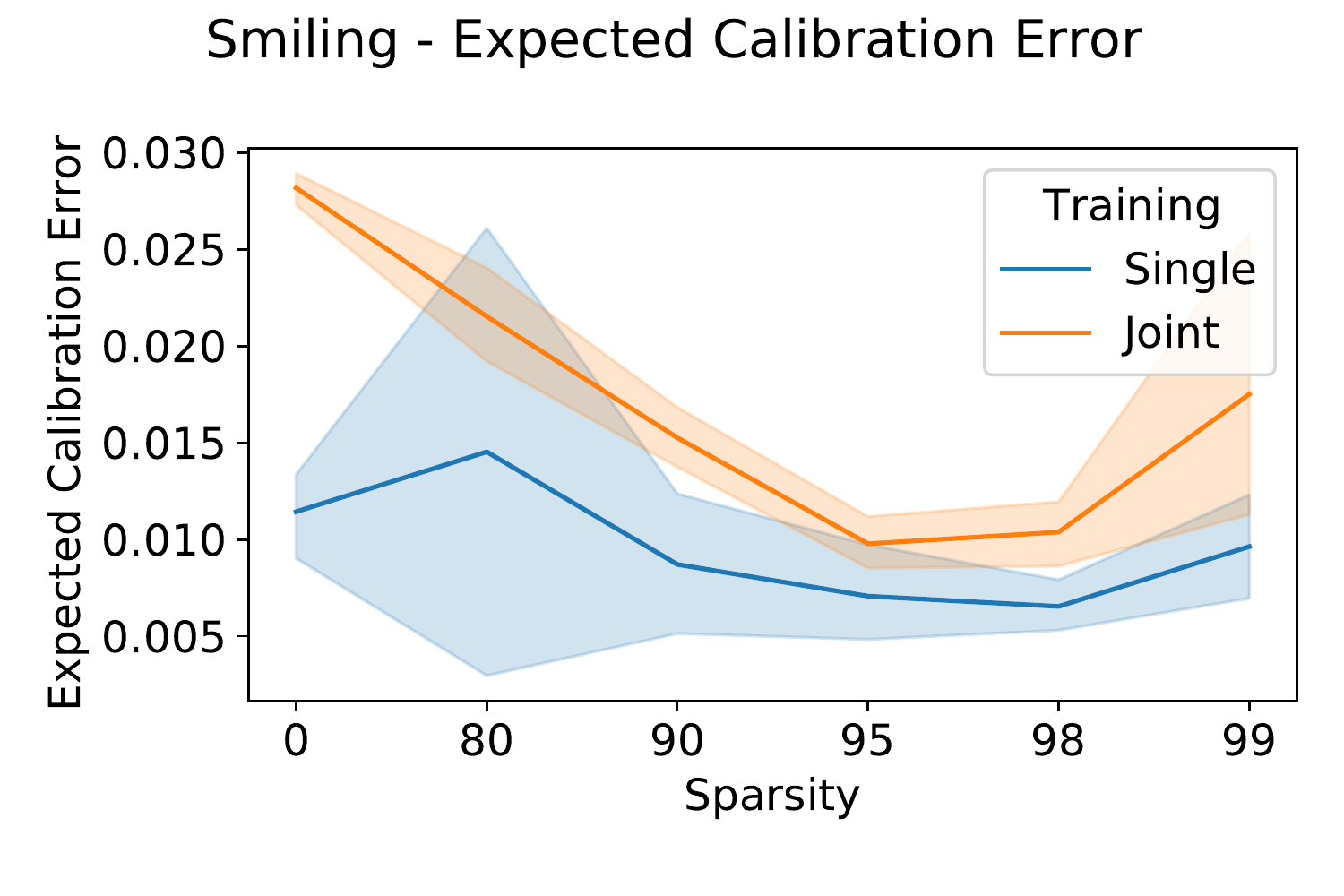}
      \\
    \includegraphics[width=0.12\textwidth]
  {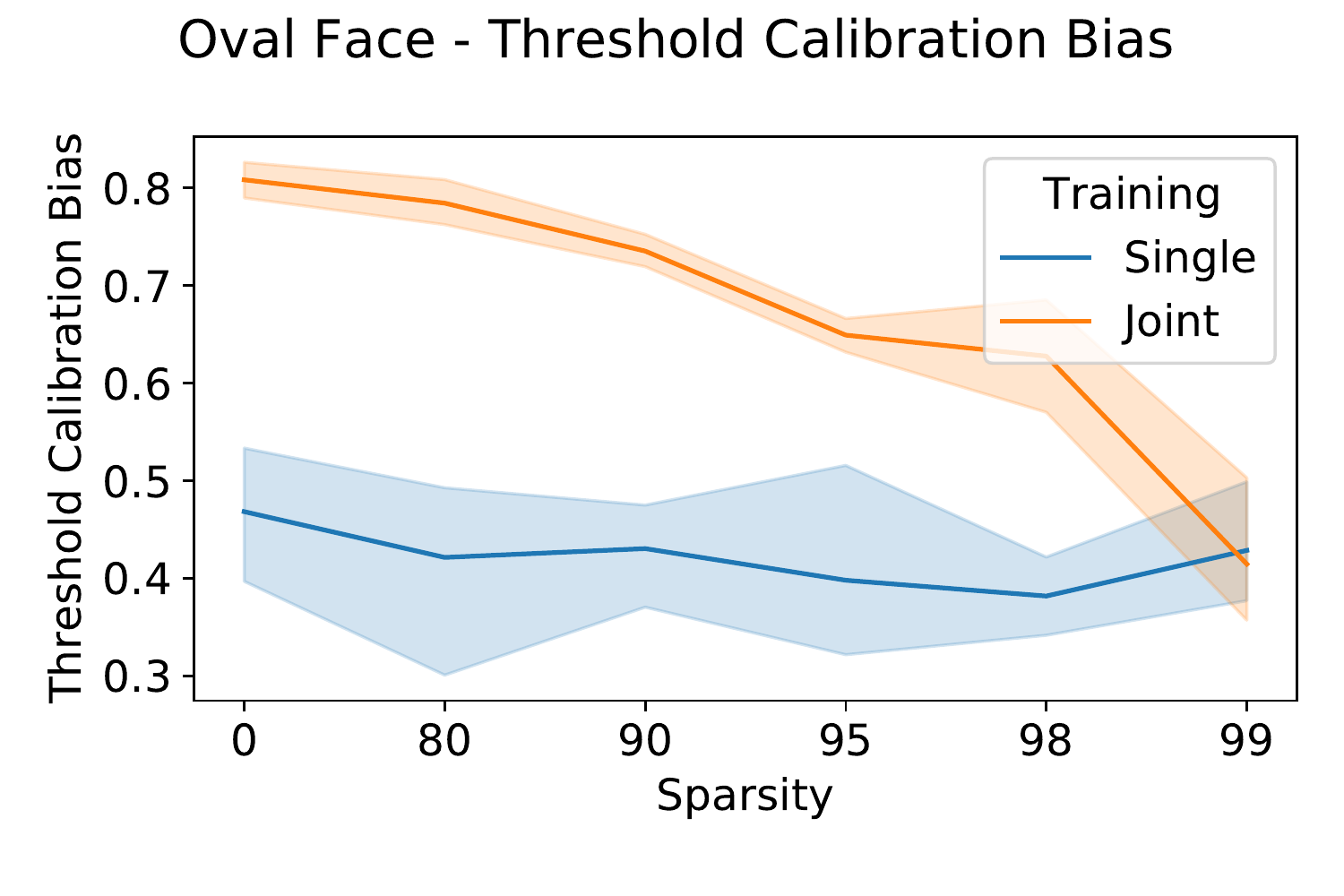} &
  \includegraphics[width=0.12\textwidth]
  {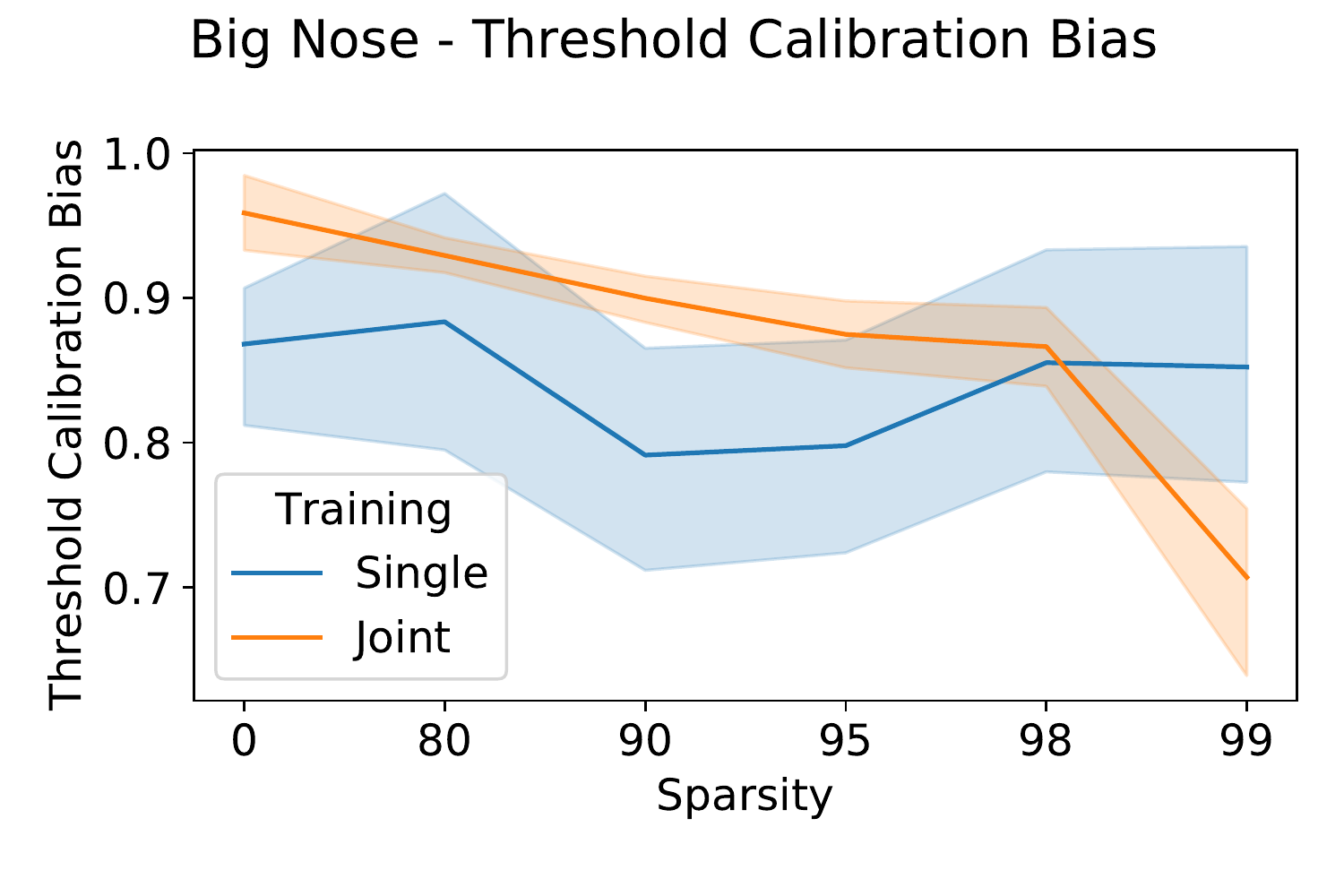} &
  \includegraphics[width=0.12\textwidth]
  {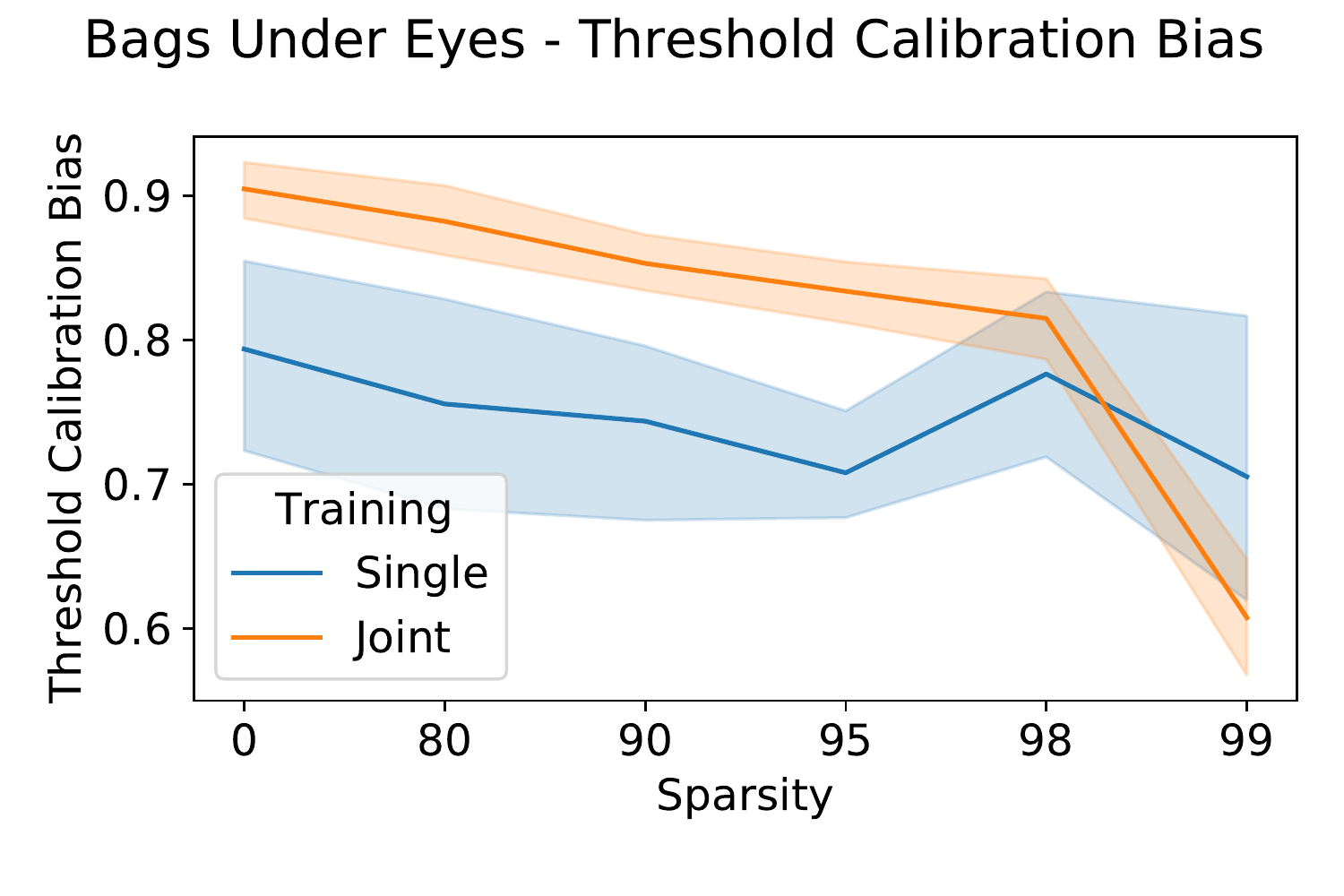} &
  \includegraphics[width=0.12\textwidth]
  {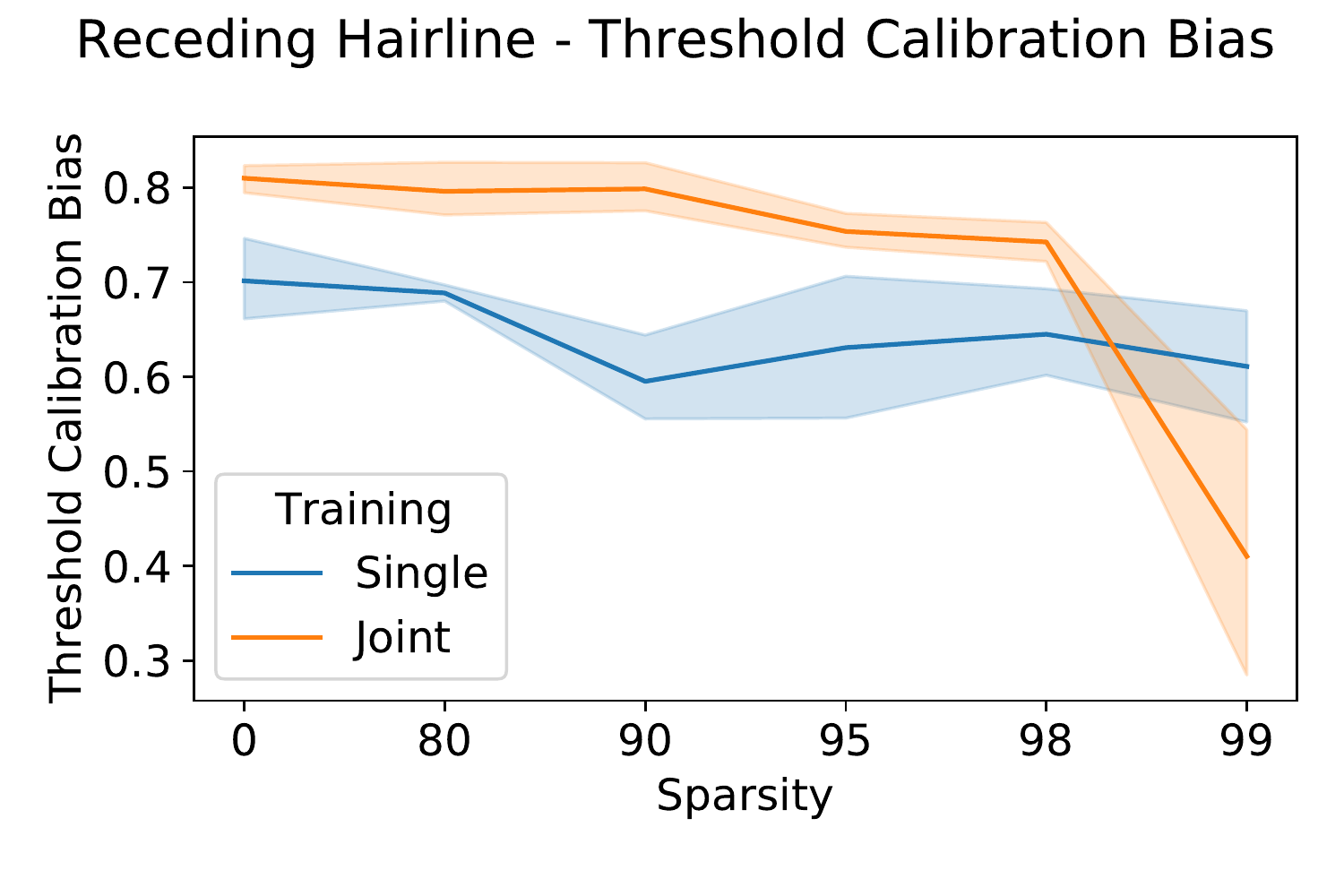} &
  \includegraphics[width=0.12\textwidth]
  {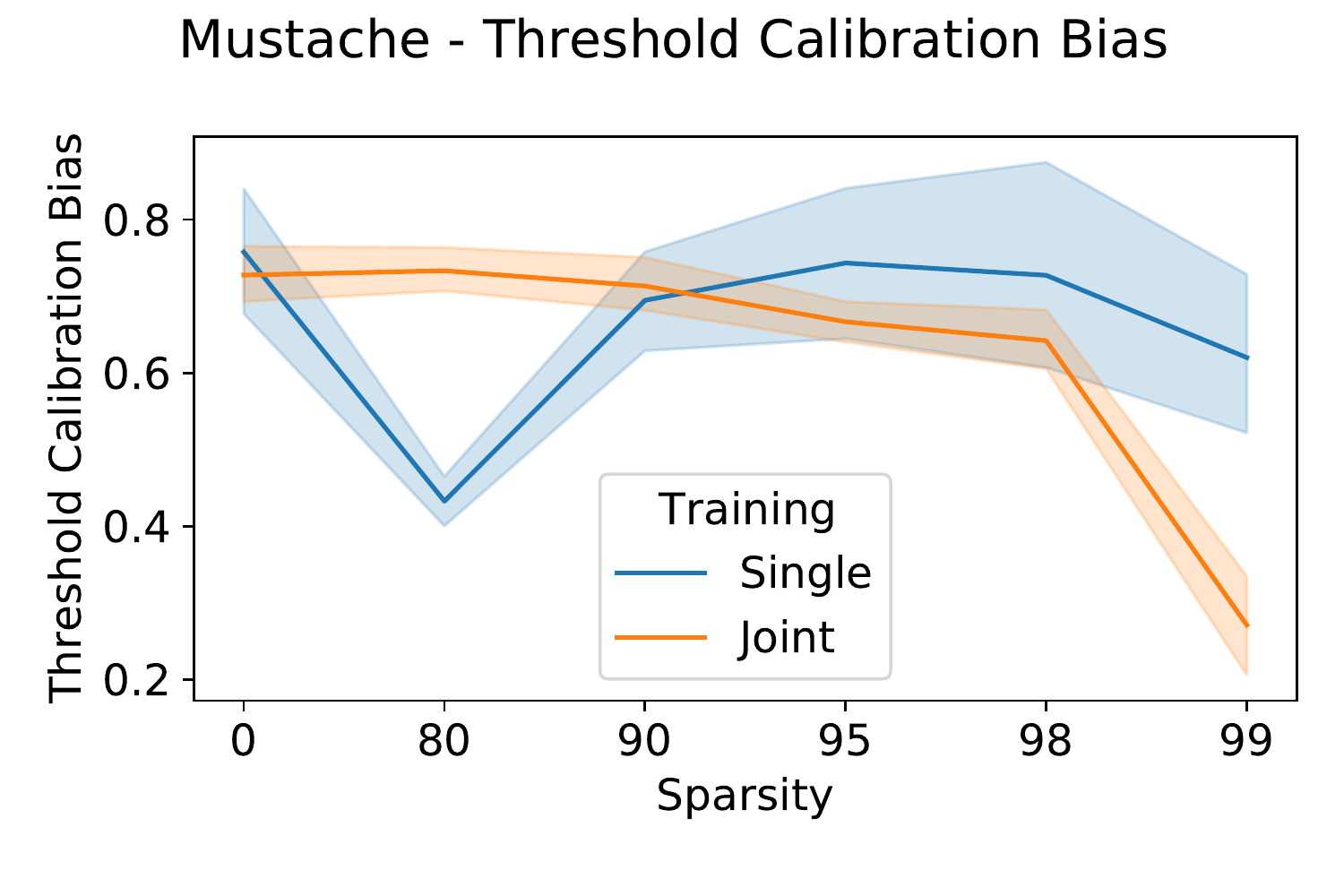} &
  \includegraphics[width=0.12\textwidth]
  {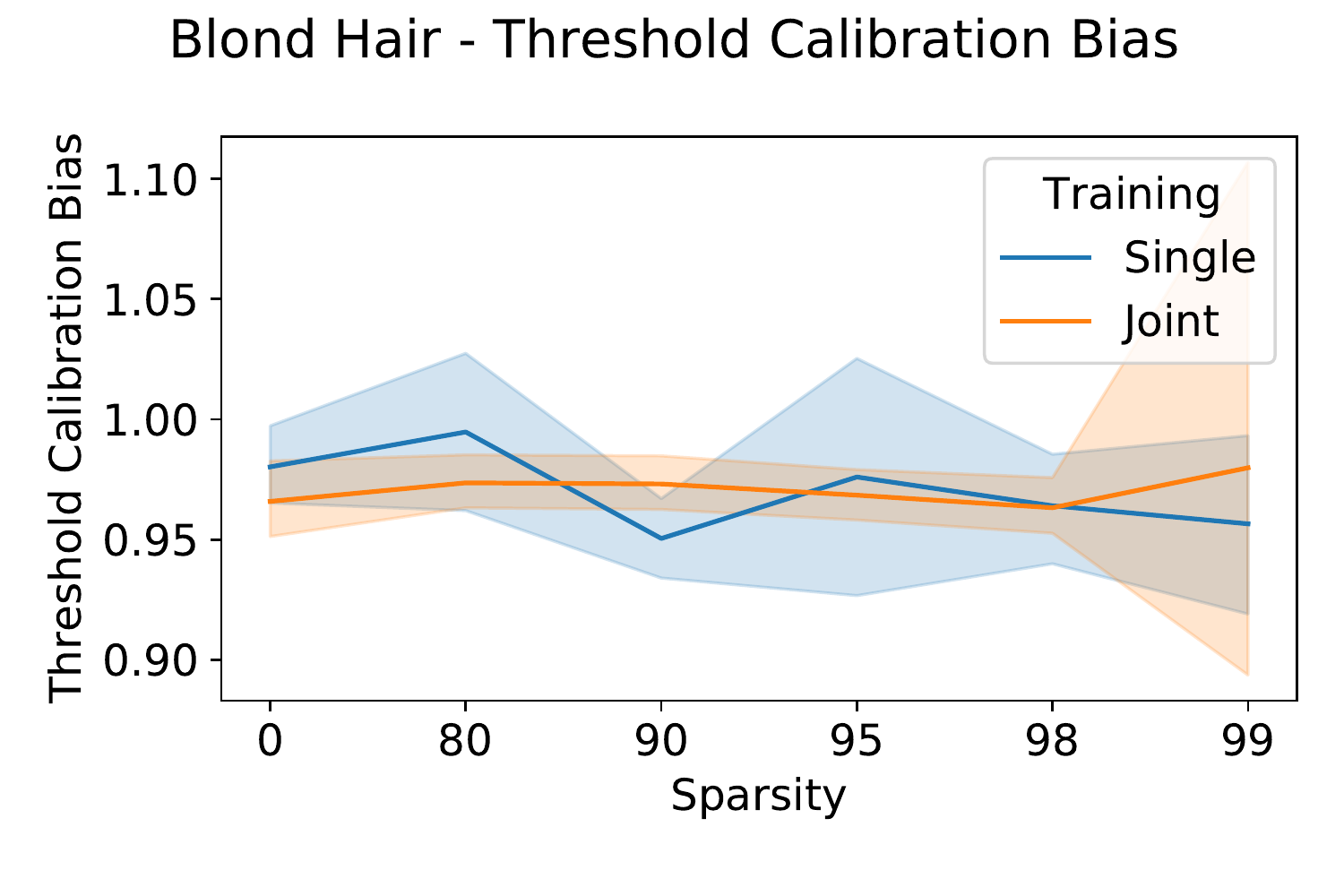} &
  \includegraphics[width=0.12\textwidth]
  {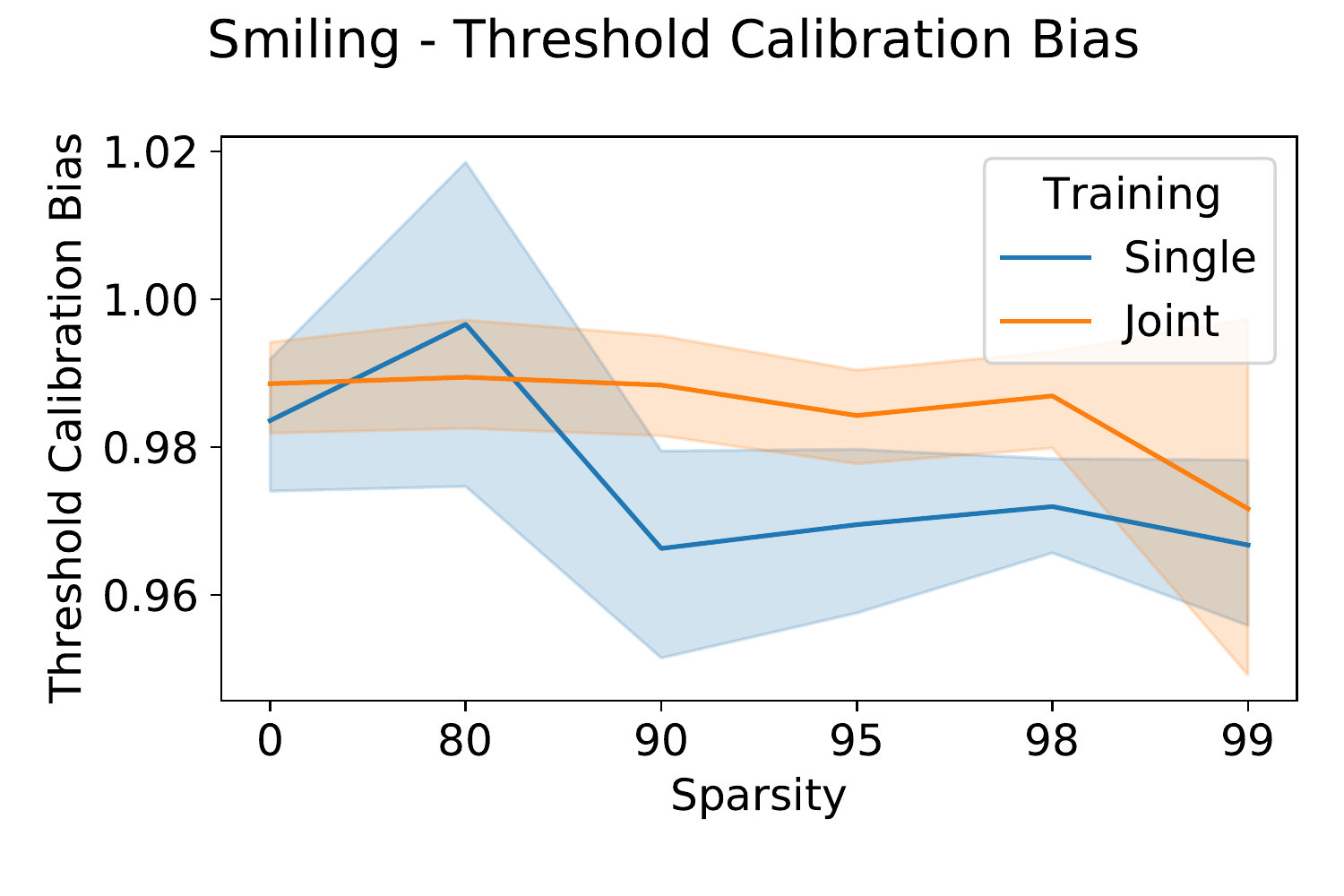}
  \\
  \includegraphics[width=0.12\textwidth]
  {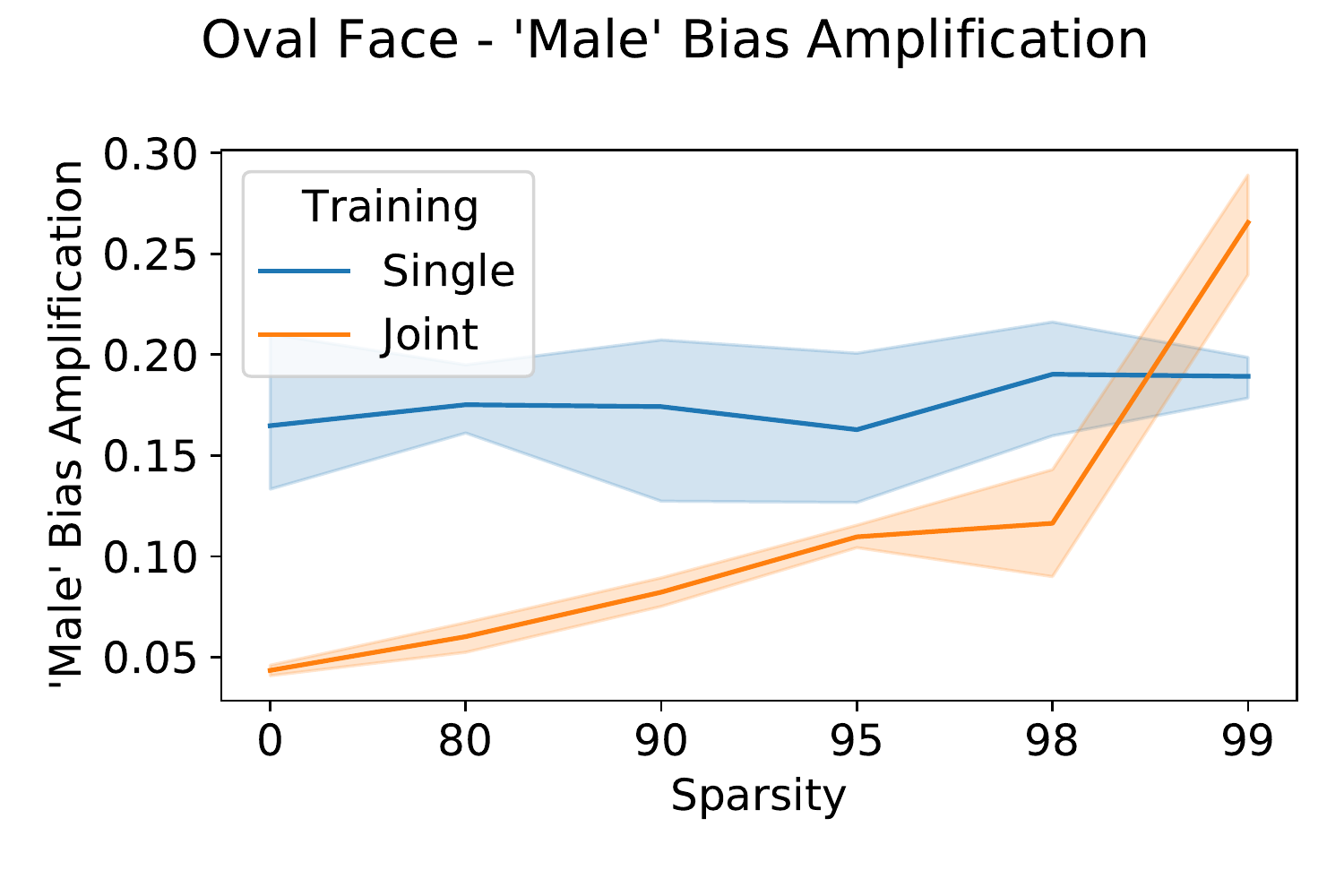} &
\includegraphics[width=0.12\textwidth]{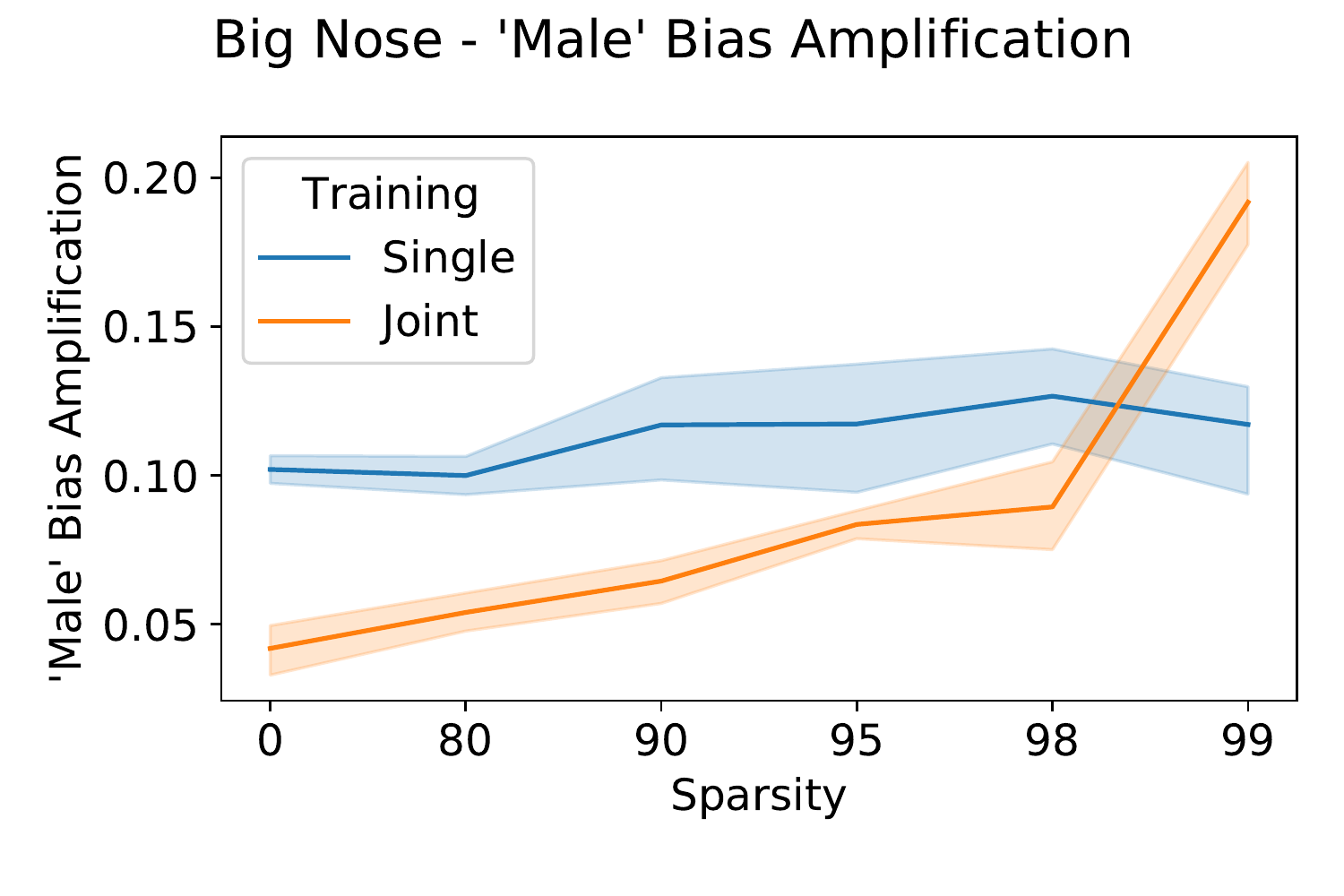} &
\includegraphics[width=0.12\textwidth]{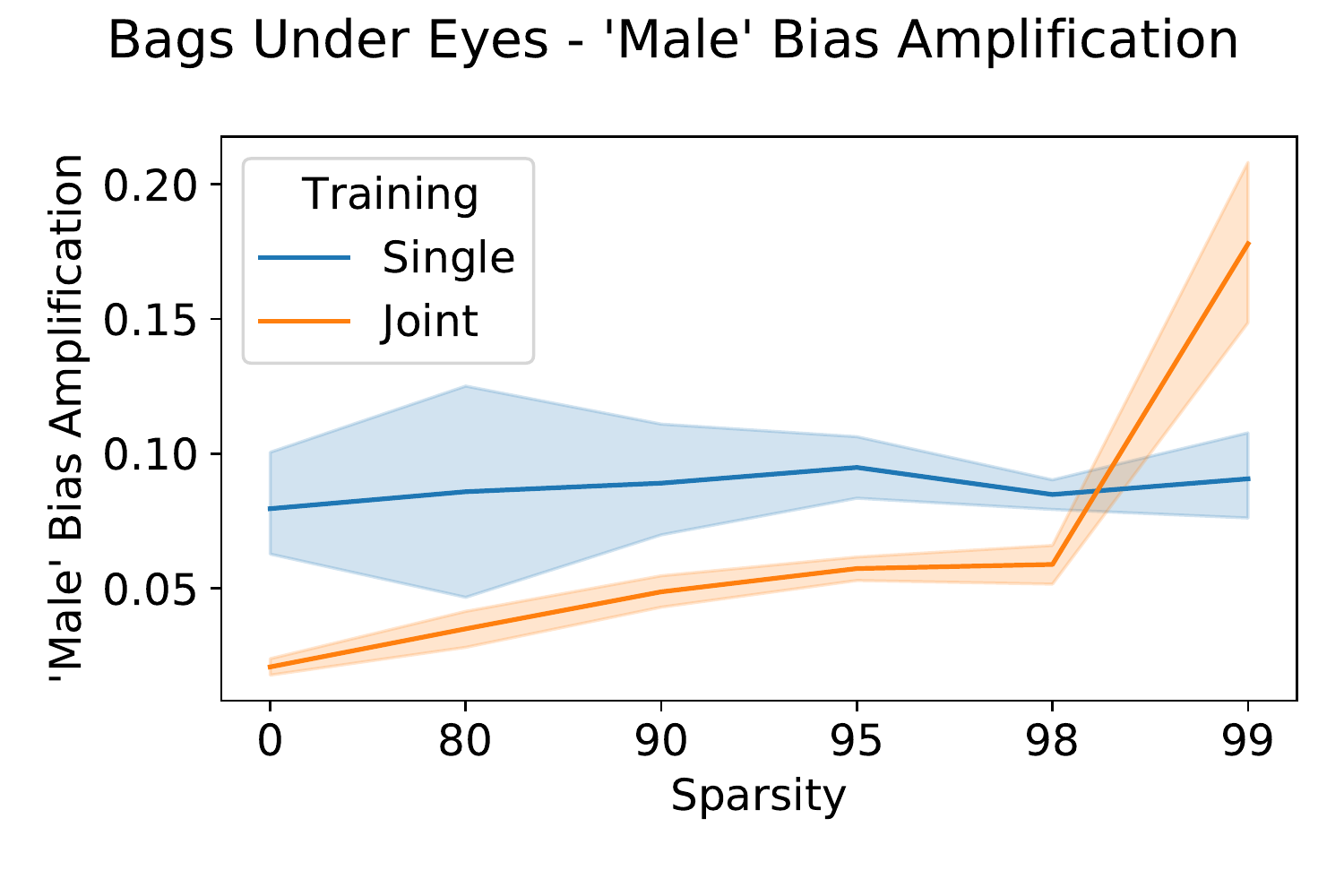} &
\includegraphics[width=0.12\textwidth]{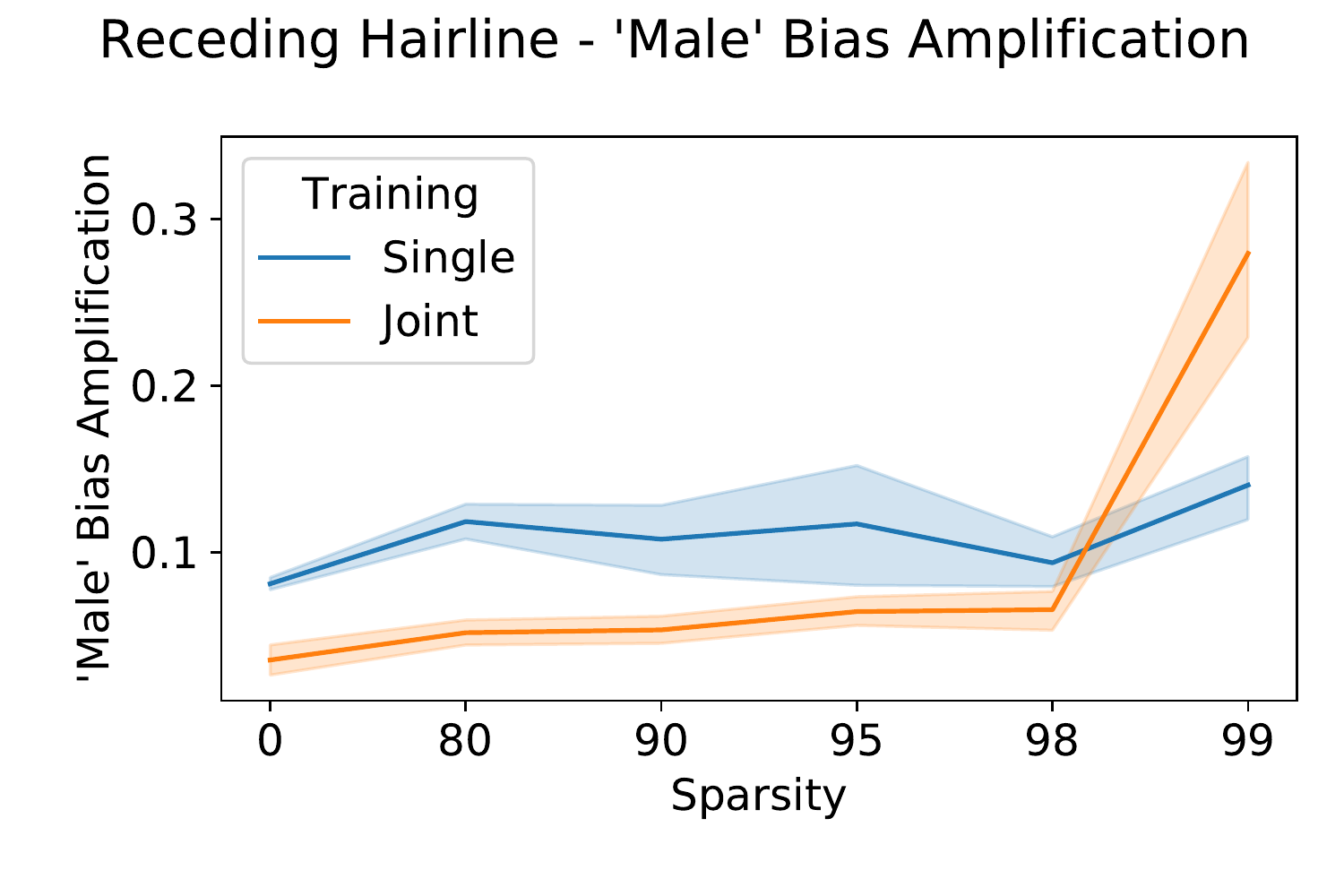} &
&%
\includegraphics[width=0.12\textwidth]{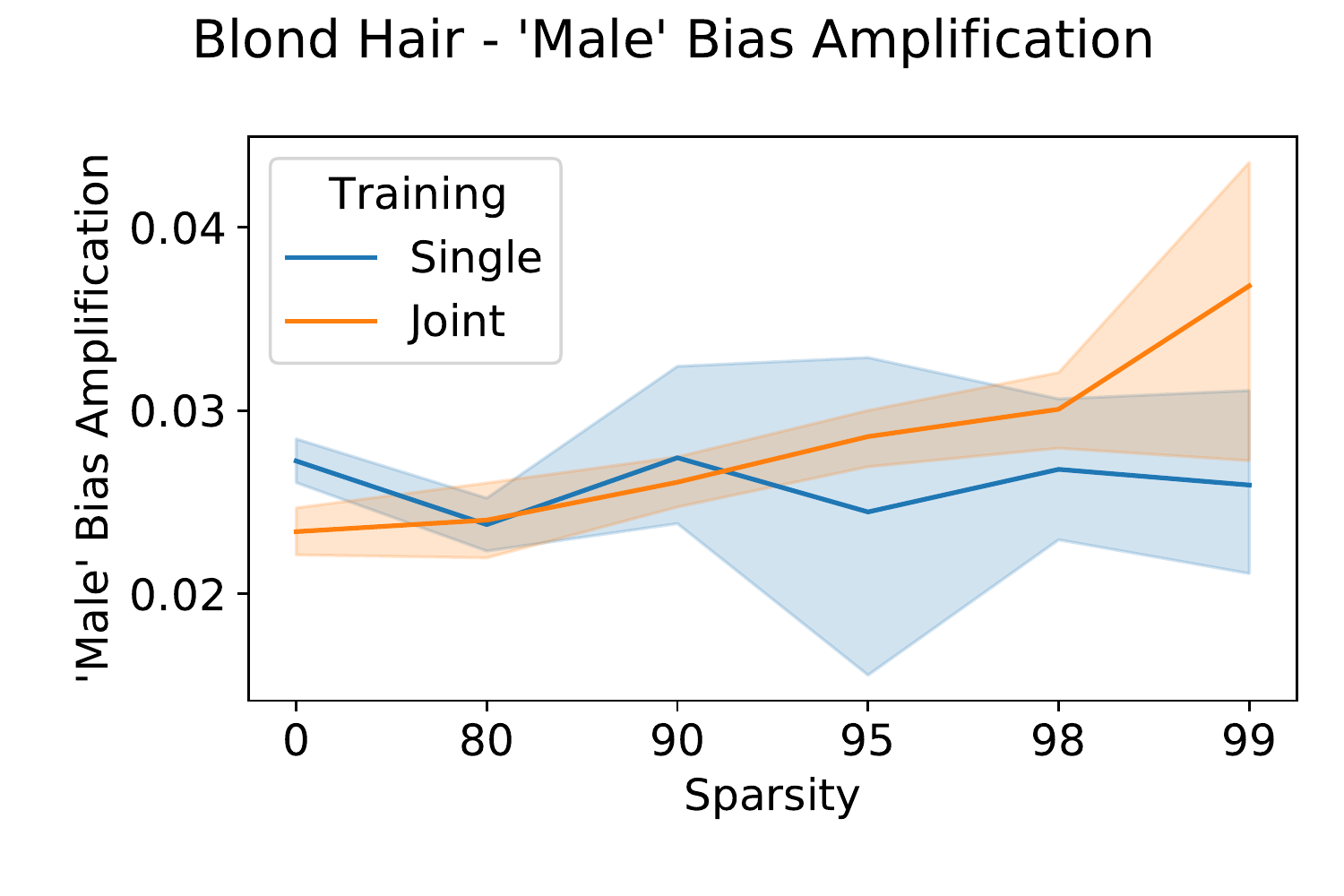} &
\includegraphics[width=0.12\textwidth]{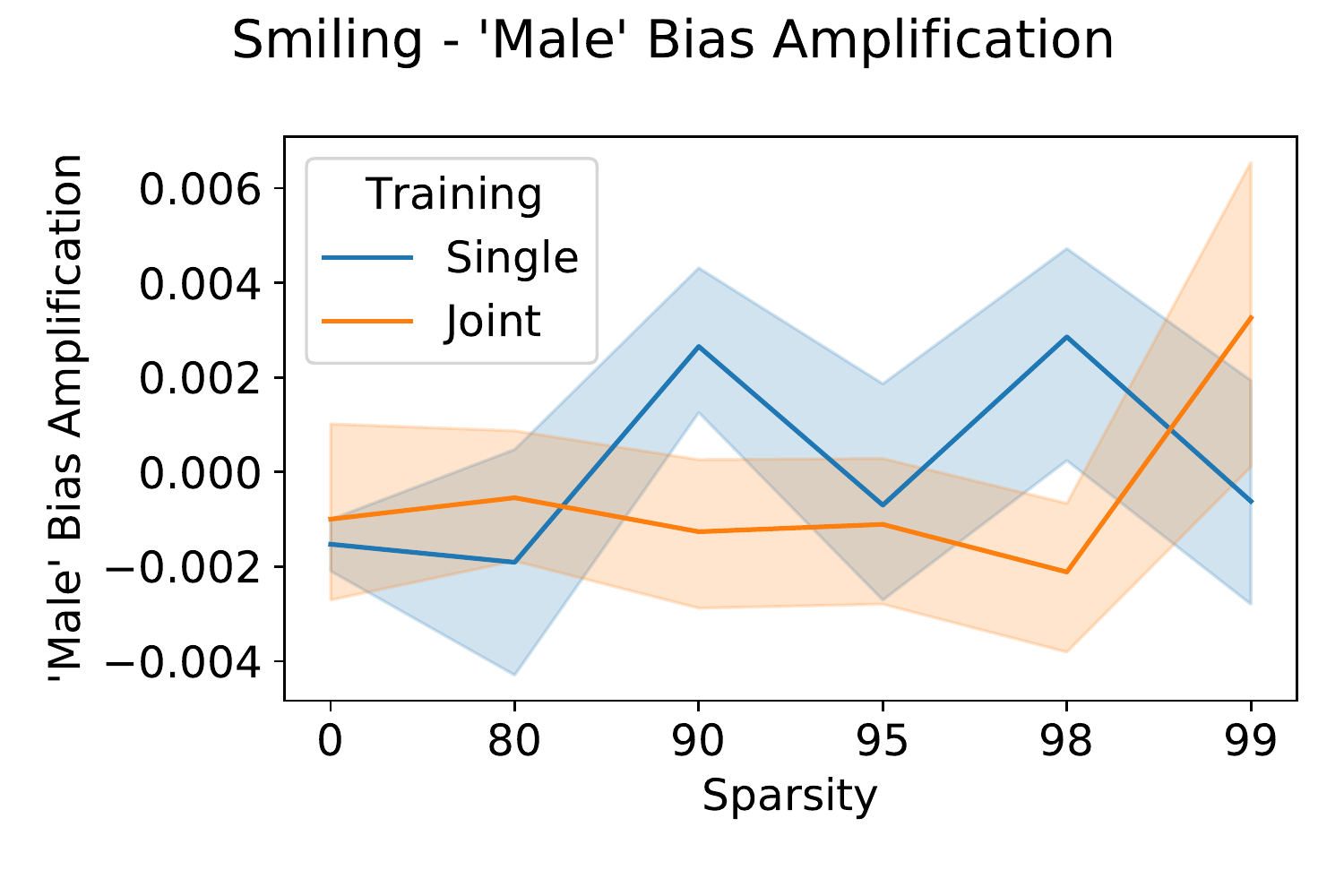}
    \\
  \includegraphics[width=0.12\textwidth]
  {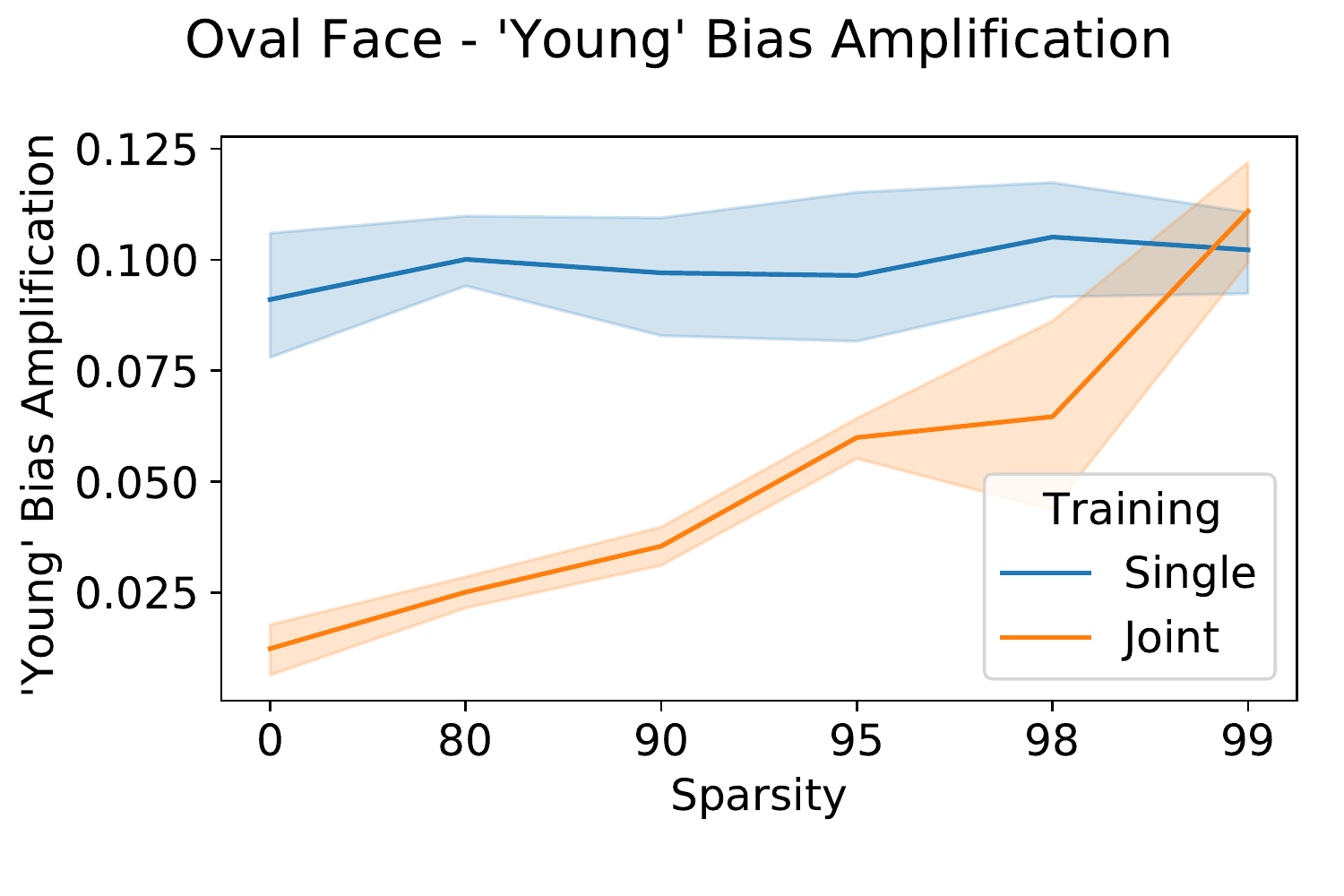} &
\includegraphics[width=0.12\textwidth]{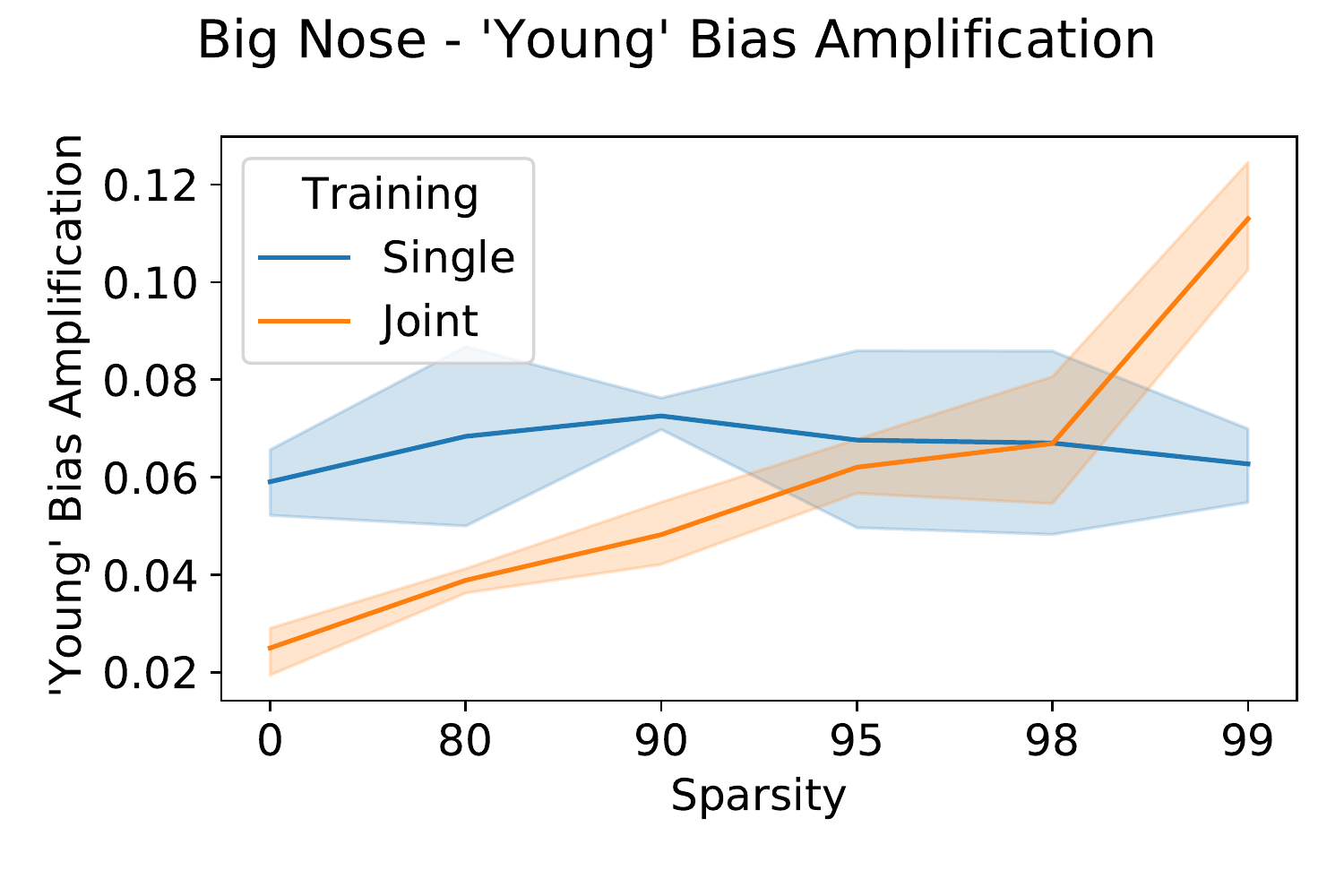} &
\includegraphics[width=0.12\textwidth]{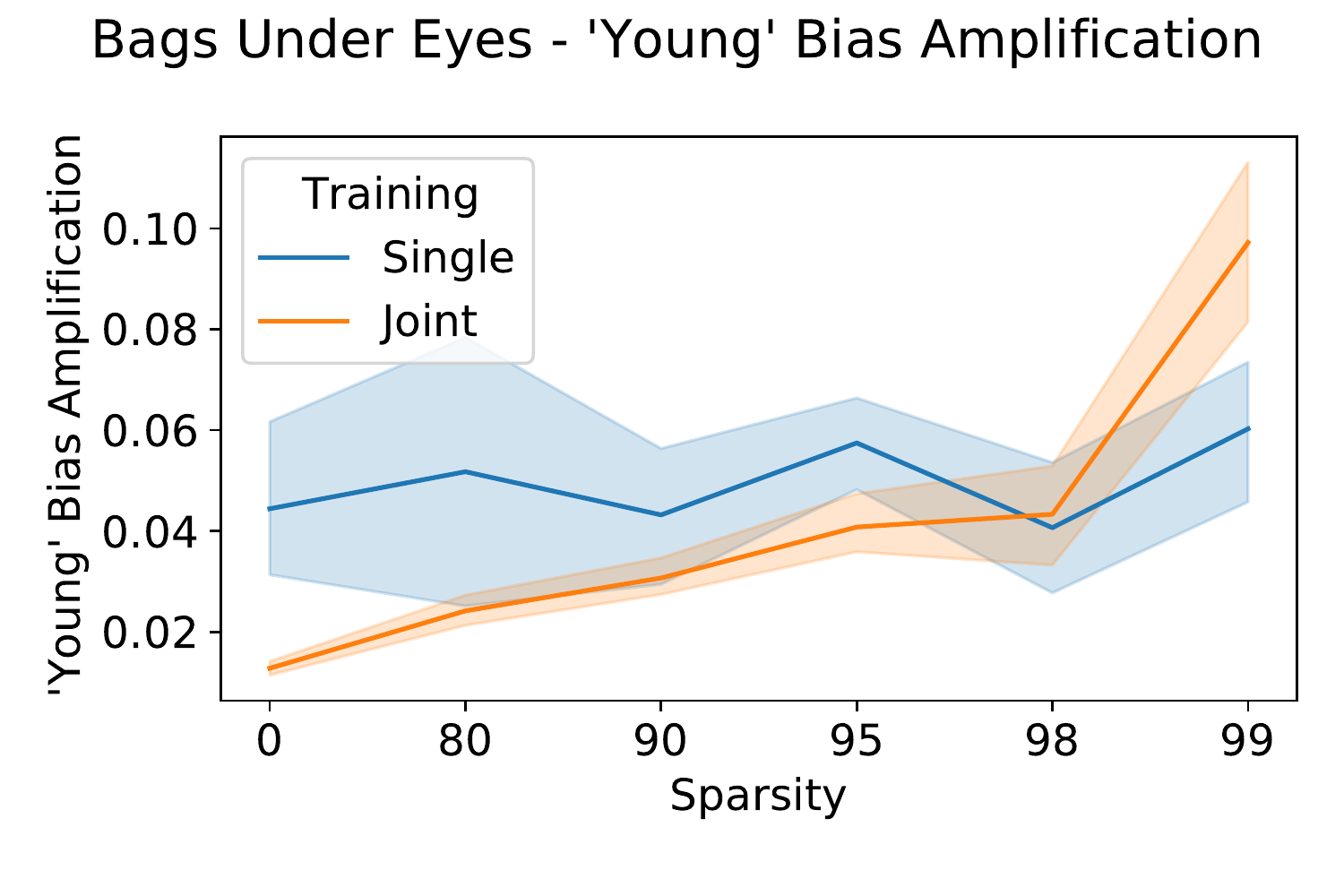} &
\includegraphics[width=0.12\textwidth]{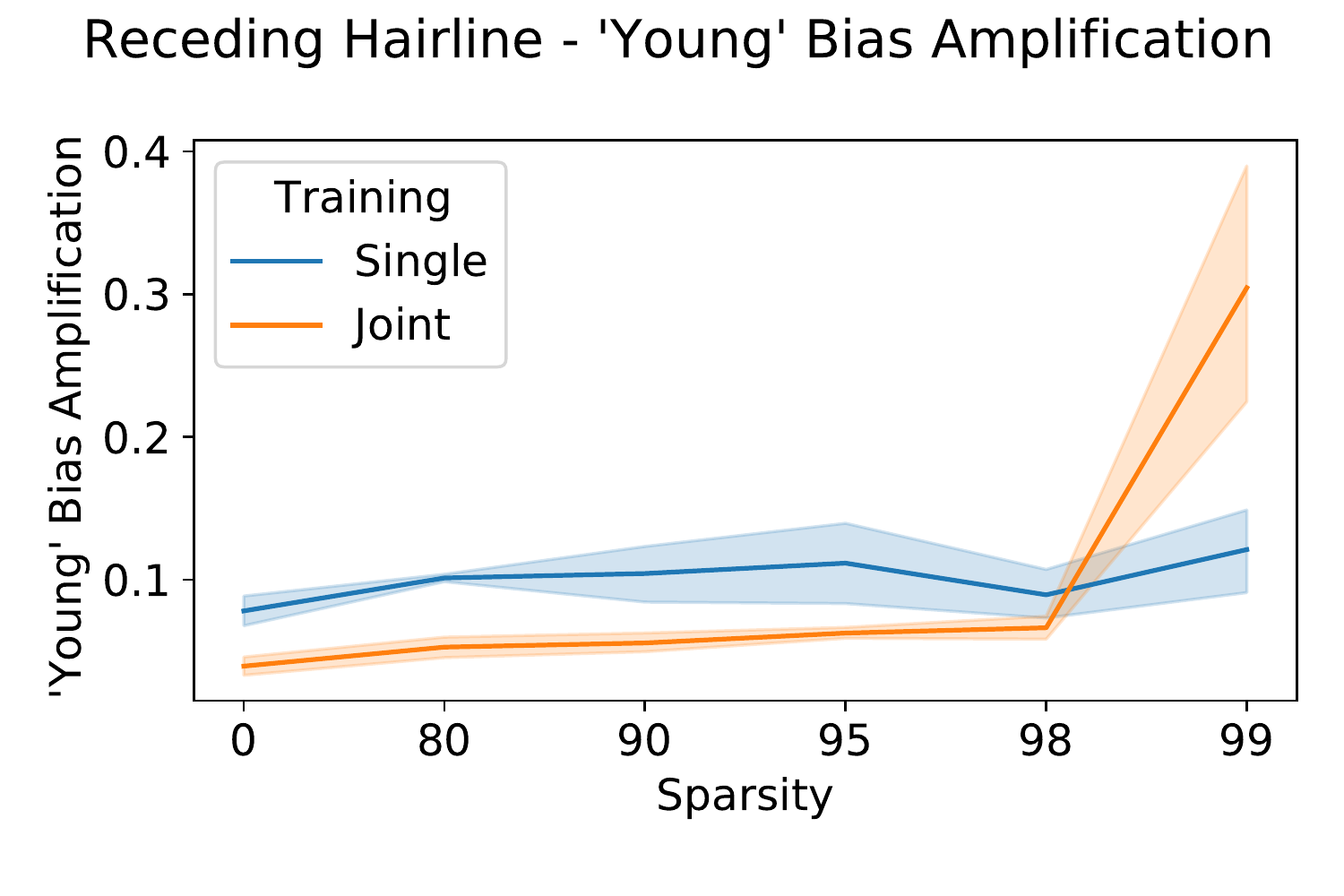} &
\includegraphics[width=0.12\textwidth]{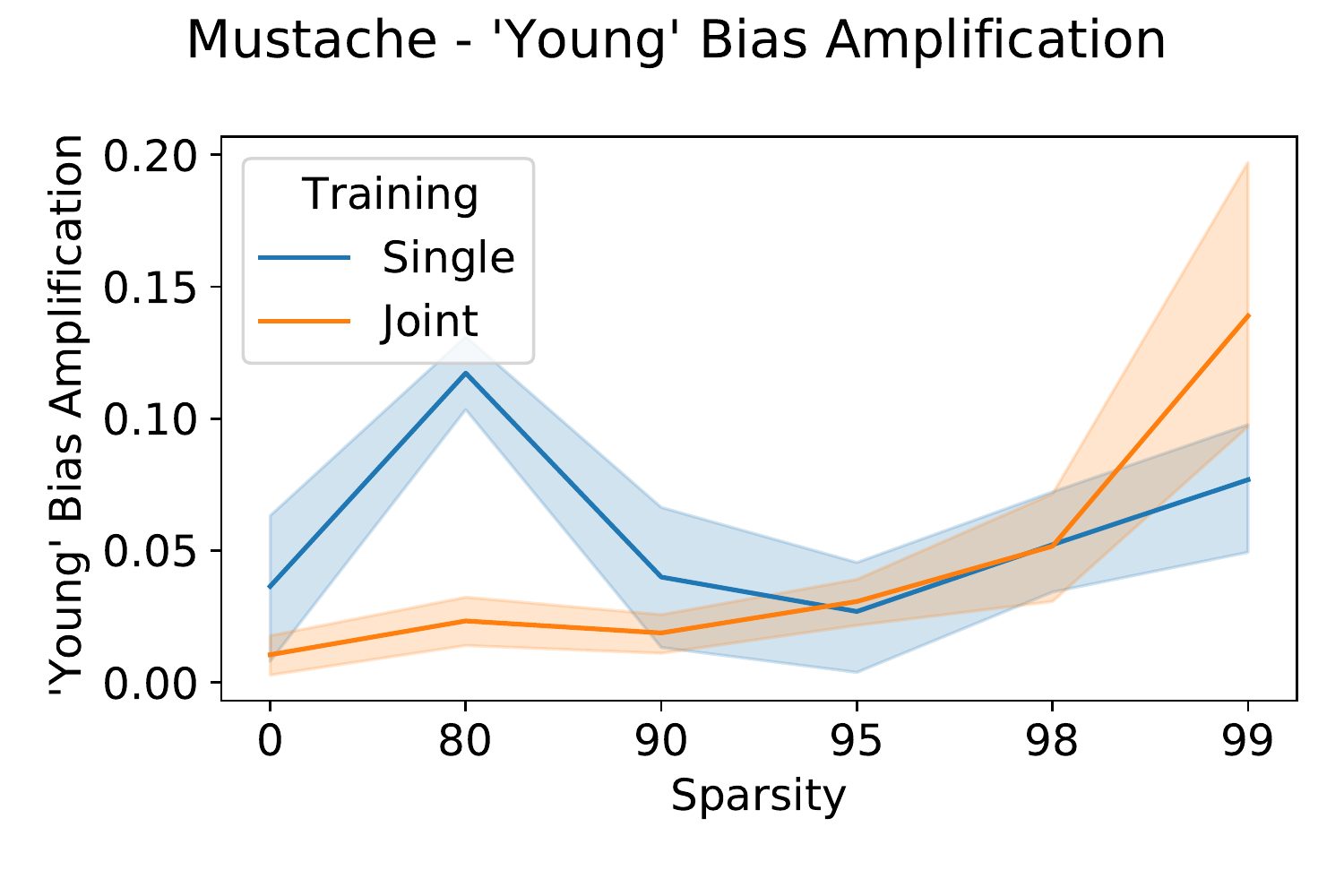} &
\includegraphics[width=0.12\textwidth]{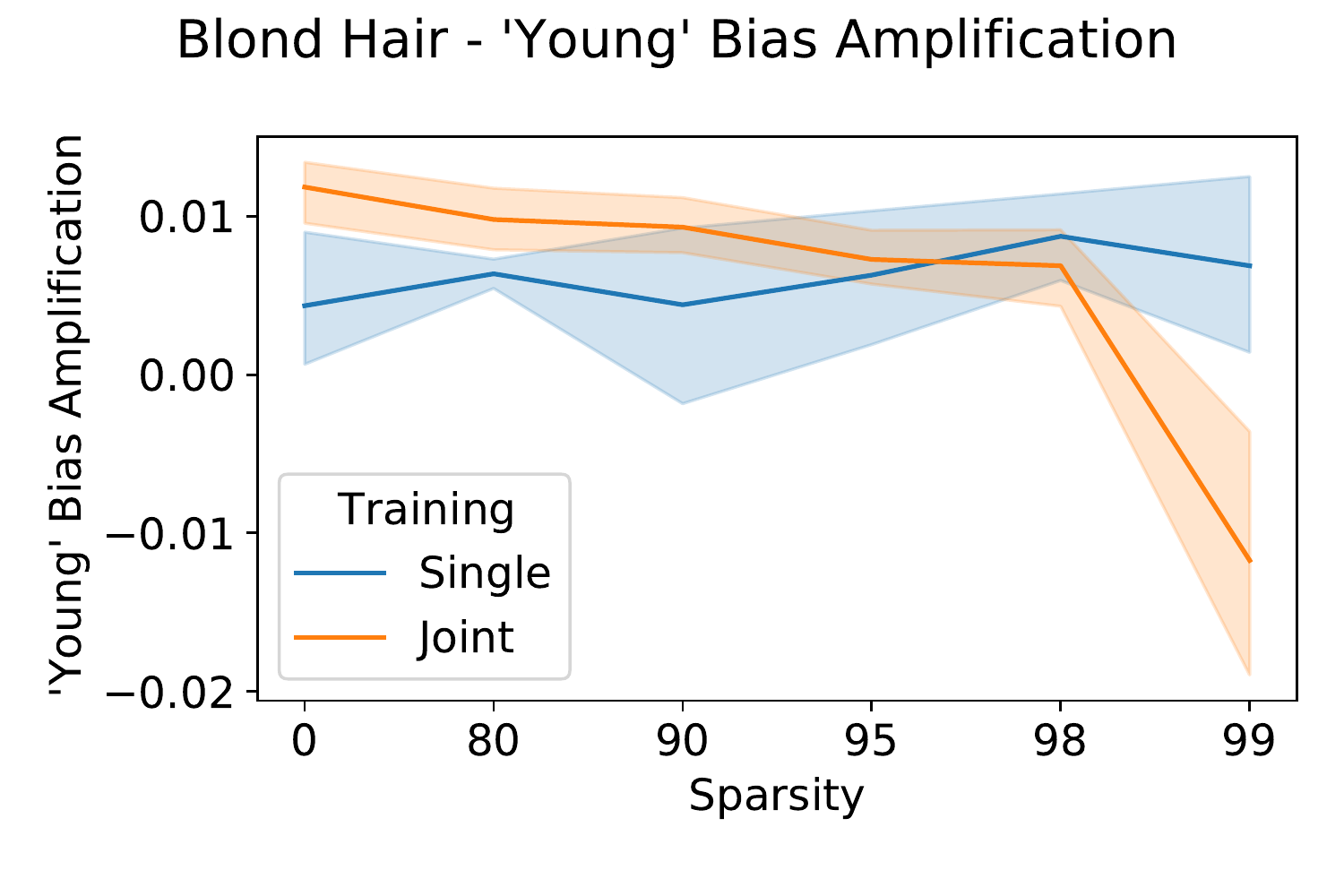} &
    \\
  \includegraphics[width=0.12\textwidth]
  {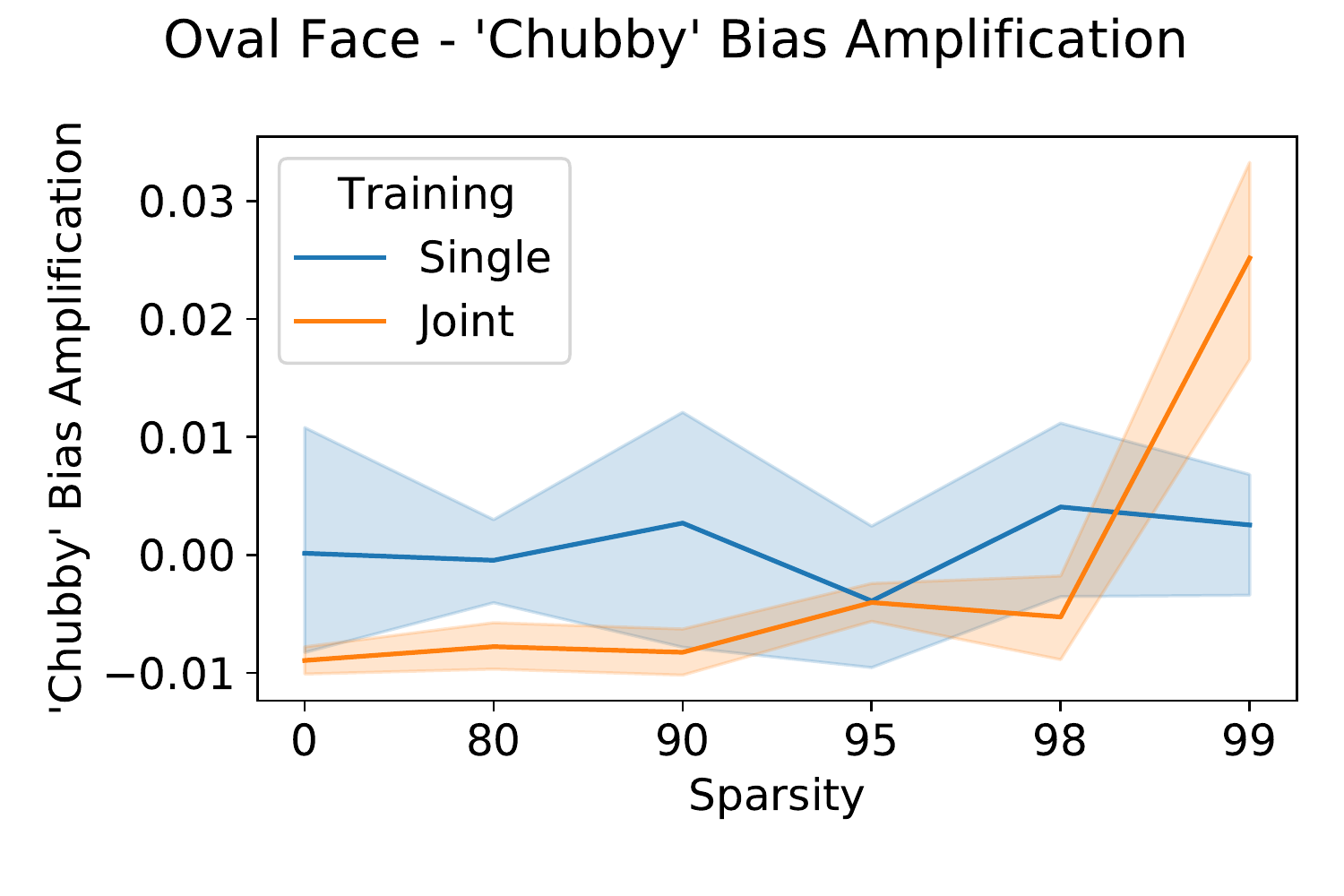} &
\includegraphics[width=0.12\textwidth]{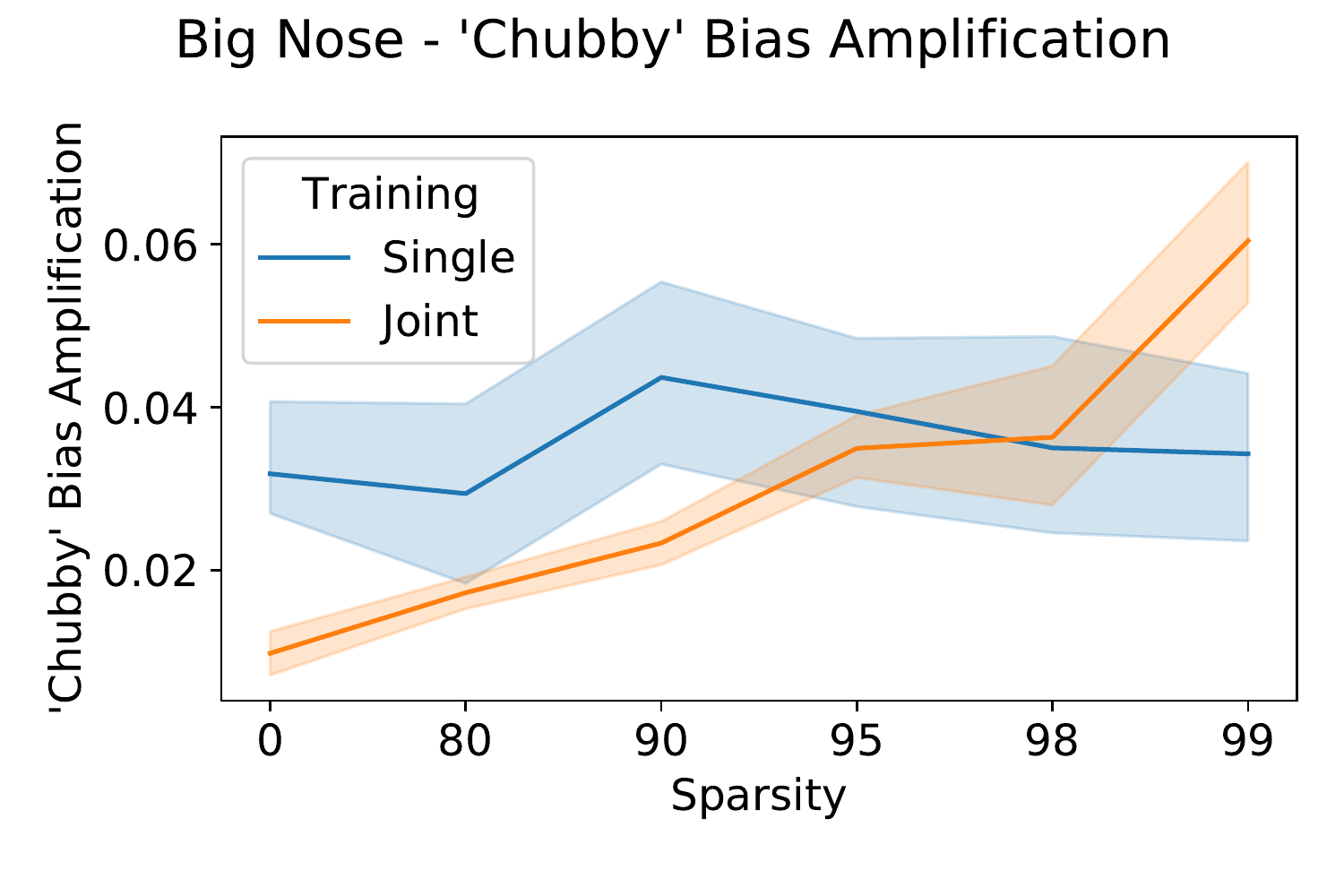} &
\includegraphics[width=0.12\textwidth]{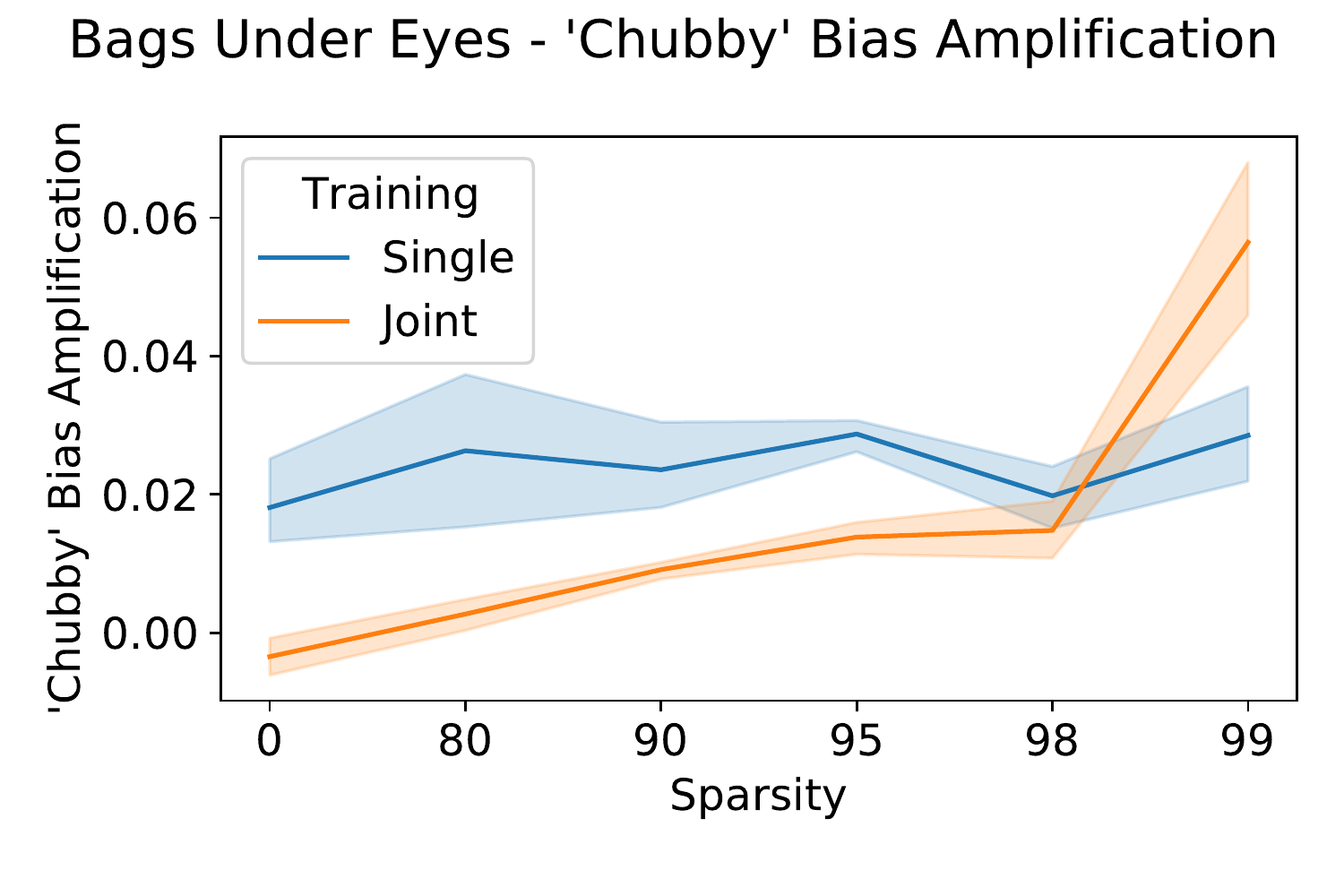} &
\includegraphics[width=0.12\textwidth]{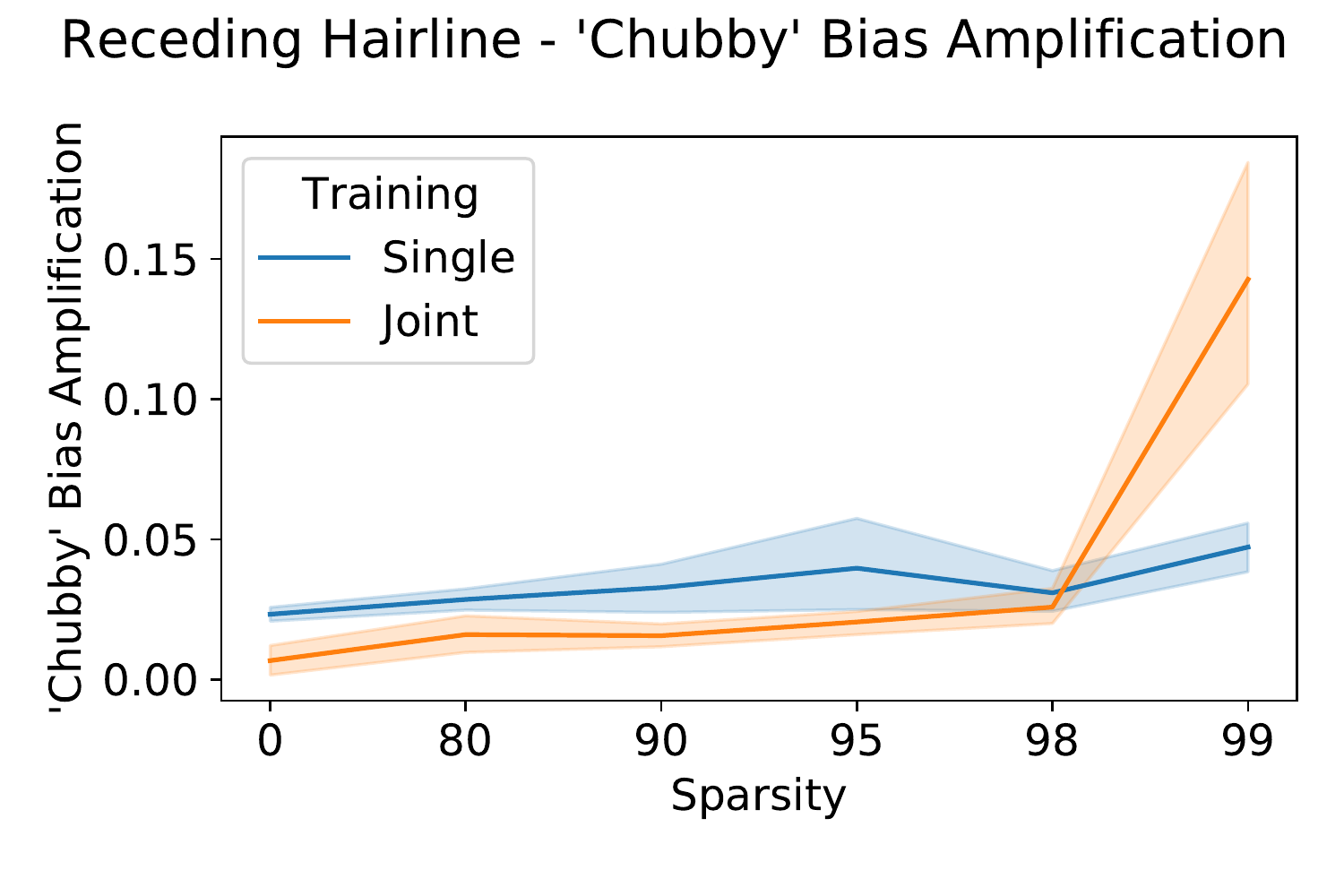} &
\includegraphics[width=0.12\textwidth]{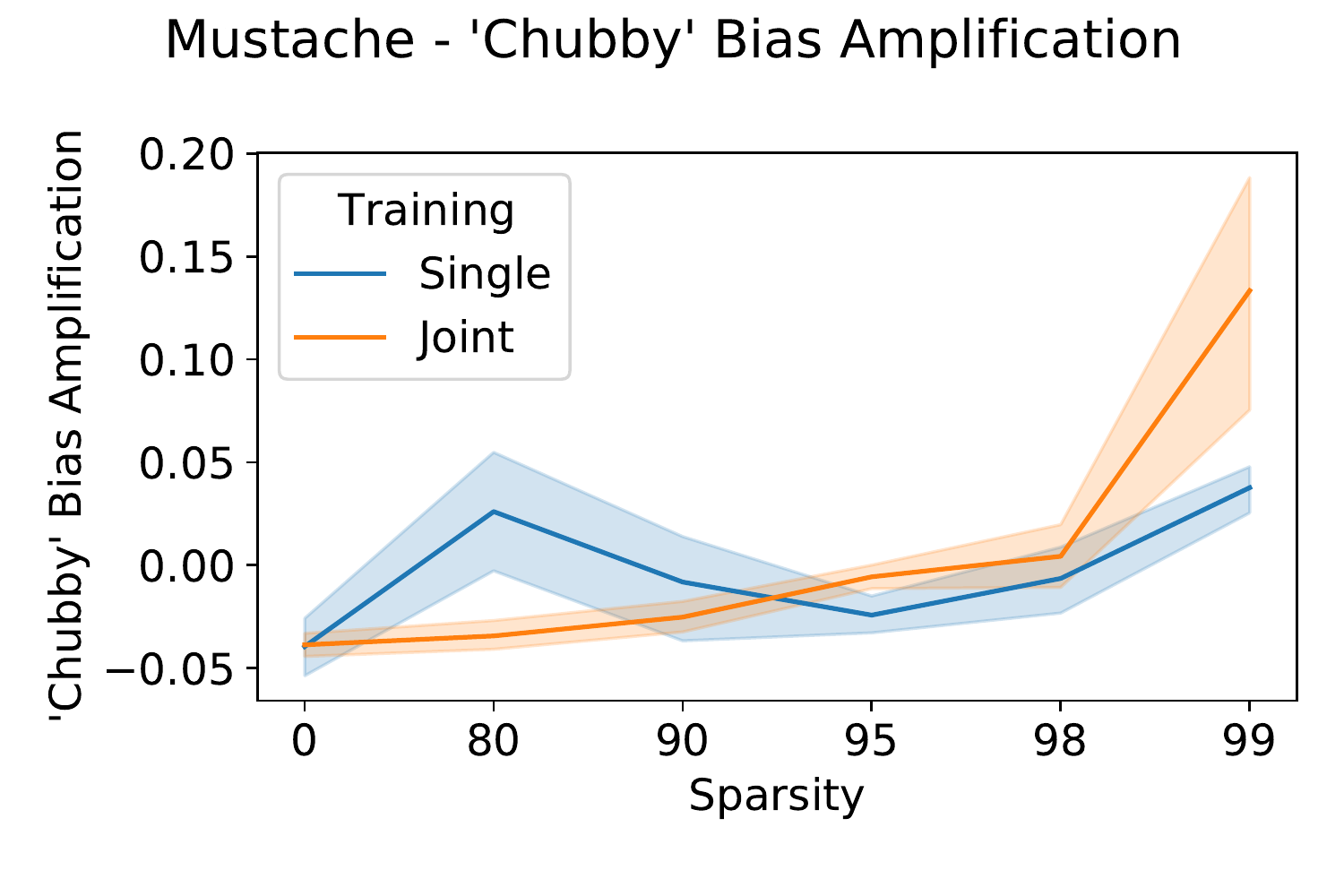} &
\includegraphics[width=0.12\textwidth]{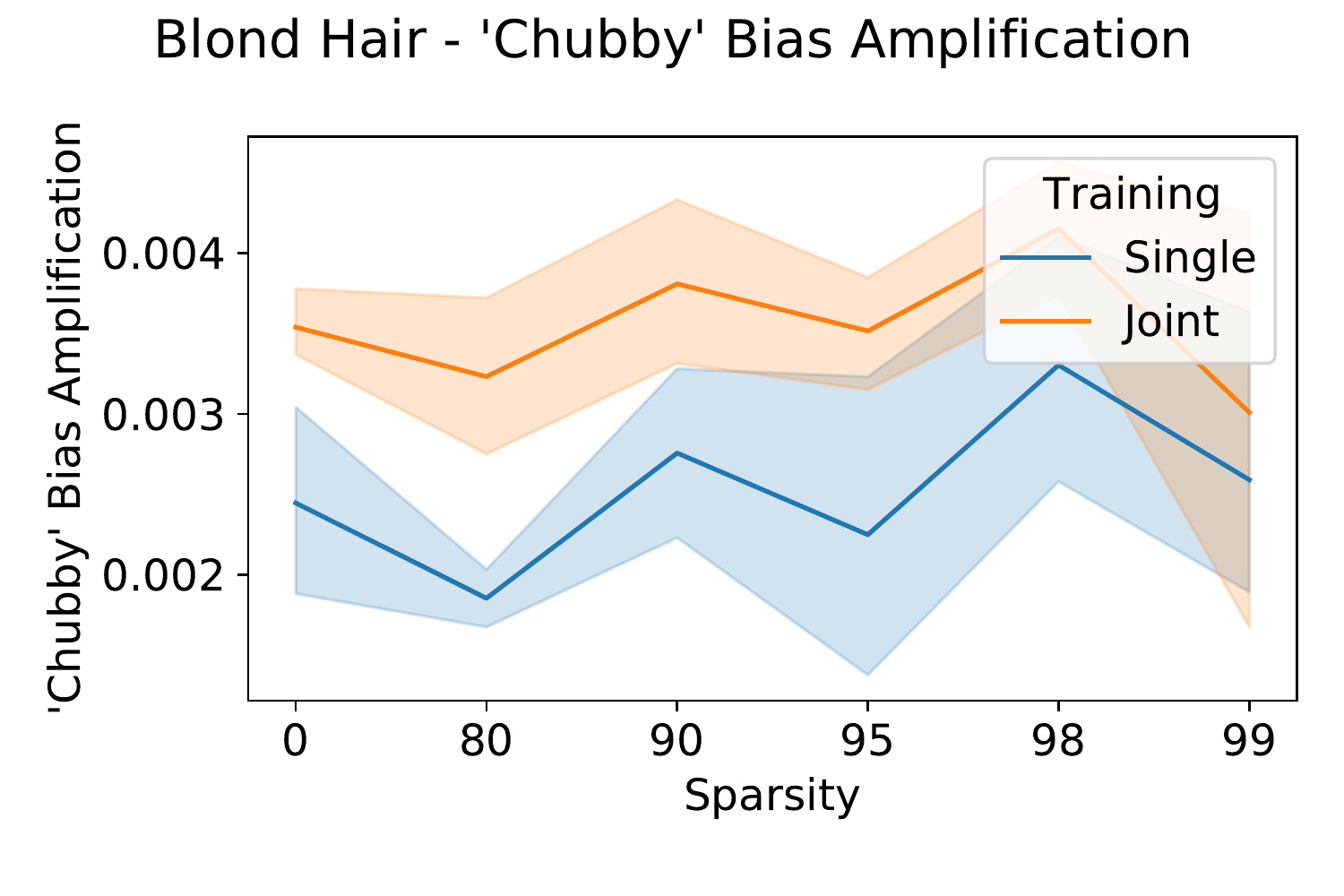} &
\includegraphics[width=0.12\textwidth]{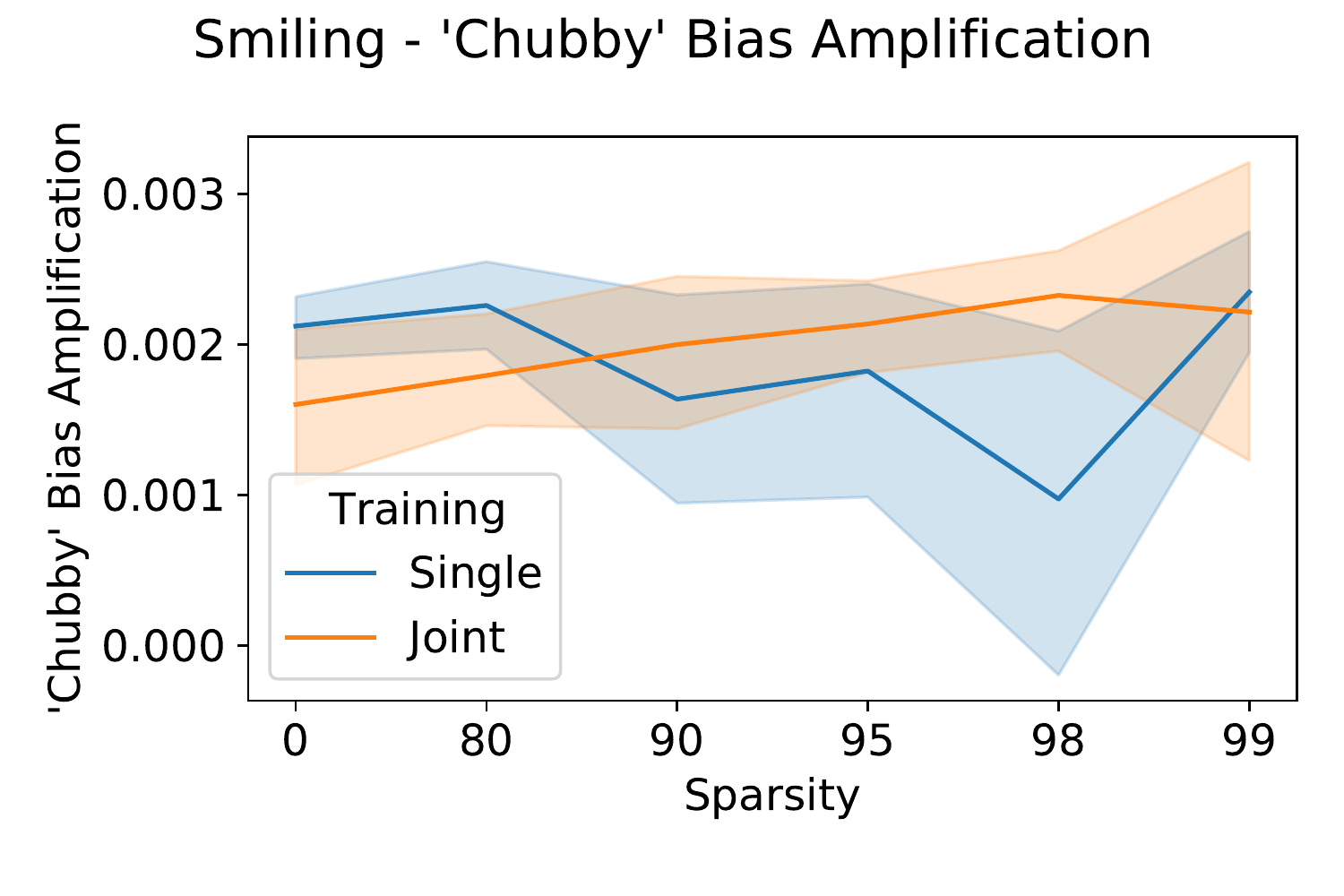}
    \\
      \includegraphics[width=0.12\textwidth]
  {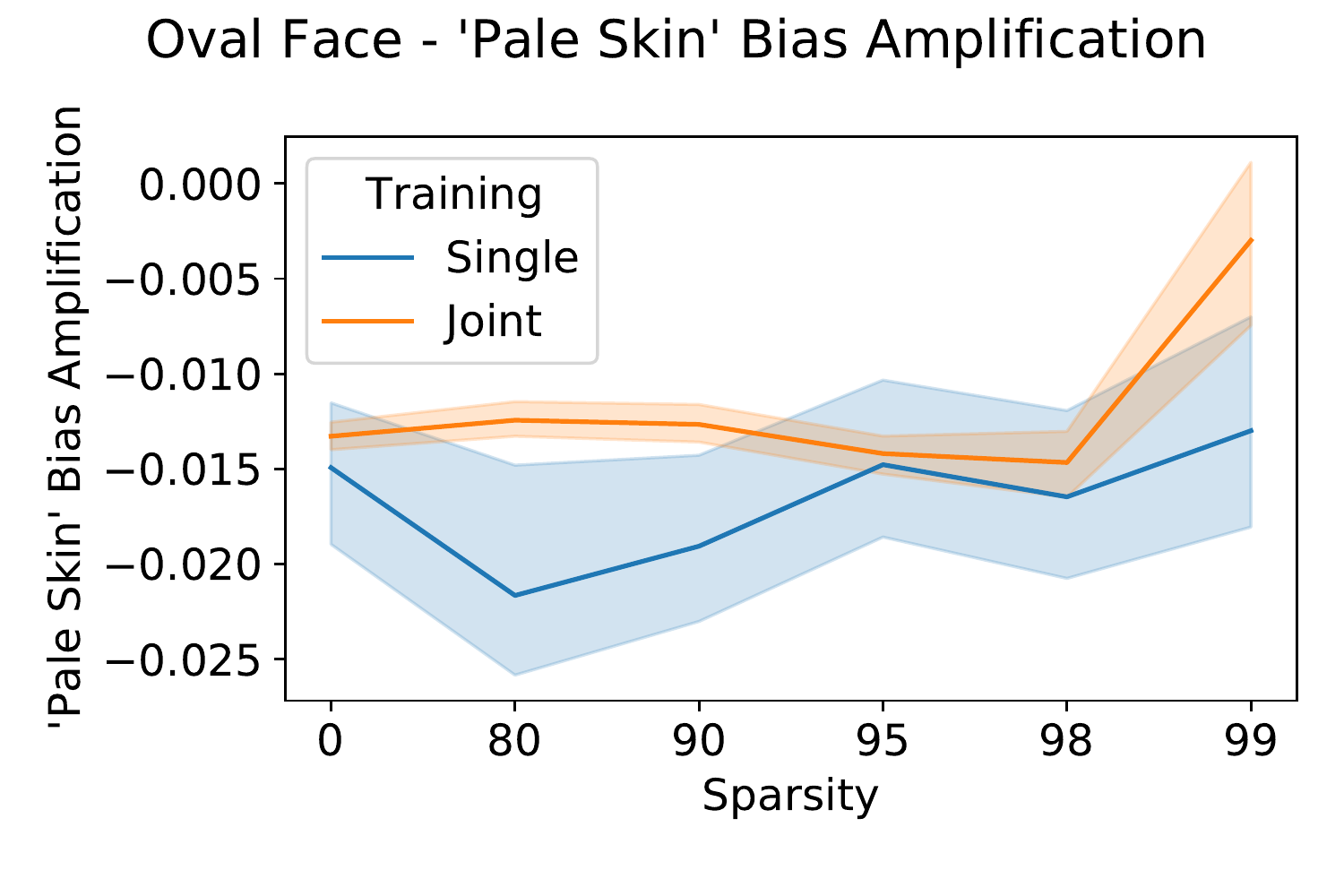} &
\includegraphics[width=0.12\textwidth]{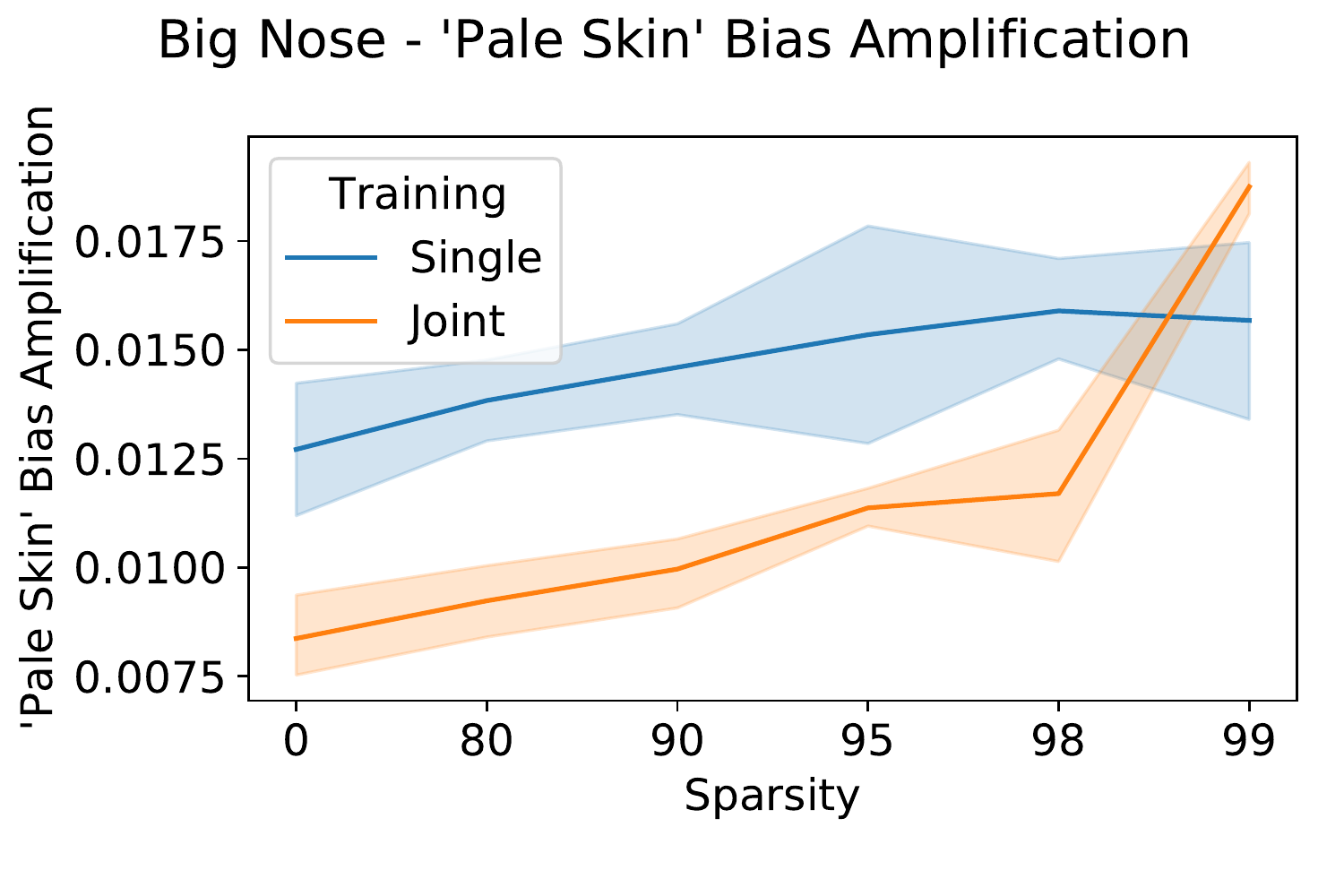} &
\includegraphics[width=0.12\textwidth]{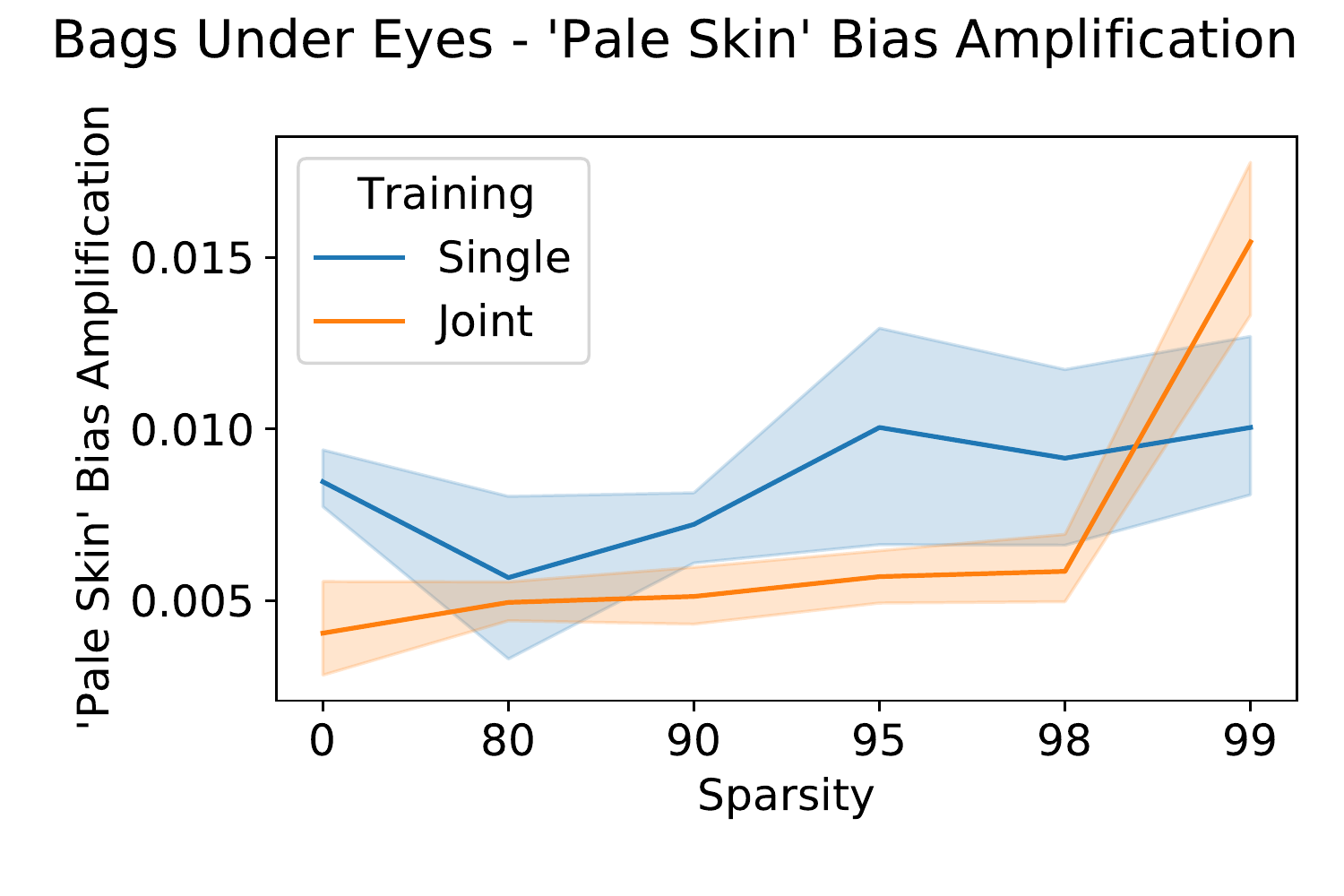} &
\includegraphics[width=0.12\textwidth]{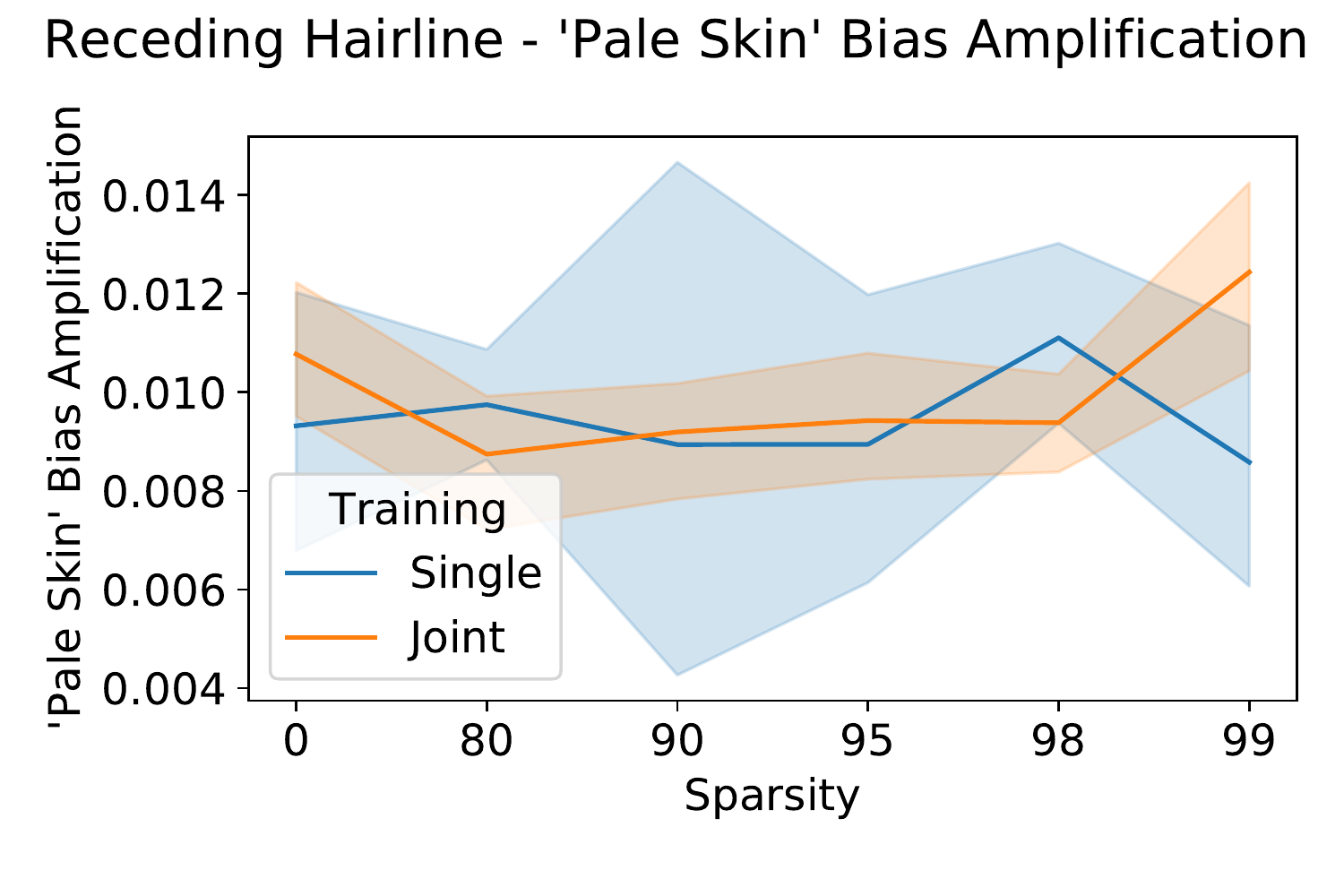} &
&%
\includegraphics[width=0.12\textwidth]{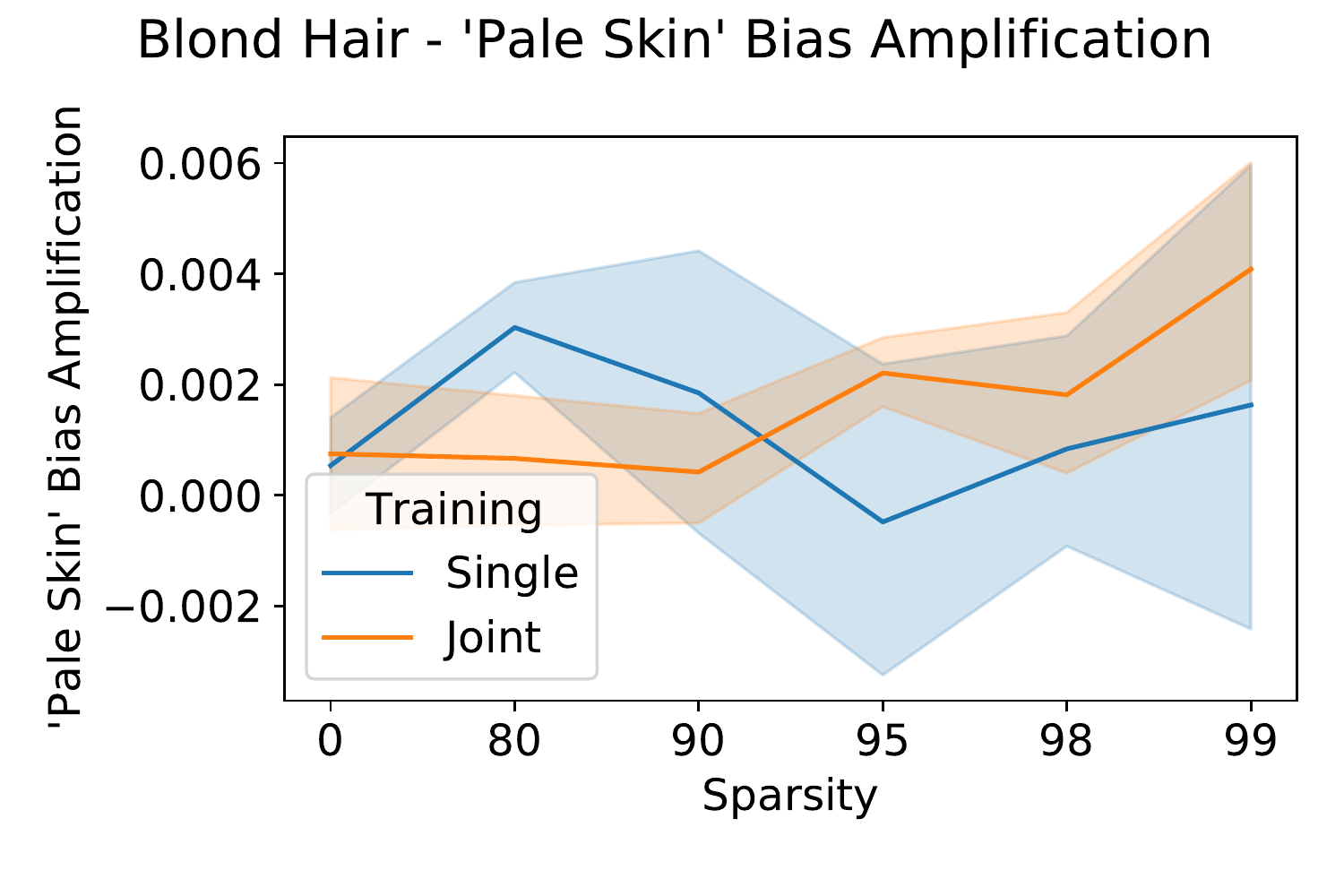} &
\includegraphics[width=0.12\textwidth]{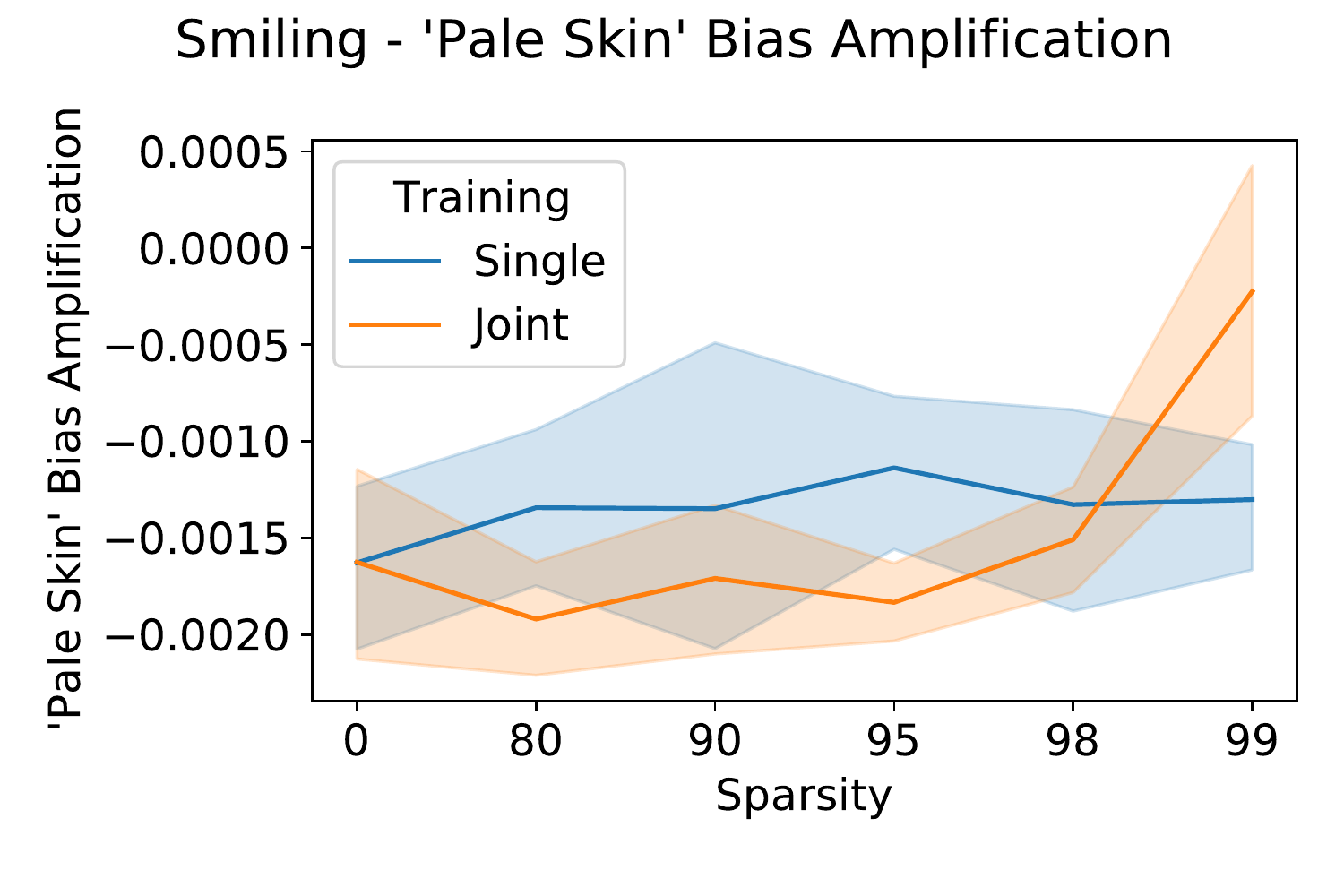}
    \\
\end{tabular}
    \caption{[CelebA / MobleNetV1 / Single Attribute / GMP-RI] Effect of single versus joint training of attributes on Accuracy (first row), Uncertainty (second row), ECE (third row), Threshold Calibration Bias (fourth row), 
    and Bias Amplification for the `Male', `Young', `Chubby', and `Pale Skin' attributes (fifth-eighth rows), on the MobileNet CelebA model, predicting, from left to right, Oval Face, Big Nose, Bags Under Eyes, Receding Hairline, Mustache, Blond Hair, and Smiling). Orange denotes results from joint runs and blue denotes results from single runs. Omitted panels are cases where BA cannot be computed, either because there is no relationship between the predicted attribute and the category, or because the attribute is not present for one of the values of the category.}
    \label{fig:celeba_mobilenet_single_full}
\end{figure}

We also repeat the single-label experiments on this architecture. Unlike the joint training, performance on singly-trained MobileNet models does not decrease at the 99\% sparsity level, which can be observed in Figure~\ref{fig:celeba_mobilenet_single_full}. Generaly, we observe similar trends in both Systematic and Categorical bias as we observe on ResNet18.

\clearpage
\section{ResNet50 Results}
\label{appendix:resnet50}

We further validate our joint training GMP-RI results on the ResNet50 architecture, which has roughly double the parameters of ResNet18 (25.529.472 versus 11.683.712). We use the same experimental settings as for the ResNet18 GMP-RI experiments, excepting that the ResNet50 experiments were performed only in triplicate (from three random seeds).

\begin{figure}[h]
    \centering
\begin{tabular}{cccc}
   \includegraphics[width=0.22\textwidth]{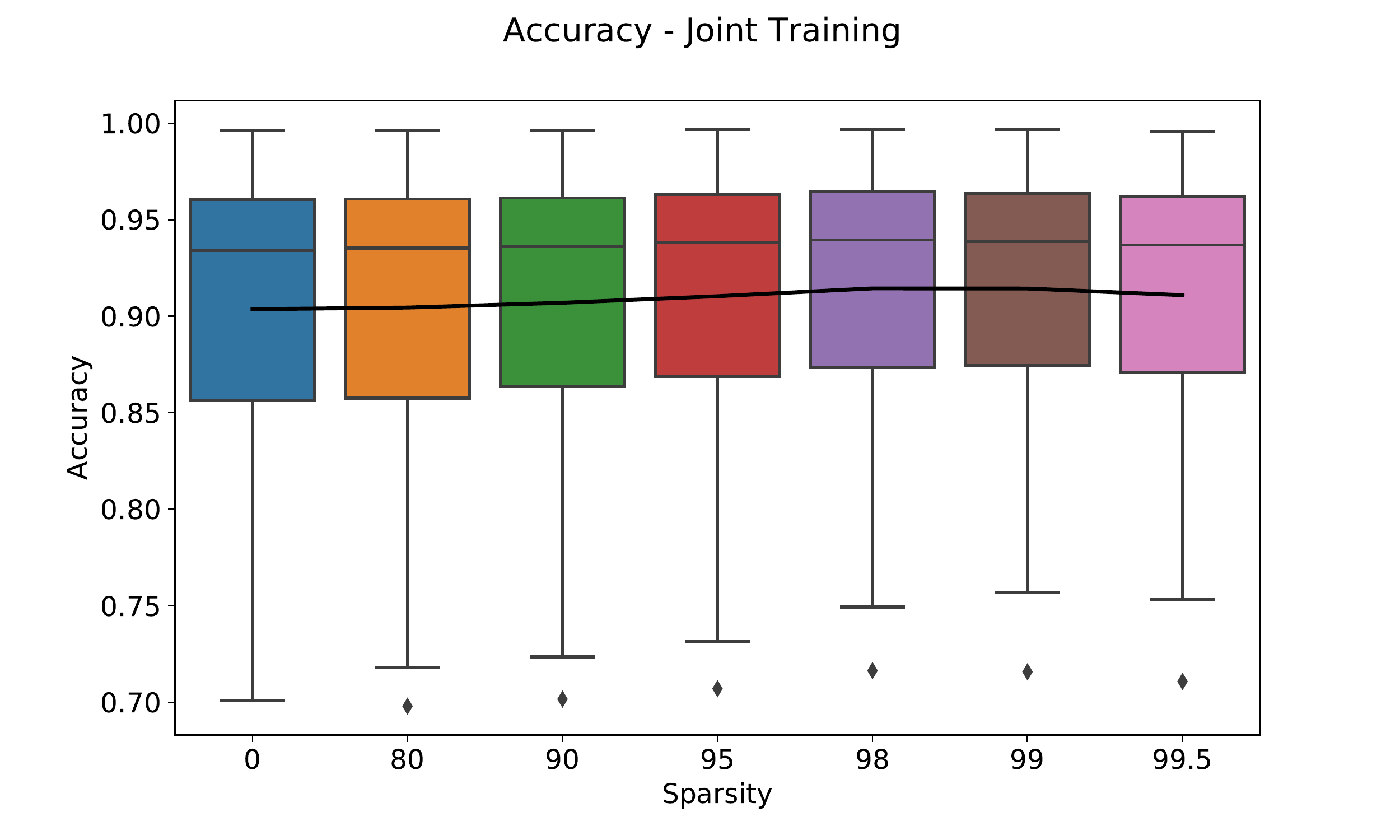} &
   \includegraphics[width=0.22\textwidth]{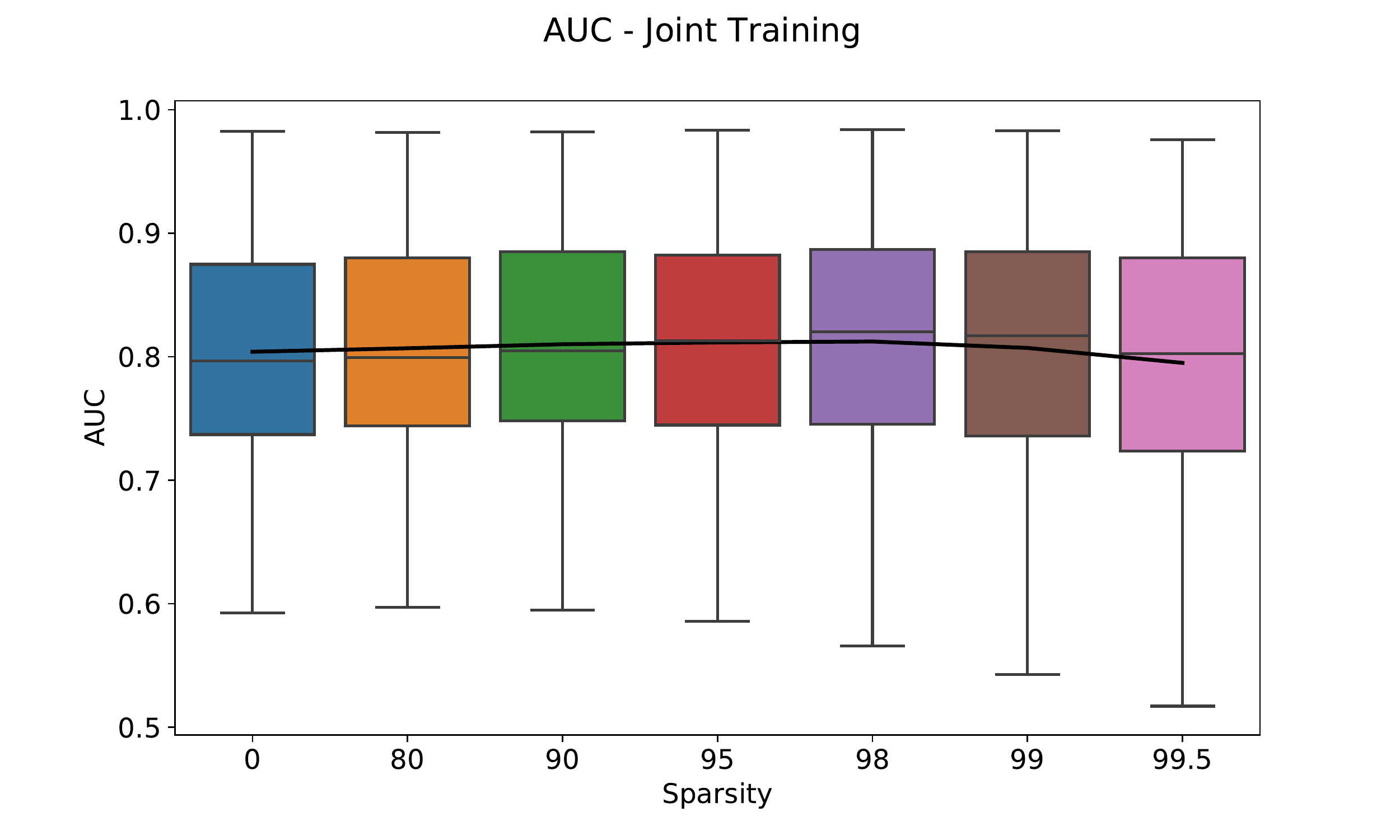} & & \\
    \includegraphics[width=0.22\textwidth]{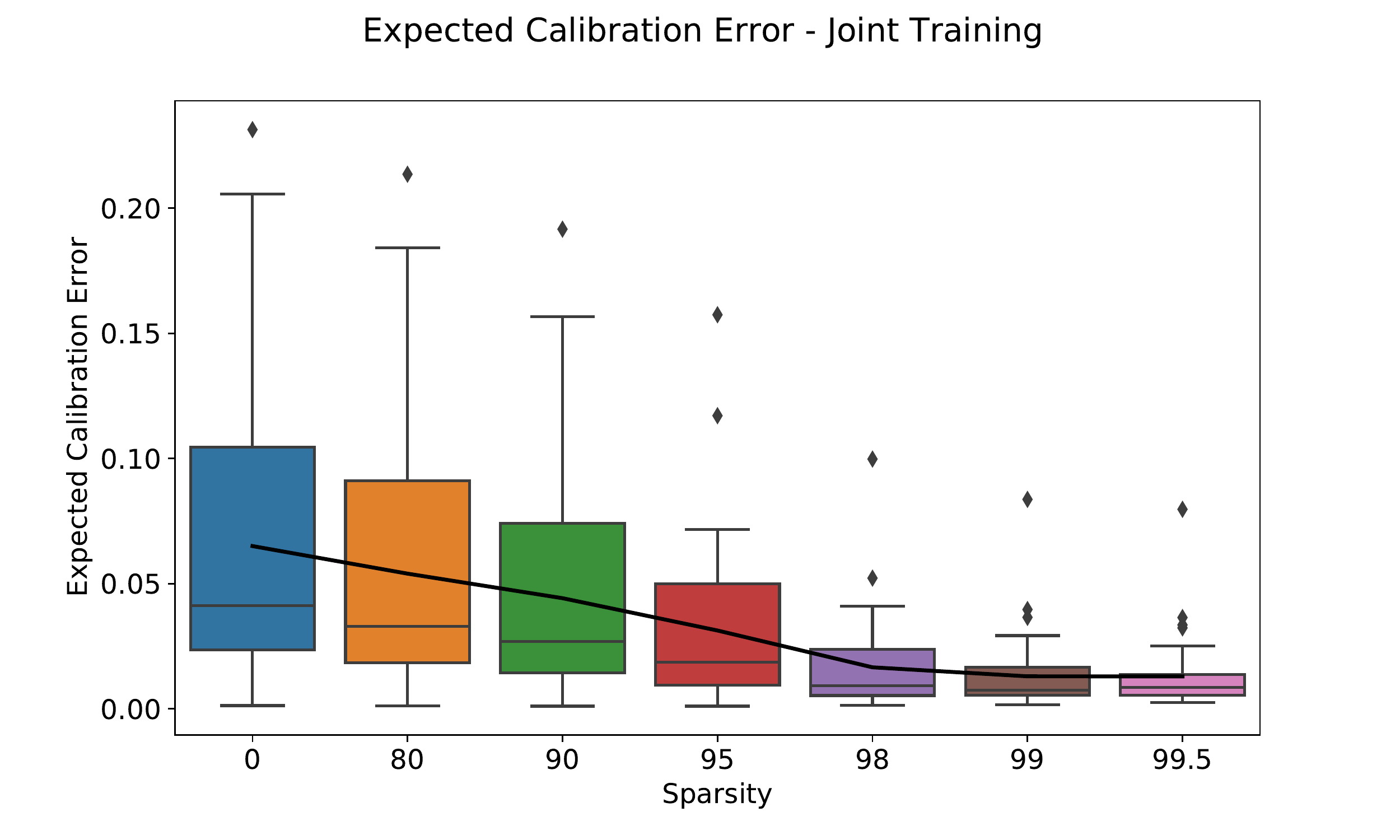} &
    \includegraphics[width=0.22\textwidth]{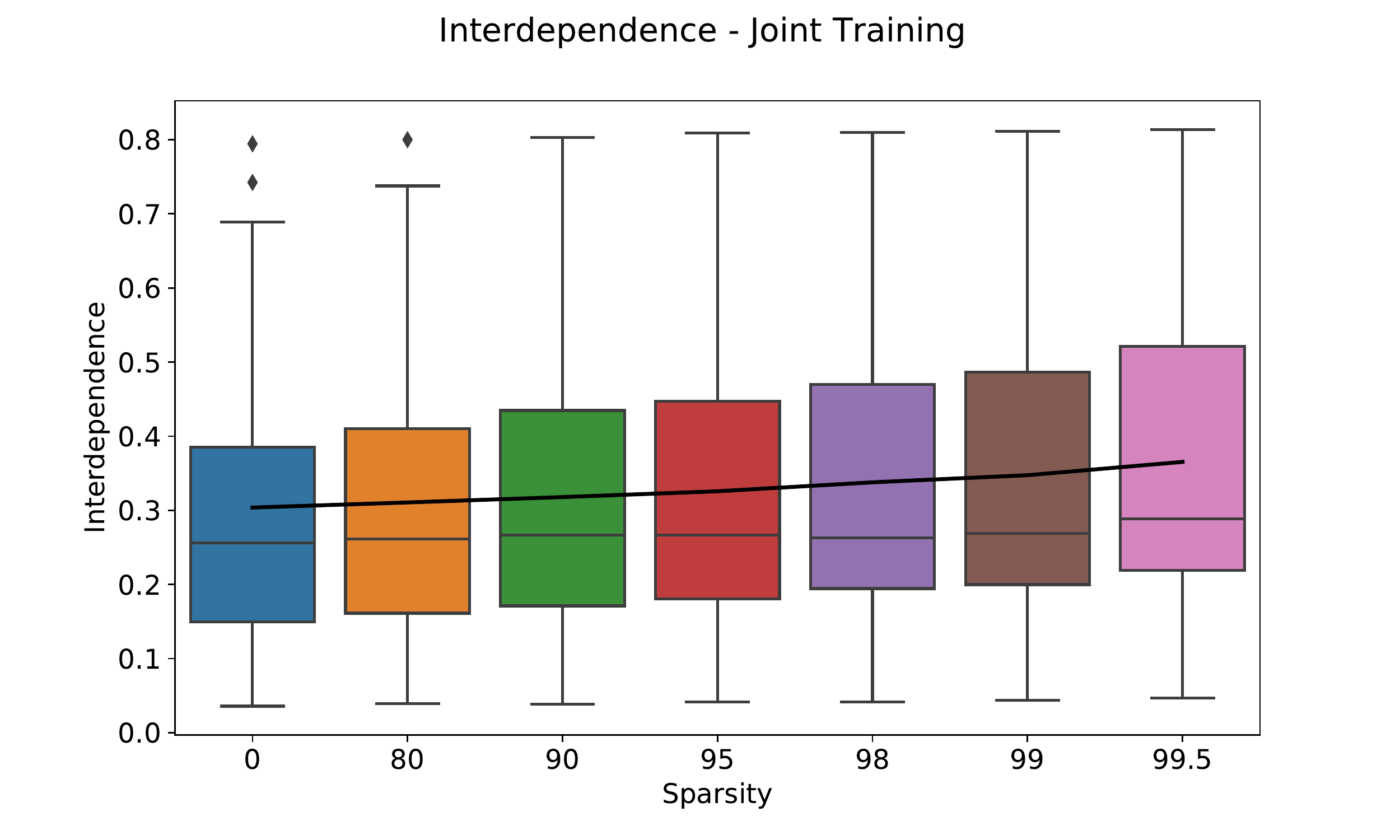} & 
    \includegraphics[width=0.22\textwidth]{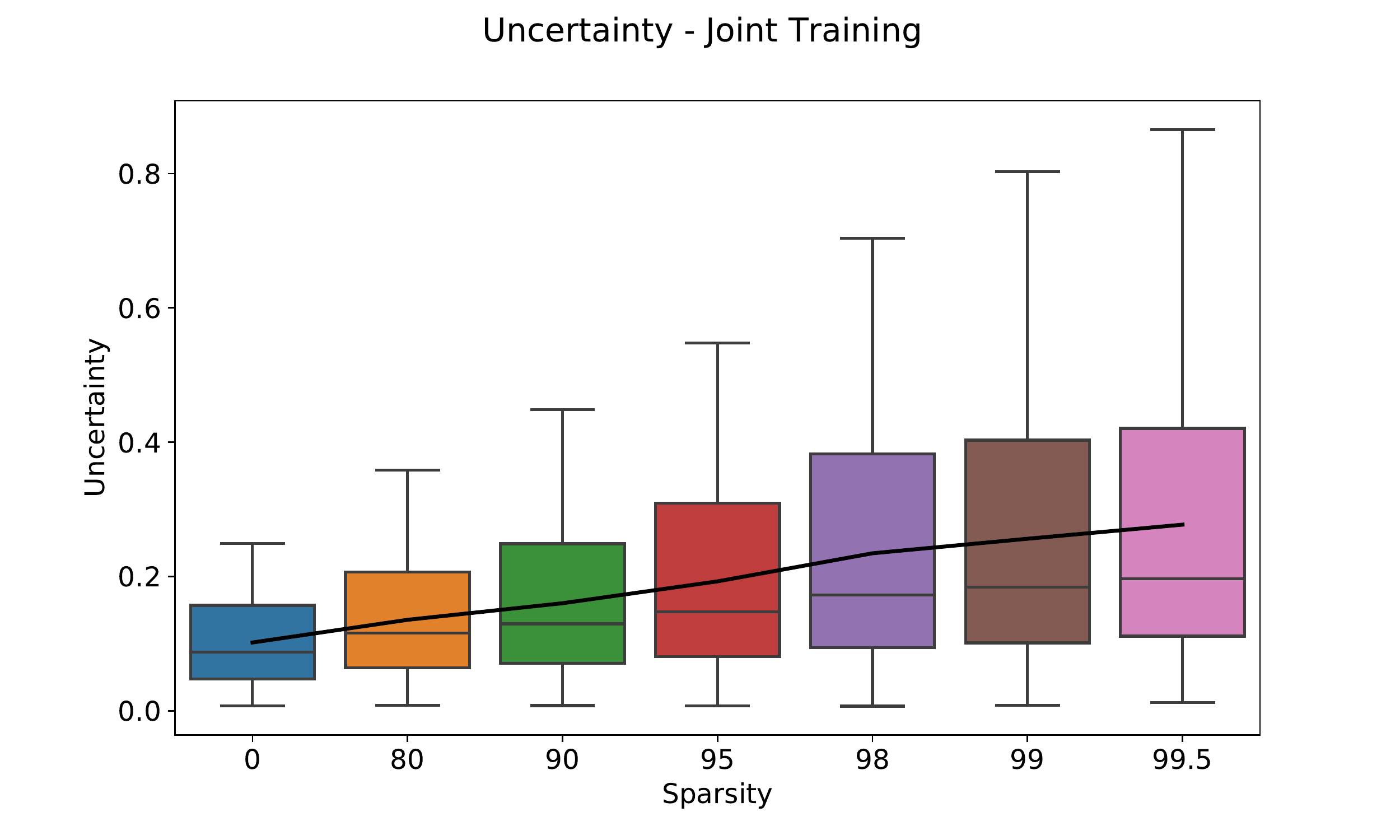} &
    \includegraphics[width=0.22\textwidth]{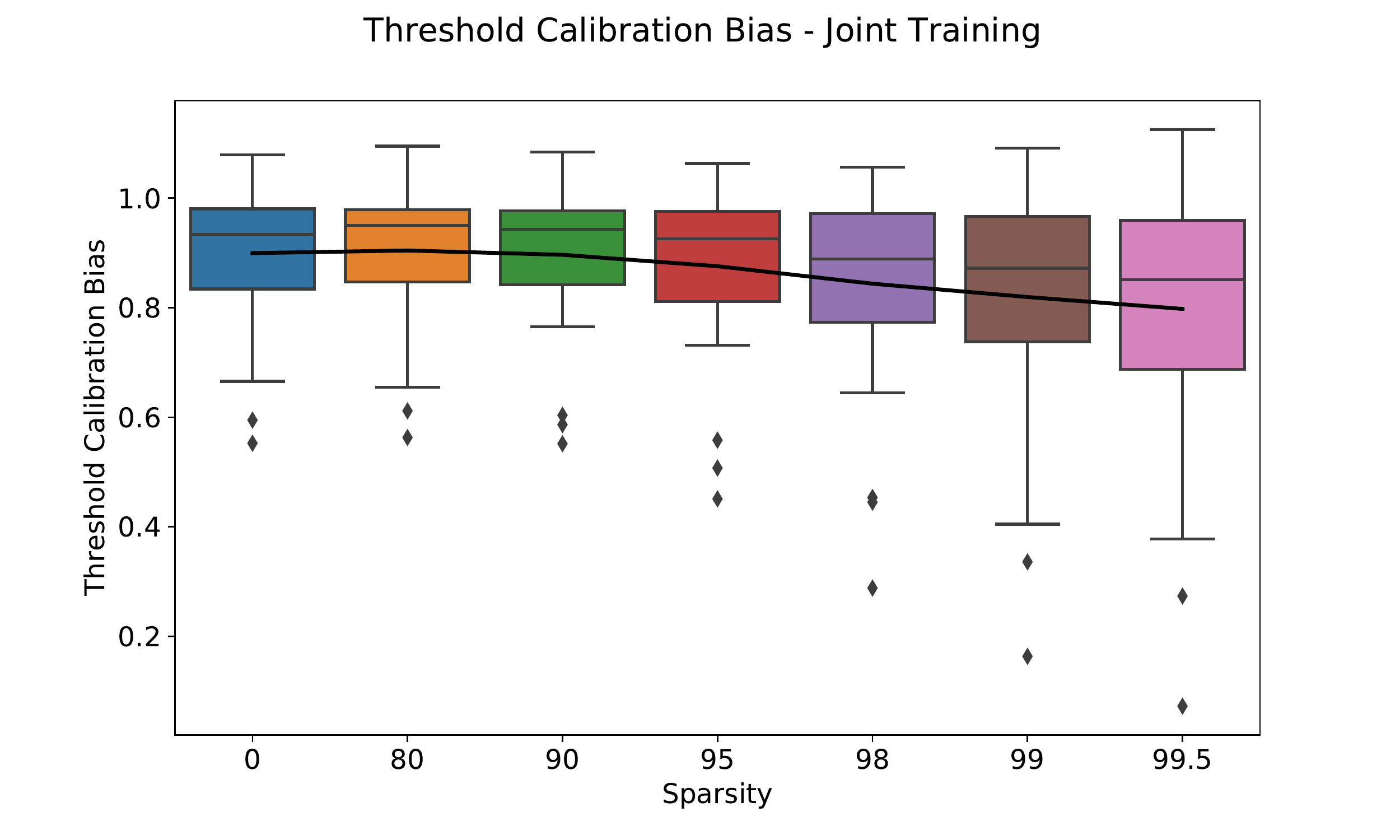}\\
    \includegraphics[width=0.22\textwidth]{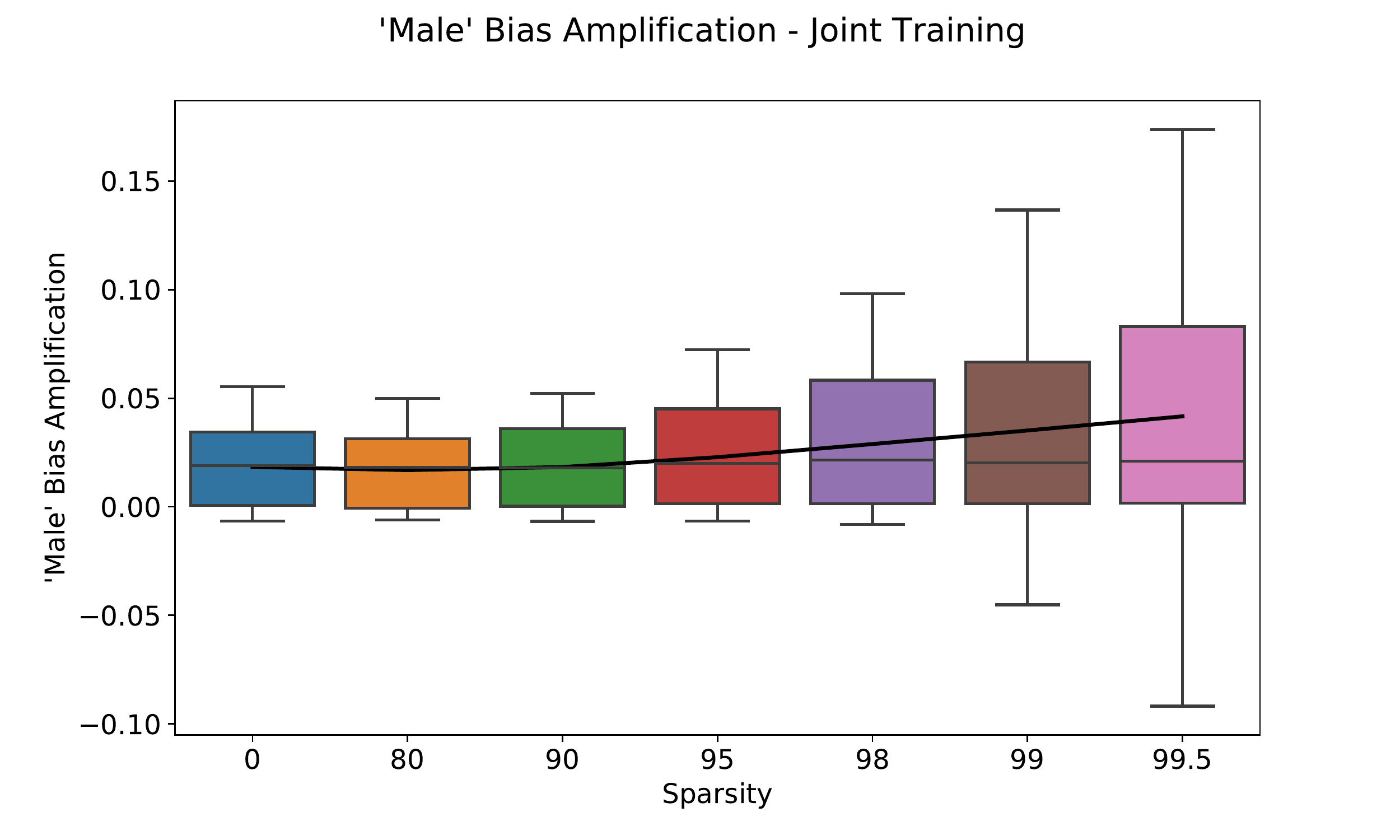} &
    \includegraphics[width=0.22\textwidth]{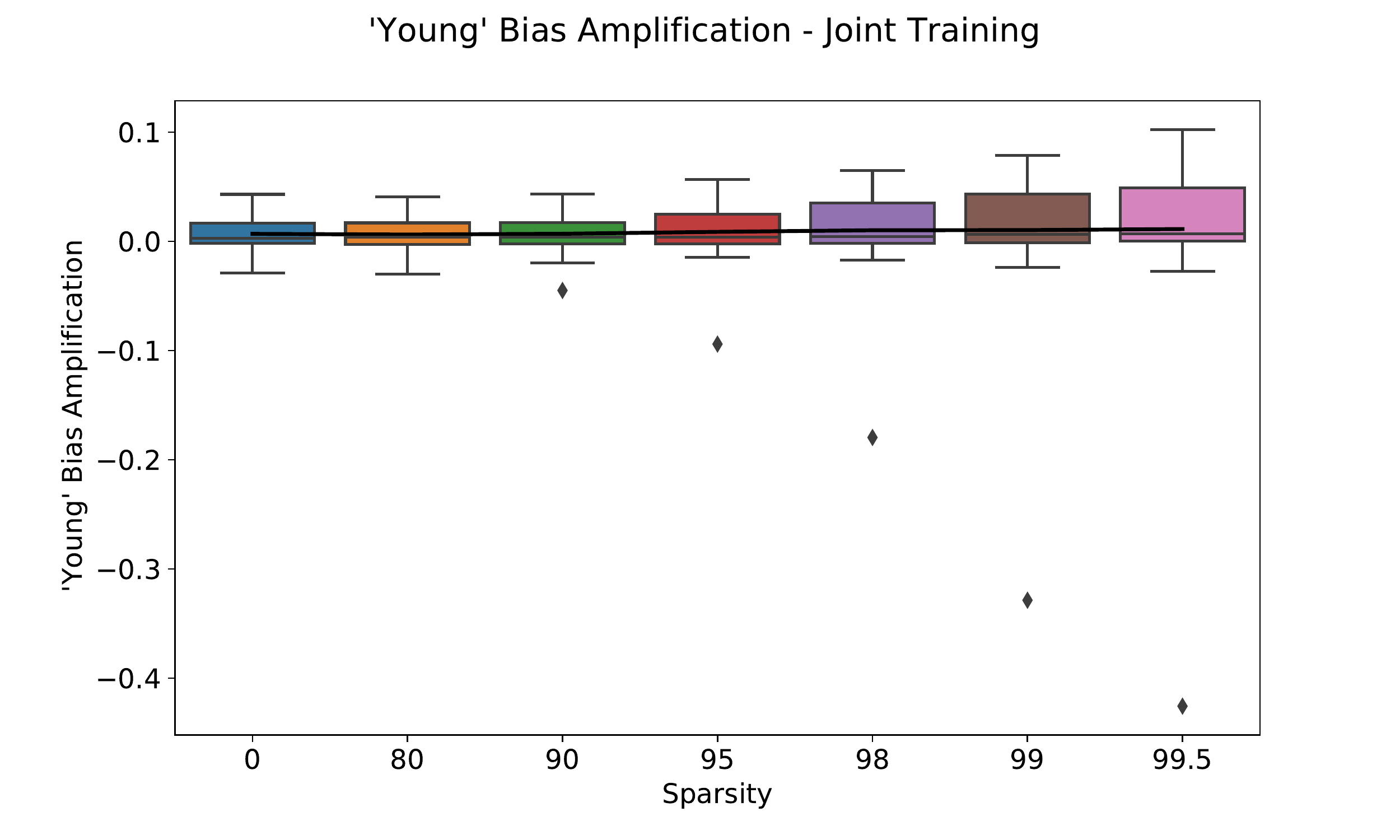} &
    \includegraphics[width=0.22\textwidth]{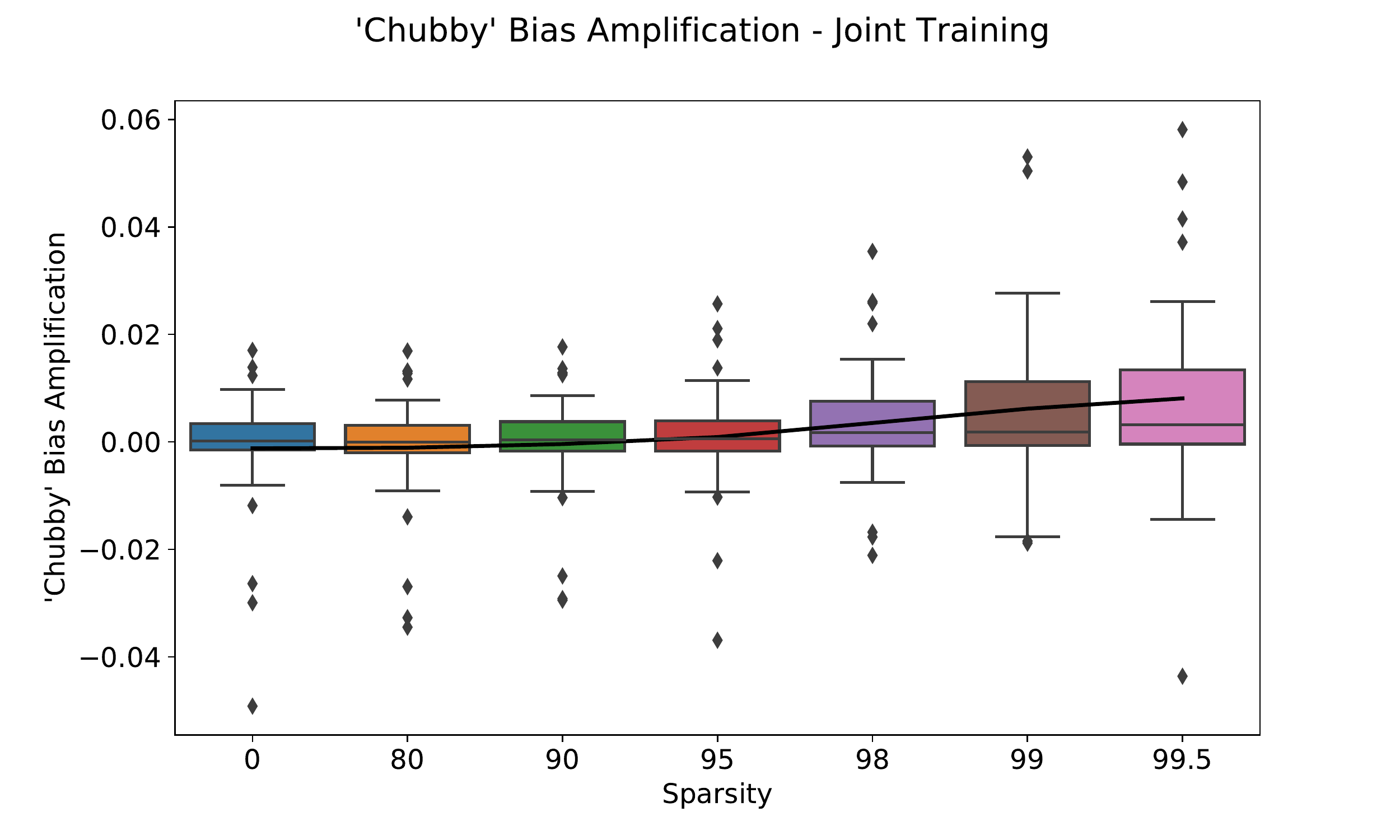} &
    \includegraphics[width=0.22\textwidth]{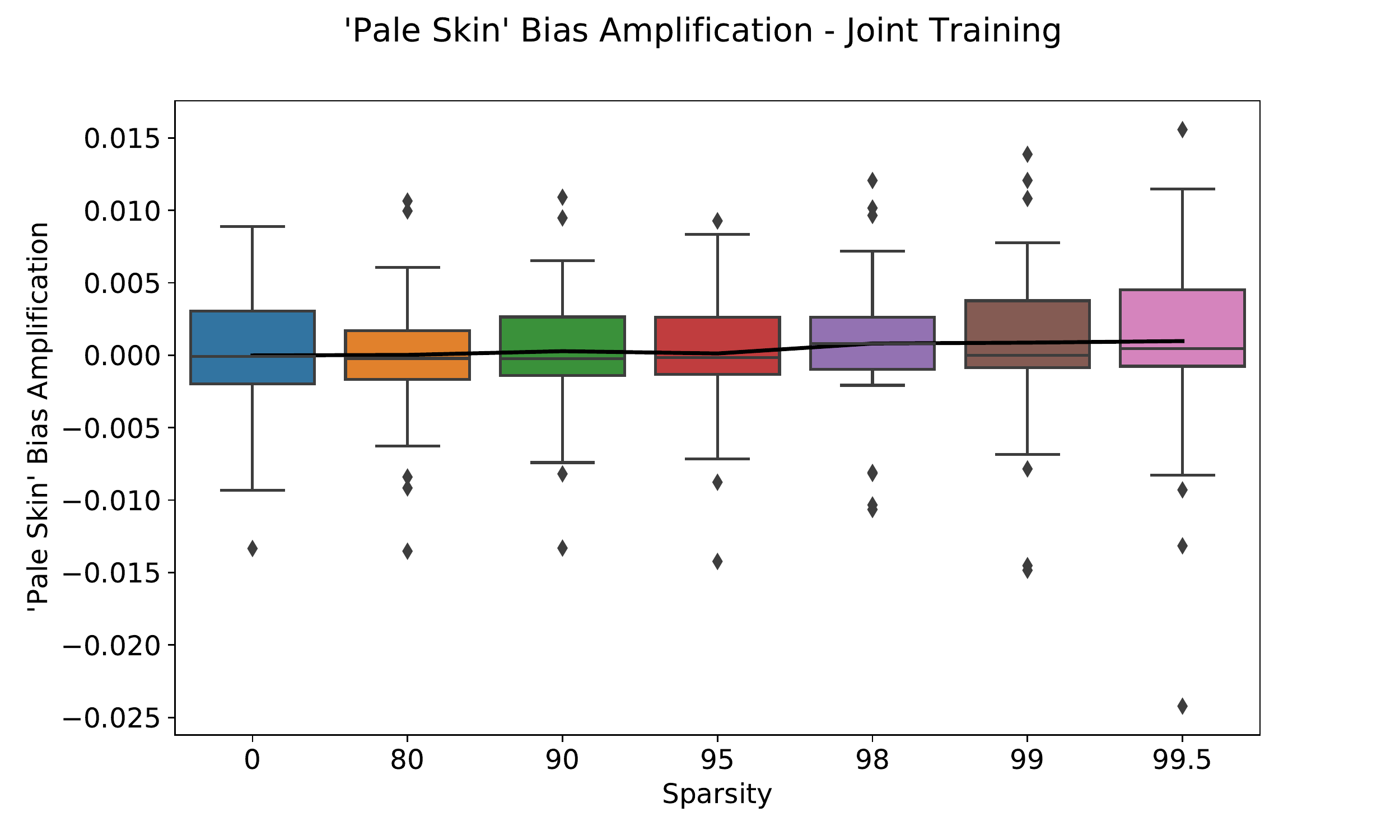}\\
\end{tabular}
    \caption{[CelebA / ResNet50 / GMP-RI] Accuracy and Systematic Bias metrics (TCB, ECE, Interdependence) of ResNet50 models jointly trained on all CelebA attributes. The thick black line denotes the mean value at each sparsity level.
    }
    \label{fig:celeba_rn50_joint_systematic}
\end{figure}

\begin{figure}[h]
\centering
\includegraphics[width=0.8\textwidth]{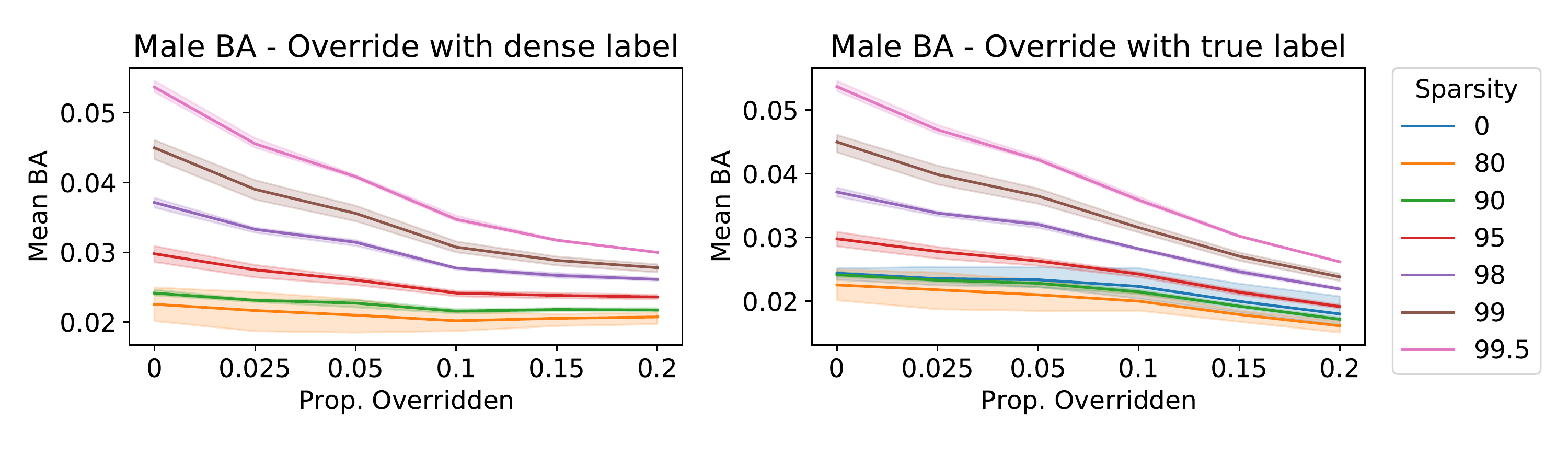}
\includegraphics[width=0.8\textwidth]{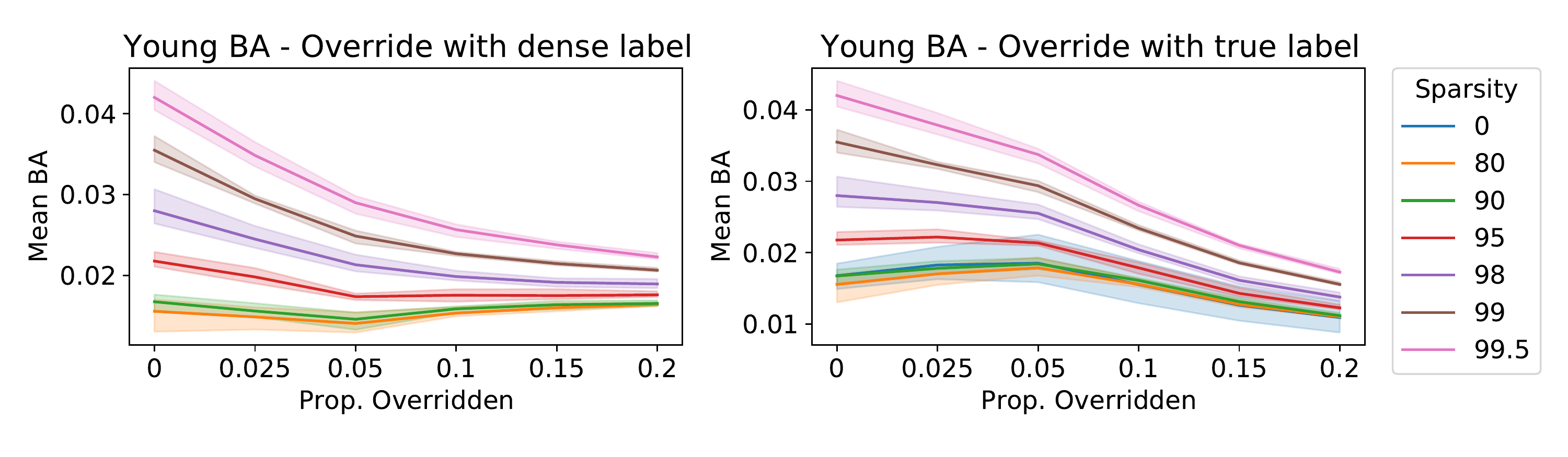}
\includegraphics[width=0.8\textwidth]{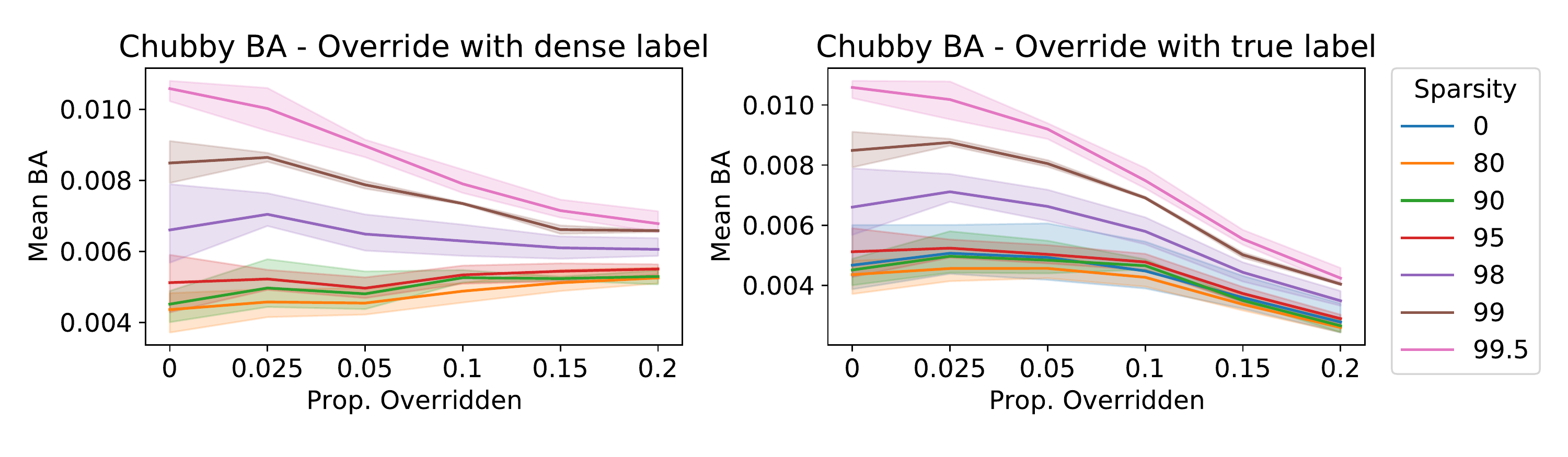}
\includegraphics[width=0.8\textwidth]{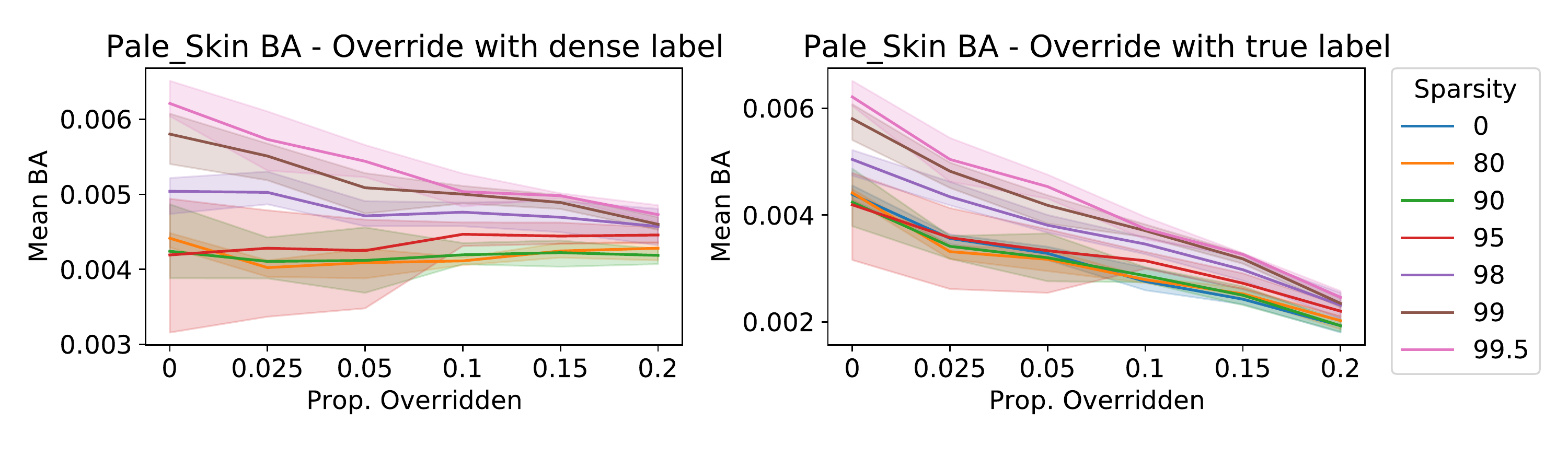}
    \caption{[CelebA / ResNet50 / GMP-RI] Effect of label overrides on Bias Amplification. In all cases, overrides are prioritized by dense model uncertainty.}
    \label{fig:overrides_rn50}
\end{figure}

The accuracy and systematic bias metrics are presented in Figure~\ref{fig:celeba_rn50_joint_systematic}. Overall, the patters we observe using the ResNet50 architecture very closely match those using ResNet18. Figure~\ref{fig:overrides_rn50} shows the impact on Bias Amplification of overriding the most uncertain predictions (closest to 0.5 probability as measured on a dense model) with either the dense prediction or the correct label. Consistent with the rest of the paper, the override is only applied if the Bias Amplification is positive on the dense model for the attribute and category in question. As in other cases, both types of overrides are effective at reducing Bias Amplification, generally when using the correct label, and when applied to high-sparsity models in the case of the dense label.

\begin{figure}[ht]
\centering
\begin{tabular}{cc}
  \includegraphics[width=0.35\textwidth]{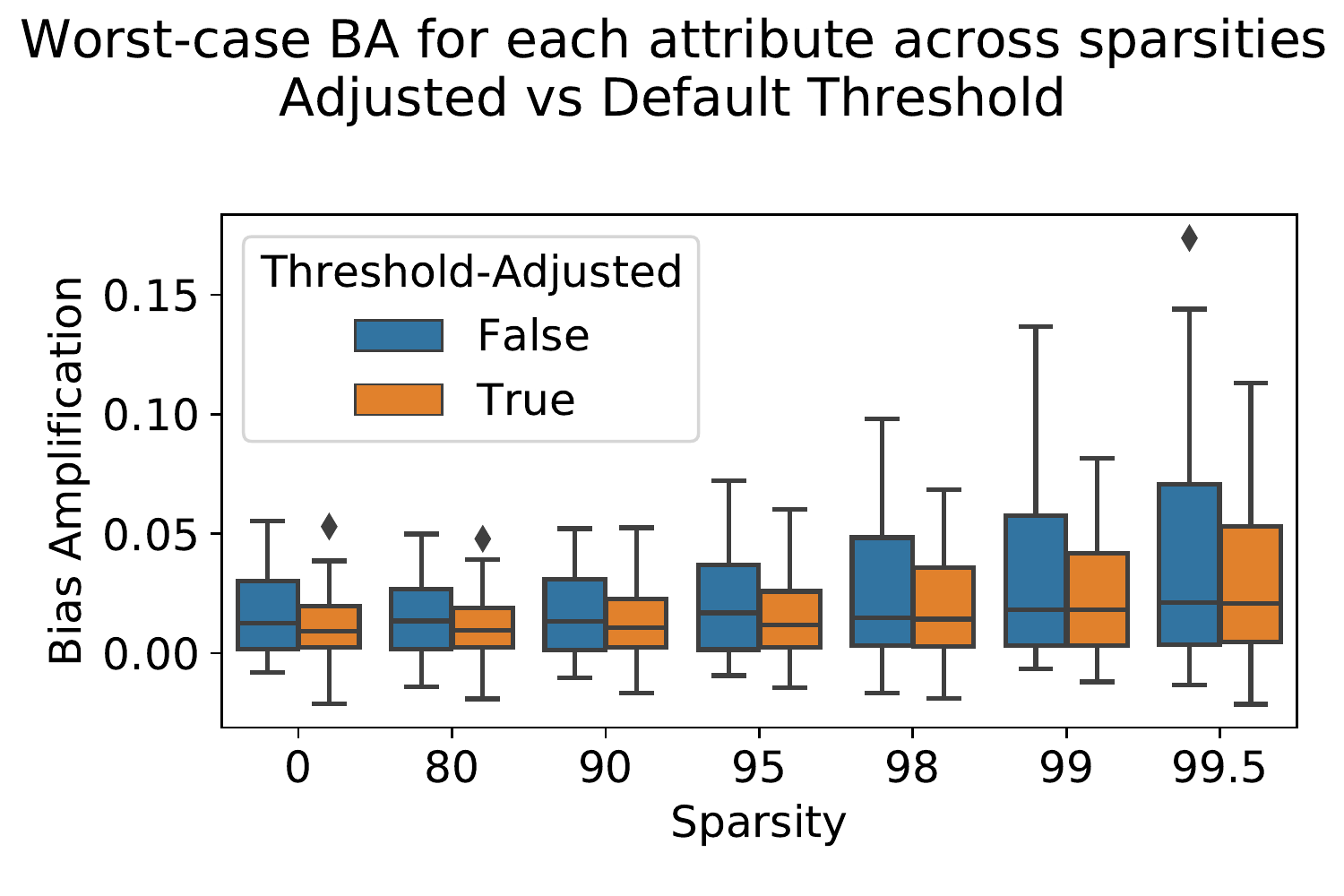} &
  \includegraphics[width=0.35\textwidth]{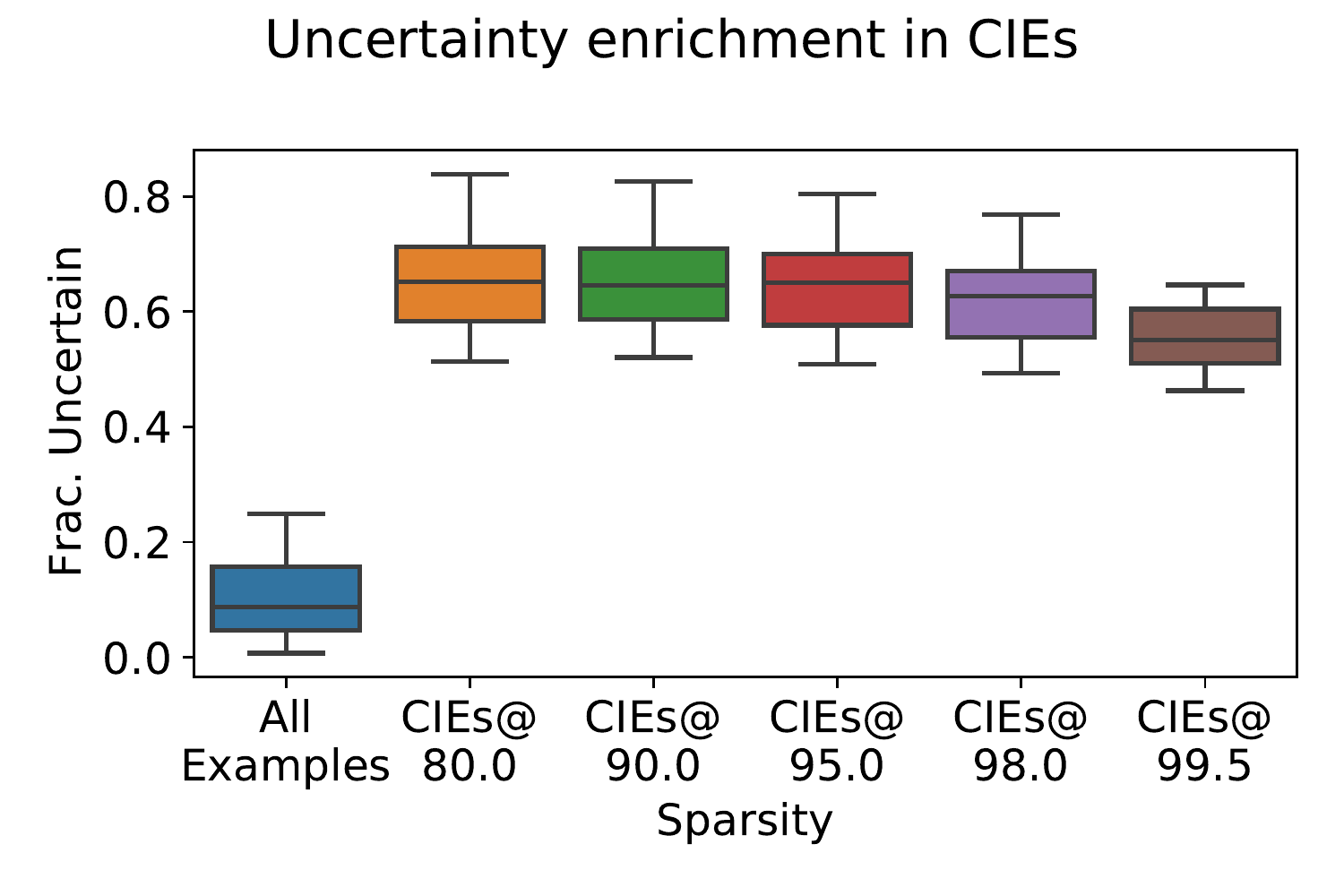}
  \end{tabular}
    \caption{[CelebA / ResNet50 / GMP-RI](Left) Effect of threshold calibration on ResNet50 models jointly trained on all attributes. %
    (Right) Proportion of uncertain predictions for \emph{dense} models across all attributes for all elements in the CelebA test set, and for Compression-Identified Exemplars at different sparsities.}
    \label{fig:celeba_rn50_threshold_adj}
\end{figure}

\clearpage
\section{Uncropped CelebA Results}
\label{appendix:uncropped_celeba}
\begin{figure}[ht]
\centering
\begin{tabular}{cc}
  \includegraphics[width=0.35\textwidth]{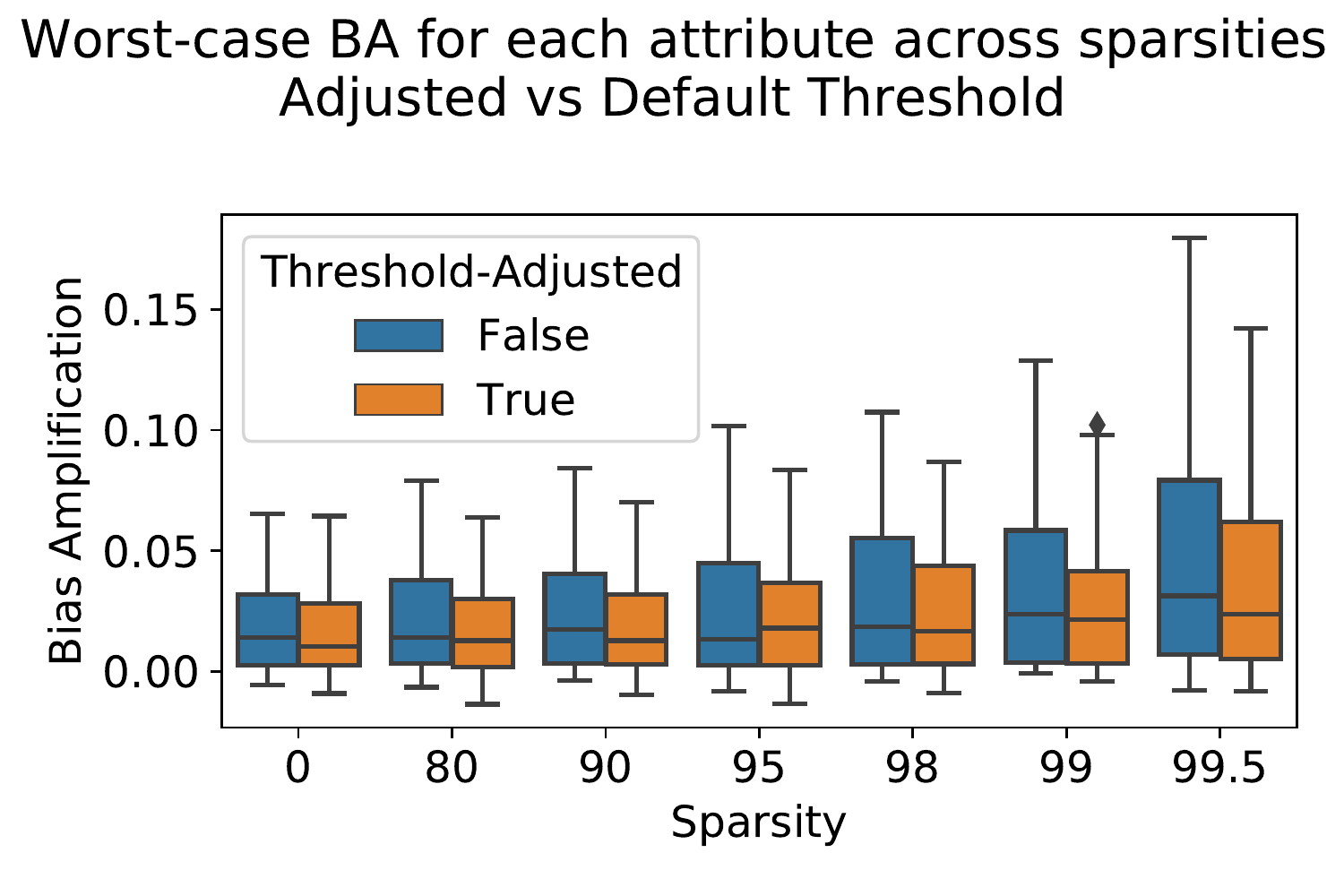} &
  \includegraphics[width=0.35\textwidth]{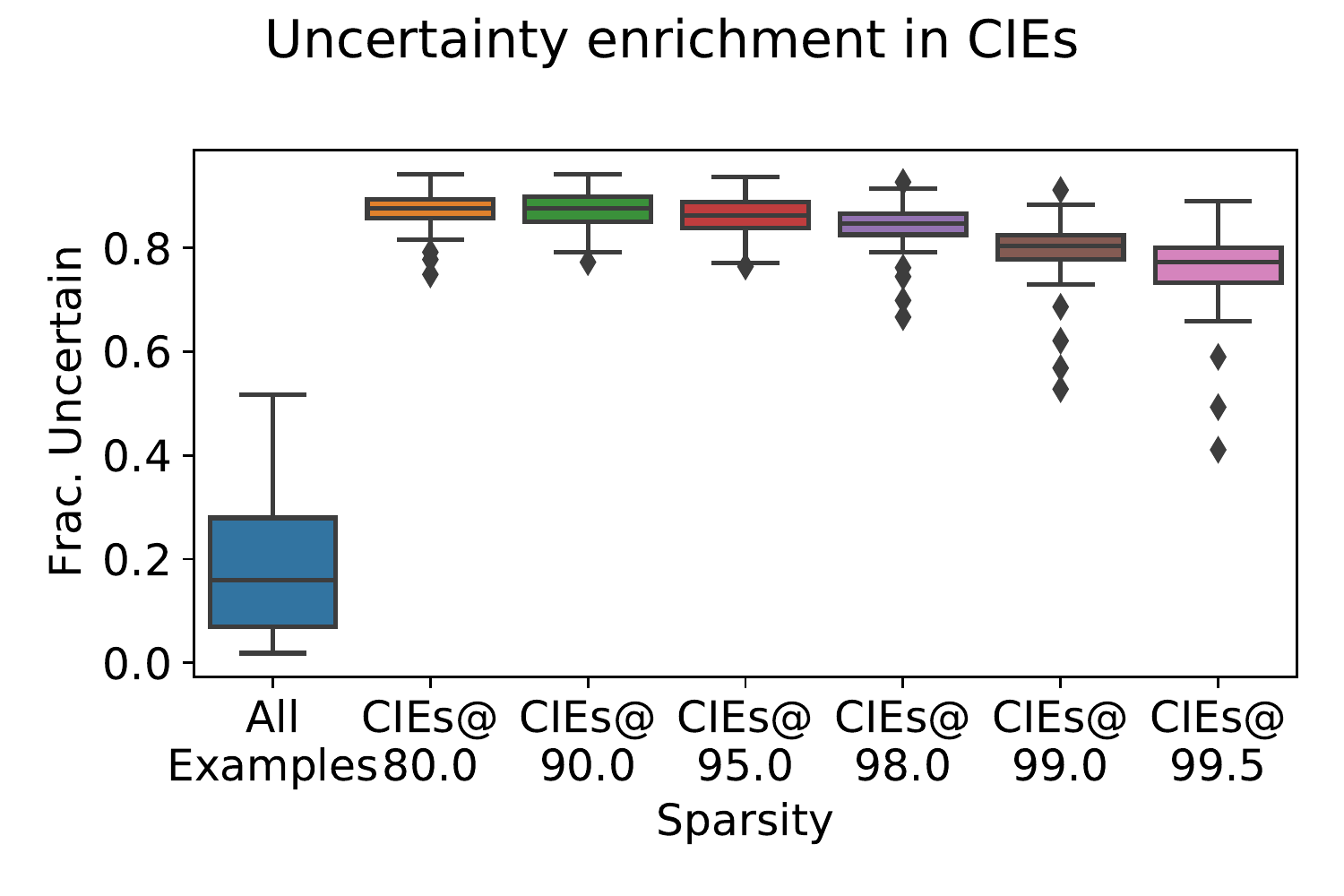}
  \end{tabular}
    \caption{[Uncropped CelebA / ResNet18 / GMP-RI] (Left) Effect of threshold calibration on models jointly trained on all attributes. %
    (Right) Proportion of uncertain predictions for \emph{dense} models across all attributes for all elements in the CelebA test set, and for Compression-Identified Exemplars at different sparsities.}
    \label{fig:uncropped_celeba_rn18_threshold_adj}
\end{figure}

\begin{figure}[h]
\centering
\includegraphics[width=0.8\textwidth]{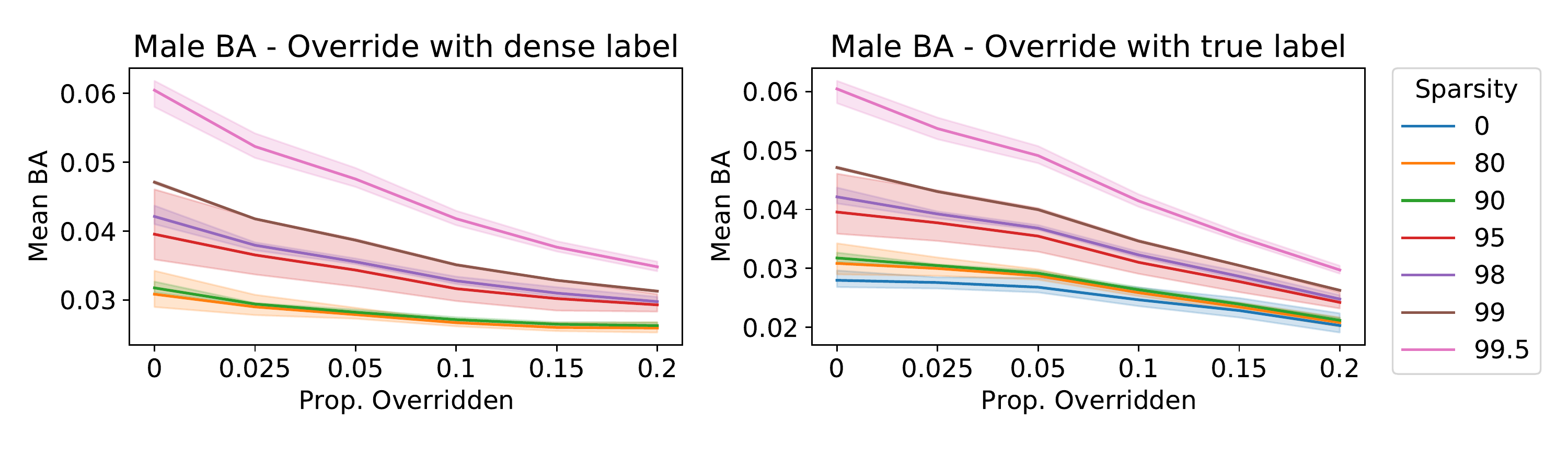}
\includegraphics[width=0.8\textwidth]{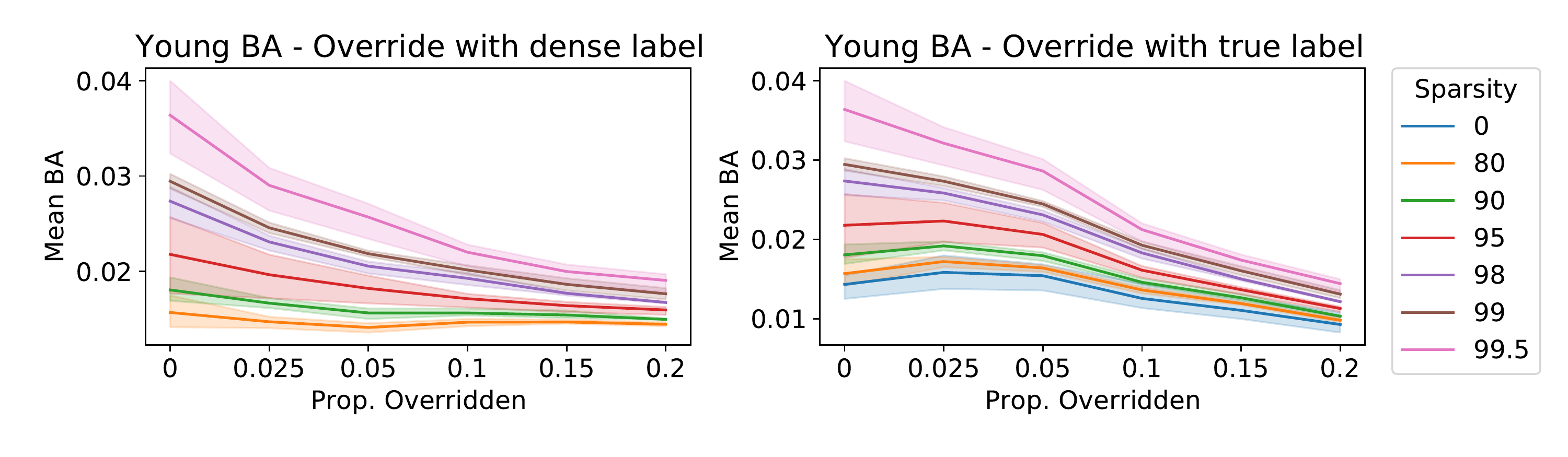}
\includegraphics[width=0.8\textwidth]{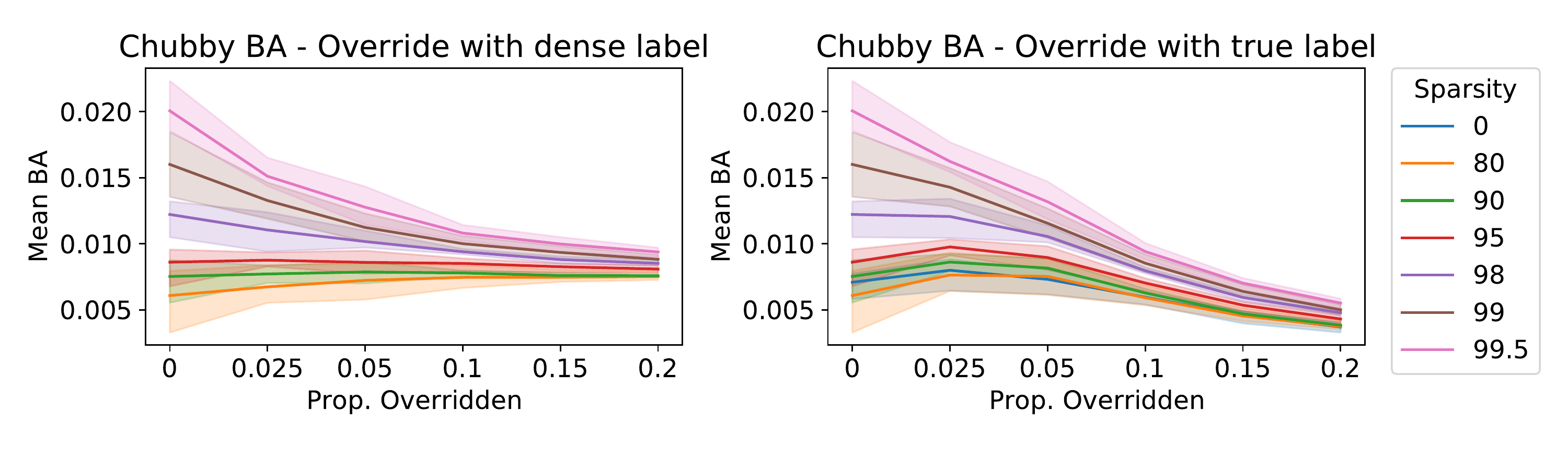}
\includegraphics[width=0.8\textwidth]{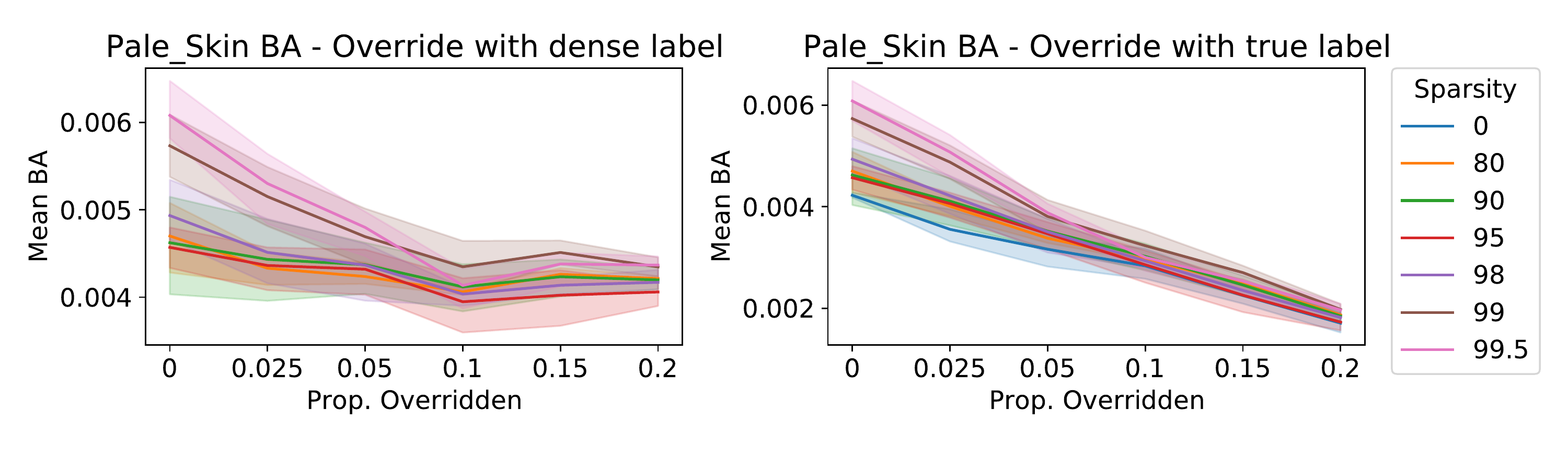}
    \caption{[Uncropped CelebA / ResNet18/ GMP-RI] Effect of label overrides on Bias Amplification. In all cases, overrides are prioritized by dense model uncertainty.}
    \label{fig:overrides_uncropped}
\end{figure}

While inspecting the CelebA samples using our visualization tool described in Appendix Section~\ref{appendix:ui_tool} we observed that some of the attributes were more prone to mislabelling, due to decisions conventionally made when training models on CelebA; for example, due to the cropping of the images in the standard CelebA version used in practice, it is often times impossible to directly observe the presence of attributes like Wearing Necktie or Wearing Necklace (see the discussion in \ref{appendix:ui_tool}, and specifically Figures \ref{fig:viwer-necklace}, \ref{fig:viwer-necktie}). In an effort to disentangle the data inherent bias, due to cropping, from Systematic or Categorical bias, we further validate our results on dense and sparse models trained on the \emph{uncropped} version of CelebA. We use the same setting for training ResNet18 GMP-RI models, as the one described in Appendix Section~\ref{appendix:training_settings}. In terms of accuracy or AUC scores, we observe a decrease in performance for very sparse ($99.5\%$ sparse) models trained on the uncropped CelebA. Otherwise, our findings in terms of systematic (ECE, TCB, Interdependence) or context (BA) bias generally confirm those on the standard CelebA dataset. It is worth noting, however, that using the uncropped CelebA version substantially reduced the Categorical bias for the problematic attributes Wearing Necklace or Wearing Necktie. For example, the BA scores for the dense model changed from 4.6 to 0.9 for Wearing Necktie and from -2.2 to -1.4 for Wearing Necklace. More importantly, the bias decreased substantially for high sparse models; for example, the interval for the BA scores for models in the 98\%-99.5\% sparsity range changed from [-34.4, -21.3] for the cropped version to [-5.8, -3.4] for uncropped, for the Wearing Necklace attribute. Similarly, the BA score for Wearing Necktie on the 99.5\% sparse model dropped from 8.7 to 3.1, and also decreased substantially for lower sparsity levels. These findings confirm our expectations that data inherent bias can play a significant role in the overall bias equation for a model, and improvements can be obtained by carefully taking the data bias into account. We further show that Categorical bias can be decreased by careful relabelling in Figure~\ref{fig:overrides_uncropped} and show the uncertainty of CIEs in Figure~\ref{fig:uncropped_celeba_rn18_threshold_adj}.

\begin{figure}[h]
    \centering
\begin{tabular}{cccc}
   \includegraphics[width=0.22\textwidth]{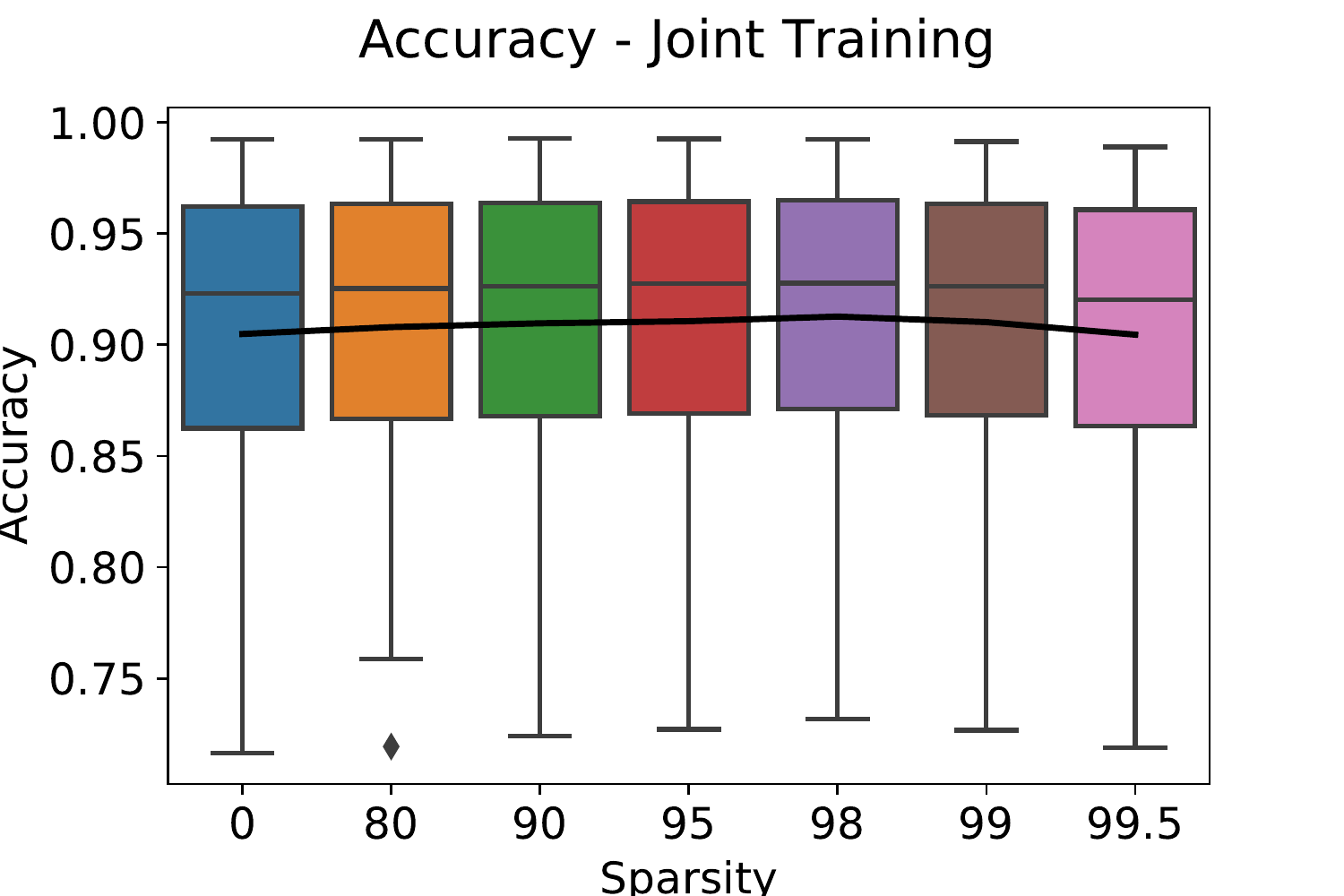} &
   \includegraphics[width=0.22\textwidth]{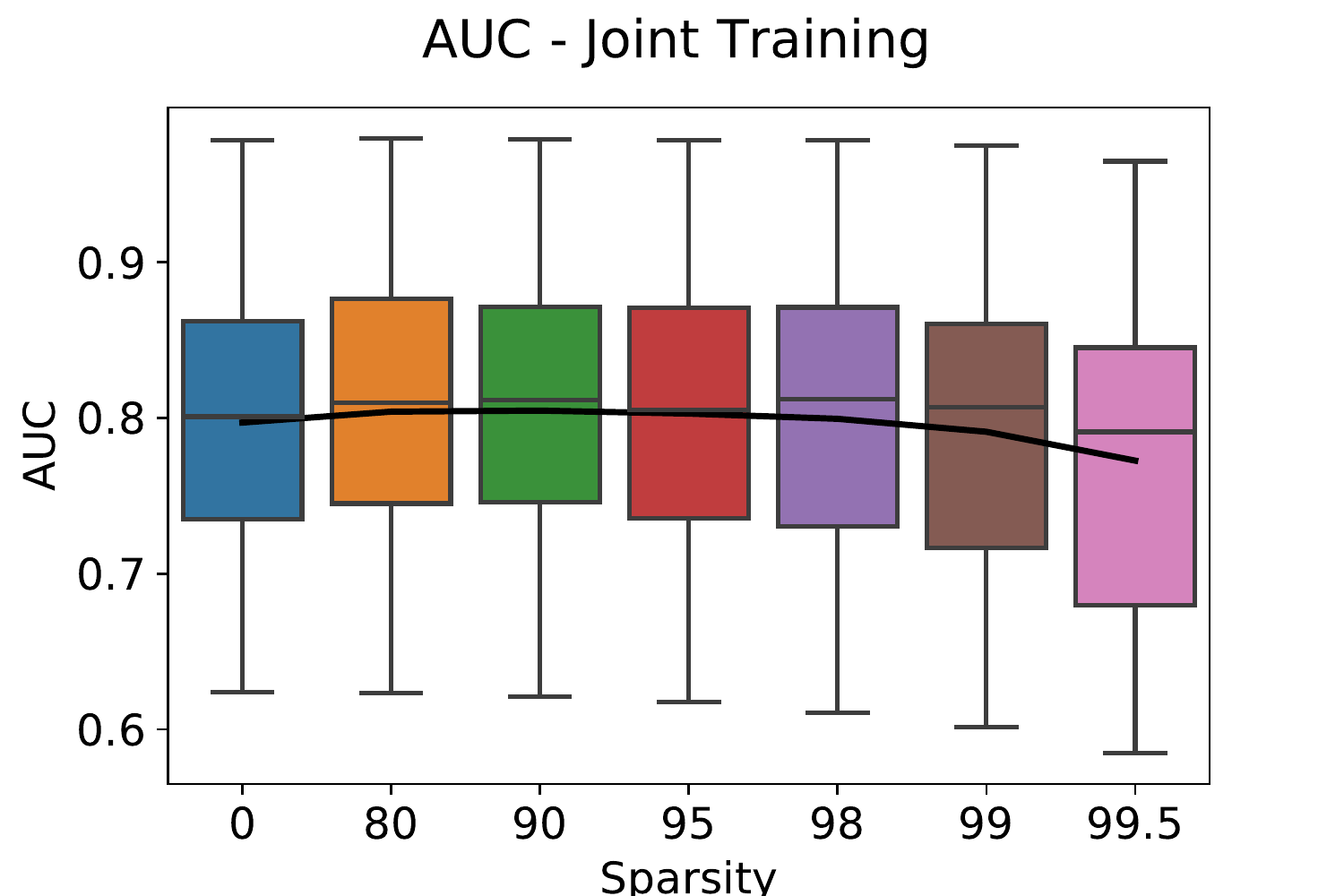} \\
    \includegraphics[width=0.22\textwidth]{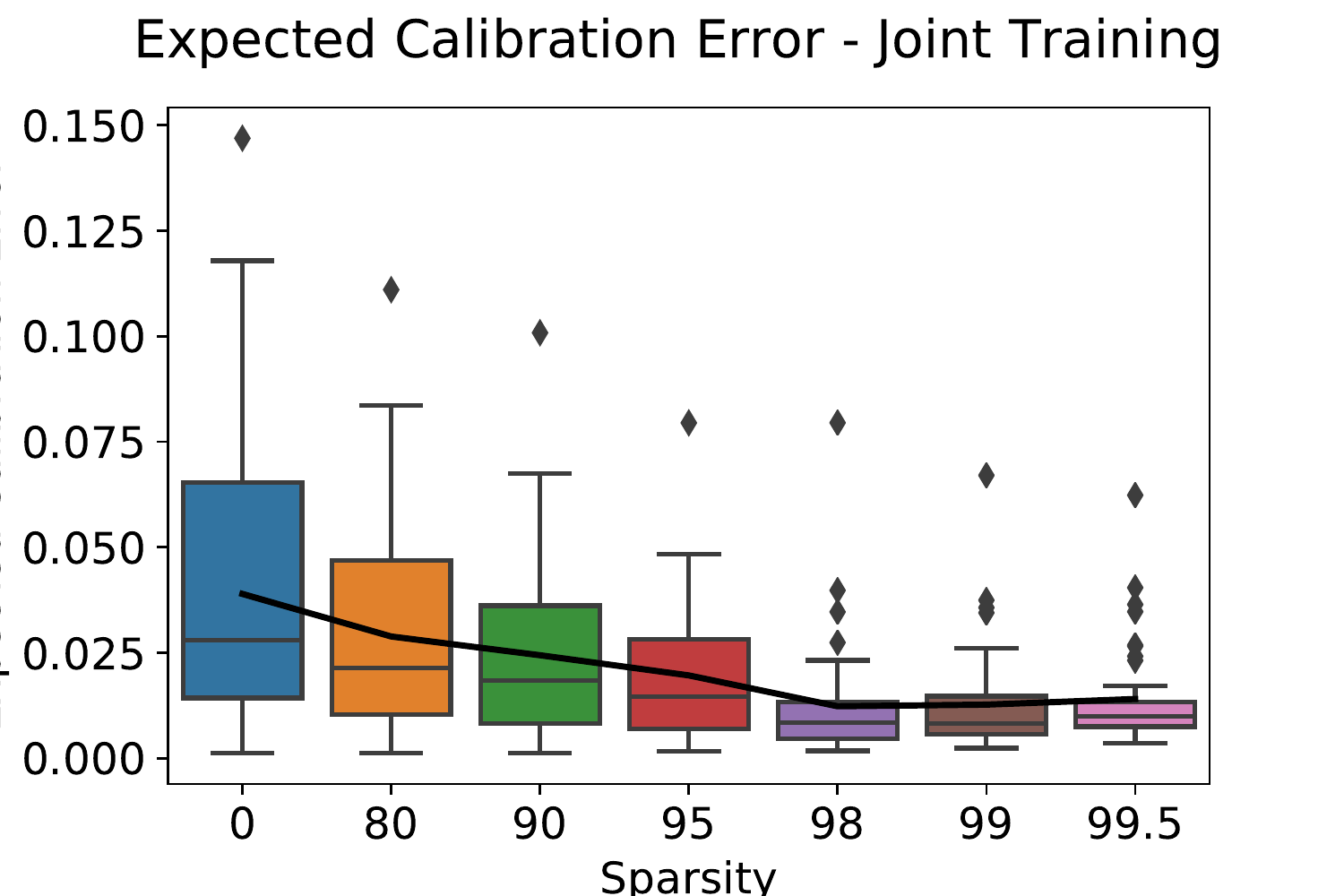} &
    \includegraphics[width=0.22\textwidth]{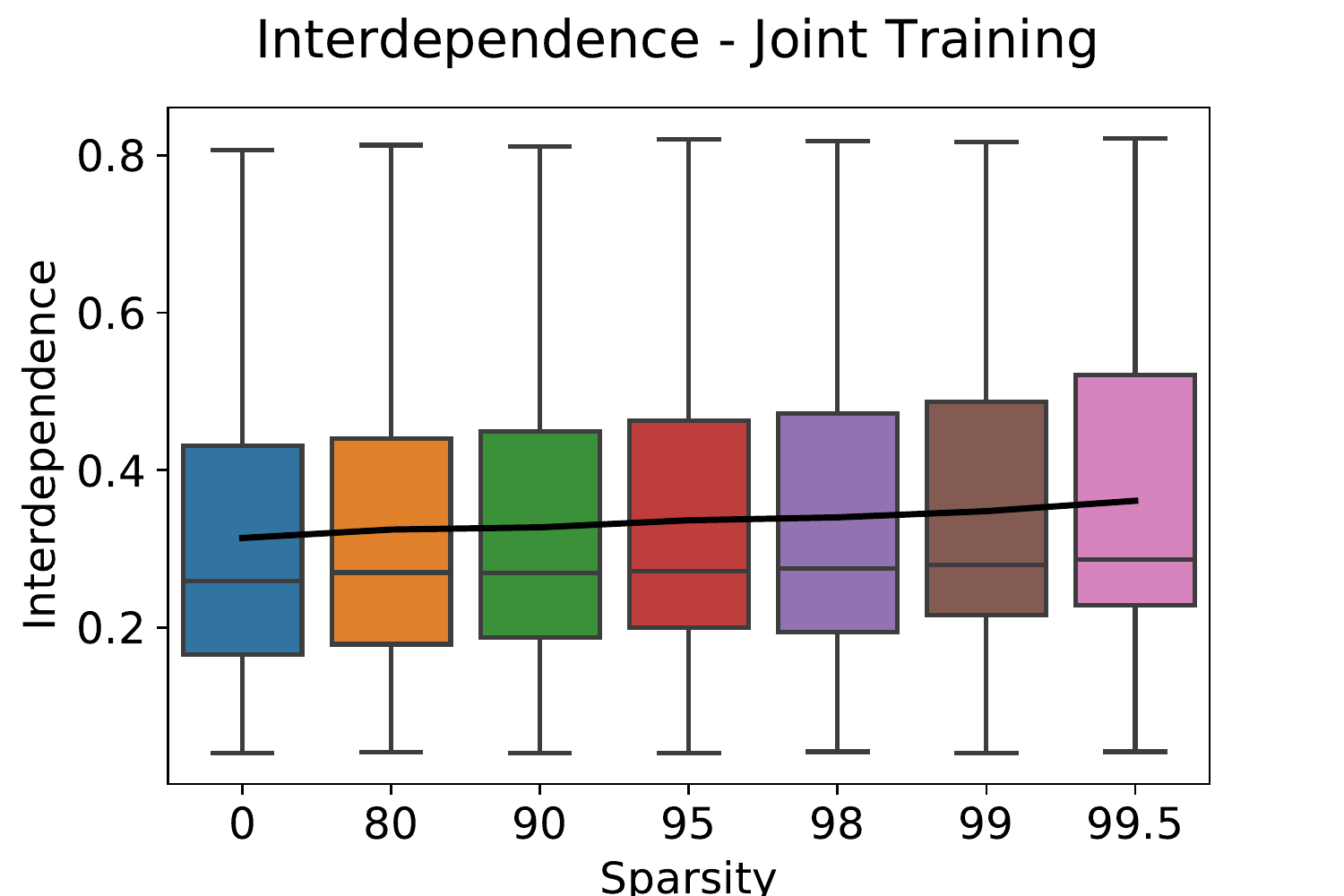} &
        \includegraphics[width=0.22\textwidth]{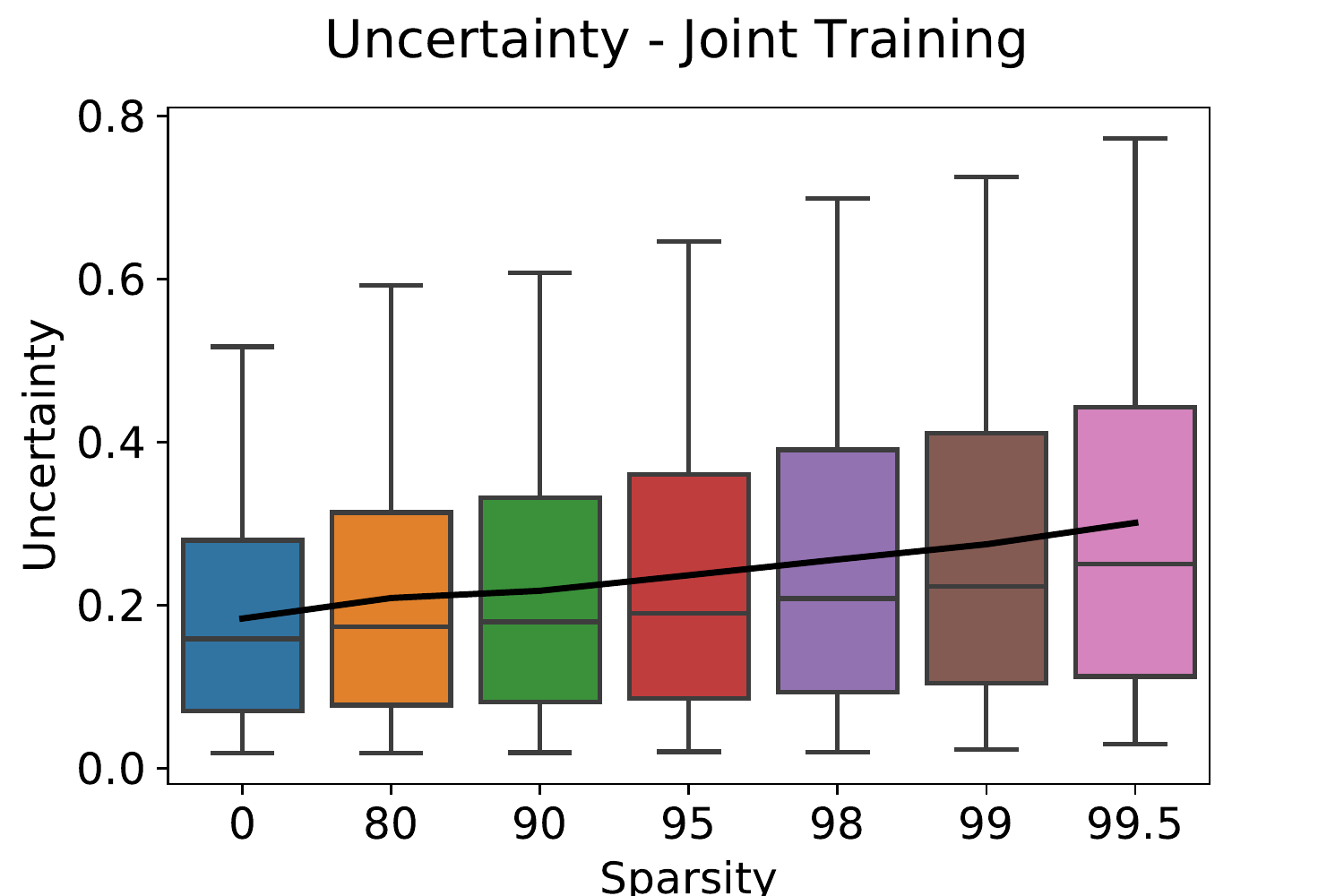} &
    \includegraphics[width=0.22\textwidth]{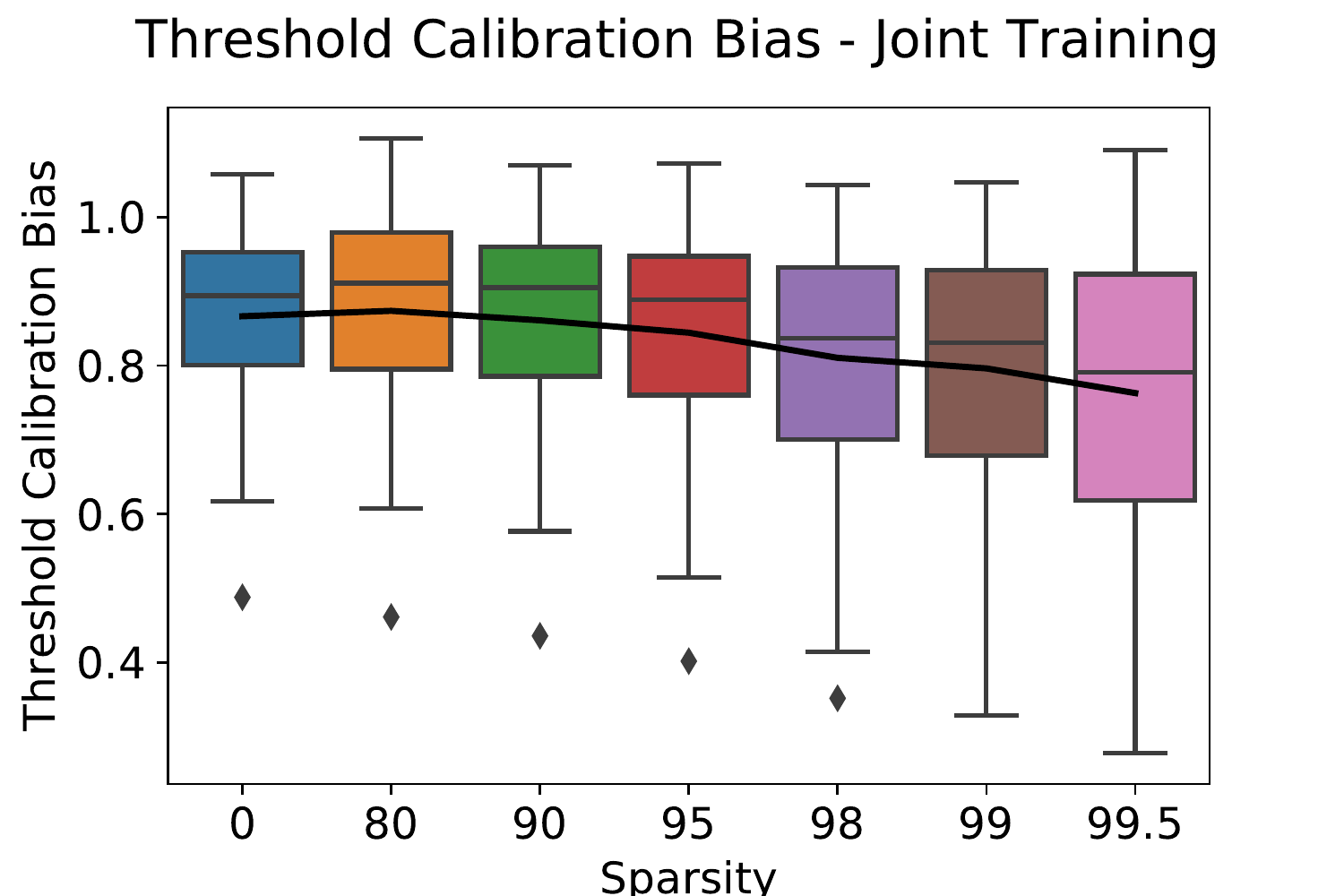}\\
    \includegraphics[width=0.22\textwidth]{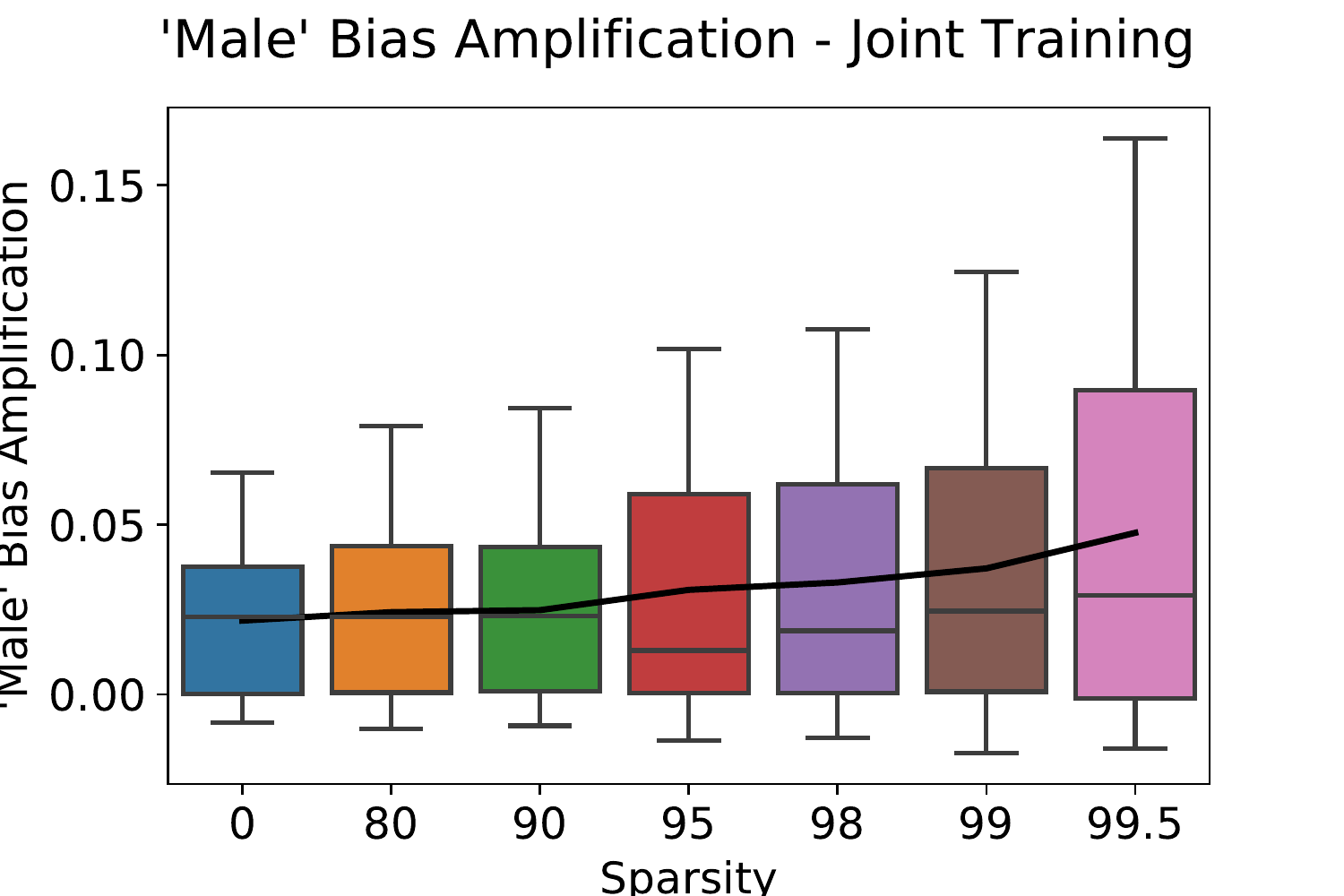} &
    \includegraphics[width=0.22\textwidth]{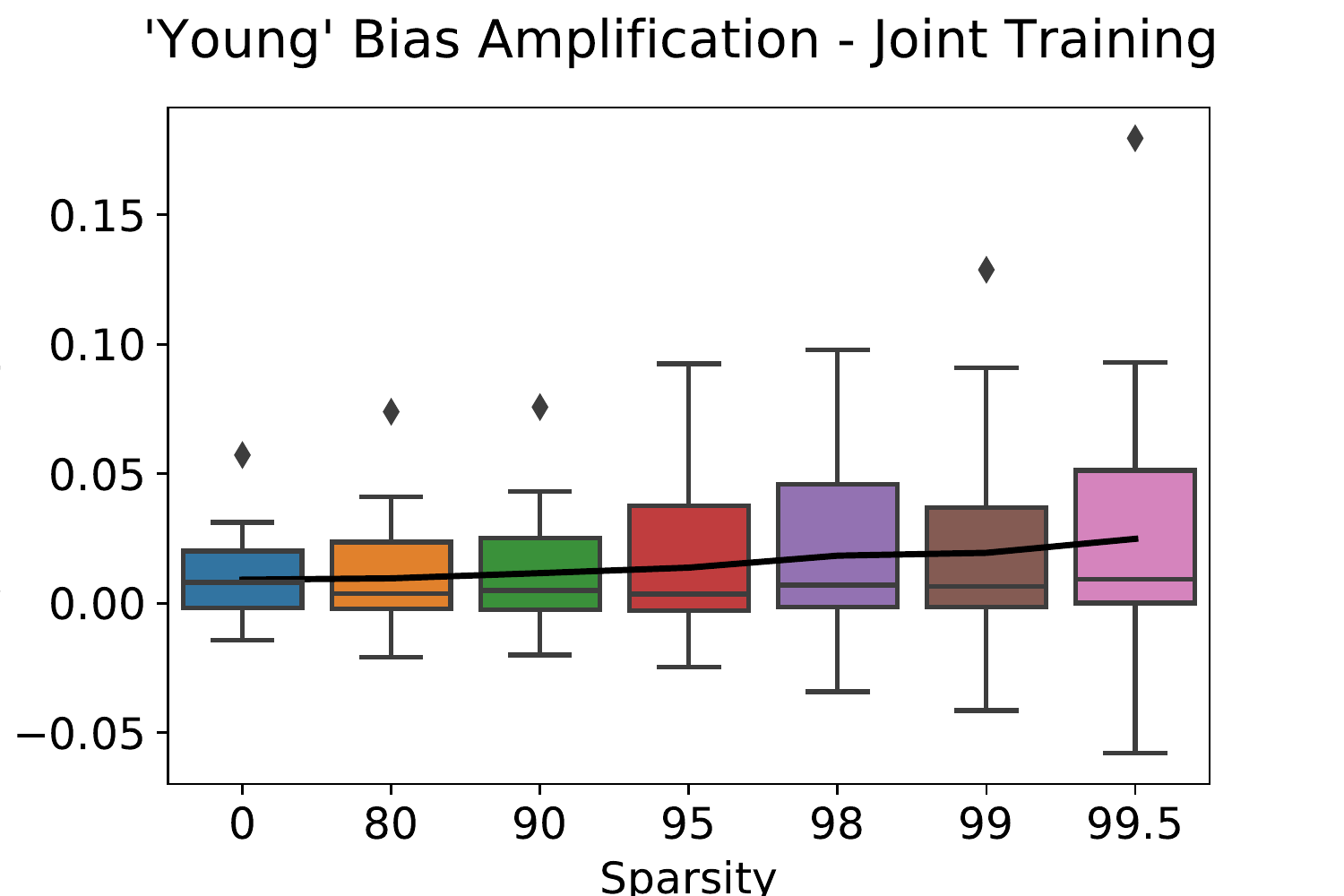} &
        \includegraphics[width=0.22\textwidth]{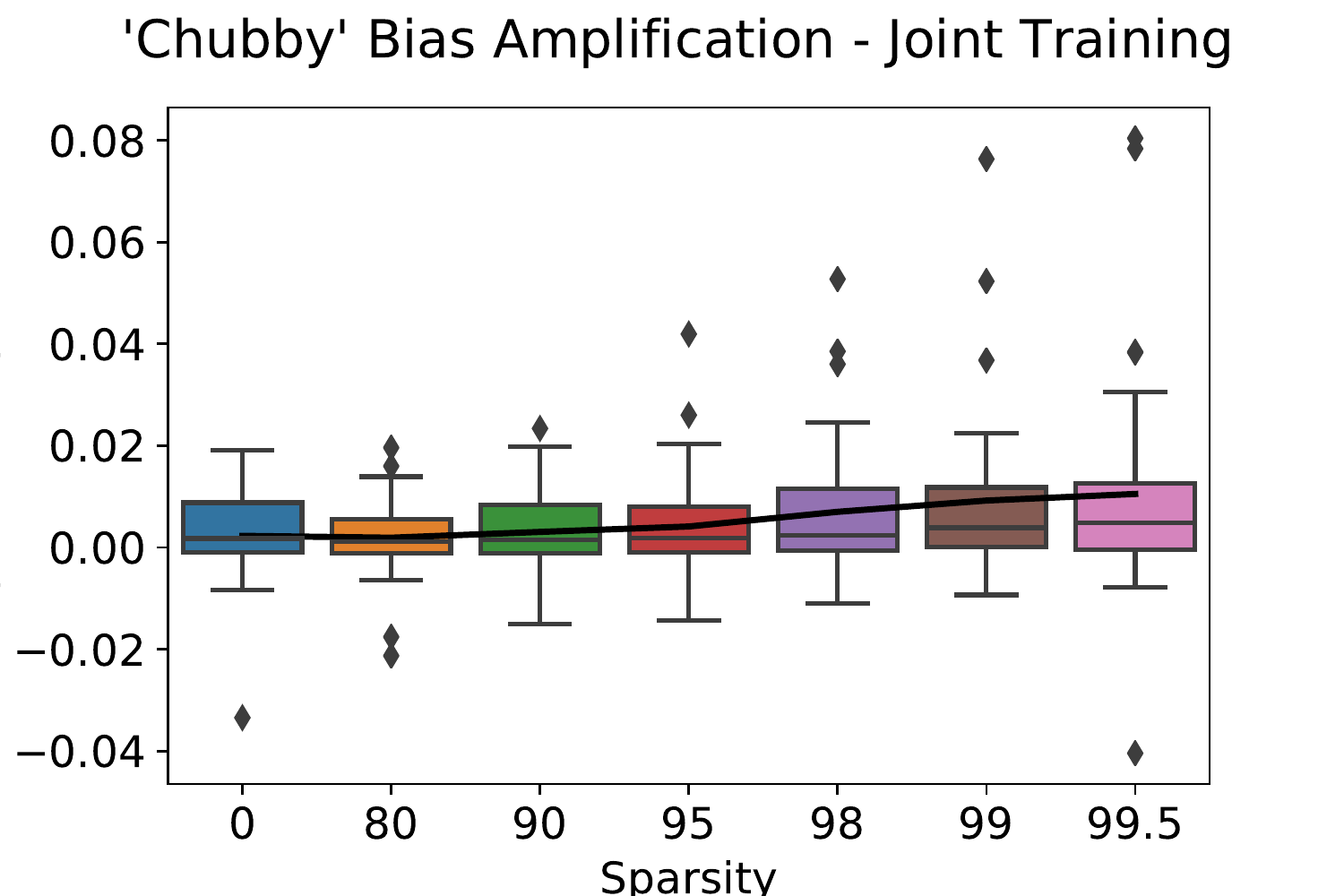} &
    \includegraphics[width=0.22\textwidth]{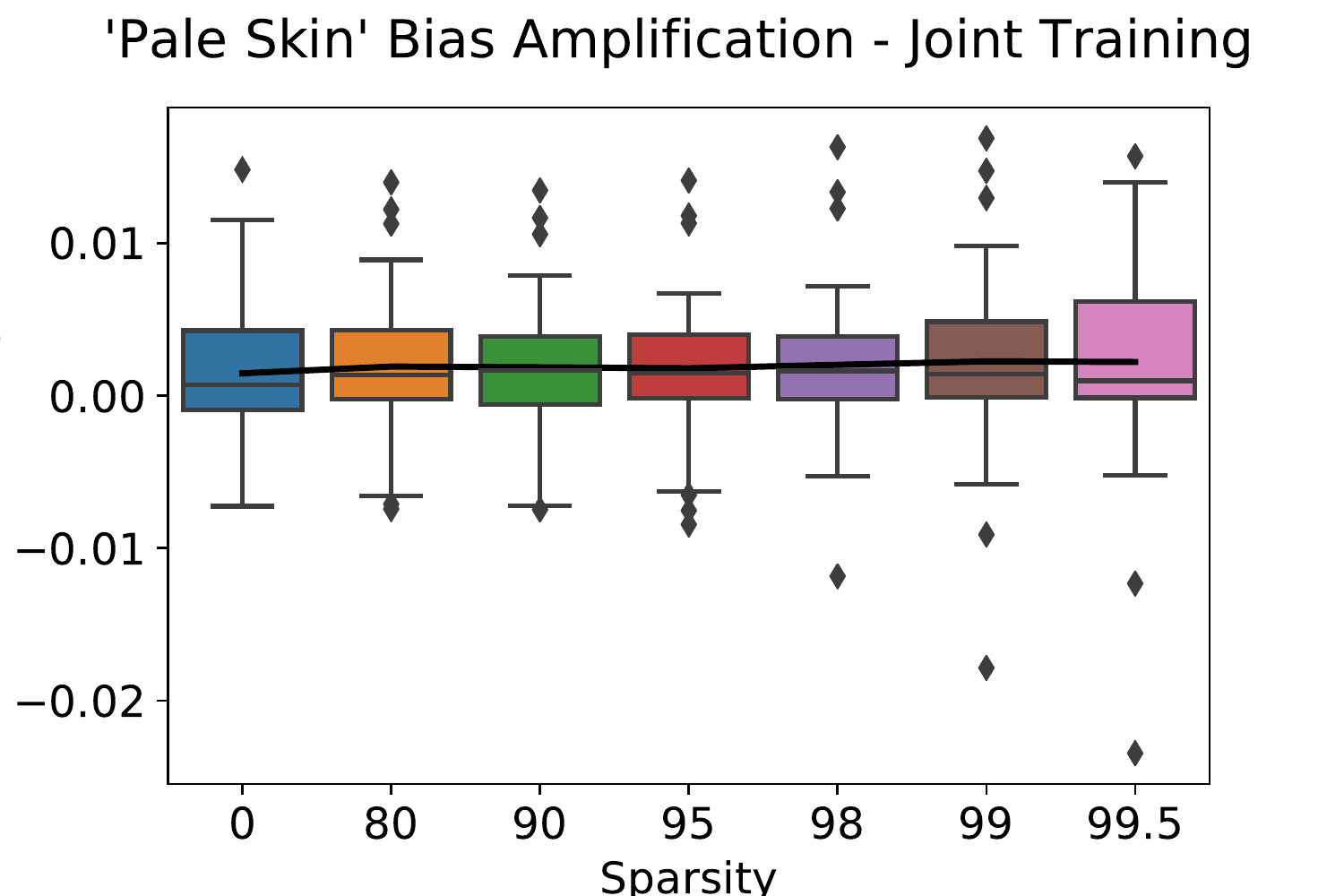}\\
\end{tabular}
    \caption{[Uncropped CelebA / ResNet18 / GMP-RI] Accuracy and Systematic Bias metrics (TCB, ECE, Interdependence) of ResNet18 models jointly trained on all CelebA attributes, using the \emph{uncropped} images for training and inference. The thick black line denotes the mean value at each sparsity level.
    }
    \label{fig:uncropped_celeba_rn18_joint_systematic}
\end{figure}

\clearpage
\section{Tabular Results for Jointly-Trained ResNet18 CelebA Models}
\label{appendix:tabular}
In this section, we present our main results for systematic and categorical bias metrics for ResNet18 CelebA models in tabular form. We first present the average values across attributes for all metrics by sparsity in Table~\ref{table:mean-celeba-rn18-joint}, then give detailed per-attribute numbers for each metric in subsequent tables. The means and standard deviations were computed from runs from five random seeds.

\begin{table}[h]
\centering
\caption{Mean Accuracy, Systematic Bias, and Categorical Bias Values, Joint CelebA Training, ResNet18}
\label{table:mean-celeba-rn18-joint}
\begin{tabular}{l|rrrrrrr}
\toprule
Sparsity &         0 &        80 &        90 &        95 &        98 &        99 &      99.5 \\
Metric                         &           &           &           &           &           &           &           \\
\midrule
Accuracy                       &  0.904 &  0.908 &  0.909700 &  0.913 &  0.915 &  0.914 &  0.911 \\
AUC                            &  0.805 &  0.810 &  0.813 &  0.815 &  0.815 &  0.810 &  0.797 \\
\midrule
Expected Calibration Error     &  0.0538 &  0.0401 &  0.0341 &  0.0254 &  0.0153 &  0.0128 &  0.0127 \\
Interdependence                &  0.310 &  0.319 &  0.324 &  0.332 &  0.341 &  0.349 &  0.361 \\
Threshold Calibration Bias     &  0.903 &  0.895 &  0.889 &  0.877 &  0.853 &  0.833 &  0.805 \\
Uncertainty                    &  0.139 &  0.172 &  0.186 &  0.207 &  0.237 &  0.256 &  0.276 \\
\midrule
'Male' Bias Amplification      &  0.0170 &  0.0180 &  0.0210 &  0.0241 &  0.0294 &  0.0337 &  0.0402 \\
'Young' Bias Amplification     &  0.00600 &  0.00663 &  0.00711 &  0.00851 &  0.00817 &  0.0101 &  0.0148 \\
'Chubby' Bias Amplification    & -0.00208 & -0.00133 & -0.000278 &  0.00106 &  0.00269 &  0.00583 &  0.00844 \\
'Pale Skin' Bias Amplification &  0.000097 & -0.000065 &  0.000323 &  0.000419 &  0.000581 &  0.000645 &  0.000935 \\
\bottomrule
\end{tabular}
\end{table}

\begin{table}[h]
\caption{Accuracy, Joint CelebA training, ResNet18 }
\label{table:acc-CelebA-ResNet18-joint}
\begin{tabular}{l|ccccccc}
\toprule
{Sparsity} & {0} & {80} & {90} & {95} & {98} & {99} & {99.5} \\
{Attribute} & {} & {} & {} & {} & {} & {} & {} \\
\midrule
5 o Clock Shadow & 94.3 $ \pm $ 0.1 & 94.5 $ \pm $ 0.0 & 94.6 $ \pm $ 0.1 & 94.8 $ \pm $ 0.1 & 95.0 $ \pm $ 0.0 & 94.9 $ \pm $ 0.1 & 94.7 $ \pm $ 0.1 \\
Arched Eyebrows & 82.3 $ \pm $ 0.1 & 82.9 $ \pm $ 0.1 & 83.2 $ \pm $ 0.2 & 83.7 $ \pm $ 0.2 & 84.1 $ \pm $ 0.1 & 84.0 $ \pm $ 0.1 & 83.2 $ \pm $ 0.1 \\
Attractive & 80.5 $ \pm $ 0.2 & 81.6 $ \pm $ 0.1 & 81.8 $ \pm $ 0.3 & 82.6 $ \pm $ 0.1 & 83.0 $ \pm $ 0.1 & 83.1 $ \pm $ 0.2 & 82.7 $ \pm $ 0.1 \\
Bags Under Eyes & 83.2 $ \pm $ 0.1 & 83.9 $ \pm $ 0.2 & 84.2 $ \pm $ 0.1 & 84.8 $ \pm $ 0.1 & 85.4 $ \pm $ 0.2 & 85.5 $ \pm $ 0.1 & 85.0 $ \pm $ 0.2 \\
Bald & 98.9 $ \pm $ 0.0 & 98.9 $ \pm $ 0.0 & 98.9 $ \pm $ 0.0 & 99.0 $ \pm $ 0.1 & 99.0 $ \pm $ 0.0 & 99.0 $ \pm $ 0.0 & 98.8 $ \pm $ 0.0 \\
Bangs & 95.5 $ \pm $ 0.1 & 95.8 $ \pm $ 0.1 & 95.9 $ \pm $ 0.0 & 96.1 $ \pm $ 0.1 & 96.2 $ \pm $ 0.0 & 96.1 $ \pm $ 0.0 & 95.9 $ \pm $ 0.1 \\
\bottomrule
\end{tabular}
\end{table}

\begin{table}[h]
\caption{AUC, Joint CelebA training, ResNet18 }
\label{table:auc-CelebA-ResNet18-joint}
\begin{tabular}{l|ccccccc}
\toprule
{Sparsity} & {0} & {80} & {90} & {95} & {98} & {99} & {99.5} \\
{Attribute} & {} & {} & {} & {} & {} & {} & {} \\
\midrule 
5 o Clock Shadow & 82.4 $ \pm $ 0.7 & 82.9 $ \pm $ 0.8 & 83.3 $ \pm $ 1.0 & 83.5 $ \pm $ 0.9 & 83.7 $ \pm $ 0.7 & 83.4 $ \pm $ 0.6 & 82.0 $ \pm $ 0.7 \\
Arched Eyebrows & 78.1 $ \pm $ 0.5 & 78.9 $ \pm $ 0.4 & 79.2 $ \pm $ 0.7 & 79.9 $ \pm $ 0.6 & 80.4 $ \pm $ 0.5 & 80.1 $ \pm $ 0.4 & 78.8 $ \pm $ 0.3 \\
Attractive & 80.5 $ \pm $ 0.2 & 81.6 $ \pm $ 0.1 & 81.8 $ \pm $ 0.3 & 82.6 $ \pm $ 0.1 & 83.1 $ \pm $ 0.1 & 83.1 $ \pm $ 0.2 & 82.7 $ \pm $ 0.1 \\
Bags Under Eyes & 72.5 $ \pm $ 0.5 & 73.4 $ \pm $ 0.7 & 73.8 $ \pm $ 0.8 & 74.6 $ \pm $ 0.5 & 74.6 $ \pm $ 0.8 & 74.4 $ \pm $ 0.7 & 72.6 $ \pm $ 0.5 \\
Bald & 85.6 $ \pm $ 1.3 & 85.6 $ \pm $ 1.6 & 86.8 $ \pm $ 1.2 & 87.4 $ \pm $ 0.6 & 86.8 $ \pm $ 0.8 & 87.3 $ \pm $ 1.5 & 83.4 $ \pm $ 1.1 \\
Bangs & 90.8 $ \pm $ 0.2 & 91.4 $ \pm $ 0.2 & 91.6 $ \pm $ 0.4 & 91.8 $ \pm $ 0.2 & 91.8 $ \pm $ 0.2 & 91.6 $ \pm $ 0.3 & 90.9 $ \pm $ 0.3 \\
\bottomrule
\end{tabular}
\end{table}

\begin{table}[h]
\caption{Expected Calibration Error, Joint CelebA training, ResNet18 }
\label{table:ece-CelebA-ResNet18-joint}
\begin{tabular}{l|ccccccc}
\toprule
{Sparsity} & {0} & {80} & {90} & {95} & {98} & {99} & {99.5} \\
{Attribute} & {} & {} & {} & {} & {} & {} & {} \\
\midrule
5 o Clock Shadow & 2.9 $ \pm $ 0.1 & 2.1 $ \pm $ 0.1 & 1.8 $ \pm $ 0.1 & 1.2 $ \pm $ 0.1 & 0.6 $ \pm $ 0.1 & 0.6 $ \pm $ 0.1 & 0.7 $ \pm $ 0.1 \\
Arched Eyebrows & 10.2 $ \pm $ 0.1 & 7.6 $ \pm $ 0.1 & 6.5 $ \pm $ 0.2 & 4.8 $ \pm $ 0.1 & 2.6 $ \pm $ 0.1 & 1.6 $ \pm $ 0.3 & 1.6 $ \pm $ 0.2 \\
Attractive & 10.6 $ \pm $ 0.2 & 6.8 $ \pm $ 0.2 & 5.5 $ \pm $ 0.4 & 3.6 $ \pm $ 0.1 & 2.0 $ \pm $ 0.2 & 1.9 $ \pm $ 0.2 & 1.9 $ \pm $ 0.2 \\
Bags Under Eyes & 10.0 $ \pm $ 0.2 & 7.4 $ \pm $ 0.1 & 6.4 $ \pm $ 0.1 & 4.4 $ \pm $ 0.2 & 1.9 $ \pm $ 0.1 & 1.4 $ \pm $ 0.2 & 1.6 $ \pm $ 0.1 \\
Bald & 0.4 $ \pm $ 0.1 & 0.3 $ \pm $ 0.0 & 0.3 $ \pm $ 0.0 & 0.3 $ \pm $ 0.1 & 0.3 $ \pm $ 0.0 & 0.3 $ \pm $ 0.1 & 0.4 $ \pm $ 0.1 \\
Bangs & 2.0 $ \pm $ 0.1 & 1.5 $ \pm $ 0.1 & 1.1 $ \pm $ 0.1 & 0.7 $ \pm $ 0.1 & 0.4 $ \pm $ 0.1 & 0.5 $ \pm $ 0.1 & 0.8 $ \pm $ 0.1 \\
\bottomrule
\end{tabular}
\end{table}

\begin{table}[h]
\caption{Threshold Calibration Bias, Joint CelebA training, ResNet18 }
\label{table:rare-val-underpredict-CelebA-ResNet18-joint}
\begin{tabular}{l|ccccccc}
\toprule
{Sparsity} & {0} & {80} & {90} & {95} & {98} & {99} & {99.5} \\
{Attribute} & {} & {} & {} & {} & {} & {} & {} \\
\midrule
5 o Clock Shadow & 92.7 $ \pm $ 4.3 & 92.1 $ \pm $ 3.6 & 91.8 $ \pm $ 3.8 & 91.1 $ \pm $ 3.6 & 89.4 $ \pm $ 2.6 & 88.6 $ \pm $ 2.4 & 84.9 $ \pm $ 3.1 \\
Arched Eyebrows & 99.4 $ \pm $ 3.6 & 99.5 $ \pm $ 3.6 & 99.1 $ \pm $ 3.9 & 99.9 $ \pm $ 3.0 & 99.0 $ \pm $ 2.8 & 98.2 $ \pm $ 2.9 & 96.1 $ \pm $ 2.3 \\
Attractive & 97.0 $ \pm $ 0.9 & 96.8 $ \pm $ 1.0 & 96.3 $ \pm $ 0.9 & 96.3 $ \pm $ 0.5 & 96.2 $ \pm $ 0.6 & 95.8 $ \pm $ 0.8 & 94.9 $ \pm $ 0.6 \\
Bags Under Eyes & 92.3 $ \pm $ 2.8 & 91.1 $ \pm $ 3.5 & 90.6 $ \pm $ 3.7 & 89.4 $ \pm $ 2.5 & 85.2 $ \pm $ 3.6 & 83.2 $ \pm $ 3.3 & 77.2 $ \pm $ 1.3 \\
Bald & 93.9 $ \pm $ 3.5 & 94.4 $ \pm $ 5.5 & 97.8 $ \pm $ 4.8 & 98.3 $ \pm $ 3.9 & 97.4 $ \pm $ 4.2 & 99.2 $ \pm $ 5.7 & 90.8 $ \pm $ 5.4 \\
Bangs & 96.7 $ \pm $ 0.9 & 96.9 $ \pm $ 0.9 & 96.7 $ \pm $ 1.5 & 96.4 $ \pm $ 0.8 & 95.6 $ \pm $ 1.0 & 94.9 $ \pm $ 1.2 & 93.7 $ \pm $ 0.5 \\
\bottomrule
\end{tabular}
\end{table}

\begin{table}[h]
\caption{Uncertainty, Joint CelebA training, ResNet18 }
\label{table:uncertainty-CelebA-ResNet18-joint}
\begin{tabular}{l|ccccccc}
\toprule
{Sparsity} & {0} & {80} & {90} & {95} & {98} & {99} & {99.5} \\
{Attribute} & {} & {} & {} & {} & {} & {} & {} \\
\midrule
5 o Clock Shadow & 9.3 $ \pm $ 0.4 & 10.9 $ \pm $ 0.4 & 11.7 $ \pm $ 0.4 & 12.9 $ \pm $ 0.4 & 14.9 $ \pm $ 0.4 & 16.5 $ \pm $ 0.4 & 18.5 $ \pm $ 0.4 \\
Arched Eyebrows & 24.3 $ \pm $ 0.3 & 30.8 $ \pm $ 0.4 & 33.4 $ \pm $ 0.1 & 37.5 $ \pm $ 0.3 & 43.4 $ \pm $ 0.5 & 47.2 $ \pm $ 0.4 & 50.5 $ \pm $ 0.4 \\
Attractive & 28.8 $ \pm $ 0.5 & 37.6 $ \pm $ 0.6 & 40.8 $ \pm $ 0.4 & 45.0 $ \pm $ 0.5 & 50.0 $ \pm $ 0.3 & 52.5 $ \pm $ 0.3 & 54.2 $ \pm $ 0.4 \\
Bags Under Eyes & 22.1 $ \pm $ 0.4 & 28.1 $ \pm $ 0.3 & 30.6 $ \pm $ 0.2 & 35.0 $ \pm $ 0.4 & 40.9 $ \pm $ 0.3 & 44.8 $ \pm $ 0.3 & 47.2 $ \pm $ 0.5 \\
Bald & 2.1 $ \pm $ 0.1 & 2.2 $ \pm $ 0.1 & 2.2 $ \pm $ 0.1 & 2.5 $ \pm $ 0.1 & 2.7 $ \pm $ 0.1 & 2.9 $ \pm $ 0.1 & 3.1 $ \pm $ 0.2 \\
Bangs & 7.5 $ \pm $ 0.2 & 8.5 $ \pm $ 0.2 & 9.0 $ \pm $ 0.1 & 10.0 $ \pm $ 0.1 & 11.0 $ \pm $ 0.2 & 11.8 $ \pm $ 0.1 & 12.7 $ \pm $ 0.3 \\
\bottomrule
\end{tabular}
\end{table}

\begin{table}[h]
\caption{Interdependence, Joint CelebA training, ResNet18 }
\label{table:interdependence-CelebA-ResNet18-joint}
\begin{tabular}{l|ccccccc}
\toprule
{Sparsity} & {0} & {80} & {90} & {95} & {98} & {99} & {99.5} \\
{Attribute} & {} & {} & {} & {} & {} & {} & {} \\
\midrule
5 o Clock Shadow & 43.7 $ \pm $ 0.5 & 44.9 $ \pm $ 1.1 & 45.3 $ \pm $ 0.6 & 45.9 $ \pm $ 0.9 & 47.1 $ \pm $ 0.5 & 48.5 $ \pm $ 0.6 & 48.8 $ \pm $ 1.1 \\
Arched Eyebrows & 34.5 $ \pm $ 0.9 & 35.4 $ \pm $ 1.1 & 36.0 $ \pm $ 1.1 & 37.3 $ \pm $ 0.7 & 38.2 $ \pm $ 0.5 & 39.3 $ \pm $ 0.7 & 42.1 $ \pm $ 1.2 \\
Attractive & 44.6 $ \pm $ 0.2 & 46.8 $ \pm $ 0.5 & 48.2 $ \pm $ 0.2 & 50.5 $ \pm $ 0.4 & 53.1 $ \pm $ 0.3 & 54.5 $ \pm $ 0.4 & 56.0 $ \pm $ 0.6 \\
Bags Under Eyes & 28.1 $ \pm $ 1.0 & 30.3 $ \pm $ 0.9 & 30.9 $ \pm $ 1.1 & 32.4 $ \pm $ 0.7 & 33.6 $ \pm $ 0.8 & 35.1 $ \pm $ 1.1 & 36.6 $ \pm $ 0.7 \\
Bald & 13.0 $ \pm $ 0.6 & 13.0 $ \pm $ 0.9 & 13.8 $ \pm $ 0.5 & 13.6 $ \pm $ 0.6 & 13.3 $ \pm $ 0.5 & 13.9 $ \pm $ 0.6 & 14.4 $ \pm $ 0.8 \\
Bangs & 10.5 $ \pm $ 0.4 & 10.9 $ \pm $ 0.4 & 11.4 $ \pm $ 0.3 & 11.6 $ \pm $ 0.2 & 12.2 $ \pm $ 0.3 & 12.5 $ \pm $ 0.2 & 12.9 $ \pm $ 0.2 \\
\bottomrule
\end{tabular}
\end{table}

\begin{table}[h]
\caption{'Male' Bias Amplification, Joint CelebA training, ResNet18 }
\label{table:Male-bas-CelebA-ResNet18-joint}
\begin{tabular}{l|ccccccc}
\toprule
{Sparsity} & {0} & {80} & {90} & {95} & {98} & {99} & {99.5} \\
{Attribute} & {} & {} & {} & {} & {} & {} & {} \\
\midrule
5 o Clock Shadow & -- & -- & -- & -- & -- & -- & -- \\
Arched Eyebrows & 2.3 $ \pm $ 0.4 & 2.4 $ \pm $ 0.2 & 2.4 $ \pm $ 0.2 & 2.7 $ \pm $ 0.3 & 3.0 $ \pm $ 0.5 & 3.8 $ \pm $ 0.5 & 5.1 $ \pm $ 0.2 \\
Attractive & -0.0 $ \pm $ 0.3 & 0.1 $ \pm $ 0.2 & 0.3 $ \pm $ 0.2 & 0.6 $ \pm $ 0.2 & 0.8 $ \pm $ 0.2 & 1.0 $ \pm $ 0.3 & 1.6 $ \pm $ 0.2 \\
Bags Under Eyes & 1.9 $ \pm $ 0.5 & 2.4 $ \pm $ 0.8 & 3.1 $ \pm $ 0.5 & 4.1 $ \pm $ 0.2 & 5.4 $ \pm $ 0.9 & 6.6 $ \pm $ 0.6 & 8.8 $ \pm $ 0.8 \\
Bald & -- & -- & -- & -- & -- & -- & -- \\
Bangs & -0.1 $ \pm $ 0.2 & -0.0 $ \pm $ 0.4 & 0.1 $ \pm $ 0.4 & 0.3 $ \pm $ 0.1 & 0.1 $ \pm $ 0.2 & 0.1 $ \pm $ 0.2 & -0.3 $ \pm $ 0.2 \\
\bottomrule
\end{tabular}
\end{table}

\begin{table}[h]
\caption{'Young' Bias Amplification, Joint CelebA training, ResNet18 }
\label{table:Young-bas-CelebA-ResNet18-joint}
\begin{tabular}{l|ccccccc}
\toprule
{Sparsity} & {0} & {80} & {90} & {95} & {98} & {99} & {99.5} \\
{Attribute} & {} & {} & {} & {} & {} & {} & {} \\
\midrule
5 o Clock Shadow & -- & -- & -- & -- & -- & -- & -- \\
Arched Eyebrows & 0.1 $ \pm $ 0.2 & 0.0 $ \pm $ 0.2 & -0.0 $ \pm $ 0.3 & 0.2 $ \pm $ 0.1 & 0.2 $ \pm $ 0.2 & 0.3 $ \pm $ 0.1 & 0.7 $ \pm $ 0.2 \\
Attractive & 0.1 $ \pm $ 0.1 & 0.4 $ \pm $ 0.3 & 0.6 $ \pm $ 0.1 & 0.8 $ \pm $ 0.2 & 1.2 $ \pm $ 0.1 & 1.1 $ \pm $ 0.2 & 0.7 $ \pm $ 0.1 \\
Bags Under Eyes & 1.4 $ \pm $ 0.6 & 1.9 $ \pm $ 0.6 & 2.1 $ \pm $ 0.4 & 3.0 $ \pm $ 0.8 & 3.7 $ \pm $ 0.5 & 4.7 $ \pm $ 0.8 & 6.3 $ \pm $ 0.7 \\
Bald & -1.3 $ \pm $ 0.5 & -1.3 $ \pm $ 0.4 & -1.6 $ \pm $ 0.8 & -1.7 $ \pm $ 1.1 & -2.3 $ \pm $ 0.3 & -1.7 $ \pm $ 0.5 & -1.3 $ \pm $ 0.6 \\
Bangs & 0.3 $ \pm $ 0.4 & 0.3 $ \pm $ 0.2 & 0.2 $ \pm $ 0.2 & 0.0 $ \pm $ 0.4 & 0.0 $ \pm $ 0.2 & 0.0 $ \pm $ 0.3 & -0.2 $ \pm $ 0.3 \\
\bottomrule
\end{tabular}
\end{table}

\begin{table}[h]
\caption{'Chubby' Bias Amplification, Joint CelebA training, ResNet18 }
\label{table:Chubby-bas-CelebA-ResNet18-joint}
\begin{tabular}{l|ccccccc}
\toprule
{Sparsity} & {0} & {80} & {90} & {95} & {98} & {99} & {99.5} \\
{Attribute} & {} & {} & {} & {} & {} & {} & {} \\
\midrule 
5 o Clock Shadow & 1.1 $ \pm $ 0.3 & 1.0 $ \pm $ 0.3 & 1.3 $ \pm $ 0.4 & 1.5 $ \pm $ 0.2 & 1.6 $ \pm $ 0.2 & 1.6 $ \pm $ 0.2 & 1.7 $ \pm $ 0.3 \\
Arched Eyebrows & 0.4 $ \pm $ 0.0 & 0.4 $ \pm $ 0.1 & 0.4 $ \pm $ 0.1 & 0.3 $ \pm $ 0.1 & 0.3 $ \pm $ 0.1 & 0.3 $ \pm $ 0.1 & 0.4 $ \pm $ 0.1 \\
Attractive & -0.1 $ \pm $ 0.0 & -0.1 $ \pm $ 0.0 & -0.1 $ \pm $ 0.0 & -0.0 $ \pm $ 0.1 & -0.0 $ \pm $ 0.0 & -0.0 $ \pm $ 0.0 & -0.1 $ \pm $ 0.0 \\
Bags Under Eyes & -0.1 $ \pm $ 0.2 & -0.1 $ \pm $ 0.3 & 0.1 $ \pm $ 0.3 & 0.5 $ \pm $ 0.2 & 1.2 $ \pm $ 0.4 & 1.8 $ \pm $ 0.2 & 2.4 $ \pm $ 0.1 \\
Bald & -2.0 $ \pm $ 1.2 & -2.0 $ \pm $ 1.8 & -2.3 $ \pm $ 0.8 & -2.1 $ \pm $ 0.4 & -3.2 $ \pm $ 0.8 & -3.1 $ \pm $ 0.1 & -2.4 $ \pm $ 0.9 \\
Bangs & -0.2 $ \pm $ 0.1 & -0.2 $ \pm $ 0.1 & -0.1 $ \pm $ 0.1 & -0.1 $ \pm $ 0.0 & -0.1 $ \pm $ 0.1 & -0.1 $ \pm $ 0.1 & -0.1 $ \pm $ 0.1 \\
\bottomrule
\end{tabular}
\end{table}

\begin{table}[h]
\caption{'Pale Skin' Bias Amplification, Joint CelebA training, ResNet18 }
\label{table:Pale Skin-bas-CelebA-ResNet18-joint}
\begin{tabular}{l|ccccccc}
\toprule
{Sparsity} & {0} & {80} & {90} & {95} & {98} & {99} & {99.5} \\
{Attribute} & {} & {} & {} & {} & {} & {} & {} \\
\midrule
5 o Clock Shadow & 0.0 $ \pm $ 0.1 & -0.0 $ \pm $ 0.1 & -0.0 $ \pm $ 0.1 & 0.0 $ \pm $ 0.2 & 0.0 $ \pm $ 0.1 & -0.0 $ \pm $ 0.1 & -0.0 $ \pm $ 0.1 \\
Arched Eyebrows & -1.0 $ \pm $ 0.2 & -1.1 $ \pm $ 0.2 & -0.9 $ \pm $ 0.1 & -0.9 $ \pm $ 0.1 & -0.8 $ \pm $ 0.1 & -0.9 $ \pm $ 0.1 & -1.0 $ \pm $ 0.2 \\
Attractive & 0.1 $ \pm $ 0.1 & 0.1 $ \pm $ 0.1 & 0.1 $ \pm $ 0.1 & 0.2 $ \pm $ 0.1 & 0.3 $ \pm $ 0.1 & 0.4 $ \pm $ 0.0 & 0.4 $ \pm $ 0.1 \\
Bags Under Eyes & 0.5 $ \pm $ 0.1 & 0.5 $ \pm $ 0.1 & 0.5 $ \pm $ 0.2 & 0.6 $ \pm $ 0.1 & 0.5 $ \pm $ 0.1 & 0.7 $ \pm $ 0.1 & 0.9 $ \pm $ 0.1 \\
Bald & -- & -- & -- & -- & -- & -- & -- \\
Bangs & -0.1 $ \pm $ 0.1 & -0.1 $ \pm $ 0.2 & -0.1 $ \pm $ 0.1 & -0.1 $ \pm $ 0.1 & -0.0 $ \pm $ 0.1 & -0.1 $ \pm $ 0.2 & -0.0 $ \pm $ 0.1 \\
\bottomrule
\end{tabular}
\end{table}

\clearpage
\section{Results on the Animals with Attributes Dataset}
\label{appendix:awa}
In our efforts of investigating the exacerbation of bias in sparse models, we further validate our results on CelebA on the Animals with Attributes (AwA2)~\cite{Xian2019AwAZeroShotLC} dataset, which consists of 37 322 images of animals belonging to 50 different classes. Each class is annotated using 85 binary attributes, which indicate the presence or absence of different characteristics in each species. We note that AwA2 is not as suited for the study of bias as CelebA, for two important reasons: first, there is a reduced sociological incentive of studying bias, compared to a dataset consisting of human subjects; furthermore, the attributes are labelled at species level, rather than individually per sample, which makes it more difficult to disambiguate between different sources of bias. Nonetheless, we believe AwA2 still serves as a useful validation for our findings on CelebA. 

In our experiments with AwA2, we train dense and GMP-RI models at $\{80\%, 90\%, 95\%, 98\%, 99\%, 99.5\%\}$ sparsities 
to predict the 85 binary attributes. For both the dense and sparse models we use the same training setup and hyperparameters as for CelebA. We follow the original dataset split~\cite{Xian2019AwAZeroShotLC}, where the train and test set classes are disjoint: 40 classes are used for training and validation, and the remaining 10 we leave for testing. We follow a different split for train and validation, compared to~\cite{Xian2019AwAZeroShotLC}; namely, we randomly select 80\% of the samples for training and the remaining 20\% for validation. Our choice is motivated by the fact that further splitting the classes between train and validation would make it more likely to exclude certain attributes from the train set; this would be detrimental to our analysis, as we want to measure the presence of bias on certain attributes. The categories under which it is most sensible to study Categorical bias are not well-established for Animals with Attributes; here we use Furry, Bipedal, Domestic, and Water, where the last refers to the animal's natural habitat.

\begin{figure}[h]
    \centering
\begin{tabular}{cccc}
   \includegraphics[width=0.22\textwidth]{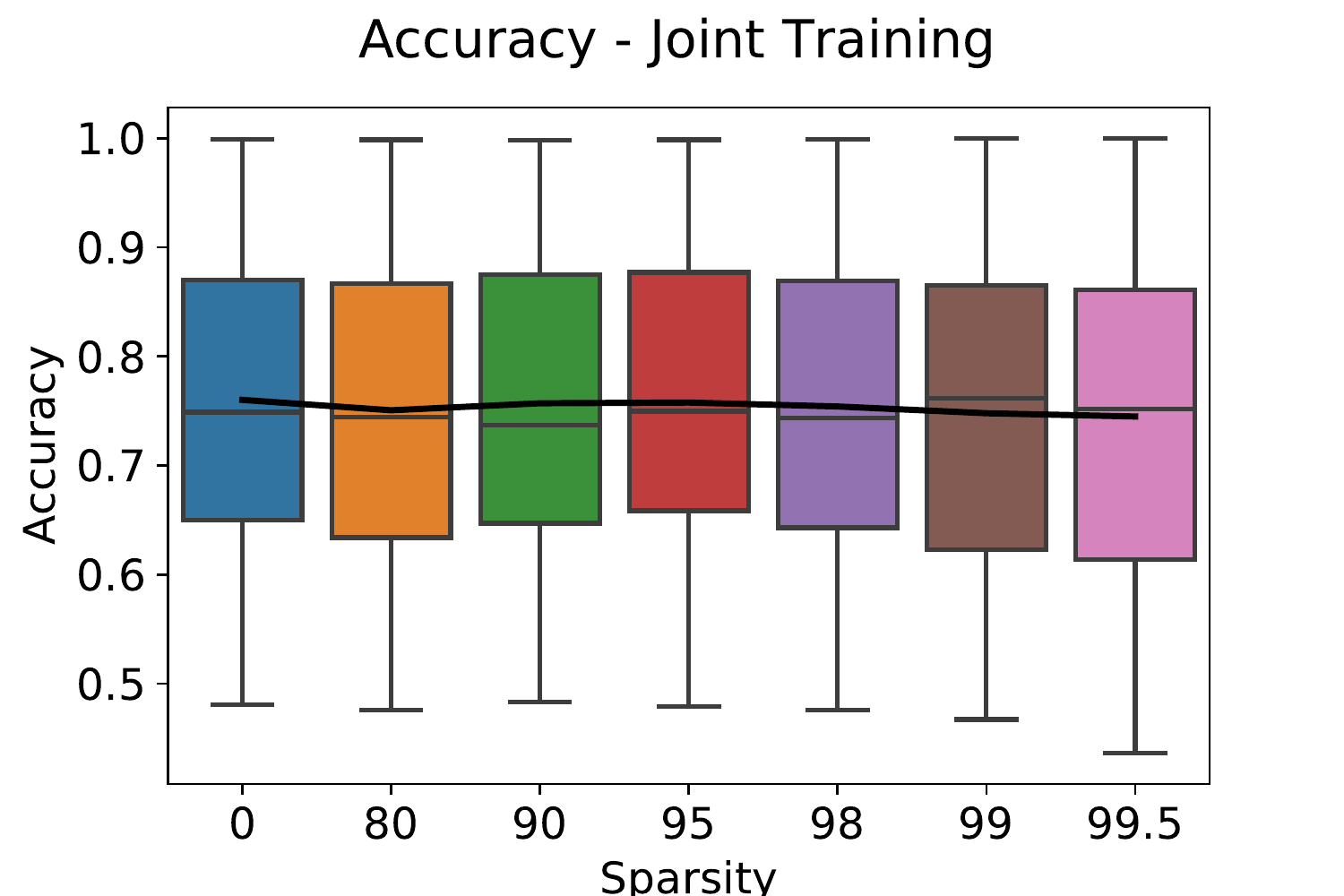} &
   \includegraphics[width=0.22\textwidth]{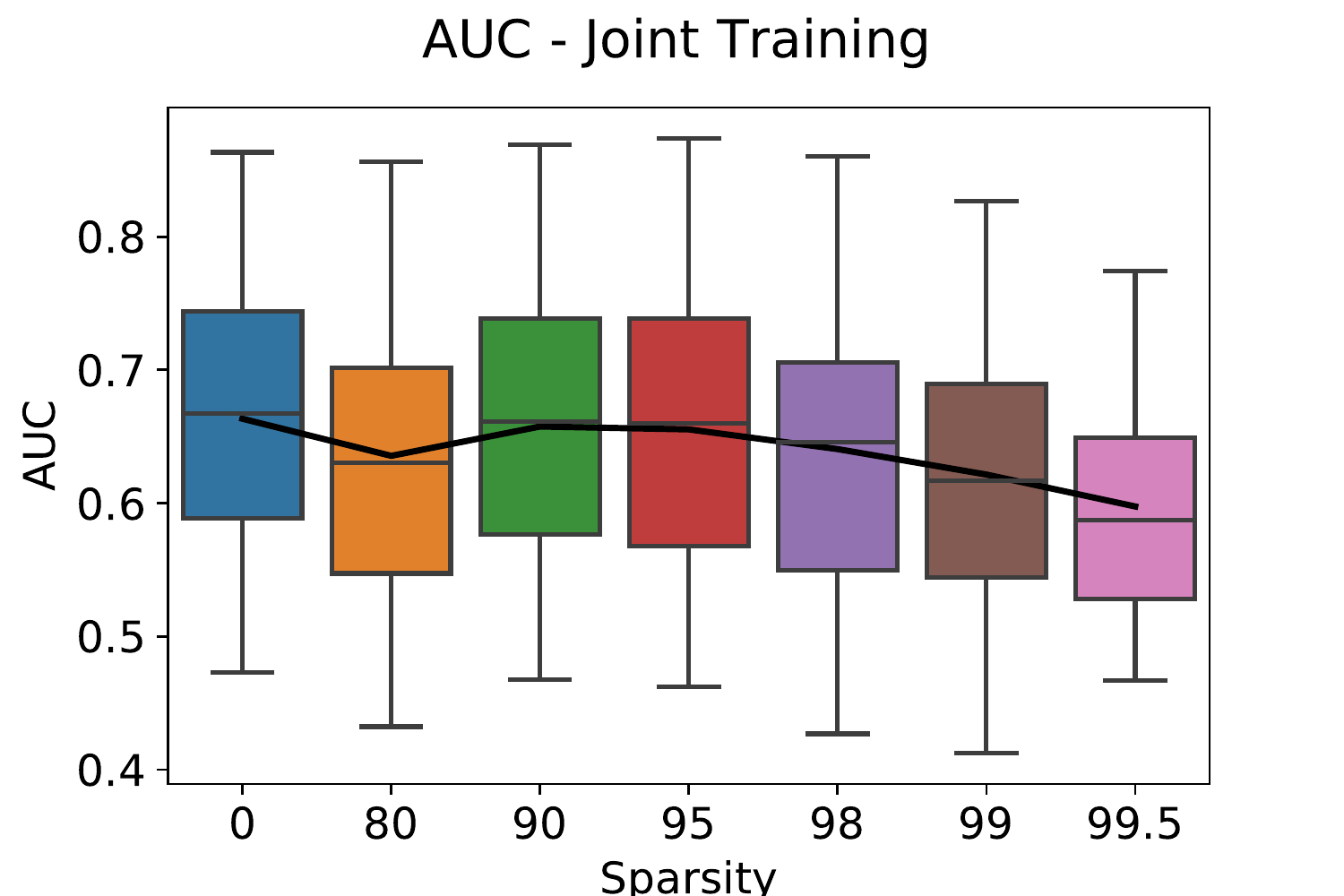} & & \\
    \includegraphics[width=0.22\textwidth]{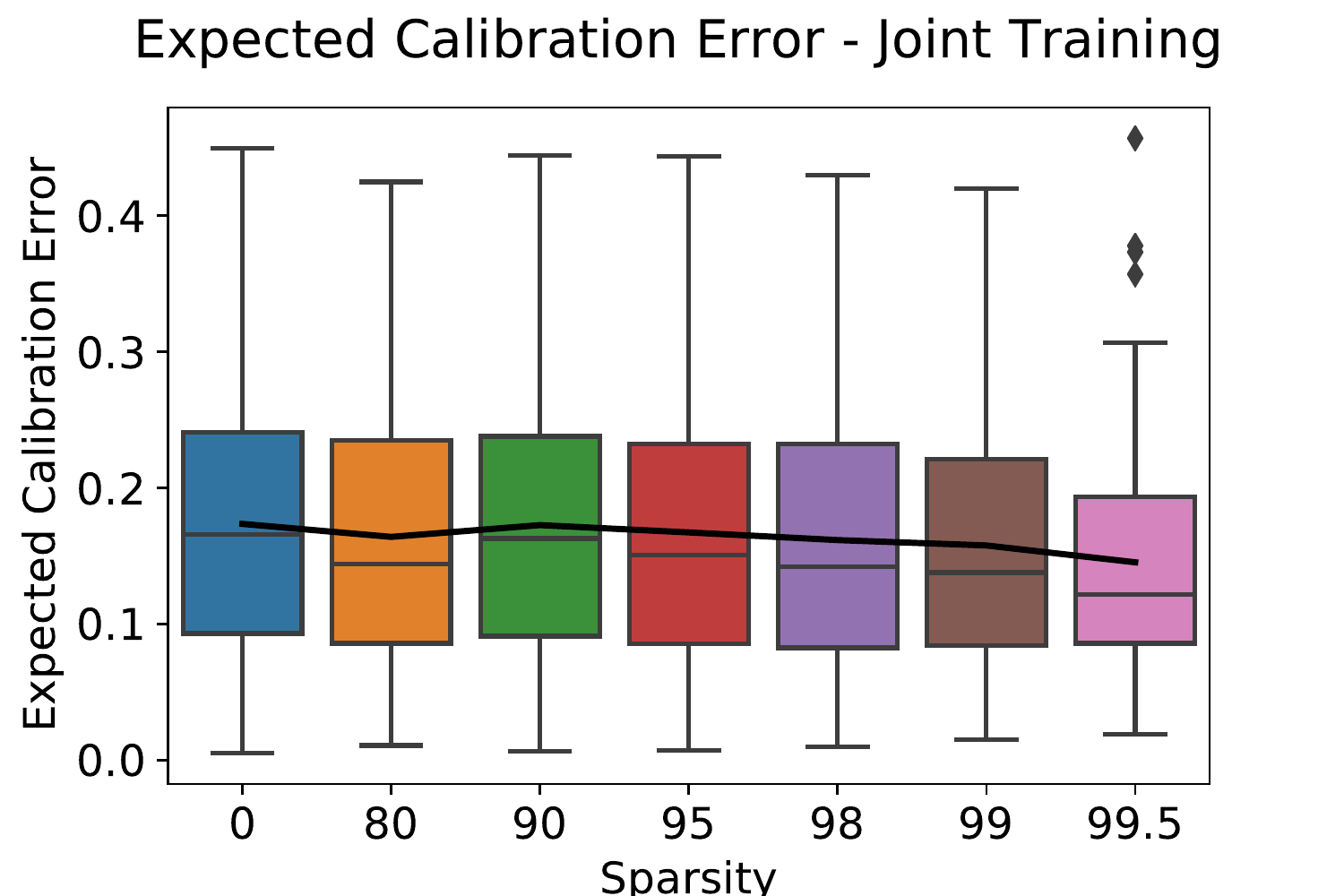} &
    \includegraphics[width=0.22\textwidth]{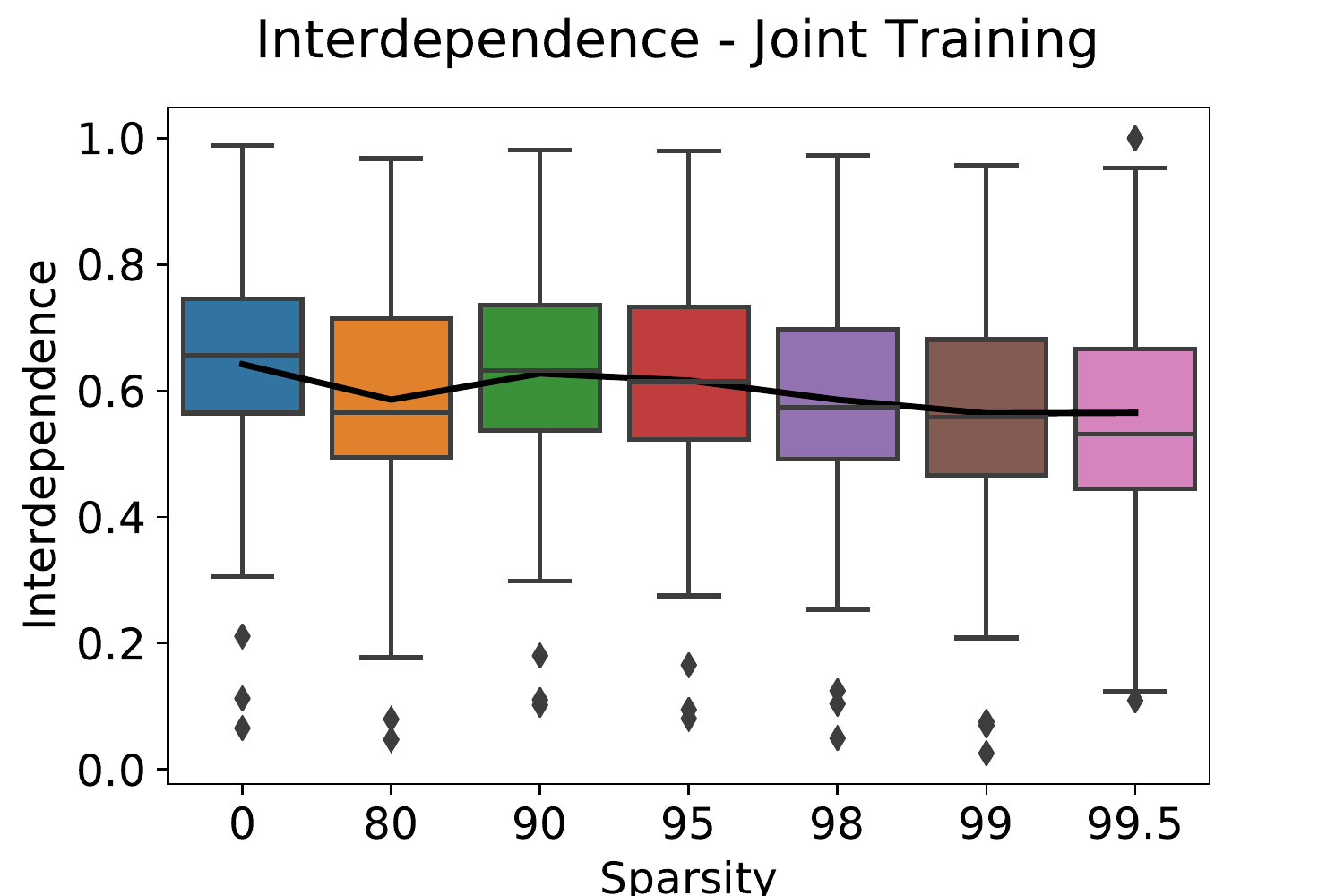} & 
    \includegraphics[width=0.22\textwidth]{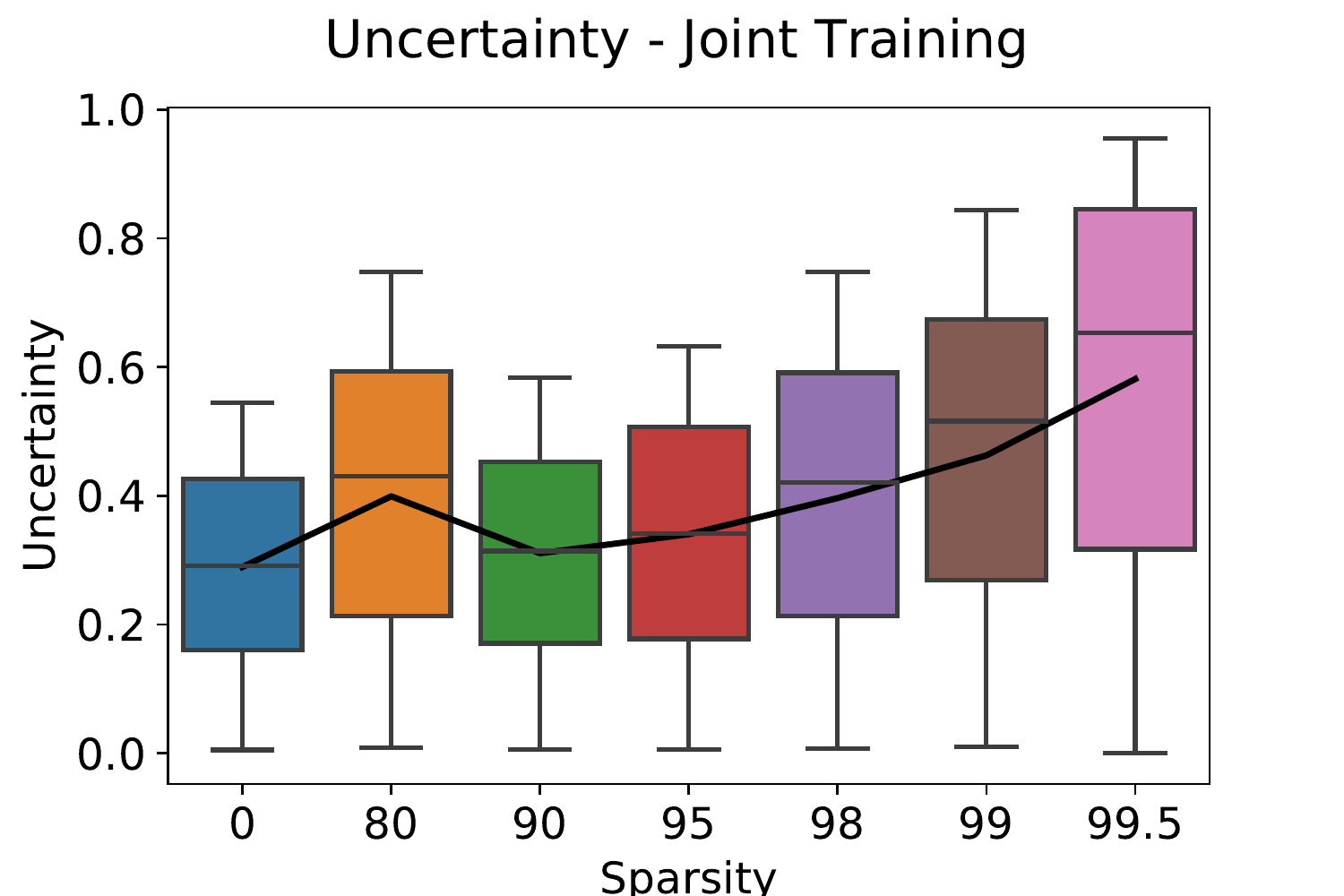} &
    \includegraphics[width=0.22\textwidth]{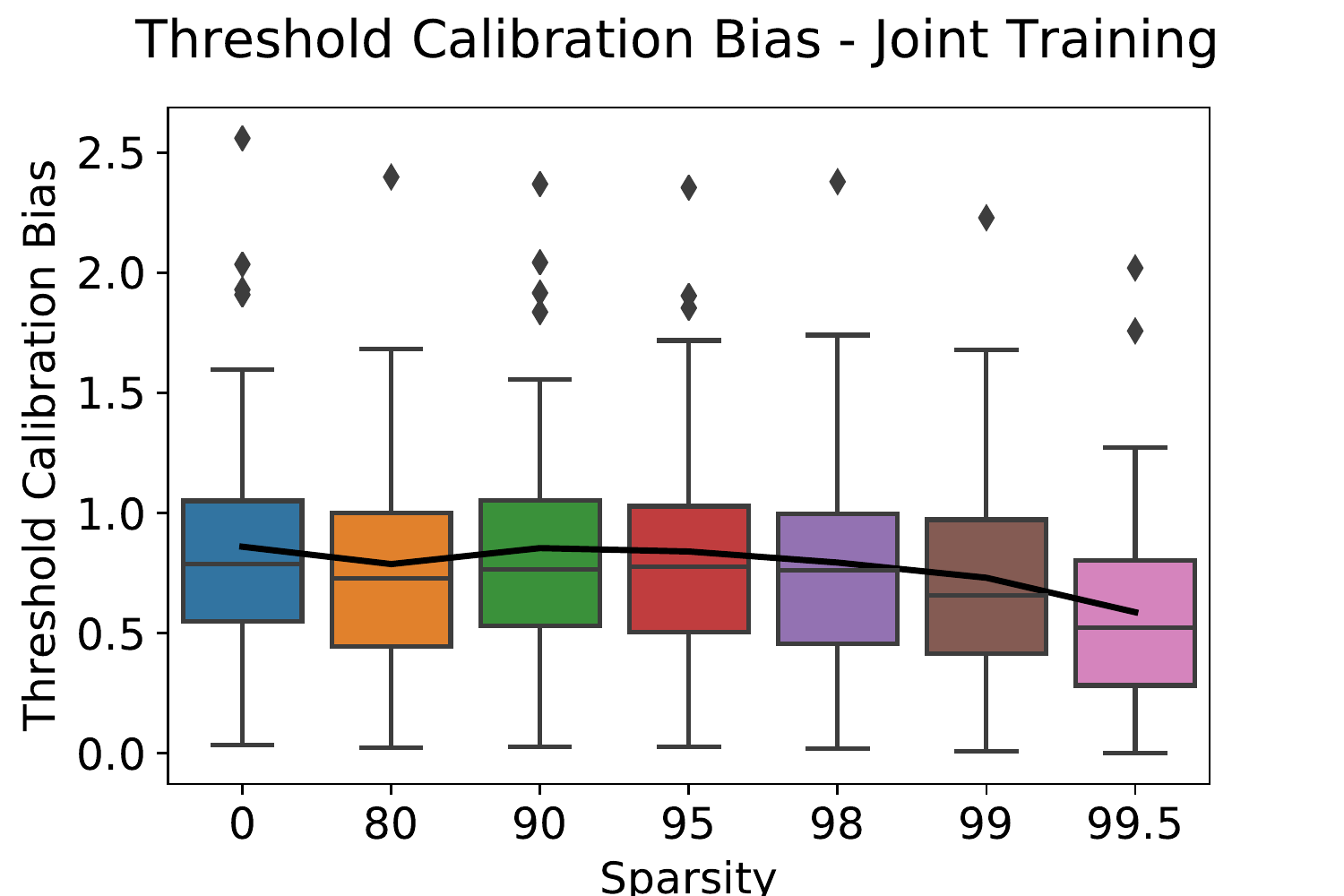}\\
    \includegraphics[width=0.22\textwidth]{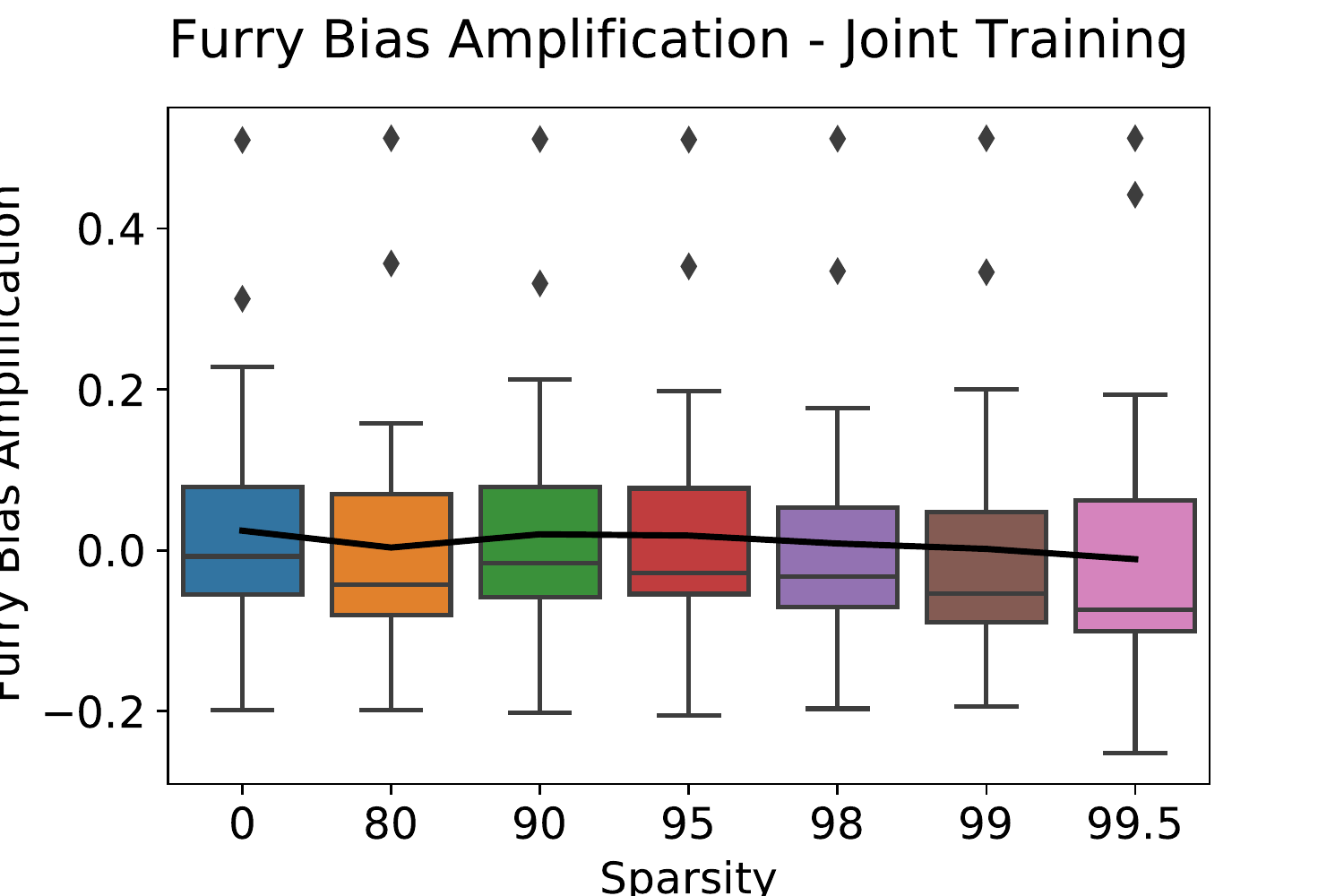} &
    \includegraphics[width=0.22\textwidth]{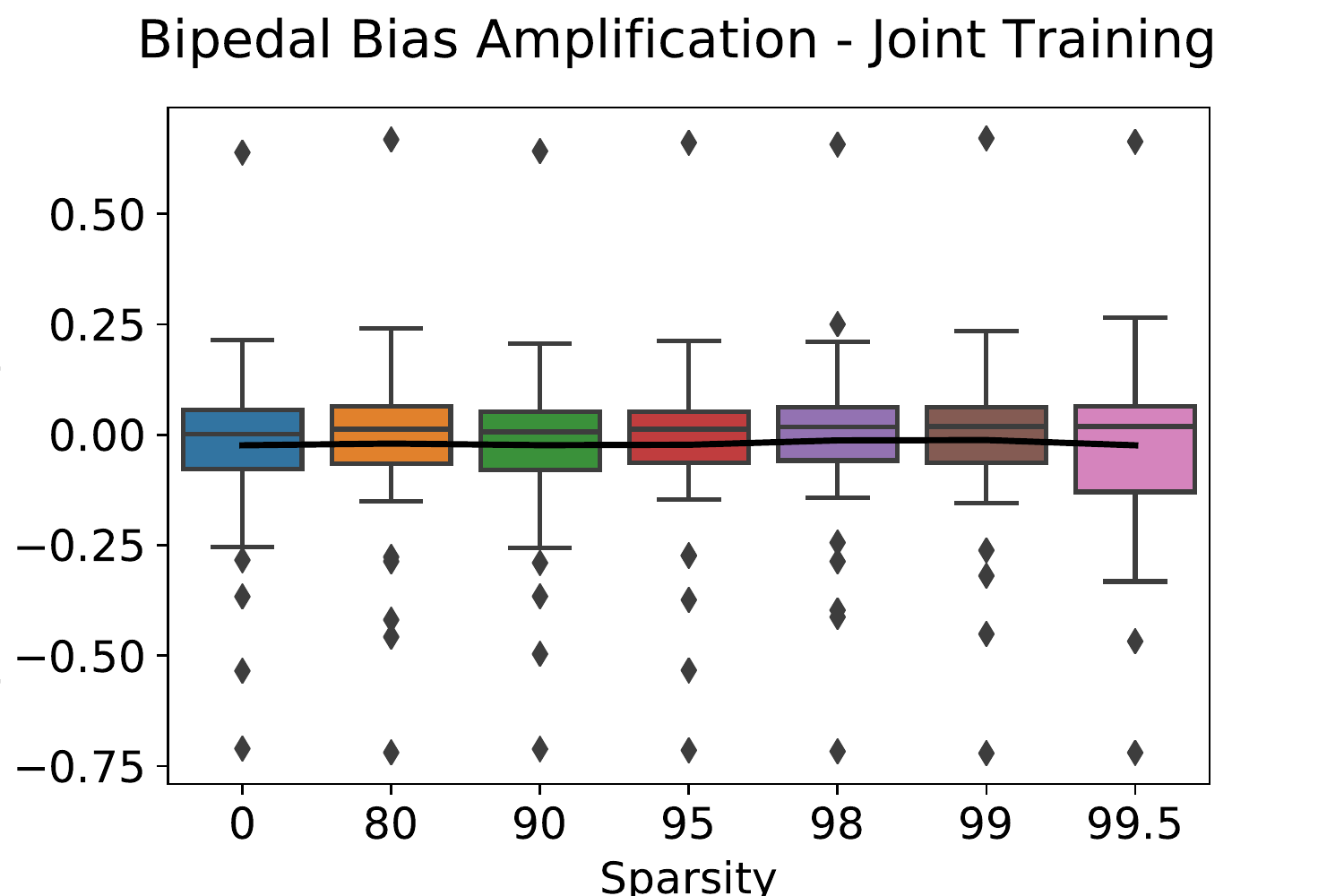} &
        \includegraphics[width=0.22\textwidth]{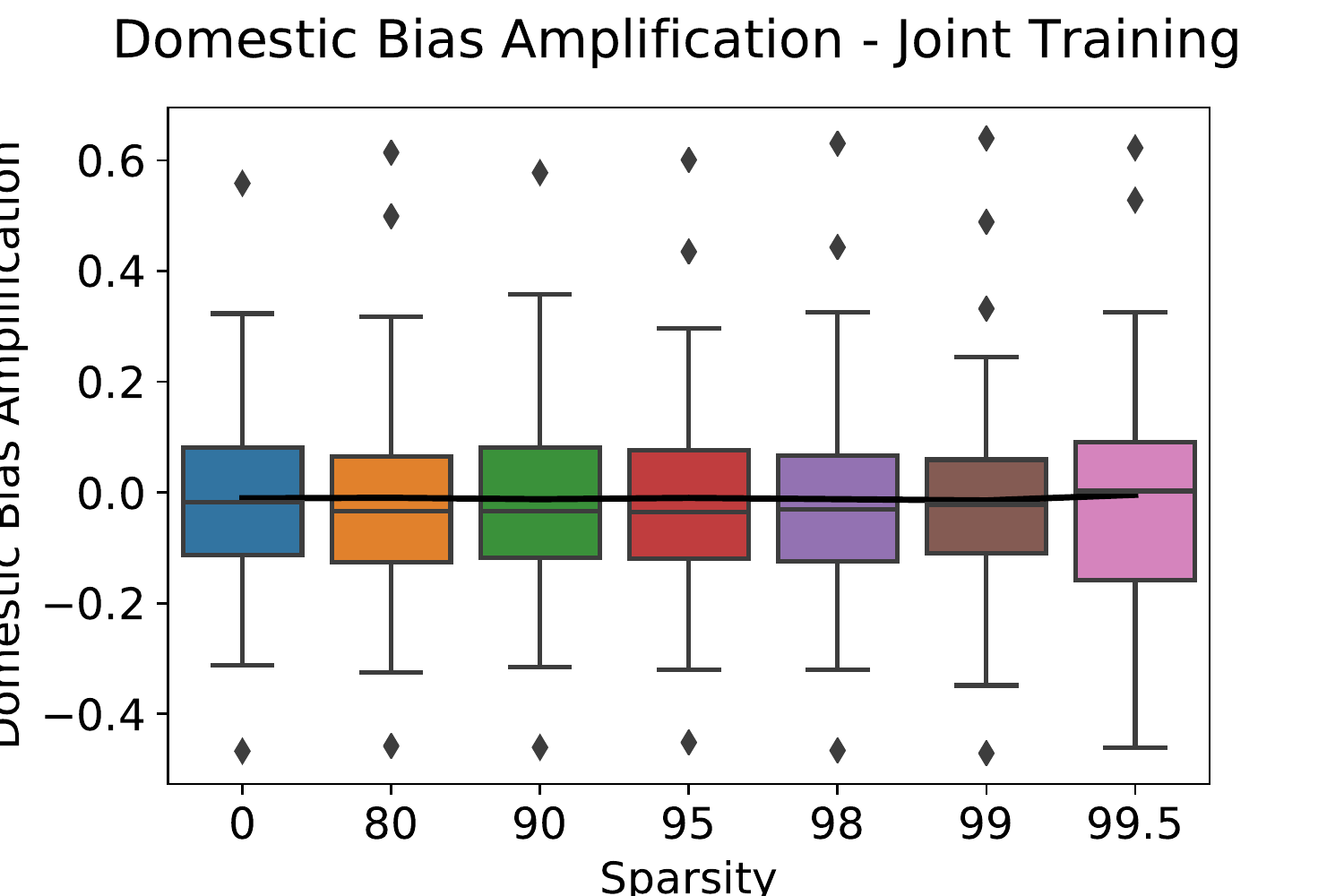} &
    \includegraphics[width=0.22\textwidth]{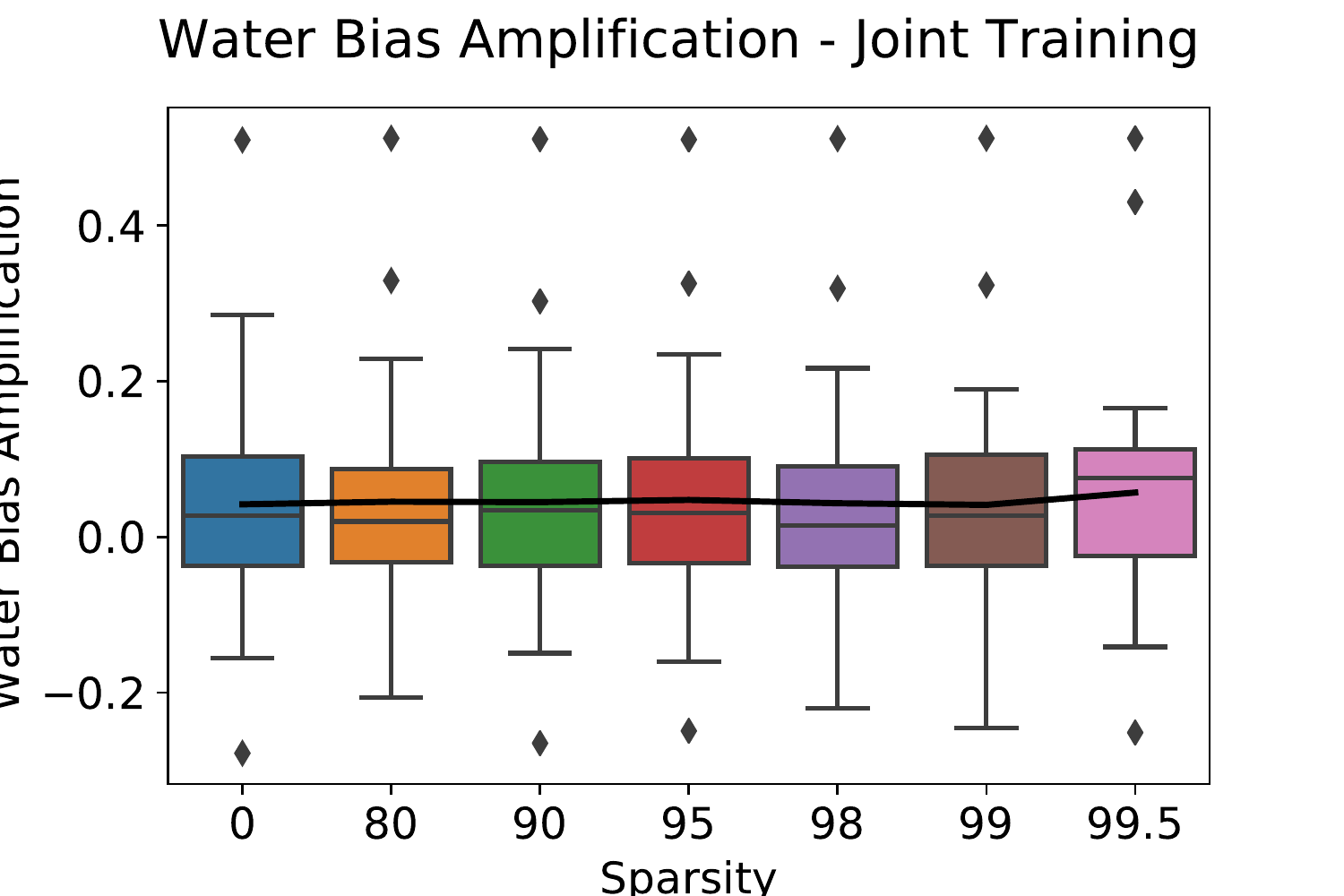}\\
\end{tabular}
    \caption{[Animals With Attributes2 / ResNet18 / GMP-RI] Accuracy and Systematic Bias metrics (TCB, ECE, Interdependence) of ResNet18 models jointly trained on all AwA2 attributes. The thick black line denotes the mean value at each sparsity level.
    }
    \label{fig:awa_rn18_bias}
\end{figure}

Our results are shown in Figure~\ref{fig:awa_rn18_bias}. We observe a degradation in AUC scores for models at $\geq 98\% $sparsity, whereas the accuracy does not decrease significantly even at $99.5\%$ sparsity. Moreover, the fraction of uncertain samples increases substantially at $\geq 98\%$ sparsity, and roughly doubles compared to the dense model at $99.5\%$ sparsity. Other metrics, such as TCB or interdependence, decrease slightly with sparsity, compared to the dense model; however, in the case of Systematic (and, to a large extent, Categorical) bias, the fact that the attributes are labeled at the species level - and therefore the model need only learn the species to also learn all the labels - makes this result difficult to interpret. We further study the amplification of bias with sparsity, by following a similar approach to the one on CelebA: namely, we select four category identity attributes with respect to which we compute bias amplification on the remaining attributes. On all attributes considered we did not observe a significant increase in bias induced by sparsity. Generally, our observations on AwA2 seem to validate our findings from CelebA: good quality models even at high sparsity, and substantially increased uncertainty with sparsity.

\clearpage
\section{iWildcam Results}
\label{appendix:iwildcam}

The iWildCam dataset\cite{beery2020iwildcam} is a set of images collected from wildlife-spotting camera traps provided by the Wildlife Conservation Society (WCS). Each image contains at least one animal, and is annotated with a single animal label (there is an extension of this dataset containing unlabelled images, but we do not use it here). In total, the dataset contains 203 029 labelled images, divided between a training set, in-distribution (ID) validation and test sets, and out-of-distribution(OOD) validation and test sets. The train (129 809 images),  ID validation (7 134 images), and ID test (8154 images) sets were obtained by splitting the photographs from 243 cameras, while the OOD validation (14 961 images) and test (42 791 images) sets were obtained using images from an additional 32 and 48 cameras, respectively. The iWildCam dataset contains images of 182 different animals and is highly unbalanced in terms of class sizes, with some classes having less than 10 images in the training data, and some over 1000. For this reason, the dataset is frequently used to study rare-subgroup performance, as in \cite{beery2020iwildcam}.

We study compression-induced bias on the iWildcam dataset by measuring the performance degradation for rarer classes. It is postulated in, e.g, \cite{hooker_characterising_2020} that features that distinguish rare examples may be cannibalized by larger classes, leading to degraded performance for those classes. To conduct our study, we trained models at 0\%, 80\%, 90\%, 95\%, and 98\% sparsity. All models used the training settings and hyperparameters (including data augmentations, batch size, epoch number, optimizer, and learning rates) used in \cite{beery2020iwildcam} for plain ERM. The pruning was done using the GMP-RI variant of Global Magnitude Pruning, with pruning beginning at epoch 2 and ending at epoch 11, with another 2 epochs afterwards for fine-tuning. We use the metrics of Macro Precision, Recall, and F1-Score used in \cite{beery2020iwildcam}; these metrics assign equal weight to each class when computing the aggregate values. Additionally, we measure the softmax entropy across classes of the predictions as a measure of uncertainty. Ths measure is computed by first computing the softmax per-class prediction for each example, $$\sigma(z)_i = \frac{e^{z_i}}{\sum_j e^{z_j}},$$ where the sum is taken over all classes. As these values sum up to 1 for each example, they may be loosely interpreted as the probabilities for each class; thus, their entropy $$H(X) = -\sum_{i}\sigma(x_i)\log(\sigma(x_i))$$ may be interpreted as a measure of uncertainty as to the correct class (where the sum is once again taken over that example's predictions for every class). To stay ideologically consistent with the Macro metrics used to evaluate accuracy, we compute the average entropy across examples by upweighting rare class examples, so that each class has equal weight in determining the average entropy.

We report our accuracy and bias results in Table~\ref{tab:iwildcam-acc}. Following convention, we report Precision, Recall, and F1-score in \%, even though F1-score is a hyperbolic mean of the first two.  We observe that the Macro F1-Score, precision, and recall stay fairly constant between Dense, 80\%, and 90\% sparse models, but then decay fairly rapidly after that, with a ID F1-Score drop of 6.4\% between 90\% sparse and 98\% sparse models =, and an OOD F1-Score drop of 5.4\%. We also note that precision and recall are fairly well balanced in the models. The dense results are a fairly close match to the results obtained in \cite{beery2020iwildcam}; we attribute the difference primarily to the choice of random seed.
\begin{table}[t]
    \centering
    \scalebox{0.8}{
    \begin{tabular}{cccccc}
        \toprule
         \multirow{2}{*}{Metric} &  \multirow{2}{*}{Dense} & \multicolumn{4}{c}{Sparsity (\%)}\\
         & & 80 & 90 & 95 & 98\\
         \midrule 
         ID F1 Score (\%) & 50.1$\pm$0.3  & 51.6$\pm$0.6 & 50.1$\pm$1.8  & 48.2$\pm$1.6 & 43.7$\pm$1.2 \\
        OOD F1 Score (\%) & 38.5$\pm$1.3 & 39.8$\pm$ 0.7 & 38.9 $\pm$ 1.9 & 37.2 $\pm$ 1.6 & 33.4 $\pm$1.3\\
        \midrule
        ID Precision (\%) & 54.1$\pm$0.8  & 55.3$\pm$0.4 & 54.6$\pm$2.2  & 52.7$\pm$2.5 & 49.3$\pm$1.9 \\
        OOD Precision (\%) & 41.5$\pm$0.8 & 43.2$\pm$0.4 & 43.4 $\pm$ 2.2 & 41.5 $\pm$2.5 & 38.2 $\pm$1.9\\
        \midrule
        ID Recall (\%) & 53.1$\pm$0.6  & 53.3$\pm$0.7 & 51.2$\pm$1.7  & 50.4$\pm$2.9 & 45.4$\pm$5.6 \\
        OOD Recall (\%) & 39.6$\pm$0.7 & 40.1$\pm$ 0.7 & 40.0 $\pm$ 1.6 & 38.6 $\pm$ 1.3 & 35.5 $\pm$1.7\\

        \bottomrule 
        
    \end{tabular}
    }
    \caption{Average ID and OOD Test Accuracy and for iWildcam models}
    \label{tab:iwildcam-acc}
\end{table}

\begin{figure}[h]
\centering
  \includegraphics[width=0.85\textwidth]{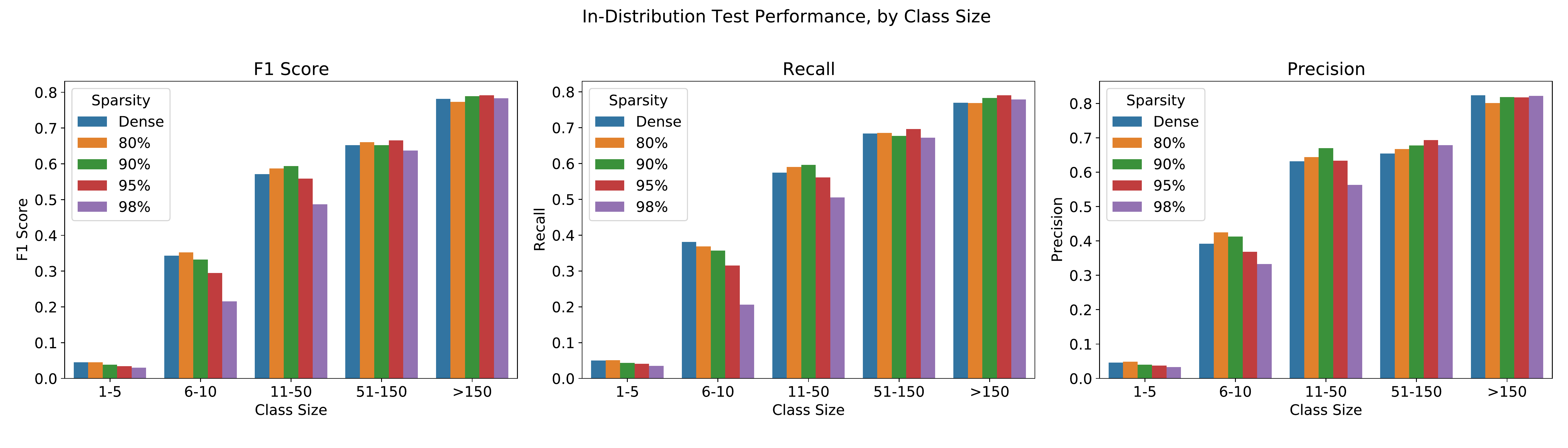} \\
  \includegraphics[width=0.85\textwidth]{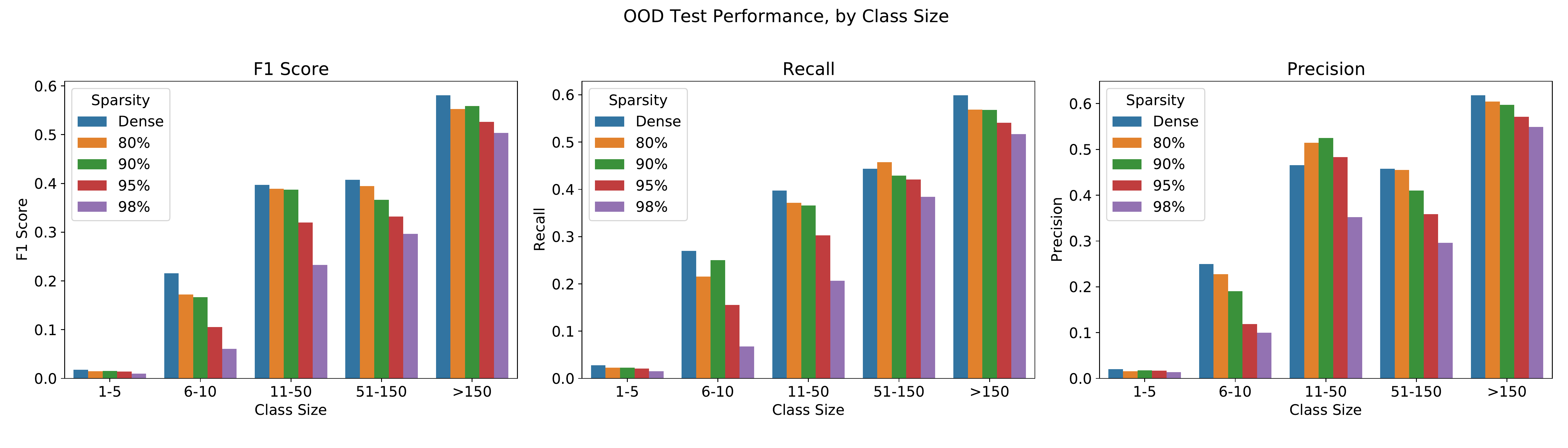} \\
  \caption{[iWildCam / ResNet18 / GMP-RI] Macro F1-Score, Precision, and Recall by sparsity and size of test class.} 
    \label{fig:iwildcam-macros}
\end{figure}

We additionally break down the dense and sparse F1-Score, Precision, and Recall by the size of the class in the test data, as shown in  Figure~\ref{fig:iwildcam-macros}. We observe that class size has a very large impact on all three metrics, with very small classes having extremely low performance as compared to larger classes. We further observe that, outside of the very low-performant 0-5 class size, sparsity disproportionately affects the performance of smaller classes, with F1-Score decreasing substantially with sparsity for classes containing 6-50 examples, but remaining nearly constant for classes of over 50 elements on ID test data. On OOD data, the performance decreases with sparsity on all class sizes (again, over 5 examples), but the decrease is greater on smaller class sizes. This experiments provides further evidence for the hypothesis outlined in \cite{hooker_characterising_2020} that ERM with sparsity can sacrifice smaller group performance to preserve accuracy on larger groups. However, we note that on the ID test data, we do not see this effect until the higher sparsity levels of 95\% and 98\%, where overall F1 score also starts to drop. 

\begin{figure}[h]
\centering
  \includegraphics[width=0.7\textwidth]{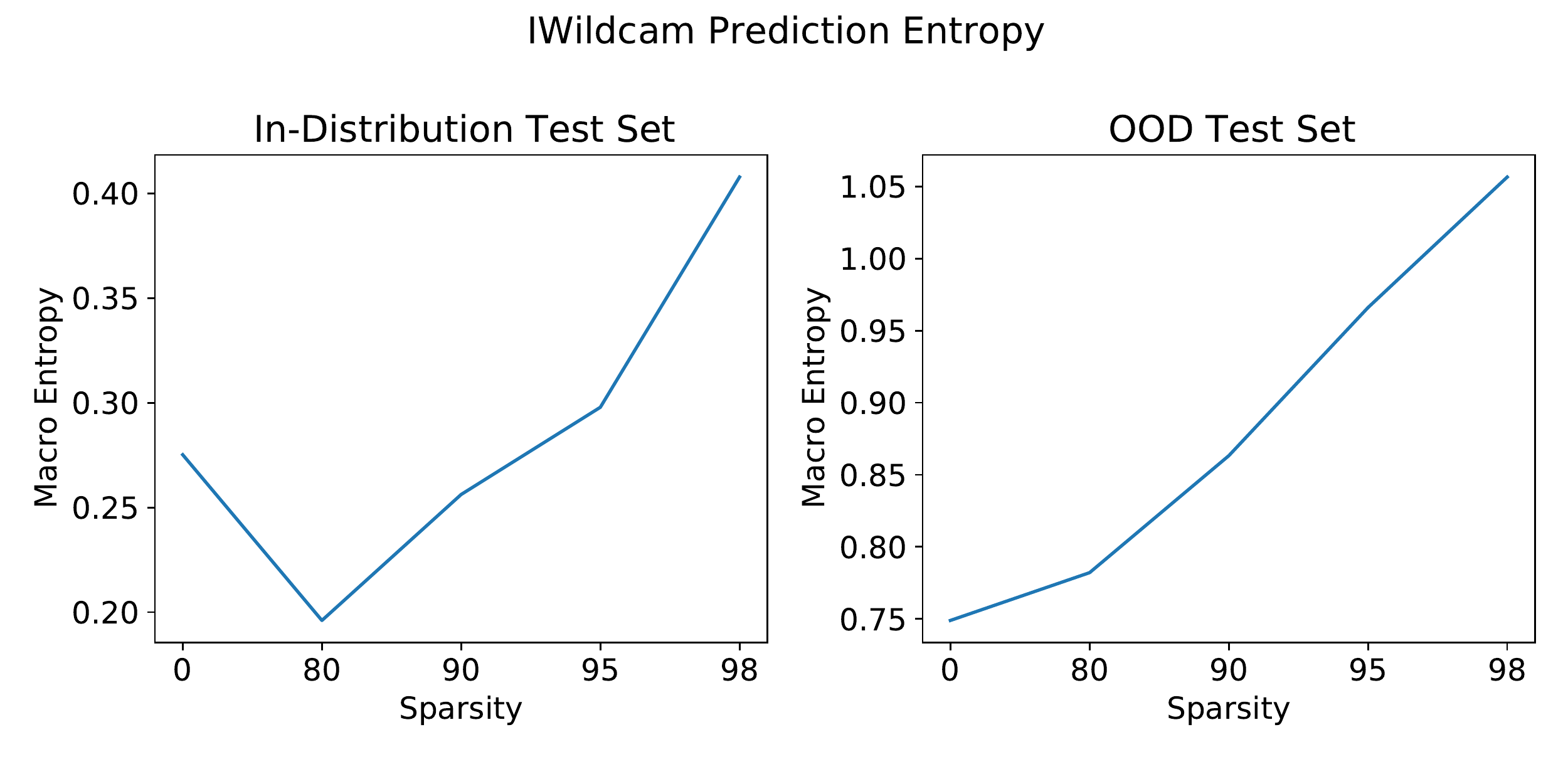} 

  \caption{[iWildCam / ResNet18 / GMP-RI] Average prediction Entropy across sparsities.} 
    \label{fig:iwildcam-entropy}
\end{figure}

The entropy of the models is shown in Figure~\ref{fig:iwildcam-entropy}. We observe that the entropy of the models increases with sparsity when measured on the OOD test set; on the ID test set, the entropy also increases, but only for high-sparsity models where the accuracy is also lower, and the smaller classes' performance is largely decayed. This adds confirmatory evidence that increased uncertainty is related to increased bias as sparsity increases.

\clearpage
\section{Example Viewer}
\label{appendix:ui_tool}

As part of our contributions, we provide a simple UI tool that allows the people working with a dataset, for example engineers or scientists who build models, to quickly and easily examine a small subset of the data. This tool is not meant to be a replacement for external review, such as example relabeling, or an audit of the data collection pipeline; if these tools are available than we strongly recommend they be used; however, they can be expensive and difficult to implement; our Example Viewer can serve as a minimum check in case that more extensive review is impossible. Further, our tool relies primarily on random sampling to choose examples to examine. This may cause users of the tool to miss small effects in the data, which may be surfaced by tools using more sophisticated error metrics to choose examples. We also note that other tools already exist that allow for model and dataset exploration, for instance the Kaggle dataset viewer, or HuggingFace Hub. However, unlike these tools, the Example Viewer runs locally. This design choice confers the advantage that neither data nor models need be uploaded to a third-party tool; in addition to increased privacy, this means that it is very easy to integrate the Example Viewer into a research pipeline, where tens or even hundreds of types models may be created as part of the study, and any of them may be instantly auditable through the tool. Finally, the tool is web-based using the popular Flask framework, and so can be run on a development machine (e.g., a laptop), on a development server while still allow for local viewing, or on a world-open server as a regular website. We provide the tool as code, which requires only Python and a few additional packages to run. It is available at [will be made available upon acceptance].

The tool has two core functionalities: viewing a random sample of positive and negative examples for a binary prediction task, and viewing a random selection of true positives, false negatives, false positives, and true negatives for a binary prediction task. These are further stratified by high and low certainty examples, using the definition in section~\ref{sec:methods}. In all cases, reloading the page produces a new random sample.

\begin{figure}[h]
\centering
  \includegraphics[width=0.7\textwidth]{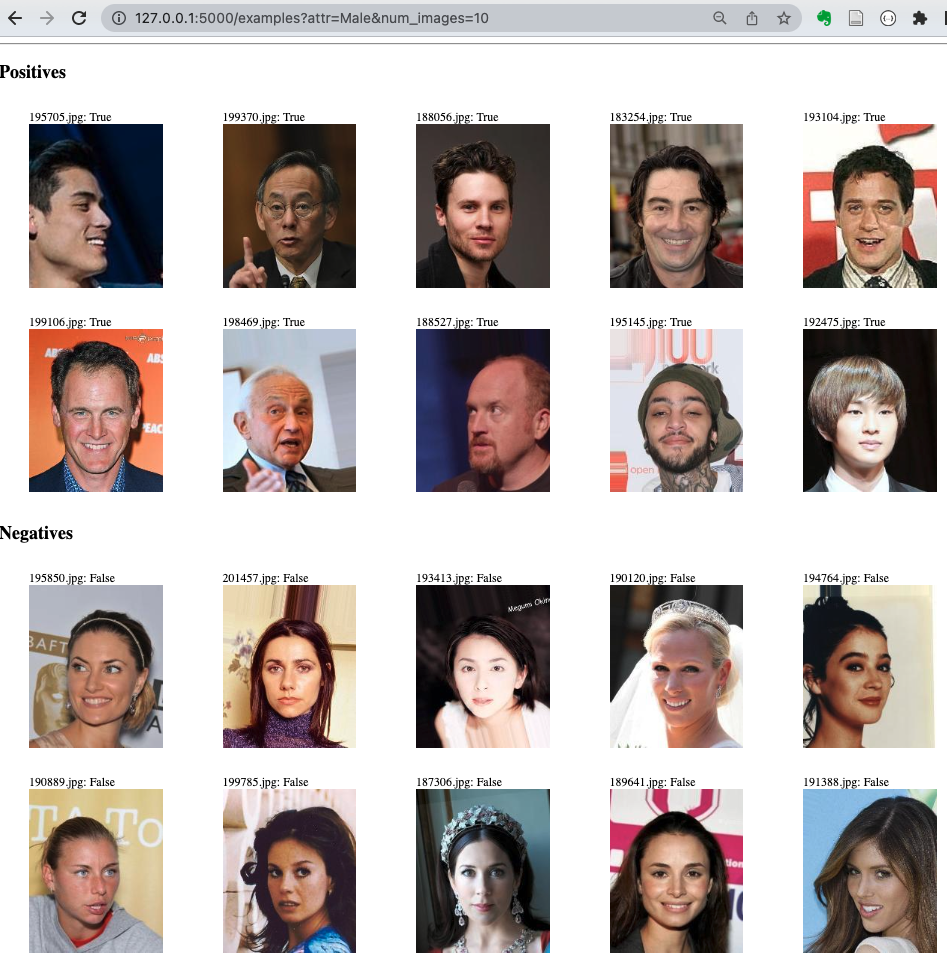} 
  \caption{ Examples of images that are Positive and Negative for Male.} 
    \label{fig:viwer-male}
\end{figure}

\begin{figure}[h]
\centering
  \includegraphics[width=0.7\textwidth]{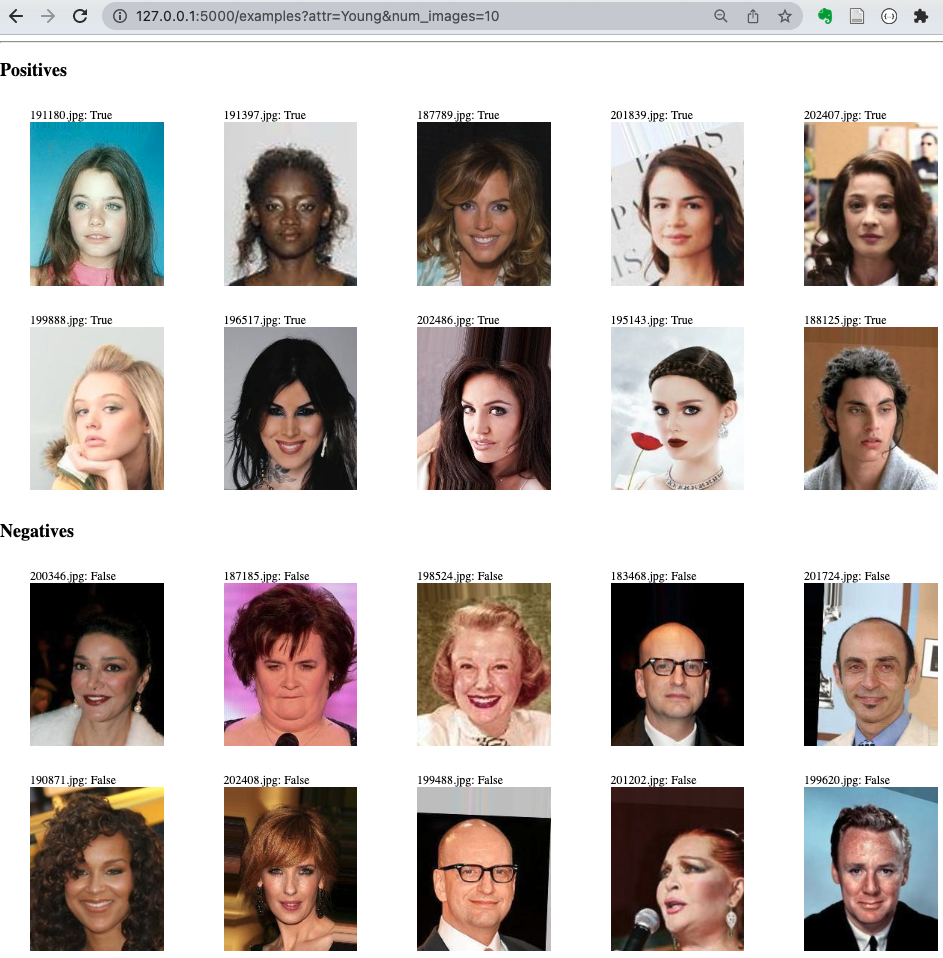} 
  \caption{ Examples of images that are Positive and Negative for Young.} 
    \label{fig:viwer-young}
\end{figure}

\begin{figure}[h]
\centering
  \includegraphics[width=0.7\textwidth]{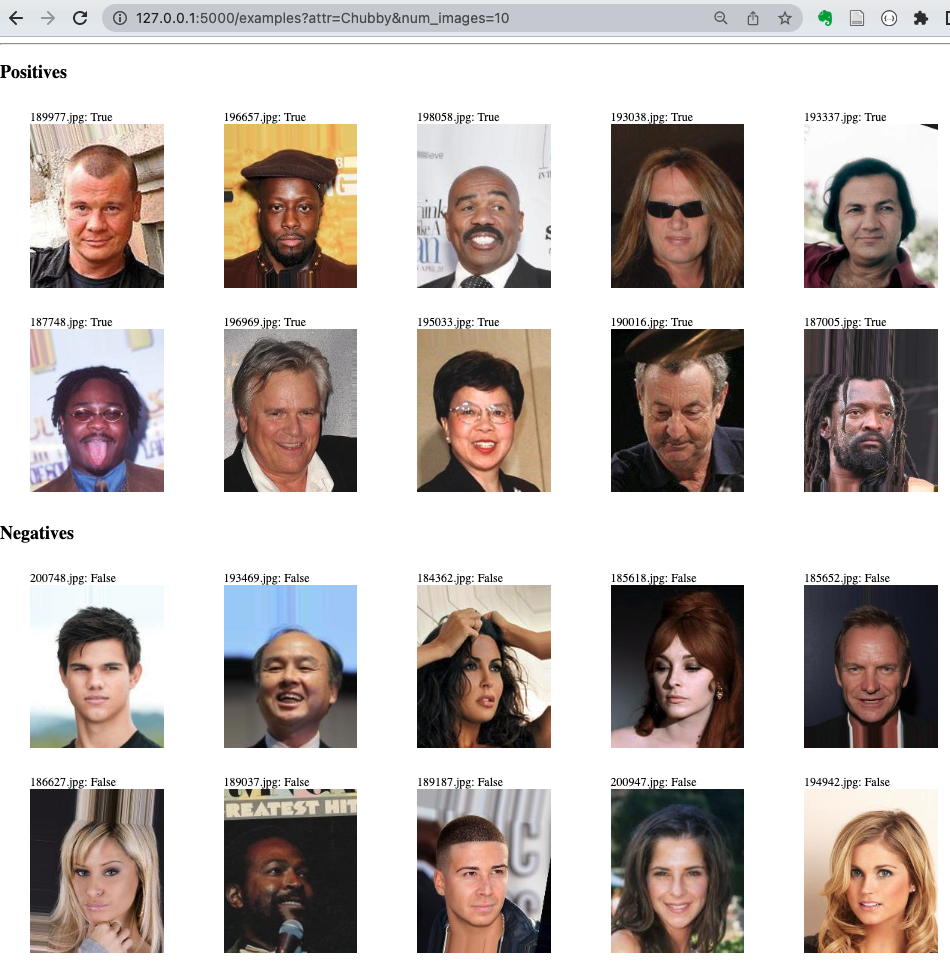} 
  \caption{ Examples of images that are Positive and Negative for Chubby.} 
    \label{fig:viwer-chubby}
\end{figure}

\begin{figure}[h]
\centering
  \includegraphics[width=0.7\textwidth]{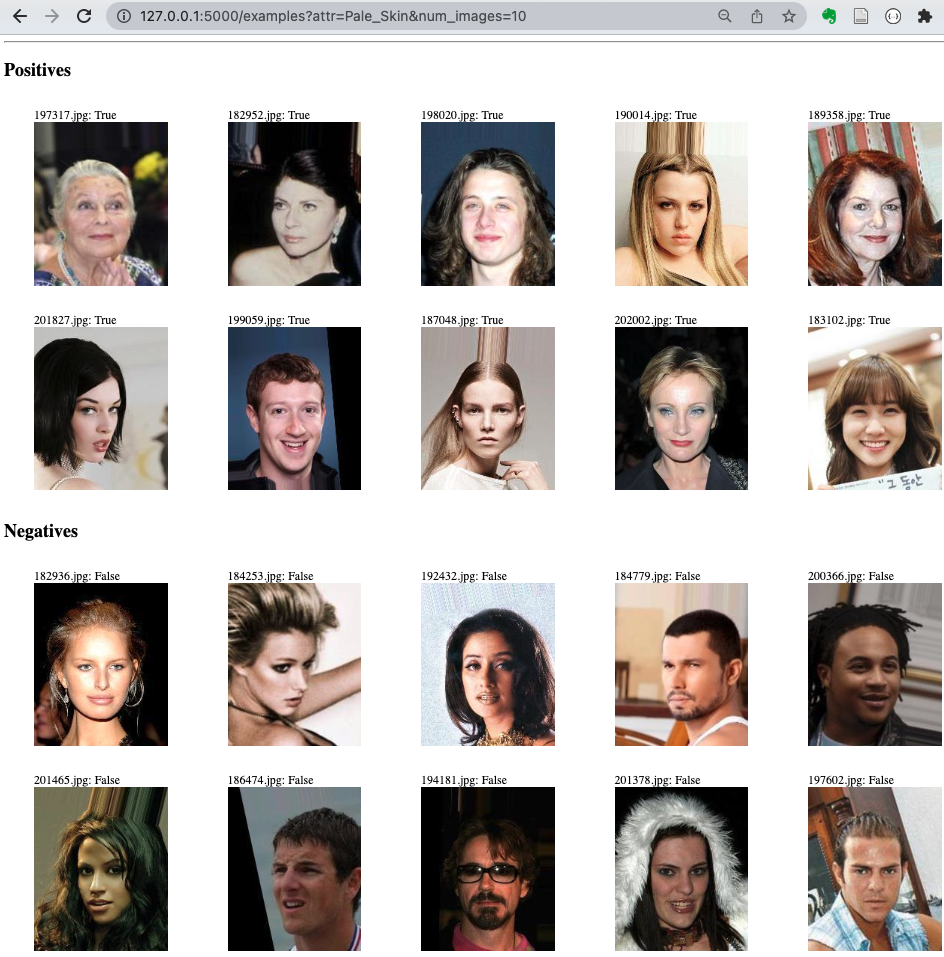} 
  \caption{ Examples of images that are Positive and Negative for Pale Skin.} 
    \label{fig:viwer-pale-skin}
\end{figure}

\begin{figure}[h]
\centering
  \includegraphics[width=0.7\textwidth]{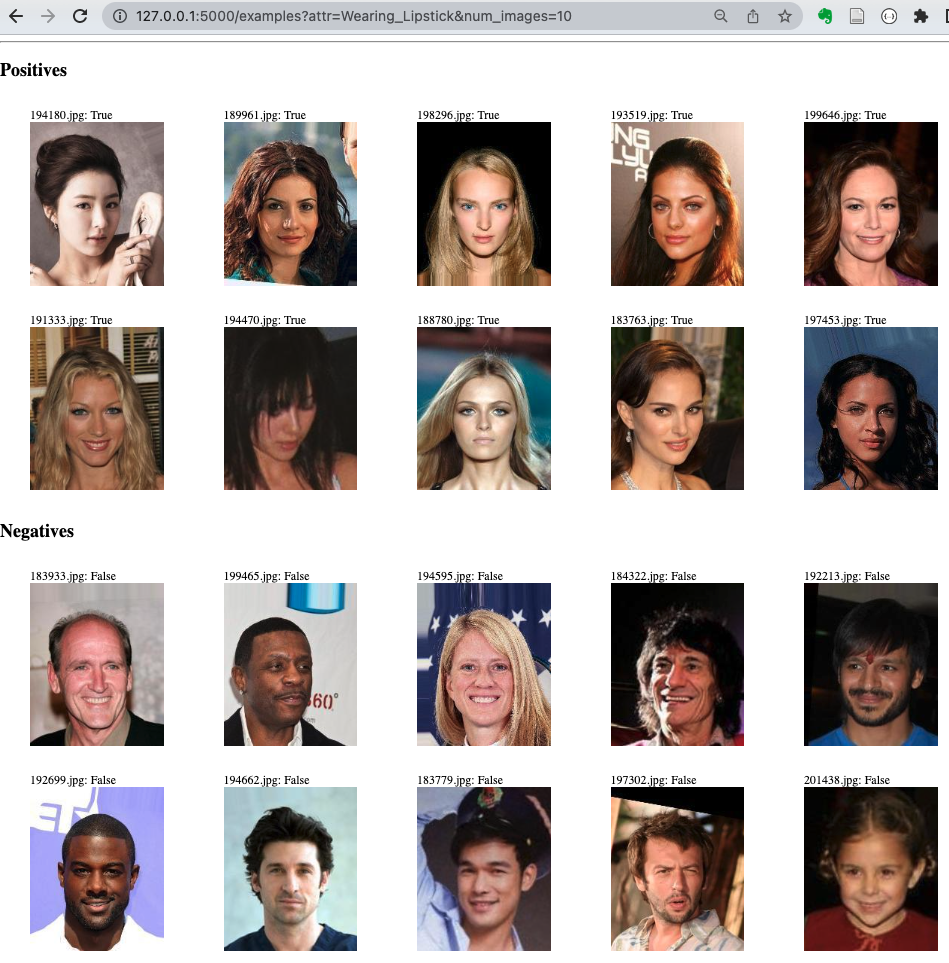} 
  \caption{ Examples of images that are Positive and Negative for Wearing Lipstick.} 
    \label{fig:viwer-lipstick}
\end{figure}

\begin{figure}[h]
\centering
  \includegraphics[width=\textwidth]{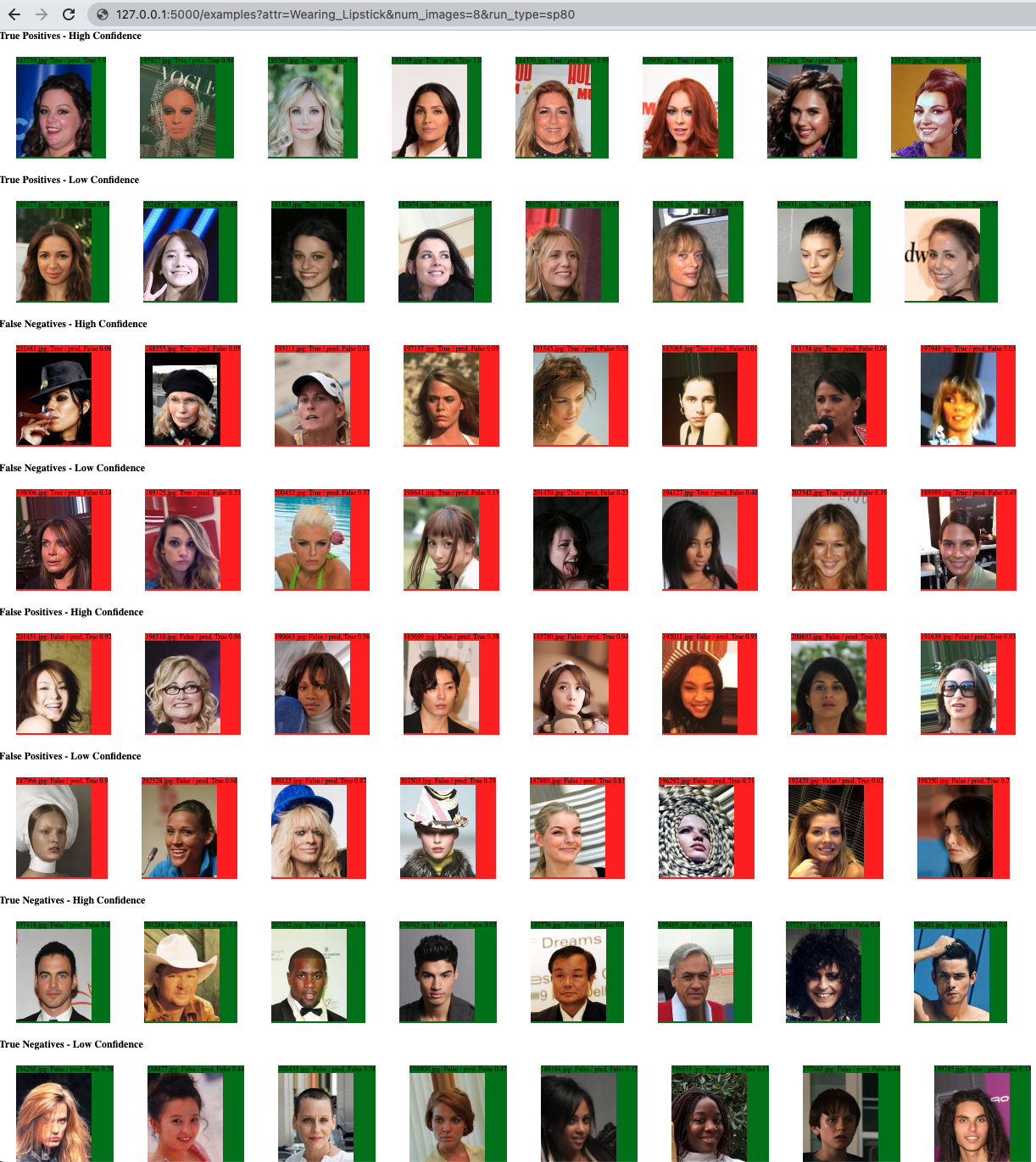} 
  \caption{ Examples of 80\% sparse model performance on images that are Positive and Negative for Wearing Lipstick.} 
    \label{fig:viwer-lipstick-sp80}
\end{figure}

\begin{figure}[h]
\centering
  \includegraphics[width=0.7\textwidth]{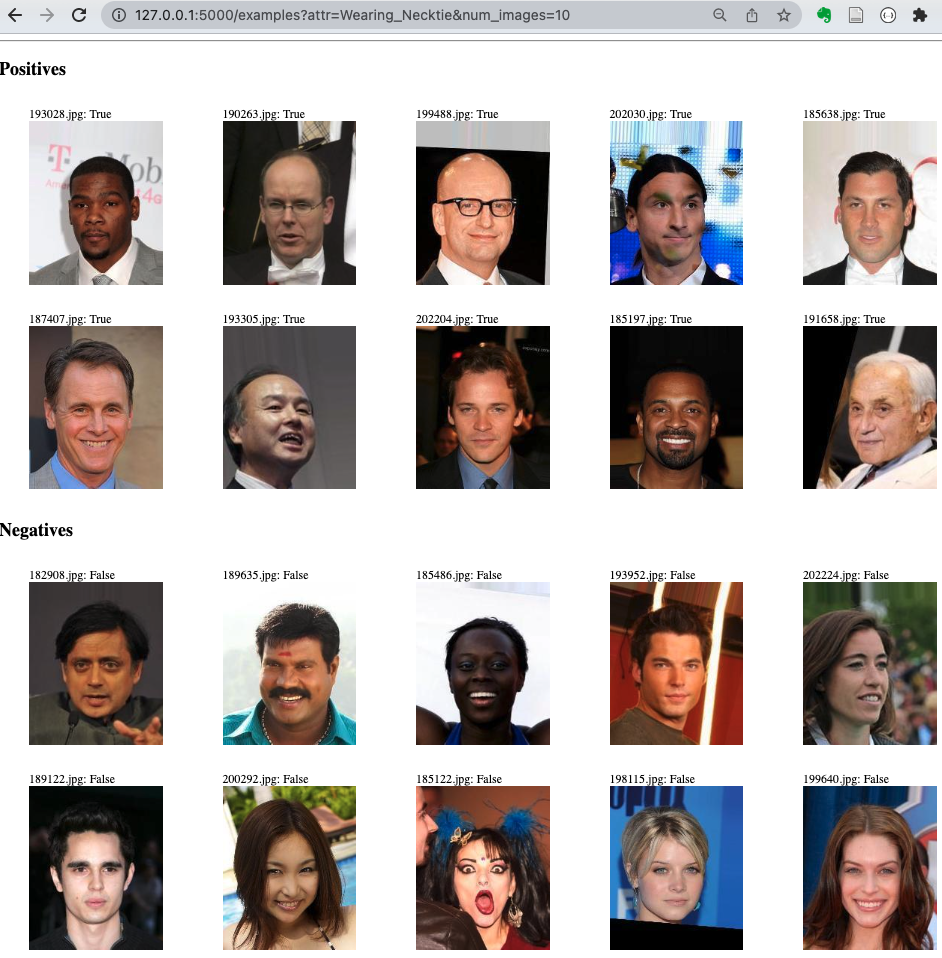} 
  \caption{ Examples of images that are Positive and Negative for Wearing Necktie.} 
    \label{fig:viwer-necktie}
\end{figure}

\begin{figure}[h]
\centering
  \includegraphics[width=\textwidth]{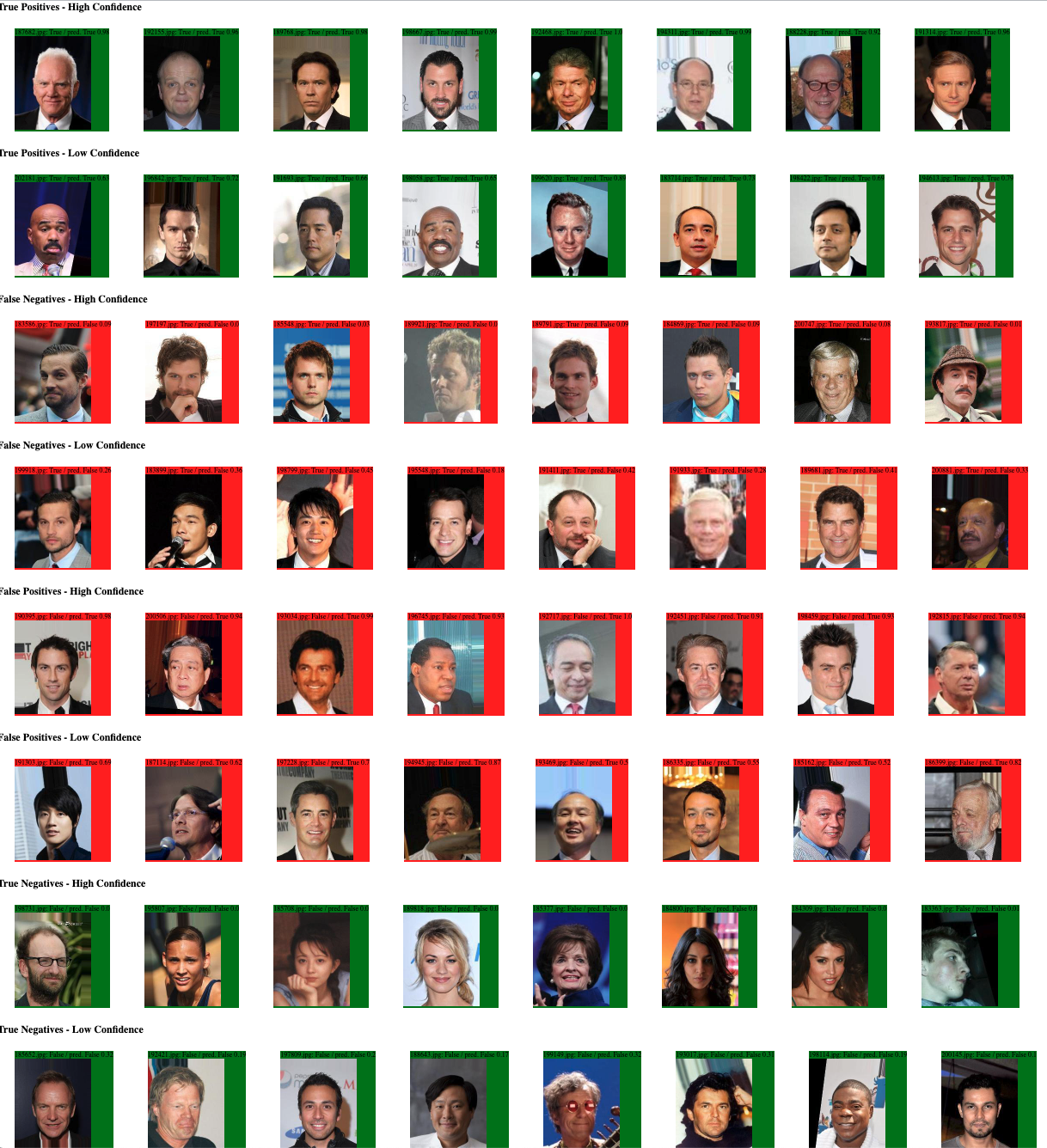} 
  \caption{ Examples of 80\% sparse model performance on images that are Positive and Negative for Wearing Necktie.} 
    \label{fig:viwer-necktie-sp80}
\end{figure}

\begin{figure}[h]
\centering
  \includegraphics[width=0.7\textwidth]{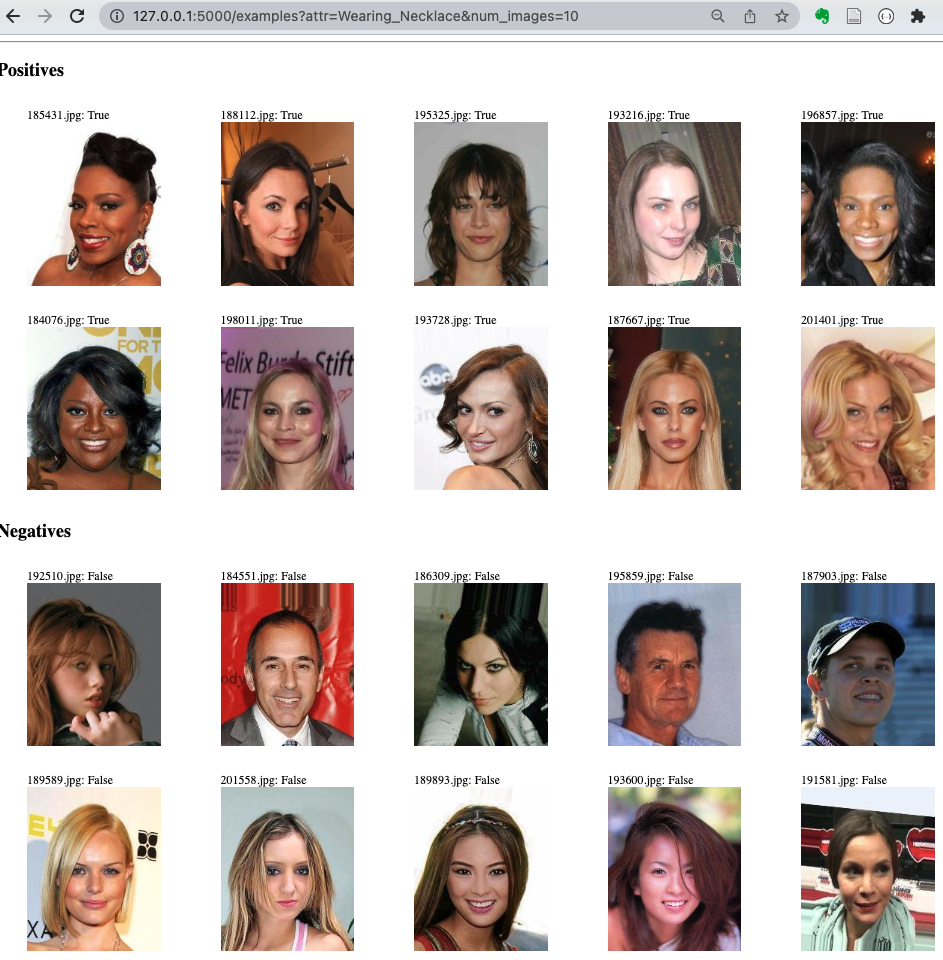} 
  \caption{ Examples of images that are Positive and Negative for Wearing Necklace.} 
    \label{fig:viwer-necklace}
\end{figure}

\begin{figure}[h]
\centering
  \includegraphics[width=\textwidth]{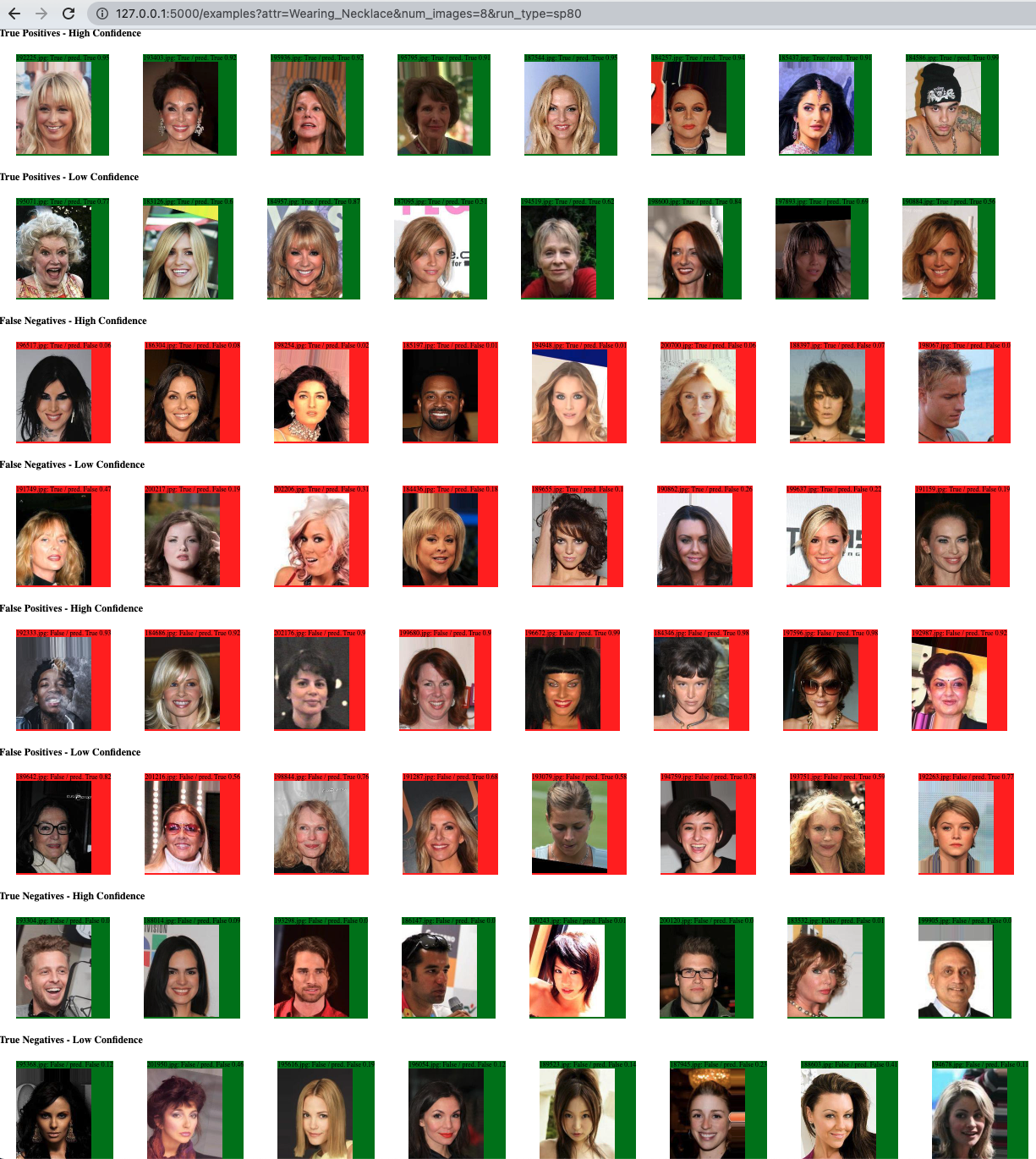} 
  \caption{ Examples of 80\% sparse model performance on images that are Positive and Negative for Wearing Necklace.} 
    \label{fig:viwer-necklace-sp80}
\end{figure}

Despite its simplicity, a quick examination can yield clues to defects in the dataset. As case studies, we first present the viewer showing positive and negative examples for the four CelebA identity categories - Male (Figure~\ref{fig:viwer-male}), Young (Figure~\ref{fig:viwer-young}), Chubby (Figure~\ref{fig:viwer-chubby}), and Pale Skin (Figure~\ref{fig:viwer-pale-skin}). Then, we show three case studies that demonstrate problems in the dataset that can easily be detected from the Example Viewer. Please note that in all illustrations, we avoid cherry-picking by taking the screenshot of the very first returned random set. First, we demonstrate that the categories "Wearing Necklace" (Figures~\ref{fig:viwer-necklace}, \ref{fig:viwer-necklace-sp80}) and "Wearing Necktie" (Figures~\ref{fig:viwer-necktie}, \ref{fig:viwer-necktie-sp80}) often cannot be inferred from the cropped version of the CelebA dataset, due to the fact that images are generally cropped at the neck, between the chin and the clavicle. The cropping frequently removes or largely reduces direct visual evidence of the presence or absence of the attribute, leaving the model to use other, correlated features, even though the human raters had access to the full version of the image. Additionally, we show a view of positive and negative examples of the Wearing Lipstick attribute (Figures~\ref{fig:viwer-lipstick}, \ref{fig:viwer-lipstick-sp80}). These examples readily show that in many cases it is very difficult to determine whether the person in the photograph is wearing lipstick by only examining the mouth. Rather, it appears far more likely that the human raters used other information in the photograph, such as the gender, clothes, and other makeup of the subject as additional information in choosing the correct label. relying heavily on this information can naturally lead to bias in the human labels, thus making any bias (and accuracy) measurement of the predictions unreliable. A closer examination of the viewer output that also shows correct and incorrect high and low-certainty predictions of the GMP-RI 80\% sparse model on these attributes (Figures~\ref{fig:viwer-necklace-sp80}, \ref{fig:viwer-necktie-sp80}, and \ref{fig:viwer-lipstick-sp80}) confirms this observation. Additionally, we note that in the case of Wearing Lipstick and Wearing Necklace, the high-certainty True Negatives appear to skew much more heavily Male than do the low-certainty True Negatives, and the opposite is true for Wearing Necktie. This suggests that the Male attribute and markers of this attribute are used heavily by the model in order to make these predictions.

 \end{document}